\ifdefined\KAUSTOfficialVariant
\documentclass[onecolumn, 12 pt, doublespace, fullpage, a4paper]{report}
\else
\documentclass[onecolumn, 12 pt, doublespace, fullpage, letterpaper]{report}
\fi

\usepackage{amssymb}
\usepackage{textcomp}
\usepackage{graphicx}
\usepackage{graphics}
\usepackage{epsfig}
\usepackage{epstopdf}
\usepackage{float}
\usepackage{placeins}
\usepackage{xcolor}
\usepackage[cmex10]{amsmath}
\usepackage{latexsym,amsfonts}
\usepackage{amsthm}
\usepackage{url}
\usepackage{longtable}
\usepackage[figuresright]{rotating}
\usepackage{listings}
\usepackage{etoolbox}
\usepackage{sectsty}

\chapterfont{\fontsize{14}{15}\selectfont}
\sectionfont{\fontsize{14}{15}\selectfont}
\subsectionfont{\fontsize{14}{15}\selectfont}
\subsubsectionfont{\fontsize{14}{15}\selectfont}

\usepackage{fancyhdr}       
\fancyheadoffset[L]{0.25in}

\makeatletter
\patchcmd{\@makechapterhead}{50\p@}{20pt}{}{}
\patchcmd{\@makeschapterhead}{50\p@}{20pt}{}{}
\makeatother
\fancypagestyle{plain}{
\fancyhf{} %
\fancyhead[C]{\thepage} %

}

\usepackage{setspace}
\usepackage{subfigure}
\usepackage{wrapfig}
\usepackage{overpic}
\usepackage{adjustbox}
\usepackage{threeparttable}
\usepackage{makecell}
\usepackage{colortbl}

   \usepackage{ifthen,xkeyval,xfor,amsgen}
   \usepackage[acronym,toc, nogroupskip]{glossaries}
   \newglossary[slg]{symbols}{syi}{sbl}{List of Symbols}

\usepackage[round,authoryear]{natbib}
\usepackage[pagebackref]{hyperref}
\definecolor{thesislink}{HTML}{243B55} %
\definecolor{thesiscite}{HTML}{5A2A2A} %
\definecolor{thesisurl}{HTML}{2F5D62}  %
\definecolor{thesisfile}{HTML}{4B4453} %
\hypersetup{
    colorlinks=true,
    linkcolor=thesislink,
    citecolor=thesiscite,
    urlcolor=thesisurl,
    filecolor=thesisfile,
    pdftitle={Understanding Practical Optimization in Machine Learning},
    pdfauthor={Egor Shulgin},
    pdfsubject={PhD Dissertation in Computer Science},
    pdfkeywords={optimization, machine learning, distributed optimization, differential privacy, Muon, linear minimization oracles}
}
\renewcommand*{\backrefalt}[4]{%
    \ifcase #1 \footnotesize{(Not cited.)}%
    \or        \footnotesize{(Cited on page~#2)}%
    \else      \footnotesize{(Cited on pages~#2)}%
    \fi}
\usepackage{amsmath,amssymb,amsthm}
\usepackage{color,mathtools}

\usepackage{multirow}
\usepackage{enumitem}
\usepackage{nicefrac}

\usepackage{tablefootnote}

\usepackage{algorithm}
\usepackage{algorithmic}
\usepackage{nicefrac}
\usepackage{xspace}
\usepackage{subfigure}

\usepackage{bm}

\usepackage{amssymb,amsmath,amsthm,dsfont}

\providecommand{\lin}[1]{\ensuremath{\left\langle #1 \right\rangle}}
  \providecommand{\R}{\mathbb{R}} %

  \renewcommand{\epsilon}{\varepsilon}

  \DeclareMathOperator*{\argmin}{arg\,min}

  \providecommand{\bb}{\mathbf{b}}

  \let\ggg\gg
  \renewcommand{\gg}{\mathbf{g}}

  \let\lll\ll
  \renewcommand{\ll}{\mathbf{l}}

  \providecommand{\tt}{\mathbf{t}}

  \providecommand{\mA}{\mathbf{A}}
  \providecommand{\mB}{\mathbf{B}}
  \providecommand{\mC}{\mathbf{C}}
  \providecommand{\mD}{\mathbf{D}}

  \providecommand{\mI}{\mathbf{I}}

  \providecommand{\mL}{\mathbf{L}}

  \providecommand{\mS}{\mathbf{S}}

  \providecommand{\mW}{\mathbf{W}}

\theoremstyle{plain}
\newtheorem{theorem}{Theorem}[section]
\newtheorem{proposition}[theorem]{Proposition}
\newtheorem{lemma}[theorem]{Lemma}
\newtheorem{corollary}[theorem]{Corollary}
\theoremstyle{definition}
\newtheorem{definition}[theorem]{Definition}
\newtheorem{assumption}[theorem]{Assumption}
\theoremstyle{remark}
\newtheorem{remark}[theorem]{Remark}
\newtheorem{example}[theorem]{Example}

\DeclarePairedDelimiterX{\inp}[2]{\langle}{\rangle}{#1, #2}
\DeclarePairedDelimiterX{\abs}[1]{\lvert}{\rvert}{#1}
\DeclarePairedDelimiterX{\norm}[1]{\lVert}{\rVert}{#1}
\DeclarePairedDelimiterX{\cbr}[1]{\{}{\}}{#1} %
\DeclarePairedDelimiterX{\rbr}[1]{(}{)}{#1} %
\DeclarePairedDelimiterX{\sbr}[1]{[}{]}{#1} %

\definecolor{mydarkblue}{rgb}{0,0.08,0.45}

\makeatletter
\providecommand{\thesis@cite@plain}[1]{\cite{#1}}
\providecommand{\thesis@cite@opt}[1][]{\@ifnextchar[{\thesis@cite@opt@two}{\thesis@cite@opt@one}}
\providecommand{\thesis@cite@opt@one}[1]{\cite{#1}}
\providecommand{\thesis@cite@opt@two}[2][]{\cite{#2}}
\@ifundefined{citep}{\DeclareRobustCommand{\citep}{\@ifnextchar[{\thesis@cite@opt}{\thesis@cite@plain}}}{}
\@ifundefined{citet}{\DeclareRobustCommand{\citet}{\citep}}{}
\@ifundefined{citealp}{}{}
\@ifundefined{citealt}{}{}
\@ifundefined{citeauthor}{\DeclareRobustCommand{\citeauthor}{\citep}}{}
\@ifundefined{citeyear}{\DeclareRobustCommand{\citeyear}{\citep}}{}
\@ifundefined{citeyearpar}{\DeclareRobustCommand{\citeyearpar}{\citep}}{}
\makeatother
\providecommand{\Cref}[1]{\ref{#1}}
\providecommand{\cref}[1]{\ref{#1}}

\usepackage{xcolor}
\usepackage{color}
\providecolor{midnightblue}{HTML}{0059b3}
\usepackage{graphicx}
\usepackage{verbatim}
\usepackage{multirow}

\newcommand{\cB}{{\cal B}}
\newcommand{\cC}{{\cal C}}
\newcommand{\cD}{{\cal D}}

\newcommand{\cF}{{\cal F}}

\newcommand{\cO}{{\cal O}}

\newcommand{\cQ}{{\cal Q}}

\newcommand{\cS}{{\cal S}}

\newcommand{\cX}{{\cal X}}

\newcommand{\eqdef}{\overset{\text{def}}{=}} 
\newcommand{\ve}[2]{\langle #1 ,  #2 \rangle} %
\DeclareMathOperator{\Prob}{Prob}

\newcommand{\Diag}[1]{\mathbf{Diag}\left( #1\right)}

\newcommand{\Exp}[1]{{\mathbb{E}}\left[#1\right] }    %
\newcommand{\E}[1]{{\mathbb{E}}\left[#1\right] } 
\newcommand{\EE}[2]{{\mathbb{E}}_{#1}\left[#2\right] }

\usepackage{url}

\usepackage{xspace}

\newcommand{\sgd}{{\tt SGD}\xspace}

\newcommand{\bU}{\mathbf{U}}

\usepackage{booktabs}
\usepackage{tabularx}

\def\<{\left\langle}
\def\>{\right\rangle}
\def\({\left(}
\def\){\right)}

\usepackage{pifont}
\newcommand{\cmark}{{\color{green}\ding{51}}}%
\newcommand{\xmark}{{\color{red}\ding{55}}}%

\usepackage{bbm}

\NewDocumentCommand{\rot}{O{0} O{2em} m}{\makebox[#2][l]{\rotatebox{#1}{#3}}}%

\ifdefined\KAUSTOfficialVariant
\usepackage[lmargin=40mm, rmargin=25mm, vmargin=25mm, headsep=2.5mm]{geometry}
\else
\usepackage[lmargin=1.5in, rmargin=1in, vmargin=1in, headsep=0.083334in]{geometry}
\fi
\newcommand{\mathsym}[1]{{}}
\newcommand{\unicode}[1]{{}}

\renewcommand\bibname{\centering BIBLIOGRAPHY}

\begin{document}

\ifdefined\KAUSTOfficialVariant
  \thispagestyle{empty}
  \addvspace{5mm}
  \begin{center}
    \begin{doublespace}
      {\textbf{{\large Understanding Practical Optimization in Machine Learning}}}
    \end{doublespace}
    \vspace{10mm}
    {Dissertation by}\\
    {Egor Shulgin}
    \vspace{30mm}

    {In Partial Fulfillment of the Requirements}\\[12pt]
    {For the Degree of}\\[12pt]
    {Doctor of Philosophy}
    \vfill
    {King Abdullah University of Science and Technology}\\
    {Thuwal, Kingdom of Saudi Arabia}
    \vfill
    \begin{onehalfspace}
      {\copyright{} July 2026}\\
      {Egor Shulgin}\\
      {All rights reserved}\\
      {\small\href{https://shulgin-egor.github.io/}{\textcolor{black}{https://shulgin-egor.github.io/}}}
    \end{onehalfspace}
  \end{center}
\else
  \vspace{2pt}
  \thispagestyle{empty}
  \addvspace{10mm}

  \begin{center}

  {\textbf{{\large Understanding Practical Optimization in Machine Learning}}}\vfill
  {Dissertation by}\\
  {Egor Shulgin}\vfill

  { In Partial Fulfillment of the Requirements}\\[12pt]
  { For the Degree of}\\[12pt]
  {Doctor of Philosophy} \vfill
  {King Abdullah University of Science and Technology }\\
  {Thuwal, Kingdom of Saudi Arabia}
  \vfill
  {July 2026}

  \end{center}
\fi

\newpage

\begin{center}

\end{center}

\begin{center}

{ \textbf{{\large EXAMINATION COMMITTEE PAGE}}}\\\vspace{1cm}

\end{center}
\noindent{The dissertation of Egor Shulgin is approved by the examination committee}
\phantomsection
\addcontentsline{toc}{chapter}{Examination Committee Page}

\vspace{4\baselineskip}

\begin{onehalfspacing}
\noindent{Committee Chairperson: Peter Richt\'{a}rik}\\
Committee Members: Marco Canini, Ra\'{u}l Fidel Tempone, H. Brendan McMahan, Shiqiang Wang, Tong Zhang\vfill
\end{onehalfspacing}

\ifdefined\KAUSTOfficialVariant
\else
  \newpage
  \phantomsection
  \addcontentsline{toc}{chapter}{Copyright}
  \vspace*{\fill}
  \begin{center}
  { \copyright{} July 2026}\\
  {Egor Shulgin}\\
  {Licensed under \href{https://creativecommons.org/licenses/by/4.0/}{\textcolor{black}{CC BY 4.0}}}\\[6pt]
  {\href{https://shulgin-egor.github.io/}{\textcolor{black}{https://shulgin-egor.github.io/}}}
  \end{center}
\fi

\begin{center}

\end{center}

\begin{center}
{{\bf\fontsize{14pt}{14.5pt}\selectfont \uppercase{ABSTRACT}}}
\end{center}

\doublespacing
\phantomsection
\addcontentsline{toc}{chapter}{Abstract}

\begin{center}
{{\fontsize{14pt}{14.5pt}\selectfont {Understanding Practical Optimization in Machine Learning\\
Egor Shulgin}}}
\end{center}

\singlespacing

Modern machine learning relies on optimization methods whose empirical success often precedes their theoretical understanding. They frequently differ from the idealized algorithms studied in classical analyses: they train submodels rather than the full model, clip or normalize gradients, update each layer in its own non-Euclidean geometry, and replace expensive inner computations with cheap approximations. This thesis analyzes the procedure that is actually implemented rather than a convenient proxy, and shows that such modifications are rarely harmless: they can change the objective, the fixed point, the appropriate assumptions, and the best hyperparameters.

The first part studies submodel training. We show that Independent Subnetwork Training, which computes gradients after sketching the model, can converge to a shifted fixed point and incur an irreducible error on the original objective, because sketching and differentiation do not commute. Model-Agnostic Sparsified Training instead places the sketch distribution inside the objective. For this objective, sketched gradients are unbiased, smoothness and convexity are inherited from the original problem, and we prove convergence guarantees in convex, nonconvex, distributed, and variance-reduced settings.

The second part studies differentially private optimization. We show that adaptive quantile clipping, which avoids tuning a fixed clipping threshold, still retains clipping bias at any fixed quantile. A time-varying quantile and stepsize schedule removes this bias and attains nonconvex rates that are optimal up to logarithmic factors under heavy-tailed noise. We then combine smoothed normalization with error compensation and obtain convergence guarantees for differentially private distributed optimization under standard smoothness assumptions, without assuming bounded gradients.

The third part studies Muon- and Scion-style optimizers based on linear minimization oracles. We introduce a layer-wise generalized smoothness model, validate it along NanoGPT training trajectories, and build on it the Gluon framework, whose guarantees depend on per-layer rather than global smoothness constants. We also analyze the approximate orthogonalization used in practical Muon: less precise orthogonalization calls for smaller stepsizes, so its precision should be tuned together with the learning rate, as NanoGPT experiments confirm.

The thesis thus shows that practical optimization methods are better understood, and can often be improved, when theory retains the features that determine their practical behavior.

\begin{center}

{\bf\fontsize{14pt}{14.5pt}\selectfont \uppercase{Acknowledgements}}\\\vspace{1cm}
\end{center}

\phantomsection
\addcontentsline{toc}{chapter}{Acknowledgements}

First and foremost, I am deeply grateful to my advisor, Peter Richtárik. I started working with Peter in 2018, while I was an undergraduate at the Moscow Institute of Physics and Technology, and later joined his group at KAUST for my master’s degree and PhD. Especially in the early years, his guidance, support, and our technical discussions helped me develop the skills and confidence to work more independently. As my interests evolved, Peter gave me the freedom to explore different topics, change direction when needed, and eventually focus on the questions that I found most exciting. I am also grateful for the research group he built and for the atmosphere around it, which shaped a large part of my experience at KAUST.

I am very grateful to Alexander Gasnikov. During my undergraduate studies at MIPT, he was one of the people who made optimization look like an exciting and active area of research. He helped connect me with Peter and continued to support me during my early years in research. I also owe a lot to the Department of Control and Applied Mathematics and the Department of Intelligent Systems at MIPT, where my interest in research first developed.

I sincerely thank the members of my dissertation committee: Marco Canini, Raúl Fidel Tempone, H. Brendan McMahan, Shiqiang Wang, and Tong Zhang. I am grateful for their time, careful consideration of my work, and thoughtful questions and discussions. I had followed their research and learned from it for many years, and I was honored that they agreed to serve on my committee.

I have been fortunate to collaborate with many excellent researchers over these years. In addition to Peter and Alexander, I would like to thank Robert Mansel Gower, Nicolas Loizou, Xun Qian, Alibek Sailanbayev, Konstantin Mishchenko, Dmitry Kovalev, Yura Malitsky, Anastasiya Ivanova, Dmitry Pasechnyuk, Dmitry Grishchenko, Vladislav Matyukhin, Mher Safaryan, Alexander Rogozin, Vladislav Lukoshkin, Motasem Alfarra, Juan C. Pérez, Bernard Ghanem, Yury Demidovich, Grigory Malinovsky, Sarit Khirirat, Avetik Karagulyan, Abdurakhmon Sadiev, Artem Riabinin, Kaja Gruntkowska, Sultan AlRashed, Francesco Orabona, Dimitri von Rütte, Tianyue H. Zhang, Niccolò Ajroldi, Bernhard Schölkopf, Antonio Orvieto, Tamaz Gadaev, Mohamed Awad, Eduard Gorbunov, Sam Laing, Aliaksandr Haponau, and Zhirayr Tovmasyan. I learned a great deal from these collaborations, and working with people who approached problems differently was one of the parts of research I enjoyed most.

Konstantin Mishchenko was one of my earliest inspirations in optimization and one of the reasons I decided to continue toward a PhD in the field. His work and our discussions taught me a lot, and I especially enjoyed the “Nintendo” seminars he organized. Grigory Malinovsky has been a collaborator and friend since our undergraduate years at MIPT, when we first began working in optimization. I am grateful for our joint projects and for our many debates about research, politics, philosophy, and much else---sometimes heated, always respectful. I owe special thanks to Sarit Khirirat for our long research collaboration. I always looked forward to our regular meetings, where conversations could easily drift from technical questions to history and politics. I am grateful to Avetik Karagulyan for a quick tour of Langevin Monte Carlo and for the many research discussions and projects that were always fun to work on. I thank Konstantin Burlachenko for the seemingly endless discussions in the lab about research, technology, and life. I am also grateful to Samuel Horváth for the friendship that continued long after his time at KAUST, and for all our meetings at conferences and elsewhere over the years. I thank Abdurakhmon Sadiev for being a collaborator, flatmate, and friend, with special thanks for the amazing plov and lagman.

I am thankful to everyone who was part of the Optimization and Machine Learning group during my years at KAUST. I learned a lot from the seminars, reading groups, and countless discussions in the office, and the trips and informal gatherings made the group a great place to spend these years. I was fortunate to be in the group with Laurent Condat, Zhize Li, El Houcine Bergou, Alexander Tyurin, Filip Hanzely, Adil Salim, Aritra Dutta, Kai Yi, Lukang Sun, Hanmin Li, Igor Sokolov, Ivan Ilin, Omar Shaikh Omar, Alyazeed Basyoni, Igor Klimczak, Majied Ammar Mahran, Yassine Maziane, and Alina Abdikarimova. I also enjoyed getting to know Hussein Rammal, Adrien Fradin, Philip Zmushko, Aadi Rane, Wojciech Anyszka, Dymitr Lubczyk, Simone Maria Giancola, Robin Yadav, Kirill Acharya, Georg Meinhardt, Ákos Záhorský, Marta Pozzi, Andrei Panferov, Ahmad Rammal, Babis Kostopoulos, Dinis Seward, Michał Grudzień, Rafał Szlendak, Muhammad Harun Ali Khan, Ilyas Fatkhullin, Rustem Islamov, and Ahmed Khaled. The group lunches and long tea breaks, often accompanied by far too many sweets, made everyday life in the lab a lot more fun.

I am also grateful to the people I worked with during my internships and research visits. At Samsung AI Center in Cambridge, I am especially grateful to Stefanos Laskaridis for hosting me and for our work together, and to Nicholas D. Lane for welcoming me into his group. I also thank Łukasz Dudziak, Javier Fernández-Marqués, Victor Escorcia, and Tae Kwon Chang (TK) for the many discussions and for making my time in Cambridge fun both in and out of the office.

At Apple in Cambridge, I am grateful to Áine Cahill for her mentorship and to Filip Granqvist, Mona Chitnis, and Vojta Jína for their support and for welcoming me into the Private Machine Learning team. I also thank Eeshan Gunesh Dhekane and Max Müller-Eberstein for the conversations and the many tea breaks during the internship.

I am grateful to Eduard Gorbunov for hosting my research visit at the Mohamed bin Zayed University of Artificial Intelligence in Abu Dhabi and for our collaboration. I also thank Antonio Orvieto for our many discussions and joint work during the final part of my PhD and for hosting me in Tübingen. I am grateful to Sam Laing and Tianyue (Helen) Zhang for their hospitality and for making my stay there so pleasant.

I am deeply grateful to KAUST for making it possible to focus so much of my time and energy on research. The campus, offices, excellent sports facilities, and the fact that so many everyday practical matters were taken care of made KAUST an unusually comfortable place to live and work. I also thank the Visual Computing Center and the KAUST AI Initiative for their support, as well as the administrative and technical staff who helped me throughout these years.

My first years at KAUST coincided with the COVID-19 pandemic. When travel was restricted and borders were often closed, the friendships on campus became particularly important. I am grateful to Grisha, Anastasia, and Zohra for the time we spent together during those years and for our many discussions about politics, religion, and more.

Many other friends made my years at KAUST memorable. I am especially grateful to Jonathan Klein for the many ambitious hikes and backpacking trips over the years. I thank Kymbat, Eliza, and Avetik for our wonderful trips around Saudi Arabia, and Arto for many other trips and adventures. I owe special thanks to Begimai for somehow making great cooking look effortless and for the postcard tradition that I happily picked up from her. I am grateful to Nadya Briantceva for her constant positivity and laughter, to Slavo for the fun board-game evenings, and to Elnur for the lively Russian parties.

The Ultimate Frisbee community also became an important part of my life at KAUST. I am especially grateful to Nursulu for all the training sessions and games, and to everyone I played with at KAUST and in tournaments.

Saudi Arabia gradually became home to me during my years at KAUST. I am grateful for the hospitality I experienced and for the opportunity to explore so much of the country beyond the campus: from the Red Sea coast to mountains, deserts, canyons, and many places I would probably never have discovered otherwise. Some of my favorite memories from these years come from those trips and the people I shared them with.

During the later years of my PhD, I started returning to Kazakhstan regularly, and Almaty became home to me again in a new way. I was born there, but during these years I got to know the city, and especially the surrounding mountains, much better. Alpine skiing, hiking, and backpacking became a welcome balance to research. The mountains around Almaty will always have a special place in my life.

Finally, I would like to thank my partner, my friends, and my family in Russia and Kazakhstan for their love, patience, encouragement, and support throughout these years. It would have been much harder, and perhaps even impossible, without them.

\begin{onehalfspacing}%

\renewcommand{\contentsname}{\centerline{\textbf{{\large TABLE OF CONTENTS}}}}
\begingroup
\hypersetup{linkcolor=black}
\tableofcontents
\endgroup
\hypersetup{
    linkcolor=thesislink,
    citecolor=thesiscite,
    urlcolor=thesisurl,
    filecolor=thesisfile
}
\cleardoublepage

\cleardoublepage
\phantomsection
\addcontentsline{toc}{chapter}{\listfigurename} 
\renewcommand*\listfigurename{\centerline{LIST OF FIGURES}} 
\listoffigures

\cleardoublepage
\phantomsection
\addcontentsline{toc}{chapter}{\listtablename}
\renewcommand*\listtablename{\centerline{LIST OF TABLES}} 
\listoftables

\end{onehalfspacing}
\singlespacing

\definecolor{midnightblue}{HTML}{0059b3}
\newcommand{\algname}[1]{{\color{midnightblue!40!black}\small\sf#1}\xspace}

\chapter{Introduction}
\label{chapter:introduction}

Over the last decade, machine learning has moved from a specialized technical field to one of the main drivers of progress in computing, science, engineering, and beyond. Modern machine-learning systems are now used not only for classification, translation, and recommendation, but also for coding, search, scientific exploration, and mathematical work. Several milestones illustrate the breadth of this progress, from game playing and scientific prediction to recent examples of AI-assisted mathematical discovery \citep{silver2016mastering,jumper2021highly,openai2026erdos}. These systems still have important limitations, but their usefulness is no longer easy to dismiss.

Behind this progress is a simple fact: useful machine-learning models have to be trained. Training is an optimization process, and often one of the most expensive parts of developing a model. In large-scale language modeling, model quality has improved dramatically with scale, but making this progress practical requires algorithms and training rules that use compute efficiently \citep{hoffmann2022training}. Scaling is also not just a matter of spending more compute: without appropriate optimization and scaling rules, increasing model size or training budget can even degrade performance in some modern architectures \citep{vankadara2024feature}. At the same time, recent estimates suggest that the cost of training frontier AI models has increased rapidly over the last several years \citep{cottier2024rising}. Improvements in optimization can therefore translate directly into cheaper training, faster experimentation, easier tuning, and a lower barrier to applying machine learning in new domains.

This is one reason optimization remains central to machine learning. Another reason is that the methods most important in practice are often not the methods that are easiest to analyze. Stochastic gradient methods have long been the standard theoretical workhorse for large-scale machine learning \citep{bottou2018optimization}. At the same time, much of deep learning practice has relied on adaptive methods, most prominently \algname{Adam} \citep{kingma2015adam}. \algname{Adam} became one of the default optimizers for neural-network training and received the ICLR 2025 Test-of-Time Award \citep{iclr2025adam_tot}. Yet its theoretical understanding was initially much less clean than its empirical success: the original convergence analysis had gaps, and later work was needed to clarify convergence and failure modes of \algname{Adam}-type methods \citep{reddi2019convergence}.

Practical training methods are often introduced to solve an immediate problem: reduce tuning, fit a model across several devices, control privacy sensitivity, or replace an expensive computation with a cheaper approximation. The corresponding theory usually comes later. Sometimes it explains why the method works; sometimes it exposes a limitation; and sometimes it points to a modified algorithm. In each case, optimization provides a way to reason about training rather than treat the optimizer as a black-box engineering choice.

This thesis studies several examples in which the standard model omits a feature that matters. Submodel training can change the effective objective or fixed point. Clipping and normalization alter the update before privacy noise is added. \algname{Muon} acts layer by layer and relies on an approximate inner computation. These discrepancies also matter for the broader goal of understanding deep learning, which still lacks a comprehensive theory of why training works, when it fails, and how its behavior depends on algorithmic and modeling choices. Optimization offers concrete tools for studying these questions.

More faithful theory can help explain why a method is robust or brittle, why it needs delicate tuning, why it converges to the wrong object, or why an implementation detail changes performance. Rather than building one universal theory of practical optimization, the thesis studies several concrete cases and asks the same question each time: what do we learn when we model the objective, assumptions, update, and computation actually used in practice, rather than a more convenient proxy?

\section{A common viewpoint: from objective to implementation}
\label{sec:common_viewpoint}

Two mathematical objects recur throughout this thesis. The first is the optimization problem itself: what objective is induced by the practical training procedure, and which assumptions appropriately describe it? The second is the implemented method: how is the update formed, and is any inner computation performed exactly? Keeping these objects separate is useful because a practical procedure can depart from its idealized model at either level.

A common mathematical starting point for machine learning is the optimization problem
\begin{equation*}
    \min_{x\in\mathbb{R}^d} f(x),
\end{equation*}
where $x$ denotes the parameters of a model and $f$ is the objective being optimized. This compact notation hides many modeling choices: the objective may come from data, from a surrogate loss, from a distributed system, or from a practical modification of the training procedure. In many learning problems, the objective is defined through data or randomness, for example
\begin{equation*}
    f(x)=\mathbb{E}_{\xi\sim\mathcal D}\big[f_\xi(x)\big],
\end{equation*}
where $\xi$ represents a random data point or another source of stochasticity. After replacing the data distribution by a finite dataset, one often obtains the finite-sum objective
\begin{equation*}
    f(x)=\frac1n\sum_{i=1}^n f_i(x).
\end{equation*}
This formulation underlies much of large-scale optimization for machine learning \citep{bottou2018optimization}. It is also flexible enough to describe centralized, stochastic, distributed, and federated learning problems, depending on how the component functions $f_i$ or the random variable $\xi$ are interpreted.

In modern machine learning, however, the objective is usually nonconvex. Thus, writing $\min_x f(x)$ should not be understood as claiming that the algorithms studied in this thesis find global minimizers. The usual goal is more modest: to find an approximate stationary point. In the simplest differentiable setting, this means producing an iterate $\hat x$ such that
\begin{equation*}
    \mathbb{E}\|\nabla f(\hat x)\|^2 \le \varepsilon^2,
\end{equation*}
or proving a bound of this form for an output iterate sampled from the optimization trajectory. Intuitively, such a guarantee says that the algorithm reaches a point where the first-order signal is small on average. In later chapters, the exact stationarity measure may change with the structure of the method, but the same basic idea remains: the analysis tries to show that the relevant first-order quantity becomes small.

The classical update associated with the objective $f$ is gradient descent,
\begin{equation*}
    x^{k+1}=x^k-\gamma_k\nabla f(x^k).
\end{equation*}
In large-scale settings, the full gradient is often replaced by a stochastic estimator,
\begin{equation*}
    x^{k+1}=x^k-\gamma_k g^k,
    \qquad
    \mathbb{E}\!\left[g^k\mid x^k\right]=\nabla f(x^k).
\end{equation*}
The stochastic-gradient viewpoint has its roots in stochastic approximation \citep{robbins1951stochastic}, and remains one of the central abstractions in optimization for machine learning \citep{bottou2018optimization}. In this standard model, the update direction is either the gradient of the stated objective or a random approximation to it.

The assumptions used to analyze such updates are also part of the mathematical model. A typical baseline assumption is $L$-Lipschitz continuity of the gradient:
\[
    \|\nabla f(y)-\nabla f(x)\|
    \le
    L\|y-x\|.
\]
By the descent lemma, this implies
\[
    f(y)
    \le
    f(x)
    +
    \langle \nabla f(x),y-x\rangle
    +
    \frac{L}{2}\|y-x\|^2.
\]
The latter inequality upper-bounds $f(y)$ by its first-order Taylor approximation at $x$ plus a quadratic remainder. Together, Lipschitz continuity of the gradient and a lower bound $f(x)\ge f^{\inf}$ form a standard assumption pair in nonconvex descent analysis: smoothness controls the local change of the objective, while lower boundedness prevents infinite descent.

A stronger simplifying assumption is a uniform gradient bound, $\|\nabla f(x)\|\le G$ for all $x$. This assumption can be useful for obtaining clean bounds, but it is quite restrictive: it fails even for unconstrained quadratic objectives on $\mathbb{R}^d$. More importantly for this thesis, it can hide the effect of the practical modification being analyzed. If gradients are assumed bounded and the clipping threshold is chosen large enough, clipping becomes essentially inactive. Similarly, bounding the gap between a gradient computed on a sketched model and the original gradient by a uniform constant can hide whether the update is aligned with the nominal objective at all. Several chapters therefore avoid relying on such assumptions when the goal is precisely to understand the effect of the modified update.

The assumptions are necessarily chapter-dependent. Some chapters use standard smoothness, while others study objectives or updates for which the usual conditions are too restrictive or too coarse. The thesis therefore does not impose one global model on all six technical chapters.

The methods in this thesis often start from this gradient-based picture but replace the direction. Write
\begin{equation*}
    x^{k+1}=x^k-\gamma_k u^k,
\end{equation*}
where $u^k$ denotes the direction actually produced by the algorithm. For \algname{SGD}, $u^k$ is an unbiased stochastic gradient. In this thesis, it may instead be computed on a submodel, clipped or normalized for privacy, corrected through memory, or obtained approximately from an inner routine. This notation separates the implemented direction from the gradient that appears in the nominal optimization problem. If

\begin{equation*}
    \mathbb{E}\!\left[u^k\mid x^k\right]\neq \nabla f(x^k),
\end{equation*}
then the method is not \algname{SGD} on the nominal objective. It may converge to a shifted point, leave an irreducible error term, or require a different interpretation. For random sketches or submodels, changing the objective itself may give the more useful description.

The same four questions recur throughout the thesis: What objective is induced by the practical procedure? Which assumptions capture its structure? How is the update formed? Is the update or inner computation exact? A useful model need not reproduce every implementation detail, but it should retain the features that determine the optimization behavior and still lead to meaningful convergence guarantees.

Although the chapters differ in notation and assumptions, the modeling step is the same: identify the feature of the procedure that matters and analyze its consequences. The next section develops this view for the three directions of the thesis.

\section{Main directions of the thesis}

The six technical chapters are organized around three departures from standard optimization models. In submodel and sketch-aware training (Chapters~\ref{chapter2:ist} and~\ref{chapter3:mast}), the model is modified before differentiation and the nominal objective may no longer be the effective target. In private optimization, clipping or normalization bounds sensitivity but also changes the update and its convergence: Chapter~\ref{chapter4:adaptive_clipping} analyzes adaptive quantile clipping, while Chapter~\ref{chapter5:smoothed_normalization} develops smoothed normalization with error compensation. The final pair concerns LMO-based optimizers. Chapter~\ref{chapter6:gluon} develops layer-wise assumptions and geometry, and Chapter~\ref{chapter7:beyond_the_ideal} analyzes an approximate inner computation. In each direction, the analysis keeps the feature that departs from the standard model and studies its effect directly.

\subsection{Submodel and sketch-aware optimization}

In submodel and sparse training, each step operates on only part of the model. Dropout and random masking can be viewed as training with random submodels \citep{srivastava2014dropout}. Sparse and masked methods reduce computational or memory cost by activating only part of the model \citep{mohtashami2022masked}. In distributed and federated settings, model heterogeneity and resource constraints motivate training or deploying different submodels across devices \citep{caldas2018expanding,diao2020heterofl,jiang2022model}. The aim is to reduce memory, communication, or device requirements while still training a larger shared model. Whether the resulting procedure still optimizes the same objective is a separate question.

In synchronous data parallelism, every worker stores the same full model $x$. For the distributed finite-sum problem
\begin{equation*}
    f(x)=\frac1n\sum_{i=1}^n f_i(x),
\end{equation*}
a deterministic baseline is distributed gradient descent:
\begin{equation*}
    x^{k+1}
    =
    x^k
    -
    \gamma_k
    \frac1n
    \sum_{i=1}^n
    \nabla f_i(x^k).
\end{equation*}
Each worker evaluates its local gradient at the same shared iterate $x^k$, and the server averages the resulting directions. Stochastic or mini-batch variants replace the local gradients by stochastic estimators. This distributes data and computation, but not the model itself.

Submodel training changes something more fundamental: the object on which each worker computes a gradient. If worker $i$ evaluates its loss on a restricted or sketched model, the resulting direction is not necessarily a compressed version of the full gradient. It is the gradient of a different local object.

The next two chapters take complementary views. Chapter~\ref{chapter2:ist} shows that submodel gradients can shift the fixed point for the original objective. Chapter~\ref{chapter3:mast} puts the sketching mechanism into the objective, making sketched updates natural for the modified problem.

\subsubsection{Independent Subnetwork Training: Biased Submodel Update}

Independent Subnetwork Training (\algname{IST}) is a distributed method in which workers train different subnetworks and periodically combine their parameters. \algname{IST} was proposed to combine data parallelism and model parallelism: instead of requiring each device to train the full model, each worker trains a smaller submodel, and the full model is assembled from these independently trained parts \citep{yuan2022distributed}. Unlike pure data parallelism, this splits the model itself, not only the data. Related work studied masked and subnetwork training in neural-network and federated settings \citep{liao2022on,dun2022resist,wolfe2021gist}. Such methods can work well empirically, but standard distributed-\algname{SGD} or compressed-gradient theory does not fully explain their optimization behavior.

\algname{IST} may initially look like another form of compressed distributed optimization. Compressed \algname{SGD} computes or estimates a full-model gradient and then compresses the communicated vector. Under standard unbiased-compression assumptions, this still gives an unbiased estimator of the true gradient \citep{alistarh2017qsgd}. \algname{IST} reverses the order: it sketches the model before differentiating. The gradient is therefore taken at a worker-specific submodel, not at the common full-model iterate. In the notation of Chapter~\ref{chapter2:ist}, a simplified update direction has the form
\begin{equation*}
    g^k
    =
    \frac{1}{n}\sum_{i=1}^{n}
    \mathbf C_i^k
    \nabla f_i(\mathbf C_i^k x^k),
\end{equation*}
where $\mathbf C_i^k$ is the submodel-selection or sketch matrix used by worker $i$ at iteration $k$. Even if the sketches are unbiased in a suitable sense, the nonlinearity of the gradient map means that $g^k$ need not be an unbiased estimator of $\nabla f(x^k)$. Sketching and differentiation do not commute in general. The question is not only how much communication or memory is saved, but whether the method optimizes the original objective.

A quadratic finite-sum model makes the mismatch explicit. In interpolation-style quadratic settings, \algname{IST} can behave well and can converge under appropriate matrix conditions. In heterogeneous settings, however, the expected update may contain an additional offset. Informally, one can think of the expected direction as having the form
\begin{equation*}
    \mathbb E[g^k\mid x^k]
    =
    \mathbf P\nabla f(x^k)+h,
\end{equation*}
where $\mathbf P$ is a preconditioning-like matrix and $h$ is a nonzero term caused by the interaction between sketching and heterogeneity. When $h\neq 0$, the fixed point of the method need not coincide with the minimizer of the original finite-sum objective. As a result, the method can exhibit an irreducible error term that does not disappear merely by decreasing the stepsize.

\algname{IST} therefore cannot, in general, be interpreted as \algname{SGD} with a different communication pattern. The gradient is evaluated on a different model, and this can change the limiting point. Chapter~\ref{chapter2:ist} uses this observation to identify when the original empirical-risk objective is an inadequate description of the method.

\begin{samepage}
\medskip
\noindent\textbf{Chapter~\ref{chapter2:ist} is based on:}
\par\smallskip

\noindent Egor Shulgin and Peter Richt{\'a}rik, ``Towards a Better Theoretical Understanding of Independent Subnetwork Training,'' \emph{International Conference on Machine Learning (ICML)}, 2024.
\end{samepage}

\subsubsection{Model-Agnostic Sparsified Training: A Sketch-Aware Objective}

Chapter~\ref{chapter3:mast} takes a different route. Model-Agnostic Sparsified Training (MAST) does not treat sketching as noise around the original objective; it places the sketch distribution inside the objective itself.

Let $\mathbf S$ be a random sketch matrix representing sparsification, masking, dropout, or submodel selection. For a reference point $v$, MAST defines the sketched function
\begin{equation*}
    f_{\mathbf S}(x)
    =
    f\bigl(v+\mathbf S(x-v)\bigr),
\end{equation*}
whose gradient is
\begin{equation*}
    \nabla f_{\mathbf S}(x)
    =
    \mathbf S^\top
    \nabla f\bigl(v+\mathbf S(x-v)\bigr).
\end{equation*}
The MAST objective is then the average sketched objective
\begin{equation*}
    f_{\mathcal D}(x)
    =
    \mathbb E_{\mathbf S\sim\mathcal D}
    \bigl[
        f_{\mathbf S}(x)
    \bigr],
\end{equation*}
where $\mathcal D$ is the sketch distribution. The reference point $v$ remains part of the formulation, so the sketch acts on the displacement $x-v$ rather than on $x$ alone. If a model is trained or evaluated through sketches, the optimization problem itself then includes that mechanism. The resulting objective connects to dropout, sparse training, resource-constrained learning, and the \algname{IST} analysis of Chapter~\ref{chapter2:ist} \citep{srivastava2014dropout,caldas2018expanding,mohtashami2022masked,shulgin2023towards}.

The corresponding double-sketched gradient method makes the contrast with \algname{IST} explicit. A simple iteration has the form
\begin{equation*}
    x^{k+1}
    =
    x^k
    -
    \gamma_k
    \nabla f_{\mathbf S^k}(x^k)
    =
    x^k
    -
    \gamma_k
    (\mathbf S^k)^\top
    \nabla f\bigl(v+\mathbf S^k(x^k-v)\bigr),
    \qquad
    \mathbf S^k\sim\mathcal D.
\end{equation*}
For the original objective $f$, such a sketch-induced direction may be biased. For the MAST objective, however, it is unbiased:
\begin{equation*}
    \mathbb E_{\mathbf S^k\sim\mathcal D}
    \left[
        \nabla f_{\mathbf S^k}(x^k)
    \right]
    =
    \nabla f_{\mathcal D}(x^k).
\end{equation*}

The MAST objective remains tractable. Smoothness, convexity, and strong convexity of the original objective are inherited by $f_{\mathcal D}$, with constants controlled by spectral properties and moments of the sketch distribution. Chapter~\ref{chapter3:mast} then proves convergence guarantees in nonconvex, convex, strongly convex, stochastic, distributed, and variance-reduced settings. Sparse or dropout-like training can therefore be viewed as solving a different, well-defined optimization problem rather than as an uncontrolled perturbation of full-model training.

\begin{samepage}
\medskip
\noindent\textbf{Chapter~\ref{chapter3:mast} is based on:}
\par\smallskip

\noindent Yury Demidovich, Grigory Malinovsky, Egor Shulgin, and Peter Richt{\'a}rik, ``MAST: Model-Agnostic Sparsified Training,'' \emph{International Conference on Learning Representations (ICLR)}, 2025.
\end{samepage}

Chapters~\ref{chapter2:ist} and~\ref{chapter3:mast} separate two questions: whether a submodel update solves the nominal objective, and what objective it solves naturally. The \algname{IST} analysis identifies the mismatch; MAST makes the induced target explicit.

\subsection{Private optimization: adaptive clipping and smoothed normalization}

Differential privacy provides a standard mathematical framework for learning from sensitive data \citep{dwork2014algorithmic}. In deep learning, the canonical optimization method is \algname{DP-SGD} \citep{abadi2016deep}. It bounds each per-example contribution before adding noise, so the privacy mechanism becomes part of the optimization update. In a simplified form, one iteration can be written as
\begin{equation*}
    g_j^k = \nabla f_{\xi_j}(x^k),
    \qquad
    \bar g^k
    =
    \frac{1}{B}
    \sum_{j\in \mathcal B_k}
    \operatorname{clip}_{\tau}(g_j^k),
    \qquad
    x^{k+1}
    =
    x^k-\gamma_k(\bar g^k+z^k).
\end{equation*}
Here $\mathcal B_k$ is the minibatch sampled at iteration $k$, $B=|\mathcal B_k|$, and $\xi_j$ denotes the data sample associated with index $j$. Thus, $g_j^k$ is the corresponding per-example gradient. The vector $z^k$ is Gaussian noise calibrated to the clipping threshold and the desired privacy guarantee. The clipping operator is
\begin{equation*}
    \operatorname{clip}_{\tau}(g)
    =
    g\min\left\{1,\frac{\tau}{\|g\|}\right\}.
\end{equation*}
The clipping step is essential for bounding sensitivity, but it is also an optimization operation: it changes the update direction before the privacy noise is added.

Choosing the clipping threshold $\tau$ is one of the central practical difficulties in DP learning. If $\tau$ is too small, the method clips aggressively and the update can be strongly distorted. If $\tau$ is too large, the sensitivity bound and the privacy noise increase. The problem is compounded by the fact that hyperparameter tuning is itself delicate in private learning: tuning may require access to sensitive data, repeated training runs may consume additional privacy budget or require a separate private model-selection procedure, and evaluating many candidate hyperparameters can be computationally expensive \citep{papernot2021hyperparameter,koskela2023practical}. This makes adaptive or nearly tuning-free clipping mechanisms especially attractive.

\subsubsection{Adaptive Clipping for Differentially Private SGD}

Adaptive quantile clipping was introduced in differentially private federated learning to reduce the need to tune a fixed clipping threshold manually \citep{andrew2021differentially}. Unlike standard clipped SGD, where $\tau$ is chosen in advance, it sets the threshold from the current distribution of stochastic-gradient norms. For a target quantile $p\in(0,1)$, define
\begin{equation*}
    \tau_p(x)
    =
    \inf\left\{
        t\ge 0:
        \mathbb{P}\{\|g(x)\|\le t\}\ge p
    \right\}.
\end{equation*}
For example, $p=1/2$ sets the threshold to the median stochastic-gradient norm, so about half of the stochastic gradients are clipped.

Chapter~\ref{chapter4:adaptive_clipping} analyzes \algname{SGD} with this adaptive threshold in the smooth nonconvex setting under bounded-moment assumptions that allow heavy-tailed stochastic-gradient noise. It covers quantile-clipped \algname{SGD} (\algname{QC-SGD}) and its differentially private extension \algname{DP-QC-SGD}. Compared with fixed-threshold clipped \algname{SGD}, the analysis must account for a threshold that changes with the stochastic-gradient distribution.

Earlier work showed that clipping can prevent \algname{SGD} from reaching a stationary point and can introduce a systematic bias \citep{chen2020understanding,koloskova2023revisiting}. Chapter~\ref{chapter4:adaptive_clipping} gives an explicit example showing an analogous obstruction for \algname{QC-SGD}: with a fixed quantile, the method can retain a non-vanishing clipping bias. The convergence upper bounds contain a corresponding irreducible term.

The bias can be controlled by varying the quantile over time. With an appropriate quantile and stepsize schedule, \algname{QC-SGD} achieves nonconvex stationarity rates that are optimal up to logarithmic factors under heavy-tailed stochastic-gradient noise. The schedule gradually changes how much of the stochastic-gradient distribution is clipped, rather than committing to one fixed fraction throughout training. Quantile clipping can also serve as a robust stochastic optimization method outside differential privacy. In the private extension, the quantile also sets the Gaussian noise scale, because the noise is calibrated to the adaptive threshold; the choice of quantile therefore trades clipping bias against privacy noise.

\begin{samepage}
\medskip
\noindent\textbf{Chapter~\ref{chapter4:adaptive_clipping} is based on:}
\par\smallskip

\noindent Egor Shulgin and Peter Richt{\'a}rik, ``On the Convergence of DP-SGD with Adaptive Clipping,'' \emph{NeurIPS Workshop on Optimization for Machine Learning}, 2024.
\end{samepage}

\subsubsection{Smoothed Normalization for Distributed Private Optimization}

Chapter~\ref{chapter5:smoothed_normalization} changes the bounded operator itself. Returning to the distributed finite-sum model
\begin{equation*}
    f(x)=\frac1n\sum_{i=1}^n f_i(x),
\end{equation*}
where worker $i$ has access to $f_i$, a natural clipped distributed-gradient method performs
\begin{equation*}
    x^{k+1}
    =
    x^k
    -
    \gamma_k
    \frac1n
    \sum_{i=1}^n
    \operatorname{clip}_{\tau}\bigl(\nabla f_i(x^k)\bigr).
\end{equation*}
However, clipping the local gradients can distort the average direction and prevent convergence under standard assumptions; analyses based on uniformly bounded gradients can hide this effect by making clipping essentially inactive or by producing bounds that do not capture the true difficulty \citep{khirirat2023clip21}. The directional distortion appears even in a two-worker deterministic example. Let
\[
    g_1=(4,0),\qquad g_2=(0,1),\qquad \tau=1.
\]
Then
\[
    \frac{g_1+g_2}{2}
    =
    \left(2,\frac12\right),
    \qquad
    \frac{\operatorname{clip}_1(g_1)
    +\operatorname{clip}_1(g_2)}{2}
    =
    \left(\frac12,\frac12\right).
\]
The two averages are not parallel, so local clipping can change the aggregate descent direction even without stochastic noise. This motivates the question studied in Chapter~\ref{chapter5:smoothed_normalization}: can one design a bounded transformation whose effect is easier to control in distributed private optimization?

Smoothed normalization replaces hard clipping by the smoother transformation
\begin{equation*}
    \operatorname{Norm}_{\alpha}(g)
    =
    \frac{g}{\alpha+\|g\|},
    \qquad \alpha>0,
\end{equation*}
which has also been studied as an alternative to clipping in private optimization \citep{yang2022normalized,yu2023smoothed,bu2024automatic}. Like clipping, this transformation bounds the message magnitude and can control sensitivity. Unlike hard clipping, it changes smoothly with $\|g\|$, which gives a contractive-like property and makes its distortion easier to track analytically. The parameter $\alpha$ controls the transition between an approximately linear response near the origin and normalization for large inputs. Normalization still distorts the update, so the method combines it with error compensation, building on error feedback and its later theory in communication-efficient distributed optimization \citep{Seide2014,karimireddy2019error,EF21}.

A compact form of \algname{(DP-)$\alpha$-NormEC} is
\begin{equation*}
    \Delta_i^k
    =
    \operatorname{Norm}_{\alpha}\!\left(\nabla f_i(x^k)-g_i^k\right),
    \qquad
    g_i^{k+1}
    =
    g_i^k+\beta\Delta_i^k,
    \qquad i=1,\ldots,n,
\end{equation*}
and
\begin{equation*}
    \widehat g^{k+1}
    =
    \widehat g^k
    +
    \frac{\beta}{n}
    \sum_{i=1}^n
    \Delta_i^k
    +
    z^k,
    \qquad
    x^{k+1}
    =
    x^k
    -
    \gamma_k
    \frac{\widehat g^{k+1}}{\|\widehat g^{k+1}\|}.
\end{equation*}
Here $g_i^k$ are local memory variables, $\widehat g^k$ is the server-side aggregate memory, and $z^k$ denotes the aggregate Gaussian noise added in the private version. The memory variables store information that is not fully incorporated in one step, so the distortion introduced by normalization is not simply discarded. This is analogous in spirit to error feedback for compression, but the bounded operator here is motivated by privacy and sensitivity control.

Chapter~\ref{chapter5:smoothed_normalization} proves convergence guarantees for \algname{$\alpha$-NormEC} in smooth nonconvex distributed optimization and extends them to the differentially private setting. The analysis neither assumes bounded gradients, which could make the operator effectively inactive, nor drops the operator from the communicated update. This gives the first convergence guarantees for a differentially private distributed optimization method under standard smoothness assumptions.

\begin{samepage}
\medskip
\noindent\textbf{Chapter~\ref{chapter5:smoothed_normalization} is based on:}
\par\smallskip

\noindent Egor Shulgin, Sarit Khirirat, and Peter Richt{\'a}rik, ``Smoothed Normalization for Efficient Distributed Private Optimization,'' \emph{arXiv preprint arXiv:2502.13482}, 2025. A version appeared as ``Private Federated Learning with Provable Convergence via Smoothed Normalization'' at the \emph{ICML Workshop on Collaborative and Federated Agentic Workflows (CFAgentic)}, 2025.
\end{samepage}

Chapters~\ref{chapter4:adaptive_clipping} and~\ref{chapter5:smoothed_normalization} address the same distortion from two sides. Chapter~\ref{chapter4:adaptive_clipping} quantifies the bias of an existing adaptive clipping rule. Chapter~\ref{chapter5:smoothed_normalization} replaces hard clipping by a smooth bounded operator and uses memory to compensate for the remaining distortion. Privacy-motivated transformations should therefore be analyzed as optimization operators, not as harmless preprocessing steps.

\subsection{LMO-based optimizers and inexact updates}

The final pair of chapters concerns deep-learning optimizers that are not standard Euclidean gradient methods. For many years, \algname{Adam} and \algname{AdamW} have been the default choices in practice \citep{kingma2015adam,loshchilov2019decoupled}. Their adaptive updates act coordinate-wise, treating parameters as arrays of scalar entries. More recently, \algname{Shampoo} \citep{gupta2018shampoo}, \algname{Muon} \citep{jordan2024muon}, \algname{Scion} and related norm-constrained LMO methods \citep{pethick2025training}, and large-scale \algname{Muon} variants \citep{liu2025muon,team2025kimi} have drawn attention to updates that use the matrix structure of neural-network layers. This structure changes both the geometry of the outer step and the computation needed to produce its direction.

Below, capital letters denote matrix-valued quantities and collections of matrix blocks, reflecting the structure of the parameters and updates; the summary of Chapter~\ref{chapter7:beyond_the_ideal} follows that chapter's lowercase notation.

A useful abstraction for this family is a linear minimization oracle (LMO), familiar from conditional-gradient or Frank--Wolfe-type methods \citep{frank1956algorithm,jaggi2013revisiting}. Given a gradient or momentum-like matrix $M^k$, an LMO direction over a norm ball of radius $\rho>0$ can be written as
\begin{equation*}
    D^k
    \in
    \argmin_{\|D\|\le \rho}
    \langle M^k,D\rangle .
\end{equation*}
The choice of norm determines the update direction. The Euclidean norm ball gives a normalized-gradient direction, while the $\ell_{\infty}$-ball gives a sign-type direction. For matrix parameters, the spectral-norm ball leads to an orthogonalized direction. If $M=U\Sigma V^\top$ is an SVD, then an LMO minimizer is
\begin{equation*}
    D=-\rho UV^\top,
\end{equation*}
while $UV^\top$ is the polar factor of $M$. Up to sign and scaling, this polar factor is the core of \algname{Muon}-style updates. In practice, \algname{Muon} typically applies the polar factor to a momentum-like matrix rather than directly to the raw gradient.

\algname{Muon} differs from \algname{SGD} and \algname{Adam} in two ways that matter here. It treats weight matrices as matrices rather than as unstructured vectors or coordinate-wise arrays, and it requires an inner polar-factor computation before the outer step. Chapters~\ref{chapter6:gluon} and~\ref{chapter7:beyond_the_ideal} address these two issues in sequence.

\subsubsection{Gluon: Layer-Wise Generalized Smoothness for Muon-Style LMO Optimizers}

\algname{Muon} updates whole weight matrices layer by layer, but early analyses of it relied on a single global smoothness constant and an idealized description of the optimizer. Chapter~\ref{chapter6:gluon} starts from the smoothness model: what assumptions are appropriate when different layers use different geometries and behave differently along the training trajectory? The chapter then develops an LMO optimizer framework on top of this model.

\algname{Gluon} models the parameters as blocks or layers, $X=(X_1,\ldots,X_p)$. Each block can have its own norm, radius, and LMO direction, so a matrix layer need not share its geometry or update scale with an embedding, bias, or normalization parameter. A representative layer-wise direction has the form
\begin{equation*}
    D_i^k
    \in
    \argmin_{\|D_i\|_{(i)}\le t_i^k}
    \langle M_i^k,D_i\rangle,
    \qquad i=1,\ldots,p,
\end{equation*}
where $M_i^k$ is a gradient or momentum-like matrix for block $i$, $\|\cdot\|_{(i)}$ is the block norm, and $t_i^k$ is the corresponding update radius. This LMO viewpoint recovers normalized and sign-type methods with momentum, as well as \algname{Muon}-style matrix updates.

Coordinate-wise theory provides a useful precedent. Coordinate-descent analyses use coordinate-specific smoothness constants \citep{nesterov2012efficiency}, while adaptive online methods and \algname{AdaGrad} scale coordinates according to the observed gradients \citep{mcmahan2010adaptive,duchi2011adaptive}. More recent analyses under anisotropic or refined smoothness conditions obtain more favorable guarantees for \algname{AdaGrad} than models based on one global constant \citep{liu2024adagradanisotropicsmoothness,jiang2024convergence}. Related geometry-aware work explains \algname{Adam} through favorable $\ell_\infty$-geometry \citep{xie2024adam}.

\algname{Muon} instead transforms entire weight matrices. Its update calls for a layer-wise and non-Euclidean model, with potentially different geometry and smoothness scales across blocks. \algname{Gluon} combines this layer-wise view with gradient-dependent generalized smoothness, proposed as a better model for neural-network training than uniform smoothness \citep{zhang2020why}. A representative layer-wise condition is
\begin{equation*}
    \|\nabla_i f(Y)-\nabla_i f(X)\|_{(i),*}
    \le
    \left(
        L_i^0+
        L_i^1\|\nabla_i f(X)\|_{(i),*}
    \right)
    \|Y_i-X_i\|_{(i)} ,
\end{equation*}
for every block $i$ and all $X$ and $Y$; the right-hand side depends only on the change $Y_i-X_i$ in block $i$. Compared with standard smoothness, this condition is both layer-wise and gradient-dependent: different matrices and layers can have different constants, and the effective smoothness can vary along the training trajectory.

Under this model, Chapter~\ref{chapter6:gluon} proves deterministic and stochastic convergence guarantees for \algname{Gluon}-type methods.
With the adaptive radii
\[
    t_i^k
    =
    \frac{\|\nabla_i f(X^k)\|_{(i),*}}
    {L_i^0+L_i^1\|\nabla_i f(X^k)\|_{(i),*}},
\]
deterministic \algname{Gluon} guarantees
\[
    \min_{0\le k<K}
    \sum_{i=1}^p
    \frac{1/L_i^1}{\frac1p\sum_{j=1}^p 1/L_j^1}
    \|\nabla_i f(X^k)\|_{(i),*}
    \le
    \varepsilon
\]
after
\begin{equation*}
    K
    =
    O\!\left(
    \frac{
        \Delta^0
        \sum_{i=1}^p L_i^0/(L_i^1)^2
    }{
        \varepsilon^2
        \left(\frac1p\sum_{j=1}^p 1/L_j^1\right)^2
    }
    +
    \frac{
        \Delta^0
    }{
        \varepsilon
        \left(\frac1p\sum_{j=1}^p 1/L_j^1\right)
    }
    \right),
\end{equation*}
where $\Delta^0=f(X^0)-f^{\inf}$. The key feature of this bound is that it depends on the full collection of layer-wise smoothness quantities $\{L_i^0,L_i^1\}_{i=1}^p$, rather than on one worst-case global constant.

\begin{figure}[H]
\centering
\begin{minipage}[t]{0.46\textwidth}
    \centering
    \includegraphics[width=\linewidth,trim={605pt 0pt 555pt 0pt},clip]{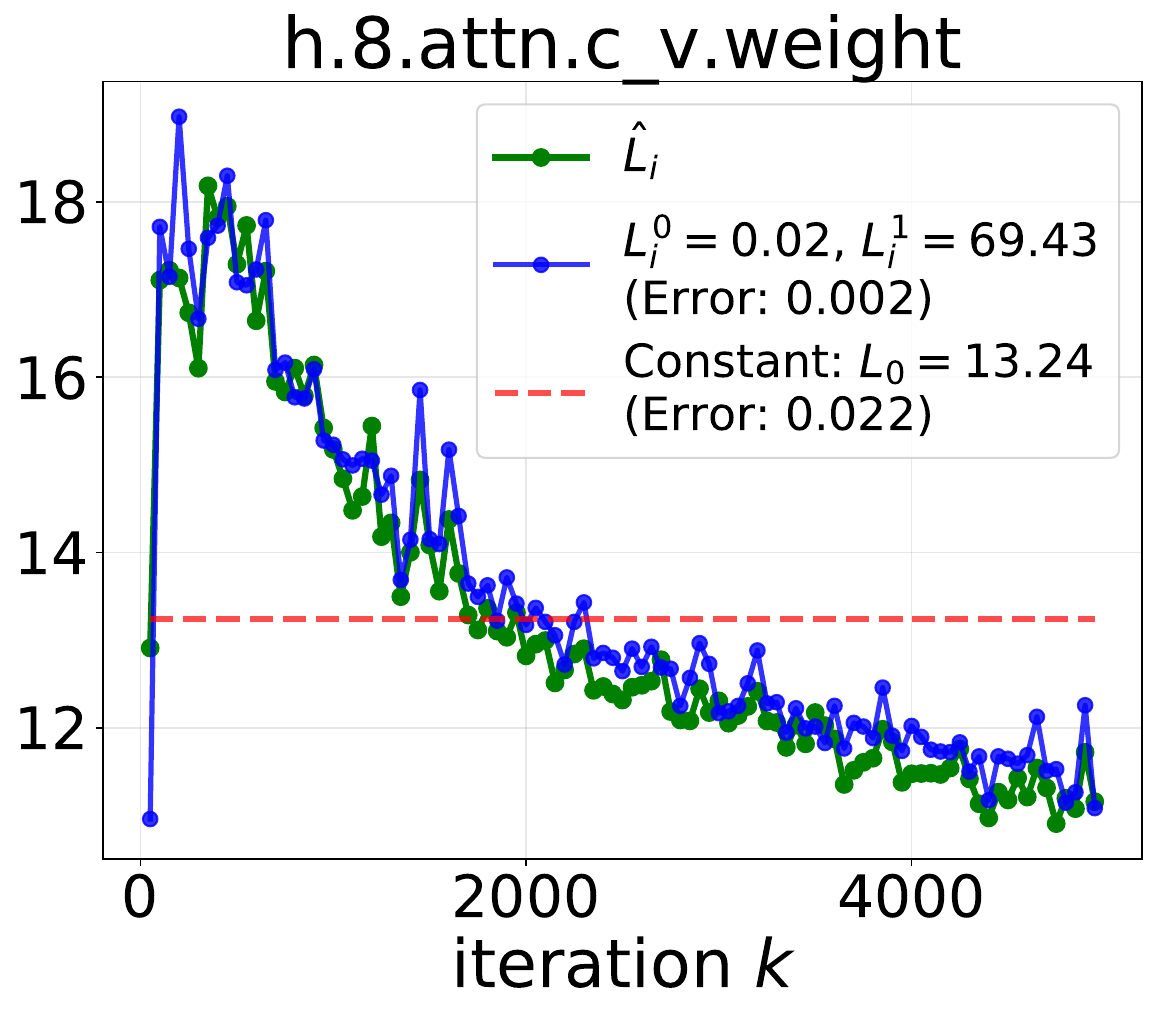}
\end{minipage}\hfill
\begin{minipage}[t]{0.52\textwidth}
    \centering
    \includegraphics[width=\linewidth]{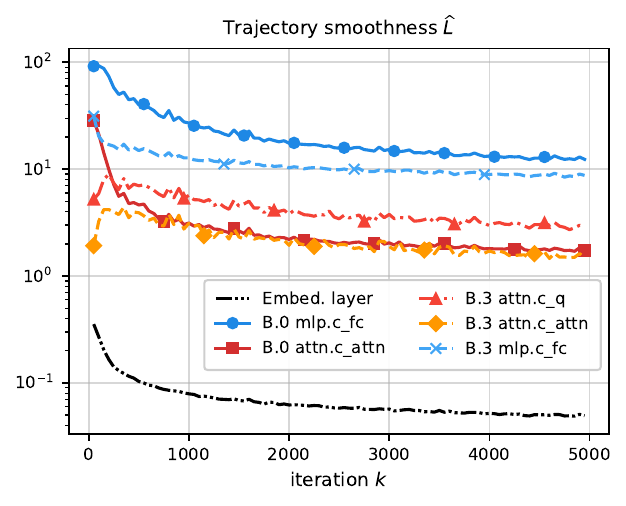}
\end{minipage}
\caption{Representative trajectory-smoothness measurements from Chapter~\ref{chapter6:gluon}. Left: the fitted layer-wise $(L_i^0,L_i^1)$ model tracks observed trajectory smoothness better than a constant-smoothness proxy. Right: trajectory smoothness varies substantially across layers and matrix groups, motivating layer-wise constants rather than a single global smoothness parameter.}
\label{fig:intro_gluon_smoothness}
\end{figure}

Figure~\ref{fig:intro_gluon_smoothness} shows representative trajectory measurements. The fitted layer-wise model closely follows the observed smoothness, while a constant-smoothness proxy can be substantially less accurate. The estimates also vary strongly across layers and matrix types. The variation appears within a single training run, not only across models or datasets.

The bound and the measurements complement each other. The adaptive radii $t_i^k$ vary across layers and along the trajectory, so they act as layer-wise learning rates. The two terms in the complexity bound describe different regimes. The $1/\varepsilon^2$ term is weighted by $\sum_{i=1}^p L_i^0/(L_i^1)^2$, whereas the $1/\varepsilon$ term depends on the constants $L_j^1$ only through their harmonic mean $\bigl(\frac1p\sum_{j=1}^p 1/L_j^1\bigr)^{-1}$.
For several measured layers, the fitted $L_i^0$ is close to zero. When this makes the first term small at the target accuracy, the $1/\varepsilon$ term can dominate, giving a potentially sharper dependence than a standard analysis based on one global smoothness constant. The comparison is not one-to-one, since the stationarity measure and norms are layer-wise, but it shows how exploiting layer heterogeneity can improve the guarantee.

The contribution of Chapter~\ref{chapter6:gluon} is therefore first a layer-wise smoothness model and second the \algname{Gluon} optimizer framework built on top of that model. The trajectory measurements support the model, while the convergence results show how the resulting layer-wise quantities can guide radii or learning-rate scales.

\begin{samepage}
\medskip
\noindent\textbf{Chapter~\ref{chapter6:gluon} is based on:}
\par\smallskip

\noindent Artem Riabinin, Egor Shulgin, Kaja Gruntkowska, and Peter Richt{\'a}rik, ``From Muon to Gluon: Bridging Theory and Practice of LMO-Based Optimizers for LLMs,'' \emph{International Conference on Machine Learning (ICML)}, 2026.
\end{samepage}

\subsubsection{Beyond the Ideal: Analyzing the Inexact Muon Update}

One idealization remains: practical \algname{Muon} does not compute an exact polar direction. Exact polar-factor computation is too expensive at every iteration for every matrix, so implementations use approximate inner routines such as Newton--Schulz or PolarExpress \citep{amsel2025polar,grishina2025accelerating}. Spectral or polar-type updates had been studied before, but fast approximate orthogonalization made them cheap enough for modern training.

The chapter represents the implemented update through an inexact LMO, measuring error directly in direction space without requiring the approximate output to remain feasible for the exact norm ball. In the notation of Chapter~\ref{chapter7:beyond_the_ideal}, if $d^k$ is the exact LMO direction, the implemented direction $\hat d^k$ satisfies
\begin{equation*}
    \|\hat d^k-d^k\|\le \delta_k.
\end{equation*}
This condition is motivated by the behavior of practical polar-factor approximation routines such as PolarExpress. It abstracts the fact that the optimizer does not use the exact oracle direction, while keeping the model simple enough for convergence analysis.

The analysis tracks how this inexactness propagates through convergence. In the deterministic setting with a constant inexactness level $\delta$, one representative result states that the optimal constant stepsize is
\begin{equation*}
    \gamma^\star = \frac{1}{1+\delta} \sqrt{\frac{2\Delta^0}{KL}},
\end{equation*}
and the average gradient norm satisfies
\begin{equation*}
    \frac1K\sum_{k=0}^{K-1}\|\nabla f(x^k)\|_*
    \le
    \frac{1+\delta}{1-\delta}
    \sqrt{\frac{2\Delta^0 L}{K}}.
\end{equation*}
The exact-oracle behavior is recovered as $\delta\to0$, while larger inexactness worsens the constant and changes the optimal stepsize. In the stochastic setting with momentum, less precise orthogonalization calls for a smaller stepsize and a larger momentum parameter, that is, a shorter momentum memory. The number of inner orthogonalization steps is therefore an optimization hyperparameter, not just an implementation detail.

NanoGPT and CIFAR-10 experiments show that the precision of the approximate orthogonalization affects convergence and hyperparameter sensitivity. Fewer inner iterations can shift the best learning rate and narrow the range of stable choices, consistent with the dependence on inexactness in the bounds. Additional diagnostics find substantial variation in approximation difficulty across layers and matrix families: attention and MLP matrices need not reach comparable errors after the same number of PolarExpress steps. A single inner-iteration budget may therefore be unnecessarily rigid.

\begin{samepage}
\medskip
\noindent\textbf{Chapter~\ref{chapter7:beyond_the_ideal} is based on:}
\par\smallskip

\noindent Egor Shulgin, Sultan AlRashed, Francesco Orabona, and Peter Richt{\'a}rik, ``Beyond the Ideal: Analyzing the Inexact Muon Update,'' \emph{International Conference on Artificial Intelligence and Statistics (AISTATS)}, 2026.
\end{samepage}

Chapters~\ref{chapter6:gluon} and~\ref{chapter7:beyond_the_ideal} remove two simplifications in sequence. \algname{Gluon} models the layer-wise geometry and smoothness of the objective; Chapter~\ref{chapter7:beyond_the_ideal} models the approximate inner computation. Both affect the convergence guarantees and the hyperparameters used in practice.

\subsubsection*{Notation conventions}

The technical chapters follow the notation of the source papers, and notation is intentionally local. Throughout this Introduction, $f$ denotes the relevant objective or surrogate objective, $x^k$ a generic iterate, $\mathbb E$ expectation, and $\|\cdot\|$ a norm whose precise meaning depends on context. In the submodel and MAST chapters, bold matrices such as $\mathbf C$ and $\mathbf S$ denote sketching or submodel-selection operators. In the private-optimization chapters, $(\varepsilon,\delta)$ denotes the privacy budget, while clipping thresholds and normalization parameters are defined locally. In the LMO-based chapters, capital letters such as $X$, $M$, and $D$ often denote block or matrix-valued quantities; in Chapter~\ref{chapter7:beyond_the_ideal}, $\delta$ denotes oracle inexactness rather than a privacy parameter. Appendix~\ref{appendix:common_facts} collects reusable basic inequalities; all chapter-specific assumptions and notation are introduced in the corresponding chapters.

\section{Papers not included as full chapters}

The thesis does not include all of the work I coauthored during my time at KAUST.
The six technical chapters were selected to form a coherent narrative around
practical optimization methods whose behavior is only partly captured by standard
theoretical models. The remaining papers span several related research directions.

Some of this work concerns stochastic and convex optimization, including stochastic
extragradient methods and adaptive acceleration for smooth convex problems
\citep{mishchenko2020revisiting,ivanova2021adaptive}. A separate line studies
decentralized optimization over time-varying networks, including accelerated methods
and optimal convergence rates \citep{ADOM,rogozin2021towards}. Communication-efficient
distributed optimization forms another direction, with work on fundamental limits
and optimal constructions for compression operators and on the shifted-compression
framework \citep{UP2020,shulgin2022shifted}.

The federated-learning papers address several different questions. \algname{SPAM} studies
nonconvex optimization for cross-device federated learning
\citep{karagulyan2024spam}, while a separate work develops certified robustness
guarantees for federated learning \citep{alfarra2022certified}. \algname{Fed-$\alpha$-NormEC}
directly extends Chapter~\ref{chapter5:smoothed_normalization} to a more practical
federated setting with partial participation and local computation
\citep{shulgin2025practicalprivatefl}.

Several recent papers continue the optimizer-theory direction of
Chapters~\ref{chapter6:gluon} and~\ref{chapter7:beyond_the_ideal}. One derives
scaling laws for the learning rate, momentum, and batch size of LMO-based methods
as the training horizon or token budget changes
\citep{shulgin2026scaling_laws}. Another develops a more general stochastic analysis
under the ABC/expected-smoothness condition, going beyond the bounded-variance
models used in the thesis and exposing additional batch-size--momentum trade-offs
\citep{shulgin2026general_lmo}. The \algname{MARS} analysis identifies when scaling the
momentum-correction term can provide a provable improvement
\citep{shulgin2026mars}. Finally, a quadratic analysis of \algname{Muon} studies polar gradient
descent on matrix-quadratic problems and compares Euclidean, spectral, and entrywise
$\ell_\infty$ geometries \citep{shulgin2026quadratic_muon}.

\FloatBarrier
\section{Contributions and thesis organization}

Table~\ref{tab:thesis_map} maps the six technical chapters to the modeling questions of Section~\ref{sec:common_viewpoint} and summarizes their main messages.

\begin{table}[!ht]
\centering
\small
\caption{Thesis map: main objects, modeled features, and messages.}
\label{tab:thesis_map}
\renewcommand{\arraystretch}{1.12}
\begin{tabularx}{\textwidth}{
@{}
>{\centering\arraybackslash}p{0.04\textwidth}
>{\raggedright\arraybackslash}p{0.22\textwidth}
>{\raggedright\arraybackslash}p{0.155\textwidth}
>{\raggedright\arraybackslash}X
@{}}
\toprule
Ch. & Object & Modeled feature & Main message \\
\midrule
\hyperref[chapter2:ist]{2} &
\strut\algname{IST} / submodel-gradient training &
Objective &
Sketching before differentiation biases the update, and the method can converge to a shifted fixed point. \\
\specialrule{0.1pt}{2pt}{2pt}

\hyperref[chapter3:mast]{3} &
MAST / sketch-aware objective &
Objective &
Placing the sketch distribution in the objective makes sketched gradients unbiased for the induced target. \\
\specialrule{0.4pt}{2pt}{2pt}

\hyperref[chapter4:adaptive_clipping]{4} &
Adaptive quantile clipping &
Update &
Fixed quantiles retain clipping bias; a time-varying quantile schedule removes it and recovers optimal rates. \\
\specialrule{0.1pt}{2pt}{2pt}

\hyperref[chapter5:smoothed_normalization]{5} &
Smoothed normalization / \algname{$\alpha$-NormEC} &
Update &
Smoothed normalization with error compensation gives private distributed guarantees without bounded gradients. \\
\specialrule{0.4pt}{2pt}{2pt}

\hyperref[chapter6:gluon]{6} &
\strut\algname{Gluon} / layer-wise generalized smoothness &
Assumptions &
Layer-wise generalized smoothness matches training trajectories and yields guarantees with per-layer constants. \\
\specialrule{0.1pt}{2pt}{2pt}

\hyperref[chapter7:beyond_the_ideal]{7} &
Inexact \algname{Muon} / LMO updates &
Update &
Orthogonalization precision changes the convergence bound and the best stepsize. \\
\bottomrule
\end{tabularx}
\end{table}
\FloatBarrier

\noindent
The main technical contributions, listed in chapter order, are:

\begin{enumerate}

\item \textbf{Independent Subnetwork Training
(Chapter~\ref{chapter2:ist}).}
We analyze \algname{IST} in representative quadratic models and show that gradients
computed after sketching can bias the update for the original finite-sum
objective. Under heterogeneity, the expected dynamics can converge to a shifted
fixed point, producing an irreducible error that is not removed by decreasing
the stepsize.

\item \textbf{Model-Agnostic Sparsified Training
(Chapter~\ref{chapter3:mast}).}
We introduce the MAST objective, which incorporates the sketch distribution
directly into the optimization problem. The resulting sketched-gradient
estimator is unbiased for this objective. We establish its smoothness and
convexity properties and prove convergence guarantees in nonconvex, convex,
strongly convex, stochastic, distributed, and variance-reduced settings.

\item \textbf{Adaptive quantile clipping
(Chapter~\ref{chapter4:adaptive_clipping}).}
We provide a convergence analysis of \algname{QC-SGD} and \algname{DP-QC-SGD} under heavy-tailed
stochastic-gradient noise. An explicit example shows that a fixed quantile can
retain non-vanishing clipping bias, while a time-varying quantile and stepsize
schedule recovers optimal nonconvex stationarity rates. The private analysis
shows how the choice of quantile trades clipping bias against privacy noise.

\item \textbf{Smoothed normalization and \algname{$\alpha$-NormEC}
(Chapter~\ref{chapter5:smoothed_normalization}).}
We show that bounded client transformations can change the aggregate descent
direction in distributed optimization. We combine smoothed normalization with
error compensation and obtain nonconvex convergence guarantees, including
differentially private guarantees under standard smoothness assumptions without
assuming uniformly bounded gradients.

\item \textbf{\algname{Gluon}
(Chapter~\ref{chapter6:gluon}).}
We introduce layer-wise $(L_i^0,L_i^1)$-smoothness and validate the model using
trajectory measurements from neural-network training. We then develop a
layer-wise LMO framework whose deterministic and stochastic guarantees depend
on the collection of layer-wise smoothness constants rather than on one
worst-case global constant.

\item \textbf{Inexact \algname{Muon}/LMO updates
(Chapter~\ref{chapter7:beyond_the_ideal}).}
We formulate an additive inexact-LMO model for practical approximate
orthogonalization and quantify how the approximation error degrades convergence.
Less precise orthogonalization calls for smaller stepsizes and, in the stochastic
setting, a larger momentum parameter. The experiments show that orthogonalization
precision also changes learning-rate stability and hyperparameter sensitivity.

\end{enumerate}

Chapters~\ref{chapter2:ist}--\ref{chapter7:beyond_the_ideal} present these contributions in three pairs: submodel training (Chapters~\ref{chapter2:ist} and~\ref{chapter3:mast}), private optimization (Chapters~\ref{chapter4:adaptive_clipping} and~\ref{chapter5:smoothed_normalization}), and LMO-based optimizers (Chapters~\ref{chapter6:gluon} and~\ref{chapter7:beyond_the_ideal}). Each chapter keeps the notation of its source paper. Its proofs, auxiliary results, and additional experiments are in Appendices~\ref{appendix:chapter2_ist}--\ref{appendix:chapter7_beyond_the_ideal}, and Appendix~\ref{appendix:common_facts} collects facts used across chapters. Chapter~\ref{chapter8:conclusion} concludes the thesis and discusses future research directions.

Taken together, the chapters support a simple conclusion:

\begin{center}
\setlength{\fboxsep}{10pt}
\fcolorbox{thesislink!45}{thesislink!4}{%
\begin{minipage}{0.88\textwidth}
\setstretch{1.25}
\emph{Practical optimization methods are often better understood by
modeling the objective, assumptions, update, and computation that define
the practical procedure, rather than forcing that procedure into a simplified model.}
\end{minipage}%
}
\end{center}

\chapter{Independent Subnetwork Training: Biased Submodel Update}
\label{chapter2:ist}

\begingroup
\input{KAUST_Thesis/imported/02_ist/import_setup}

\section*{Abstract}
Modern advancements in large-scale machine learning would be impossible without the paradigm of data-parallel distributed computing. Since distributed computing with large-scale models places excessive pressure on communication channels, significant recent research has been directed toward co-designing communication compression strategies and training algorithms with the goal of reducing communication costs. While pure data parallelism supports scaling to larger datasets, it has poor model-scaling properties. Indeed, compute nodes are severely limited by memory constraints, preventing further increases in model size. For this reason, the latest achievements in training giant neural network models also rely on some form of model parallelism. In this work, we take a closer theoretical look at Independent Subnetwork Training (IST), which is a recently proposed and highly effective technique for solving the aforementioned problems. We identify fundamental differences between IST and alternative approaches, such as distributed methods with compressed communication, and provide a precise analysis of its optimization performance on a quadratic model.

\begin{quote}
\small
\noindent\textit{This chapter is based on:}\\
Egor Shulgin and Peter Richt{\'a}rik, ``Towards a Better Theoretical Understanding of Independent Subnetwork Training,'' \emph{International Conference on Machine Learning (ICML)}, 2024.
\nocite{shulgin2023towards}
\end{quote}

\section{Introduction}

\begin{algorithm*}[h]%
\begin{algorithmic}[1]
\caption{Distributed Submodel (Stochastic) Gradient Descent} \label{ist:alg:IST}
    \STATE \textbf{Parameters:} learning rate $\gamma>0$; sketches $\mC_1, \ldots, \mC_n$; initial model $x^0 \in \bR^d$
    \FOR{$k = 0, 1, 2 \ldots$}
    \STATE Select submodels $w_i^k = \mC_i^k x^k$ for $i \in [n]$ and broadcast to all computing nodes
    \FOR{$i = 1, \ldots, n$ in parallel}
    \STATE Compute local (stochastic) gradient w.r.t. submodel: $\mC_i^k \nabla f_i(w_i^k)$
    \STATE Take (maybe multiple) gradient descent step $z_i^+ = w_i^k - \gamma \mC_i^k \nabla f_i(w_i^k)$
    \STATE Send $z_i^+$ to the server
    \ENDFOR
    \STATE Aggregate/merge received submodels: $x^{k+1} = \frac{1}{n} \sum_{i=1}^n z_i^+$
    \ENDFOR
\end{algorithmic}
\end{algorithm*}

A huge part of today's machine learning success is driven by the possibility of building more and more complex models and training them on increasingly larger datasets. This rapid progress has become feasible due to advancements in distributed optimization, which is necessary for proper scaling when the size of the training data grows~\citep{zinkevich2010parallelized}. In a typical scenario, data parallelism is used for efficiency and implies sharding the dataset across computing devices. This allowed very efficient scaling and acceleration of training moderately sized models by using additional hardware \citep{goyal2018accurate}. However, this data-parallel approach can suffer from a communication bottleneck, which has sparked extensive research on distributed optimization with compressed communication between nodes \citep{alistarh2017qsgd,konecny2017federated,Seide2014}.

\subsection{The need for model parallelism}
Despite its efficiency, data parallelism has some fundamental limitations when it comes to scaling up the model size. As the dimensions of a model increase, the amount of memory required to store and update the parameters also increases, which becomes problematic due to resource constraints on individual devices. This has led to the development of model parallelism \citep{dean2012large,Hydra}, which splits a large model across multiple nodes, with each node responsible for computations of parts of the model \citep{farber1997parallel,zhang1989efficient}.
However, naive model parallelism also poses challenges because each node can only update its portion of the model based on the data it has access to. This creates a need for very careful management of communication between devices. Thus, a combination of both data and model parallelism is often necessary to achieve efficient and scalable training of huge models.

\textbf{IST.}
Independent Subnetwork Training (IST) is a technique that suggests dividing a neural network into smaller subparts, training them in a distributed parallel fashion, and then aggregating the results to update the weights of the whole model. In IST, every subnetwork can operate independently and has fewer parameters than the full model, which not only reduces the load on computing nodes but also results in faster synchronization. A generalized analog of the described method is formalized as an iterative procedure in Algorithm \ref{ist:alg:IST}. %
The IST paradigm was pioneered by \citet{yuan2022distributed} for networks with fully connected layers and was later extended to ResNets \citep{dun2022resist} and graph architectures \citep{wolfe2021gist}. Previous experimental studies have shown that IST is a promising approach for various applications because it allows practitioners to combine data and model parallelism and train larger models with limited compute.
In addition, \citet{liao2022on} performed a theoretical analysis of IST for overparameterized single-hidden-layer neural networks with ReLU activations. The idea of IST was also extended to the federated setting via an asynchronous distributed dropout technique \citep{dun2023efficient}.

\textbf{Federated Learning.} Another important setting in which data are distributed, often for privacy reasons, is federated learning \citep{kairouz2021advances,konecny2017federated,mcmahan2017communication}. In this scenario, computing devices (e.g., mobile phones) are often heterogeneous and more resource-constrained than devices in data-center settings \citep{caldas2018expanding}. Such challenges have prompted extensive research efforts into selecting smaller and more efficient submodels for local on-device training \citep{alam2022fedrolex,charles2022federated,chen2022fedobd,diao2020heterofl,horvath2021fjord,jiang2022model,lin2022federated,qiu2022zerofl,wen2022federated,yang2022partial}.
Many of these works propose approaches to adapt submodels, often tailored to specific neural network architectures, based on the capabilities of individual clients for various machine learning tasks.
However, the theoretical properties of these methods remain poorly understood.

\subsection{Summary of contributions}
After reviewing the literature, we identified a glaring gap in the rigorous understanding of IST convergence, directly motivating our research. The main contributions of this chapter include:
\begin{itemize}
    \item A novel approach to analyzing distributed methods that combine data and model parallelism by operating with sparse submodels for a quadratic model.

    \item The first analysis of independent subnetwork training in homogeneous and heterogeneous scenarios without restrictive assumptions on gradient estimators.

    \item Identification of settings in which IST optimizes efficiently and settings in which it converges only to a tightly characterized irreducible neighborhood of the optimal solution.

    \item Empirical validation of the proposed theory through experiments for several practical settings.
\end{itemize}

\section{Formalism and setup} \label{ist:sec:setup}

We consider the standard optimization formulation of a distributed/federated learning problem \citep{wang2021field}
\begin{equation} \label{ist:eq:general_problem}
  \min \limits_{x \in \bR^d}~\sbr{f(x) \eqdef \frac{1}{n} \sum \limits_{i=1}^n f_i(x)},
\end{equation}
where $n$ is the number of clients/workers, and each $f_i: \bR^d \to \bR$ represents the loss of the model parameterized by vector $x \in \bR^d$ on the data of client $i$.

A typical Stochastic Gradient Descent (SGD)-type method for solving this problem has the form
\begin{equation} \label{ist:eq:SGD_generic}
  x^{k+1} = x^k - \gamma g^k, \qquad g^k = \frac{1}{n}\sum \limits_{i=1}^n g_i^k,
\end{equation}
where $\gamma>0$ is the stepsize and $g_i^k$ is a suitably constructed estimator of $\nabla f_i(x^k)$. In the distributed setting, computation of gradient estimators $g_i^k$ is typically performed by clients, and the results are sent to the server, which subsequently performs aggregation via averaging $g^k = \frac{1}{n}\sum_{i=1}^n g_i^k$. The average is then used to update the model $x^{k+1}$ via a gradient-type method \eqref{ist:eq:SGD_generic}, and at the next iteration, the model is broadcasted back to the clients. The process is repeated iteratively until a suitable model is found.

One of the main techniques used to accelerate distributed training is lossy \textit{communication compression} \citep{alistarh2017qsgd,konecny2017federated,Seide2014},
which suggests applying a (possibly randomized) lossy compression mapping $\cC$ to a vector/matrix/tensor $x$ before broadcasting. This reduces the bits sent per communication round at the cost of transmitting a less accurate estimate $\cC(x)$ of $x$.
This technique can be formalized in the following definition.

\begin{definition}[Unbiased compressor] \label{ist:def:unbiased_compressor}
A randomized mapping $\cC: \bR^d \to \bR^d$ is an {\bf unbiased compression operator} ($\cC \in \bU(\omega)$ for brevity) if for some $\omega \geq 0$ and $\forall x \in \bR^d$
\begin{equation} \label{ist:eq:unbiased_compressor}
    \Exp{\cC(x)} = x, \qquad \Exp{\norm{\cC(x) - x}^2} \leq \omega \norm{x}^2.
\end{equation}
\end{definition}
A notable example of a mapping from this class is the \textit{random sparsification} (\texttt{Rand-q} for $q \in \{1, \dots, d\}$) operator defined by
\begin{equation} \label{ist:eq:rand-k}
    \cC_{\texttt{Rand-q}}(x) \eqdef \mC_q x = \frac{d}{q} \sum\limits_{i\in S} e_i e_i^\top x ,
\end{equation}
where $e_1,\dots,e_d \in\bR^d$ are standard unit basis vectors, and $S$ is a random subset of $[d] \eqdef \{1, \dots, d\}$ sampled from the uniform distribution on the all subsets of $[d]$ with cardinality $q$. \texttt{Rand-q} belongs to $\bU\br{d/q-1}$, which means that the more elements are \say{dropped} (lower $q$), the higher the variance $\omega$ of the compressor.

In this work, we are mainly interested in a somewhat more general class of operators than mere sparsifiers. In particular, we are interested in compressing via the application of random matrices, i.e., via {\em sketching}. A sketch $\mC_i^k \in \R^{d\times d}$ can be used to represent submodel computations in the following way:
\begin{equation*} %
    g_i^k \eqdef \mC_i^k \nabla f_i(\mC_i^k x^k),
\end{equation*}
where we require $\mC_i^k$ to be a symmetric positive semi-definite matrix. Such gradient estimates correspond to computing the local gradient with respect to a sparse submodel $\mC_i^k x^k$, and additionally sketching the resulting gradient with the same matrix $\mC_i^k$ to guarantee that the resulting update lies in the lower-dimensional subspace.

Using this notion, Algorithm \ref{ist:alg:IST} (with one local gradient step) can be represented in the following form:
\begin{equation} \label{ist:eq:IST}
    x^{k+1} = \frac{1}{n} \sum \limits_{i=1}^n \sbr{\mC_i^k x^k - \gamma \mC_i^k \nabla f_i(\mC_i^k x^k)},
\end{equation}
which is equivalent to the SGD-type update \eqref{ist:eq:SGD_generic} when the
\emph{perfect reconstruction} property holds  (with probability one)
$$\mC^k \eqdef \frac{1}{n}\sum \limits_{i=1}^n \mC_i^k = \mathbf{I},$$ where $\mathbf{I}$ is the identity matrix.
This property is inherent for a specific class of compressors that are particularly useful for capturing the concept of an \textit{independent} subnetwork partition.

\begin{definition}[Permutation sketch] \label{ist:def:permutation}
Assume that model size is greater than the number of clients $d \geq n$ and $d=q n$, where $q \geq 1$ is an integer\footnote{While this condition may look restrictive, it naturally holds for distributed learning in a data-center setting. Permutation sparsifiers were introduced by \citet{szlendak2022permutation} and generalized to other scenarios (like $n \geq d$).}.
Let $\pi=(\pi_1, \ldots, \pi_d)$ be a random permutation of $[d]$. Then for all $i \in [n]$, we define $\texttt{Perm-q}$ operator
\begin{equation} \label{ist:eq:permutation}
    \mC_i \eqdef n \cdot \sum \limits_{j=q(i-1)+1}^{q i} e_{\pi_j} e_{\pi_j}^\top.
\end{equation}
\end{definition}
$\texttt{Perm-q}$ is unbiased and can be conveniently used for representing a structured decomposition of the model, such that every client $i$ is responsible for computations over a submodel $\mC_i x^k$.

Our convergence analysis relies on the assumption that was previously used for coordinate descent-type methods.
\begin{assumption}[Matrix smoothness] \label{ist:ass:matrix-smoothness}
 A differentiable function $f: \bR^d \to \bR$ is $\mathbf{L}$-smooth, if there exists a positive semi-definite matrix $\mL \in \bR^{d \times d}$ such that $\forall x, h \in \Rd$
\begin{equation}\label{ist:eq:L-matrix-smooth}
   f(x + h) \leq f(x) + \left\langle\nabla f(x), h \right\rangle + \frac{1}{2} \left\langle \mL h, h \right\rangle.
\end{equation}
\end{assumption}
A standard $L$-smoothness condition is obtained as a special case of \eqref{ist:eq:L-matrix-smooth} for $\mL = L \cdot \mI$.
Matrix smoothness was previously used for designing data-dependent gradient sparsification to accelerate optimization in communication-constrained settings \citep{safaryan2021smoothness,wang2022theoretically}.

\subsection{Issues with existing approaches}

Consider the simplest gradient descent method with a compressed model in the single-node setting:
\begin{equation} \label{ist:eq:CMGD}
    x^{k+1} = x^k - \gamma \nabla f (\cC(x^k)).
\end{equation}

Algorithms belonging to this family require a different analysis in comparison to SGD \citep{gorbunov2020unified,gower2019sgd}, Distributed Compressed Gradient Descent \citep{alistarh2017qsgd,khirirat2018distributed}, and Randomized Coordinate Descent \citep{nesterov2012efficiency,richtarik2014iteration}-type methods because the gradient estimator is no longer unbiased
\begin{equation*}
    \Exp{\nabla f (\cC(x))} \neq \nabla f(x) = \Exp{\cC(\nabla f(x))}.
\end{equation*}

This is why algorithms of this kind, such as \eqref{ist:eq:CMGD}, are harder to analyze. Therefore, prior results for \textit{unbiased} SGD \citep{khaled2020better} cannot be directly reused.
Furthermore, the nature of the bias in this type of gradient estimator does not exhibit additive noise, thereby preventing the application of previous analyses for biased SGD \citep{ajalloeian2020convergence}.

An assumption like the bounded stochastic gradient norm extensively used in previous works \citep{lin2019dynamic,zhou2022convergence} hinders an accurate understanding of such methods. This assumption hides the fundamental difficulty of analyzing a biased gradient estimator:
\begin{equation} \label{ist:eq:bounded_gradient}
    \Exp{\sqN{\nabla f(\cC(x))}} \leq G
\end{equation}
and may not hold, even for quadratic functions $f(x) = x^\top \mA x$. In addition, in the distributed setting, such a condition can result in vacuous bounds \citep{khaled2020tighter} as it does not capture heterogeneity accurately.

\subsection{Simplifying assumptions}

To conduct a thorough theoretical analysis of methods that combine data with model parallelism, we simplify the algorithm and problem setting to isolate the unique effects of this approach. The following considerations are made:

\begin{enumerate}
    \item[\textbf{(a)}] We assume that every node $i$ computes the true gradient at the submodel $\mC_i \nabla f_i(\mC_i x^k)$.
    \item[\textbf{(b)}] A notable difference compared to the original IST Algorithm \ref{ist:alg:IST} is that workers perform a single gradient descent step (or just gradient computation).
    \item[\textbf{(c)}] Finally, we consider a special case of a quadratic model \eqref{ist:eq:het_gen_quad_problem} as a loss function \eqref{ist:eq:general_problem}.
\end{enumerate}

Condition \textbf{(a)} is mainly for the sake of simplicity and clarity of exposition and can be generalized to stochastic gradient estimators with bounded variance.
Condition \textbf{(b)} is imposed because local steps did not bring any theoretical efficiency improvements for heterogeneous settings until very recently \citep{mishchenko2022proxskip}, and even then, only with the introduction of additional control variables, which goes against the requirements of resource-constrained device settings. The reason behind \textbf{(c)} is that despite its apparent simplicity, the quadratic problem has been used extensively to study properties of neural networks \citep{zhang2019algorithmic,zhu2022quadratic}. Moreover, it is a non-trivial model, which makes it possible to understand complex optimization algorithms \citep{arjevani2020tight,cunha2022only,goujaud2022super}.
The quadratic problem is suitable for observing complex phenomena and providing theoretical insights, which can also be observed in practical scenarios. Finally, Appendix \ref{ist:sec:generalization} presents a generalization for smooth functions.

Having said that, we consider a special case of problem \eqref{ist:eq:general_problem} for symmetric matrices $\mL_i$
\begin{equation} \label{ist:eq:het_gen_quad_problem}
      f(x) = \frac{1}{n} \sum \limits_{i=1}^n f_i(x), \quad f_i(x) \equiv \frac{1}{2} x^\top \mL_i x - x^\top \bb_i.
\end{equation}

In this case, $f(x)$ is $\moL$-smooth, and $\nabla f(x) = \moL x - \ob$, where $\moL = \frac{1}{n} \sum_{i=1}^n \mL_i$ and $\ob \eqdef \frac{1}{n} \sum_{i=1}^n \bb_i$.

\section{Results in the interpolation case}

First, let us examine the case of $\bb_i \equiv 0$, which we call interpolation for quadratics, and perform the analysis for general sketches $\mC_i^k$.
In this case, the gradient estimator \eqref{ist:eq:SGD_generic} takes the form
\begin{equation} \label{ist:eq:het_estimator}
   g^k = \frac{1}{n} \sum \limits_{i=1}^n \mC_i^k \nabla f_i(\mC_i^k x^k) %
   = \moB^k x^k,
\end{equation}
where $\moB^k \eqdef \frac{1}{n}\sum_{i=1}^n  \mC_i^k \mL_i \mC_i^k$.
We prove the following result for a method with such an estimator.

\begin{theorem} \label{ist:thm:het_quad}
Consider the method \eqref{ist:eq:SGD_generic} with estimator \eqref{ist:eq:het_estimator} for a quadratic problem \eqref{ist:eq:het_gen_quad_problem} with $\moL \succ 0$ and $\bb_i \equiv 0$. Then if $\moW \eqdef \frac{1}{2} \Exp{\moL \moB^k + \moB^k \moL} \succeq 0$ and there exists a constant $\theta > 0$:
\begin{equation} \label{ist:eq:het_exp_sep}
    \Exp{\moB^k \moL \moB^k} \preceq \theta \moW,
\end{equation}
and the step size is chosen as $0 < \gamma \leq \frac{1}{\theta}$, the iterates satisfy
\begin{equation} \label{ist:eq:het_res}
    \frac{1}{K} \sum \limits_{k=0}^{K-1} \Exp{\sqN{\nabla f(x^k)}_{\moL^{-1} \moW \moL^{-1}}}
    \leq
    \frac{2 \delta^K}{\gamma K},
\end{equation}
for $\delta^K \eqdef f(x^0) - \Exp{f(x^{K})}$ and
\begin{equation} \label{ist:eq:het_res_it}
    \Exp{\sqn{x^k - x^\star}_{\moL}} \leq
    \br{1 - \gamma \Lambda}^k \sqn{x^0- x^\star}_{\moL},
\end{equation}
where $\Lambda \eqdef \lambda_{\min} \br{\moL^{-\frac{1}{2}}\moW\moL^{-\frac{1}{2}}}$.
\end{theorem}

This theorem establishes an $\cO(1/K)$ convergence rate with a constant step size up to a stationary point \eqref{ist:eq:het_res} and linear convergence \eqref{ist:eq:het_res_it} for the expected distance to the optimum $x^\star \eqdef \argmin f(x)$.
Note that we employ weighted norms in our analysis, as the considered class of loss functions satisfies the matrix $\moL$-smoothness Assumption \ref{ist:ass:matrix-smoothness}.
The use of standard Euclidean distance may result in loose bounds that do not recover correct rates for special cases like gradient descent.

It is important to highlight that the inequality \eqref{ist:eq:het_exp_sep} may not hold (for any $\theta > 0$) in the general case as the matrix $\moW$ is not guaranteed to be positive (semi-)definite in the case of general sampling.
The intuition behind this issue is that arbitrary sketches $\mC_i^k$ can result in the gradient estimator $g^k$, which is misaligned with the true gradient $\nabla f(x^k)$. Specifically, the inner product $\left\langle\nabla f(x^k), g^k \right\rangle$ can be negative, and there is no expected descent after one step.

Next, we give examples of samplings for which the inequality \eqref{ist:eq:het_exp_sep} can be satisfied.

\textbf{1.} {\bf Identity.} Consider $\mC_i\equiv \mI$. Then $\moB^k = \moL$, $\moB^k\moL\moB^k = \moL^3, \moW = \moL^2 \succ 0$ and hence \eqref{ist:eq:het_exp_sep} is satisfied for $\theta = \lambda_{\max}(\moL)$. So, \eqref{ist:eq:het_res} says that if we choose $\gamma = \nicefrac{1}{\theta}$, then
\[
    \frac{1}{K} \sum \limits_{k=0}^{K-1} \sqN{\nabla f(x^k)}_{\mI }
    \leq
    \frac{2 \lambda_{\max}(\moL)\br{f(x^0) - f(x^{K})}}{K},
\]
which exactly matches the rate of gradient descent in the non-convex setting.
As for convergence of the iterates,
the rate in \eqref{ist:eq:het_res_it} is $\nicefrac{\lambda_{\max}(\moL)}{\lambda_{\min}(\moL)}$ which corresponds to the precise gradient descent result for strongly convex functions.

\textbf{2.} {\bf Permutation.} Assume\footnote{This is mainly done to simplify the presentation. Results can be generalized to the case of $n \neq d$ in a similar manner as in \citet{szlendak2022permutation}; the generalization is given in Appendix~\ref{appendix:chapter2_ist}.} $n=d$ and the use of \texttt{Perm-1} (special case of Definition \ref{ist:def:permutation}) sketch $\mC^k_i = n e_{\pi^k_i} e_{\pi^k_i}^\top$, where $\pi^k = (\pi^k_1, \dots, \pi^k_n)$ is a random permutation of $[n]$. Then
\begin{equation*}
    \Exp{\moB^k} = \frac{1}{n} \sum \limits_{i=1}^n \Exp{\mC_i^k \mL_i \mC_i^k} = \frac{1}{n} \sum \limits_{i=1}^n n \Diag(\mL_i) = n \moD
\end{equation*}

where $\moD \eqdef \frac{1}{n} \sum_{i=1}^n \mD_i, \mD_i \eqdef \Diag(\mL_i)$. Then inequality \eqref{ist:eq:het_exp_sep} leads to
\begin{equation*} %
   n \moD \moL \moD \preceq \frac{\theta}{2} \br{\moL \moD + \moD \moL},
\end{equation*}
which may not always hold as  $\moL \moD + \moD \moL$ is not guaranteed to be positive-definite---even in the case of $\moL \succ 0$. However, such a condition can be enforced via a slight modification of the permutation sketches, which is done in Section \ref{ist:sec:precond_het_quad}. The limitation of such an approach is that the resulting compressors are no longer unbiased.

\begin{remark}
Matrix $\moW$ in the case of permutation sketches may not be positive-definite. Consider the following example of a homogeneous ($\mL_i \equiv \mL$) two-dimensional problem:
\begin{equation*}
\mL =
    \left[\begin{array}{cccc}
        a & c \\
        c & b
    \end{array}\right].
\end{equation*}
Then $\moW$ equals to
\begin{equation*}
    \frac{1}{2} \sbr{\moL \moD + \moD \moL} =
    \left[\begin{array}{cccc}
    a^2 & c(a+b)/2 \\
    c(a+b)/2 & b^2
    \end{array}\right],
\end{equation*}
which for $c > \frac{2 a b}{a + b}$ has $\mathrm{det}(\moW) < 0$, and thus $\moW\nsucc 0$ according to Sylvester's criterion.
\end{remark}

Next, we focus on the particular case of \textbf{permutation} sketches, which are the most suitable for model partitioning according to Independent Subnetwork Training (IST). In the rest of this section, we discuss how the condition \eqref{ist:eq:het_exp_sep} can be enforced via a specially designed preconditioning of the problem \eqref{ist:eq:het_gen_quad_problem} or modification of the sketch mechanism \eqref{ist:eq:permutation}.

\subsection{Homogeneous problem preconditioning}
\label{ist:sec:precond_homo_quad}
To start, consider a homogeneous setting $f_i(x) = \frac{1}{2} x^\top \mL x$, so $\mL_i \equiv \mL$.
Now define $\mD = \Diag (\mL)$ -- a diagonal matrix with elements equal to the diagonal of $\mL$. Then, the problem can be converted to
\begin{equation*} %
    f_i(\mD^{-\frac{1}{2}}x) = \frac{1}{2} \br{\mD^{-\frac{1}{2}}x}^\top \mL \br{\mD^{-\frac{1}{2}} x} =
    \frac{1}{2} x^\top \mcL x,
\end{equation*}
where $\mcL \eqdef \mD^{-\frac{1}{2}} \mL \mD^{-\frac{1}{2}}$.
It is equivalent to the original problem after changing the variables $\tilde{x} \eqdef \mD^{-\frac{1}{2}}x$. Note that $\mD = \Diag (\mL)$ is positive-definite as $\mL \succ 0$, and therefore $\mcL \succ 0$. Moreover, the preconditioned matrix $\mcL$ has all ones on the diagonal: $\Diag(\mcL) = \mI$. If we now combine reformulated problem %
with \texttt{Perm-1} sketches
\begin{equation*}
   \Exp{\moB^k} = \Exp{\frac{1}{n} \sum_{i=1}^n \bfC_i \mcL \bfC_i} = n \Diag(\mcL) = n \mI.
\end{equation*}
Therefore, inequality \eqref{ist:eq:het_exp_sep} takes the form $\tilde{\mW} = n \mcL \succeq \frac{1}{\theta} n^2 \mcL$, which holds for $\theta \geq n$, and the left-hand side of \eqref{ist:eq:het_res} can be transformed (for an accurate comparison to standard methods) in the following way:
\begin{equation*}
\begin{aligned}
    \sqN{\nabla f(x^k)}_{\mcL^{-1} \tilde{\mW} \mcL^{-1}}
    &\geq n \lambda_{\min} \br{\mcL^{-1}} \sqN{\nabla f(x^k)}_{\mI}
    \\&=
    n \lambda_{\max} (\mcL) \sqN{\nabla f(x^k)}_{\mI}
\end{aligned}
\end{equation*}
The resulting convergence guarantee is
\begin{equation*}
    \frac{1}{K} \sum \limits_{k=0}^{K-1} \Exp{\sqN{\nabla f(x^k)}_{\mI}}
    \leq
    \frac{2\lambda_{\max} (\mcL) \delta^K}{K},
\end{equation*}
which matches classical gradient descent.

\subsection{Heterogeneous sketch preconditioning} \label{ist:sec:precond_het_quad}
In contrast to the homogeneous case, the heterogeneous problem $f_i(x) = \frac{1}{2} x^\top \mL_i x$ cannot be so easily preconditioned by a simple change of variables $\tilde{x} \eqdef \mD^{-\frac{1}{2}}x$, as every client $i$ has its own matrix $\mL_i$. However, this problem can be fixed via the following modification of \texttt{Perm-1}, which scales the output according to the diagonal elements of the local smoothness matrix $\mL_i$:
\begin{equation} \label{ist:eq:imp_perm}
    \tilde{\bfC}_i \eqdef \sqrt{n/\sbr{\mL_i}_{\pi_i, \pi_i}} e_{\pi_i} e_{\pi_i}^\top.
\end{equation}
In this case,  $\Exp{\tilde{\mC}_i \mL_i \tilde{\mC}_i} = \mI$,  $\Exp{\moB^k} = \mI$, and $\moW = \moL$.
Then inequality \eqref{ist:eq:het_exp_sep} is satisfied for $\theta \geq 1$.

If one inputs these results into \eqref{ist:eq:het_res}, such convergence guarantee can be obtained
\begin{equation*}
    \frac{1}{K} \sum \limits_{k=0}^{K-1} \Exp{\sqN{\nabla f(x^k)}_{\mI}}
    \leq
    \frac{2 \lambda_{\max} (\moL) \delta^K}{K},
\end{equation*}
which matches the gradient descent result as well. Thus, we can conclude that heterogeneity does not bring such a fundamental challenge in this scenario. In addition, the method with \texttt{Perm-1} is significantly better in terms of computational and communication complexity, as it requires calculation of the local gradients with respect to much smaller submodels and transmits only sparse updates.

This construction also shows that for $\gamma = 1/\theta = 1$
\begin{equation*}
    \gamma \lambda_{\min} \br{\moL^{-\frac{1}{2}}\moW\moL^{-\frac{1}{2}}} %
    = 1,
\end{equation*}
which, after plugging into the bound for the iterates \eqref{ist:eq:het_res_it}, shows that the method basically converges in one iteration. This observation indicates that sketch preconditioning can be extremely efficient, although it uses only the diagonal elements of matrices $\mL_i$.

Now that we understand that the method can perform very well in the special case of $\cb_i \equiv 0$, we can move on to a more complicated situation.

\section{Irreducible bias in the general case}

Now we look at the most general heterogeneous case with different matrices and linear terms
$f_i(x) \equiv \frac{1}{2} x^\top \mL_i x - x^\top \bb_i.$
In this instance, the gradient estimator \eqref{ist:eq:SGD_generic} takes the form
\begin{equation} \label{ist:eq:het_gen_quad_est}
    g^k = %
    \frac{1}{n} \sum \limits_{i=1}^n \mC_i^k \br{\mL_i \mC_i^k x^k - \mathrm{b}_i} = \moB^k x^k - \oCb,
\end{equation}
where $\oCb = \frac{1}{n}\sum_{i=1}^n \mC_i^k \bb_i$. Herewith let us use a heterogeneous permutation sketch preconditioner \eqref{ist:eq:imp_perm}, as in Section \ref{ist:sec:precond_het_quad}.
Then $\Exp{\moB^k} = \mI$ and $\Exp{\oCb} = \frac{1}{\sqrt{n}} \cDb$, where $\cDb \eqdef \frac{1}{n} \sum_{i=1}^n \mD_i^{-\frac{1}{2}} \bb_i$.
Furthermore, the expected gradient estimator \eqref{ist:eq:het_gen_quad_est} results in $\Exp{g^k} = x^k - \frac{1}{\sqrt{n}} \cDb$ and can be transformed in the following manner:
\begin{equation} \label{ist:eq:het_gen_estimator}
\begin{aligned}
    \squeeze
    \Exp{g^k}
    &=
    \moL^{-1} \moL x^k \pm \moL^{-1} \ob - \frac{1}{\sqrt{n}} \cDb
    \\&=
    \moL^{-1} \nabla f(x^k) + \underbrace{\moL^{-1}\ob - \frac{1}{\sqrt{n}}\cDb}_{h},
\end{aligned}
\end{equation}
which reflects the decomposition of the estimator into the optimally preconditioned true gradient and a bias, depending on the linear terms $\bb_i$.

\subsection{Bias of the method} \label{ist:sec:het_method_bias}

Estimator \eqref{ist:eq:het_gen_estimator} can be directly plugged (with proper conditioning) into the general SGD update \eqref{ist:eq:SGD_generic}
\begin{equation}\label{ist:eq:het_iterates}
    \Exp{x^{k+1}} = (1 - \gamma) x^k + \frac{\gamma}{\sqrt{n}} \cDb
    = \br{1 - \gamma}^{k+1} x^0 + \frac{\gamma}{\sqrt{n}} \cDb \sum \limits_{j=0}^k (1 - \gamma)^j.
\end{equation}

The resulting recursion \eqref{ist:eq:het_iterates} is exact, and its asymptotic limit can be analyzed.
Thus, for constant $\gamma < 1$, by using the formula for the sum of the first $k$ terms of a geometric series, one gets
\begin{eqnarray*}
    \Exp{x^k} =
    \br{1 - \gamma}^k x^0 + \frac{1-(1 - \gamma)^k}{\sqrt{n}} \cDb \underset{k \to \infty}{\longrightarrow} \frac{1}{\sqrt{n}} \cDb,
\end{eqnarray*}
which shows that in the limit, the first initialization term (with $x^0$) vanishes while the second converges to $\frac{1}{\sqrt{n}} \cDb$.
This reasoning shows that the method does not converge to the exact solution
$$\Exp{x^k} \to x^\infty \neq x^\star
\in \argmin\limits_{x\in \bR^d} \cbr{\frac{1}{2} x^\top \moL x - x^\top \ob }, $$
which for the positive-definite $\moL$ can be defined as
\[
x^\star = \moL^{-1} \ob,
\qquad
x^\infty = \frac{1}{n\sqrt{n}} \sum_{i=1}^n \mD_i^{-\frac{1}{2}} \bb_i.
\]
So, in general, there is an unavoidable bias. However, in the limit case: $n = d \to \infty$, the bias diminishes.

\subsection{Generic convergence analysis} \label{ist:sec:gen_convergence}
While the analysis in Section \ref{ist:sec:het_method_bias} is precise, it does not allow us to compare the convergence of IST to standard optimization methods. Therefore, we also analyze the non-asymptotic behavior of the method to understand the convergence speed. Our result is formalized in the following theorem:

\begin{theorem} \label{ist:thm:het_gen_gradient}
\emergencystretch=2em
Consider the method \eqref{ist:eq:SGD_generic} with the estimator \eqref{ist:eq:het_gen_quad_est} for the quadratic problem \eqref{ist:eq:het_gen_quad_problem} with the positive-definite matrix $\moL \succ 0$. Assume that for every $\mD_i \eqdef \Diag(\mL_i)$ matrices $\mD_i^{-\frac{1}{2}}$ exist, scaled permutation sketches \eqref{ist:eq:imp_perm}
are used, and heterogeneity is bounded as $\Exp{\sqN{g^k - \Exp{g^k}}_{\moL}} \leq \sigma^2$. Then, for the step size chosen as follows:
\begin{equation*}
    0 < \gamma \leq \gamma_{\beta} \eqdef \frac{1/2 - \beta}{\beta + 1/2},
\end{equation*}
where $\gamma_{c, \beta} \in (0, 1]$ for $\beta \in (0, 1/2)$, the iterates satisfy
\begin{equation} \label{ist:eq:het_gen_grad_res}
\frac{1}{K} \sum \limits_{k=0}^{K-1} \Exp{\sqN{\nabla f(x^k)}_{\moL^{-1}}}
    \leq
    \frac{f(x^0) - \Exp{f(x^K)}}{0.5\gamma K} + \br{\frac{1 - \gamma}{0.5\beta} + \gamma} \sqn{h}_{\moL} + \gamma \sigma^2,
\end{equation}
where $\moL = \frac{1}{n} \sum_{i=1}^n \mL_i, h = \moL^{-1}\ob - \frac{1}{n^{3/2}} \sum_{i=1}^n \mD_i^{-\frac{1}{2}} \bb_i$ and $\ob = \frac{1}{n} \sum_{i=1}^n \bb_i$.
\end{theorem}

Note that the derived convergence upper bound has a neighborhood proportional to the bias of the gradient estimator $h$ and level of heterogeneity $\sigma^2$. Some of these terms with factor $\gamma$ can be eliminated by decreasing the learning rate (e.g., $\sim 1/\sqrt{K}$). However, such a strategy does not diminish the term with a multiplier $2 \beta^{-1} \br{1 - \gamma}$, making the neighborhood irreducible. Moreover, this term can be eliminated for $\gamma = 1$, which also minimizes the first term that decreases as $1/K$. However, this step size choice maximizes the terms with factor $\gamma$.
Thus, there exists an inherent trade-off between convergence speed and the size of the neighborhood.

In addition, convergence to the stationary point is measured by the weighted $\moL^{-1}$ squared norm of the gradient. At the same time, the neighborhood term depends on the weighted by $\moL$ norm of $h$. This fine-grained decoupling is achieved by carefully applying the Fenchel-Young inequality and provides a tighter characterization of the convergence compared to using standard Euclidean distances.

\paragraph{Homogeneous Case.} In this scenario, every worker has access to all data $f_i(x) \equiv \frac{1}{2} x^\top \mL x - x^\top \bb$. Then diagonal preconditioning of the problem can be used, as in the previous Section \ref{ist:sec:precond_homo_quad}. This results in a gradient
$\nabla f(x) = \mcL x - \cb$ for $\mcL = \mD^{-\frac{1}{2}} \mL \mD^{-\frac{1}{2}}$ and $\cb = \mD^{-\frac{1}{2}} \bb$. If this expression is further combined with a permutation sketch scaled by $1/\sqrt{n}$
$\mC_i' \eqdef \sqrt{n} e_{\pi_i} e^{\top}_{\pi_i}$,
the resulting gradient estimator is:
\begin{equation} \label{ist:eq:hom_gen_estimator}
    g^k = x^k - \frac{1}{\sqrt{n}} \cb
    =
    \mcL^{-1} \nabla f(x^k) + \tilde{h},
\end{equation}
for $\tilde{h} = \mcL^{-1}\cb - \frac{1}{\sqrt{n}}\cb$.
In this case, the heterogeneity term $\sigma^2$ from the upper bound \eqref{ist:eq:het_gen_grad_res} disappears as $\Exp{\sqN{g^k - \Exp{g^k}}_{\moL}} = 0$, which significantly decreases the neighborhood size. However, the bias term depending on $\tilde{h}$ still remains, as the method does not converge to the exact solution
$x^k \to x^\infty \neq x^\star = \mcL^{-1} \cb$ for positive-definite $\mcL$. Nevertheless the method's fixed point $x^\infty = \cb / \sqrt{n}$ and solution $x^\star$ can coincide when $\mcL^{-1} \cb = \frac{1}{\sqrt{n}} \cb$, which means that $\cb$ is the right eigenvector of matrix $\mcL^{-1}$ with eigenvalue $\frac{1}{\sqrt{n}}$.

Let us contrast the obtained result \eqref{ist:eq:het_gen_grad_res} with the non-convex rate of SGD \citep{khaled2020better} with constant step size $\gamma$ for $L$-smooth and lower-bounded $f$
\begin{equation} \label{ist:eq:SGD_rate}
    \min_{k<K-1} \Exp{\sqN{\nabla f(x^k)}} %
    \leq
    \frac{f(x^0) - \inf f}{0.1\gamma K} + \gamma L C,
\end{equation}
where constant $C$ depends, for example, on the variance of the stochastic gradient estimator.
Observe that the first term in the compared upper bounds \eqref{ist:eq:SGD_rate} and \eqref{ist:eq:het_gen_grad_res} is almost identical and decreases with speed $1/K$. However, unlike \eqref{ist:eq:het_gen_grad_res}, the neighborhood for SGD can be completely eliminated by reducing the step size $\gamma$. This highlights a fundamental difference between our results and unbiased methods.
The intuition behind this issue is that for SGD-type methods like compressed gradient descent
\begin{equation*}
    x^{k+1} = x^k - \cC(\nabla f(x^k))
\end{equation*}
the gradient estimate is unbiased and enjoys the property that variance
\begin{equation*}
    \Exp{ \sqn{\cC(\nabla f(x^k)) - \nabla f(x^k)}} \leq \omega \sqn{\nabla f(x^k)}
\end{equation*}
goes down to zero as the method progresses because $\nabla f(x^k) \to \nabla f(x^\star) = 0$ in the unconstrained case.
In addition, any stationary point $x^\star$ ceases to be a fixed point of the iterative procedure as
\begin{equation*}
    x^\star \neq x^\star - \nabla f(\cC(x^\star)),
\end{equation*}
in the general case, unlike for compressed gradient descent
with both biased and unbiased compressors $\cC$. Thus, even if the method---computing the gradient with a sparse model---is initialized from the \emph{solution} after one gradient step, the method may get away from the optimum.

\subsection{Comparison to previous works}

\paragraph{Independent Subnetwork Training \citep{yuan2022distributed}.} There are several improvements over the previous works that tried to theoretically analyze the convergence of distributed IST.

The first difference is that our results allow for an almost arbitrary level of model sparsification, i.e., will work for any $\omega \geq 0$ as permutation sketches can be viewed as a special case of compression operators \eqref{ist:def:unbiased_compressor}. This represents a significant improvement over the work of \citet{yuan2022distributed}, which demands\footnote{$\mu$ refers to a constant from the Polyak-Łojasiewicz (or strong convexity) condition. In case of a quadratic problem with positive-definite matrix $\mA$  constant $\mu$ equals to $\lambda_{\min} (\mA)$} $\omega \lesssim \nicefrac{\mu^2}{L^2}$. Such a requirement is very restrictive as the condition number $L/\mu$ of the loss function $f$ is typically very large for any non-trivial optimization problem. Thus, the sparsifier's \eqref{ist:eq:rand-k} variance $\omega = d/q-1$ has to be very close to 0 and $q \approx d$. Thus, the previous theory allows almost no compression (sparsification) because it is based on the analysis of gradient descent with compressed iterates \citep{khaled2019gradient}.

The second distinction is that the original IST work \citep{yuan2022distributed} considered a single node setting, and thus their convergence bounds did not capture the effect of heterogeneity, which we believe is of crucial importance for distributed settings \citep{chraibi2019distributed,shulgin2022shifted}.
Moreover, the original work considers the Lipschitz continuity of the loss function $f$, which is not satisfied for a simple quadratic model. A more detailed comparison, including additional assumptions on the gradient estimator made by \citet{yuan2022distributed}, is presented in the Appendix \ref{ist:sec:app_comparison}.

\paragraph{FL with Model Pruning.}
In a recent work, \citet{zhou2022convergence} made an attempt to analyze a variant of the FedAvg algorithm with sparse local initialization and compressed gradient training (pruned local models). They considered a case of $L$-smooth loss and a sparsification operator satisfying a similar condition to \eqref{ist:def:unbiased_compressor}. However, they also assumed that the squared norm of the stochastic gradient is uniformly bounded \eqref{ist:eq:bounded_gradient}, which is \say{pathological} \citep{khaled2020tighter}---especially in the case of local methods---as it does not allow the analysis to capture the very important effect of heterogeneity and can result in vacuous bounds.

In the Appendix \ref{ist:sec:app_comparison}, we present some limitations of other relevant previous approaches to training with compressed models: too restrictive assumptions on the algorithm \citep{mohtashami2022masked} or non-applicability in our problem setting \citep{chayti2022optimization}. In addition, we discuss the differences between IST and 3D parallelism \citep{shoeybi2019megatron}.

\section{Experiments} \label{ist:sec:experiments}

To empirically validate our theoretical framework and its implications, we focus on carefully controlled setting that satisfy the assumptions of our work. Specifically, we consider a quadratic problem defined in \eqref{ist:eq:het_gen_quad_problem},
where $\mL_i = \mB_i^\top \mB_i$. Entries of the matrices $\mB_i \in \bR^{d \times d}$, vectors $\bb_i \in \Rd$, and initialization $x^0\in\Rd$ are generated from a standard Gaussian distribution $\mathcal{N}(0, 1)$.

\begin{figure}[H]%
\centering
\subfigure[Function convergence for heterogeneous case.]{%
\label{ist:fig:hetero_convergence}%
\includegraphics[width=0.49\linewidth]{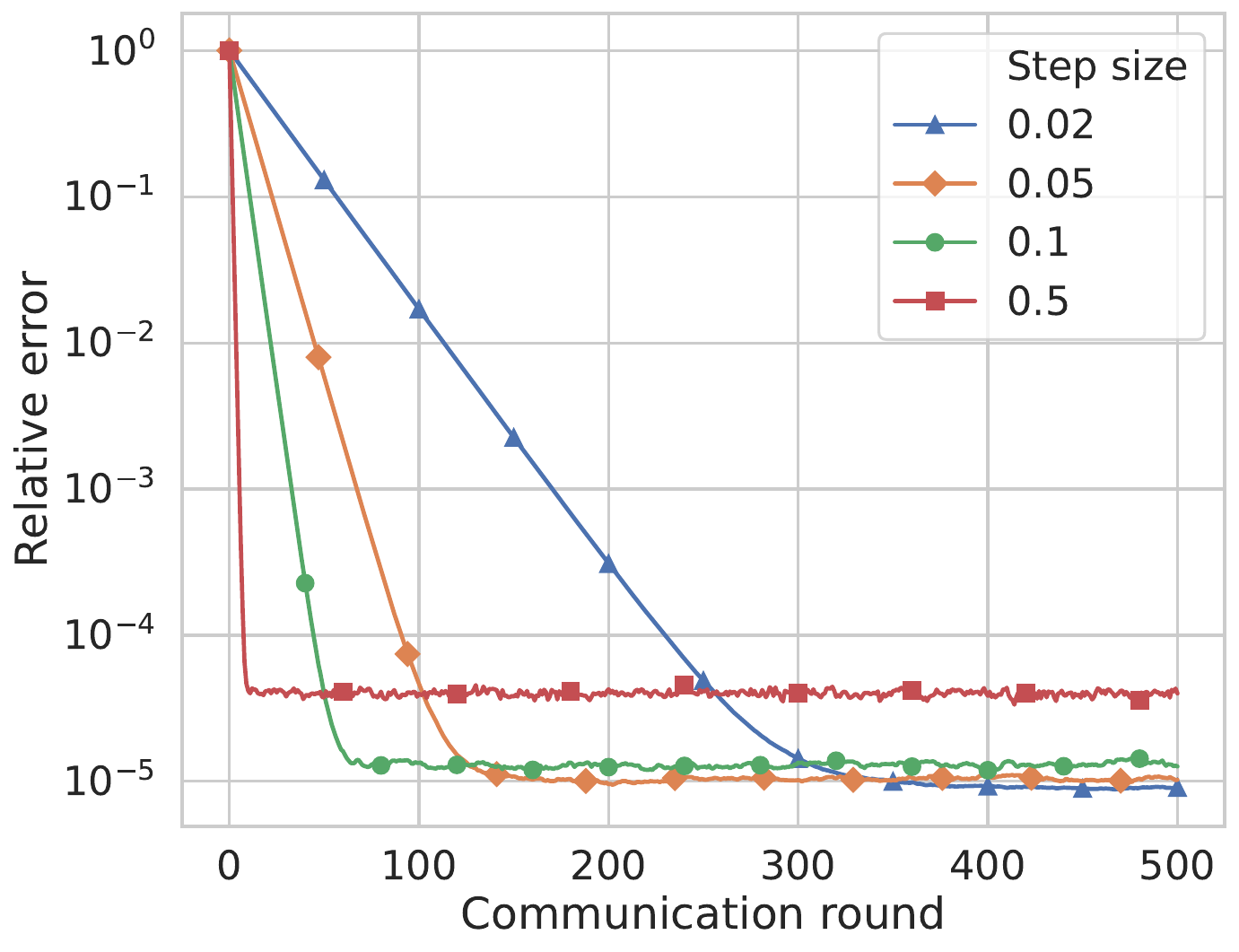}}%
\hfill
\subfigure[Iterates convergence for homogeneous case.]{%
\label{ist:fig:homo_stationary}%
\includegraphics[width=0.49\linewidth]{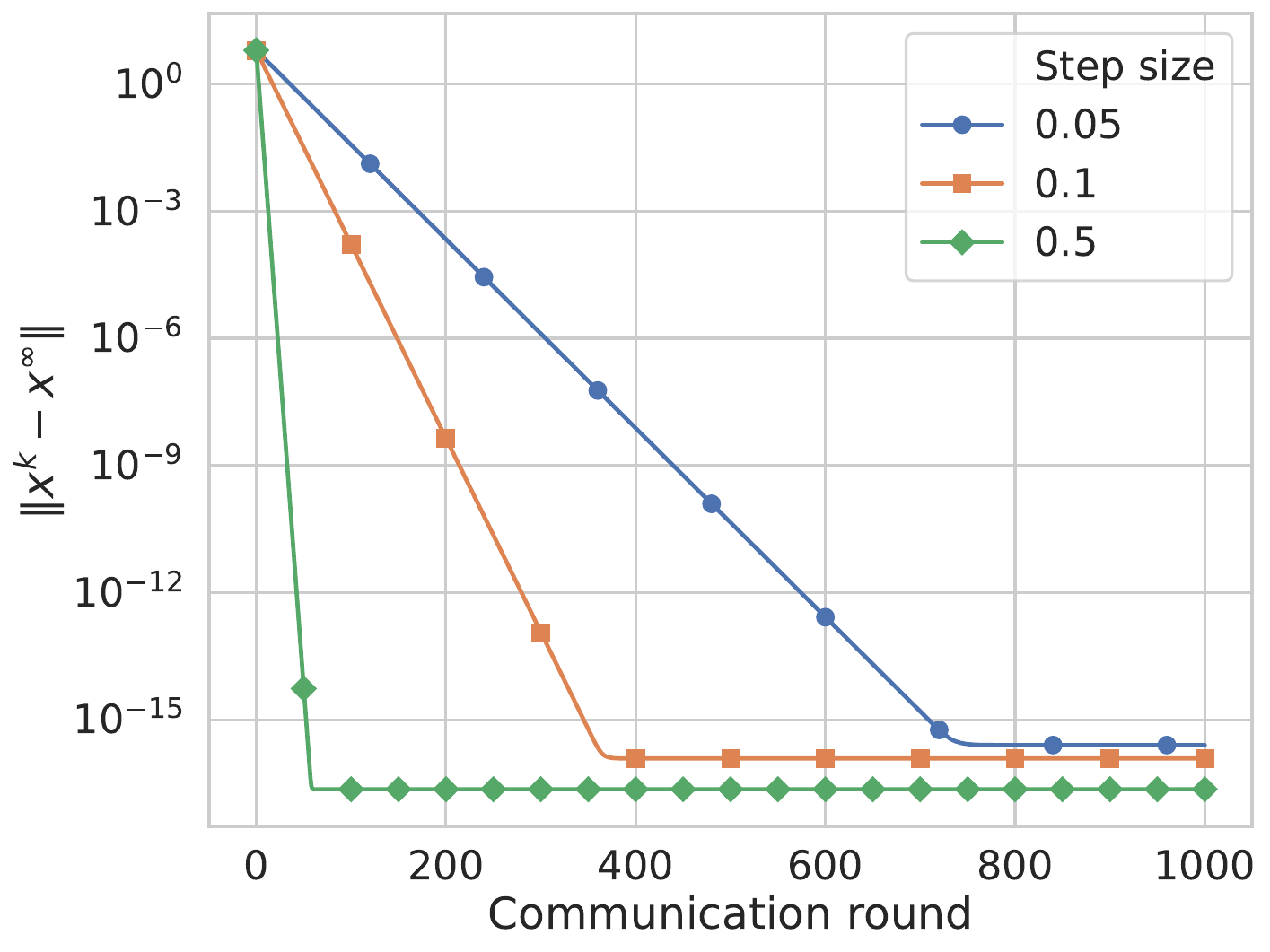}}%
\caption{Performance of simplified IST on a quadratic problem for varying step size values.}
\label{ist:fig:varying_ss}
\end{figure}

\textbf{Heterogeneous setting.}
In Figure \ref{ist:fig:hetero_convergence}, we present the performance of the simplified Independent Subnetwork Training (IST) algorithm (update \eqref{ist:eq:SGD_generic} with estimator \eqref{ist:eq:het_gen_quad_est}) for a heterogeneous problem. We fix the dimension $d$ to 1000 and the number of computing nodes $n$ to 10. We evaluate the logarithm of a relative functional error $\log (f(x^k) - f(x^\star)) / (f(x^0) - f(x^\star))$, while the horizontal axis denotes the number of communication rounds required to achieve a certain error tolerance. According to our theory \eqref{ist:eq:het_gen_grad_res} and \eqref{ist:eq:gen_func_res}, the method converges to a neighborhood of the solution, which depends on the chosen step size. Specifically, a larger step size allows for faster convergence but results in a larger neighborhood.

\textbf{Homogeneous setting.}
In Figure \ref{ist:fig:homo_stationary}, we demonstrate the convergence of the iterates $x^k$ for a homogeneous problem with $d = n = 50$.
The results are in close agreement with our theoretical predictions for the estimator \eqref{ist:eq:hom_gen_estimator}. We observe that the distance to the method's expected fixed point $x^\infty = \cb / \sqrt{n}$ decreases linearly for different step size values. This confirms that IST may not converge to the optimal solution $x^\star = \mcL^{-1} \cb$ of the original problem \eqref{ist:eq:het_gen_quad_problem} in general cases without interpolation. In addition, there are no visible oscillations in comparison to the heterogeneous case.

\subsection{Neural network results}
\begin{figure*}[h]%
\centering
\subfigure[Comparison of IST and Distributed Gradient Descent (DGD).]{%
\label{ist:fig:ist_vs_dgd}%
\includegraphics[width=0.49\linewidth]{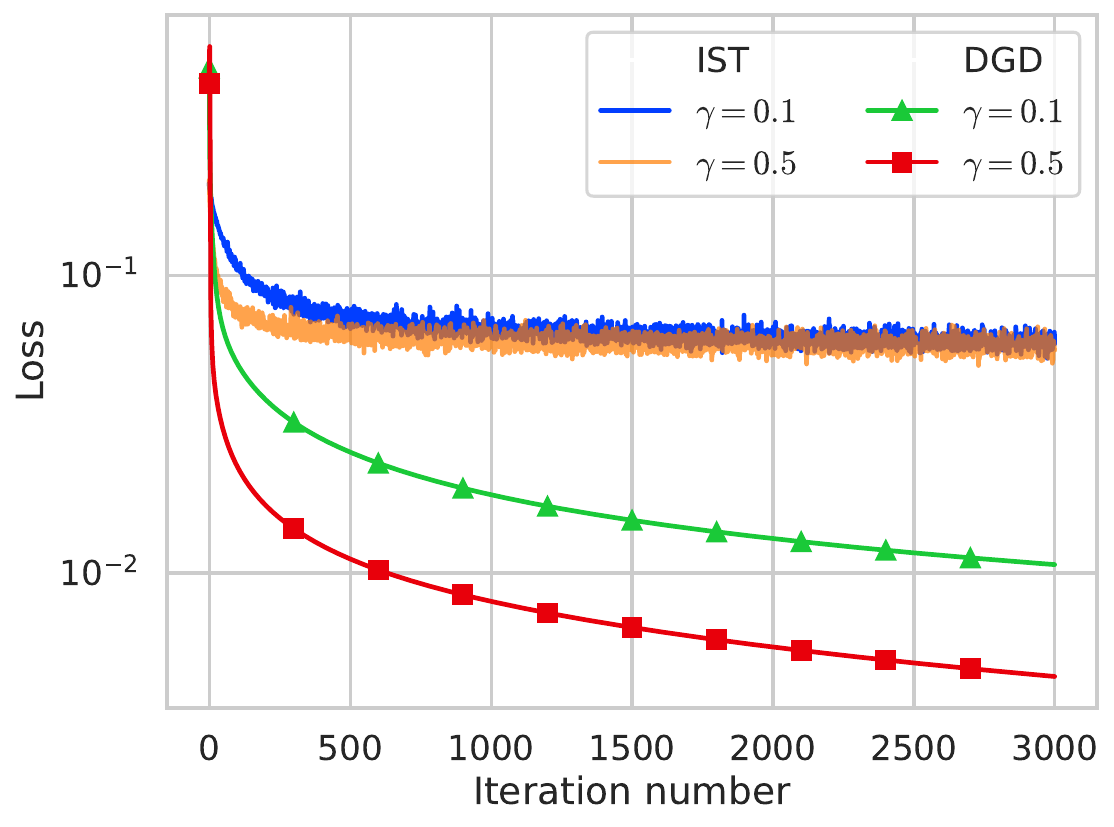}}%
\hfill
\subfigure[Performance of IST with different step sizes.]{%
\label{ist:fig:ist_decr}%
\includegraphics[width=0.49\linewidth]{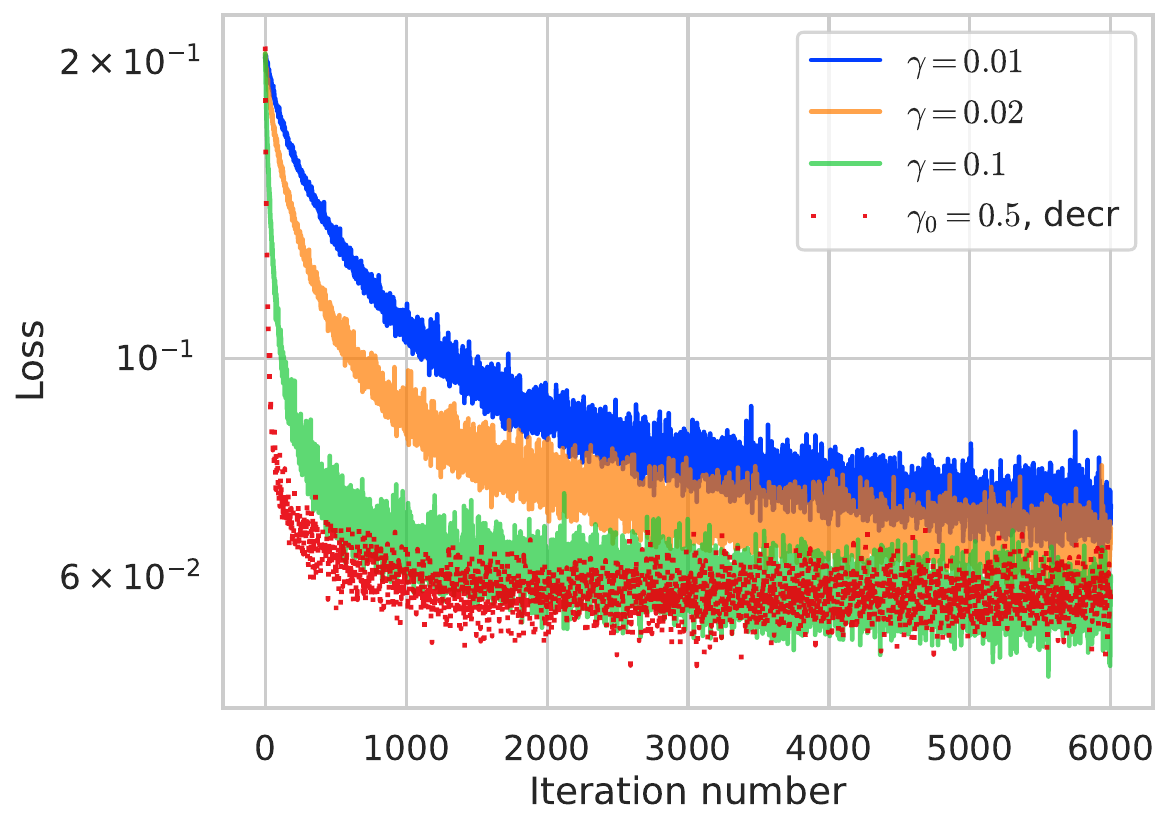}}%
\caption{Experimental study of IST on a neural network problem.}
\label{ist:fig:nn_experiments}
\end{figure*}

We closely follow the experimental setup of \citet{liao2022on} and use a one-hidden-layer neural network with ReLU activations. For completeness, we repeat some of the details here.
ResNet-50 model \citep{he2016deep} pre-trained on ImageNet is used as a feature extractor and concatenated with two fully connected layers. The resulting model is then trained on the CIFAR-10 \citep{krizhevsky2009cifar} dataset. We take the outputs of the re-trained ResNet-50 as the input features resulting in $d = 2048$, and the logit outputs of the combined model are used as the labels.

The goal of the experiment is to study the optimization performance of the IST method described in Section \ref{ist:sec:setup}. Namely, we consider Algorithm \ref{ist:alg:IST} with 1) $\mC_i$ chosen as \texttt{Perm-q} \eqref{ist:eq:permutation} for IST and 2) $\mC_i = \mI$ for Distributed Gradient Descent (DGD). Both methods are implemented across $n=10$ nodes, employing constant step sizes $\gamma$, and one local step per communication round.

Figure \ref{ist:fig:ist_vs_dgd} shows the logarithm of the average loss of submodels after every iteration.
The main observation is that IST's performance stagnates (starts oscillating) at some point, unlike DGD's. Namely, the studied method converges to the neighborhood whose size is basically the same for different step size values. This phenomenon distinguishes the method from standard SGD.
Moreover, larger $\gamma$ speeds up DGD and allows reaching the error floor (oscillation level) faster for IST. These observations agree well with our Theorems \ref{ist:thm:het_gen_gradient} and \ref{ist:thm:gen_function} as the convergence upper bound \eqref{ist:eq:het_gen_grad_res} has an irreducible term proportional to the bias norm $\sqN{h}_{\moL}$ of the gradient estimator. This term cannot be eliminated completely by decreasing the step size $\gamma$, unlike SGD \eqref{ist:eq:SGD_rate}.

In Figure \ref{ist:fig:ist_decr}, we take a closer look at the training loss during IST optimization. For this problem, the situation differs from Figure \ref{ist:fig:hetero_convergence}: even for smaller step size values ($\gamma \in \{0.01, 0.02\}$), the method converges to a higher error floor. Interestingly, if $\gamma$ is decreased by a factor of 10 every 1000 iterations, the method's performance ({\color{red} red} dotted curve) changes very little. This can be explained by the second term $\br{1-\gamma}\beta^{-1} \sqN{h}_{\moL}$ in the convergence bound \eqref{ist:eq:het_gen_grad_res} from Theorem \ref{ist:thm:het_gen_gradient}, which increases for smaller $\gamma$.
The observed effect distinguishes IST from the typical training situation in deep learning.

\section{Conclusions and future work}

In this study, we introduced a novel approach to understanding training with combined model and data parallelism for a quadratic model. Our framework sheds light on distributed submodel optimization, which reveals the advantages and limitations of Independent Subnetwork Training (IST). Moreover, we accurately characterized the behavior of the considered method in both homogeneous and heterogeneous scenarios without imposing restrictive assumptions on the gradient estimators.

In future research, it would be valuable to explore extensions of our findings to settings that are closer to scenarios, such as cross-device federated learning. This could involve investigating partial participation support, leveraging local training benefits, and ensuring robustness against stragglers.

It would be interesting to generalize our analysis to non-quadratic scenarios without relying on pathological assumptions.
This work shows a somewhat negative result regarding worst-case IST performance for standard Empirical Risk Minimization problem \eqref{ist:eq:general_problem}. However, IST was recently shown to be effective at solving an alternative optimization problem formulation by \citet{demidovich2023mast}.
Another potential promising research direction is algorithmic modifications of the original IST to solve the fundamental problems highlighted in this work and acceleration of training.
\endgroup

\chapter{Model-Agnostic Sparsified Training: A Sketch-Aware Objective}
\label{chapter3:mast}

\begingroup
\input{KAUST_Thesis/imported/03_mast/import_setup}

\section*{Abstract}
 We introduce a novel optimization problem formulation that departs from the conventional way of minimizing machine-learning model loss as a black-box function. Unlike traditional formulations, the proposed approach explicitly incorporates an initially pre-trained model and random sketch operators, allowing both the model and gradient to be sparsified during training. We establish properties of the proposed objective function and highlight its connections to the standard formulation. Furthermore, we present several variants of the Stochastic Gradient Descent (\SGD) method adapted to the new problem formulation, including \SGD\text{ }with general sampling, a distributed version, and \SGD\text{ }with variance reduction techniques. Our convergence guarantees use weaker assumptions or yield tighter rates in several settings and cover important techniques such as Dropout and sparse training. This work provides a basis for improving the theoretical understanding of model training through a sparsification-aware optimization approach.

\begin{quote}
\small
\noindent\textit{This chapter is based on:}\\
Yury Demidovich, Grigory Malinovsky, Egor Shulgin, and Peter Richt{\'a}rik, ``MAST: Model-Agnostic Sparsified Training,'' \emph{International Conference on Learning Representations (ICLR)}, 2025.
\nocite{demidovich2023mast}
\end{quote}

\section{Introduction}\label{mast:sec:introduction}

Efficient optimization methods have played an essential role in the advancement of modern Machine Learning (ML), given that many supervised problems ultimately reduce to the task of minimizing an abstract loss function --- a process that can be formally expressed as:
\begin{equation} \label{mast:eq:main}
	\min_{x\in \R^d} f(x),
\end{equation}
where $f:\R^d\to \R$ is a loss function of a model with parameters/weights $x \in \R^d$. The statement~\eqref{mast:eq:main} has been comprehensively analyzed within the domain of optimization literature. Within the ML community, substantial attention has been directed towards studying this problem, particularly in the context of the Stochastic Gradient Descent (\SGD) method \citep{robbins1951stochastic}. \SGD\text{ }stands as a foundational and highly effective optimization algorithm within the realm of ML \citep{bottou2018optimization}.
The pervasive use of \SGD\text{ } in the field attests to its versatility and success in training a diverse array of models \citep{goodfellow2016deep}. Notably, contemporary deep learning practices owe a substantial debt to \SGD, as it serves as the cornerstone for many state-of-the-art training techniques \citep{sun2020optimization}.

While problem \eqref{mast:eq:main} has been of primary interest in mainstream optimization research, this approach is not always best suited for representing recent ML techniques, such as sparse/quantized training \citep{wu2016quantized,hoefler2021sparsity},
resource-constrained distributed learning \citep{caldas2018expanding,wang2018edge}, model personalization \citep{smith2017federated,hanzely2020federated,mansour2020three}, and meta-learning \citep{schmidhuber1987evolutionary,finn2017model}.
Although various attempts have been made to analyze some of these settings \citep{khaled2019gradient,lin2019dynamic,mohtashami2022masked}, there is still no satisfactory optimization theory that can explain the success of these techniques in deep learning. Previous works analyze variants of \SGD\text{ }trying to solve problem~\eqref{mast:eq:main}, which often results in vacuous convergence bounds and/or overly restrictive assumptions on the class of functions and algorithms involved \citep{shulgin2023towards}. We assert that these issues arise due to mismatches between the method used and the problem formulation being solved.

In this work, to address the issues mentioned above, we propose a new optimization problem formulation called Model-Agnostic Sparsified Training (MAST):
\begin{equation}\label{mast:eq:pretrained_compressed_problem}
	\min_{x\in \R^d} \left[ \tf(x) \eqdef \Exp{f_{\mS}(x)}\right],
\end{equation}
where $f_{\mS}(x)\eqdef f(\shift +\mS (x-\shift))$ for $\shift\in \R^d$ (e.g., a pre-trained model, possibly compressed), $\mS \in \R^{d \times d}$ is a random matrix (i.e., a \textbf{sketch}) sampled from distribution $\cD$.

When $\shift = 0$, the function takes the following form: $f_{\mS}(x) = f(\mS x)$. This scenario may be considered a process in which the architecture requires training from scratch, with quantization as a central consideration. Such an approach involves a substantial increase in training time and the necessity of hyperparameter tuning to achieve effectiveness.
In terms of theory, formulation \eqref{mast:eq:pretrained_compressed_problem} can be viewed as a nested (stochastic) composition optimization \citep{bertsekas1977approximation,poljak1979bertsekas,ermoliev1988stochastic}.
Moreover, our formulation is partially related to the setting of splitting methods \citep{condat2023proximal}.
However, due to their generality other setups do not take the peculiarities of the considered problem instance into account and are focused on other areas of application. Conversely, when $\shift$ is non-zero and pre-trained weights are utilized ($\shift \neq 0$), the newly formulated approach can be interpreted as a process of acquiring a \say{meta-model} $x$. Solving problem \eqref{mast:eq:pretrained_compressed_problem} then ensures
that the sketched model $\shift + \mS(x - \shift)$ exhibits strong performance on average.
This interpretation shares many similarities with Model-Agnostic Meta-Learning \citep{finn2017model}.

We argue that this framework may be better suited for modeling various practical ML techniques, as discussed below. Furthermore, our proposed framework facilitates thorough theoretical analysis and holds potential independent interest for a broader audience.
While our analysis and algorithms work for a quite general class of sketches, we focus on applications that are relevant for sparse $\mS$.

\subsection{Motivating examples}

\textbf{Dropout} \citep{hanson1990stochastic,hinton2012improving,frazier2018dropout} is a regularization technique originally introduced to prevent overfitting in neural networks by dropping some of the model's units (activations) during training. Later, this approach was generalized to incorporate Gaussian noise \citep{wang2013fast,srivastava2014dropout} (instead of Bernoulli masks) and zeroing the weights via DropConnect \citep{wan2013regularization} or entire layers \citep{fan2019reducing}. It was also observed \citep{srivastava2014dropout} that training with Dropout induces sparsity in the activations. In addition, using the Dropout-like technique \citep{gomez2019learning,lejeune2021flip} can make the resulting network more amenable to subsequent sparsification (via pruning) before deployment. Modern DL has seen a huge increase in the size of models \citep{villalobos2022machine}. This has resulted in growing energy and computational costs, necessitating optimizations of neural networks' training pipeline \citep{yang2017method}. Among others, pruning and sparsification were proposed as effective techniques due to the overparametrization properties of large models \citep{chang2021provable}.

\textbf{Sparse training} algorithms \citep{mocanu2016topological,guo2016dynamic,mocanu2018scalable}, in particular, suggest working with a smaller subnetwork during every optimization step. This increases efficiency by reducing the model size (via compression), which naturally brings memory and computation acceleration benefits to the inference stage. Moreover, sparse training has recently been shown to speed up optimization by leveraging sparse backpropagation \citep{nikdan2023sparseprop}.

\textbf{On-device learning} also creates a need for sparse or submodel computations due to the memory, energy, and computational constraints of edge devices. In settings like cross-device federated learning \citep{konecny2017federated,mcmahan2017communication,kairouz2021advances}, models are trained in a distributed way across a population of heterogeneous nodes, such as mobile phones.
The heterogeneity of the clients' hardware makes it necessary to adapt the (potentially large) server model to the needs of low-tier devices \citep{caldas2018expanding,bouacida2021adaptive,horvath2021fjord}.

\paragraph{Contributions.}
The main results of this work include:

$\bullet$ A rigorous formalization of a new optimization formulation, as shown in Equation \eqref{mast:eq:pretrained_compressed_problem}, which can encompass various important practical settings as special cases, such as Dropout and Sparse training.

$\bullet$ In-depth theoretical characterization of the proposed problem's properties, highlighting its connections to the standard formulation in Equation \eqref{mast:eq:main}. Notably, our problem is efficiently solvable with practical methods.

$\bullet$ The development of optimization algorithms that naturally emerge from the formulation in Equation \eqref{mast:eq:pretrained_compressed_problem}, along with insightful convergence analyses, both in non-convex and (strongly) convex settings.

$\bullet$ The generalization of the problem and methods to the distributed scenario, expanding the range of applications even further, including scenarios like IST and Federated Learning.

$\bullet$ An experimental study of the proposed algorithms and MAST properties in the context of machine learning, highlighting the advantages of the proposed approach.

\textbf{Chapter organization.} We introduce the basic formalism and discuss sketches in Section \ref{mast:sec:sketches}. The properties of the suggested formulation \eqref{mast:eq:pretrained_compressed_problem} are analyzed in Section \ref{mast:sec:properties}. Section \ref{mast:sec:convergence} contains convergence results for full-batch, stochastic, and variance-reduced methods for solving problem \eqref{mast:eq:pretrained_compressed_problem}. Extensions to the distributed case are presented in Section \ref{mast:sec:distributed}. Section~\ref{mast:sec:experiments} describes some of our experimental results.
The last Section~\ref{mast:sec:conclusion} concludes the chapter and outlines potential directions of future work. All proofs are provided in Appendix~\ref{appendix:chapter3_mast}.

\section{Sketches}
\label{mast:sec:sketches}
(Stochastic) gradient-based methods are mostly used to solve problem \eqref{mast:eq:main}. Applying this paradigm to our formulation \eqref{mast:eq:pretrained_compressed_problem} requires computing the gradient of $\tf = \Exp{f_{\mS}}$. In the case of the general distribution $\mS \sim \cD$, such computation may not be possible due to expectation $\mathbb{E}$. Therefore, practical algorithms typically rely on gradient estimates. An elegant property of the MAST problem \eqref{mast:eq:pretrained_compressed_problem} is that the gradient estimator takes the form
\begin{equation} \label{mast:eq:grad_est}
	\nabla f_{\mS}(x) = \mS^\top \nabla f(\shift + \mS(x - \shift)),
\end{equation}
due to the chain rule, as matrix $\mS$ is independent of $x$. Sketches/matrices $\mS$ are random and sampled from distribution $\cD$.
Note that estimator \eqref{mast:eq:grad_est} sketches both the model $x$ and the gradient of $f$.
Next, we explore some of the sketches' important properties and give some practical examples.
\begin{assumption} \label{mast:ass:C}
	The sketching matrix $\mS$ satisfies:
	\begin{equation} \label{mast:eq:sketch_def}
		\Exp{\mS } = \mI, \qquad \text{and} \qquad \Exp{\mS^\top \mS} \; \text{is finite},
	\end{equation}
	where $\mI$ is the identity matrix. Note that $\mS^\top \nabla f(x)$ is an unbiased estimator of the gradient $\nabla f(x)$.
\end{assumption}
Denote $L_{\mS}\eqdef \lambda_{\max}(\mS^\top \mS)$, $\mu_{\mS}\eqdef \lambda_{\min}(\mS^\top \mS)$, and $L_{\mathcal{D}} \eqdef \lambda_{\max}(\Exp{\mS^\top \mS})$, $ \mu_{\mathcal{D}} \eqdef \lambda_{\min}(\Exp{\mS^\top \mS})$,  where $\lambda_{\max}$ and $\lambda_{\min}$ represent the largest and smallest eigenvalues.
Clearly, $L_{\mS}\geq \mu_{\mS}\geq 0$ and $L_{\mathcal{D}} \geq \mu_{\mathcal{D}} \geq 0$.
If Assumption~\ref{mast:ass:C} is satisfied, then $\Exp{\mS^\top \mS} \succeq \mI$, which means that $\mu_{\mathcal{D}} \geq 1$ and
\[ \norm{x}^2 \leq \mu_{\mathcal{D}} \norm{x}^2  \leq \Exp{\norm{\mS x}^2} \leq L_{\mathcal{D}} \norm{x}^2.\]

\subsection{Diagonal sketches}

Let $c_1, c_2, \dots, c_d$ be a collection of random variables and define a matrix with $c_i$-s on the diagonal
\begin{equation} \label{mast:eq:diag_sketch}
	\mS = \Diag (c_1, c_2, \dots, c_d),
\end{equation}
which satisfies Assumption \ref{mast:ass:C} when $\Exp{c_i} = 1$ and $\Exp{c_i^2}$ is finite for every $i$.

The following example illustrates how our framework can be used for modeling Dropout.
\begin{example}
	The \textbf{independent Bernoulli sparsification} operator is defined as a diagonal sketch  \eqref{mast:eq:diag_sketch}, where every $c_i$ is an (independent) scaled Bernoulli random variable:
	\begin{equation} \label{mast:eq:ind_sparse}
		c_i =
		\left\{\begin{array}{ll}
			1/p_i, & \text{with probability }  p_i \\
			0, & \text{with probability }  1-p_i
		\end{array} , \right.
	\end{equation}
	for $p_i \in (0, 1]$ and $i \in [d] \eqdef \{1, \dots, d\}$.
\end{example}

It can be shown that for independent Bernoulli sparsifiers,
\begin{equation}
	L_{\mathcal{D}} = \max_i p_i^{-1} \eqdef p^{-1}_{\min}, \quad \mu_{\mathcal{D}} = \min p_i^{-1} \eqdef p^{-1}_{\max}.
\end{equation}
Notice that when $p_i \equiv p$, $i\in[d]$, and $\shift = 0$, gradient estimator \eqref{mast:eq:grad_est} results in a sparse update as $\mS^\top \nabla f(\mS x)$ drops out $d (1-p)$ (on average) components of model weights and the gradient.

Next, we show another practical example of a random sketch often used for reducing communication costs in distributed learning \citep{konecny2017federated,wangni2018gradient,stich2018sparsified}.
\begin{example}
	\textbf{Random $K$ sparsification} (in short, \rnd{\text{$K$}} for $K \in [d]$) operator is defined by
	\begin{equation} \label{mast:eq:rand-k}
		\mS_{\rnd{\text{$K$}}} \eqdef \frac{d}{K} \sum\limits_{i\in S} e_i e_i^\top,
	\end{equation}
	where $e_1, \dots, e_d \in \R^d$ are standard unit basis vectors, and $S$ is a random subset of $[d]$ sampled from the uniform distribution over all subsets of $[d]$ with cardinality $K$.
\end{example}

$\rnd{\text{$K$}}$ belongs to the class of diagonal sketches \eqref{mast:eq:diag_sketch}. $\mS x$ preserves $K$ non-zero coordinates out of total $d$ coordinates. Since $\mathbb{E}\left[\mS^\top\mS\right] = \mI \cdot \sfrac{d}{K}$, this sketch satisfies $L_{\mathcal{D}} = \mu_{\mathcal{D}} = d/K$. This example is suitable for modeling fixed budget sparse training when the proportion ($K/d$) of network parameters being updated remains constant during optimization \citep{mocanu2018scalable,evci2020rigging}.

\section{Problem properties} \label{mast:sec:properties}
We show that the proposed formulation \eqref{mast:eq:pretrained_compressed_problem} inherits the smoothness and convexity properties of the original problem \eqref{mast:eq:main}. Let us introduce the most standard assumptions in the optimization field.
\begin{assumption} \label{mast:ass:l_f_smooth}
	Function $f$ is differentiable and $L_f$-\textbf{smooth}, i.e., there is $L_f>0$ such that
	\[f(x+h) \leq f(x) + \ve{\nabla f(x)}{h} + \frac{L_f}{2}\norm{h}^2 \quad  \forall x, h \in \R^d.\]
	We also require $f$ to be lower bounded by $f^{\inf} \in \mathbb{R}$.
\end{assumption}
\begin{assumption} \label{mast:ass:mu_f-convex}
	Function $f$ is differentiable and $\mu_f$-\textbf{strongly convex}, i.e., there is $\mu_f>0$ such that %
	\[f(x+h) \geq f(x) + \ve{\nabla f(x)}{h} + \frac{\mu_f}{2}\norm{h}^2 \quad  \forall x, h \in \R^d.\]
\end{assumption}

Next, we show how the choice of the sketch $\mS$ affects the smoothness parameters of $f_{\mS}$ and $\tf$.%

\begin{lemma}[Consequences of $L_f$-smoothness] \label{mast:lemma:smoothness}
	If $f$ is $L_f$-smooth, then
	\begin{itemize}
		\item[(i)] $f_{\mS}$ is $L_{f_\mS}$-smooth with $L_{f_\mS} \leq L_{\mS} L_f $.

		\item[(ii)] $\tf$ is $L_{\tf}$-smooth with $L_{\tf}\leq L_{\mathcal{D}} L_f $.

		\item[(iii)] $\tf(x) \leq f(x) + \frac{(L_{\mathcal{D}}-1) L_f}{2} \norm{x-\shift}^2 \quad \forall x\in \R^d.$
	\end{itemize}
\end{lemma}

In particular, property~($iii$) in Lemma~\ref{mast:lemma:smoothness} demonstrates that the gap between the sketched loss $\tf$ and the original function $f$ depends on the model weights and the smoothness parameter of function $f$.

\begin{lemma}[Consequence of Convexity] \label{mast:lem:convex-only}
	If $f$ is convex, then $\tf$ is convex and $\tf(x)\geq f(x)$.%
\end{lemma}
It shows that the convexity of $f$ is preserved, and the \say{sketched} loss is always greater than the original loss.
Moreover, Lemma \ref{mast:lem:convex-only} (along with other results in this section) offers a huge advantage of the proposed problem formulation over the sparsification-promoting alternatives based on $\ell_0$-norm regularization \citep{louizos2018learning,peste2021ac}, that make the original problem hard to solve.

\begin{lemma}[Consequences of $\mu_f$-convexity] \label{mast:lemma:consequences_strongly_convex}
	If $f$ is $\mu_f$-convex, then
	\begin{itemize}
		\item[(i)] $f_{\mS}$ is $\mu_{f_\mS}$-convex with $\mu_{f_\mS} \geq \mu_{\mS} \mu_f $.

		\item[(ii)] $\tf$ is $\mu_{\tf}$-convex with $\mu_{\tf} \geq \mu_{\mathcal{D}} \mu_f $.

		\item[(iii)] $\tf(x) \geq f(x) + \frac{ (\mu_{\mathcal{D}}-1) \mu_f}{2} \norm{x-\shift}^2 \quad  \forall x\in \R^d.$
	\end{itemize}
\end{lemma}

As a consequence, we get the following result for the condition number of the proposed problem:
\begin{equation}
	\kappa_{\tf} \eqdef \frac{L_{\tf}}{\mu_{\tf}} \leq
	\frac{L_{\mathcal{D}} L_f}{\mu_{\mathcal{D}} \mu_f} =
	\frac{L_{\mathcal{D}}}{\mu_{\mathcal{D}}} \kappa_f.
\end{equation}

Therefore, $\kappa_{\tf} \leq \kappa_f \cdot L_{\cD}$
as $\mu_{\mathcal{D}} \geq 1$. Thus, the resulting condition number may increase, which indicates that $\tf$ may be harder to optimize, which agrees with the intuition that compressed training is harder \citep{evci2019difficulty}. In addition, for independent Bernoulli sparsifiers~\eqref{mast:eq:ind_sparse},
\begin{equation}
    \kappa_{\mathcal{D}} \eqdef L_{\mathcal{D}}/ \mu_{\mathcal{D}} = p_{\max}/p_{\min},
\end{equation}
which shows that the upper bound on the ratio $\kappa_{\tf} / \kappa_f $ can be made as large as possible by choosing a small enough $p_{\min}$. At the same time, $\kappa_{\cD} = 1$ for classical Dropout: $p_i \equiv p$, $i\in[d],$ indicating that training with Dropout may be no harder than optimizing the original model.

\textbf{Relation between $f$ and $\tf$ minima.} Let $\cX^\star$ be the solutions to problem~\ref{mast:eq:main}, and $\cX^\star_{\mathcal{D}}$ the solutions of the new MAST problem~\eqref{mast:eq:pretrained_compressed_problem}.
We now show that a solution $x_\cD^\star\in \cX_\cD^\star$ of \eqref{mast:eq:pretrained_compressed_problem} is an approximate solution of the original problem \eqref{mast:eq:main}.
\begin{theorem} \label{mast:thm:approx} Let Assumptions~\ref{mast:ass:l_f_smooth}~and~\ref{mast:ass:mu_f-convex} hold, and let $x_\cD^\star \in \cX_\cD^\star$ and $x^\star \in \cX^\star$. Then
	\begin{align*}
		f(x^\star) \leq f(x_\cD^\star) \leq f(x^\star) + \frac{(L_{\mathcal{D}}-1)L_f}{2}\norm{x^\star-\shift}^2
		 - \frac{(\mu_{\mathcal{D}}-1)\mu_f}{2}\norm{x_\cD^\star-\shift}^2;
	\end{align*}
	\begin{align*}
		f(x^\star) +
		\frac{(\mu_{\mathcal{D}}-1)\mu_f}{2}\norm{x_\cD^\star-\shift}^2
		\leq \tf(x_\cD^\star)
		\leq f(x^\star)
		+ \frac{(L_{\mathcal{D}}-1)L_f}{2}\norm{x^\star-\shift}^2.
	\end{align*}
\end{theorem}

Consider \rnd{\text{$K$}} as a sketch.
If $K = (1 - \varepsilon) d$ for some $\varepsilon \in [0, 1)$, which corresponds to dropping roughly an $\varepsilon$ share of coordinates, then $L_{\mathcal{D}}-1 = \mu_{\mathcal{D}}-1 = \nicefrac{d}{K}-1 = \nicefrac{\varepsilon}{(1-\varepsilon)}.$ Theorem \ref{mast:thm:approx} then states that
$$
	f(x^\star) \leq f(x_\cD^\star) \leq f(x^\star) + \frac{\varepsilon}{2(1-\varepsilon)} \left(L_f \norm{x^\star-\shift}^2 - \mu_f \norm{x_\cD^\star-\shift}^2\right);
$$
$$
	f(x^\star) +
	\frac{\varepsilon\mu_f}{2(1-\varepsilon)}\norm{x_\cD^\star-\shift}^2
	\leq \tf(x_\cD^\star)\leq f(x^\star)
	+ \frac{\varepsilon L_f}{2(1-\varepsilon)}\norm{x^\star-\shift}^2.
$$
If $\varepsilon$ is small (a ``light'' sparsification), or if the pre-trained model $\shift$ is close to $x^\star$, then  $x_\cD^\star$ will have a small loss, comparable to the loss of the optimal model~$x^\star;$ $x_\cD^\star$ will be close to the pre-trained model~ $\shift;$ MAST loss  will be small comparable to the loss of the optimal uncompressed model~$x^\star.$

\section{Individual node setting} \label{mast:sec:convergence}

In this section, we discuss the properties of the \SGD-type algorithm applied to problem \eqref{mast:eq:pretrained_compressed_problem}
\begin{equation}
	\label{mast:eq:GD}
	x^{t+1} = x^t - \gamma \nabla f_{\mS^{t}} (x^t) = x^t - \gamma \left(\mS^{t}\right)^\top \nabla f(y^t)
\end{equation}
for $y^t = \shift + \mS^t (x^t - \shift)$, where $\mS^t$ is sampled from $\cD$.
One advantage of the proposed formulation \eqref{mast:eq:pretrained_compressed_problem} is that it naturally gives rise to the described method, which generalizes standard (Stochastic) Gradient Descent.
As noted before, due to the properties of the gradient estimator \eqref{mast:eq:grad_est}, recursion \eqref{mast:eq:GD} defines an Algorithm \ref{mast:alg:SGD} (I) that sketches both the model and the gradient of $f$.

Let us introduce a notation frequently used in our convergence results. This quantity is determined by the spectral properties of the sketches being used.
\begin{equation}
	L_{\mS}^{\max} \eqdef \sup_{\mS} L_{\mS} = \sup_{\mS} \lambda_{\max}\left(\mS^{\top}\mS\right),
\end{equation}
where $\sup_{\mS} L_{\mS}$ represents the tightest constant such that $L_{\mS} \leq \sup_{\mS} L_{\mS}$ almost surely.
For independent random sparsification sketches \eqref{mast:eq:ind_sparse}:
$L_{\mS}^{\max} = 1 / p^2_{\min}$. If $\mS$ is \rnd{\text{$K$}} \eqref{mast:eq:rand-k} then $L_{\mS}^{\max} = d^2/K^2$. Our convergence analysis relies on the following Lemma:

\begin{lemma} \label{mast:lemma:ac}
	Assume that $f$ is $L_f$-smooth \eqref{mast:ass:l_f_smooth} and $\mS$ satisfies Assumption \ref{mast:ass:C}. Then we have that $\forall x\in\mathbb{R}^d$
	\begin{equation*}
		\begin{split}
			\Exp{\left\|\nabla f_{\mS}(x)\right\|^2} & \leq 2L_f L_{\mS}^{\max} \left(\tf(x) - f^{\inf}\right), %
		\end{split}
	\end{equation*}
	where the expectation is taken with respect to $\mS.$
\end{lemma}

It generalizes a standard property of smooth functions often used in the non-convex analysis of \SGD\text{ }\citep{khaled2020better}. Now, we are ready to present our first convergence result.
\begin{theorem}
	\label{mast:thm:str_conv}
	Assume that $f$ is $L_f$-smooth (\ref{mast:ass:l_f_smooth}), $\mu_f$-strongly convex \eqref{mast:ass:mu_f-convex}, and $\mS$ satisfies Assumption \ref{mast:ass:C}.
	Then, for stepsize $\gamma \leq 1 / (L_fL_{\mS}^{\max})$, the iterates of Algorithm \ref{mast:alg:SGD} (I)
	satisfy
	\begin{equation} \label{mast:eq:SGD_cvx_res}
			\Exp{\left\|x^T - x_\cD^\star\right\|^2}\leq
			\left(1 - \gamma \mu_{\mathcal{D}} \mu_f\right)^T \sqn{x^0 - x_\cD^\star} + \frac{2\gamma L_fL_{\mS}^{\max}}{\mu_f\mu_{\mathcal{D}}} \left(\tf^{\inf} - f^{\inf}\right).
	\end{equation}
\end{theorem}
This theorem establishes a linear convergence rate with a constant stepsize up to a neighborhood of the solution to the MAST problem~\eqref{mast:eq:pretrained_compressed_problem}. Our result is similar to the convergence of \SGD\text{ } \citep{gower2019sgd} for the standard formulation \eqref{mast:eq:main}, with two differences discussed below.

\textbf{1.} Both terms of the upper bound \eqref{mast:eq:SGD_cvx_res} depend not only on the smoothness and convexity parameters of the original function $f$, but also on the \textit{spectral properties} of sketches $\mS$. Thus, for independent Bernoulli sparsification sketches \eqref{mast:eq:ind_sparse} with $p_i \equiv p$ (or \rnd{\text{$K$}}), the linear convergence term deteriorates and the neighborhood size is increased by $1/p^2$ ($d^2/K^2$ respectively). Therefore, we conclude that higher sparsity makes optimization harder.

\textbf{2.} Interestingly, the neighborhood size of \eqref{mast:eq:SGD_cvx_res} depends on the difference between the minima of $\tf$ and $f$, rather than on the variance of stochastic gradients at the optimum, as is typical for \SGD~\citep{gower2019sgd}. Thus, the method may even linearly converge to the exact solution when $\tf^{\inf} = f^{\inf}$, which we refer to as the \textit{interpolation} condition. This condition may naturally hold when the original and sketched models are sufficiently overparametrized (allowing minimization of the loss to zero). Notable examples in which similar phenomena have been observed in practice are training with Dropout \citep{srivastava2014dropout} and the \say{lottery ticket hypothesis} \citep{frankle2018lottery}.

Next, we provide results in the non-convex setting.
\begin{theorem} \label{mast:thm:sgd_nonconvex}
	Assume that $f$ is $L_f$-smooth \eqref{mast:ass:l_f_smooth} and $\mS$ satisfies Assumption \ref{mast:ass:C}. Then, for the stepsize $\gamma\leq 1/(L_f\sqrt{L_{\mathcal{D}}L_{\mS}^{\max} T})$,
	the iterates of Algorithm \ref{mast:alg:SGD} (I) satisfy
	\begin{equation*} \label{mast:eq:SGD_noncvx_res}
			\min \limits_{0\leq t < T} \Exp{\left\|\nabla \tf(x^t)\right\|^2} \leq \frac{3 \br{\tf(x^0) - \tf^{\inf}}}{\gamma T}
			 + \gamma L_f^2 L_{\mathcal{D}} L_{\mS}^{\max} \br{\tf^{\inf} - f^{\inf}}.
	\end{equation*}
\end{theorem}

This theorem shows an $\cO(1/\sqrt{T})$ convergence rate for reaching a stationary point. Our result shares similarities with the theory of \SGD~\citep{khaled2020better} for problem~\eqref{mast:eq:main}, with the main difference that the rate depends on the distribution of sketches through $\sqrt{L_{\mathcal{D}} L_{\mS}^{\max}}$.
Moreover, the second term depends on the difference between the minima of $\tf$ and $f$, as in the strongly convex case. However, the first term depends on the gap between the initialization and the lower bound of the loss function, which is more common in non-convex settings \citep{khaled2020better}.

\begin{corollary}[Informal] \label{mast:cor:sgd_nonconvex}
	Fix $\varepsilon > 0$ and denote
	$\delta^t \eqdef \Exp{\tf(x^t) - \tf^{\inf}},  r^t \eqdef \Exp{\left\|\nabla \tf(x^t)\right\|^2}.$
	Then, for the stepsize	$
		\gamma = \min\left\lbrace \frac{1}{\sqrt{DT}}, \frac{\varepsilon^2}{2D\left(\tf^{\inf} - f^{\inf}\right)} \right\rbrace,
	$
	where $D \eqdef L_f \sqrt{L_{\mathcal{D}}L_{\mS}^{\max} }$, Algorithm \ref{mast:alg:SGD} (I) needs
	$
		T\geq \frac{12 \delta^0 D}{\varepsilon^4} \max\left\lbrace 3\delta^0, \tf^{\inf} - f^{\inf}\right\rbrace
	$
	iterations to reach a stationary point $\min_{0 \leq t < T} r^t \leq \varepsilon^2$, which is order $\cO(\varepsilon^{-4})$ optimal \citep{ghadimi2013stochastic,drori2020complexity}.
\end{corollary}
In Appendix~\ref{appendix:chapter3_mast}, we also provide a general convex analysis.

\begin{algorithm}[t]
	\begin{algorithmic}[1]
		\caption{Double Sketched $\mathsf{(S)GD}$} \label{mast:alg:SGD}
		\STATE \textbf{Parameters:} learning rate $\gamma > 0$; distribution $\mathcal{D}$; initial model and shift $x^0, \shift \in \R^d$.
		\FOR{$t = 0, 1, 2 \ldots$}
		\STATE Sample a sketch: $\mS^t \sim \cD$
		\STATE Form a gradient estimator: \\
		$g^t =
		\left\{\begin{array}{l}
			\nabla f_{\mS^t} (x^t) \hspace{20pt} \rhd\text{exact (I)} \\
			g_{\mS^t}\left(x^t\right) \hspace{27pt} \rhd\text{(stochastic) inexact (II)} \\
		\end{array} \right.$
		\STATE Perform a gradient-type step: $x^{t+1} = x^t - \gamma g^t$
		\ENDFOR
	\end{algorithmic}
\end{algorithm}
\subsection{(Stochastic) inexact gradient}
Algorithm \ref{mast:alg:SGD}\,(I) is probably the simplest approach for solving the MAST problem~\eqref{mast:eq:pretrained_compressed_problem}. Analyzing this algorithm isolates and highlights the unique properties of the proposed problem formulation. Algorithm \ref{mast:alg:SGD}\,(I) requires exact (sketched) gradient computations at every iteration, which may not be feasible/efficient for modern ML applications. Hence, we consider the following generalization:
\begin{equation} \label{mast:eq:abc_sgd}
	x^{t+1} = x^t - \gamma g_{\mS^t}(x^t),
\end{equation}
where $g_{\mS^t}(x^t)$ is a gradient estimator satisfying
\begin{align} \label{mast:eq:unbiased_abc}
	\Exp{g_{\mS}(x)} &= \nabla f_{\mS}(x),\\
	\label{mast:eq:abc}
	\Exp{\left\|g_{\mS}(x)\right\|^2} &\leq 2A\left(f_{\mS}(x) - f^{\inf}_{\mS}\right) + B\left\|\nabla f_{\mS}(x)\right\|^2 + C,
\end{align}
for $\forall x \in\mathbb{R}^d$ and some constants $A, B, C \geq 0$.

The first condition \eqref{mast:eq:unbiased_abc} is an unbiasedness assumption standard for the analysis of \SGD-type methods. The second (so-called \say{ABC}) inequality \eqref{mast:eq:abc} is one of the most general assumptions covering bounded stochastic gradient variance, subsampling/minibatching of data, and gradient compression \citep{khaled2020better,demidovich2023guide}. Note that the expectation in \eqref{mast:eq:unbiased_abc} and \eqref{mast:eq:abc} is taken with respect only to the randomness of the stochastic gradient estimator and not the sketch $\mS$.

Algorithm \ref{mast:alg:SGD}\,(II) describes the resulting method in detail. We state the convergence result for it.

\begin{theorem} \label{mast:thm:abc_ncvx}
	Assume that $f$ is $L_f$-smooth \eqref{mast:ass:l_f_smooth}, $\mS$ satisfies Assumption \ref{mast:ass:C}, and the gradient estimator $g(x)$ satisfies conditions~\ref{mast:eq:unbiased_abc},~\ref{mast:eq:abc}.
	Denote $D_{A, B} = A + B L_f L_{\mS}^{\max}$ then, for stepsize $\gamma \leq 1/\sqrt{L_f L_{\mathcal{D}} D_{A, B} T}$, the iterates of Algorithm~\ref{mast:alg:SGD} (II) satisfy
	\begin{equation*} \label{mast:eq:ABC_noncvx_res}
			\min_{0 \leq t < T} \Exp{\left\|\nabla \tf(x^t)\right\|^2} \leq \frac{3 \br{\tf(x^0) - \tf^{\inf}}}{\gamma T} + \frac{\gamma L_f L_{\mathcal{D}}}{2}
			\Bigl\{C + 2 D_{A, B} \left(\tf^{\inf} - f^{\inf}\right)\Bigr\}.
	\end{equation*}
\end{theorem}

Similarly to Theorem \ref{mast:thm:sgd_nonconvex}, this result establishes an $\cO(1/\sqrt{T})$ convergence rate. However, the upper bound in \eqref{mast:eq:ABC_noncvx_res} is affected by constants $A, B, C$ due to the inexactness of the gradient estimator. The case of $A = C = 0, B = 1$ sharply recovers our previous Theorem \ref{mast:thm:sgd_nonconvex}. When $B = 1, C = \sigma^2$, we obtain convergence of \SGD\text{ } with bounded (by $\sigma^2$) variance of stochastic gradients. Moreover, when the loss is represented as a finite-sum
\begin{equation} \label{mast:eq:sketch_fin_sum}
	\tf (x) = \frac{1}{n} \sum\limits_{i=1}^n \Exp{f_{i}(\shift + \mS(x - \shift))},
\end{equation}
where each $f_i$ is $L_{f_i}$-smooth and lower-bounded by $f_i^{\inf}$, then $A = \max_i L_{f_i}, C = 2A \br{\frac{1}{n} \sum_{i=1}^n f^{\inf} - f_i^{\inf}}$ if losses $f_i$ are sampled uniformly at every iteration.
Finally, our result \eqref{mast:eq:ABC_noncvx_res} guarantees optimal $\cO(\varepsilon^{-4})$ complexity in a similar manner to Corollary \ref{mast:cor:sgd_nonconvex}.

We direct the reader to Appendix~\ref{appendix:chapter3_mast} for the results in convex and strongly convex cases.

\subsection{Discussion of related works}

\textbf{Compressed (sparse) model training.}
To our knowledge, the first work that analyzed convergence of (full batch) Gradient Descent with compressed iterates (model updates) is the work of \citet{khaled2019gradient}. They considered general unbiased compressors and, in the strongly convex setting, showed linear convergence to the irreducible neighborhood, depending on the norm of the model at the optimum $\|x^\star\|^2$. In addition, their analysis requires the variance of the compressor ($d/K$ for \rnd{\text{$K$}}) to be lower than the inverse condition number of the problem $\mu_f/L_f$, which basically means that the compressor has to be close to the identity mapping in practical settings. These results were extended using a modified method to distributed training with compressed model updates \citep{chraibi2019distributed,shulgin2022shifted}. \citet{lin2019dynamic} consider dynamic pruning with feedback inspired by the Error Feedback mechanism \citep{Seide2014,Alistarh-EF-NIPS2018,stich2019error}. Their result is similar to \citep{khaled2019gradient}, as the method also converges only to the irreducible neighborhood, the size of which is proportional to the norm of model weights. However, in \citet{lin2019dynamic} the norm of stochastic gradients is required to be uniformly upper-bounded, narrowing the class of losses. The partial \SGD\text{ }method proposed in \citet{mohtashami2022masked} allows general perturbations of the model weights where the gradient (additionally sparsified) is computed. Unfortunately, their analysis \citep{wang2022progfed} was recently shown to be vacuous \citep{szlendak2023understanding}.

\textbf{Dropout convergence analysis.}
Despite the wide empirical success of Dropout, there is limited theoretical understanding of its behavior and success. A few recent works \citep{mianjy2020convergence} suggest convergence analysis of this technique. However, these attempts typically focus on a certain class of models, such as shallow linear Neural Networks (NNs) \citep{senen2022asymptotic} or deep NNs with ReLU activations \citep{senen2020almost}.
Moreover, \citet{liao2022on} analyzed overparameterized single-hidden layer perceptron with a regression loss in the context of Dropout.
In contrast, our approach is model-agnostic and requires only mild assumptions like smoothness of the loss \eqref{mast:ass:l_f_smooth} (and convexity \eqref{mast:ass:mu_f-convex} for some of the results).

\begin{algorithm}[tb]
	\begin{algorithmic}[1]
		\caption{Distributed Double Sketched GD} \label{mast:alg:distributed_GD}
		\STATE \textbf{Parameters:} learning rate $\gamma > 0$; sketch distributions $\mathcal{D}_1, \ldots, \mathcal{D}_M$; initial model and shift $x^0, \shift \in \R^d$
		\FOR{$t = 0, 1, 2 \ldots$}
		\STATE Sample sketches: $\mS_i^t \sim \cD_i$
		\STATE
		Compute $y_i^t = \shift + \mS_i^t
		(x^t - \shift)$ for $i \in [M]$ and broadcast to corresponding nodes
		\FOR{$i = 1, \ldots, M$ in parallel}
		\STATE Compute local gradient: $\nabla f_i(y_i^t)$
		\STATE Send gradient $(\mS_i^t)^\top \nabla f_i(y_i^t)$ to the server
		\ENDFOR
		\STATE Aggregate messages and make a gradient-type step: $x^{t+1} = x^t - \frac{\gamma}{M} \sum_{i=1}^M (\mS_i^t)^\top \nabla f_i(y_i^t)$
		\ENDFOR
	\end{algorithmic}
\end{algorithm}

\section{Distributed setting} \label{mast:sec:distributed}

Consider $f$ being a finite sum over a number of machines, i.e., in the distributed setup:
\begin{equation} \label{mast:eq:fin_sketch_sum}
	\min\limits_{x\in \R^d} \frac{1}{M}\sum\limits_{i=1}^M f_{i, \cD_i} (x),
\end{equation}
where $f_{i, \cD_i} (x) \eqdef \Exp{f_{i, \mS_i}(x)} = \Exp{f_i(\shift + \mS_i (x-\shift))}$.
This setting is more general than problem \eqref{mast:eq:fin_sum} as every node $i$ has its own distribution of sketches $\mS_i \sim \cD_i$. Every machine performs local computations with a model of different size, which is crucial for scenarios with heterogeneous computing hardware.
The shift model $\shift$ is shared across all $f_{i, \mathcal{D}_i}$. We solve~\eqref{mast:eq:fin_sketch_sum} with the method
\begin{equation} \label{mast:alg:finsum_noncvx}
	x^{t+1} = x^t - \frac{\gamma}{M} \sum_{i=1}^{M}\left(\mS_i^t\right)^{\top} \nabla f_i(y^t_i),
\end{equation}
where $y^t_i = \shift + \mS_i^t (x^t - \shift)$.
Algorithm \ref{mast:alg:distributed_GD} describes the proposed approach in detail. Local gradients can be computed for sketched (sparse) model weights, which decreases the computational load on the computing nodes. Moreover, the local gradients are sketched as well, which brings communication efficiency in the case of sparsifiers $\mS_i$.

Recursion \eqref{mast:alg:finsum_noncvx} is closely related to the distributed Independent Subnetwork Training (IST) framework \citep{yuan2022distributed}. At every iteration of IST, a large model $x^t$ is decomposed into submodels $\mS^t_i x^t$ for independent computations (e.g. local training), which are then aggregated on the server to update the whole model.
IST efficiently combines model and data parallelism allowing training of huge models that cannot fit onto the single device. IST was shown to be very successful for a range of DL applications \citep{dun2022resist,wolfe2021gist}. \citet{shulgin2023towards} analyzed the convergence of IST for a quadratic model decomposed with permutation sketches \citep{szlendak2022permutation}, which satisfy Assumption~\ref{mast:ass:C}. They also showed that naively applying IST to standard distributed optimization problems (\eqref{mast:eq:fin_sketch_sum} for $\mS_i \equiv \mI$) results in a biased method and may not converge.

Resource-constrained Federated Learning (FL) \citep{kairouz2021advances,konecny2017federated,mcmahan2017communication} is another important practical scenario covered by Algorithm \ref{mast:alg:distributed_GD}. In cross-device FL, local computations are typically performed by edge devices (e.g.\, mobile phones), which have limited memory, computational power, and energy \citep{caldas2018expanding}. Thus, this forces practitioners to rely on smaller (potentially less capable) models or use techniques as Dropout in distributed setting \citep{alam2022fedrolex,bouacida2021adaptive,charles2022federated,chen2022fedobd,diao2020heterofl,horvath2021fjord,jiang2022model,qiu2022zerofl,wen2022federated,yang2022partial,dun2023efficient}. Despite extensive experimental studies of this problem setting, the principled theoretical understanding remains minimal. Our work can be considered as the first rigorous analysis in the most general setting without restrictive assumptions.
\begin{theorem} \label{mast:thm:distributed_ncvx}
	Assume that each $f_i$ is $L_{f_i}$-smooth \eqref{mast:ass:l_f_smooth} and sketches $\mS_i$ satisfy Assumption \ref{mast:ass:C}. Let $D_{\max} \eqdef \max_i \br{L_{f_i}^2 L_{\cD_i} L_{\mS_i}^{\max}}$.
	Then, for $\gamma \leq 1 / \sqrt{D_{\mathrm{\max}} T}$, the iterates of Algorithm~\ref{mast:alg:distributed_GD} satisfy
	\begin{equation*}
			\min_{0 \leq t < T} \Exp{\left\|\nabla \tf(x^t)\right\|^2} \leq
			\frac{3 \br{\tf(x^0) - \tf^{\inf}}}{\gamma T} + \gamma D_{\max} \biggl(\tf^{\inf} - \frac{1}{M} \sum_{i=1}^{M} f^{\inf}_i \biggr).
	\end{equation*}
\end{theorem}

This theorem resembles our previous Theorem \ref{mast:thm:sgd_nonconvex} in the single-node setting. Namely, Algorithm~\ref{mast:alg:distributed_GD}
reaches a stationary point at rate $\cO(1/\sqrt{T})$. But due to the distributed setup, convergence depends on $D_{\max}$ expressed as the \textit{maximum} product of local smoothness and constants related to the sketches' properties. Thus, clients with more aggressive sparsification may slow down the method, given the same local smoothness constant $L_{f_i}$. Yet, \say{easier} local problems (with smaller $L_{f_i}$) can allow the use of \say{harsher} sparsifiers (with larger $L_{\cD_i} L_{\mS_i}^{\max}$) without negatively affecting the convergence.

\subsection{Discussion of related works}

A notable distinction between our result and the theory of methods like Distributed Compressed Gradient Descent \citep{khirirat2018distributed} lies in the second convergence term of Theorem \ref{mast:thm:distributed_ncvx}. Instead of relying on the variance of local gradients at the optimum, given by $\delta^2 = \frac{1}{n} \sum_{i=1}^n \|\nabla f_i (x^\star)\|^2$, our result depends on the average difference between the lower bounds of the global and local losses: $\tf^{\inf} - f^{\inf}_i$. This term can be interpreted as a measure of heterogeneity within a distributed setting \citep{khaled2020tighter}.
Furthermore, our findings may provide a better explanation of the empirical efficacy of distributed methods. Namely, $\delta^2$ is less likely to be equal to zero, unlike our term, which can be very small when models are over-parameterized allowing local losses to be minimized to zero. %

\citet{yuan2022distributed} analyzed convergence of Independent Subnetwork Training in their original work using the framework of \citet{khaled2019gradient}. Their analysis was performed in the single-node setting and required additional assumptions on the gradient estimator, which were recently shown to be problematic \citep{shulgin2023towards}.
In a federated setting, \citet{zhou2022convergence} suggested a method that combines model pruning with local compressed Gradient Descent steps. They provided non-convex convergence analysis relying on bounded stochastic gradient assumption, which results in \say{pathological} bounds \citep{khaled2020tighter} for a heterogeneous distributed case.

\section{Experiments} \label{mast:sec:experiments}

To empirically validate our theoretical framework and its implications, we focus on carefully controlled settings that satisfy the assumptions of our work. Specifically, we consider an $\ell_2$-regularized logistic regression optimization problem with the \textit{a5a} dataset from the LibSVM repository~\citep{chang2011libsvm}.
See Appendix \ref{mast:apdx:experiments} for further details and Appendix \ref{mast:sec:add_experiments} for more results on other methods and sketches. %

\begin{figure}[t]
\centering
\includegraphics[width=0.62\textwidth]{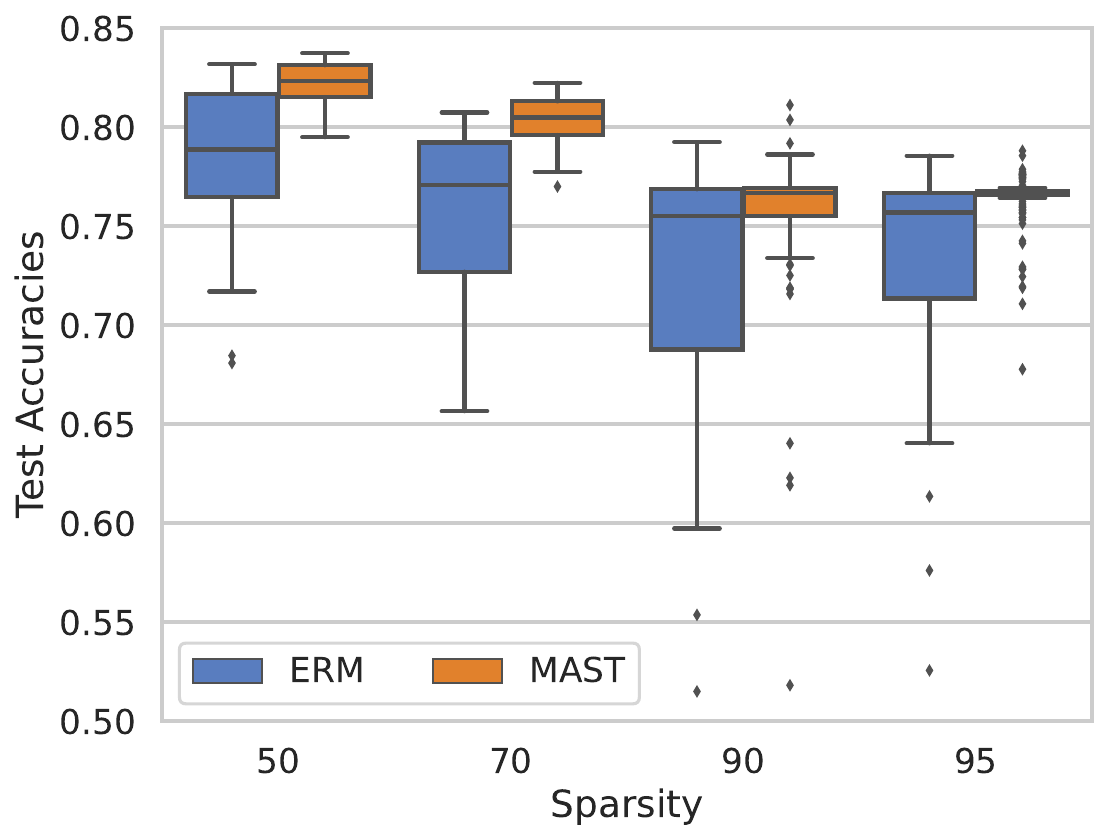}%
\caption{Distributions of test accuracies of sparsified solutions for the ERM formulation \eqref{mast:eq:main} and MAST problem \eqref{mast:eq:pretrained_compressed_problem}. ``Sparsity'' corresponds to the percentage of zeroed weights.}
\label{mast:fig:test_acc_box}
\end{figure}

In Figure \ref{mast:fig:test_acc_box} we compare the test accuracy of sparsified solutions for the standard (ERM) problem \eqref{mast:eq:main} and introduced MAST formulation \eqref{mast:eq:pretrained_compressed_problem}.
Visualization is performed using the \href{https://seaborn.pydata.org/archive/0.11/generated/seaborn.boxplot.html#seaborn.boxplot}{boxplot} method from Seaborn (version 0.11.0) library \citep{Waskom2021} with default parameters.
For ERM, we find the exact (up to machine precision) optimum, which is subsequently used for the accuracy evaluation. For the MAST optimization problem, we run $\mathsf{DSGD}$ with exact sketched gradient $\nabla \tf$ for every sparsity level.
After the ERM and MAST models ($x^T$) are obtained, we apply partition sketches to model weights and evaluate the test accuracy of the sparsified solutions ($\mS x^T$).

Figure \ref{mast:fig:test_acc_box} reveals that models obtained using the MAST approach exhibit greater robustness to random pruning compared to their ERM counterparts given the same sparsity. Moreover, the ERM model suffers from greater accuracy variability, while the median test accuracies of the MAST models are markedly higher.
Increasing the sparsity leads to the degradation of the performance of both approaches.

\begin{figure*}[h]
\centering
\subfigure[High sparsity]{%
\label{mast:fig:1a_}%
\includegraphics[width=0.333\linewidth]{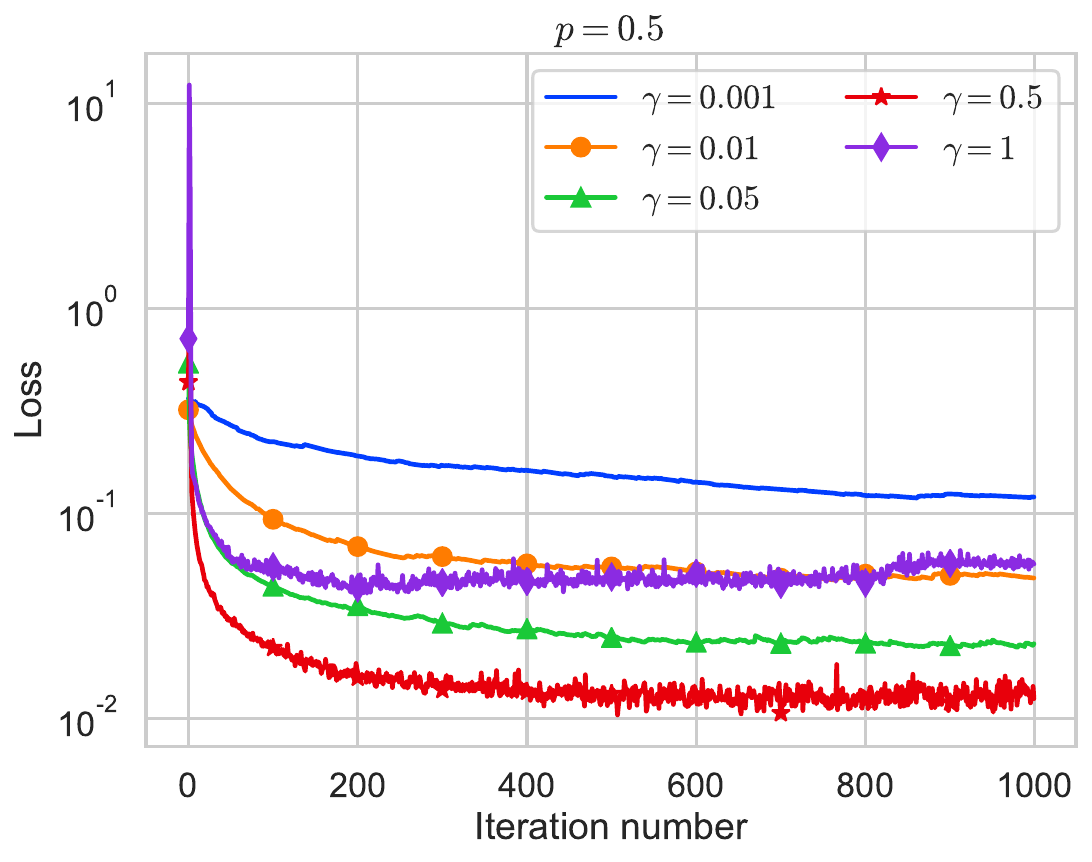}}%
\hfill
\subfigure[Mid sparsity]{%
\label{mast:fig:1b_}%
\includegraphics[width=0.333\linewidth]{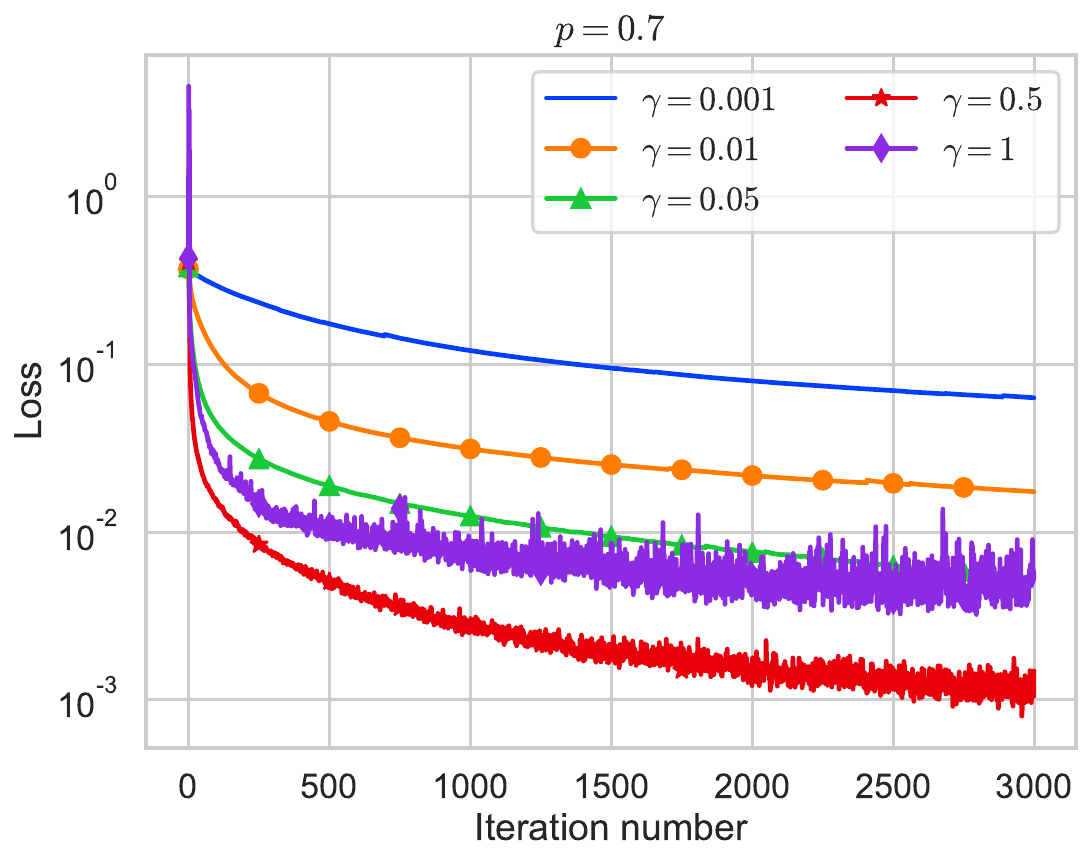}}%
\hfill
\subfigure[Low sparsity]{%
\label{mast:fig:3c_}%
\includegraphics[width=0.333\linewidth]{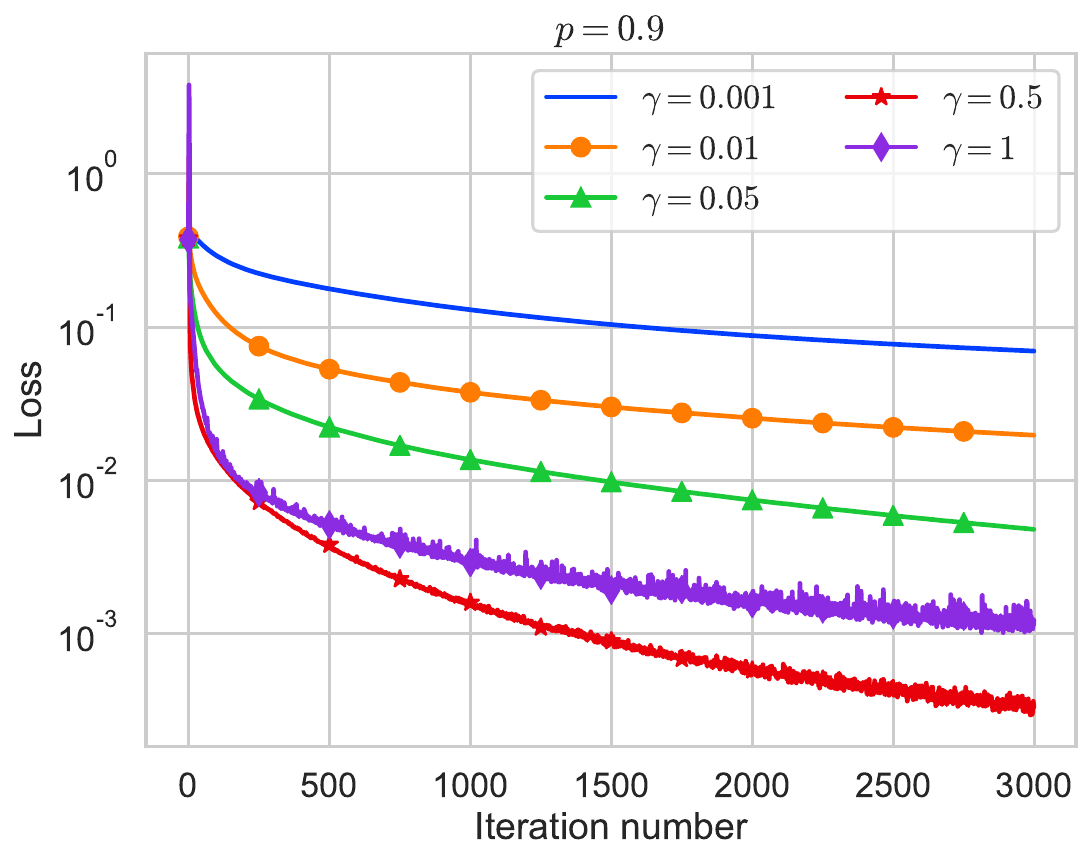}}%

\caption[Performance of Algorithm \ref{mast:alg:distributed_GD} with Bernoulli sketches]{
 Performance of Algorithm \ref{mast:alg:distributed_GD} with Bernoulli sketches \eqref{mast:eq:ind_sparse} on standard loss \eqref{mast:eq:fin_sketch_sum} (for $\mS_i \equiv \mI$).
}
\label{mast:fig:lr_p_main}
\end{figure*}

\paragraph{Neural network results.} %
Figure \ref{mast:fig:lr_p_main} presents a subset of our distributed deep learning results (full details are provided in Appendix \ref{mast:apdx:nn_exp}). Our experimental setup closely follows that of \citet{liao2022on}, which is based on the ResNet-50 model \citep{he2016deep}.
We study the Algorithm \ref{mast:alg:distributed_GD} with Bernoulli sketches \eqref{mast:eq:ind_sparse} and $p_i \equiv p$ for the standard (ERM) loss \eqref{mast:eq:fin_sketch_sum} with $\mS_i \equiv \mI$.

Figure \ref{mast:fig:lr_p_main} illustrates the impact of sparsity level ($p$) and step size ($\gamma$) on the method's performance. Across all sparsity levels, we observe an optimal \say{sweet spot} ($\gamma=0.5$) for the step size, beyond which increasing $\gamma$ results in slower convergence. Crucially, there exists a nuanced interplay of $\gamma$ with sparsity level. Namely, at $\gamma=1$, convergence slows down for $p=0.9$, while for $p=0.7$, performance degrades due to high variance, eventually being outperformed by a smaller step size.

Notably, high sparsity ($p=0.5$) leads to a quick stagnation of the loss even with a small step size $\gamma = 0.01$, unlike for $p \in\{0.7, 0.9\}$.
Remarkably, Figure \ref{mast:fig:1a_} illustrates that an excessively large step size ($\gamma=1$) may even lead to divergence of the method.
This can indicate that high sparsity significantly alters the minimized loss, confirming that Sparse/Dropout training indeed optimizes a formulation distinct from standard ERM.
In general, larger step sizes and more aggressive sparsification (lower $p$) result in increased loss variance, aligning with our theoretical predictions from Sections \ref{mast:sec:sketches} and \ref{mast:sec:convergence}.

One of the key practical insights derived from our theoretical analysis is that the step size $\gamma$ (learning rate) must be decreased for sparse optimization and training with Dropout. Our results demonstrate that this insight is applicable not only to convex models (Figure \ref{mast:fig:mast_loss_perm}) but also to a broader range of non-convex neural networks.

\section{Conclusions and future work} \label{mast:sec:conclusion}

In this work, a novel theoretical framework for sketched model learning was introduced. We rigorously formalized a new optimization paradigm that captures practical scenarios like Dropout and Sparse training.
Efficient optimization algorithms tailored to the proposed formulation were developed and analyzed in multiple settings. We expanded this methodology to distributed environments, encompassing areas such as IST and Federated Learning, underscoring its broad applicability.

In future research, it would be interesting to expand the class of linear matrix sketches to encompass other compression techniques, particularly those exhibiting conic variance property (contractive compressors). Such an extension might offer insights into methods such as (magnitude-based) pruning and quantized training. Nevertheless, a potential challenge to be considered is the non-differentiability of such compression techniques.

\endgroup

\chapter{Adaptive Clipping for Differentially Private SGD}
\label{chapter4:adaptive_clipping}

\begingroup
\input{KAUST_Thesis/imported/04_adaptive_clipping/import_setup}

\section*{Abstract}
Stochastic Gradient Descent (\sgd) with gradient clipping is a powerful technique for enabling differentially private optimization. Although prior work has extensively investigated clipping with a constant threshold, private training remains highly sensitive to threshold selection, which can be expensive or even infeasible to tune. This sensitivity motivates the development of adaptive approaches, such as quantile clipping, which have demonstrated empirical success but lack a solid theoretical understanding. This paper provides the first comprehensive convergence analysis of \sgd with quantile clipping (\alg). We demonstrate that \alg suffers from a bias problem similar to constant-threshold clipped \sgd but show how this can be mitigated through a carefully designed quantile and step size schedule. Our analysis reveals crucial relationships between quantile selection, step size, and convergence behavior, providing practical guidelines for parameter selection. We extend these results to differentially private optimization, establishing the first theoretical guarantees for \algname{DP-QC-SGD}.
Our findings provide theoretical foundations for a widely used adaptive clipping heuristic and highlight open avenues for future research.

\begin{quote}
\small
\noindent\textit{This chapter is based on:}\\
Egor Shulgin and Peter Richt{\'a}rik, ``On the Convergence of DP-SGD with Adaptive Clipping,'' \emph{NeurIPS Workshop on Optimization for Machine Learning}, 2024.
\nocite{shulgin2024convergence}
\end{quote}

\section{Introduction}
It is hard to imagine the success of modern machine learning without effective optimization, the cornerstone of which is stochastic gradient descent (\sgd)-type methods \citep{robbins1951stochastic,bottou2018optimization}.
However, \sgd is not perfect, particularly in the context of deep learning.
Efficient neural network training often requires modifications of \sgd to stabilize optimization. For example, the problem of exploding gradients \citep{pascanu2013difficulty} is often addressed using a \emph{gradient-clipping} operator, which rescales the input vector's norm if it exceeds a certain threshold.
Moreover, gradient clipping is vital in privacy-preserving machine learning \citep{dwork2006calibrating,dwork2014algorithmic}.
Rigorous differential privacy guarantees are usually established by relying on the Gaussian mechanism \citep{dwork2014algorithmic}, which requires bounded sensitivity to control the amount of noise added.
The most commonly used Differentially Private SGD (\algname{DP-SGD}) method enforces such a bound by clipping the per-example gradients \citep{abadi2016deep}.

Clipped \sgd was shown to be superior to vanilla \sgd for minimizing generalized smooth functions \citep{zhang2020why} and when stochastic gradient noise is heavy-tailed \citep{zhang2020adaptive}.
However, the effectiveness of gradient clipping hinges critically on the choice of the clipping threshold, denoted by $\tau$. This introduces an additional hyperparameter that requires careful tuning, a challenge especially pronounced in private optimization settings
where performance can be highly sensitive to this threshold \citep{kurakin2022toward,bu2024automatic}. Furthermore, each training run incurs an additional privacy loss, making extensive hyperparameter search prohibitively expensive from a privacy perspective \citep{papernot2021hyperparameter}.

\begin{figure*}[!hb]
\centering
\includegraphics[width=\textwidth]{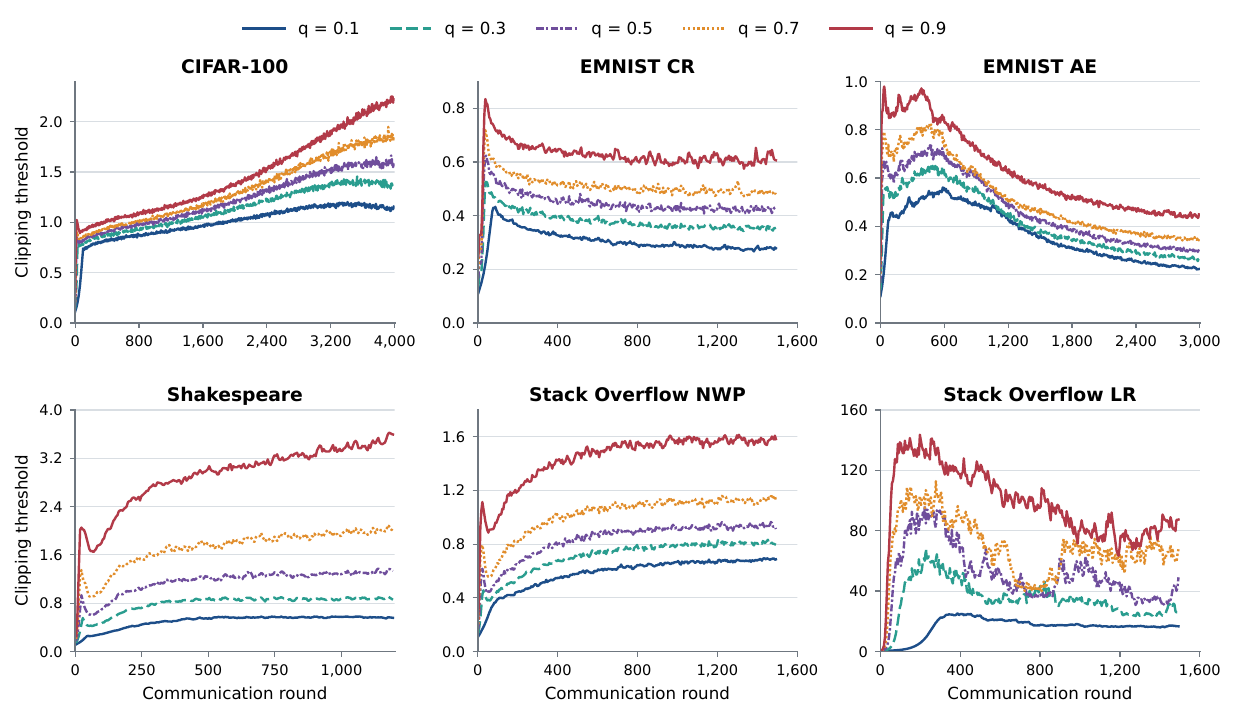}
\caption{Adaptive clipping thresholds for five target quantiles on six federated learning tasks without privacy noise. The trajectories vary substantially across tasks and may increase, decrease, or change nonmonotonically. The curves were redrawn for this thesis from the vector trajectories in Figure~4 of \citet{andrew2021differentially}; the original clipping norms were estimated using geometric updates with $\eta_C=0.2$ and $C_0=0.1$.}
    \label{adaptive:fig:quantile_clip}
\end{figure*}

\paragraph{Adaptive clipping.} The problem described above has been addressed by \citet{andrew2021differentially} in the setting of private Federated Learning \citep{konecny2017federated,mcmahan2017communication,mcmahan2018learning,kairouz2021advances}. Specifically, \citet{andrew2021differentially} proposed to adaptively select the clipping threshold based on the distribution of gradient norms (or updates) of the participating clients. Their method (named \say{adaptive quantile clipping}) efficiently estimates a quantile and applies it as the clipping threshold. Crucially, privacy analysis by \citet{andrew2021differentially} revealed that this adaptive approach incurs only negligible additional privacy loss.
Extensive empirical evaluations have shown that quantile clipping is competitive with and often outperforms carefully tuned constant clipping baselines \citep{andrew2021differentially}. This adaptive strategy offers the additional benefit of adjusting to the evolving gradient distribution throughout the federated optimization process. This adaptability is particularly valuable given the significant variability observed across different machine learning tasks and datasets (see Figure \ref{adaptive:fig:quantile_clip}), suggesting that a uniform clipping schedule may be suboptimal.
The success of the adaptive clipping technique has led to its widespread adoption for multiple applications \citep{song2022flair,xu2023federated}, even beyond privacy-constrained settings \citep{charles2021large}, and implementation in Federated Learning libraries \citep{TensorFlow_Federated2018,granqvist2024pfl}.

However, despite its practical success and adoption, the theoretical properties of adaptive quantile clipping remain largely unexplored.
This research gap motivates our current study. We aim to provide a comprehensive optimization analysis of \sgd with quantile clipping (\alg in short).
In particular, by building upon recent work \citep{merad2023robust}, we demonstrate that \sgd with quantile clipping suffers from a bias problem that can prevent convergence, similar to that observed in constant-threshold clipped \sgd. We design a quantile and step size schedule that effectively eliminates the identified bias problem. Our analysis reveals the crucial relationship between the chosen quantile value and the step size in \alg, providing insights into how this interplay affects convergence. Finally, we extend the results to differentially private optimization.

\section{Problem and assumptions}

We consider a classical stochastic optimization problem
\begin{equation} \label{adaptive:eq:problem}
	\min_{x \in \RR^d} \Bigl[ f(x) \eqdef \EE_{\xi\sim \cD}[f_{\xi}(x)]\Bigr],
\end{equation}
where $f_\xi: \RR^d \to \RR$ is the loss of a machine-learning model parameterized by $x \in \RR^d$ on data point $\xi$. Thus, $f$ is the expected loss over the data distribution $\xi \sim \cD$.
We rely on standard assumptions for non-convex stochastic optimization to enable optimization analysis.
\begin{assumption}[Stochastic gradient] \label{adaptive:ass:b_variance}
The stochastic gradient estimator is unbiased, $\Exp{\nabla f_{\xi}(x)} = \nabla f(x)$, and has a \textbf{bounded $q$-th moment} for $q \in (1, 2]$:
\begin{equation} \label{adaptive:eq:bound_var}
    \br{\EE_{\xi \sim \cD} \left[ \left\| \nabla f_{\xi}(x) - \nabla f(x) \right\|^q \right]}^{1/q} \leq \sigma_q, \qquad \forall x\in \R^d.
\end{equation}
\end{assumption}
Condition \eqref{adaptive:eq:bound_var} is usually referred to as heavy-tailed noise \citep{zhang2020adaptive} for $q \in (1, 2)$. For $q=2$, it recovers the classical bounded-variance assumption \citep{nemirovski2009robust,ghadimi2012optimal}.

\begin{assumption}[Function] \label{adaptive:ass:l_smooth}
    The function $ f $ is differentiable and $ L $-\textbf{smooth}, meaning there exists $ L > 0 $ such that
    \begin{equation}
        f(x+u) \leq f(x) + \langle \nabla f(x), u \rangle + \frac{L}{2} \|u\|^2, \qquad \forall x,u \in \R^d.
    \end{equation}
    Additionally, $ f $ is lower-bounded by $ f^{\inf} \in \mathbb{R} $.
\end{assumption}

\section{Stochastic gradient descent with quantile clipping}

We consider the following stochastic-gradient-descent-type method with step size $\gamma_t > 0$:
\begin{equation} \label{adaptive:eq:sgd_qc}
    x^{t+1} = x^t - \gamma_t g^t,
\end{equation}
where $g^t = g(x^t)$ has the form of a clipped stochastic gradient estimator $g^t = \alpha_{\xi^t}(x^t) \nabla f_{\xi^t}(x^t)$ for
\begin{equation} \label{adaptive:eq:alpha}
    \alpha_{\xi}(x) = \min \cbr{1, \frac{\tau(x)}{\norm{\nabla f_{\xi}(x)}}},
\end{equation}
and $\tau(x)$ is $p$-th quantile (of random variable $\norm{\gf(x)}$) clipping threshold, defined as
\begin{equation} \label{adaptive:eq:quantile}
    \mathrm{Prob}(\norm{\gf(x)} \leq \tau(x)) = p.
\end{equation}
We call the described algorithm SGD with Quantile Clipping (\alg). Note that clipped \sgd is a special case of \alg for $\tau(x) \equiv \tau$.
This algorithm was originally introduced in the seminal work by \citet{merad2023robust} in the context of robust optimization with a corrupted oracle. They analyzed it as a Markov chain, while we are interested in optimization properties with a focus on differentially private settings.

\subsection{Preliminaries}

At first, we present some crucial properties of the described gradient estimator $g(x^t)$ and clipping threshold $\tau(x)$ needed for convergence analysis.

\begin{lemma}[\citet{merad2023robust}] \label{adaptive:lemma:key_lemma}
Assume that stochastic gradient estimator $\gf(x)$ satisfies Assumption \ref{adaptive:ass:b_variance}, $\alpha_{\xi}(x)$ is chosen as \eqref{adaptive:eq:alpha}, and $p$-th quantile clipping threshold $\tau(x)$ satisfies \eqref{adaptive:eq:quantile}. Then for all $x \in \RR^d$,
\begin{equation} \label{adaptive:lemma:tau}
    \tau(x) \leq \norm{\nabla f(x)} + \sigma_q \br{1-p}^{-1/q},
\end{equation}
\begin{equation} \label{adaptive:lemma:bias_}
    \norm{\Exp{\alpha_{\xi}(x) \gf(x)} - \oalpha(x) \nabla f(x)} \leq \sigma_q \br{1-p}^{1-1/q},
\end{equation}
where $\oalpha(x) \eqdef \Exp{\alpha_{\xi}(x) }$.
\end{lemma}
Note that \eqref{adaptive:lemma:bias_} is different from typical results \citep{zhang2020adaptive,koloskova2023revisiting} characterizing the bias of gradient clipping due to the additional multiplier $\oalpha(x)$ before the gradient $\nabla f(x)$. This significantly impacts the convergence analysis of \alg and makes it different from Clipped \sgd.

\begin{lemma}[Second moment of QC estimator] \label{adaptive:lemma:second_moment_qc}
Assume Assumption~\ref{adaptive:ass:b_variance} holds for some $q\in(1,2]$ and let $h\eqdef 1-p$.
Let $g(x)=\alpha_{\xi}(x)\nabla f_{\xi}(x)$ with $\alpha_{\xi}(x)$ defined in \eqref{adaptive:eq:alpha}, and let $\tau(x)$ be the $p$-quantile threshold \eqref{adaptive:eq:quantile}.
Then, for all $x\in\RR^d$,
\begin{equation}\label{adaptive:eq:second_moment_qc}
    \Exp{\|g(x)\|^2} \;\le\; 6\,\|\nabla f(x)\|^2 \;+\; 8\,\sigma_q^2\, h^{-(2-q)/q}.
\end{equation}
In particular, when $q=2$ the exponent satisfies $(2-q)/q=0$ and \eqref{adaptive:eq:second_moment_qc} becomes $\Exp{\|g(x)\|^2}\le 6\|\nabla f(x)\|^2+8\sigma_2^2$ (no dependence on $p$).
\end{lemma}

\subsection{Convergence analysis}

Our analysis relies on the following recursion.
\begin{lemma} \label{adaptive:lemma:recursion}
Suppose $f$ is $L$-smooth (Assumption~\ref{adaptive:ass:l_smooth}) and stochastic gradients satisfy Assumption~\ref{adaptive:ass:b_variance} for some $q\in(1,2]$.
Then, for any $\beta>0$, the iterates of \alg \eqref{adaptive:eq:sgd_qc} satisfy
\begin{align} \label{adaptive:eq:core_recursion}
    \Exp{f(x^{t+1}) \mid x^t}
    &\leq f(x^t)
    - \gamma_t \br{\oalpha(x^t) - \beta/2 - 3L\gamma_t}\,\sqn{\nabla f(x^t)}
    \\
    &{}
    + \frac{\gamma_t}{2}\beta^{-1}\sigma_q^2 \br{1-p}^{2-2/q}
    + 4L\gamma_t^2 \sigma_q^2 \br{1-p}^{-(2-q)/q}. \notag
\end{align}
\end{lemma}

Equipped with Lemma \ref{adaptive:lemma:recursion}, we provide a general convergence result for \alg.

\begin{theorem}[General case] \label{adaptive:thm:non_cvx_gen}
Suppose $f$ is $L$-smooth (Assumption~\ref{adaptive:ass:l_smooth}) and stochastic gradients satisfy Assumption~\ref{adaptive:ass:b_variance} for some $q\in(1,2]$.
Let $h\eqdef 1-p$ and $\Gamma_T\eqdef \sum_{t=0}^{T-1}\gamma_t$.
If the step sizes satisfy
\begin{equation} \label{adaptive:eq:step_size}
    0 < \gamma_t \leq \frac{2p - \beta - c}{6L},
\end{equation}
where $c,\beta\in(0,1)$ and $\beta + c \le 2p$, then the iterates of \alg satisfy
\begin{equation} \label{adaptive:eq:gen_res}
    \frac{c}{\Gamma_T} \sum_{t=0}^{T-1} \gamma_t \Exp{\sqn{\nabla f(x^t)}}
    \leq
    \frac{2\br{f(x^0) - f^{\inf}}}{\Gamma_T}
    + \frac{\sigma_q^2}{\Gamma_T}\sum_{t=0}^{T-1}\br{\beta^{-1}\gamma_t h^{2-2/q} + 8L\gamma_t^2 h^{-(2-q)/q}}.
\end{equation}
\end{theorem}
Theorem \ref{adaptive:thm:non_cvx_gen} indicates that \alg can find an approximate stationary point. Importantly, more aggressive clipping (smaller $p$) requires decreasing the step size $\gamma_t$ according to \eqref{adaptive:eq:step_size}.
However, overall performance heavily depends on $h$ and the sequence $\gamma_t$. Next, we discuss how two possible choices affect convergence.

\begin{corollary}[Constant parameters] \label{adaptive:cor:const_res}
Under the conditions of Theorem~\ref{adaptive:thm:non_cvx_gen}, if $\gamma_t\equiv \gamma$ and $p_t\equiv p$, then
\begin{equation} \label{adaptive:eq:constant_res}
    \frac{c}{T}\sum_{t=0}^{T-1}\Exp{\sqn{\nabla f(x^t)}}
    \leq
    \underbrace{\frac{2\br{f(x^0) - f^{\inf}}}{\gamma T}}_{\circledOne}
    + \underbrace{8\gamma L\sigma_q^2\,h^{-(2-q)/q}}_{\circledTwo}
    + \underbrace{\beta^{-1}\sigma_q^2\,h^{2-2/q}}_{\circledThree}.
\end{equation}
\end{corollary}

The obtained result shares similarities with classical \sgd convergence guarantees (under bounded
stochastic gradient variance assumption) as it indicates convergence to a neighborhood of a stationary
point. Namely, the first term \circledOne in the upper bound \eqref{adaptive:eq:constant_res} decreases with rate
$\mathcal{O}(1/T)$.

The second term \circledTwo scales as $h^{-(2-q)/q}$, where $h=1-p$. In particular, for the bounded-variance
case $q=2$ this term is independent of $h$, while for $q\in(1,2)$ the exponent $(2-q)/q>0$ implies that
more aggressive clipping (smaller $p$, larger $h$) \emph{decreases} this stochastic neighborhood term.
In contrast, the third term \circledThree scales as $h^{2-2/q}$ and therefore \emph{increases} with more
aggressive clipping. Hence, for fixed $(p,\gamma)$ the method exhibits a variance--bias trade-off:
stronger clipping reduces the stochastic term but increases the irreducible bias term.

Moreover, exact convergence to a stationary point cannot be guaranteed for any constant stepsize
$\gamma$, as the bias term \circledThree in \eqref{adaptive:eq:constant_res} cannot be driven to zero without letting
$h\to 0$ (i.e., $p\to 1$). This is in contrast to standard \sgd, where one can achieve
$\mathcal{O}(1/\sqrt{T})$ by choosing $\gamma=\mathcal{O}(1/\sqrt{T})$.

In addition, larger $\beta$ decreases the bias neighborhood \circledThree but requires choosing a smaller
stepsize via \eqref{adaptive:eq:step_size}, roughly $\gamma \lesssim (p-\beta/2)/L$ (up to constants), leading to
slower decay of the term \circledOne.

\paragraph{Time-varying parameters.}
More generally, one can allow iteration-dependent quantiles $p_t = 1-h_t$ by defining $\tau_t(x^t)$ as the $p_t$-quantile of $\|\nabla f_{\xi}(x^t)\|$ at iteration $t$.
Then the proof of Theorem~\ref{adaptive:thm:non_cvx_gen} yields the bound
\begin{equation}\label{adaptive:eq:gen_res_tv}
    \frac{c}{\Gamma_T} \sum_{t=0}^{T-1} \gamma_t \Exp{\sqn{\nabla f(x^t)}}
    \leq
    \frac{2\br{f(x^0) - f^{\inf}}}{\Gamma_T}
    + \frac{\sigma_q^2}{\Gamma_T}\sum_{t=0}^{T-1}\br{\beta^{-1}\gamma_t h_t^{2-2/q} + 8L\gamma_t^2 h_t^{-(2-q)/q}},
\end{equation}
provided $0<\gamma_t \le (2p_t-\beta-c)/(6L)$ for all $t$.

For $\gamma_t = \tilde{\cO}(t^{\theta-1})$ and $h_t=\tilde{\cO}(t^{\nu})$ we have $\Gamma_T=\tilde{\cO}(T^\theta)$ and the right-hand side in \eqref{adaptive:eq:gen_res_tv} is of order
\begin{equation}\label{adaptive:eq:rate}
    \tilde{\cO}\!\br{T^{-\theta} + T^{\theta-1-\nu(2-q)/q} + T^{2\nu(1-1/q)}}.
\end{equation}
Balancing the three terms yields the choice
$\theta = 2(q-1)/(3q-2), \nu = -q/(3q-2)$,
resulting in the rate $\tilde{\cO}\!\bigl(T^{-2(q-1)/(3q-2)}\bigr)$.
In particular, for the bounded-variance case $q=2$ we obtain $\theta=1/2$ and $\nu=-1/2$, i.e.,
$
\gamma_t=\tilde{\cO}(t^{-1/2})
$
and
$
p_t = 1-\tilde{\cO}(t^{-1/2}),
$
which gives the standard $\tilde{\cO}(T^{-1/2})$ convergence behavior.

\paragraph{Iteration and sample complexity; relation to minimax lower bounds.}
Let $\Delta \eqdef f(x^0)-f^{\inf}$ and fix a target accuracy $\varepsilon>0$.
Corollary~\ref{adaptive:cor:const_res} implies that for constant parameters $(\gamma,p)$ (with $h\eqdef 1-p$),
\[
\frac{1}{T}\sum_{t=0}^{T-1}\EE\|\nabla f(x^t)\|^2
\;\leq\;
\cO\!\left(\frac{\Delta}{\gamma T} + \gamma L\sigma_q^2 h^{-(2-q)/q} + \sigma_q^2 h^{2-2/q}\right).
\]
In particular, the last term yields an irreducible bias floor for any fixed $p\in(0,1)$.
To obtain an $\varepsilon$-stationary point guarantee, one may tune the quantile via
$h \asymp (\varepsilon/\sigma_q)^{q/(q-1)}$ (equivalently $p=1-h$) so that
$\sigma_q^2 h^{2-2/q} \asymp \varepsilon^2$, and then choose
$\gamma \asymp \varepsilon^{q/(q-1)}/(L\sigma_q^{q/(q-1)})$.
This yields the iteration complexity
\[
T = \tilde{\cO}\!\left(\frac{L\Delta}{\varepsilon^2}\left(\frac{\sigma_q}{\varepsilon}\right)^{q/(q-1)}\right)
= \tilde{\cO}\!\left(\frac{L\Delta\,\sigma_q^{q/(q-1)}}{\varepsilon^{(3q-2)/(q-1)}}\right),
\]
and sample complexity $N=BT$ for mini-batch size $B$.
This dependence matches the known lower bounds for stochastic nonconvex optimization under
bounded $q$-th central moment (p--BCM) noise and standard smoothness (up to log factors).

\subsection{Comparison to fixed clipping}

The latest analysis on \sgd with constant clipping ($\tau(x) \equiv \tau$) we are aware of is due to \citet{koloskova2023revisiting}. Their (simplified) result indicates that with a proper step size choice, for $q=2, \sigma_q = \sigma$, and for $L$-smooth function, the expected squared gradient norm is upper bounded by
\begin{equation} \label{adaptive:eq:const_clip}
\cO \br{\left(\frac{F^0}{\gamma T \tau}\right)^2 + \frac{F^0}{\gamma T} + \gamma L \sigma^2 + \min \br{\sigma^2, \frac{\sigma^4}{\tau^2}}},
\end{equation}
where $F^0 \eqdef f(x^0) - f^{\inf}$.
This result is fundamentally similar to quantile clipping \eqref{adaptive:eq:constant_res} as the last term is also irreducible via decreasing step size $\gamma$.
However, upper bound \eqref{adaptive:eq:const_clip} can be made arbitrarily small by setting the step size to $\gamma = \tilde{\cO}(T^{-1/2})$ and the clipping threshold to $\tau = \tilde{\cO}(T^\lambda), \lambda \in (0, 1/4)$. While increasing $\tau$ can solve the problem in theory, it is unsatisfactory from a practical perspective. As shown by \citet{andrew2021differentially}, the distribution of update norms (or pseudo-gradient norms \citep{reddi2021adaptive}) may behave very differently across federated training tasks. Figure \ref{adaptive:fig:quantile_clip} shows that these norms may, in fact, increase during optimization.
In addition, for differentially private settings, bigger $\tau$ requires adding larger noise at every iteration, resulting in the degraded utility of the model \citep{bu2024automatic}.

\subsection{Bias due to clipping}
In the discussion after Corollary~\ref{adaptive:cor:const_res}, we mentioned that for any non-trivial fixed quantile $p\in(0,1)$, exact convergence to a stationary point cannot, in general, be guaranteed.
To show that this is not merely an artifact of the analysis, we give simple examples where \alg admits fixed points $x^\dagger$ satisfying $\Exp{g(x^\dagger)}=0$ while $\nabla f(x^\dagger)\neq 0$.

\begin{example} \label{adaptive:ex:qc_bias_general_quantile}
Fix $r>0$ and $\omega\in(0,1)$ and define the stochastic objective
\begin{equation}\label{adaptive:eq:example_f}
f_{\xi}(x)=\frac{1}{2}
\left\{\begin{array}{cc}
    (x+r)^2, &\text{with probability } \omega,\\
    x^2,     &\text{with probability } 1-\omega.
\end{array}\right.
\end{equation}
Then $\nabla f(x) = \Exp{\nabla f_{\xi}(x)} = x + r\omega$, and the minimizer of $f(x)=\Exp{f_{\xi}(x)}$ is $x^\star = -r\omega$.

Consider any $x\ge 0$ and choose the quantile level $p = 1-\omega$.
Since $\nabla f_{\xi}(x)\in\{x+r,\,x\}$ and $x < x+r$, we have
\[
\Prob\bigl(\,|\nabla f_{\xi}(x)| \le x\,\bigr) = 1-\omega = p,
\]
hence the $p$-quantile of $|\nabla f_{\xi}(x)|$ equals $\tau(x)=x$ (which matches the quantile definition \eqref{adaptive:eq:quantile}).

Since $\tau(x)=x$ and $x\ge 0$, for both outcomes we have
\[
g(x)=\min\{|\nabla f_\xi(x)|,\tau(x)\}\,\mathrm{sign}(\nabla f_\xi(x))
      =\min\{|\nabla f_\xi(x)|,x\}=x \quad \text{a.s.}
\]
so $g(x)=x$ almost surely and $\Exp{g(x)}=x$.
Therefore, $x^\dagger=0$ is the unique fixed point of the mean dynamics, while
\[
\nabla f(x^\dagger)=\nabla f(0)=r\omega \neq 0.
\]
In particular, \alg can converge to a non-stationary point (e.g., if initialized at $x^0\ge 0$ and using a stable stepsize schedule).
\end{example}

\begin{example}[Median clipping] \label{adaptive:ex:qc_bias_median}
In the setting of Example~\ref{adaptive:ex:qc_bias_general_quantile}, take $\omega=\tfrac12$ and choose the median $p=\tfrac12$.
For any $x\ge 0$, we again have $\tau(x)=x$ because $\Prob(|\nabla f_{\xi}(x)|\le x)=\tfrac12$.
Hence $g(x)=x$ almost surely, the fixed point is $x^\dagger=0$, but the minimizer is $x^\star=-r/2$ and $\nabla f(0)=r/2\neq 0$.
Thus, the bias phenomenon persists even for median quantile clipping.
\end{example}

\section{Differentially private extension} \label{adaptive:sec:dp_sgd}
The most standard way to make clipped \sgd $\br{\epsilon, \delta}$-Differentially Private (DP) is by adding isotropic Gaussian noise with variance proportional to the clipping threshold \citep{abadi2016deep} (along with subsampling/mini-batching). This approach applied to \alg results in the following update (\algname{DP-QC-SGD} for short)
\begin{equation} \label{adaptive:eq:dp_sgd_qc}
    x^{t+1} = x^t - \gamma_t \frac{1}{B} \br{\sum_{j=1}^B g^t_j + z^t}, \qquad
    g^t_j = \min \biggl\{1, \frac{\tau(x^t)}{\bigl\|\nabla f_{\xi^t_j}(x^t)\bigr\|}\biggr\} \nabla f_{\xi^{t}_j}(x^t),
\end{equation}
where $z^t \sim \mathcal{N} (0, \br{\tau(x^t)}^2 \sigma^2_{\mathrm{DP}} \mathbf{I})$ and $\sigma_{\mathrm{DP}} \geq C \sqrt{T \log(\nicefrac{1}{\delta})} \epsilon^{-1}$ for some universal constant $C$ independent of $T, \delta, \epsilon$. For simplicity, we assume that $\xi_j^t$ are uniformly and independently sampled.
In contrast to \algname{DP-SGD} with fixed clipping, the noise variance in our approach varies throughout training because it depends on $\tau(x^t)$. This variability is crucial to the convergence result in the following theorem and distinguishes our setting from earlier studies \citep{koloskova2023revisiting}.

\begin{theorem}[\algname{DP-QC-SGD}] \label{adaptive:thm:dp_sgd}
Suppose $f$ is $L$-smooth (Ass. \ref{adaptive:ass:l_smooth}) and stochastic gradients satisfy Assumption~\ref{adaptive:ass:b_variance}. Then, for the step size chosen as
\begin{equation} \label{adaptive:eq:dp_step_size}
    \gamma_t \leq \frac{2p - \beta - c}{2 L (2 +\mathfrak{S})},
\end{equation}
where $\mathfrak{S} \eqdef 1 + d\sigma^2_{\mathrm{DP}}/B^2$, and $c, \beta \in (0, 1)$, and $\beta + c \leq 2p$ the iterates of \algname{DP-QC-SGD} \eqref{adaptive:eq:dp_sgd_qc} satisfy
\begin{equation} \label{adaptive:eq:dp_sgd_res}
\begin{aligned}
    \frac{c}{\Gamma_T} \sum_{t=0}^{T-1} \gamma_t \Exp{\sqn{\nabla f(x^t)}}
    \leq
    \frac{2(f(x^0) - f^{\inf})}{\Gamma_T}
    \\+ \frac{\sigma_q^2}{\Gamma_T} \sum_{t=0}^{T-1} \gamma_t h^{-2/q}
    \br{\beta^{-1}h^{2} + 2 \gamma_t L \br{4h +\mathfrak{S}-1}},
\end{aligned}
\end{equation}
where $\Gamma_T \eqdef \sum_{t=0}^{T-1} \gamma_t$ and $h \eqdef 1 - p$.
\end{theorem}

Analogously to Corollary \ref{adaptive:cor:const_res}, we simplify the result for constant step size $\gamma_t \equiv \gamma$.

\begin{equation}\label{adaptive:eq:const_dp_sgd_res}
\frac{c}{T}\sum_{t=0}^{T-1}\Exp{\|\nabla f(x^t)\|^2}
\le
\frac{2\bigl(f(x^0)-f^{\inf}\bigr)}{\gamma T}
+ \beta^{-1}\sigma_q^2 h^{2-2/q}
+ 2\gamma L \sigma_q^2 h^{-2/q} \br{4h + \mathfrak{S}-1}
\end{equation}

Theorem~\ref{adaptive:thm:dp_sgd} is similar to the non-private result \eqref{adaptive:eq:gen_res} in nature as it shows convergence to a neighborhood of a stationary point. However, there is a significant difference expressed in the factor $\mathfrak{S}$ appearing in the denominator of the stepsize condition \eqref{adaptive:eq:dp_step_size} and in the corresponding neighborhood term in \eqref{adaptive:eq:const_dp_sgd_res}. In particular, $\mathfrak{S}$ captures the inflation of the second-moment term due to the injected Gaussian noise; under the standard \algname{DP-SGD} construction \eqref{adaptive:eq:dp_sgd_qc}, its DP component decays with the cohort size $B$ (as the noise is added to the sum and then averaged). Consequently, in private federated learning with a sufficiently large cohort size $B$ and/or a relatively small number of communication rounds $T$ (hence a smaller $\sigma_{\mathrm{DP}}$), the private bound approaches the non-private one. On the other hand, for centralized DP training, $\sigma_{\mathrm{DP}}$ is typically larger, which results in a smaller admissible stepsize and a larger neighborhood. The latter can be reduced via a standard diminishing stepsize strategy, as the term involving $\mathfrak{S}$ depends on $\gamma$.

\paragraph{Effect of DP noise.}
Compared to the non-private bound, Theorem~\ref{adaptive:thm:dp_sgd} introduces an additional term scaling as
$\gamma L\sigma_q^2 (\mathfrak{S}-1)h^{-2/q}$, where $\mathfrak{S}-1=d\sigma_{\mathrm{DP}}^2/B^2$.
In particular, for $q=2$ the DP contribution scales like $h^{-1}$, highlighting that driving the
clipping bias to zero by taking $p\uparrow 1$ (i.e., $h\downarrow 0$) can be in tension with DP utility.

\section{Conclusion and future directions}
Our work provides the first theoretical analysis of \sgd with quantile clipping, advancing the understanding of adaptive clipping methods. We established convergence guarantees for \alg under standard smoothness and heavy-tailed noise conditions.
Our analysis revealed that \alg suffers from an inherent bias problem analogous to that observed in fixed clipping \sgd. Furthermore, we designed a time-varying quantile and step-size schedule that could fix the discovered limitation in a practical way.
Finally, we proposed and analyzed a differentially private extension, \algname{DP-QC-SGD}.

Moving forward, the observed limitations of the \alg (with fixed quantile) raise the question of possible improvements via algorithmic modifications.
The current analysis also assumes an idealized setting in which the exact quantile $\tau(x)$ is available. In practical scenarios such as cross-device FL, only an approximation is typically available.
Moreover, despite the empirical success of adaptive clipping, there are scenarios where it performs suboptimally \citep{xu2023federated}, motivating future research.
Addressing these open questions can lead to more robust and efficient private learning methods.

\endgroup

\chapter{Smoothed Normalization for Distributed Private Optimization}
\label{chapter5:smoothed_normalization}

\begingroup
\input{KAUST_Thesis/imported/05_smoothed_normalization/import_setup}

\section*{Abstract}
  Federated learning enables the training of machine-learning models while preserving the privacy of participants. Surprisingly, there is no differentially private distributed method for smooth, non-convex optimization problems with convergence guarantees under standard assumptions. The reason is that standard privacy techniques require bounding the participants' contributions, usually enforced by \textit{clipping} the updates. Existing literature typically ignores the effect of clipping by assuming bounded gradient norms or analyzes distributed algorithms with clipping while ignoring DP constraints. In this work, we study an alternative based on \textit{smoothed normalization} of the updates, motivated by its favorable performance in the single-node setting. By integrating smoothed normalization with an error-compensation mechanism, we design a new distributed algorithm \alg and prove that it achieves a faster convergence rate than prior methods. By extending \alg to the DP setting, we obtain the first differentially private distributed optimization algorithm with provable convergence guarantees under standard assumptions.
Finally, our empirical results from neural network training indicate robust convergence of \alg across different parameter settings.

\begin{quote}
\small
\noindent\textit{This chapter is based on:}\\
Egor Shulgin, Sarit Khirirat, and Peter Richt{\'a}rik, ``Smoothed Normalization for Efficient Distributed Private Optimization,'' \emph{arXiv preprint arXiv:2502.13482}, 2025. A version appeared as ``Private Federated Learning with Provable Convergence via Smoothed Normalization'' at the \emph{ICML Workshop on Collaborative and Federated Agentic Workflows (CFAgentic)}, 2025.
\end{quote}

\section{Introduction}\label{smooth:sec:intro}

Federated Learning (FL) has become a viable approach for distributed collaborative
training of machine learning models \citep{konecny2017federated,mcmahan2017communication,mcmahan2018learning}.
This growing interest has spurred the development of novel distributed optimization methods tailored for FL, focusing on
ensuring high \textit{communication efficiency}~\citep{kairouz2021advances}.
Although FL optimization methods ensure that private data is never directly transmitted, \citet{boenisch2023curious} demonstrated that the global models produced through FL can still enable the reconstruction of participants' data.
Therefore, it is essential to study \emph{differentially private} distributed optimization methods for differentially private training \citep{dwork2014algorithmic,mcmahan2018learning,sun2019can}.

To mitigate emerging privacy risks in FL, differential privacy (DP)~\citep{dwork2014algorithmic} has become the standard for providing theoretical privacy guarantees in machine learning.
DP is often enforced by a clipping operator.
It bounds gradient sensitivity, allowing the addition of DP noise to the updates before communication.
While gradient clipping enables DP as in Differentially Private Stochastic Gradient Descent (\algname{DP-SGD})~\citep{abadi2016deep}, it also introduces a bias that can impede convergence
~\citep{chen2020understanding,koloskova2023revisiting}.
Often, distributed DP gradient methods with clipping have been studied under assumptions that are unrealistic for heterogeneous FL environments, such as
bounded gradient norms~\citep{li2022soteriafl,pmlr-v216-wang23b,lowy2023private,zhang2020private} or effectively ignoring the impact of clipping bias.
To our knowledge, convergence guarantees for distributed DP methods remain elusive unless the impact of clipping bias is explicitly considered.

Error feedback (EF), also known as error compensation (EC), has been used to alleviate clipping bias and establish convergence for non-private distributed methods. Examples include \algname{EF21}~\citep{EF21} and its use in methods with clipped or normalized updates~\citep{khirirat2023clip21,yu2023smoothed}.
However, extending these methods to the private setting remains an open problem.
Furthermore, optimizing the convergence of distributed DP clipping methods is challenging because the clipping threshold significantly influences both the convergence speed and DP noise variance.
Extensive grid search for the optimal clipping threshold is computationally expensive~\citep{andrew2021differentially} and leads to additional privacy loss~\citep{papernot2021hyperparameter}. Two major approaches have emerged to address the need to manually tune the clipping threshold. The first is to use adaptive clipping techniques, such as adaptive quantile clipping, initially proposed by \citet{andrew2021differentially} and further analyzed by \citet{merad2023robust,shulgin2024convergence}. The second, which is the focus of this chapter, is to replace clipping with a normalization operator.

{Smoothed normalization}, introduced by \citet{bu2024automatic,yang2022normalized}, is a normalization operator that offers an alternative to clipping.
Unlike clipping, smoothed normalization eliminates the need to tune the clipping threshold. By ensuring that the Euclidean norm of the normalized gradient is bounded above by one, smoothed normalization provides robust performance for \algname{DP-SGD} in the private setting.
However, little work characterizes the properties of smoothed normalization, and a rigorous convergence analysis of \algname{DP-SGD} using this operator is still lacking, especially in the distributed setting. While the method has been studied in the single-node setting, the convergence results rely on unrealistic or restrictive assumptions, such as symmetric gradient noise~\citep{bu2024automatic} and almost-sure bounds on the gradient-noise variance~\citep{yang2022normalized}.

\subsection{Contributions}
We propose \alg, a distributed gradient method leveraging smoothed normalization and error compensation.
It is the first algorithm to achieve provable convergence for DP, non-convex distributed optimization without relying on restrictive assumptions. Our key contributions are:

\textbf{$\bullet$ Favorable properties of smoothed normalization.}
In Section~\ref{smooth:sec:not_def}, we show that smoothed normalization exhibits a contractive-like property similar to that of biased compression operators \citep{beznosikov2023biased,shulgin2022shifted}. Combined with a novel induction-based proof technique, this property allows us to analyze \alg without ignoring the impact of bias.

\begin{table*}
\centering
\begingroup
\resizebox{\textwidth}{!}{%
\begin{tabular}{lccc}
\hline
{\textbf{Property}} & {\algname{DP-SGD}} & {\algname{Clip21}} & \alg (\textbf{Ours}) \\
\hline
Non-private Convergence  & \xmark & \cmark  & {\color{green}\cmark} \\
Private Convergence     & \xmark & \xmark  & {\color{green}\cmark}  \\
Operators & Clipping / Normalization & Clipping & {\color{green}Normalization} \\
Additional Assumptions & Bounded gradient or heterogeneity & No & {\color{green}No} \\
\hline
\end{tabular}
}
\caption{
Comparison of \emph{distributed} private optimization methods. Unlike prior work, which either lacks private convergence guarantees or requires restrictive assumptions like bounded gradients, \alg is the first to provide these guarantees under standard smoothness conditions alone.
}
\label{smooth:tab:comparison}
\endgroup
\end{table*}

\textbf{$\bullet$ Convergence for non-convex, smooth problems without restrictive assumptions.}
In Section~\ref{smooth:sec:norm21}, we prove that
\alg achieves the optimal convergence rate \citep{carmon2020lower}
for minimizing non-convex, smooth functions without imposing additional restrictive assumptions, such as bounded gradient norms or bounded heterogeneity. Furthermore, \alg achieves a faster convergence rate than \algname{Clip21}~\citep{khirirat2023clip21} and does not require inaccessible knowledge of the value $f(x^0)-f^{\inf}$.

\textbf{$\bullet$
The first provable convergence in the private setting.}
In Section~\ref{smooth:sec:dp_norm21}, we extend \alg to the DP setting.
Specifically, \alg achieves the first convergence guarantees for DP, non-convex, smooth problems \emph{without} ignoring the bias introduced by smoothed normalization.
This is the first provably efficient distributed method in the DP setting under standard assumptions, thus addressing the gap left by prior work \citep{khirirat2023clip21}.
Theoretical comparisons between \algname{DP-SGD}, \algname{Clip21}, and \alg are summarized in~\Cref{smooth:tab:comparison}.

\textbf{$\bullet$ Robust empirical performance  of \alg.}
In Section~\ref{smooth:sec:exp}, we verify the theoretical benefits of \alg in both non-private and private training via experiments. %
Our algorithm demonstrates robust convergence across different parameter values and benefits from error compensation that enables superior performance over vanilla distributed gradient normalization methods (such as \algname{DP-SGD}).
In private training, server normalization enhances the robustness of \algname{DP-}\alg across tuning parameters. Finally, \algname{DP-}\alg without server normalization outperforms \algname{DP-Clip21}.

\subsection{Related work}\label{smooth:sec:related_work}

\textbf{Clipping and normalization.}
Clipping and normalization
address many key challenges in machine learning.
They mitigate the problem of exploding gradients in recurrent neural networks \citep{pascanu2013difficulty}, enhance neural network training for tasks in natural language processing \citep{merity2017regularizing,brown2020language} and computer vision \citep{brock2021high}, enable differentially private machine learning \citep{abadi2016deep,mcmahan2018learning}, and provide robustness in the presence of misbehaving or adversarial workers \citep{karimireddy2021learning,ozfatura2023byzantines,malinovsky2023byzantine}.
In this chapter, we consider smoothed normalization, introduced by \citet{yang2022normalized,bu2024automatic}, as an alternative to clipping, given its robust empirical performance and hyperparameter tuning benefits in the DP setting.

\textbf{Private optimization methods.}
\algname{DP-SGD}~\citep{abadi2016deep} is the standard distributed first-order method that achieves the DP guarantee by clipping the gradient
before adding noise scaled with the clipped gradient's sensitivity.
However, existing \algname{DP-SGD} convergence analyses often neglect the clipping bias.
Specifically, convergence results for smooth functions under differential privacy often require either the assumption of bounded gradient norms~\citep{zhang2020private,li2022soteriafl,zhang2022understanding,pmlr-v216-wang23b,lowy2023private,murata2023diff2,wang2024efficient} or conditions where clipping is effectively inactive~\citep{zhang2024private,noble2022differentially}.
Thus, the convergence behavior of \algname{DP-SGD} in the presence of clipping bias remains poorly understood.

\textbf{Single-node non-private methods with clipping. }
The impact of clipping bias has been extensively studied in single-node gradient methods for non-private optimization.
Numerous works have shown strong convergence guarantees of clipped gradient methods under various conditions, including nonsmooth, rapidly growing convex functions~\citep{shor2012minimization,ermoliev1988stochastic,alber1998projected}, generalized smoothness~\citep{zhang2020why,koloskova2023revisiting,gorbunov2024methods,vankov2025optimizing,lobanov2024linear,hubler2024parameter}, and heavy-tailed noise~\citep{gorbunov2020stochastic,nguyen2023improved,gorbunov2024highprobability,hubler2024gradient,chezhegov2024gradient}.

\textbf{Distributed non-private methods with clipping.}
Applying gradient clipping in the distributed setting is challenging.
Existing convergence analyses often rely on bounded-heterogeneity assumptions, which may not hold for arbitrarily diverse clients.
For example, federated optimization methods with clipping have been analyzed under the bounded difference between the local and global gradients~\citep{wei2020federated,liu2022communication,crawshaw2023episode,li2024an}.
However, even in the non-private setting, these distributed clipping methods do not converge for simple problems~\citep{chen2020understanding,khirirat2023clip21} with arbitrary clipping thresholds.
To address the convergence issue, one approach is to use error feedback mechanisms, such as \algname{EF21} \citep{EF21}, that are employed by \citet{khirirat2023clip21,yu2023smoothed} to compute local gradient estimators and alleviate clipping bias. However, these distributed clipping methods using error feedback are limited to the non-private setting, and extending them to the DP setting is still an open problem.
In this chapter, we propose a distributed method that replaces clipping with smoothed normalization in the \algname{EF21} mechanism.
Our method provides the first provable convergence in the DP setting and empirically outperforms the distributed version of \algname{DP-SGD} with smoothed normalization \citep{bu2024automatic,yang2022normalized}, a special case of \citet{das2021convergence}.

\textbf{Error feedback.}
Error feedback, or error compensation, has been applied to improve the convergence of distributed methods with gradient compression for communication-efficient learning.
First introduced by \citet{Seide2014}, \algname{EF14} was extensively analyzed for first-order methods in both single-node settings~\citep{stich2018sparsified,karimireddy2019error,stich2019error,khirirat2019convergence} and distributed settings~\citep{wu2018error,Alistarh-EF-NIPS2018,gorbunov2020linearly,qian2021error,tang2019doublesqueeze,danilova2022distributed,qian2023catalyst}.
Another error feedback variant is \algname{EF21} proposed by \citet{EF21} that ensures strong convergence under any contractive compression operator for non-convex, smooth problems.
Recent variants, e.g. \algname{EF21-SGD2M}~\citep{fatkhullin2024momentum} and \algname{EControl}~\citep{gao2023econtrol}, have been developed to obtain the lower iteration and communication complexities than \algname{EF21} for stochastic optimization.

\subsection{Notation}
We use $[a,b]$ to denote the set
$\{a,a+1,a+2,\ldots,b\}$ for integers $a,b$ such that $a \leq b$, and $\Exp{u}$ to represent the expectation of a random variable $u$.
For vectors $x,y\in\R^d$, $\inp{x}{y}$ denotes their inner product, and $\norm{x} \eqdef \sqrt{\inp{x}{x}}$ denotes the Euclidean norm of $x$.
Finally, for functions $f,g:\R^d\rightarrow\R$, we write $f(x)=\cO(g(x))$ if $f(x) \leq M \cdot g(x)$ for some $M>0$.

\section{Preliminaries}\label{smooth:sec:preliminaries}

We focus on solving the
finite-sum optimization problem:
\begin{align}\label{smooth:eqn:problem}
    \underset{x\in\R^d}{\min}  \  \biggl\{ f(x) \eqdef \frac{1}{n}\sum_{i=1}^n f_i(x) \biggr\},
\end{align}
where $x\in\R^d$ is the vector of model parameters of dimension $d$, and $f_i:\R^d\rightarrow\R$ is either the loss function for client $i \in [1,n]$ (distributed setting) or the loss associated with data point $i$ (single-node setting).
Moreover, we impose the following assumption on objective functions commonly used for analyzing the convergence of first-order optimization algorithms~\citep{nesterov2018lectures}.

\begin{assumption}\label{smooth:assum:lowebound_f}
Consider Problem~\eqref{smooth:eqn:problem}.
\begin{enumerate}
    \item  Let  $f:\R^d\rightarrow\R$ be bounded from below by a  finite constant $f^{\inf}$, i.e. $f(x) \geq f^{\inf} > -\infty$ for all $x\in\R^d$, and be $L$-smooth, i.e. $\norm{\nabla f(x) - \nabla f(y)} \leq L \norm{x-y}$ for all $x,y\in\R^d$.
    \item Let each $f_i:\R^d\rightarrow\R$ be $L_i$-smooth, i.e. $\norm{\nabla f_i(x) - \nabla f_i(y)} \leq L_i \norm{x-y}$ for all $x,y\in\R^d$.
\end{enumerate}
\end{assumption}

\subsection{DP-SGD}\label{smooth:sec:not_def}

To solve Problem~\eqref{smooth:eqn:problem}, the most common approach that ensures the approximate $(\epsilon, \delta)$-differential privacy~\citep{dwork2006calibrating} is via the \algname{DP-SGD} method~\citep{abadi2016deep}
\begin{align}\label{smooth:eqn:DP_biased_GD}
    x^{k+1} = x^k - \frac{\gamma}{B} \Bigl( \sum_{i\in\cB^k} \Psi(\nabla f_i(x^k)) + z^k \Bigr),
\end{align}
where $\gamma>0$ is the step size, $\cB^k$ is a subset of $\{1,2,\ldots,n\}$ with cardinality $\vert \cB^k \vert = B$, $z^k \in \R^d$ is the DP noise, and $\Psi:\R^d\rightarrow\R^d$ is an operator with bounded norm, i.e.  $\norm{\Psi(g)} \leq \Phi$ for some $\Phi>0$ and  any $g\in\R^d$.
The method~\eqref{smooth:eqn:DP_biased_GD} achieves $(\epsilon,\delta)$-DP~\citep{abadi2016deep} if  $z^k$  is zero-mean Gaussian noise with variance
\begin{align}\label{smooth:eqn:variance_dp_sgd}
\sigma_{\text{DP}}^2 \geq  \Phi^2 \cdot    \frac{cB^2}{n^2}\frac{{K \log(1/\delta)}}{\epsilon^2},
\end{align}
where $c>0$ is a constant, and $K>0$ is the total number of iterations.
To obtain reasonable DP guarantees, we usually choose $\epsilon \leq 10$ and $\delta \ll 1/n$, where $n$ is the number of data points \citep{ponomareva2023dp}. Notice that the DP Gaussian noise variance \eqref{smooth:eqn:variance_dp_sgd} is scaled with the \textit{sensitivity} $\Phi$.

The method \eqref{smooth:eqn:DP_biased_GD} has been analyzed, e.g. by \citet{zhang2020private,zhang2022understanding}, under the bounded gradient norm assumption
\begin{eqnarray} \label{smooth:eq:bounded_grad}
    \norm{\nabla f_i(x)} \leq \Phi \quad \text{ for all \ $i$ \ and \ } x \in \R^d.
\end{eqnarray}
However, this assumption has several limitations.
First, the sensitivity $\Phi$ is typically infeasible to compute for many loss functions used in training machine learning models. Even when it can be estimated, the resulting upper bound is often overly pessimistic, leading to excessively large DP noise and thus significantly degrading the convergence performance.
Second, this assumption restricts the class of models and loss functions~$f$, as it excludes simple quadratic functions over unbounded domains.
Third, this assumption is \say{pathological} in the distributed setting because it restricts the heterogeneity between different clients and can result in vacuous bounds \citep{khaled2020tighter}.

To enforce bounded sensitivity without imposing the bounded gradient norm, \citet{abadi2016deep} suggested clipping with threshold $\tau>0$
\begin{equation} \label{smooth:eq:clipping}
    \Clip{\tau}{g} \eqdef \min \br{1,\frac{\tau}{\norm{g}}} g.
\end{equation}

Here, the sensitivity $\Phi$ is the clipping threshold $\tau$, as $\norm{\Clip{\tau}{g}} \leq \tau = \Phi$.
In general, the method \eqref{smooth:eqn:DP_biased_GD} that uses clipping \eqref{smooth:eq:clipping} is referred to as \algname{DP-SGD} in the literature.
However, it is challenging to analyze the convergence of \algname{DP-SGD} without additional restrictive assumptions such as the symmetric noise assumption \citep{chen2020understanding,qian2021understanding}.
Even without DP noise, \algname{DP-SGD} fails to converge due to the clipping bias~\citep{koloskova2023revisiting}.
Furthermore, since smaller values of $\tau$ imply stronger privacy but larger bias, jointly optimizing convergence and privacy of \algname{DP-SGD} by carefully tuning $\tau$ and $\gamma$ in the DP setting is a difficult task \citep{kurakin2022toward,bu2024automatic}.

\textbf{Smoothed normalization as an alternative to clipping.}
To eliminate the need to tune the clipping threshold $\tau$, smoothed normalization is an operator alternative~\citep{bu2024automatic,yang2022normalized}:
\begin{align}\label{smooth:eqn:normalize}\Normalize{\alpha}{g} \eqdef \frac{1}{\alpha + \norm{g}} g,
\end{align}
for some $\alpha  \geq 0$
and satisfies the following property.
\begin{lemma}\label{smooth:lemma:prop_norm_vr2}
For any $\alpha  \geq 0$, $\beta>0$, and $g\in\R^d$,
\begin{align}
\norm{ \Normalize{\alpha}{g} }  \leq & 1, \quad \text{and} \notag\\ \sqnorm{g -   \beta  \Normalize{\alpha}{g}} =& \left( 1 - \frac{\beta}{\alpha+\norm{g}}\right)^2 \sqnorm{g}. \label{smooth:eq:norm_2}
\end{align}
\end{lemma}
Clearly, smoothed normalization ensures the following conditions. First, the norm of the normalized vector is bounded above by $1$. Second, the distance between the true vector and a $\beta$-multiple of the normalized vector is bounded by a function of $\beta$, $\alpha$, and $\norm{g}$.
Although smoothed normalization with $\alpha=0$ recovers  standard normalization ${g}/{\norm{g}}$~\citep{nesterov1984minimization,hazan2015beyond,levy2016power}, smoothed normalization with $\alpha>0$  improves the contraction factor, compared to standard normalization.
Specifically, as $\norm{g}\rightarrow 0$, the contraction factor of smoothed normalization approaches $(1-\beta/\alpha)^2$, while standard normalization lacks this contraction property.

Although \algname{DP-SGD} \eqref{smooth:eqn:DP_biased_GD} with smoothed normalization achieves robust empirical performance in the DP setting~\citep{bu2024automatic},
its theoretical convergence is limited to the single-node setting and relies on restrictive assumptions like the central symmetry of stochastic gradients.%

\subsection{Limitations of DP distributed gradient methods}
Extending the convergence results of \algname{DP-SGD} to the distributed setting poses significant challenges due to {potential client heterogeneity}.
Existing results often address the bias introduced by the operator (clipping or normalization) by relying on restrictive assumptions, such as imposing bounded gradient norms~\citep{li2022soteriafl,zhang2022understanding,murata2023diff2,wang2024efficient}, or assuming that clipping is effectively turned off~\citep{zhang2024private,noble2022differentially}. {Recently, \citet{li2024an} extended the analysis of \citet{koloskova2023revisiting} to a distributed private setting under strong gradient dissimilarity condition. However, their method fails to converge due to the clipping bias, as discussed in the previous section.}
More importantly, even in the absence of the DP noise ($z^k=0$), the inherent bias in the gradient estimator can severely impact the convergence.
For instance, \algname{DP-SGD}~\eqref{smooth:eqn:DP_biased_GD} diverges exponentially when $\Psi(\cdot)$ is a Top-$1$ compressor~\citep{beznosikov2023biased} and fails to converge when $\Psi(\cdot)$ is clipping~\citep{chen2020understanding,khirirat2023clip21}.

Also, \algname{DP-SGD} that uses smoothed normalization \eqref{smooth:eqn:normalize} directly fails to converge, as shown in the example below:

\begin{example} \label{smooth:ex:counter}
Consider Problem~\eqref{smooth:eqn:problem} with $n=2, d=1$, $f_1 (x) = \frac{1}{2}\br{x-3}^2$ and $f_2 (x) = \frac{1}{2}\br{x+3}^2$. Then $f(x) = \frac{1}{2}(f_1(x) + f_2(x))$ satisfies Assumption \ref{smooth:assum:lowebound_f} and is minimized at $x^\star = 0$. The iterates $\{x^k\}$ generated by \eqref{smooth:eqn:DP_biased_GD} (for $B=2$) with $z^k=0$ and $\alpha=0$ do not progress when $x^0 = 2$, as the gradient estimator $\Normalize{\alpha}{\nabla f_1(x^k)} + \Normalize{\alpha}{\nabla f_2(x^k)}$ results in
\begin{equation*}
    \frac{\nabla f_1(x^0)}{\|\nabla f_1(x^0)\|} + \frac{\nabla f_2(x^0)}{\|\nabla f_2(x^0)\|} = -1/1 + 5/5 = 0.
\end{equation*}
\end{example}
Naively applying normalization to the gradients in \algname{DP-SGD} leads to a method that does not converge in the distributed setting without further assumptions.
This fundamental limitation affects federated averaging algorithms that apply normalization on the client updates~\citep{das2021convergence}.

\subsection{EF21 mechanism}

One approach to resolve the convergence issues of distributed gradient methods with biased operators is to use \algname{EF21}, an error feedback mechanism developed by \citet{EF21}.
Instead of directly applying the biased gradient estimator $\Psi$ to the gradient, \algname{EF21} applies $\Psi$ to the \emph{difference} between the true gradient and the current error feedback (memory) vector.
At iteration $k=0,\ldots,K$, each client $i$ receives the current iterate $x^k$ from the central server, and computes its local update $g_i^{k+1}$ via
\begin{eqnarray}\label{smooth:eqn:g_i_k_dist_biased_gradient}
    g_i^{k+1} = g_i^k + \beta \Psi(\nabla f_i(x^k)- g_i^k),
\end{eqnarray}
where $\beta > 0$. Next, the central server receives the average of local error feedback vectors that are communicated by all clients $(\nicefrac{1}{n})\sum_{i=1}^n\Psi(\nabla f_i(x^k)-g_i^k)$, computes the global gradient estimator
\begin{eqnarray}\label{smooth:eqn:g_k_dist_biased_gradient}
    g^{k+1} = g^k + \frac{\beta}{n} \sum_{i=1}^n\Psi(\nabla f_i(x^k)-g_i^k),
\end{eqnarray}
and updates the next iterate via $x^{k+1} = x^k - \gamma g^{k+1}$.
{\color{black}This method generalizes \algname{EF21} by replacing a contractive compressor\footnote{\color{black}A contractive compressor~\citep{stich2018sparsified} is defined by  $\sqnorm{g - \cC(g)} \leq (1-\eta)^2 \sqnorm{g}$ for $\eta\in (0,1]$.} with other biased estimators, such as  clipping in \algname{Clip21} proposed by \citet{khirirat2023clip21}.
}
{\color{black}
Although \algname{Clip21} attains the $\cO(1/\sqrt{K})$ convergence in the non-private setting, deriving its convergence in the presence of the DP noise is a challenging task.
Because clipping does not satisfy the contractive property required of compressors by
\algname{EF21} (\Cref{smooth:tab:clipping_normalization}), its convergence analysis relies heavily on separate descent inequalities when clipping is turned on and off (\Cref{smooth:app:ef21_clip_norm}). Also, the clipping threshold $\tau$ intricately influences both privacy and convergence guarantees. A sufficiently large $\tau$ is needed to achieve the descent inequality, but this condition requires adding large Gaussian noise, which may in turn prevent convergence as the noise accumulates.
}

\section{\texorpdfstring{$\alpha$-NormEC}{alpha-NormEC}}%
\label{smooth:sec:norm21}

To address the convergence challenges of \algname{Clip21}, we propose \alg presented in Algorithm~\ref{smooth:alg:norm_norm21_dp_v1}.
Note that \alg recovers \algname{EF21} and \algname{Clip21}, when we replace smoothed normalization with contractive compressor and clipping, respectively. Additionally we employ \textit{Server Normalization} (SN) $x^{k+1}=x^k-\gamma \hat g^{k+1}/\norm{\hat g^{k+1}}$ and adopt notation $0/0=0$. In the main part we focus on Algorithm~\ref{smooth:alg:norm_norm21_dp_v1}, and in \Cref{smooth:app:convg_norm21_no_server_norm} variation of \alg without SN is analyzed.
Our experiments (in~\Cref{smooth:app:exp_dp}) indicate that, while \alg with SN offers robust performance, the variant without it achieves fastest convergence.

\begin{table*}[t]
    \centering
    \resizebox{\textwidth}{!}{%
    \begin{tabular}{cc}
     \hline
      \bf Operator   &  \bf Property
      \\ \hline
      \begin{tabular}{c}
        Contractive compressor
        { $\cC:\R^d\rightarrow\R^d$}
      \end{tabular}
      &
     $\sqnorm{\cC(g)-g} \leq (1-\eta)^2 \sqnorm{g}$
     \\
      \begin{tabular}{c}
        Clipping
        { $\Clip{\tau}{g} \eqdef \min\left( 1, \frac{\tau}{\norm{g}}\right) g$}
      \end{tabular}
    &   $\sqnorm{\Clip{\tau}{g}-g} \leq \max( 0 , \norm{g} - \tau)^2$
    \\
      \begin{tabular}{c}
        Smoothed normalization
        { $\Normalize{\alpha}{g}\eqdef \frac{1}{\alpha + \norm{g}} g$}
      \end{tabular}
      &   $\sqnorm{\Normalize{\alpha}{g}-g} \leq \left(1- \frac{1}{\alpha+\norm{g}} \right)^2 \sqnorm{g}$
      \\ \hline
    \end{tabular}
    }
    \caption{Comparison of contractive compressors, clipping, and smoothed normalization with their properties. Smoothed normalization, unlike clipping, satisfies a contractive-like property similar to that of compressors.
    }
    \label{smooth:tab:clipping_normalization}
\end{table*}

\begin{algorithm}
\caption{\algname{(DP-)}\alg}
\label{smooth:alg:norm_norm21_dp_v1}
\begin{algorithmic}[1]
\STATE \textbf{Input:} {Step size} $\gamma>0$; $\beta>0$; normalization parameter $\alpha>0$;  initial points $x^0,{g_i^{0}} \in \R^d$ for $i\in [1,n]$; $\hat g^0 = \frac{1}{n}\sum_{i=1}^n g_i^0$; $z_i^k \in \R^d$ {are sampled from Gaussian distribution with zero mean and $\sigma_{\rm DP}^2$-variance.}
\FOR{each iteration $k = 0, 1, \dots, K$}
    \FOR{each client $i = 1, 2, \dots, n$ in parallel}
        \STATE Compute local  gradient $\nabla f_i(x^k)$
        \STATE Compute $\Delta_i^k = \Normalize{\alpha}{\nabla f_i(x^k) - g_i^{k}}$
        \STATE Update  {$g_i^{k+1} = g_i^{k} + \beta \Delta_i^k$}
        \STATE \textbf{Non-private setting:}  Transmit $\hat \Delta_i^k = \Delta_i^k$
        \STATE \textbf{Private setting:}  Transmit $\hat \Delta_i^k = \Delta_i^k + z_i^k$
    \ENDFOR
    \STATE Server computes $\hat g^{k+1} = \hat g^{k} + ({\beta}/{n})\sum_{i=1}^n \hat \Delta_i^k$
    \STATE Server updates {$x^{k+1} = x^k - \gamma {\hat g^{k+1}}/{\norm{\hat g^{k+1}}}$}
\ENDFOR
\STATE \textbf{Output:} $x^{K+1}$
\end{algorithmic}
\end{algorithm}

{\color{black}
Although  \alg and \algname{Clip21} use error compensation,
\alg provides stronger guarantees than \algname{Clip21} in the non-private setting, and achieves  convergence under only standard assumptions unlike \algname{Clip21} in the DP setting.
These  advantages result from the contractive-like property of smoothed normalization, which distinguishes it from clipping (see~\Cref{smooth:tab:clipping_normalization}).
In contrast to \algname{EF21}, which relies on the strong contractivity of its compressors, \alg leverages the monotonicity of smoothed normalization to obtain the descent inequality.
}

The first theorem presents the convergence of \alg in the non-private setting.

\begin{theorem}[Non-private setting]\label{smooth:thm:norm_ef21_nondp}
Consider  \alg (Algorithm~\ref{smooth:alg:norm_norm21_dp_v1}) for solving Problem~\eqref{smooth:eqn:problem}, where Assumption~\ref{smooth:assum:lowebound_f} holds.
{\color{black}
For any $D>0$, choose $\beta,\gamma>0$  such that
\begin{eqnarray*}
    \beta < \alpha + \frac{D}{(K+1)^{1/2}}, \ \text{and} \ \gamma \leq \frac{\beta}{L_{\max}} \frac{D}{\alpha+D} \frac{1}{(K+1)^{1/2}},
\end{eqnarray*}
where $L_{\max}=\max_{i\in[1,n]} L_i$,
and initialize $g_i^0\in\R^d$ such that
\[
\max_{i\in[1,n]} \norm{\nabla f_i(x^0)-g_i^{0}} = \frac{D}{(K+1)^{1/2}}.
\]
Then,
\begin{eqnarray*}
     \min_{k\in[0,K]} \norm{\nabla f(x^k)}   \leq   \frac{C}{(K+1)^{1/2}},
\end{eqnarray*}
where $C = \frac{L_{\max} (\alpha + D)}{\beta D} (f(x^0)-f^{\inf}) + 2 D+ \frac{L}{2} \frac{\beta D}{L_{\max}(\alpha+D)}$.}
\end{theorem}

From Theorem~\ref{smooth:thm:norm_ef21_nondp},~\alg attains the same $\cO(1/\sqrt{K+1})$ rate  in the gradient norm as classical gradient descent.
By further choosing $\beta$ properly according to~\Cref{smooth:app:derivation_sublinear_nonprivate}, \alg attains:
\begin{eqnarray}\label{smooth:eqn:final_bound_norm21}
      \min_{k\in[0,K]} \norm{\nabla f(x^k)}
     \leq  \frac{\sqrt{2L (f(x^0)-f^{\inf})} + 2D}{(K+1)^{1/2}}.
\end{eqnarray}
The first term $\sqrt{2L (f(x^0)-f^{\inf})}(K+1)^{-1/2}$ matches classical gradient descent~\citep{carmon2020lower}, while the second term $2D(K+1)^{-1/2}$ comes from initializing $x^0,g_i^0 \in \R^d$ such that $\max_{i\in[1,n]} \norm{\nabla f_i(x^0)-g_i^{0}}  = D(K+1)^{-1/2}$ with $D>0$.
To ensure the upper-bound, one choice is $g_i^0 = \nabla f_i(x^0) + e$, where $e = (D/\sqrt{K+1},0,\ldots,0)\in\R^d$. %

\paragraph{Comparison to \algname{Clip21}.}
In the non-private setting, \alg
{\color{black}~has a simpler proof and} provides better convergence guarantees than \algname{Clip21}.
Specifically, the convergence bound of \alg in \eqref{smooth:eqn:final_bound_norm21} exhibits a smaller factor than that of \algname{Clip21}, as explained in Appendix~\ref{smooth:app:compare_norm21_clip21}.
The hyperparameter choices for \alg ($\beta, \alpha, \gamma$) in Theorem~\ref{smooth:thm:norm_ef21_nondp} are also easier to implement.
In particular, the step-size $\gamma$ for \alg avoids reliance on the practically inaccessible sub-optimality gap $f(x^0)-f(x^\star)$, in contrast to \algname{Clip21}, whose step-size (Theorem 5.6 of \citet{khirirat2019convergence}) also depends on $\max_{i\in[1,n]}\|\nabla f_i(x^0)\|$.

{\color{black}
\paragraph{Comparison to \algname{EF21}.}
Although \alg incorporates Error Compensation similarly to \algname{EF21}, these two methods rely on fundamentally different operator conditions and proof techniques. \algname{EF21} requires a strong \textit{fixed} contractivity condition of compressors.
Smoothed normalization violates this assumption as it satisfies \textit{state-dependent} contractive-like property \eqref{smooth:eq:norm_2}.
That is why we rely on a novel induction-based proof for \alg as detailed in~\Cref{smooth:app:alg_and_EF21}.
}

\paragraph{Extensions of \alg.}
Our  analysis can be extended to establish the convergence of \alg without server normalization in \Cref{smooth:app:convg_norm21_no_server_norm}, and \alg using stochastic local gradients in \Cref{smooth:app:norm21_stochastic}.

\subsection{\texorpdfstring{$\alpha$-NormEC in the DP setting}{alpha-NormEC in the DP setting}}\label{smooth:sec:dp_norm21}

Next, we demonstrate that \alg achieves provable convergence guarantees in the DP setting, the feature that \algname{Clip21} lacks. As shown in~\Cref{smooth:alg:norm_norm21_dp_v1}, each client applies smoothed normalization to its gradient before injecting DP noise. Because smoothed normalization ensures the sensitivity is always 1, we can prove that \alg achieves utility guarantees by appropriately choosing the variance of the DP noise according to~\citet{abadi2016deep}.

Unlike \algname{Clip21}~\citep{khirirat2023clip21}, \alg achieves $(\epsilon,\delta)$-DP\footnote{Privacy guaranty follows from Theorem 1 of \citet{abadi2016deep}.} and comes with utility guarantees.
We show this by setting the standard deviation of the DP noise according to Theorem 1 of \citet{abadi2016deep}, i.e., $\sigma_{\rm DP}=\cO(\sqrt{(K+1)\log(1/\delta)}\epsilon^{-1})$, which yields the utility bound $\cO\left(\Delta\sqrt[4]{\frac{d \log(1/\delta)}{n\epsilon^2}} \right)$ with constant $\Delta = \sqrt{ L_{\max} (f(x^0)-f^{\inf})}$, as shown below:

\begin{theorem}[DP setting]\label{smooth:thm:dp_n_norm21}
Consider Problem~\eqref{smooth:eqn:problem}, where Assumption~\ref{smooth:assum:lowebound_f} holds.
Let
\[
\Delta = \sqrt{ L_{\max} (f(x^0)-f^{\inf})},\quad
A = \frac{\sqrt{\log(1/\delta)}}{\sqrt{n}\epsilon},\quad
\sigma_{\rm DP} = \cO\left(\frac{\sqrt{(K+1)\log(1/\delta)}}{\epsilon}\right).
\]
Also, choose $\beta = \frac{\beta_0}{K+1}$ with
\[
\beta_0 = \cO\left( \sqrt{\frac{\Delta}{A}} \right),\quad
\alpha = R = \cO\left( \sqrt[4]{d}\sqrt{\Delta A} \right),\quad
\gamma \leq \frac{\beta_0 R}{\alpha + R}\frac{1}{L_{\max}} \frac{1}{K+1}.
\]
Then, \algname{DP-}\alg (\ref{smooth:alg:norm_norm21_dp_v1}) satisfies $(\epsilon,\delta)$-DP and attains
\begin{align*}
  \min_{k\in[0,K]} \Exp{\norm{\nabla f(x^k)}}
 \leq \cO\left(  \Delta \sqrt[4]{d\frac{\log(1/\delta)}{n \epsilon^2}} \right).
\end{align*}
\end{theorem}

\textbf{Utility guarantees.}
Unlike \algname{Clip21}, \alg provides the utility bound in the DP distributed setting that accounts for the effect of bounding sensitivity, a factor often neglected in the existing literature.
Our obtained utility bound applies for smooth problems without the bounded gradient norm assumption, the limitation present in prior works for analyzing \algname{DP-SGD}, e.g., by \citet{li2022soteriafl,pmlr-v216-wang23b,lowy2023private,zhang2020private}.
Furthermore, to ensure the condition on $R$, we can use secure aggregation techniques to initialize the memory vector $\hat{g}^0 = \frac{1}{n}\sum_{i=1}^n g_i^0$ at the server, without revealing clients' local gradients $g_i^0$ (details in~\Cref{smooth:app:private_initial}).

\FloatBarrier
\section{Experiments}\label{smooth:sec:exp}

We evaluate the performance of \alg
for deep neural network training in both non-private and private settings.
We conduct experiments on the CIFAR-10 \citep{krizhevsky2009learning} dataset using the ResNet20 \citep{he2016deep} model for the image classification task. The compared methods are run for 300 communication rounds. The convergence plots present results for tuned step size $\gamma$.
Additional experimental details and results are provided in Appendix \ref{smooth:app:experiments}.

\subsection{Non-private training}

\textbf{Stable convergence of \alg across the normalization parameter values $\alpha$, and robustness across  $\beta$ values.}
From Figure~\ref{smooth:fig:accuracy_heatmap}, we observe that convergence of \alg is stable with respect to a wide range of $\alpha$ values and robust to variations in $\beta$.

\begin{wrapfigure}{r}{0.44\textwidth}
\centering
\includegraphics[width=\linewidth]{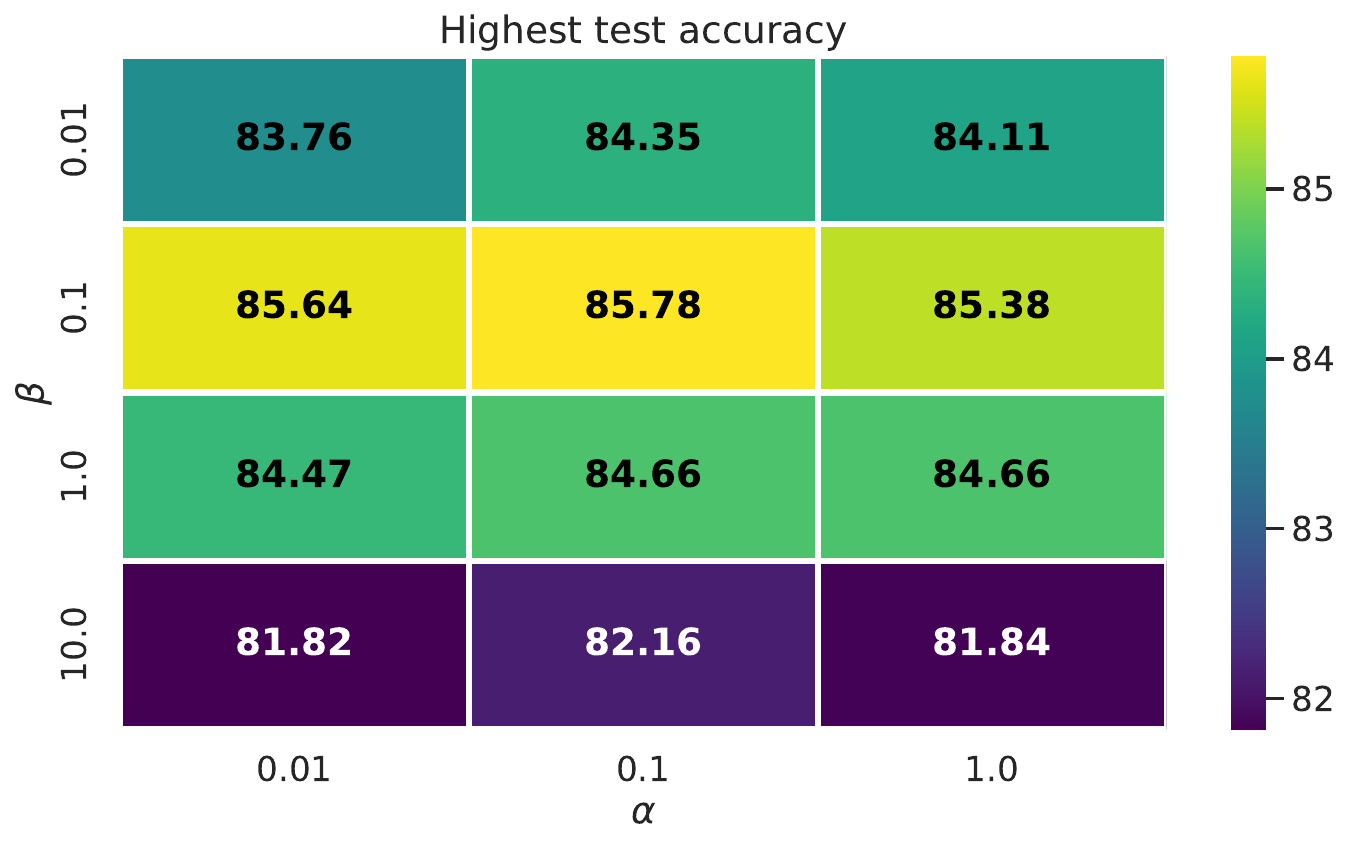}
\caption{The highest test accuracy achieved by \alg with different $\alpha$ and $\beta$ values.}
\label{smooth:fig:accuracy_heatmap}
\end{wrapfigure}
The performance of \alg is primarily governed by the choice of $\beta$.
From Figure~\ref{smooth:fig:accuracy_heatmap},  optimal performance ($85$-$86\%$ accuracy) is observed when $\beta$ is around $0.1$.
While \alg is stable with respect to $\alpha$, extreme values of $\beta$ lead to suboptimal performance: very large values ($\beta=10.0$) result in lower accuracy ($81$-$82\%$),
while very small values ($\beta=0.01$) achieve moderate performance ($83$-$84\%$).
The optimal configuration, achieving the highest $85.78\%$ accuracy, is  $\beta=0.1$ and $\alpha=0.1$.
Further analysis of the algorithm's stability with respect to $\alpha$ and robustness to $\beta$, including additional metrics (along with convergence curves) is provided in Appendix \ref{smooth:app:exp_sensitivity}.
For subsequent experiments, we set $\alpha=0.01$, which is consistent with recommendations from prior work in the single-node setting \citep{bu2024automatic}.

\textbf{Benefit of error compensation on \alg over \algname{DP-SGD}.}
From Figure \ref{smooth:fig:ef21_effect_loss}, \alg outperforms \algname{DP-SGD} with smoothed normalization (defined by Equation\eqref{smooth:eqn:DP_biased_GD} with $B = n$ and $z \equiv 0$).
This improvement is attributed to error compensation (EC), as confirmed by running \alg without server normalization (Line 11 of Algorithm~\ref{smooth:alg:norm_norm21_dp_v1}).
From Figure \ref{smooth:fig:ef21_effect_loss}, \alg achieves faster convergence than \algname{DP-SGD} with smoothed normalization for most $\beta$ values, with the exception of $\beta=10$.
However, such a large $\beta$ is impractical for differentially private training due to the resulting increase in noise variance.
Moreover, while our algorithm demonstrates robust performance across varying $\beta$ values, \algname{DP-SGD} with smoothed normalization exhibits greater sensitivity to this parameter, notably struggling with convergence at $\beta = 0.01$. This comparison underscores how EC not only accelerates convergence but also improves the algorithm's stability with respect to parameter selection.
Appendix \ref{smooth:app:exp_ef} presents further details (such as accuracy convergence curves in Figure \ref{smooth:fig:ef21_effect_acc}) and optimal parameters with corresponding final accuracies (Table \ref{smooth:tab:norm_comparison}).

An ablation study examining the impact of server normalization is provided in Appendix \ref{smooth:app:exp_serv_norm}. Furthermore, a comparison between \alg and \algname{Clip21} is presented in Appendix \ref{smooth:app:exp_clip_norm}.

\begin{figure}[tbp]
\centering
\begin{minipage}[t]{0.48\textwidth}
\centering
\includegraphics[width=\linewidth]{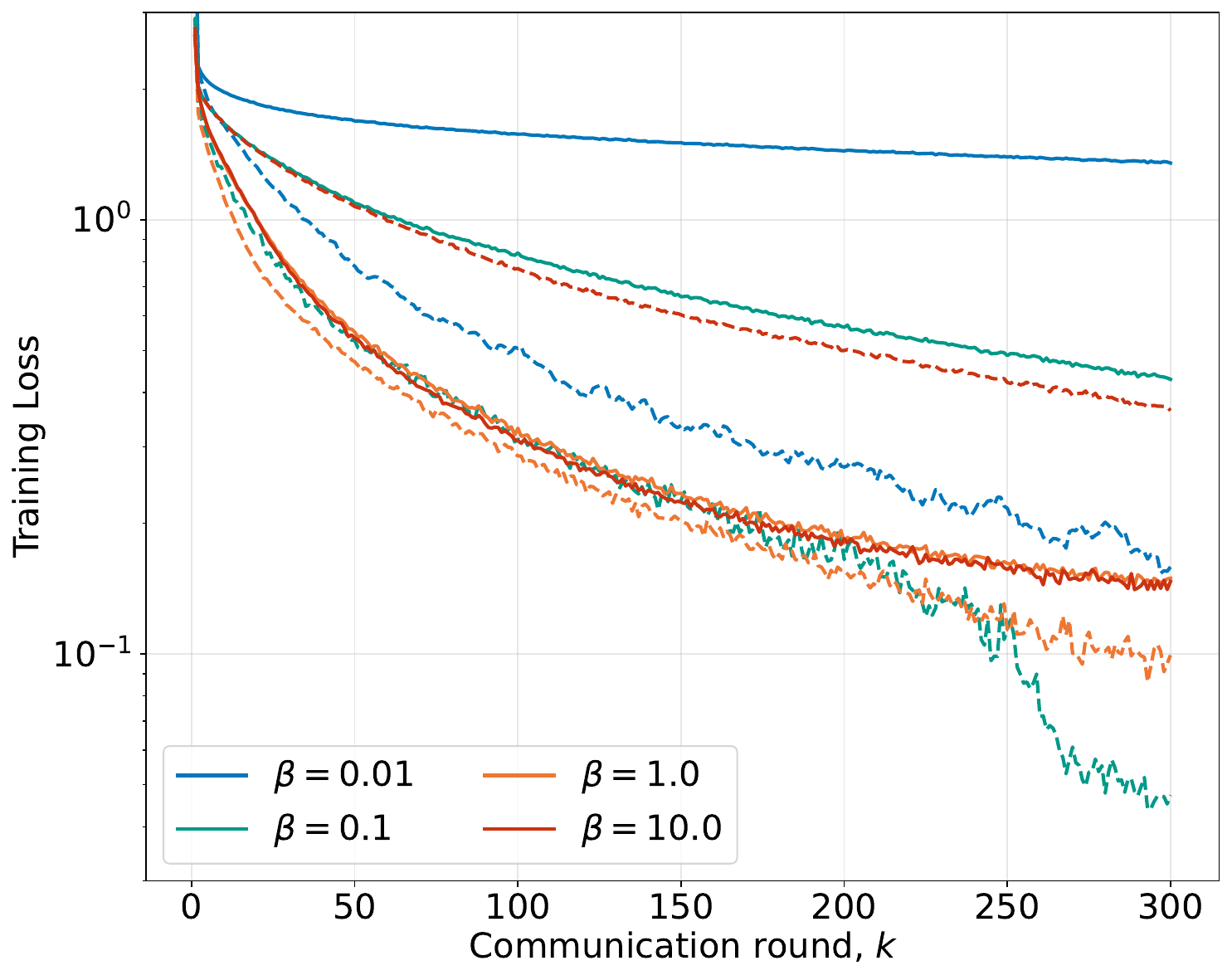}
\caption{Comparison of \algname{DP-SGD} \eqref{smooth:eqn:DP_biased_GD} [solid] and \alg (\ref{smooth:alg:norm_norm21_dp_v1}) [dashed] without server normalization. \alg converges faster for every $\beta$ value.}
\label{smooth:fig:ef21_effect_loss}
\end{minipage}
\hfill
\begin{minipage}[t]{0.48\textwidth}
\centering
\begin{overpic}[width=\linewidth]{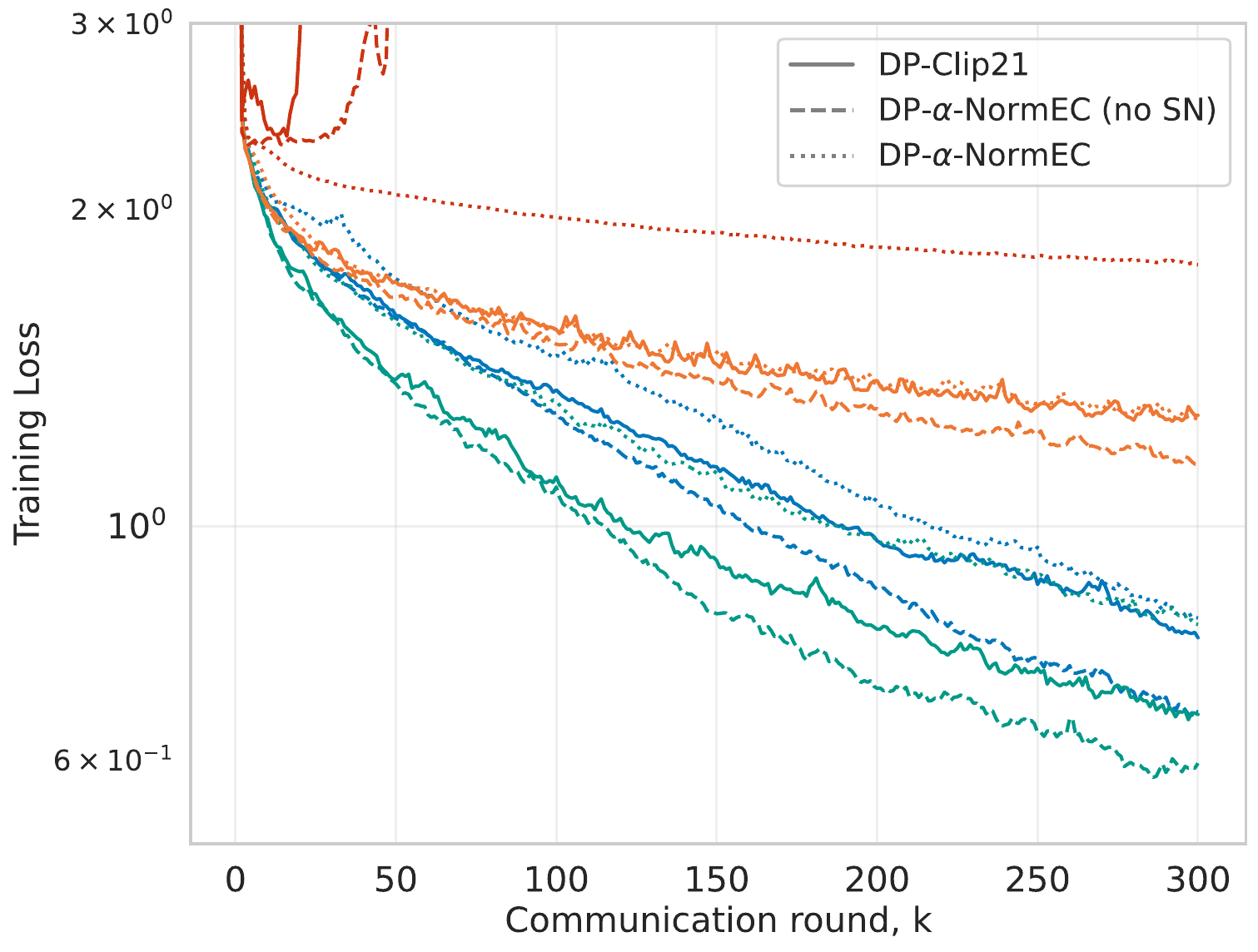}
    \put(16.5,9.5){\includegraphics[width=0.50\linewidth]{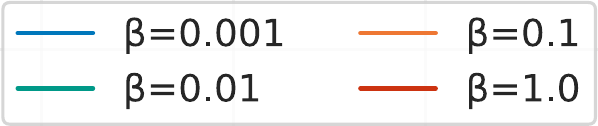}}
\end{overpic}
\caption{Comparison of methods in the Differentially Private (DP) setting across different $\beta$ values.}
\label{smooth:fig:dp_accuracy}
\end{minipage}
\end{figure}
\FloatBarrier

\subsection{Private training} \label{smooth:sec:private_exp}

We analyze the performance of \alg in the differentially private setting by
setting the standard deviation of added noise at $\beta\sqrt{K \log(1/\delta)}\epsilon^{-1}$ and vary $\beta$ to simulate different noise levels.
Following common DP deep learning evaluation conventions \citep{abadi2016deep,tang2023dprandp}, we report results at $\epsilon=8$ alongside stronger privacy regimes ($\epsilon=1$) in the Appendix \ref{smooth:app:small_eps}.
Figure~\ref{smooth:fig:dp_accuracy} shows the training loss curves for \algname{DP-}\alg (with and without server normalization) and \algname{DP-Clip21}\footnote{\color{black}\algname{DP-Clip21}, unlike \algname{Clip21}, does not have theoretical convergence guarantees.}.
Notably, compared to the non-private case, convergence in the DP setting is slower and requires a smaller $\beta$ (e.g., 0.01) for best performance.
From Figure~\ref{smooth:fig:dp_accuracy} we observe three key findings:
\textbf{(1)} \algname{DP-}\alg without Server Normalization converges significantly faster than \algname{DP-Clip21} at all privacy levels ($\beta \in \{0.001, 0.01, 0.1\}$);
\textbf{(2)} Server normalization (SN) provides crucial stability at high noise levels at $\beta=1.0$, only \algname{DP-}\alg with SN maintains convergence;
\textbf{(3)} While SN improves robustness, it comes with a slight reduction in convergence speed at lower noise levels.

The complete analysis, including test accuracy results across different hyperparameter combinations and detailed performance comparisons, is presented in Appendix~\ref{smooth:app:exp_dp}. Notably, server normalization significantly reduces performance variation across different learning rates ($\gamma$), with at most $6\%$ variation compared to $40\%$ without normalization, demonstrating improved hyperparameter robustness.
The superiority of \algname{DP-}\alg is further confirmed in a higher privacy regime with a stricter budget of $\epsilon=1$ in Appendix \ref{smooth:app:small_eps}.

\textbf{Logistic regression problem.}\par\noindent
\begin{wrapfigure}{r}{0.44\textwidth}
\centering
\includegraphics[width=\linewidth]{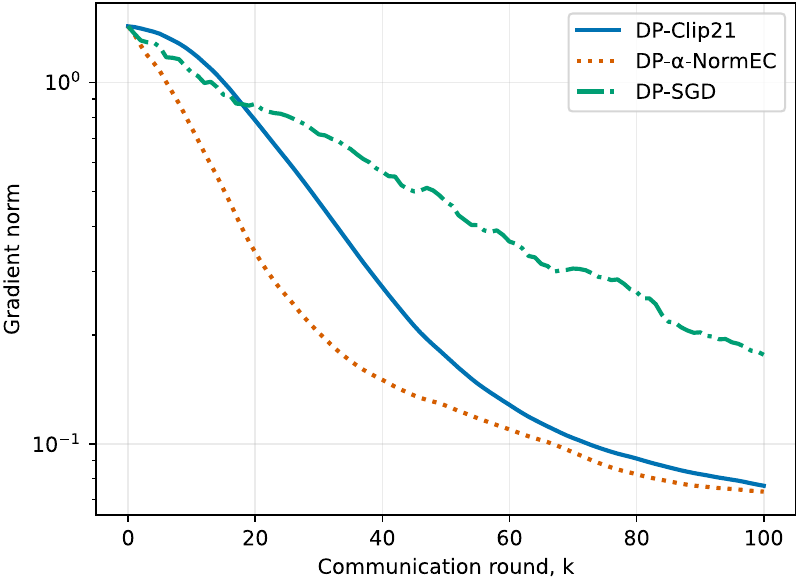}
\caption{Gradient norm $\|\nabla f(x^k)\|$ across methods on a logistic regression problem. Curves are averaged over $5$ seeds.}
\label{smooth:fig:log_reg_grad_norm}
\end{wrapfigure}
To complement the deep-learning experiments, we also evaluate the methods on a logistic regression task under $(\epsilon,\delta)$-DP with $\epsilon=3$ and $\delta=n^{-1.1}$ for $K=100$ rounds.

Figure~\ref{smooth:fig:log_reg_grad_norm} shows the evolution of the global gradient norm (log-scale): both error-feedback methods, \algname{DP-Clip21} and \algname{DP-}\alg, reduce the gradient norm $\|\nabla f(x^k)\|$ substantially faster than \algname{DP-SGD}, indicating more stable optimization under privatized updates.
Among the compared methods, \algname{DP-}\alg achieves the lowest final gradient norm and the highest final test accuracy (see Table~\ref{smooth:tab:reg_table}).
We report curves averaged over $5$ random seeds. Full experimental details, including the precise setup and hyperparameter-tuning grid, are provided in Appendix~\ref{smooth:sec:log_reg}.

\FloatBarrier
\section{Conclusion}

We have proposed and analyzed \alg, a novel distributed algorithm that integrates smoothed normalization with the \algname{EF21} mechanism for solving non-convex, smooth optimization problems in both non-private and private settings. Unlike \algname{Clip21}, \alg achieves strong convergence guarantees that match those of classical gradient descent for non-private training and provides the first utility bound for private training without relying on restrictive assumptions such as bounded gradient norms.
In neural network training, \alg achieves robust convergence across varying hyperparameters and significantly stronger convergence (due to error compensation) compared to \algname{DP-SGD} with smoothed normalization. In private training, \algname{DP-}\alg benefits from server normalization for increased robustness and outperforms \algname{DP-Clip21}.

 Our work implies many promising  directions.
 One direction is to extend \alg to accommodate the partial participation case, where the central server receives the updates from a subset of clients. %
 Another important direction is to modify \alg to solve federated learning problems, where the clients run their local updates before the local updates are normalized and transmitted to the central server.

\endgroup

\chapter{Gluon: Layer-Wise Generalized Smoothness for Muon-Style LMO Optimizers}
\label{chapter6:gluon}

\begingroup
\input{KAUST_Thesis/imported/06_gluon/import_setup}

\section*{Abstract}
    Recent developments in deep learning optimization have brought about radically new algorithms based on the Linear Minimization Oracle (LMO) framework, such as \algname{Muon} \citep{jordan2024muon} and \algname{Scion} \citep{pethick2025training}. After over a decade of \algname{Adam}'s dominance, these LMO-based methods are emerging as viable replacements, offering several practical advantages such as improved memory efficiency, better hyperparameter transferability, and most importantly, superior empirical performance on large-scale tasks, including LLM training. However, a significant gap remains between their practical use and our current theoretical understanding: prior analyses (1) overlook the layer-wise application of LMOs in these optimizers and (2) rely on an unrealistic smoothness assumption, leading to impractically small stepsizes. To address both issues, we propose a new LMO-based framework called \algname{Gluon}, which captures previously analyzed methods as special cases, and introduce a refined generalized smoothness model that captures the layer-wise geometry of neural networks, matches the layer-wise implementation of \algname{Muon} and \algname{Scion}, and leads to state-of-the-art convergence guarantees.
    Our experiments with \texttt{NanoGPT} and \texttt{CNN} confirm that our assumption holds along the optimization trajectory, ultimately closing the gap between theory and practice.

\begin{quote}
\small
\noindent\textit{This chapter is based on:}\\
Artem Riabinin, Egor Shulgin, Kaja Gruntkowska, and Peter Richt{\'a}rik, ``From Muon to Gluon: Bridging Theory and Practice of LMO-Based Optimizers for LLMs,'' \emph{International Conference on Machine Learning (ICML)}, 2026.
\nocite{riabinin2025gluon}
\end{quote}

\section{Introduction}

The success of deep learning models across a wide range of challenging domains is inseparable from the optimization algorithms used to train them. As neural networks have grown deeper and datasets larger, optimization has quietly become one of the most consequential components of modern machine learning (ML). Nowhere is this more evident than in the training of large language models (LLMs), which routinely consume thousands of GPU-hours.
\algname{Adam} \citep{kingma2015adam} (and lately \algname{AdamW} \citep{loshchilov2019decoupled})---being effective, relatively reliable, and widely adopted---has for over a decade served as the default choice for this task. While this reliance has powered much of deep learning’s progress, it has also exposed the shortcomings of adaptive moment estimation as a one-size-fits-all solution--namely, sensitivity to learning rate schedules, heavy tuning requirements \citep{wilson2017marginal}, and poor generalization when not carefully calibrated~\citep{zou2021understanding}. However, a shift may now be underway. Recent optimizers, such as \algname{Muon} \citep{jordan2024muon} and \algname{Scion} \citep{pethick2025training}, represent a significant departure from \algname{Adam}-type methods: they forgo the adaptive moment estimation in favor of a geometry-aware approach inspired by Frank-Wolfe algorithms \citep{frank1956algorithm,pokutta2024frank}. These optimizers are not only simpler to implement and easier to tune, but also appear empirically stronger, outperforming \algname{AdamW} in LLM training \citep{liu2025muon,pethick2025training}.

Yet, despite their potential, these new methods are still in their infancy, and our understanding of their theoretical foundations and practical utility in LLM training remains incomplete. Prior convergence guarantees in realistic nonconvex regimes are still far from satisfactory. Indeed, as we argue in \Cref{gluon:sec:impl_vs_an}, the (very few) existing theoretical analyses \emph{fail to capture the true algorithms used in practice}, focusing instead on simplified variants that diverge from actual implementations.
We identify two key mismatches---\emph{neglect of layer-wise structure} (\Cref{gluon:sec:layerwise}) and flawed stepsize choices stemming from an \emph{inaccurate smoothness model} (\Cref{gluon:sec:orig_steps})---and close this gap with a \emph{solution to both}. We elaborate on these advances in the remainder of the chapter.

\begin{algorithm}[t]
\caption{\algname{Gluon:} Stochastic Adaptive Layer-Wise LMO-based Optimizer with Momentum}
\label{gluon:algo:stochastic}
\begin{algorithmic}[1]
\STATE \textbf{Input:} Initial model parameters $X^0 = [X_1^0, \dots, X_p^0] \in \cS$, momentum $M^0=[M_1^0, \dots, M_p^0]\in \cS$, momentum decay factors $\beta^k \in [0,1)$ for all iterations $k\geq 0$
\FOR{$k = 0, 1, 2, \dots, K-1$}
    \STATE Sample $\xi^k \sim \mathcal{D}$
    \FOR{$i = 1, 2, \dots, p$}
        \STATE Compute stochastic gradient $\nabla_i f_{\xi^k}(X^k)$ for layer~$i$
        \STATE \label{gluon:step:momentum}Update momentum for layer $i$ via
        $$M_i^k=\beta^k M^{k-1}_i+(1-\beta^k)\nabla_i f_{\xi^k}(X^k)$$
        \STATE Choose adaptive stepsize/radius $t_i^k>0$ for layer $i$
        \STATE Update parameters for layer $i$ via LMO over $\mathcal{B}_i^k \eqdef \{ X_i \in \cS_i : \|X_i - X_i^k\|_{(i)} \leq t_i^k \}$ via
        \begin{equation}\label{gluon:eq:update_rule_stoch}
            X_i^{k+1} = \mathrm{LMO}_{\mathcal{B}_i^k} \left( M_i^k \right) := \underset{X_i \in \mathcal{B}_i^k}{\argmin} \ \langle M_i^k, X_i \rangle_{(i)}
        \end{equation}
    \ENDFOR
    \STATE Update full parameter vector
    $
    X^{k+1} = [X_1^{k+1}, \ldots, X_p^{k+1}]
    $
\ENDFOR
\end{algorithmic}
\end{algorithm}

Our goal is to solve the general optimization problem
\begin{eqnarray}\label{gluon:eq:problem}
    \min_{X \in\cS} \brac{f(X) := \ExpSub{\xi \sim \mathcal{D}}{f_{\xi}(X)}},
\end{eqnarray}
where $\cS$ is a finite-dimensional vector space and $f_{\xi}: \cS \to \mathbb{R}$ are potentially non-convex, continuously differentiable functions that are not necessarily smooth in the standard Lipschitz-gradient sense. Here, $f_\xi(X)$ represents the loss of the model parameterized by $X$ associated with a training data point $\xi$ sampled from the probability distribution~$\mathcal{D}$. To make the problem meaningful, we assume that $f^{\inf} \eqdef \inf_{X \in \cS} f(X) > - \infty$. In this work, we are particularly interested in the scenario in which the parameter vector $X\in \cS$ is obtained by collecting the matrices $X_i \in \cS_i \eqdef \R^{m_i \times n_i}$ of trainable parameters across all layers $i = 1,\ldots, p$ of a deep model. For simplicity, we therefore write $X = [X_1, \ldots, X_p]$. This means that, formally, $\cS$ is the $d$-dimensional product space
\begin{align*}
    \cS \eqdef \prod_{i = 1}^{p} \cS_i \equiv \cS_1 \times \cdots \times \cS_p,
\end{align*}
where $d\eqdef \sum_{i=1}^p m_i n_i $. With each space $\cS_i$ we associate the trace inner product  $\langle X_i, Y_i \rangle_{(i)} \eqdef \operatorname{tr}(X_i^{\top} Y_i)$ for $X_i,Y_i \in \cS_i$, and an arbitrary norm $\|\cdot\|_{(i)}$, not necessarily induced by the inner product.

\section{Theory vs. practice of Muon and Scion}\label{gluon:sec:impl_vs_an}

In this work, we focus on an algorithm based on iteratively calling linear minimization oracles (LMOs) across all layers, formalized in \Cref{gluon:algo:stochastic}, for which we coin the name \algname{Gluon}. In particular, for each layer $i$, independently across all layers, \algname{Gluon} iteratively updates the parameters  via
\begin{align*}
    X_i^{k+1} = \lmo{\mathcal{B}_i^k}{M_i^k}
    \eqdef \underset{X_i \in \mathcal{B}_i^k}{\arg\min} \ \langle M_i^k, X_i \rangle_{(i)},
\end{align*}
where $\cB_i^k \eqdef \{X_i \in  \cS_i : \|X_i - X_i^k\|_{(i)} \leq t_i^k \}$ and $t_i^k>0$ is an adaptively chosen stepsize/radius/learning rate.\footnote{In this context, the radii defining the norm balls in the LMOs effectively act as stepsizes--see \Cref{gluon:sec:step_note}. Accordingly, we use the terms \emph{radius}, \emph{stepsize}, and \emph{learning rate} interchangeably throughout.}
Note that the momentum  $M^k = [M_1^k, \dots, M_p^k] \in \cS$ accumulates the contributions from the stochastic gradients
\[
\nabla f_{\xi^k}(X^k)
= [\nabla_1 f_{\xi^k}(X^k),\ldots,\nabla_p f_{\xi^k}(X^k)] \in \cS
\]
(see Step \ref{gluon:step:momentum} of \Cref{gluon:algo:stochastic}).

The \algname{Gluon} framework generalizes a range of methods, including \algname{Muon} and \algname{Scion}, which are recovered as special cases under specific norm choices (see \Cref{gluon:sec:det_examples} and \Cref{gluon:sec:lmo_examples}).
Beyond their ability to outperform \algname{AdamW} on large-scale benchmarks, these optimizers offer a number of attractive properties: improved memory efficiency, greater robustness to hyperparameter settings, and the ability to transfer those settings across model sizes \citep{pethick2025training,shah2025practical}. Moreover, in contrast to \algname{Adam}, they were theoretically analyzed shortly after release and are guaranteed to converge under standard assumptions of Lipschitz smoothness\footnote{A function $f:\cS\to\R$ is $L$-smooth if $\norm{\nabla f(x) - \nabla f(y)}_{\star} \leq L \norm{x-y}$ for all $x,y\in \cS$, where $\cS$ is a finite-dimensional vector space equipped with a norm $\|\cdot\|$ and $\|\cdot\|_{\star}$ is the dual norm associated with $\|\cdot\|$.} and bounded variance of stochastic gradients \citep{kovalev2025muon,li2025noteconvergencemuon,pethick2025training}.

\algname{Gluon} presents the method that is deployed in practice \citep{jordan2024moddednanogpt,code_scion} and has proven highly effective. That said, we argue that existing analyses \citep{kovalev2025muon,li2025noteconvergencemuon,pethick2025training} do \emph{not} accurately reflect this implementation, diverging from it in two key ways. As such, they \emph{fail to explain why the algorithm performs so well}. Let us detail why.

\subsection{Layer-wise structure}\label{gluon:sec:layerwise}

\begin{figure}[t]
    \centering
    \makebox[\textwidth][c]{%
    \subfigure[Token embedding matrix from the first/last layer.]{%
        \label{gluon:fig:embedding_ex}
        \begin{minipage}[t]{0.315\textwidth}
            \centering
            \includegraphics[width=\linewidth,height=0.17\textheight,keepaspectratio]{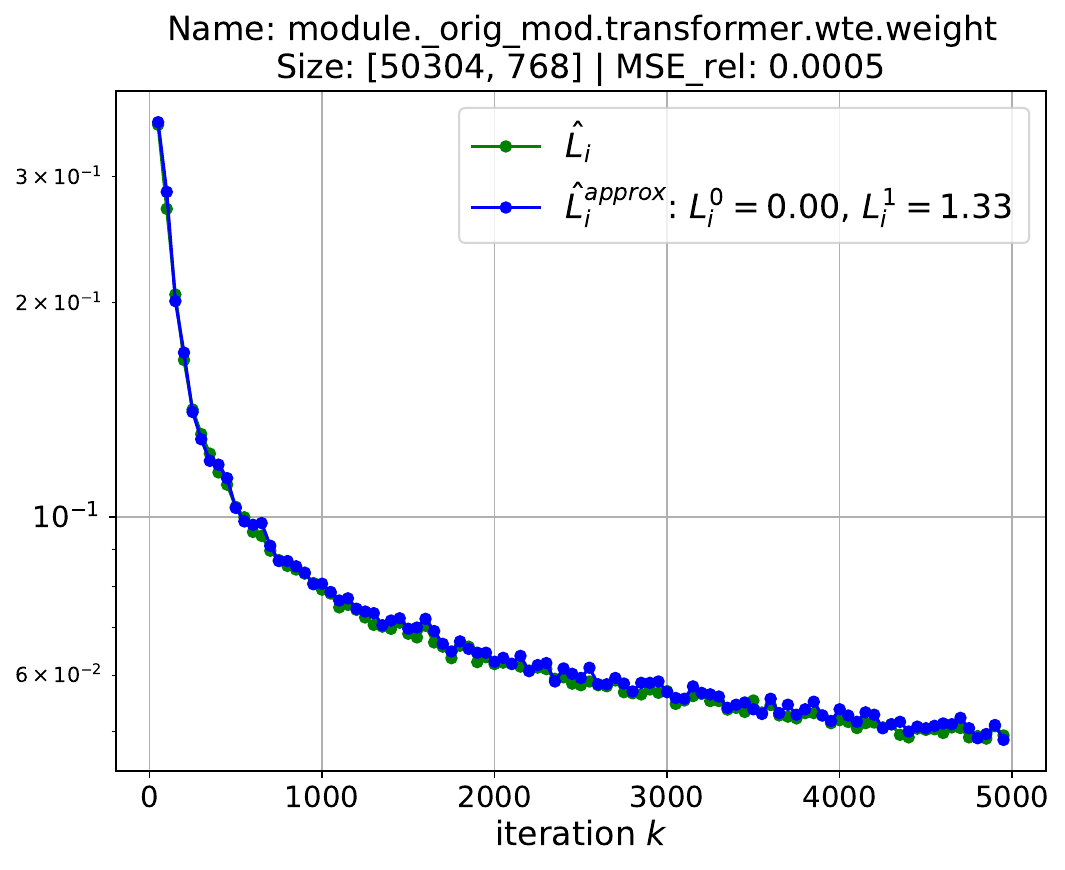}
        \end{minipage}}%
    \hspace{0.01\textwidth}%
    \subfigure[Self-attention query matrix from the 4th transformer block.]{%
        \label{gluon:fig:q_ex}
        \begin{minipage}[t]{0.315\textwidth}
            \centering
            \includegraphics[width=\linewidth,height=0.17\textheight,keepaspectratio]{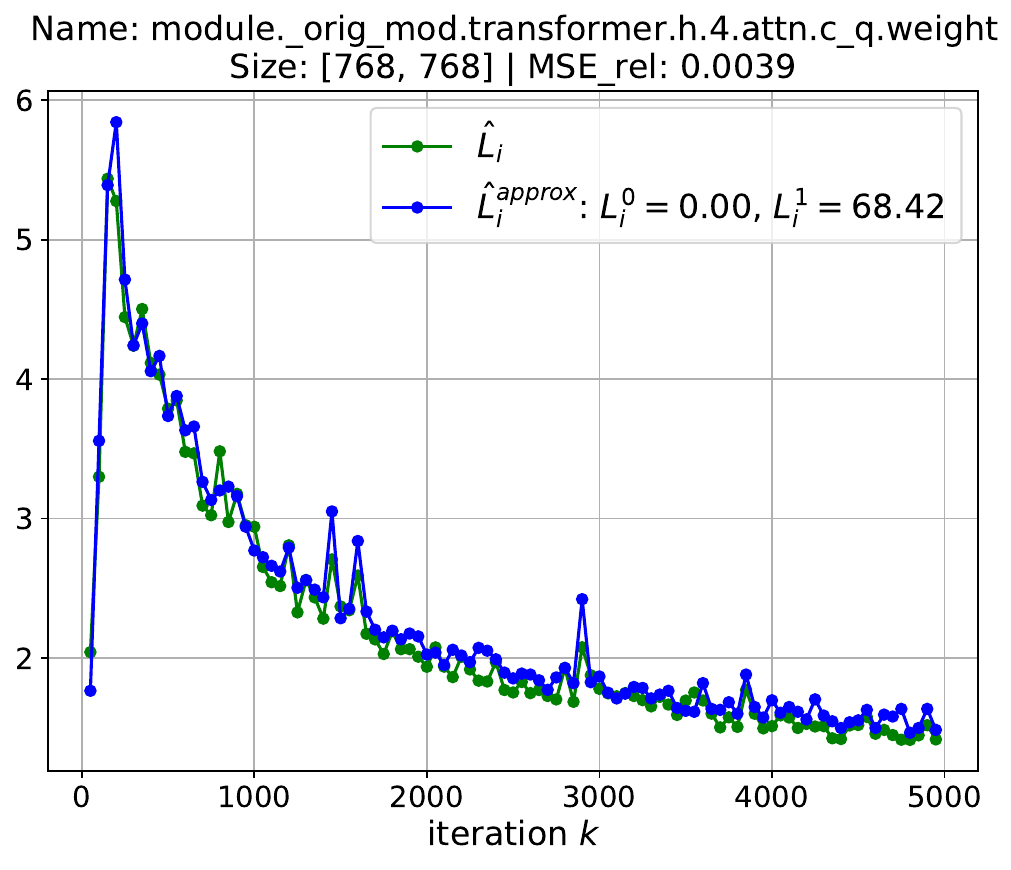}
        \end{minipage}}%
    \hspace{0.01\textwidth}%
    \subfigure[Trajectory smoothness across different blocks (B.$i$) and layers.]{%
        \label{gluon:fig:layer_smooth}
        \begin{minipage}[t]{0.315\textwidth}
            \centering
            \includegraphics[width=\linewidth,height=0.17\textheight,keepaspectratio]{KAUST_Thesis/imported/06_gluon/plots/nanogpt/smoothness_selected_layers.pdf}
        \end{minipage}}%
    }%
    \caption{Training \texttt{NanoGPT} on \texttt{FineWeb} validates our layer-wise $(L^0, L^1)$-smoothness model.}
    \label{gluon:fig:intro}
\end{figure}

First, we briefly walk through the theoretical understanding offered by previous studies. \algname{Muon} is an optimizer specifically designed for hidden layers, leaving the first and last layers to be handled by some other optimizer, e.g., \algname{Adam(W)}. Its original introduction by \citet{jordan2024muon} was purely empirical, with no attempt at theoretical analysis. The first convergence result came from \citet{li2025noteconvergencemuon}, who analyzed the smooth nonconvex setting but focused solely on problem~\eqref{gluon:eq:problem} with $p = 1$, effectively limiting the scope to the single-layer case. The \algname{Scion}\footnote{\citet{pethick2025training} introduce two variants of the \algnamefootnote{Scion} optimizer: one for constrained optimization, called simply ``\algnamefootnote{Scion}'', and another for unconstrained problems, referred to as ``unconstrained \algnamefootnote{Scion}''. In this work, ``\algnamefootnote{Scion}'' refers to either variant, and ``\algnamefootnote{unScion}'' is used when referring to the unconstrained version.} optimizer (a special case of \algname{Gluon}) proposed by \citet{pethick2025training} improves upon \algname{Muon} by applying the LMO-based rule to all layers, ultimately achieving better empirical performance. Both this work and that of \citet{kovalev2025muon} analyze (a variant of) the general update rule
\begin{align} \label{gluon:eq:orig_muon_upd}
    \begin{split}
    M^k &= \beta^k M^{k-1} + (1-\beta^k) \nabla f_{\xi^k}(X^k), \\
    X^{k+1} &= \lmo{\cB^k}{M^k},
    \end{split}
\end{align}
where $\beta^k \in [0,1)$ is momentum, $\nabla f_{\xi^k}(X^k)$ is the stochastic gradient sampled at iteration $k$, and $\cB^k \eqdef \{X \in \cS: \|X - X^k\| \leq t^k \}$ is a norm ball centered at $X^k$ with stepsize $t^k>0$. This setup closely resembles the structure of \algname{Gluon}, but is \emph{not} exactly the same.
Indeed, \algname{Gluon} updates the parameters \emph{layer-wise}, not jointly over the full vector~$X$.
This distinction is critical since, for practical, extremely high-dimensional models, calculating a single global LMO for the entire parameter vector is prohibitively expensive, while breaking the problem into ``smaller'', per-layer {\rm LMOs} restores computational feasibility.

Motivated by this disconnect, we formulate our analysis in the matrix product space $\cS$, explicitly honoring the layer-wise structure. This enables us to study the actual per-layer updates \eqref{gluon:eq:update_rule_stoch}, with assumptions and hyperparameters adapted to each layer.

\subsection{A theory with predictive power}\label{gluon:sec:orig_steps}

All prior works claiming to guarantee convergence of \Cref{gluon:algo:stochastic} come with several serious analytical shortcomings--and these directly translate into practical deficiencies. Concretely, all existing analyses of \algname{Muon}/\algname{Scion} are built on the classical $L$-smoothness assumption, imposing a uniform smoothness constant across all layers. This is problematic, as \emph{different layers have different geometries}, and thus \emph{should be treated differently}.

But the issue runs much deeper. These algorithms are built for deep learning, where the objective functions are already well known \emph{not} to be smooth \citep{crawshaw2022robustness,zhang2020why}. This mismatch has consequences: prior convergence analyses prescribe \emph{tiny constant stepsizes} (see \Cref{gluon:table1}), uniform across all parameter groups, which bear little resemblance to the tuned learning rates that yield state-of-the-art empirical performance in practice. Consequently, they completely fail to explain why these methods perform so well empirically.
In other words, the theory falls short at the one thing it should do best: guiding practical choices, leaving practitioners reliant on costly manual tuning.

Our result in \Cref{gluon:theorem:1} shows this mismatch is \emph{not} inevitable. To better reflect the behavior of deep models, we introduce a more expressive regularity condition: the \emph{layer-wise $(L^0, L^1)$-smoothness}\footnote{While we state \Cref{gluon:ass:generalized-smoothness} in this general form, it is worth noting that the proofs do not rely on its full strength. In all cases, we only require the assumption to hold for pairs $X$, $Y$ such that $\norm{X-Y} < c$ for some constant $c\geq0$ (where $\norm{\cdot}$ is any norm on $\cS$). Specifically, the assumption is only invoked with $X=X^k$, $Y=X^{k+1}$, and since the stepsizes we use are bounded, the distances between consecutive iterates remain bounded as well. For clarity and consistency across results--since the relevant constants vary by theorem--we state the assumption in its stronger, global form, even though the local version suffices for all proofs.}--an extension of the generalized smoothness model of \citet{zhang2020why}, applied at the layer level.

\begin{assumption}[Layer-wise $(L^0, L^1)$-smoothness]\label{gluon:ass:generalized-smoothness}
    The function $f: \cS \to \mathbb{R}$ is layer-wise $(L^0, L^1)$-smooth with constants $L^0 \eqdef (L^0_1, \ldots, L^0_p)\in \R^p_+$ and $L^1 \eqdef (L^1_1, \ldots, L^1_p)\in \R^p_+$. That is, the inequality
    \begin{equation}
       \| \nabla _i f(X) - \nabla _i f(Y) \|_{(i) \star}
       \leq \left( L^0_i + L^1_i \| \nabla _i f(X) \|_{(i) \star} \right) \|X_i - Y_i \|_{(i)} \label{gluon:eq:09yhfd98y9fd}
    \end{equation}
    holds for all $i=1,\dots,p$ and all $X = [X_1, \ldots, X_p]\in \cS$, $Y = [Y_1, \ldots, Y_p] \in \cS$, where $\|\cdot\|_{(i) \star}$ is the dual norm associated with~$\|\cdot\|_{(i)}$ (i.e., $\|X_i\|_{(i) \star} \eqdef \sup_{\norm{Z_i}_{(i)} \leq 1} \inp{X_i}{Z_i}_{(i)}$ for any $X_i\in \cS_i$).
\end{assumption}

Assumption~\ref{gluon:ass:generalized-smoothness} can be viewed as a generalization of the anisotropic \say{vector} $(L^0, L^1) $--smoothness introduced by \citet{liu2024adagradanisotropicsmoothness} (now framed in terms of arbitrary norms), which itself is a generalization of the $(L^0, L^1) $--smoothness model of \citet{zhang2020why}. As such, our analysis of \algname{Gluon} goes beyond all existing results, which have only considered the classical $L$-smooth setting.
Crucially, however, this is \emph{not} generalization for its own sake--we argue that this is in fact \emph{the right} model for the problem setting at hand. Why? There are (at least) two reasons.

First, unlike classical $L$-smoothness, our formulation \emph{aligns very closely with empirical observations}.
In Figures \ref{gluon:fig:embedding_ex} and \ref{gluon:fig:q_ex}, we validate Assumption~\ref{gluon:ass:generalized-smoothness} in the context of training \texttt{NanoGPT} on the \texttt{FineWeb} dataset. We plot estimated \emph{trajectory smoothness} $\hat{L}_{i}[k]$ (defined in \eqref{gluon:eq:dir_smooth}) alongside the approximation $\hat{L}_{i}^{\text{approx}}[k] \eqdef L_i^0 + L_i^1 \|\nabla_i f_{\xi^{k+1}} (X^{k+1})\|_{(i) \star}$, where $L^0_i, L^1_i$ are layer-specific parameters estimated from the training run.
The figures show these quantities for parameters from the embedding layer and one of the transformer blocks. The close correspondence between $\hat{L}_i[k]$ and $\hat{L}_{i}^{\text{approx}}[k]$ provides strong evidence that Assumption~\ref{gluon:ass:generalized-smoothness} holds approximately along the training trajectory.
In \Cref{gluon:sec:experiments}, we further corroborate this finding, showing that our assumption is satisfied \emph{across the entire model architecture} for both the \texttt{NanoGPT} language modeling task and a \texttt{CNN} trained on \texttt{CIFAR-10}. In all cases, we find that $L^0_i\approx 0$ for all $i$, again highlighting the limitations of classical smoothness. Moreover, as shown in \Cref{gluon:fig:layer_smooth}, trajectory smoothness varies substantially across blocks and layers, underscoring the need for per-layer treatment. Complementary experiments using \algname{AdamW} as the optimizer (\Cref{gluon:fig:Adam_app}) confirm that this heterogeneity is an intrinsic property of the loss landscape.
Together, these results suggest that layer-wise $(L^0, L^1)$-smoothness offers a \emph{significantly more realistic model of the loss landscape in modern deep learning}.

Second, \Cref{gluon:ass:generalized-smoothness} not only better captures the geometry of the models, but also \emph{directly informs the design of adaptive and practically effective stepsizes}. In \Cref{gluon:theorem:1}, we derive learning rates that reflect the local geometry of each parameter group, guided by our layer-wise smoothness model.
As demonstrated in \Cref{gluon:sec:experiments}, our theoretically grounded stepsizes turn out to accurately capture the relative magnitudes of the layer-wise learning rates obtained by \citet{pethick2025training} via hyperparameter tuning--a striking validation of our approach, which further highlights the need for layer-wise reasoning.
This proves that \emph{theoretical stepsizes can have predictive power} and effectively guide hyperparameter tuning.

\section{Contributions}

We present a comprehensive theoretical and empirical study of a broad class of layer-wise LMO-based optimization algorithms. Our key contributions can be summarized as follows:

$\diamond$ \textbf{A new generalized smoothness framework for neural networks.}
We introduce \emph{layer-wise $(L^0, L^1)$-smoothness} (\Cref{gluon:ass:generalized-smoothness}), a novel non-Euclidean generalized smoothness condition that reflects the anisotropic, layer-wise structure of modern deep networks. This framework extends the standard $(L^0, L^1)$-smoothness assumption \citep{zhang2020why} to arbitrary norms while capturing per-layer variation, offering a \emph{realistic foundation for analyzing deep learning optimizers}.

$\diamond$ \textbf{First principled analysis of layer-wise methods.}
Building on our new assumption, we develop the first faithful convergence analysis for a class of LMO-based algorithms we term \algname{Gluon} (Algorithms~\ref{gluon:algo:stochastic} and~\ref{gluon:algo:deterministic}). We recover known algorithms, including state-of-the-art \algname{Muon}-type optimizers, as special cases (\Cref{gluon:sec:det_examples} and \Cref{gluon:sec:lmo_examples}), and pinpoint why earlier theoretical works \emph{fail} to explain the empirical success of these methods (\Cref{gluon:sec:impl_vs_an}). In contrast to prior analyses that oversimplify the update rules used in practice, our framework directly aligns with real-world implementations, bridging a critical gap between theory and application.

$\diamond$ \textbf{Sharper and more general convergence theory.}
We develop a convergence theory that extends prior work in both scope and sharpness. In the deterministic case (\Cref{gluon:algo:deterministic}), we establish convergence for general non-convex objectives under our \Cref{gluon:ass:generalized-smoothness} (\Cref{gluon:theorem:1}), and under the block-wise PŁ condition (\Cref{gluon:theorem:1-pl}). Unlike earlier analyses, our theory yields \emph{adaptive, layer-wise stepsizes} that can guide real-world training decisions (\Cref{gluon:sec:experiments}).
We next analyze the practical stochastic variant with time-varying stepsizes and momentum (\Cref{gluon:algo:stochastic}), proving convergence under a bounded-variance assumption (\Cref{gluon:theorem:1}).
In both deterministic and stochastic regimes, our guarantees offer \emph{tighter convergence rates} under \emph{more general assumptions} (\Cref{gluon:table1}), providing the first such results in the nonsmooth setting.
Moreover, we provide the first theoretical explanation of the benefits of layer-wise learning rates, clearly establishing the advantages of structured, anisotropic optimization in deep learning.

$\diamond$ \textbf{Empirical evidence.}
We validate our theoretical insights through extensive experiments (\Cref{gluon:sec:experiments} and \Cref{gluon:appendix:more_exp}) in both language modeling (\texttt{NanoGPT} on \texttt{FineWeb}) and image classification (\texttt{CNN} on \texttt{CIFAR-10}). The results confirm that our \Cref{gluon:ass:generalized-smoothness} holds approximately throughout training and demonstrate the practical utility of our theoretically prescribed stepsizes from \Cref{gluon:theorem:1}.

\begin{table*}
\caption{%
Comparison of convergence guarantees for \algname{Gluon} (Algorithms \ref{gluon:algo:stochastic} and \ref{gluon:algo:deterministic}) to achieve $\min_{k=0,\dots,K-1} \sum_{i=1}^p \mathbb{E}[\| \nabla_i f(X^k) \|_{(i)\star}] \leq \varepsilon$, where the $ \mathcal{O}(\cdot) $ notation hides logarithmic factors.
Notation: $K$ = total number of iterations, $(L^0, L^1)$ = the result holds under layer-wise $(L^0, L^1)$-smoothness, $t_i^k$ = radius/stepsize, $1-\beta^k$ = momentum.
}
\label{gluon:table1}
\centering
\begin{adjustbox}{width=\textwidth,center}
\begin{threeparttable}
\begin{tabular}[H]{cccccc}
    \toprule
    \textbf{Result} & \textbf{Stochastic?} & \textbf{$(L^0, L^1)$} & \textbf{Rate} & \textbf{Stepsizes $t_i^k$} & \textbf{$1-\beta^k$} \\
    \midrule
    \makecell{\citep[Theorem 1]{kovalev2025muon}} & \xmark & \xmark & $\mathcal{O}\parens{\frac{1}{K^{1/2}}}$ & $\textnormal{const} \propto {\frac{1}{K^{1/2}}}$\textsuperscript{(b)} & --- \\
    \makecell{\citep[Theorem 2]{kovalev2025muon}} & \cmark & \xmark & $\mathcal{O}\parens{\frac{1}{K^{1/4}}}$ & $\textnormal{const} \propto {\frac{1}{K^{3/4}}}$\textsuperscript{(b)} & $\textnormal{const} \propto {\frac{1}{K^{1/2}}}$ \\
    \makecell{\citep[Theorem 2.1]{li2025noteconvergencemuon}\textsuperscript{(a)}} & \cmark & \xmark & $\mathcal{O}\parens{\frac{1}{K^{1/4}}}$ & $\textnormal{const} \propto {\frac{1}{K^{3/4}}}$\textsuperscript{(b)} & $\textnormal{const} \propto {\frac{1}{K^{1/2}}}$ \\
    \makecell{\citep[Lemma 5.4]{pethick2025training}} & \cmark & \xmark & $\mathcal{O}\parens{\frac{1}{K^{1/4}}}$ & $\textnormal{const} \propto {\frac{1}{K^{3/4}}}$\textsuperscript{(b)} & $\propto {\frac{1}{k^{1/2}}}$ \\
    \cellcolor{bgcolor}  {\bf NEW: Theorem}~\ref{gluon:theorem:1} & \cellcolor{bgcolor} \xmark & \cellcolor{bgcolor} \cmark & \cellcolor{bgcolor} $\mathcal{O}\parens{\frac{1}{K^{1/2}}}$ & \cellcolor{bgcolor}  Adaptive & \cellcolor{bgcolor} ---\\
    \cellcolor{bgcolor} {\bf NEW: Theorem}~\ref{gluon:theorem:3}  & \cellcolor{bgcolor} \cmark & \cellcolor{bgcolor} \cmark & \cellcolor{bgcolor} $\mathcal{O}\parens{\frac{1}{K^{1/4}}}$ & \cellcolor{bgcolor} $\propto {\frac{1}{k^{3/4}}}$ & \cellcolor{bgcolor} $\propto {\frac{1}{k^{1/2}}}$ \\
    \bottomrule
    \end{tabular}
    \begin{tablenotes}
        \item [(a)] Applies only to the \algnametiny{Muon}/\algnametiny{Scion} update in \eqref{gluon:eq:muon} with $p=1$.
        \item [(b)] These stepsizes are impractically tiny since they have an inverse dependence on the total number of iterations $K$.
    \end{tablenotes}
\end{threeparttable}
\end{adjustbox}
\end{table*}

\section{Main theory and results}\label{gluon:sec:deterministic}

To gain a better intuition into the structure of the updates, we begin with a deterministic formulation of \algname{Gluon}, formalized in Algorithm~\ref{gluon:algo:deterministic}. At each iteration, the method independently minimizes a linear approximation of~$f$ around each parameter group~$X^k_i$ within a ball of radius~$t^k_i > 0$, ultimately allowing for layer-specific algorithmic design choices.

\subsection{Examples of optimizers within our framework}\label{gluon:sec:det_examples}

Deterministic \algname{Gluon} describes a general class of methods, parameterized by the choice of norms $\|\cdot\|_{(i)}$ in the LMO. To illustrate the flexibility of this framework, we highlight several notable special cases (see \Cref{gluon:sec:lmo_examples} for more details).
First, observe that the update rule \eqref{gluon:eq:update_rule} can be written as
\begin{equation}
\begin{aligned}
    X_i^{k+1} &= X_i^{k} + t_i^k \mathrm{LMO}_{\{ X_i \in \cS_i : \|X_i\|_{(i)} \leq 1 \}} \left( \nabla _i f(X^k) \right) \\
    &= X_i^{k} + t_i^k \underset{\left\|X_i\right\|_{(i)} \leq 1}{\operatorname{argmin}} \langle\nabla _i f(X^k), X_i \rangle_{(i)}. \label{gluon:eq:b9f8dgfd}
\end{aligned}
\end{equation}
For any $X_i \in \cS_i = \R^{m_i \times n_i}$, define  $\|X_i\|_{\alpha \rightarrow \beta} \eqdef \sup _{\|z\|_{\alpha}=1} \|X_i z\|_{\beta}$, where  $\|\cdot\|_{\alpha}$ and $\|\cdot\|_{\beta}$ are some (possibly non-Euclidean) norms on $\R^{n_i}$ and $\R^{m_i}$, respectively.
Note that \eqref{gluon:eq:b9f8dgfd} naturally recovers several known updates for specific choices of the layer norms, e.g., layer-wise normalized \algname{GD} \citep{wei2018blocknormalized} for Euclidean norms $ \|\cdot\|_{(i)}=\|\cdot\|_{2} $, and layer-wise \algname{signGD} \citep{balles2020geometrysigngradientdescent} for max-norms $\|\cdot\|_{(i)}=\|\cdot\|_{\infty}$.
Two special cases are particularly relevant to our analysis:

$\diamond$ \textbf{Muon} \citep{jordan2024muon} when $ \|\cdot\|_{(i)} = \|\cdot\|_{2 \rightarrow 2} $ for all hidden layers.

$\diamond$ \textbf{unScion for LLM training} \citep{pethick2025training}
when $\|\cdot\|_{(i)} = \sqrt{\nicefrac{n_i}{m_i}} \|\cdot\|_{2 \to 2}$ for $ i = 1, \dots, p-1 $, corresponding to weight matrices of transformer blocks, and $\|\cdot\|_{(p)} = n_p \|\cdot\|_{1 \to \infty}$ for the last group~$ X_p $, representing the embedding and output layers (the two coincide under the weight sharing regime\footnote{Weight sharing refers to the practice of using the same parameters (weights) for different parts of a model, rather than allowing each part to have its own unique parameters.} considered here).
In this case, update \eqref{gluon:eq:b9f8dgfd} becomes
\begin{align}\label{gluon:eq:scion_llm_main}
    X_i^{k+1} &=  X_i^k - t^k_i \sqrt{\frac{m_i}{n_i}} U_i^k \left(V_i^k\right)^{\top}, \quad i = 1, \dots, p-1, \nonumber \\
    X_p^{k+1} &= X_p^k - \frac{t^k_p}{n_p} \text{sign}\left(\nabla _p f(X^k)\right),
\end{align}
where the matrices $U^k_i, V^k_i$ are obtained from the (reduced) SVD of $\nabla _i f(X^k) = U^k_i \Sigma^k_i \left(V^k_i\right)^{\top}$.

\subsection{Convergence results}
\label{gluon:sec:convergence_deterministic}

Having demonstrated the framework’s flexibility through concrete examples, we now state a general convergence result for deterministic \algname{Gluon}.

\begin{theorem}\label{gluon:theorem:1}
    Let \Cref{gluon:ass:generalized-smoothness} hold and fix $\varepsilon>0$. Let $X^0,\ldots,X^{K-1}$ be the iterates of deterministic \algname{Gluon} (Algorithm \ref{gluon:algo:deterministic}) run with stepsizes $t^k_i = \frac{\|\nabla _i f(X^k)\|_{(i) \star}}{L^0_i + L^1_i\|\nabla _i f(X^k)\|_{(i) \star}}$.
    Then, to guarantee that
    \begin{align}\label{gluon:eq:det_precision}
        \min\limits_{k=0,\ldots,K-1} \sum\limits_{i=1}^p \left[\frac{\nicefrac{1}{L_i^1}}{\frac{1}{p} \sum_{j=1}^p \nicefrac{1}{L_j^1}} \norm{\nabla _i f(X^k)}_{(i) \star}\right] \leq \varepsilon,
    \end{align}
    it suffices to run the algorithm for
    \begin{align}\label{gluon:eq:det_rate}
        K = \left\lceil \frac{2 \Delta^0 \parens{\sum_{i=1}^p \nicefrac{L^0_i}{(L_i^1)^2}}}{\varepsilon^2 \parens{\frac{1}{p} \sum_{j=1}^p \nicefrac{1}{L_j^1}}^2}
        + \frac{2 \Delta^0}{\varepsilon \parens{\frac{1}{p} \sum_{j=1}^p \nicefrac{1}{L_j^1}}} \right\rceil
    \end{align}
    iterations, where $\Delta^0\eqdef f(X^0) - f^{\inf}$.
\end{theorem}

Several important observations follow.

\textbf{Convergence rate.} In \Cref{gluon:sec:det_proof}, we prove an additional result (\Cref{gluon:theorem:1_full}) that modifies the first term in \eqref{gluon:eq:det_rate} to $\nicefrac{2 \Delta^0 \sum_{i=1}^p L^0_i}{\epsilon^2}$, potentially leading to improvements in certain settings (depending on the relationship between the sequences $\{L^0_i\}$ and $\{L^1_i\}$--see \Cref{gluon:rem:det}). However, this introduces a dependence on $L^1_{\max}\eqdef\max_{i=1,\dots,p} L^1_i$ in the second term.
Empirically, we find that $L^0_i \approx 0$ across all layers (see \Cref{gluon:sec:experiments}), making the first term vanish in both bounds. In this case, the rate~\eqref{gluon:eq:det_rate} is clearly superior, replacing the worst-case constant $L^1_{\max}$ with the more favorable harmonic mean.

When $p=1$, our rates match the best-known complexity for finding a stationary point of $(L^0, L^1)$-smooth functions, $\mathcal{O}\left(\nicefrac{L^0 \Delta^0}{\epsilon^2} + \nicefrac{L^1 \Delta^0}{\epsilon}\right)$, as established by \citet{vankov2025optimizing} for the Gradient Method.
While no prior work has analyzed deterministic \algname{Gluon} under general $(L^0, L^1)$-smoothness, there exist analyses under classical $L$-smoothness, treating the parameters as a single vector. The analysis by \citet{kovalev2025muon} guarantees convergence in $K = \left\lceil \nicefrac{6 L \Delta^0}{\epsilon^2} \right\rceil$ iterations. The same bound appears in \citet{li2025noteconvergencemuon} and \citet{pethick2025training} (by setting $\sigma^2 = 0$). Since for $p=1$, $L$-smoothness implies \Cref{gluon:ass:generalized-smoothness} with $L^1 = 0$ (\Cref{gluon:lemma:L0vsL}), our rates match these prior results up to a constant factor. Thus, even in the smooth setting, our bounds are as tight as those derived specifically for it.

However, the real strength of our guarantees lies in their broader applicability. Our analysis is much more general than prior studies, as it extends beyond standard smoothness--allowing $L_i^1 > 0$ introduces additional terms that drive the accelerated convergence enabled by $(L^0, L^1)$-smoothness. This richer model is \emph{essential for explaining the empirical speedup} of methods like \algname{Muon}, and much more accurately reflects the geometry of neural network loss surfaces. Indeed, as we demonstrate in \Cref{gluon:sec:experiments}, the assumption typically holds with $L^0_i\approx 0$ and $L_i^1 > 0$.

\textbf{Practical radii $t^k_i$.} Unlike previous analyses \citep{kovalev2025muon,li2025noteconvergencemuon,pethick2025training}, which prescribe impractically small constant radii proportional to $ \epsilon $, our framework allows $ t^k_i $ to be \emph{adaptive} to the loss landscape. Therefore, $ t^k_i $ can be larger early in training when $ \|\nabla _i f(X^{k})\|_{(i) \star} $ is large and gradually shrink as the gradient norm decreases. In the special case in which $ L^0_i \approx 0 $ (as observed empirically), $ t^k_i \approx \nicefrac{1}{L^1_i} $, which is substantially larger than the radii dictated by earlier analyses.

\subsection{Stochastic case}
\label{gluon:sec:convergence_stochastic}

In practice, computing full gradients is often infeasible due to the scale of modern ML problems. We therefore turn to the practical \algname{Gluon} (\Cref{gluon:algo:stochastic}), a stochastic variant of \Cref{gluon:algo:deterministic} that operates with noisy gradient estimates available through a stochastic gradient oracle $\nabla f_{\xi}$, $\xi \sim \mathcal{D}$.

\begin{assumption}\label{gluon:ass:bounded_var}
    The stochastic gradient estimator $\nabla f_{\xi}: \cS \to \cS$ is unbiased and has bounded variance. That is, $\mathbb{E}_{\xi \sim \cD}[\nabla f_{\xi}(X)]=\nabla f(X)$ for all $X \in \cS$ and there exists $\sigma \geq 0$ such that $\mathbb{E}_{\xi \sim \cD} \bigl[\|\nabla_i f_{\xi}(X)-\nabla_i f(X)\|^2_2\bigr] \leq \sigma^2$ for all $X \in \cS$, $i = 1, \dots, p$.
\end{assumption}
Note that the choice of norm in \Cref{gluon:ass:bounded_var} is not restrictive: in finite-dimensional spaces, all norms are equivalent, so variance bounds remain valid up to a constant factor when compared to those based on any non-Euclidean norm.
The following result establishes the convergence properties.

\begin{theorem}\label{gluon:theorem:3}
    Let Assumptions \ref{gluon:ass:generalized-smoothness} and \ref{gluon:ass:bounded_var} hold and fix $\varepsilon>0$. Let $X^0,\allowbreak\ldots,\allowbreak X^{K-1}$ be the iterates of \algname{Gluon} (Algorithm \ref{gluon:algo:stochastic}) run with
    \[
        \beta^k = 1 - (k+1)^{-1/2}, \qquad
        t_i^k = t_i (k+1)^{-3/4}
    \]
    for some $ t_i>0$, and $M^0_i=\nabla_i f_{\xi^0}(X^0)$. Then
    \begin{equation}\label{gluon:eq:stoch_rate}
        \min\limits_{k=0,\ldots,K-1} \sum\limits_{i=1}^p \frac{1}{12L^1_i} \Exp{\| \nabla _i f(X^k) \|_{(i) \star}} \lesssim \frac{\Delta^0}{K^{1/4}} + \frac{1}{K^{1/4}} \sum\limits_{i=1}^p \left[ \frac{\sigma}{L^1_i} + \frac{L^0_i}{(L^1_i)^2}\right],
    \end{equation}
    where $\Delta^0:=f(X^0) - f^{\inf}$ and the notation $\lesssim$ omits numerical constants and logarithmic factors.
\end{theorem}

For $p=1$, our rate in \eqref{gluon:eq:stoch_rate} recovers the complexity for finding a stationary point of $(L^0, L^1)$-smooth functions established by \citet{hubler2024parameter} for normalized \algname{SGD} with momentum.
When $p\geq 1$, compared to existing guarantees for \algname{Gluon}, our \Cref{gluon:theorem:3} operates under the significantly more general \Cref{gluon:ass:generalized-smoothness} and uniquely supports training with larger, non-constant stepsizes $t_i^k \propto k^{-3/4}$. In contrast, prior analyses prescribe constant, vanishingly small stepsizes $t_i^k \equiv t_i \propto K^{-3/4}$, tied to the \emph{total} number of iterations $K$ (see \Cref{gluon:table1}).

\FloatBarrier
\section{Experiments}
\label{gluon:sec:experiments}

Below, we highlight selected experimental results for the \algname{unScion} optimizer, a special case of \algname{Gluon} (see \Cref{gluon:sec:lmo_examples}). The primary aim is to verify our layer-wise $(L^0, L^1)$-smoothness model (\Cref{gluon:ass:generalized-smoothness}). Additional details and further experiments are provided in \Cref{gluon:appendix:more_exp}.

\begin{figure*}
    \includegraphics[width=\textwidth]{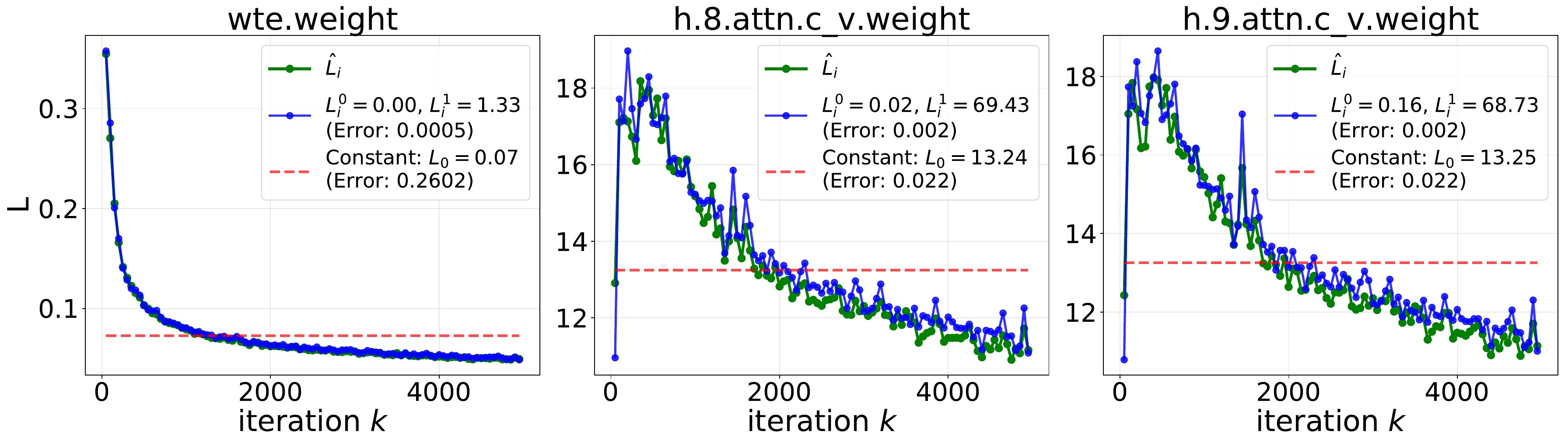}
    \caption{\textbf{Validation of \Cref{gluon:ass:generalized-smoothness}} in \texttt{NanoGPT-124M} along training trajectories of \algname{unScion}.
    The estimates $\hat{L}_i[k]$ are closely matched by $\hat{L}_i^{\text{approx}}[k]$, while a constant $L_0$ (corresponding to the classical smoothness assumption), results in a substantially poorer fit.
    }
    \label{gluon:fig:best_worst_fits_main}
\end{figure*}

\begin{figure*}[t]
    \centering
    \begin{minipage}{0.61\textwidth}
        \includegraphics[width=\textwidth]{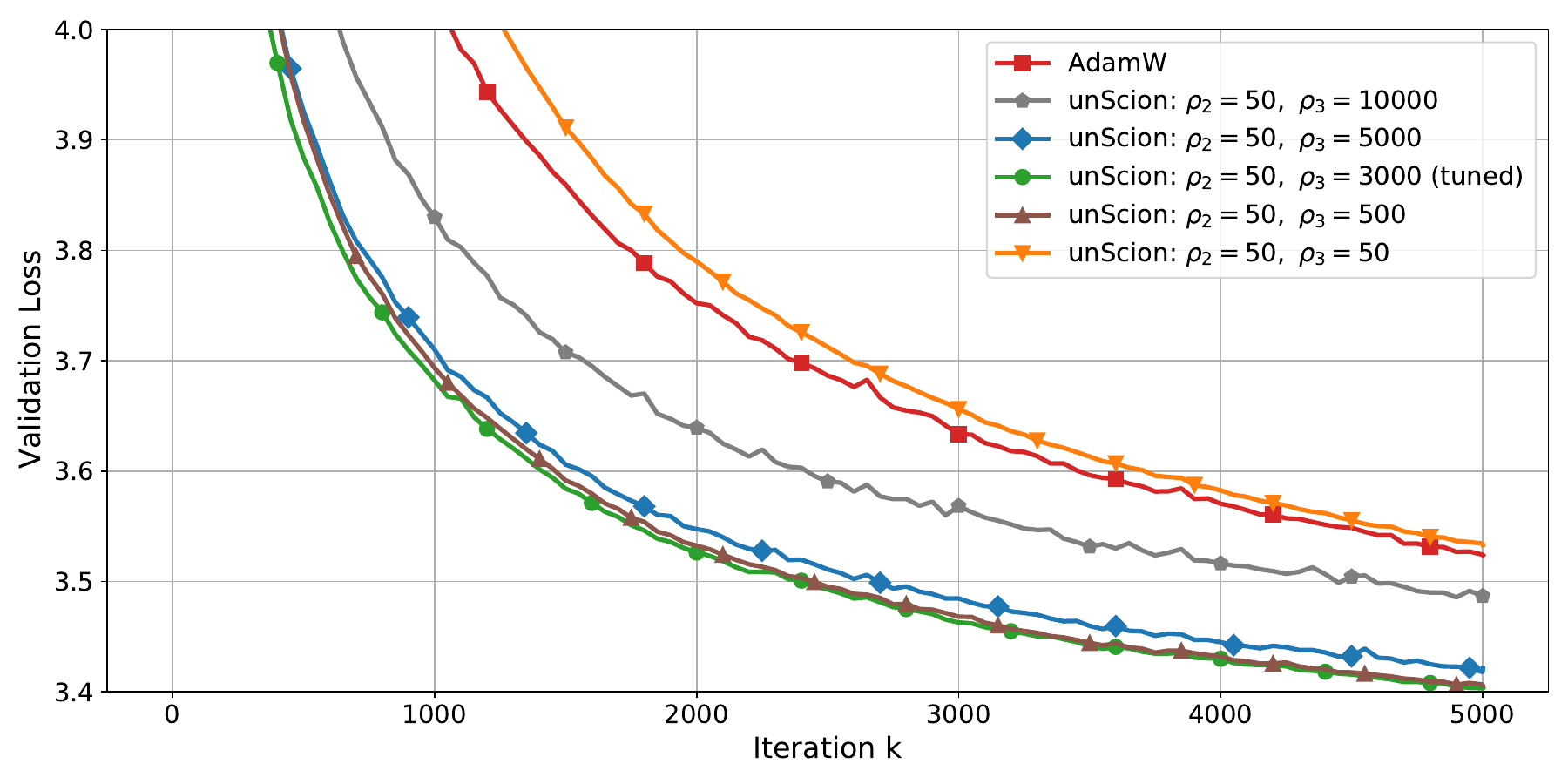}
    \end{minipage}
    \hfill
    \begin{minipage}{0.38\textwidth}
        \includegraphics[width=\textwidth]{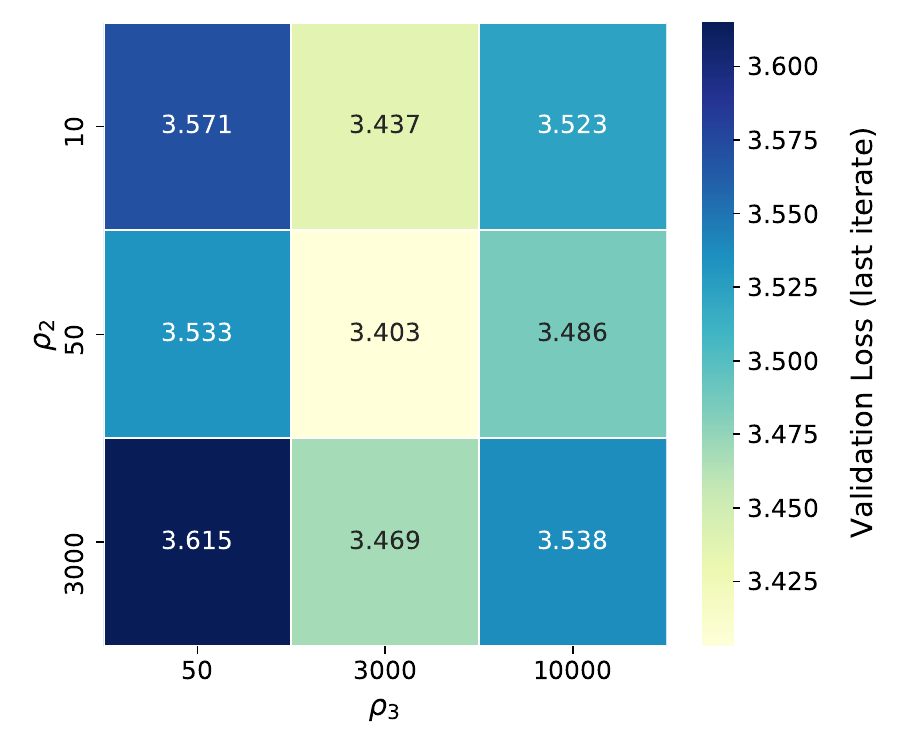}
    \end{minipage}
    \caption{(a) Validation curves for \algname{AdamW} and \algname{unScion} with varying $\rho_3$ values; (b) Heatmap of validation loss from the last iteration of \algname{unScion} across different combinations of $\rho_2$ and $\rho_3$.}
    \label{gluon:fig:effect_scaling_factors}
\end{figure*}

In the first set of experiments, we train the \texttt{NanoGPT} model with 124M parameters on the \texttt{FineWeb} dataset, leveraging two open-source GitHub repositories~\citep{jordan2024moddednanogpt,code_scion}. We use the \algname{unScion} optimizer, i.e., \algname{Gluon} with the norm choices as in \eqref{gluon:eq:scion_llm_main}.
We adopt the hyperparameters from~\citet[Table~7]{pethick2025training}, mapping their values $ \gamma = 0.00036 $, $ \rho_2 = 50 $, and $ \rho_3 = 3000 $ into our notation as follows: $ t_i^k \equiv \gamma \rho_2 = 0.018 $ for $ i = 1, \dots, p-1 $ (corresponding to the transformer block layers), and $ t_p^k \equiv \gamma \rho_3 = 1.08 $ (token embeddings and output projections, due to weight sharing).
We set the number of warmdown iterations to $0$ to keep the learning rates constant throughout training. The model is trained for 5{,}000 iterations in accordance with the Chinchilla scaling laws.%
In Figures \ref{gluon:fig:intro}, \ref{gluon:fig:1}, \ref{gluon:fig:2}, we plot the estimated \emph{trajectory smoothness} as a function of the iteration index $k$
\begin{align}\label{gluon:eq:dir_smooth}
    \hat{L}_i[k] \eqdef \frac{\|\nabla_i f_{\xi^{k+1}} (X^{k+1}) - \nabla_i f_{\xi^k} (X^{k}) \|_{(i)\star}}{\|X_i^{k+1} - X_i^{k}\|_{(i)}}
\end{align}
for parameter groups from the embedding layer and 4th and 8th transformer blocks (with similar trends observed across all blocks).
We compare this to the approximation
$$\hat{L}_i^{\text{approx}}[k] \eqdef L_i^0 + L_i^1 \|\nabla_i f_{\xi^{k+1}}(X^{k+1})\|_{(i) \star}, $$
where $L^0_i, L^1_i\geq0$ are fitted to minimize the Euclidean error between $\hat{L}_i[k]$ and $\hat{L}_i^{\text{approx}}[k]$, with hinge-like penalty on underestimation (see \Cref{gluon:sec:nanogpt_l0l1}).
The close alignment between these curves implies that \emph{Assumption~\ref{gluon:ass:generalized-smoothness} is approximately satisfied along the training trajectories}.

In \Cref{gluon:appendix:fit_quality}, we present a gallery of the best and worst per-layer smoothness fits, demonstrating that the layer-wise $(L_i^0,L_i^1)$ model achieves substantially smaller relative error than constant-$L_i$ fits (see \Cref{gluon:fig:best_worst_fits}). This aligns with prior work indicating that classical smoothness does not adequately capture the geometry of deep learning loss landscapes \citep{zhang2020why}. We further summarize the fit quality across all blocks and matrix types in \Cref{gluon:fig:fit_quality_by_matrix_type}.

\begin{figure*}[t]
    \centering
    \begin{minipage}{0.32\textwidth}
        \includegraphics[width=\textwidth]{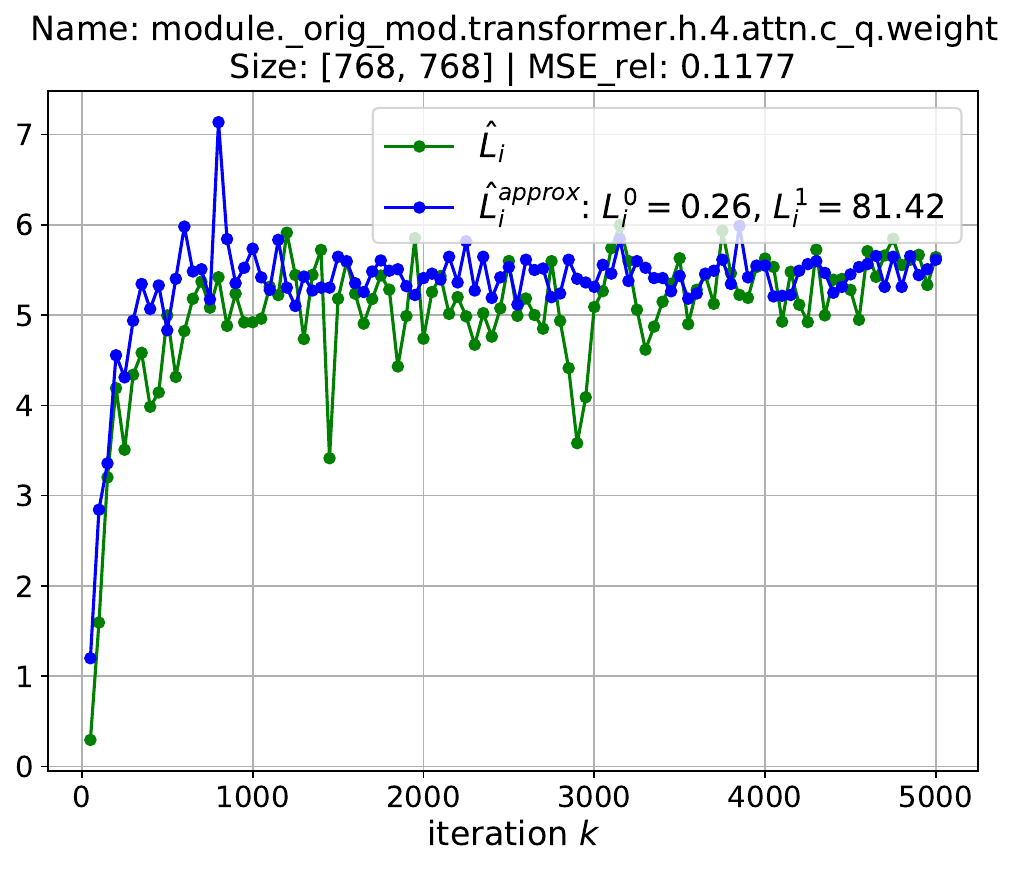}
    \end{minipage}
    \hfill
    \begin{minipage}{0.32\textwidth}
        \includegraphics[width=\textwidth]{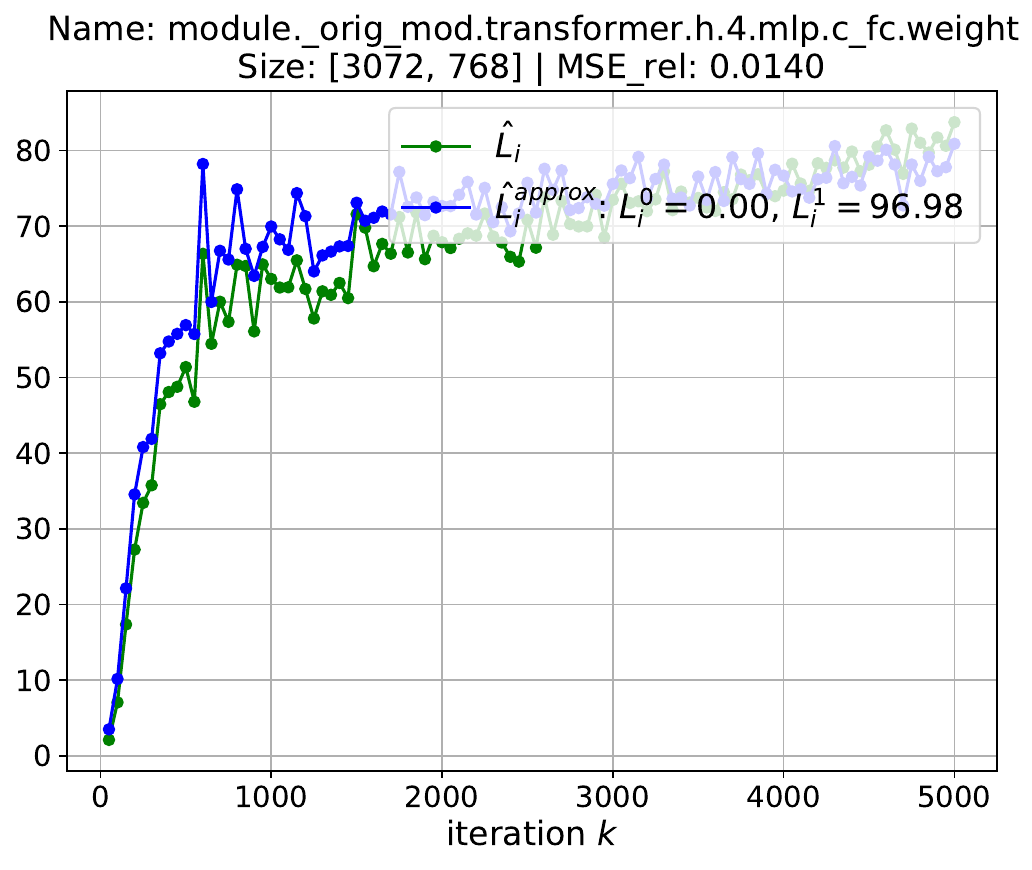}
    \end{minipage}
    \hfill
    \begin{minipage}{0.32\textwidth}
        \includegraphics[width=\textwidth]{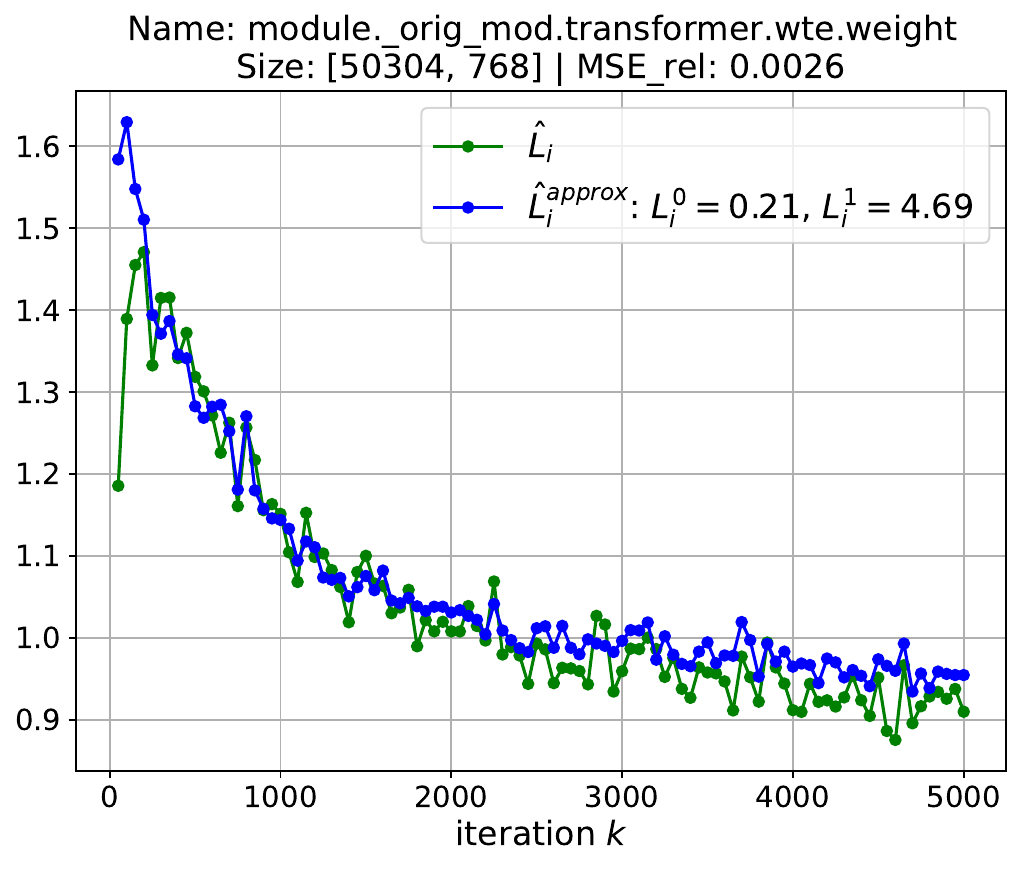}
    \end{minipage}

    \caption{\small Validation of layer-wise $(L^0, L^1)$-smoothness for different groups of parameters in \texttt{NanoGPT-124M} along \algname{AdamW} training trajectories.}
    \label{gluon:fig:Adam}
\end{figure*}

\textbf{Effect of scaling factors.}
We next evaluate the impact of the learning rate scaling factors $\rho_2$ and $\rho_3$ on the performance of the \algname{unScion} optimizer. For consistency, all other hyperparameters are fixed as described earlier. As a baseline, we include results obtained with the \algname{AdamW} optimizer, using the hyperparameter settings from Section~\ref{gluon:sec:Adam_exp}.
\Cref{gluon:fig:effect_scaling_factors} presents (a) validation curves for both optimizers, with varying $\rho_3$ in \algname{unScion}, and (b) the final validation loss for \algname{unScion} across different combinations of $\rho_2$ and $\rho_3$.
The best performance is achieved with $\rho_2 = 50$ and $\rho_3 = 3000$, i.e., $t_i^k = 0.018$ for $i = 1, \dots, p-1$ and $t^k_p = 1.08$. %
This further supports the use of non-uniform scaling across layers, with larger step sizes for the embedding layer, and provides additional evidence in favor of our layer-wise treatment.

\textbf{Learning rate transfer from AdamW.}
A careful reader may note that our estimates are computed from statistics collected along the training trajectory of \algname{Scion}, which raises the question of whether the observed phenomena reflect intrinsic properties of the loss landscape or are instead artifacts of a particular optimization trajectory induced by \algname{Scion}. To further substantiate our claims, we repeat the layer-wise $(L^0, L^1)$-smoothness verification using the same methodology as above, but employing the \algname{AdamW} optimizer.
In \Cref{gluon:fig:Adam}, we present the results for the estimated trajectory smoothness $\hat{L}_i$ and its approximation $\hat{L}_i^{\text{approx}}$ for several parameter groups along the training trajectory (additional plots are provided in \Cref{gluon:fig:Adam_app}, along with an aggregate over all layers in \Cref{gluon:fig:all_layers_scales-124M-Adam}). Notably, for the embedding layer parameters $X_p$ (last panel of Figure~\ref{gluon:fig:Adam}), the fitted value of $L^1_p$ is approximately 20–30 times smaller than in other groups. Across all plots, we observe that $L^0_i \ll L^1_i \|\nabla_i f_{\xi^k}(X^k)\|_{(i) \star}$, implying via \Cref{gluon:theorem:1} that $t^k_i \approx \nicefrac{1}{L^k_i}$. Thus, $ t^k_p $ should be 20–30 times larger than $ t^k_i $ for $ i = 1, \dots, p-1 $, consistent with the tuned parameters from \citet[Table 7]{pethick2025training}.

The fact that our layer-wise $(L^0, L^1)$-smoothness model accurately approximates trajectory smoothness during \algname{AdamW} training--and that a comparable separation between transformer and embedding layers emerges as in \algname{Scion} training--provides further support for our claims.
Moreover, it suggests that smoothness statistics from \algname{AdamW} training can guide per-layer learning rate tuning in \algname{Scion}.

Additional experimental details, conclusions, and practical recommendations are provided in \Cref{gluon:sec:Adam_exp}.
\FloatBarrier

\subsection{Additional studies}

In Appendix \ref{gluon:appendix:gensmooth_euclid}, we present an ablation study demonstrating that specialized norms provide a better approximation of trajectory smoothness compared to the standard Euclidean norm.
\Cref{gluon:appendix:layers_scales} examines whether the observations made for \texttt{NanoGPT-124M} extend to larger model scales. We report experiments with \texttt{GPT-2 Medium} ($\sim$355M parameters) and \texttt{GPT-2 Large} ($\sim$774M parameters), finding that the phenomena described above persist across all scales.
Finally, in \Cref{gluon:appendix:cifar} we validate our smoothness model by training a \texttt{CNN} model on the \texttt{CIFAR-10} dataset.

\FloatBarrier
\section{Conclusion and future work}\label{gluon:sec:conclusion}

In this work, we propose \algname{Gluon}, an LMO-based optimizer that recovers state-of-the-art optimizers such as \algname{Muon} and \algname{Scion} as special cases. We develop a principled analytical framework for layer-wise optimization based on a novel \emph{layer-wise $(L^0, L^1)$-smoothness} assumption, which captures the anisotropic structure of modern deep networks. This assumption enables sharper and more general convergence guarantees and, unlike prior analyses, leads to stepsizes with both theoretical justification and practical relevance. Our framework thus provides \emph{the first rigorous and practically useful analysis of modern layer-wise optimizers}. Experiments confirm that our assumption holds approximately throughout training. Together, these results offer a foundation for structured optimization in deep learning.

While this work resolves two key theoretical gaps (Sections~\ref{gluon:sec:layerwise} and~\ref{gluon:sec:orig_steps}), it also highlights important directions for future research.
Our analysis assumes exact LMO computations, whereas practical implementations use approximations (\Cref{gluon:sec:exp_details}). Additionally, our stochastic guarantees (\Cref{gluon:theorem:3}) rely on the widely adopted bounded variance assumption, which may not hold in certain scenarios, e.g., under subsampling \citep{khaled2020better}.
Finally, our support for adaptive stepsizes is restricted to the deterministic setting, and a complete theoretical justification in the stochastic regime remains an open challenge.

In summary, although we close the two most critical gaps--establishing a realistic generalized smoothness model and aligning analysis with actual implementations--no single work can exhaust the subject. The field remains open, with many fruitful directions left to pursue.

\endgroup

\chapter{Beyond the Ideal: Analyzing the Inexact Muon Update}
\label{chapter7:beyond_the_ideal}

\begingroup
\input{KAUST_Thesis/imported/07_beyond_the_ideal/import_setup}

\section*{Abstract}
The \algname{Muon} optimizer has rapidly emerged as a powerful, geometry-aware alternative to \algname{AdamW}, demonstrating strong performance in large-scale training of neural networks. However, a critical theory-practice disconnect exists: \algname{Muon}'s efficiency relies on fast, approximate orthogonalization, while most theoretical analyses study idealized exact-SVD updates. This work moves beyond the ideal by providing a general analysis of the \textit{inexact} orthogonalized update at \algname{Muon's} core. We develop our analysis within the general framework of Linear Minimization Oracle (LMO)-based optimization, introducing a realistic additive error model to capture the inexactness of practical approximation schemes. Our analysis yields explicit bounds that quantify performance degradation as a function of the LMO inexactness/error, $\err$. We reveal a fundamental coupling between this inexactness and the optimal step size and momentum: lower oracle precision requires a smaller step size but larger momentum parameter. These findings elevate the approximation procedure (e.g., the number of \approxalg{Newton-Schulz} steps) from an implementation detail to a critical parameter that must be \emph{co-tuned} with the learning schedule. NanoGPT experiments directly confirm the predicted coupling, with optimal learning rates clearly shifting as approximation precision changes.

\begin{quote}
\small
\noindent\textit{This chapter is based on:}\\
Egor Shulgin, Sultan AlRashed, Francesco Orabona, and Peter Richt{\'a}rik, ``Beyond the Ideal: Analyzing the Inexact Muon Update,'' \emph{International Conference on Artificial Intelligence and Statistics (AISTATS)}, 2026.
\nocite{shulgin2026beyond}
\end{quote}

\section{Introduction}

For over a decade, the landscape of deep learning optimization has been dominated by adaptive first-order methods, with \algname{AdamW}~\citep{kingma2015adam,loshchilov2019decoupled} serving as the de facto standard for training large and complex neural networks. Its robustness and general effectiveness have powered progress across numerous domains. Yet a new class of geometry-aware optimizers has recently emerged, challenging this paradigm. Among them, \algname{Muon}~\citep{jordan2024muon} has quickly gained prominence as a successor to \algname{AdamW}. By leveraging matrix structure in neural network parameters, \algname{Muon} has demonstrated superior performance and scalability, setting new training speed records for models like nanoGPT~\citep{jordan2024moddednanogpt} and enabling the efficient training of state-of-the-art Large Language Models (LLMs), such as Kimi from Moonshot AI~\citep{team2025kimi}. Benchmarking studies consistently show that \algname{Muon} can be significantly more computationally efficient than \algname{AdamW}~\citep{liu2025muon,shah2025practical,wen2025fantastic}.

However, a critical disconnect lies at the heart of \algname{Muon}'s success. The practical efficiency of the optimizer is entirely predicated on its use of fast, approximate orthogonalization methods, most notably the \approxalg{Newton-Schulz} iteration~\citep{jordan2024muon}. This iterative matrix-multiplication-based procedure provides a computationally cheap way to approximate the orthogonal factor of a matrix's polar decomposition, avoiding a full Singular Value Decomposition (SVD) which would be prohibitively expensive~\citep{grishina2025accelerating}. Yet, a significant gap exists between this practical implementation and its theoretical understanding. Existing theoretical analyses of \algname{Muon}~\citep[e.g.,][]{li2025noteconvergencemuon,shen2025convergence,chen2025muon} have largely studied an \emph{idealized}, computationally intractable version of the algorithm, assuming access to an exact orthogonalization oracle that computes the perfect SVD-based update. Hence, the existing theory still falls short of fully explaining the practical optimizer.

This work moves beyond the ideal to provide a general theoretical analysis of the \textit{inexact} \algname{Muon} update. We situate our analysis within the general framework of Linear Minimization Oracle (LMO)-based optimization~\citep{pethick2025training}, also known as the Frank-Wolfe framework~\citep{frank1956algorithm,hazan2008sparse,clarkson2010coresets,jaggi2013revisiting}. This perspective frames the core \algname{Muon} operation as an LMO call over the unit ball with respect to the spectral norm. To account for the realities of practical computation, we introduce a realistic additive error model. This model is directly motivated by the behavior of practical approximation schemes like \approxalg{Newton-Schulz}, allowing us to capture the inexactness inherent in any efficient implementation. By analyzing the algorithm under this inexact LMO, we bridge the critical gap between theory and practice.

\subsection{Contributions}

Our main contributions to the theory and practice of LMO-based optimizers are:

$\bullet$ \textbf{An analysis of the implemented inexact \algname{Muon} update.}
We analyze the \emph{implemented} \algname{Muon} update through an additive direction-space inexactness model (Assumption~\ref{beyond:assump:inexact_lmo_main}) tailored to practical polar-factor approximation schemes such as \approxalg{Newton-Schulz} and \approxalg{PolarExpress}.

$\bullet$ \textbf{Uncovering a fundamental hyperparameter coupling.}
Our analysis reveals that the LMO inexactness, $\err$, is a critical parameter that alters the optimization dynamics. We derive explicit formulas for the step size ($\gamma^* \propto (1+\err)^{-1/4}$) and momentum ($\alpha^* \propto \sqrt{1+\err}$) in the stochastic setting (Corollary~\ref{beyond:cor:stochastic_optimal_rate}), uncovering a crucial coupling: a less precise LMO (larger $\err$) requires a {\em smaller} learning rate but a {\em larger} momentum parameter. This elevates the approximation quality from an implementation detail to a core hyperparameter.

$\bullet$ \textbf{A general convergence framework.}
We establish convergence rates for both deterministic and stochastic settings. In Appendix~\ref{appendix:chapter7_beyond_the_ideal}, we further extend the \emph{deterministic} analysis to the $(L^0, L^1)$-smoothness model~\citep{zhang2020why} and to a layer-wise setting~\citep{riabinin2025gluon}. Setting the inexactness to zero recovers the corresponding exact-LMO rates in the LMO framework.

$\bullet$ \textbf{Empirical validation.}
We validate our key theoretical predictions with experiments on CIFAR-10 and NanoGPT. We confirm that performance degrades as LMO precision decreases and, crucially, our experiments with NanoGPT (Figure~\ref{beyond:fig:nanoGPT_heatmaps}) provide clear empirical evidence for the predicted hyperparameter coupling, showing that the optimal learning rate shifts to a lower value when a less precise LMO is used.

\subsection{Related work}

We survey the key developments toward a theoretical understanding of \algname{Muon}.

The conceptual basis for \algname{Muon}'s orthogonalized update was introduced by \citet{jordan2024muon} and further motivated by \citet{bernstein2024old}, who connected orthogonalized matrix updates to steepest-descent geometry under the spectral norm. The first formal convergence analyses for \algname{Muon} sought to connect it to existing theoretical frameworks. In particular, three different groups independently provided convergence guarantees for \algname{Muon}, with varying degrees of generality: \citet{li2025noteconvergencemuon}, \citet{pethick2025training}, and \citet{kovalev2025muon}.

\citet{li2025noteconvergencemuon} provided an initial convergence guarantee by viewing the \algname{Muon} update as a matrix-based generalization of normalized \algname{SGD} with momentum, building upon the analysis of \citet{cutkosky2020momentum}.

Instead, \citet{pethick2025training} and \citet{kovalev2025muon} proposed two powerful unifying frameworks, providing a more general perspective. \citet{pethick2025training} introduced the \algname{Scion} framework, which situates \algname{Muon} within the broader class of methods based on the Linear Minimization Oracle (LMO), a core component of the Frank-Wolfe algorithm~\citep{frank1956algorithm}. \citet{kovalev2025muon} developed a non-Euclidean trust-region interpretation, also for arbitrary norms. Both frameworks successfully recover the idealized \algname{Muon} update as a special case when the spectral norm is chosen and provide the first general convergence guarantees for this class of optimizers.

Beyond the Muon literature, inexact LMOs / inexact Frank--Wolfe or generalized conditional gradient methods have been studied extensively in convex and constrained optimization, under several oracle models including multiplicative and additive inaccuracies; see, e.g., \citet{lacoste2013block,locatello2017unified,locatello2017greedy,yurtsever2018conditional,silveti2021inexact}. Our setting differs in three important ways: we study unconstrained, non-convex, momentum-based \algname{Muon}-style updates; we focus on the matrix-orthogonalization regime relevant to \algname{Muon}; and we use an additive model motivated by practical polar-factor solvers such as \approxalg{Newton-Schulz} and \approxalg{PolarExpress}, without assuming that the approximate direction remains feasible.

An idealized trust-region method, called the ball-proximal (``broximal'') point method (\algname{BPM}), was concurrently developed by \citet{BPM}. \algname{BPM} has remarkable theoretical guarantees for convex optimization---it converges in finitely many steps and has a linear rate independent of the condition number, with the geometric factor improving in each iteration---without requiring differentiability, finite-valuedness, or strong convexity. Moreover, the method provably converges to the global minimizer for a certain class of non-convex problems. A non-Euclidean variant was recently proposed and analyzed by \citet{neBPM}, with the non-Euclidean norm playing the role of a hyperparameter that performs a form of geometric preconditioning.
\algname{Muon} can be seen as an approximate version of non-Euclidean \algname{BPM} for a specific norm choice, where the broximal operator is applied to a stochastic linear approximation of the loss instead of the original loss, with momentum added to the mix as a means of handling stochastic noise.

Subsequent work has focused on making these general frameworks more reflective of practical deep learning scenarios. The \algname{Gluon} framework of \citet{riabinin2025gluon} extended the analysis to a more realistic layer-wise setting and introduced a generalized non-Euclidean smoothness model to better capture the heterogeneous structure of neural networks. Concurrently, \citet{pethick2025generalized} introduced a clipped \algname{Scion} variant that provides guarantees under a similar generalized smoothness condition. Further, \citet{Drop-Muon} have shown that it may be theoretically suboptimal to update {\em all} the layers of a neural network in each iteration, and as a remedy, proposed and analyzed (also under generalized smoothness) the \algname{Drop-Muon} method, showing wall-clock improvements on toy networks.

An extension of \algname{Muon} to the distributed setting, with support for communication compression, error-feedback and generalized smoothness, was developed by \citet{EF21-Muon}, who proposed the \algname{EF21-Muon} method. The \algname{MuonBP} method of \citet{MuonBP} proposes to apply orthogonalization independently to matrix shards on each device, while periodically performing full orthogonalization to maintain training stability at scale.

Despite all this theoretical understanding, a substantial part of the literature still analyzes an idealized algorithm that assumes access to an exact, error-free LMO. This is a significant gap, as the practical efficiency and success of \algname{Muon} are entirely predicated on the use of fast but approximate solvers for the orthogonalized update. The concept of spectral descent and orthogonalized updates has historical roots in deep learning~\citep{carlson2015stochasticaistats,carlson2015stochasticieee,carlson2015preconditioned,tuddenham2022orthogonalising}. Our work addresses the remaining broader disconnect between theory and practice through a solver-agnostic inexact-LMO analysis.\footnote{Concurrent with our work, \citet{kim2026muonns} study the original Muon algorithm with finitely many Newton--Schulz steps. We defer a more detailed comparison to Appendix~\ref{beyond:app:additional_related_work}.}

\section{From idealized theory to practical implementation}
\label{beyond:sec:background}

We begin by formalizing the class of algorithms our analysis covers, starting from the \algname{Muon} optimizer.

\paragraph{The idealized update}

\algname{Muon}~\citep{jordan2024muon} is matrix-aware, leveraging the geometric structure of two-dimensional weight parameters ($W \in \mathbb{R}^{n \times m}$). Instead of element-wise scaling as in \algname{AdamW}, \algname{Muon} applies momentum and then performs an orthogonalization step on the resulting update matrix. The idealized update has the form
\begin{gather*}
M^k = (1-\alpha) M^{k-1} + \alpha G^k, \quad D^k = T(M^k),\\
 X^{k+1} = X^k - \gamma^k D^k,
\end{gather*}
where $G^k$ is a stochastic gradient, $\gamma^k>0$ is the step size, $M^k$ is the momentum matrix, $\alpha>0$ is the momentum parameter, and $T(\cdot)$ represents the projection onto the set of orthogonal matrices.

As shown by \citet{bernstein2024old}, this update is equivalent to performing steepest descent with respect to the spectral norm geometry, $\|\cdot\| \eqdef \|\cdot\|_\text{sp}$. Given a gradient matrix $G$, the steepest descent direction under this norm is equal to
\begin{equation} \label{beyond:eq:steepest_descent}
    {\operatorname{argmin}} \ \left\{\inp{G}{D} \;:\; \|D\|_\text{sp} \le 1 \right\}.
\end{equation}
The solution to \eqref{beyond:eq:steepest_descent} is the orthogonal polar factor of the negative gradient, $D = \operatorname{polar}(-G)$. If the SVD of the gradient is $G = U S V^\top$, then the solution is $D = -U V^\top$. This idealized update direction forms the theoretical basis of the \algname{Muon} optimizer. In its practical implementation, the computationally expensive polar decomposition is approximated by the efficient, SVD-free \approxalg{Newton-Schulz} iteration. This practical algorithm has demonstrated remarkable empirical success but was introduced without a formal convergence analysis.

Subsequently, \citet{pethick2025training} (using the {\em Linear Minimization Oracle} (LMO) framework) and \citet{kovalev2025muon} (via a non-Euclidean trust-region framework) provided the first meaningful convergence guarantees for this type of update. These frameworks consider a general update of the form\footnote{From now on, we drop the upper-case matrix notation in favor of simpler, lower-case vector-space notation.} $x^{k+1} = x^k + \gamma^k d^k$, where the direction $d^k$ is the solution to the LMO
\begin{equation}
    d^k \eqdef {\operatorname{argmin}} \ \left\{\inp{m^k}{d} \;:\; \|d\| \le 1 \right\},
    \label{beyond:eq:lmo_definition}
\end{equation}
where $m^k$ is the momentum term, $\inp{\cdot}{\cdot}$ refers to an inner product (trace inner product in matrix spaces), and $\|\cdot\|$ refers to an arbitrary, possibly non-Euclidean norm. These analyses recover the idealized \algname{Muon} update when $\|\cdot\|$ is chosen as the spectral norm.

\paragraph{The practical imperative of approximation}
While these frameworks provide invaluable insight, they analyze an idealized algorithm that is never run in practice.
For many norms, such as the $\ell_\infty$ norm in \algname{SignSGD}, the LMO in \eqref{beyond:eq:lmo_definition} can be computed exactly and with minimal overhead. For the spectral norm, however, the exact LMO solution requires a full, prohibitively expensive, SVD.

This computational bottleneck makes the practical implementations of \algname{Muon} entirely dependent on \textit{approximate updates}. The original optimizer employs 5 steps of the \approxalg{Newton-Schulz} iteration \citep{higham2008functions}, a matrix polynomial-based method that efficiently approximates the polar factor using only fast matrix-matrix multiplications \citep{jordan2024muon}. The importance of this approximation is underscored by an active line of research into developing superior iterative schemes \citep{cesista2025muonoptcoeffs}, such as \approxalg{PolarExpress} \citep{amsel2025polar} and \approxalg{CANS} \citep{grishina2025accelerating}, which offer better error guarantees or faster convergence. This highlights a crucial fact: the algorithm achieving state-of-the-art results is not the idealized one, but one of its many possible inexact instantiations. Much of the existing theory still focuses on the idealized case, assuming access to an error-free LMO.

\paragraph{Modeling inexactness}

To bridge this theory-practice divide, we must first establish a realistic model for the error produced by the LMO approximation. We replace the exact direction $d^k$ in the update rule with an inexact direction $\hat{d}^k$, which is assumed to satisfy the following condition.
\begin{assumption}[\textbf{Inexact LMO}]
    \label{beyond:assump:inexact_lmo_main}
    Let $d^k$ be the exact solution to \eqref{beyond:eq:lmo_definition}. The inexact solution $\hat{d}^k$ is assumed to satisfy an additive error bound for some $\err_k \ge 0$:
    \begin{equation}
        \|\hat{d}^k - d^k\| \le \err_k.
        \label{beyond:eq:inexact_assumption}
    \end{equation}
\end{assumption}

This assumption is not arbitrary; it is directly motivated by the convergence guarantees of the iterative methods used in practice. Algorithms like \approxalg{Newton-Schulz} and \approxalg{PolarExpress} produce an approximation whose error decreases with the number of iterations performed. Specifically, \citet{amsel2025polar} prove that \approxalg{PolarExpress} satisfies an error bound of the form $\|\hat{d}^k - d^k\| \le C|1-l|^{p}$, where $l<1$ and $p$ is determined by the number of iterations (see Appendix~\ref{beyond:app:error_model} for details). Since these methods are always run for a small, fixed number of steps in practice (e.g., five steps in the standard \algname{Muon} implementation \citep{jordan2024muon}), their final error is bounded by a constant. Our parameter $\err_k$ models this error, allowing our analysis to capture the behavior of the implemented algorithm. A key feature of our analysis is that we do not assume that the output $\hat{d}^k$ is feasible, i.e., we do not assume $\|\hat{d}^k\| \le 1$, which is a realistic property of these approximation schemes.

\section{Main theoretical results}
\label{beyond:sec:main_results}

Having established the practical importance of the inexact LMO, we now develop a theoretical framework for analyzing its impact on convergence. Our analysis is general and holds for any norm $\|\cdot\|$. We consider the unconstrained optimization problem
\begin{equation}
    \min \ \left\{f(x) \;:\; x \in \mathcal{X} \right\},
    \label{beyond:eq:problem_formulation}
\end{equation}
where $(\mathcal{X}, \|\cdot\|)$ is a normed space. We will use the following standard assumption on the objective function.

\begin{assumption}
\label{beyond:assump:smoothness}
The objective function $f: \mathcal{X} \to \R$ is continuously differentiable and its gradient $\nabla f$ is Lipschitz continuous with constant $L \ge 0$ with respect to the norm $\|\cdot\|$. This means for any $x, y \in \mathcal{X}$ we have
\begin{equation*}
    \|\nabla f(x) - \nabla f(y)\|_{\star} \le L \|x-y\|,
\end{equation*}
where $\|\cdot\|_{\star}$ is the dual norm of $\|\cdot\|$. Furthermore, we assume that $f$ is bounded below by $f^* > -\infty$.
\end{assumption}

We begin with the deterministic setting to build intuition, before moving to the full stochastic analysis with momentum.

\subsection{Deterministic case}
\label{beyond:sec:deterministic}

In the deterministic setting, we analyze the update
\begin{equation}
    x^{k+1} = x^k + \gamma^k \hat{d}^k,
    \label{beyond:eq:deterministic_method}
\end{equation}
where $\gamma^k > 0$ is the step size and $\hat{d}^k$ is the output of an inexact LMO for the gradient $g^k \eqdef \nabla f(x^k)$, i.e.,
\begin{equation*}
    \hat{d}^k \approx \operatorname{argmin} \ \left\{ \inp{g^k}{d} \;:\; \|d\| \le 1\right\}.
\end{equation*}
The inexact direction $\hat{d}^k$ is assumed to satisfy Assumption~\ref{beyond:assump:inexact_lmo_main} with some inexactness level $\err_k \ge 0$. We now state our first convergence result for this method.

\begin{theorem}[General Result]
\label{beyond:thm:general_deterministic}
Let Assumption~\ref{beyond:assump:smoothness} hold. Let the sequence $\{x^k\}_{k=0}^{K-1}$ be generated by the update rule \eqref{beyond:eq:deterministic_method} with step sizes $\gamma^k > 0$. Assume the inexact LMO satisfies Assumption~\ref{beyond:assump:inexact_lmo_main} with $\err_k < 1$ for all $k$ and let $\Delta^0 \eqdef f(x^0) - f^*$. Then, after $K$ iterations,
\begin{align*}
  \min \limits_{0 \le k < K} \ \norm{\nabla f(x^k)}_{\star}
    \le \frac{\Delta^0 + \frac{L}{2} \sum_{k=0}^{K-1} (\gamma^k)^2 (1+\err_k)^2}{\sum_{k=0}^{K-1} \gamma^k (1-\err_k)}.
\end{align*}
\end{theorem}

Theorem~\ref{beyond:thm:general_deterministic} provides a general convergence guarantee that explicitly characterizes how the interplay of step sizes $\{\gamma^k\}$ and inexactness levels $\{\err_k\}$ affects convergence. The bound reveals that inexactness degrades performance in two ways: the numerator is amplified by a $(1+\err_k)^2$ factor due to the potential infeasibility of the update, while the denominator, representing total progress, is diminished by a $(1-\err_k)$ factor. Condition $\err_k < 1$ is mathematically necessary to ensure that the denominator is positive, which guarantees that the algorithm makes progress on average. As shown in the proof (see Appendix~\ref{beyond:sec:proofs}), this condition is required for the per-iteration descent property and is satisfied by practical approximation schemes like \approxalg{PolarExpress}~\citep{amsel2025polar}. To gain clearer, quantitative insights into these trade-offs, we next analyze the important special case of constant parameters.

\begin{corollary}[Constant Parameters]
\label{beyond:cor:constant_deterministic}
Under the conditions of Theorem~\ref{beyond:thm:general_deterministic}, if the step size is constant, $\gamma^k = \gamma > 0$, and the LMO error is constant, $\err_k = \err < 1$, for all $k$, then after $K$ iterations we have
\begin{align*}
\frac{1}{K} \sum \limits_{k=0}^{K-1} \norm{\nabla f(x^k)}_{\star} \le \frac{\Delta^0}{K\gamma(1-\err)} + \frac{L \gamma (1+\err)^2}{2(1-\err)}.
\end{align*}
\end{corollary}

Corollary~\ref{beyond:cor:constant_deterministic} simplifies the general bound, making the trade-offs more apparent. The bound consists of two terms that exhibit the classical trade-off in the choice of step size $\gamma$: a larger $\gamma$ reduces the first term but increases the second. This structure implies that an optimal convergence rate is achieved when the step size is of the order $\mathcal{O}(1/\sqrt{K})$, which balances these two terms.

The inexactness level $\err$ degrades both terms in the bound. They are each amplified by a factor of $1/(1-\err)$, which arises from the reduced quality of the descent direction. The second term is further penalized by a $(1+\err)^2$ factor, which quantifies the cost of potential infeasibility of the update step.

Our analysis is general, holding for any norm in a framework akin to that of \citet{pethick2025training} and \citet{kovalev2025muon}. For the specific choice of the spectral norm, $\|\cdot\| = \|\cdot\|_{\star}$, this result provides a convergence guarantee for the practical, inexact \algname{Muon} update. Notably, if we set the inexactness to zero ($\err=0$), our bound recovers the standard $\mathcal{O}(1/\sqrt{K})$ rate for the exact LMO method. Our framework also accommodates more complex step-size rules; in Appendix~\ref{beyond:app:adaptive_step_size}, we analyze an adaptive step-size policy and show that it achieves the same optimal rate. To make the trade-off in the constant step size explicit, we now derive the optimal choice of $\gamma$ that minimizes this bound.

By minimizing the right-hand side of the bound in Corollary~\ref{beyond:cor:constant_deterministic} with respect to $\gamma$, we obtain the optimal constant step size and the corresponding best rate.

\begin{corollary}[Optimal Parameters]
\label{beyond:cor:optimal_deterministic}
Under the conditions of Corollary~\ref{beyond:cor:constant_deterministic}, by choosing the optimal constant step size
$\gamma^* = \frac{1}{1+\err} \sqrt{\frac{2\Delta^0}{K L}}$,
the average gradient norm is bounded as
\begin{align*}
    \frac{1}{K} \sum \limits_{k=0}^{K-1} \norm{\nabla f(x^k)}_{\star} \le \frac{1+\err}{1-\err} \sqrt{\frac{2\Delta^0 L}{K}}.
\end{align*}
\end{corollary}

Corollary~\ref{beyond:cor:optimal_deterministic} makes the theoretical trade-offs concrete, revealing a direct coupling between the LMO precision and the optimal strategy. \emph{The optimal step size, $\gamma^* \propto 1/(1+\err)$, must decrease as the oracle becomes less precise}. This has a direct practical implication for optimizers like \algname{Muon}: using a less accurate approximation of the orthogonalized update (e.g., by reducing the number of \approxalg{Newton-Schulz} iterations) might require a corresponding decrease in the step size to maintain optimal performance. This adaptation, however, does not fully mitigate the error, as the best achievable rate is degraded by a factor of $\frac{1+\err}{1-\err}$. This factor quantifies the dual cost of the potential update infeasibility (from the $1+\err$ term) and reduced descent quality (from the $1-\err$ term). The explicit dependence of the rate on this degradation factor also provides a theoretical justification for the empirical benefits of more advanced approximation schemes. In fact, methods such as \approxalg{PolarExpress}~\citep{amsel2025polar} and \approxalg{CANS}~\citep{grishina2025accelerating} are designed to achieve a smaller approximation error $\err$ more efficiently, which our analysis shows directly translates to an improved convergence rate for the overall optimization.

Our analysis is a tight generalization of prior work; setting $\err=0$ recovers the exact rate for the idealized LMO method~\citep{kovalev2025muon,pethick2025training}.
Importantly, while the constant is degraded by the inexactness, the $\mathcal{O}(1/\sqrt{K})$ rate implies an iteration complexity of $\mathcal{O}(1/\varepsilon^2)$ to find an $\varepsilon$-stationary point. This matches the optimal complexity for first-order methods on smooth non-convex problems~\citep{carmon2020lower}. Our result thus establishes that the inexact LMO method remains optimal, while characterizing the degradation as a function of the oracle's error.

\subsection{Stochastic case}
\label{beyond:sec:stochastic}

We now extend our analysis to the more practical setting where the optimizer has access only to a stochastic oracle for the gradient and incorporates momentum.

For this setting, we require two additional standard assumptions. First, we assume access to an unbiased stochastic first-order oracle with bounded variance.
\begin{samepage}
\begin{assumption}
    \label{beyond:assump:stochastic_oracle}
    The oracle returns a stochastic gradient $g^k = \nabla f(x^k)+\xi^k$ for a random variable $\xi^k$, such that it is an unbiased estimator of the true gradient and has a uniformly bounded variance,
    \[
    \begin{gathered}
        \Exp{g^k \;|\; x^k} = \nabla f(x^k), \\
        \Exp{\|g^k - \nabla f(x^k)\|_2^2 \;|\; x^k} \le \sigma^2,
        \qquad \sigma^2 \ge 0.
    \end{gathered}
    \]
\end{assumption}
\end{samepage}
Second, since our analysis is for a general norm $\|\cdot\|$ but the variance is typically assumed to be bounded in the Euclidean norm $\|\cdot\|_2$, we assume a norm compatibility condition that is always true in finite-dimensional spaces. So, we denote by $\rho > 0$ the constant such that for all $v \in \mathcal{X}$, we have $\|v\|_{\star} \le \rho \|v\|_2$.

The method we analyze is a general momentum-based algorithm with an inexact LMO, see Algorithm~\ref{beyond:alg:stochastic_inaxact_lmo}.

\begin{algorithm}[t]
\caption{Inexact Generalized \algname{Muon}}
\label{beyond:alg:stochastic_inaxact_lmo}
\begin{algorithmic}[1]
\STATE \textbf{Input:} Initial point $x^0$, momentum $m^0$
\FOR{$k = 0, 1, \dots, K-1$}
\STATE Compute stochastic gradient $g^k$
\STATE Update momentum: $m^{k+1} = (1-\alpha_k)m^k + \alpha_k g^k$
\STATE Compute inexact LMO with $m^{k+1}$:
    \begin{equation*}
    \hat{d}^k \approx \underset{\|d\| \le 1}{\operatorname{argmin}} \ \inp{m^{k+1}}{d}
\end{equation*}
\STATE Update parameters: $x^{k+1} = x^k + \gamma^k \hat{d}^k$
\ENDFOR
\end{algorithmic}
\end{algorithm}

To analyze this algorithm, our proof proceeds by establishing a per-iteration descent guarantee and then bounding the error of the momentum term. While the full proofs are deferred to Appendix~\ref{beyond:app:stochastic_proofs}, we present the key lemma for the momentum error here, as it reveals the direct impact of the inexact LMO.

\begin{lemma}
\label{beyond:lemma:inexact_momentum_error_main}
Let Assumptions~\ref{beyond:assump:inexact_lmo_main}, \ref{beyond:assump:smoothness}, \ref{beyond:assump:stochastic_oracle} hold. For Algorithm~\ref{beyond:alg:stochastic_inaxact_lmo}, the expected momentum error is bounded by
\begin{equation*}
\mathbb{E} \|m^{k+1} - \nabla f(x^k)\|_{\star}
    \le (1-\alpha_k)\mathbb{E} [\|m^k - \nabla f(x^{k-1})\|_{\star}] + \frac{\rho\sigma\sqrt{\alpha_k}}{\sqrt{2-\alpha_k}} + L\gamma^k(1+\err_k).
\end{equation*}
\end{lemma}

\textbf{Commentary on the Proof.}
This lemma bounds the expected deviation of the momentum from the true gradient. The critical difference in our analysis, compared to prior work on exact LMOs, arises when bounding the ``gradient drift'' term, which depends on the step length $\|\nabla f(x^{k-1}) - \nabla f(x^k)\|_{\star} \le L\|x^{k-1} - x^k\|$. An exact LMO guarantees a feasible direction $\|d^{k-1}\| \le 1$, yielding a step length of exactly $\gamma^{k-1}$. In our analysis, the potential infeasibility of the inexact direction, $\|\hat{d}^{k-1}\| \le 1 + \err_{k-1}$, leads to a looser step length bound of $\gamma^{k-1}(1+\err_{k-1})$. This modification introduces the crucial $(1+\err_k)$ factor into the final term of the lemma, quantifying the cost of potential infeasibility inherent to any practical LMO approximation. The full proof is provided in Appendix~\ref{beyond:app:stochastic_proofs}.

Building on this lemma, we now present the main convergence guarantee for Algorithm~\ref{beyond:alg:stochastic_inaxact_lmo}.

\begin{theorem}
\label{beyond:thm:stochastic}
Let Assumptions \ref{beyond:assump:inexact_lmo_main},\ref{beyond:assump:smoothness},\ref{beyond:assump:stochastic_oracle} hold, along with the norm compatibility condition. For Algorithm~\ref{beyond:alg:stochastic_inaxact_lmo} with parameters $\gamma > 0$, $\alpha \in (0,1)$, and $\err < 1$, after $K$ iterations, the average expected gradient norm is upper bounded by
\begin{equation*}
\begin{adjustbox}{max width=\textwidth}
$\displaystyle
\frac{1}{K}\sum_{k=1}^{K} \mathbb{E}\|\nabla f(x^k)\|_{\star}
    \le \frac{1}{1-\err}\Bigg[\frac{\Delta^0}{K\gamma}
    + 2\rho\sigma\left(\frac{1}{\alpha K} + \sqrt{\alpha}\right)
    + L\gamma(1+\delta)\left(\frac{7+3\delta}{2} + \frac{2}{\alpha}\right)
\Bigg],
$
\end{adjustbox}
\end{equation*}
where $\Delta^0 \eqdef f(x^0) - f^*$.
\end{theorem}

Theorem~\ref{beyond:thm:stochastic} provides a convergence guarantee for the practical, inexact LMO-based method with momentum. Compared to the deterministic analysis in Corollary~\ref{beyond:cor:constant_deterministic}, the bound contains additional terms dependent on the noise variance $\sigma^2$. With a constant step size, these terms establish convergence that the algorithm will minimize the expected gradient up to a floor governed by the choice of parameters.

Crucially, every term in the bound is amplified by factors involving the inexactness level $\err$, which confirms the intuition that a more precise LMO (smaller $\err$) leads to faster convergence. As a sanity check, setting the inexactness to zero ($\err=0$) recovers the corresponding convergence guarantee for the idealized, exact LMO method \citep{kovalev2025muon,pethick2025training}, demonstrating that our analysis is a strict generalization.

For clarity, we present this result for the practical case of constant parameters, the case of time-varying parameters is deferred to Appendix~\ref{beyond:app:time_varying}. To better understand the complex interplay of the constant parameters, we now derive the optimal choices of $\gamma$ and $\alpha$ that minimize this bound (details in Appendix~\ref{beyond:app:stochastic_proofs}).

\begin{corollary}
\label{beyond:cor:stochastic_optimal_rate}
Let the conditions of Theorem~\ref{beyond:thm:stochastic} hold. By choosing $\gamma$ and $\alpha$ of orders
\begin{align*}
    \gamma &\asymp \left(\frac{\Delta^0}{K}\right)^{3/4} \frac{1}{(\sigma^2 L (1+\err))^{1/4}},\\
    \alpha &\asymp \sqrt{\frac{\Delta^0 L (1+\err)}{K \sigma^2}},
\end{align*}
the average expected gradient norm is bounded as
\begin{align*}
\frac{1}{K}\sum \limits_{k=1}^{K} \mathbb{E} \norm{\nabla f(x^k)}_{\star} = \mathcal{O}\left( \frac{(L\Delta^0)^{1/4}\sigma^{1/2}(1+\err)^{1/4}}{K^{1/4}(1-\err)} \right).
\end{align*}
\end{corollary}

Corollary~\ref{beyond:cor:stochastic_optimal_rate} provides the main insights of our stochastic analysis, revealing a fundamental coupling between the LMO precision and the hyperparameter strategy. The best achievable rate is degraded by a factor of $\mathcal{O}\left(\frac{(1+\err)^{1/4}}{1-\err}\right)$. This explicit dependence on $\err$ provides a theoretical justification for the empirical benefits of more advanced approximation schemes like \approxalg{PolarExpress}~\citep{amsel2025polar} and \approxalg{CANS}~\citep{grishina2025accelerating}, as any reduction in approximation error directly improves the convergence rate. Interestingly, the degradation from inexactness is less severe than in the deterministic setting (Corollary~\ref{beyond:cor:optimal_deterministic}), where the rate is penalized by $\mathcal{O}\left(\frac{1+\err}{1-\err}\right)$. This suggests that the presence of stochastic noise, which already necessitates a degree of caution, makes the optimization dynamics less sensitive to the additional instability from the inexact LMO.

This rate is achieved when the hyperparameters are adapted to the LMO's precision. The step size, $\gamma^* \propto 1/(1+\err)^{1/4}$, must \textbf{decrease} as the oracle becomes less precise. In contrast, the momentum parameter, $\alpha^* \propto \sqrt{1+\err}$, must \textbf{increase}. This prescribes a clear strategy: a less accurate LMO (larger $\err$) requires more caution (smaller step size) but also greater agility (shorter momentum memory) to adapt to less reliable update directions.

Finally, the $\mathcal{O}(1/K^{1/4})$ rate implies an iteration complexity of $\mathcal{O}(1/\varepsilon^4)$ to find an $\varepsilon$-stationary point. This matches the established lower bounds for first-order stochastic methods on non-convex problems under our assumptions~\citep{ghadimi2013stochastic}, and this complexity is known to be unimprovable in general~\citep{arjevani2023lower}. Our analysis thus establishes that the inexact LMO method remains efficient in a formal sense, while precisely characterizing the degradation as a function of the oracle's error.

\subsection{Extensions and generalizations}

\paragraph{Beyond standard smoothness.}
Our analysis can be extended beyond the standard $L$-smoothness assumption, which can be restrictive for neural networks. In Appendix~\ref{beyond:app:l0l1_smooth}, we provide a full analysis for the more general non-Euclidean $(L^0, L^1)$-smoothness model
\citep{zhang2020why,yu2025mirror,riabinin2025gluon}. For this class of functions, we establish an iteration complexity of $$
\mathcal{O}\left(\frac{\Delta^0(1+\err)^2}{(1-\err)^2}\left(\frac{L^0}{\varepsilon^2} + \frac{L^1}{\varepsilon}\right)\right).$$
This result recovers the standard $\mathcal{O}(1/\varepsilon^2)$ rate when $L^1=0$. More importantly, it allows for an improved rate over gradient descent in regimes where $L^0$ is small, a phenomenon that has been empirically observed for the training trajectories of nanoGPT on FineWeb~\citep{riabinin2025gluon}. Our analysis shows that the dependence on the inexactness level $\err$ remains consistent, and by setting $\err=0$, our results tightly recover the prior guarantees for the exact LMO method on $(L^0, L^1)$-smooth functions~\citep{vankov2025optimizing}.

\begin{figure*}[t]
    \centering
\subfigure[Validation loss behavior; more iterations (smaller $\err$) drive faster convergence.\label{beyond:fig:nanoGPT_convergence}]{%
\begin{minipage}[t]{0.365\linewidth}
    \centering
    \includegraphics[width=\linewidth]{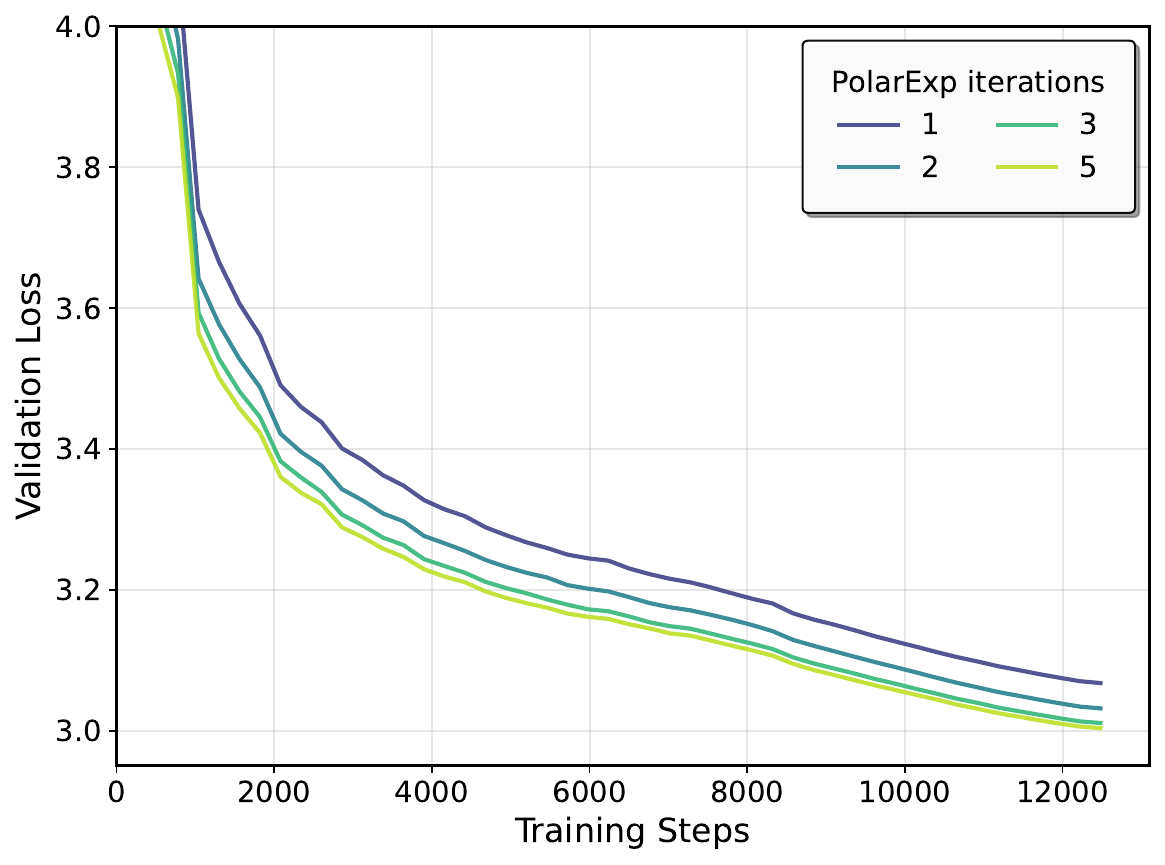}
\end{minipage}}%
\hfill
\subfigure[Hyperparameter sweeps for 2 (left) and 5 (right) \approxalg{PolarExpress} iterations; higher precision widens the stable region.\label{beyond:fig:nanoGPT_heatmaps}]{%
\begin{minipage}[t]{0.625\linewidth}
    \centering
    \includegraphics[width=\linewidth]{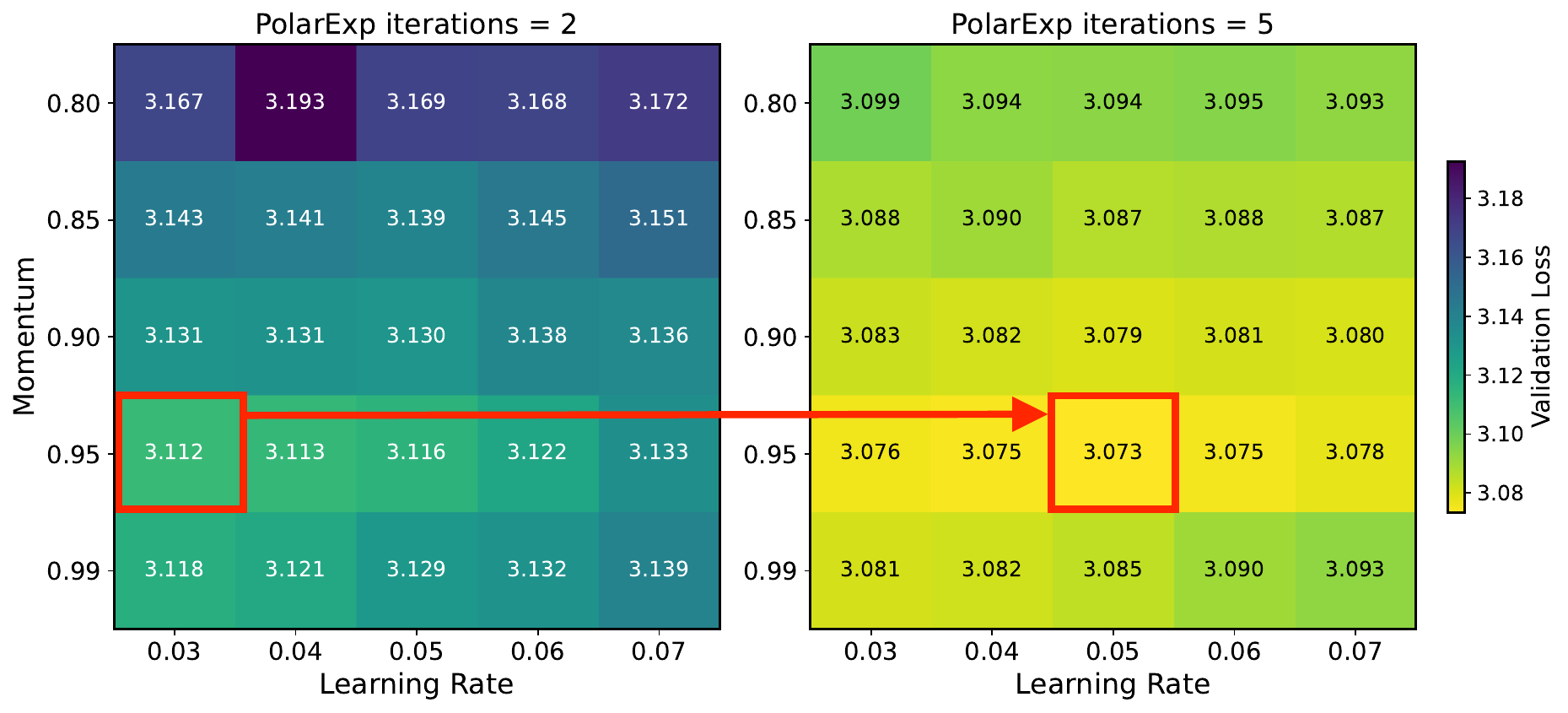}
\end{minipage}}%
\caption{NanoGPT on FineWeb: convergence (left) and hyperparameter sensitivity (right) as \approxalg{PolarExpress} (PolarExp) iterations vary.}
\label{beyond:fig:nanoGPT_combined}
\end{figure*}

\paragraph{Layer-wise analysis.}
In practice, optimizers like \algname{Muon} and \algname{Scion} are applied in a layer-wise manner, often with different norms and update rules for different parameter blocks (e.g., spectral norm for weight matrices, $\ell_\infty$ norm for biases)~\citep{pethick2025training}. This heterogeneity, along with empirical observations of varying smoothness across layers~\citep{riabinin2025gluon}, motivates a more fine-grained analysis. In Appendix~\ref{beyond:app:layerwise}, we extend our framework by considering a block-wise structure of the parameter space $x = (x_1, \dots, x_p)$, where each block $x_i$ is associated with its own norm $\|\cdot\|_{(i)}$, smoothness constant $L_i$, and inexactness $\err_i$.

Our analysis in this setting reveals that the impact of LMO inexactness is not uniform across the network. This suggests that a uniform precision level for all layers may be suboptimal and opens the door to a principled strategy for computational savings, where one could allocate fewer resources (e.g., fewer \approxalg{Newton-Schulz} iterations) to approximate the LMO for layers that are more robust to inexactness.
This type of heterogeneity is visible empirically in Figure~\ref{beyond:fig:pe_layerwise_6000}, where attention and MLP matrices exhibit markedly different \approxalg{PolarExpress} difficulty profiles even at the same training checkpoint.

\section{Experiments} \label{beyond:sec:experiments}

We now present empirical results to validate the key predictions of our theoretical analysis. Our experiments are designed to answer two main questions:
\begin{enumerate}
\item How does optimizer performance degrade as the LMO becomes less precise?
\item Do the hyperparameters for the best performance (step size and momentum) shift in response to the LMO inexactness, as predicted by our analysis?
\end{enumerate}

We use the number of iterations in the approximation algorithm (either \approxalg{Newton-Schulz} or \approxalg{PolarExpress}) as a practical proxy for the inexactness level $\err$, where fewer iterations correspond to a higher $\err$. We conduct experiments on two standard benchmarks: training a NanoGPT model on the FineWeb dataset \citep{penedo2024fineweb} and a CNN on CIFAR-10 \citep{krizhevsky2009learning}. Full details are provided in Appendix~\ref{beyond:app:exp_details}.

\subsection{NanoGPT on FineWeb}
\label{beyond:sec:exp_nanoGPT}

We train a 124M parameter nanoGPT model with \algname{Muon} with a batch size of 512,000 using the codebase by \citet{jordan2024moddednanogpt}. For the LMO approximation, we use the \approxalg{PolarExpress} algorithm~\citep{amsel2025polar}.

\textbf{Validating performance degradation.} Our theory predicts that a less precise LMO (higher $\err$) should lead to degraded convergence performance. Figure~\ref{beyond:fig:nanoGPT_convergence} confirms this trend. We trained the model for 12,000 steps and observe that increasing the precision of the LMO by using more \approxalg{PolarExpress} iterations consistently leads to a lower final validation loss. While the most significant gains are seen when moving from one to three iterations, further increases in precision continue to improve performance, though with diminishing returns (see Appendix~\ref{beyond:app:additional_nanogpt} for results with up to 8 iterations). This observation directly supports the degradation factor in our convergence bounds.

\textbf{Validating hyperparameter coupling.}
A key prediction of our stochastic analysis (Corollary~\ref{beyond:cor:stochastic_optimal_rate}) is the coupling between the LMO inexactness $\err$, the step size $\gamma^*$, and the momentum $\alpha^*$. Figure~\ref{beyond:fig:nanoGPT_heatmaps} provides empirical evidence for this theoretical insight. The figure shows the final validation loss after 6,000 training steps across a grid of step sizes and momentum values for two different levels of LMO precision: a highly inexact oracle (2 \approxalg{PolarExp} iterations, left) and a more precise one (5 \approxalg{PolarExp} iterations, right).

For the \textbf{highly inexact} case (left panel), the region of best performance (lowest loss) is concentrated at a \textbf{low step size} (around 0.03). When the LMO is made \textbf{more precise} (right panel), the optimal region not only shifts to a \textbf{higher step size} (around 0.05) but also becomes broader, indicating greater stability across different hyperparameter choices. This empirical result aligns with our theory: Corollary~\ref{beyond:cor:stochastic_optimal_rate} predicts that as inexactness $\err$ increases (fewer iterations), the optimal step size should decrease.

The learning-rate trend is much more sharply resolved than the momentum trend at the resolution of our grid. Near the optimum, the validation loss is substantially flatter in momentum than in learning rate, so discrete sweeps can easily mask the weaker momentum dependence predicted by Corollary~\ref{beyond:cor:stochastic_optimal_rate}. For this reason, we interpret the NanoGPT heatmaps primarily as evidence for the learning-rate component of the coupling.

\subsection{CNN on CIFAR-10}
\label{beyond:sec:exp_cifar10}

To test our findings in a different domain, we train a simple CNN on the CIFAR-10 dataset with a batch size of 100. Our implementation is based on the unconstrained \algname{Scion} method~\citep{code_scion}, which we run with a constant step size to isolate the effect of inexactness. For these experiments, we use the standard \approxalg{Newton-Schulz} iteration \citep{jordan2024muon} to approximate the LMO.

\begin{figure*}[t]
    \centering
    \subfigure[Training loss drops faster as \approxalg{Newton-Schulz} precision increases.\label{beyond:fig:cifar_train_loss}]{%
    \begin{minipage}[t]{0.49\linewidth}
        \centering
        \includegraphics[width=\linewidth]{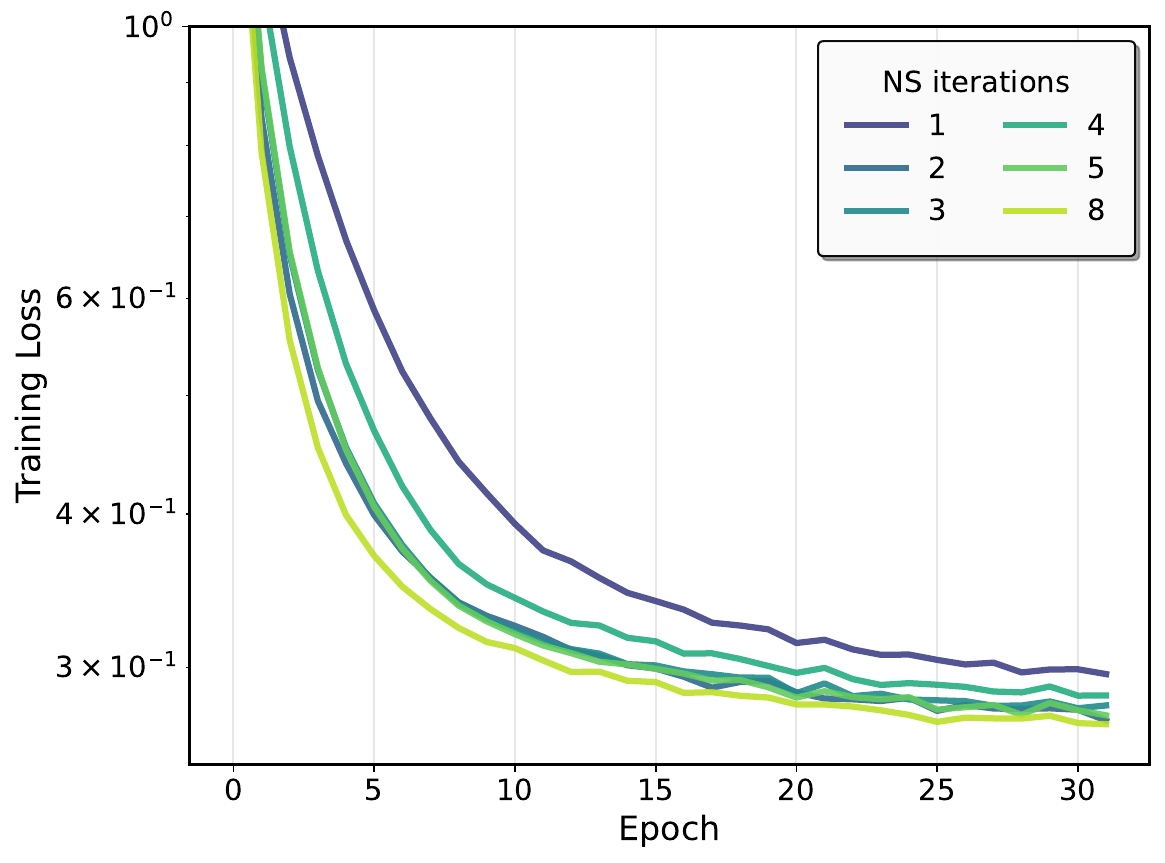}
    \end{minipage}}%
    \hfill
    \subfigure[Best (solid) vs. worst (dashed) accuracy across step sizes; higher precision narrows the gap.\label{beyond:fig:cifar_variability}]{%
    \begin{minipage}[t]{0.48\linewidth}
    \centering
        \includegraphics[width=\linewidth]{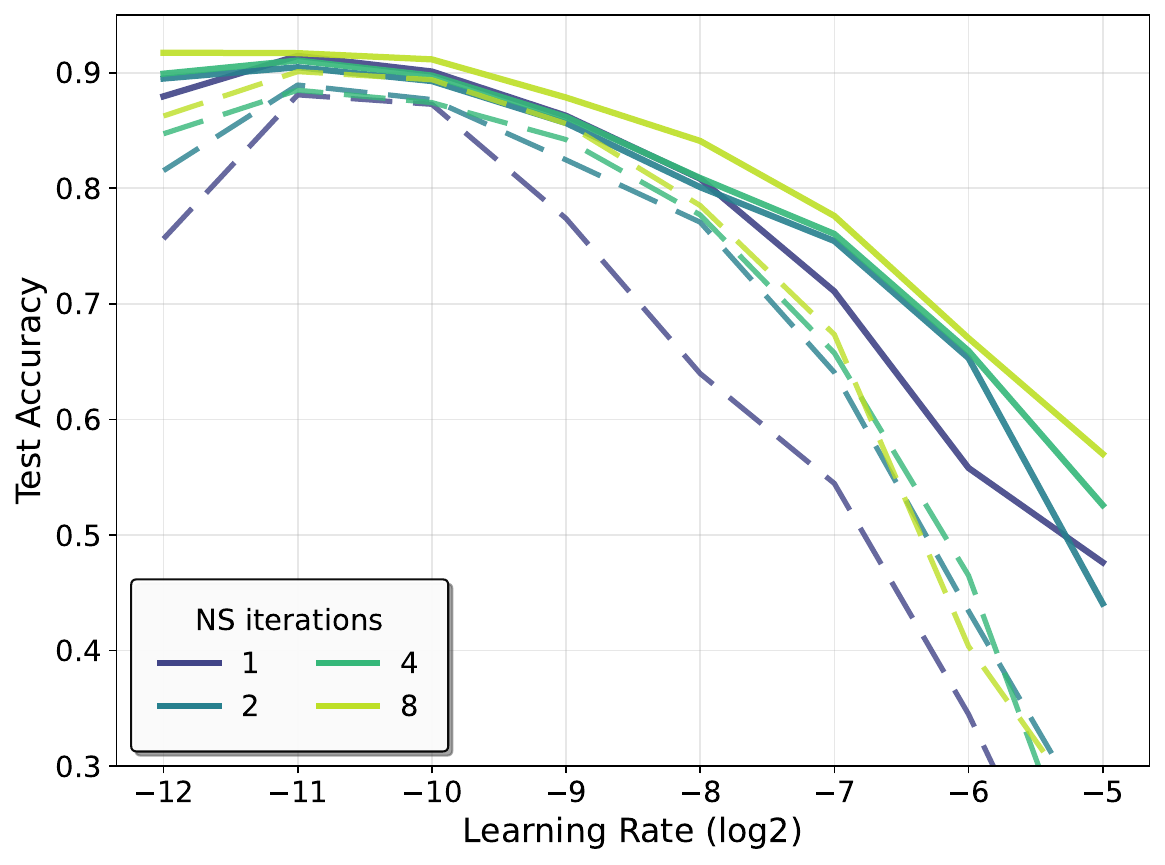}
    \end{minipage}}%
    \caption[CIFAR-10 loss and accuracy across Newton-Schulz precision levels.]{CIFAR-10 training loss and test accuracy across different \approxalg{Newton-Schulz} (NS) precision levels.}
    \label{beyond:fig:cifar_combined}
\end{figure*}

\textbf{Convergence dynamics.}
Our theory predicts that a less precise LMO should lead to degraded convergence. Figure~\ref{beyond:fig:cifar_train_loss} confirms this behavior. We plot the training loss over 32 epochs for different numbers of \approxalg{Newton-Schulz} iterations, using a fixed step size and tuned momentum for each run. The results clearly show that a more precise approximation (e.g., 8 iterations) converges faster and to a lower final training loss compared to less precise approximations (e.g., 1 or 4 iterations). This aligns with the degradation factor present in our theoretical bounds.

\textbf{Performance and stability versus inexactness.}
Beyond the training loss, we investigate how LMO inexactness affects the final test accuracy and the optimizer's stability with respect to its hyperparameters. Figure~\ref{beyond:fig:cifar_variability} shows the test accuracy achieved after a sweep over a range of step sizes for different levels of LMO precision. The solid lines represent the best accuracy achieved for a given step size (by tuning momentum), while the dashed lines show the worst.

Two key observations can be made. First, the peak of the solid lines (the best possible performance) consistently increases with the number of \approxalg{Newton-Schulz} iterations, again confirming that higher precision leads to better final performance. Second, the gap between the best (solid) and worst (dashed) performance is large for a highly inexact LMO (1 iteration) and becomes progressively smaller as precision increases. This suggests that a more precise LMO makes the optimizer more robust to the choice of other hyperparameters like momentum, a valuable practical property not explicitly captured by the convergence rate alone.

\section{Conclusion}
\label{beyond:sec:conclusion}

This work provides a formal convergence analysis of LMO-based optimizers with an inexact oracle in the unconstrained, non-convex setting, addressing the critical gap between the theory and practice of prominent algorithms like \algname{Muon}. By introducing a realistic additive error model, our analysis moves beyond the idealized assumption of a perfect oracle to study the algorithm as it is truly implemented.
Our analysis, for both deterministic and stochastic settings, reveals that there is a coupling between the LMO inexactness, $\err$, and the hyperparameter strategy. Our theory prescribes that a less precise LMO requires a more cautious (smaller) step size, a prediction that our experiments confirm.

\endgroup

\chapter{Conclusion}
\label{chapter8:conclusion}

Many of the projects in this thesis began with a practical method that had shown promising empirical behavior but whose optimization properties were not fully understood. The analysis often used simplified settings, but the simplification was selective: the aim was to preserve the feature most likely to affect the optimization behavior and abstract away the rest. Depending on the problem, this feature was the order of sketching and differentiation, the bounded transformation applied to an update, the layer-wise structure of a neural network, or an approximate inner computation. Keeping these features in the model led to different outcomes. In some cases, the theory exposed a limitation; in others, it suggested a different objective, a modified algorithm, or a more suitable set of assumptions.

The first pair of chapters began with distributed submodel training and related dropout-like methods. Chapter~\ref{chapter2:ist} isolates the defining operation of IST: the model is sketched before the gradient is computed. Even in a simplified quadratic setting, this is enough to reveal a structural limitation. The expected dynamics can converge to a fixed point that does not coincide with the minimizer of the original finite-sum objective, and the resulting error does not necessarily disappear when the stepsize is reduced. Chapter~\ref{chapter3:mast} gives a different interpretation of the same basic mechanism. Rather than viewing sketching only as an approximation to full-model optimization, MAST places the sketch distribution inside the objective itself. The sketched-gradient estimator is then unbiased for the resulting MAST objective. Optimization theory serves two roles here: it identifies when the classical problem formulation is inadequate, and it provides another formulation in which the practical mechanism becomes mathematically natural.

The second pair of chapters started from clipping mechanisms motivated by differential privacy. Chapter~\ref{chapter4:adaptive_clipping} develops a convergence analysis of adaptive quantile clipping, a method introduced because of its practical performance and its reduced reliance on a manually selected clipping threshold. The analysis reveals an obstruction similar to the one known for standard clipping: with a fixed quantile, a non-vanishing clipping bias can remain. It also shows that this bias can be controlled through a time-varying quantile and stepsize schedule, recovering optimal nonconvex rates under heavy-tailed stochastic-gradient noise. The analysis is then extended to the differentially private setting. Chapter~\ref{chapter5:smoothed_normalization} takes a more algorithmic route. Instead of changing the problem formulation, it changes the bounded operator: hard clipping is replaced by smoothed normalization and combined with error compensation. The resulting $\alpha$-NormEC method admits distributed private convergence guarantees under standard smoothness assumptions without requiring uniformly bounded gradients. Both chapters show that clipping and normalization must be included in the optimization model from the outset, rather than treated as privacy mechanisms added after the optimizer has been analyzed.

The final pair of chapters approaches the theory-practice gap from a different direction. Rather than analyzing a modern optimizer under standard assumptions, Chapter~\ref{chapter6:gluon} first asks whether those assumptions describe the structure of neural-network training. Muon-style methods act on entire weight matrices and operate layer by layer, while the measured smoothness can vary substantially across layers and along the optimization trajectory. Gluon therefore begins with a layer-wise generalized smoothness model and develops an LMO optimizer framework on top of it. The trajectory measurements support this model, and the resulting convergence guarantees explain how layer-wise smoothness quantities can guide update radii or learning-rate scales. Chapter~\ref{chapter7:beyond_the_ideal} then studies another practical component of Muon: the polar direction is computed approximately using a small number of fast inner iterations. Modeling this inexactness directly shows that it is not merely a numerical implementation detail. It affects convergence, the optimal stepsize, momentum, and the range of hyperparameters that work reliably in practice.

The common thread across these results is not that every implementation detail should be included in the theory. A useful model must remain simple enough to analyze. The important choice is what to simplify and what to preserve. In this thesis, retaining the feature that changes the optimization behavior led to a limitation for IST and fixed-quantile clipping, a new objective for MAST, an algorithmic correction in $\alpha$-NormEC, a layer-wise smoothness model for Gluon, and an explicit treatment of inexactness for Muon. Optimization is useful here not only for proving convergence of a convenient proxy, but also for identifying the objective, assumptions, update, and computation that should be analyzed in the first place.

\section{Future research directions}

The first two technical chapters suggest two related directions for submodel and
modular training. One is to correct the bias that can make the limiting point of
IST differ from the minimizer of the original objective. A stateful correction,
in the spirit of control variates or error compensation, may be able to retain
information lost when the model is sketched before the gradient is computed. It
is equally important to characterize the problem classes for which the original
IST update is already well behaved and no correction is needed. The renewed
interest in modular and path-based training, including methods such as DiPaCo,
makes both questions relevant beyond the particular IST algorithm studied in
Chapter~\ref{chapter2:ist} \citep{douillard2024dipaco}.

The complementary direction is to extend the MAST viewpoint beyond unbiased
linear sketches and, more generally, to other algorithmic families. Low-precision
training applies transformations that are often biased or nonlinear, while
parameter-efficient methods restrict training to a structured subset or a
low-dimensional parametrization of the model. In both cases, it is worth asking
whether the training mechanism implicitly defines a useful optimization problem,
which properties this problem inherits from the original objective, and whether
algorithms can be designed for that formulation directly. More broadly, the MAST
viewpoint could provide a way to study objectives induced by practical training
mechanisms beyond linear sketching.

Adaptive clipping leaves a different and particularly clear gap between the
analysis and the implemented method. Chapter~\ref{chapter4:adaptive_clipping}
assumes access to the exact population quantile of the stochastic-gradient norm,
whereas practical algorithms estimate this quantile from finite observations and
must do so privately \citep{andrew2021differentially}. The estimator has its own
estimation error and privacy cost. A more complete analysis should therefore couple
the quantile estimate with the optimization trajectory instead of treating the
threshold as an exact oracle. More broadly, private learning still needs adaptive
or nearly tuning-free optimization methods, together with private procedures for
hyperparameter and model selection
\citep{papernot2021hyperparameter,koskela2023practical}. These components should
be analyzed as part of the training procedure rather than as an external tuning
stage.

Subsequent work on Fed-$\alpha$-NormEC already moves the method of
Chapter~\ref{chapter5:smoothed_normalization} toward a more practical federated
setting \citep{shulgin2025practicalprivatefl}. The remaining difficulty is to make
stateful correction reliable in realistic cross-device training. Client-side
control variates are a standard way to correct client drift
\citep{karimireddy2020scaffold}, but persistent client state is difficult to
maintain in cross-device settings and may become stale under infrequent
participation \citep{reddi2021adaptive}. The underlying control-variate idea is
close in spirit to the error compensation used by $\alpha$-NormEC and may still be
useful if the correction is scaled according to how current and reliable it is.

One possible approach is to scale the correction instead of always applying it
in full. MARS scales a recursive momentum-correction term
\citep{yuan2025mars}, and subsequent analysis identifies settings in which this
scaling can provide a provable improvement over the corresponding unscaled MVR
correction \citep{shulgin2026mars}. It is open whether a related mechanism could
make control-variate or error-compensation ideas more robust to partial
participation, stale state, and privacy noise. Such a method would need to
determine the correction strength from information that is actually available in
a cross-device system.

The Gluon results suggest another route from measurements to algorithm design.
Chapter~\ref{chapter6:gluon} provides estimates of trajectory smoothness along
observed training runs, rather than global smoothness constants. The next question is
how to use this information during training. The estimates could be used to adapt
learning rates or LMO radii separately across layers, respond to changes along the
trajectory, and stabilize layers whose behavior differs from the rest of the model.
Recent work on noise-adaptive layer-wise learning rates shows that online
layer-specific signals can improve geometry-aware optimization
\citep{hao2025noiseadaptive}. Adaptive Gradient Descent and Prodigy
provide related examples of reducing learning-rate tuning using information
observed during optimization
\citep{MM2019,mishchenko2024prodigy}. Their standard formulations do not directly
exploit the modular structure of neural networks; an important question is how
these ideas should be adapted when the model and its smoothness are strongly
layer-dependent.

Even with layer-wise adaptation, the choice of norm and update remains open.
Muon, sign-based methods, and Euclidean methods use different geometries, and it
is still unclear which one is preferable for a given layer or stage of training.
The modular-duality viewpoint argues that norms should reflect the role of each
layer in the architecture \citep{bernstein2024modular}. Recent analyses compare
spectral and Euclidean updates in simplified matrix problems and study the
preconditioning effect of orthogonalization
\citep{shulgin2026quadratic_muon,ma2026preconditioning}. These results
provide useful examples, but not yet a general rule for attention, embedding, and
MLP layers in realistic networks. This suggests that the parametrization of a
layer, the norm used to measure its update, and the optimizer applied to it may
need to be designed together.

The inexact-Muon analysis raises a related question about computational effort.
The approximation difficulty observed in
Chapter~\ref{chapter7:beyond_the_ideal} varies substantially across layers and
matrix families, so one fixed number of Newton--Schulz or PolarExpress steps is
unlikely to be appropriate everywhere. A larger-scale study of momentum spectra
reaches a similar conclusion and gives layer- and model-size-dependent
prescriptions for Newton--Schulz precision
\citep{magakyan2026spectral}. A practical optimizer could instead monitor
spectral information or approximation quality and allocate only as many inner
iterations as each layer currently needs. The inner precision would then be
adapted jointly with the outer stepsize and momentum.

Finally, the exact polar factor is a natural reference point for analyzing
inexact Muon, but it is not necessarily the only useful matrix update. Finite
Newton--Schulz iterations define their own spectral transformation, and one can
also design smooth alternatives that retain more information about gradient
magnitudes. SoftSignum and SoftMuon provide recent examples of replacing hard
sign or polar-style maps by smooth, magnitude-sensitive transformations
\citep{feoktistov2026softsign}. Future theory should therefore compare exact polar
updates, finite polynomial approximations, and alternative matrix maps as
optimization methods in their own right. The relevant questions are not only how
closely a map approximates the polar factor, but also how much descent it produces,
how it handles conditioning and parameter heterogeneity, how stable it is, and how
much it costs to compute.

The longer-term goal is to make optimization theory predictive enough to guide
these choices: which objective to optimize, when a correction is needed, which
geometry to use for each part of the model, and how accurately an inner update
should be computed. This would make theory useful not only for analyzing an
algorithm after it has been proposed, but also for deciding what algorithm to run.

\cleardoublepage
\begin{onehalfspacing}
\renewcommand*\bibname{\centerline{REFERENCES}} 
\phantomsection
\addcontentsline{toc}{chapter}{References}
\newcommand{\BIBdecl}{\setlength{\itemsep}{0pt}}%
\bibliographystyle{thesisplainnat}
\bibliography{KAUST_Thesis/References}
\end{onehalfspacing}

\appendix

\chapter{Common Basic Facts and Auxiliary Inequalities}
\label{appendix:common_facts}

This appendix collects reusable mathematical facts used across multiple chapters.
The notation is intentionally neutral; chapter-specific constants, algorithms, and
technical lemmas remain in the corresponding chapter appendices.

\section{Inner-product and norm identities}
\label{common:sec:inner_norm}

\begin{lemma}[Polarization identity]
\label{common:lem:polarization}
For any $a,b \in \mathbb{R}^d$,
\[
    2\langle a,b\rangle
    = \|a\|^2 + \|b\|^2 - \|a-b\|^2 .
\]
\end{lemma}

\begin{lemma}[Cauchy--Schwarz inequality]
\label{common:lem:cauchy_schwarz}
For any $a,b \in \mathbb{R}^d$,
\[
    |\langle a,b\rangle| \leq \|a\|\,\|b\|.
\]
\end{lemma}

\begin{lemma}[Triangle inequality]
\label{common:lem:triangle}
For any $a,b \in \mathbb{R}^d$,
\[
    \|a+b\| \leq \|a\|+\|b\|.
\]
\end{lemma}

\begin{lemma}[Young inequalities]
\label{common:lem:young}
For any $a,b \in \mathbb{R}^d$ and any $\eta>0$,
\[
    2\langle a,b\rangle
    \leq \eta \|a\|^2 + \eta^{-1}\|b\|^2 .
\]
Consequently,
\[
    \|a+b\|^2
    \leq (1+\eta)\|a\|^2 + (1+\eta^{-1})\|b\|^2 .
\]
\end{lemma}

\begin{lemma}[Fenchel--Young inequality]
\label{common:lem:fenchel_young}
Let $\phi:\mathbb{R}^d\to(-\infty,+\infty]$ be a proper convex function and
let $\phi^\star(y):=\sup_{x\in\mathbb{R}^d}\{\langle x,y\rangle-\phi(x)\}$.
Then, for all $x,y\in\mathbb{R}^d$,
\[
    \langle x,y\rangle \leq \phi(x)+\phi^\star(y).
\]
\end{lemma}

\section{Smoothness and descent inequalities}
\label{common:sec:smoothness_descent}

\begin{lemma}[Smoothness descent inequality]
\label{common:lem:smooth_descent}
Let $f:\mathbb{R}^d\to\mathbb{R}$ be differentiable with $L$-Lipschitz
gradient. Then, for all $x,y\in\mathbb{R}^d$,
\[
    f(y) \leq f(x) + \langle \nabla f(x), y-x\rangle
    + \frac{L}{2}\|y-x\|^2 .
\]
\end{lemma}

\begin{lemma}[One-step descent for an inexact direction]
\label{common:lem:update_descent}
Let $f:\mathbb{R}^d\to\mathbb{R}$ be differentiable with $L$-Lipschitz
gradient. If $x^+ = x-\gamma g$ for $\gamma>0$, then
\[
\begin{aligned}
    f(x^+)
    &\leq f(x) - \frac{\gamma}{2}\|\nabla f(x)\|^2
    - \left(\frac{1}{2\gamma}-\frac{L}{2}\right)\|x^+-x\|^2 \\
    &\quad + \frac{\gamma}{2}\|g-\nabla f(x)\|^2 .
\end{aligned}
\]
\end{lemma}

\begin{lemma}[Gradient bound from smoothness and lower boundedness]
\label{common:lem:smooth_lower_gradient}
Let $f:\mathbb{R}^d\to\mathbb{R}$ be differentiable with $L$-Lipschitz
gradient, and assume $f(x)\geq f_{\inf}$ for all $x$. Then, for all
$x\in\mathbb{R}^d$,
\[
    \|\nabla f(x)\|^2 \leq 2L\bigl(f(x)-f_{\inf}\bigr).
\]
\end{lemma}

\section{Expectation and variance identities}
\label{common:sec:expectation_variance}

\begin{lemma}[Variance decomposition]
\label{common:lem:variance_decomposition}
Let $X$ be a random vector with finite second moment, and let
$a\in\mathbb{R}^d$ be deterministic. Then
\[
    \mathbb{E}\|X-a\|^2
    = \mathbb{E}\|X-\mathbb{E}[X]\|^2 + \|\mathbb{E}[X]-a\|^2 .
\]
\end{lemma}

\begin{lemma}[Zero-mean uncorrelated sum]
\label{common:lem:zero_mean_sum}
Let $X_1,\dots,X_n$ be random vectors with finite second moments such that
$\mathbb{E}[X_i]=0$ and $\mathbb{E}\langle X_i,X_j\rangle=0$ for all
$i\neq j$. Then
\[
    \mathbb{E}\left\|\sum_{i=1}^n X_i\right\|^2
    = \sum_{i=1}^n \mathbb{E}\|X_i\|^2 .
\]
\end{lemma}

\begin{lemma}[Jensen bound for expected norms]
\label{common:lem:jensen_expectation_norm}
Let $X$ be a random vector with finite second moment. Then
\[
    \mathbb{E}\|X\| \leq \sqrt{\mathbb{E}\|X\|^2}.
\]
\end{lemma}

\section{Averaging/Jensen-type inequalities}
\label{common:sec:averaging_jensen}

\begin{lemma}[Norm of an average]
\label{common:lem:average_norm}
For any $a_1,\dots,a_n\in\mathbb{R}^d$,
\[
    \left\|\frac{1}{n}\sum_{i=1}^n a_i\right\|
    \leq \frac{1}{n}\sum_{i=1}^n \|a_i\|.
\]
\end{lemma}

\begin{lemma}[Squared norm of an average]
\label{common:lem:average_squared_norm}
For any $a_1,\dots,a_n\in\mathbb{R}^d$,
\[
    \left\|\frac{1}{n}\sum_{i=1}^n a_i\right\|^2
    \leq \frac{1}{n}\sum_{i=1}^n \|a_i\|^2 .
\]
\end{lemma}

\begin{lemma}[Convexity under expectation]
\label{common:lem:convex_expectation}
Let $\psi(\cdot,\xi)$ be convex for every value of $\xi$, and suppose the
expectation is well defined. Then $x \mapsto \mathbb{E}_{\xi}[\psi(x,\xi)]$
is convex.
\end{lemma}

\chapter{Supplementary Material for Chapter 2 (IST)}
\label{appendix:chapter2_ist}

\begingroup
\input{KAUST_Thesis/imported/02_ist/import_setup}

\section{Matrix smoothness and weighted-norm facts}

$\mL$-matrix smoothness:
\begin{equation} \label{ist:eq:L-matrix-smooth_}
   f(x + h) \leq f(x) + \left\langle\nabla f(x), h \right\rangle + \frac{1}{2} \left\langle \mL h, h \right\rangle, \qquad \forall x, h \in \Rd.
\end{equation}

\paragraph{Basic Inequalities.}
We use the polarization identity, weighted Young inequality, variance decomposition, averaging inequality, and Fenchel--Young inequality collected in Appendix~\ref{appendix:common_facts}. The following weighted-norm specialization is used below.
\begin{equation} \label{ist:eq:fy-ineq_quad}
    \left\langle x, y \right\rangle \leq \beta \sqn{x}_{\mL} + \frac{1}{4} \beta^{-1} \sqn{y}_{\mL^{-1}}.
\end{equation}

\section{Proofs}

\subsection{Permutation sketch computations}

All derivations in this section are performed for the $n = d$ case.

\paragraph{Classical Permutation Sketching.}
\texttt{Perm-1}: $\mC_i = n e_{\pi_i} e_{\pi_i}^\top$, where $\pi = (\pi_1, \dots, \pi_n)$ is a random permutation of $[n]$.
For the homogeneous problem $\mL_i \equiv \mL$:

\begin{equation}
    \Exp{\moB^k} = \Exp{\frac{1}{n} \sum_{i=1}^n \bfC_i \mL \bfC_i} = n \Diag(\mL)
\end{equation}
Then
\begin{equation}
    2 \moW = \Exp{\mL \moB^k + \moB^k \mL} = n \br{\mL\Diag(\mL) + \Diag(\mL) \mL}
\end{equation}
and
\begin{equation}
    \Exp{\moB^k\mL\moB^k} = n^2 \Diag(\mL) \mL \Diag(\mL).
\end{equation}
By repeating basically the same calculations for $\mC'_i = \sqrt{n} e_{\pi_i} e_{\pi_i}^\top$ we have that
\begin{equation}
    \Exp{\moB^k} = \Exp{\frac{1}{n} \sum_{i=1}^n \mC'_i \mL \mC'_i} = \Diag(\mL),
\end{equation}
and
$\Exp{\moB^k\mL\moB^k} = \Diag(\mL) \mL \Diag(\mL)$, $2 \moW = \Exp{\mL \moB^k + \moB^k \mL} = \mL\Diag(\mL) + \Diag(\mL) \mL$.

\subsubsection{Heterogeneous sketch preconditioning} \label{ist:sec:het_sketch_precond}
We recall the following modification of \texttt{Perm-1}:
\begin{equation}
    \tilde{\bfC}_i \eqdef \sqrt{n/\sbr{\mL_i}_{\pi_i, \pi_i}} e_{\pi_i} e_{\pi_i}^\top.
\end{equation}
Then
\begin{equation}
    \Exp{\tilde{\mC}_i \mL_i \tilde{\mC}_i} =
    \Exp{n [\mL_i]^{-1}_{\pi_i, \pi_i} e_{\pi_i} e_{\pi_i}^\top \mL_i e_{\pi_i} e_{\pi_i}^\top} =
    \frac{1}{n} \sum_{j=1}^n n e_{j} \mathbf{I}_{j, j} e_{j}^\top =
    \mI.
\end{equation}
and
\begin{eqnarray*}
    \Exp{\moB^k} &=&
    \Exp{\frac{1}{n} \sum_{i=1}^n \tilde{\mC}_i \mL_i \tilde{\mC}_i}
    \\&=&
    \frac{1}{n} \sum_{i=1}^n \Exp{n[\mL_i]^{-1}_{\pi_i, \pi_i} e_{\pi_i} e_{\pi_i}^\top \mL_i e_{\pi_i} e_{\pi_i}^\top}
    \\&=&
    \frac{1}{n} \sum_{i=1}^n \frac{1}{n} \sum_{j=1}^n n[\mL_i]^{-1}_{j, j} e_{j} [\mL_i]_{j,j} e_{j}^\top
    \\&=&
    \frac{1}{n} \sum_{i=1}^n \sum_{j=1}^n  e_{j} e_{j}^\top
    \\&=&
    \mI.
\end{eqnarray*}
Thus $\moW = \frac{1}{2} \Exp{\moL \moB^k + \moB^k \moL} = \moL$.
On the left hand side of inequality \eqref{ist:eq:het_exp_sep}, we have
\begin{eqnarray*}
    \Exp{\moB^k \moL \moB^k}
    &=&
    \Exp{\frac{1}{n} \sum_{i=1}^n \tilde{\mC}_i \mL_i \tilde{\mC}_i \moL \frac{1}{n} \sum_{i=j}^n \tilde{\mC}_j \mL_j \tilde{\mC}_j}
    \\&=&
    \frac{1}{n^2} \sum_{i,j=1}^n \Exp{\tilde{\mC}_i \mL_i \tilde{\mC}_i \moL \tilde{\mC}_j \mL_j \tilde{\mC}_j}
    \\&=&
    \sum_{i,j=1}^n e_i e_i^\top \moL e_j e_j^\top
    \\&=&
    \mI \moL \mI
    \\&=&
    \moL.
\end{eqnarray*}

\subsection{Interpolation case: proof of theorem \ref{ist:thm:het_quad}}

In the quadratic interpolation regime, the linear term is zero
$f_i(x) = \frac{1}{2} x^\top \mL_i x$,
and the gradient estimator has the form
\begin{equation}
   g^k = \frac{1}{n} \sum \limits_{i=1}^n \mC_i^k \nabla f_i(\mC_i^k x^k) =  \frac{1}{n} \sum \limits_{i=1}^n \mC_i^k \mL_i \mC_i^k x^k = \moB^k x^k.
\end{equation}

\begin{proof}
First, we prove the \textbf{stationary point} convergence result \eqref{ist:eq:het_res}.

Using the $\moL$-smoothness of function $f$, we get
\begin{eqnarray*}
    f(x^{k+1}) \overset{\eqref{ist:eq:SGD_generic} }{=} f(x^k - \gamma g^k)
    & \overset{\eqref{ist:eq:L-matrix-smooth}}{\leq} &
    f(x^k) - \left\langle\nabla f(x^k), \gamma g^k \right\rangle + \frac{\gamma^2}{2} \sqN{g^k}_{\moL} \\
    &\overset{\eqref{ist:eq:het_estimator}}{=} &
    f(x^k) - \gamma \left\langle \moL x^k, \moB^k x^k \right\rangle + \frac{\gamma^2}{2} \sqN{\moB^k x^k}_{\moL} \\
    &= & f(x^k) - \gamma (x^k)^\top \moL \moB^k x^k  + \frac{\gamma^2}{2}
 (x^k)^\top \moB^k \moL \moB^k x^k.
\end{eqnarray*}
After applying conditional expectation, using its linearity, and the fact that
$$x^\top \mA x = \frac{1}{2} x^\top \br{\mA + \mA^\top}x$$ we get
\begin{eqnarray*}
    \Exp{ f(x^{k+1}) \;|\; x^k}
    & \leq & f(x^k) - \gamma (x^k)^\top \Exp{\moL\moB^k} x^k  + \frac{\gamma^2}{2}
    (x^k)^\top \Exp{\moB^k \moL \moB^k } x^k\\
    &=&f(x^k) - \gamma (x^k)^\top \moW x^k  + \frac{\gamma^2}{2}
    (x^k)^\top \Exp{\moB^k \moL \moB^k } x^k\\
    &=&f(x^k) - \gamma ( \nabla f(x^k))^\top  \moL^{-1}\moW \moL^{-1} \nabla f(x^k)
    \\&& \qquad + \frac{\gamma^2}{2}
    (\nabla f(x^k))^\top \moL^{-1}\Exp{\moB^k \moL \moB^k} \moL^{-1}  \nabla f(x^k) \\
    &\overset{\eqref{ist:eq:het_exp_sep}}{\leq}&f(x^k) - \gamma \norm{\nabla f(x^k)}^2_{\moL^{-1}\moW \moL^{-1}}  + \frac{\theta \gamma^2}{2} \norm{\nabla f(x^k)}^2_{\moL^{-1}\moW \moL^{-1}} \\
    &=&
    f(x^k) - \gamma \br{1 - \nicefrac{\theta\gamma}{2}} \norm{\nabla f(x^k)}^2_{\moL^{-1}\moW \moL^{-1}}
    \\ & \leq &
    f(x^k) - \frac{\gamma}{2} \sqn{\nabla f(x^k)}_{\moL^{-1}
    \moW \moL^{-1}},
\end{eqnarray*}
where the last inequality holds for the stepsize $\gamma \leq \frac{1}{\theta}$.

Rearranging gives
\begin{equation*}
    \sqN{\nabla f(x^k)}_{\moL^{-1} \moW \moL^{-1}} \leq
    \frac{2}{\gamma} \br{f(x^k) - \Exp{ f(x^{k+1}) \;|\; x^k}},
\end{equation*}
which after averaging gives the desired result
\begin{equation}
\begin{aligned}
    \frac{1}{K} \sum_{k=0}^{K-1} \Exp{\sqN{\nabla f(x^k)}_{\moL^{-1} \moW \moL^{-1}}}
    \leq
    \frac{2}{\gamma K} \sum_{k=0}^{K-1} (f(x^{k}) - \Exp{f(x^{k+1})})
    \\=
    \frac{2 \br{f(x^0) - \Exp{f(x^K)}}}{\gamma K}.
\end{aligned}
\end{equation}

Now we show the result for the \textbf{iterates convergence} \eqref{ist:eq:het_res_it}.

Expectation conditioned on $x^k$:
\begin{align*}
\Exp{\sqn{x^{k+1} - x^\star}_{\moL}}
&=
\Exp{\sqn{x^k - \gamma g^k - x^\star}_{\moL}}
\\&=
\sqn{x^k - x^\star}_{\moL}
- 2 \gamma \left\langle x^k - x^\star, \Exp{\moL\moB^k} (x^k - x^\star) \right\rangle
\\&\qquad
+ \gamma^2 \left\langle \Exp{\moB^k \moL \moB^k}(x^k - x^\star), x^k - x^\star \right\rangle
\\&\overset{x^\star=0}{=}
\sqn{x^k - x^\star}_{\moL}
- 2 \gamma \left\langle x^k - x^\star, \moW (x^k - x^\star) \right\rangle
\\&\qquad
+ \gamma^2 \left\langle x^k - x^\star, \Exp{\moB^k \moL \moB^k} (x^k - x^\star) \right\rangle
\\&\overset{\eqref{ist:eq:het_exp_sep}}{\leq}
\sqn{x^k - x^\star}_{\moL}
- 2 \gamma \left\langle x^k - x^\star, \moW (x^k - x^\star) \right\rangle
\\&\qquad
+ \theta\gamma^2 \left\langle x^k - x^\star, \moW (x^k - x^\star) \right\rangle
\\&=
\sqn{x^k - x^\star}_{\moL}
- 2\gamma \br{1 - \theta\gamma/2}
\sqN{\moL^{\frac{1}{2}} (x^k - x^\star)}_{\moL^{-\frac{1}{2}}\moW\moL^{-\frac{1}{2}}}
\\&\overset{\gamma \leq \nicefrac{1}{\theta}}{\leq}
\sqn{x^k - x^\star}_{\moL}
- \gamma \sqN{\moL^{\frac{1}{2}}(x^k - x^\star)}_{\moL^{-\frac{1}{2}}\moW\moL^{-\frac{1}{2}}}
\\&\leq
\sqn{x^k - x^\star}_{\moL}
- \gamma \lambda_{\min} \br{\moL^{-\frac{1}{2}}\moW\moL^{-\frac{1}{2}}}
\sqN{\moL^{\frac{1}{2}}(x^k - x^\star)}
\\&=
\br{1 - \gamma \lambda_{\min} \br{\moL^{-\frac{1}{2}}\moW\moL^{-\frac{1}{2}}}} \sqn{x^k - x^\star}_{\moL}.
\end{align*}
After unrolling the recursion we obtain the convergence result
\begin{equation*}
    \Exp{\sqn{x^{k+1} - x^\star}_{\moL}} \leq
    \br{1 - \gamma \lambda_{\min} \br{\moL^{-\frac{1}{2}}\moW\moL^{-\frac{1}{2}}}}^{k+1} \sqn{x^0 - x^\star}_{\moL}.
\end{equation*}

\end{proof}

\subsection{Nonzero solution}

As a reminder, in the most general case, the problem has the form
\begin{equation*}
   f(x) = \frac{1}{n} \sum \limits_{i=1}^n f_i(x), \qquad f_i(x) \equiv \frac{1}{2} x^\top \mL_i x - x^\top \bb_i.
\end{equation*}

with the gradient estimator
\begin{equation}
   g^k = \frac{1}{n} \sum \limits_{i=1}^n \mC_i^k \nabla f_i(\mC_i^k x^k) =  \frac{1}{n} \sum \limits_{i=1}^n \mC_i^k \br{\mL_i \mC_i^k x^k - \mathrm{b}_i} = \moB^k x^k - \frac{1}{n} \sum \limits_{i=1}^n \mC_i^k \bb_i.%
\end{equation}

\paragraph{General Calculations for Estimator \eqref{ist:eq:het_gen_quad_est}.}
In the heterogeneous case, the following sketch preconditioner is used
\begin{equation*}
    \tilde{\bfC}_i \eqdef \sqrt{n/\sbr{\mL_i}_{\pi_i, \pi_i}} e_{\pi_i} e_{\pi_i}^\top.
\end{equation*}
Then $\Exp{\moB^k} = \mI$ (calculation was done as in Section \ref{ist:sec:het_sketch_precond}) and
\begin{eqnarray*}
    \Exp{\oCb}
    &=&
    \frac{1}{n}\sum_{i=1}^n \Exp{\tilde{\mC}_i^k \bb_i}
    \\&=&
    \frac{1}{n}\sum_{i=1}^n \Exp{\sqrt{n} [\mL_i]^{-\frac{1}{2}}_{\pi_i, \pi_i} e_{\pi_i} e_{\pi_i}^\top \bb_i} \\&=&
    \frac{1}{n} \sum_{i=1}^n \frac{1}{n} \sum_{j=1}^n \sqrt{n} [\mL_i]^{-\frac{1}{2}}_{j,j} e_{j} [\bb_i]_j
    \\&=&
    \frac{1}{n} \sum_{i=1}^n \frac{1}{n} \sqrt{n} \mD_i^{-\frac{1}{2}} \bb_i
    \\&=&
    \frac{1}{\sqrt{n}} \frac{1}{n} \sum_{i=1}^n \mD_i^{-\frac{1}{2}} \bb_i
    \\&=&
    \frac{1}{\sqrt{n}} \underbrace{\oDb}_{\cDb}
\end{eqnarray*}

\subsubsection{Convergence analysis for heterogeneous case: proof of theorem \ref{ist:thm:het_gen_gradient}}

Here we formulate and further prove a more general version of Theorem \ref{ist:thm:het_gen_gradient}, which is obtained as a special case of the next result for $c=1/2$.

\begin{theorem}
Consider the method \eqref{ist:eq:SGD_generic} with estimator \eqref{ist:eq:het_gen_quad_est} for a quadratic problem \eqref{ist:eq:het_gen_quad_problem} with positive-definite matrix $\moL \succ 0$. Then, if for every $\mD_i \eqdef \Diag(\mL_i)$ matrices $\mD_i^{-\frac{1}{2}}$ exist, scaled permutation sketches $\bfC_i  \eqdef \sqrt{n} [\mL_i^{-\frac{1}{2}}]_{\pi_i, \pi_i} e_{\pi_i} e_{\pi_i}^\top$ are used and heterogeneity is bounded as $\Exp{\sqN{g^k - \Exp{g^k}}_{\moL}} \leq \sigma^2$. Then, the step size is chosen as
\begin{equation}
    0 < \gamma \leq \gamma_{c, \beta} \eqdef \frac{1 - c - \beta}{\beta + 1/2},
\end{equation}
where $\gamma_{c, \beta} \in (0, 1]$ for $\beta + c < 1$, the iterates satisfy
\begin{equation}
    \frac{1}{K} \sum_{k=0}^{K-1} \Exp{\sqN{\nabla f(x^k)}_{\moL^{-1}}}
    \leq
    \frac{f(x^0) - \Exp{f(x^K)}}{c\gamma K} + \br{\frac{1 - \gamma}{c\beta} + \frac{\gamma}{2c}} \sqn{h}_{\moL} + \frac{\gamma}{2c} \sigma^2.
\end{equation}
where $\moL = \frac{1}{n} \sum_{i=1}^n \mL_i, h = \moL^{-1}\ob - \frac{1}{\sqrt{n}} \frac{1}{n} \sum_{i=1}^n \mD_i^{-\frac{1}{2}} \bb_i$ and $\ob = \frac{1}{n} \sum_{i=1}^n \bb_i$.
\end{theorem}

\begin{proof}
By using $\mL$-smoothness
\begin{align*}
\Exp{f(x^{k+1}) \;|\; x^k}
&\overset{\eqref{ist:eq:L-matrix-smooth}}{\leq}
f(x^k) - \gamma \left\langle \nabla f(x^k), \Exp{g^k} \right\rangle
+ \frac{\gamma^2}{2} \Exp{\sqn{g^k}_{\moL}}
\\&\overset{\eqref{ist:eq:het_gen_estimator},\,\text{Lemma~\ref{common:lem:variance_decomposition}}}{=}
f(x^k) - \gamma \left\langle \nabla f(x^k), \moL^{-1} \nabla f(x^k) + h \right\rangle
\\&\qquad +
\frac{\gamma^2}{2} \br{\sqN{\Exp{g^k}}_{\moL} + \Exp{\sqN{g^k - \Exp{g^k}}_{\moL}}}
\\&\overset{\eqref{ist:eq:het_gen_estimator}}{=}
f(x^k) - \gamma \br{\left\langle \nabla f(x^k), \moL^{-1} \nabla f(x^k) \right\rangle
+ \left\langle \nabla f(x^k), h \right\rangle}
\\&\qquad +
\frac{\gamma^2}{2} \br{\sqN{\moL^{-1} \nabla f(x^k) + h}_{\moL}
+ \Exp{\sqN{g^k - \Exp{g^k}}_{\moL}}}
\\&\overset{\text{Lemma~\ref{common:lem:polarization}}}{=}
f(x^k) - \gamma \br{\sqN{\nabla f(x^k)}_{\moL^{-1}}
+ \left\langle \nabla f(x^k), h \right\rangle}
+ \frac{\gamma^2}{2} \Exp{\sqN{g^k - \Exp{g^k}}_{\moL}}
\\&\qquad +
\frac{\gamma^2}{2} \br{\sqN{\nabla f(x^k)}_{\moL^{-1}}
+ 2\left\langle \nabla f(x^k), h \right\rangle + \sqn{h}_{\moL}}
\\&\leq
f(x^k) - \gamma \br{1 - \nicefrac{\gamma}{2}} \sqN{\nabla f(x^k)}_{\moL^{-1}}
+ \frac{\gamma^2}{2} \sigma^2
\\&\qquad
- \gamma \br{1 - \gamma} \left\langle \nabla f(x^k), h \right\rangle
+ \frac{\gamma^2}{2} \sqn{h}_{\moL}.
\end{align*}
where the last inequality follows from the grouping of similar terms and bounded heterogeneity
\begin{eqnarray} \label{ist:eq:hetero_bound}
    \Exp{\sqN{g^k - \Exp{g^k}}_{\moL}} &=&
    \Exp{\sqN{g^k - \br{\moL^{-1} \nabla f(x^k) + h}}_{\moL}}
    \\&=&
    \Exp{\sqN{\moB^k x^k - \oCb - \br{x^k - \frac{1}{\sqrt{n}}\cDb}}_{\moL}} \leq \sigma^2.
\end{eqnarray}

Next, using a Fenchel-Young inequality \eqref{ist:eq:fy-ineq_quad} for  $\left\langle \nabla f(x^k), -h \right\rangle$ and $1 - \gamma \geq 0$
\begin{eqnarray}  \label{ist:eq:het_prec_gen_descent}
\Exp{f(x^{k+1}) \;|\; x^k}
&\leq&
f(x^k) - \gamma \br{1 - \nicefrac{\gamma}{2}} \sqN{\nabla f(x^k)}_{\moL^{-1}} + \frac{\gamma^2}{2} \br{\sqn{h}_{\moL} +  \sigma^2} \notag
\\&& \qquad + \gamma \br{1 - \gamma} \sbr{\beta \sqn{\nabla f(x^k)}_{\mcL^{-1}} + 0.25 \beta^{-1}\sqn{h}_{\mcL}} \notag
\\ &\leq&
f(x^k) - \gamma \br{1 - \nicefrac{\gamma}{2} - \beta \br{1 - \gamma}} \sqN{\nabla f(x^k)}_{\moL^{-1}} \notag
\\&& \qquad +
\gamma \cbr{\br{\beta^{-1} \br{1 - \gamma} + \frac{\gamma}{2}}\sqn{h}_{\moL} + \frac{\gamma}{2} \sigma^2},
\end{eqnarray}
where in the last inequality we grouped similar terms and used the fact that $0.25 < 1$.

Now to guarantee that $1 - \nicefrac{\gamma}{2} - \beta(1 - \gamma) \geq c >0$, we choose the step size using
\begin{equation}
    0 < \gamma \leq \gamma_{c, \beta} \eqdef \frac{1 - c - \beta}{\beta + 1/2},
\end{equation}
where $\gamma_{c, \beta} > 0$ for $\beta + c < 1$. This means that $\beta$ cannot arbitrarily grow to diminish $\beta^{-1}$. \\
Then, after standard manipulations and unrolling the recursion
\begin{equation}
     \gamma c \sqN{\nabla f(x^k)}_{\moL^{-1}} \leq f(x^k) - \Exp{f(x^{k+1}) \;|\; x^k} + \gamma \br{\beta^{-1}\br{1 - \gamma} + \gamma/2} \sqn{h}_{\moL} + \frac{\gamma^2}{2} \sigma^2
\end{equation}
we obtain
\begin{equation}
    \frac{c}{K} \sum_{k=0}^{K-1} \Exp{\sqN{\nabla f(x^k)}_{\moL^{-1}}}
    \leq
    \frac{f(x^0) - \Exp{f(x^K)}}{\gamma K} + \br{\beta^{-1}\br{1 - \gamma} + \gamma/2} \sqn{h}_{\moL} + \frac{\gamma}{2} \sigma^2.
\end{equation}

\end{proof}

\subsubsection{Homogeneous case}
The main difference compared to the result in the previous subsection is that the gradient estimator expression \eqref{ist:eq:hom_gen_estimator} holds deterministically (without expectation $\mathbb{E}$). That is why $g^k = \Exp{g^k}$ and heterogeneity term $\sigma^2$ equals to 0.

We provide the full statement and proof for the homogeneous result discussed in \ref{ist:sec:gen_convergence}.
\begin{theorem} \label{ist:thm:gen_gradient}
Consider the method \eqref{ist:eq:SGD_generic} with estimator \eqref{ist:eq:hom_gen_estimator} for a homogeneous quadratic problem \eqref{ist:eq:het_gen_quad_problem} with positive-definite matrix $\mL_i \equiv \mL \succ 0$. Then if exists $\mD^{-\frac{1}{2}}$ for $\mD \eqdef \Diag(\mL)$, scaled permutation sketch $\mC'_i = \sqrt{n} e_{\pi_i} e_{\pi_i}^\top$ is used and the step size is chosen as
\begin{equation}
    0 < \gamma \leq \gamma_{c, \beta} \eqdef \frac{1 - c - \beta}{\beta + 1/2},
\end{equation}
where $\gamma_{c, \beta} > 0$ for $\beta + c < 1$. Then the iterates satisfy
\begin{equation} \label{ist:eq:gen_grad_res}
    \frac{1}{K} \sum_{k=0}^{K-1} \Exp{\sqN{\nabla f(x^k)}_{\mcL^{-1}}}
    \leq
    \frac{f(x^0) - \Exp{f(x^K)}}{c \gamma K} + \br{\frac{1 - \gamma}{c\beta} + \frac{\gamma}{2c}} \sqn{h}_{\mcL},
\end{equation}
where $\mcL = \mD^{-\frac{1}{2}} \mL \mD^{-\frac{1}{2}}, h = \mcL^{-1}\cb - \frac{1}{\sqrt{n}}\cb$ and $\cb = \mD^{-\frac{1}{2}} \bb$.
\end{theorem}

\begin{proof}
By using $\mL$-smoothness
\begin{align*}
\Exp{f(x^k - \gamma g^k) \;|\; x^k}
    &\overset{\eqref{ist:eq:L-matrix-smooth}}{\leq}
    f(x^k) - \left\langle\nabla f(x^k), \gamma \Exp{g^k} \right\rangle + \frac{\gamma^2}{2} \Exp{\sqN{g^k}_{\mcL}} \\
&\leq
f(x^k) - \gamma \left\langle \nabla f(x^k), \mcL^{-1} \nabla f(x^k) + h \right\rangle
+ \frac{\gamma^2}{2} \sqN{\mcL^{-1} \nabla f(x^k) + h}_{\mcL}
 \\&\overset{\text{Lemma~\ref{common:lem:polarization}}}{=}
 f(x^k) - \gamma \br{\left\langle \nabla f(x^k), \mcL^{-1} \nabla f(x^k) \right\rangle
 + \left\langle \nabla f(x^k), h \right\rangle}
 \\&\qquad +
 \frac{\gamma^2}{2} \br{\sqN{\nabla f(x^k)}_{\mcL^{-1}}
 + 2\left\langle \nabla f(x^k), h \right\rangle + \sqn{h}_{\mcL}}
 \\&=
 f(x^k) - \gamma \br{1 - \nicefrac{\gamma}{2}} \sqN{\nabla f(x^k)}_{\mcL^{-1}}
 \\&\qquad
+ \frac{\gamma^2}{2} \sqn{h}_{\mcL}
- \gamma \br{1 - \gamma} \left\langle \nabla f(x^k), h \right\rangle
\end{align*}
Next by using a Fenchel-Young inequality \eqref{ist:eq:fy-ineq_quad} for  $\left\langle \nabla f(x^k), -h \right\rangle$ and $1 - \gamma \geq 0$

\begin{align*}
\Exp{f(x^{k+1}) \;|\; x^k}
&\leq
f(x^k) - \gamma \br{1 - \nicefrac{\gamma}{2}} \sqN{\nabla f(x^k)}_{\mcL^{-1}}  + \frac{\gamma^2}{2} \sqn{h}_{\mcL}
\\ &\qquad + \gamma \br{1 - \gamma} \sbr{\beta \sqn{\nabla f(x^k)}_{\mcL^{-1}} + 0.25\beta^{-1}\sqn{h}_{\mcL}}
\\ &=
f(x^k) - \gamma \br{1 - \nicefrac{\gamma}{2} - \beta(1 - \gamma)} \sqN{\nabla f(x^k)}_{\mcL^{-1}}
\\ &\qquad + \gamma \br{\beta^{-1}\br{1 - \gamma} + \gamma/2} \sqn{h}_{\mcL}.
\end{align*}
Now to guarantee that $1 - \nicefrac{\gamma}{2} - \beta(1 - \gamma) \geq c > 0$ we choose the step size as
\begin{equation}
    0 < \gamma \leq \gamma_{c, \beta} \eqdef \frac{1 - c - \beta}{\beta + 1/2},
\end{equation}
where $\gamma_{c, \beta} \geq 0$ for $\beta + c < 1$.
\\
Then after standard manipulations and unrolling the recursion
\begin{equation} \label{ist:eq:descent_recursion}
     \gamma c \sqN{\nabla f(x^k)}_{\mcL^{-1}} \leq f(x^k) - \Exp{f(x^{k+1}) \;|\; x^k} + \gamma \br{\beta^{-1}\br{1 - \gamma} + \gamma/2} \sqn{h}_{\mcL}
\end{equation}
we obtain the formulated result
\begin{equation}
    \frac{c}{K} \sum_{k=0}^{K-1} \Exp{\sqN{\nabla f(x^k)}_{\mcL^{-1}}}
    \leq
    \frac{f(x^0) - \Exp{f(x^K)}}{\gamma K} + \br{\beta^{-1}\br{1 - \gamma} + \gamma/2} \sqn{h}_{\mcL}.
\end{equation}
\end{proof}

\begin{remark}
\textbf{1)} The first term in the convergence upper bound \eqref{ist:eq:gen_grad_res} is minimized by maximizing product $c\cdot\gamma$, which motivates to choose $c>0$ and $\gamma \leq 1$ as large as possible. Although due to the constraint on the step size (and $\beta > 0$)
\begin{equation}
    0 < \gamma \leq \gamma_{c, \beta} \eqdef \frac{1 - c - \beta}{\beta + 1/2},
\end{equation}
constant $c \in (0, 1)$. So, by maximizing $c$ the value $\gamma_{c, \beta}$ becomes smaller, thus there is a trade-off.

\textbf{2)} The second term or the neighborhood size (multiplier in front of $\sqn{h}_{\mcL}$)
\begin{equation}
    \Psi(\beta, \gamma) \eqdef \frac{\beta^{-1}\br{1 - \gamma} + \gamma/2}{c} = \frac{\beta^{-1}\br{1 - \gamma} + \gamma/2}{1 - \gamma/2 - \beta (1 - \gamma)}
\end{equation}
can be numerically minimized (e.g.~by using WolframAlpha) with constraints $\gamma\in(0, 1]$ and $\beta > 0$. The solution of such optimization problem is $\gamma^\star \approx 1$ and $\beta^\star \approx \xi \in \{3.992, 2.606, 2.613\}$. In fact, $\Psi(\beta^\star, \gamma^\star) \approx 0.5$.
\end{remark}

\paragraph{Functional Gap Convergence.}
Note that for the quadratic optimization problem \eqref{ist:eq:het_gen_quad_problem}
\begin{equation} \label{ist:eq:grad_func}
    \sqN{\nabla f(x^k)}_{\mcL^{-1}} = \left\langle \mcL x^k - \cb, \mcL^{-1} \br{\mcL x^k - \cb} \right\rangle = 2 \br{f(x^k) - f(x^\star)}.
\end{equation}
Then by rearranging and subtracting $f^\star \eqdef f(x^\star)$ from both sides of inequality \eqref{ist:eq:descent_recursion} we obtain
\begin{align*}
\Exp{f(x^{k+1}) \;|\; x^k} - f^\star
&\leq
f(x^k) - f^\star - \gamma c \sqN{\nabla f(x^k)}_{\mcL^{-1}}
+ \gamma \br{\beta^{-1}\br{1 - \gamma} + \gamma/2} \sqn{h}_{\mcL}
\\ &\overset{\eqref{ist:eq:grad_func}}{=}
\br{f(x^k) - f^\star} - \gamma c \cdot 2\br{f(x^k) - f^\star}
+ \gamma \br{\beta^{-1}\br{1 - \gamma} + \gamma/2} \sqn{h}_{\mcL}
\\ &=
\br{1 - 2 \gamma c} \br{f(x^k) - f^\star} + \gamma \br{\beta^{-1}\br{1 - \gamma} + \gamma/2} \sqn{h}_{\mcL}.
\end{align*}
After unrolling the recursion
\begin{align*}
\Exp{f(x^{k+1}) \;|\; x^k} - f^\star
&\leq
\br{1 - 2 \gamma c}^k \br{f(x^0) - f^\star}
+ \gamma \br{\beta^{-1}\br{1 - \gamma} + \gamma/2} \sqn{h}_{\mcL}
\sum_{i=0}^k \br{1 - 2 \gamma c}^i
\\ &\leq
\br{1 - 2 \gamma c}^k \br{f(x^0) - f^\star}
+ \frac{1}{2c} \br{\beta^{-1}\br{1 - \gamma} + \gamma/2} \sqn{h}_{\mcL}.
\end{align*}

This result is formalized in the following Theorem.
\begin{theorem} \label{ist:thm:gen_function}
Consider the method \eqref{ist:eq:SGD_generic} with estimator \eqref{ist:eq:hom_gen_estimator} for a homogeneous quadratic problem \eqref{ist:eq:het_gen_quad_problem} with positive-definite matrix $\mL_i \equiv \mL \succ 0$. Then if exists $\mD^{-\frac{1}{2}}$ for $\mD \eqdef \Diag(\mL)$, scaled permutation sketch $\mC'_i = \sqrt{n} e_{\pi_i} e_{\pi_i}^\top$ is used and the step size is chosen as
\begin{equation}
    0 < \gamma \leq \gamma_{c, \beta} \eqdef \frac{1 - c - \beta}{\beta + 1/2},
\end{equation}
where $\gamma_{c, \beta} >0$ for $\beta + c < 1$. Then the iterates satisfy
\begin{equation} \label{ist:eq:gen_func_res}
    \Exp{f(x^k)} - f^\star \leq \br{1 - 2 \gamma c}^k \br{f(x^0) - f^\star} + \frac{1}{2c} \br{\beta^{-1}\br{1 - \gamma} + \gamma/2} \sqn{h}_{\mcL},
\end{equation}
where $h = \mcL^{-1}\cb - \frac{1}{\sqrt{n}}\cb$ and $\mcL = \mD^{-\frac{1}{2}} \mL \mD^{-\frac{1}{2}}, \cb = \mD^{-\frac{1}{2}} \bb$.
\end{theorem}

This result shows that for a proper choice of the step size $\gamma = 1$ and constant $c=1/2$, the functional gap can converge in basically one iteration to the neighborhood of size
\begin{equation*}
    \sqn{h}_{\mcL} = \left\langle \mcL \br{\mcL^{-1}\cb - \frac{1}{\sqrt{n}}\cb}, \mcL^{-1} \cb - \frac{1}{\sqrt{n}}\cb \right\rangle,
\end{equation*}
which equals zero if $\mcL^{-1} \cb = \frac{1}{\sqrt{n}}\cb$. This condition is the same as the condition we obtained at the end of Subsection \ref{ist:sec:gen_convergence} with asymptotic analysis of the iterates in the homogeneous case.

\paragraph{Discussion of the Trace.}
Consider a positive-definite $\mL \succ 0$ such that $\exists \mD^{-\frac{1}{2}}$. Thus $\mcL = \mD^{-\frac{1}{2}} \mL \mD^{-\frac{1}{2}}$ has only ones on the diagonal and $\tr(\mcL) = n$. Then
\begin{equation*}
    n \cdot \tr(\mcL^{-1}) = \tr(\mcL) \tr(\mcL^{-1}) = \br{\lambda_1 + \dots + \lambda_n} \br{\frac{1}{\lambda_1} + \dots + \frac{1}{\lambda_n}} \geq n^2,
\end{equation*}
where the last inequality is due to the relation between harmonic and arithmetic means. Therefore $\tr(\mcL^{-1}) = \lambda_1^{-1} + \dots + \lambda_n^{-1} \geq n$ and sum of $\mcL^{-1}$ eigenvalues has to be greater than $n$.

\subsection{\texorpdfstring{Generalization to the $n \neq d$ case}{Generalization to the n != d case}}
Our results can be generalized in a similar way as in \citet{szlendak2022permutation}.

\textbf{1)} $d = qn$, for integer $q \geq 1$. Let $\pi = \br{\pi_1, \dots, \pi_d}$ be a random permutation of $\{1, \dots,d\}$. Then for each $i \in \{1, \dots, n\}$ define
\begin{equation}
    \mC_i' := \sqrt{n} \cdot \sum_{j=q(i-1)+1}^{q i} e_{\pi_j} e_{\pi_j}^{\top}.
\end{equation}
Matrix $\Exp{\moB^k}$ for the homogeneous preconditioned case can be computed as follows:
\begin{eqnarray*}
    \Exp{\moB^k} &=&
    \Exp{\frac{1}{n} \sum_{i=1}^n \mC_i' \mcL \mC_i'}
    \\&=&
    \frac{1}{n} \sum_{i=1}^n\Exp{\sum_{j=q(i-1)+1}^{q i} n e_{\pi_j} e_{\pi_j}^{\top} \mcL e_{\pi_j} e_{\pi_j}^{\top}}
    \\&=&
    \sum_{i=1}^n \sum_{j=q(i-1)+1}^{q i} \Exp{e_{\pi_j} e_{\pi_j}^{\top} \mcL e_{\pi_j} e_{\pi_j}^{\top}}
    \\&=&
    \sum_{i=1}^n \sum_{j=q(i-1)+1}^{q i} \frac{1}{d} \sum_{l=1}^d e_{l} e_{l}^{\top} \mcL e_{l} e_{l}^{\top}
    \\&=&
    \sum_{i=1}^n \sum_{j=q(i-1)+1}^{q i} \frac{1}{d} \Diag(\mcL)
    \\&=&
    n \frac{q}{d} \Diag(\mcL)
    \\&=&
    \Diag(\mcL) \\&=& \mI.
\end{eqnarray*}
As for the linear term
\begin{eqnarray*}
\Exp{\mC'\bb} &=&
\Exp{\frac{1}{n} \sum_{i=1}^n \mC'_i \cb} =
\frac{1}{n} \sum_{i=1}^n \Exp{\sum_{j=q(i-1)+1}^{q i} \sqrt{n} e_{\pi_j} e_{\pi_j}^{\top} \cb}
\\&=&
\frac{1}{\sqrt{n}} \sum_{i=1}^n \sum_{j=q(i-1)+1}^{q i} \frac{1}{d} \mI \cb = \frac{\sqrt{n}q}{d} \mI \cb = \frac{1}{\sqrt{n}} \cb.
\end{eqnarray*}

\textbf{2)} $n = qd$, for integer $q \geq 1$. Define the multiset $S \eqdef \{1, \ldots, 1, 2, \ldots, 2, \ldots, d, \ldots, d\}$, where each number occurs precisely $q$ times. Let $\pi = \br{\pi_1, \ldots, \pi_n}$ be a random permutation of $S$. Then for each $i \in \{1, \ldots, n\}$ define
\begin{equation}
    \mC_i' := \sqrt{d} \cdot e_{\pi_i} e_{\pi_i}^{\top}.
\end{equation}

\begin{eqnarray*}
    \Exp{\moB^k} &=&
    \Exp{\frac{1}{n} \sum_{i=1}^n \mC_i' \mcL \mC_i'}
    =
    \frac{1}{n} \sum_{i=1}^n \Exp{d e_{\pi_i} e_{\pi_i}^{\top} \mcL e_{\pi_i} e_{\pi_i}^{\top}}
    \\&=&
    \frac{1}{n} \sum_{i=1}^n \frac{1}{d} \sum_{j=1}^d d e_{j} e_{j}^{\top} \mcL e_{j} e_{j}^{\top}
    =
    \frac{1}{n} \sum_{i=1}^n \Diag(\mcL)
    =
    \mI.
\end{eqnarray*}
The linear term
\begin{eqnarray*}
    \Exp{\mC'\bb}=
    \Exp{\frac{1}{n} \sum_{i=1}^n \mC'_i \cb} =
    \frac{1}{n} \sum_{i=1}^n \Exp{\sqrt{d} e_{\pi_i} e_{\pi_i}^{\top} \cb}
    =
    \frac{\sqrt{d}}{n} \sum_{i=1}^n \frac{1}{d} \mI \cb = \frac{1}{\sqrt{d}} \cb.
\end{eqnarray*}

To sum up both cases, in a homogeneous preconditioned setting
$\Exp{\moB^k} = \mI$ and
$$\Exp{\mC'\bb} = \Exp{\frac{1}{n} \sum_{i=1}^n \mC'_i \bb} = \cb/\sqrt{\min(n, d)}.$$

Similar modifications and calculations can be performed for heterogeneous scenarios.
The cases in which $n$ does not divide $d$, or vice versa, can be handled using constructions from \citet{szlendak2022permutation}.

\section{Generalization beyond quadratics} \label{ist:sec:generalization}

In this section, we revisit our analysis for the more general case of smooth (non-convex) functions and a class of compressors. Although this result is less detailed than the result for the quadratic model, it still does not require restrictive assumptions on the gradient estimator, unlike prior work.

\subsection{Preliminary facts}
Our convergence analysis relies on the following Lemma.

\begin{lemma}[Descent Lemma \citep{li2021page}] \label{ist:lem:descent_lemma_appendix}
Suppose that function $f$ is $L$-smooth \eqref{ist:eq:L-matrix-smooth} and let $x^{k+1} \eqdef x^k - \gamma g^k$. Then for any $g^k \in \mathbb{R}^{d}$ and $\gamma>0$, we have
\begin{equation} \label{ist:eq:descent_lemma}
    f(x^{k+1}) \leq
    f(x^k) - \frac{\gamma}{2} \sqN{\nabla f(x^k)}
    - \br{\frac{1}{2\gamma} - \frac{L}{2}} \sqn{x^{k+1} - x^k}
    + \frac{\gamma}{2} \sqN{g^k - \nabla f(x^k)},
\end{equation}
\end{lemma}

If a continuously differentiable function $f$ is $L$-smooth \eqref{ist:eq:L-matrix-smooth}, then for any $x, y \in \mathbb{R}^{d}$, the following inequality holds:
\begin{equation} \label{ist:eq:smoothness}
    \norm{\nabla f(x) - \nabla f(y)} \leq L \norm{x - y}.
\end{equation}

\begin{lemma}[\citet{khaled2020better}]
    Let $f$ be $L$-smooth \eqref{ist:eq:L-matrix-smooth} and $f^{\inf}$-lower bounded. Then for any $x\in \bR^d$ we have:
    \begin{equation} \label{ist:eq:grad_f_lemma}
        \sqN{\nabla f(x)} \leq 2 L (f(x) - f^{\inf})
    \end{equation}
\end{lemma}

We also use the notion of the $AB$ inequality for a collection of (correlated) compressors $\cC_i$, introduced by \citet{szlendak2022permutation}.

\begin{definition}[$AB$ inequality]  \label{ist:def:AB_inequality}
A collection of random unbiased operators $\cC_1, \dots, \cC_n : \bR^d \to \bR^d$ satisfies the $AB$ inequality if there exist constants $A, B \geq 0$ such that
\begin{equation} \label{ist:eq:AB_inequality}
\begin{aligned}
    \E\sqN{ \frac{1}{n}\sum_{i=1}^n \cC_i(x_i) - \frac{1}{n}\sum_{i=1}^n x_i}  &\leq A \frac{1}{n}\sum_{i=1}^n \norm{x_i}^2 - B \sqN{\frac{1}{n}\sum_{i=1}^n x_i}
\end{aligned}
\end{equation}
for all $x_1, \dots, x_n \in \bR^d$. For brevity: $\{\cC_i\}_{i=1}^n \in \mathbb{C}(A,B)$.
\end{definition}
This definition is useful for algorithms such as IST because it extends the analysis to dependent compressors $\cC_1, \dots, \cC_n$, as occurs when the model parameters are randomly decomposed in a non-overlapping fashion. Moreover, the $AB$ inequality is satisfied for unbiased compressors \eqref{ist:def:unbiased_compressor}: $\cC_i \in \bU(\omega_i)$ for all $i$.

Denote $w_i^k \eqdef \cC_i^k (x^k)$.
If compressors $\cC_i^k$ satisfy the \emph{perfect reconstruction} property $\frac{1}{n} \sum_{i=1}^n \cC^k_i(x) = x$, which makes sense for the IST formulation, then method \eqref{ist:eq:IST} can be reformulated in the following way
\begin{eqnarray} \label{ist:eq:IST_general}
    x^{k+1} &=&
    \frac{1}{n} \sum_{i=1}^n \br{w_i^k - \gamma \cQ_i^k \br{\nabla f_i(w_i^k)}} \notag
    \\ &=&
    \frac{1}{n} \sum_{i=1}^n \cC_i^k (x^k) - \gamma \frac{1}{n} \sum_{i=1}^n \cQ_i^k \br{\nabla f_i(w_i^k)} \notag
    \\ &=&
    x^k - \gamma \underbrace{\frac{1}{n} \sum_{i=1}^n \cQ_i^k \br{\nabla f_i(\cC_i^k (x^k))}}_{g^k}.
\end{eqnarray}
This reformulation makes the algorithm amenable to analysis using Lemma \ref{ist:lem:descent_lemma_appendix}.

\subsection{Convergence analysis}

\begin{theorem}
Let $f$ and $f_i$ be $L$ and $L_i$-smooth \eqref{ist:eq:L-matrix-smooth} respectively. Collection of unbiased compressors $\cC_i^k, \cQ_i^k$ satisfy $AB$-inequality \eqref{ist:eq:AB_inequality} and property $\frac{1}{n} \sum_{i=1}^n \cC^k_i(x) = x$ for every $k$. Then for step size chosen as
$$\gamma \leq \min \left\{\frac{1}{L}, \frac{\sqrt{1 + \nicefrac{2}{(A L_m K)}} - 1}{2}\right\}$$
the iterations of Algorithm~\eqref{ist:eq:IST_general} for any $K \geq 1$ satisfy
\begin{equation} \label{ist:eq:general_res}
\begin{aligned}
    \min_{0\leq k \leq K-1} \sqN{\nabla f(x^k)}
    \leq
    \frac{6 \br{f(x^0) - \finf}}{\gamma K} + 2\br{1+(1+\beta^{-1})A} \max_k \sqn{x^k} \overline{L^2_\omega} %
    \\+ 4 A L_{\max} (1+\beta) \Delta,
\end{aligned}
\end{equation}
where
$\beta \leq \gamma, \overline{L^2_\omega} = \frac{1}{n} \sum_{i=1}^n L_i^2 \omega_i, L_{\max} = \max_i L_i$, and $\Delta = \frac{1}{n} \sum_{i=1}^n \br{\finf - f_i^{\inf}}$.
\end{theorem}
Note that the minimum over the squared gradient norm $\sqN{\nabla f(x^k)}$ and the maximum over the model weights $\sqn{x^k}$ in \eqref{ist:eq:general_res} can be replaced with weighted sums due to proof step \eqref{ist:eq:weighted_sum}.
Let us contrast the obtained convergence result with what we have for the quadratic case in \eqref{ist:eq:het_gen_grad_res}. The first term is basically the same and decreases as $1/K$ for constant step size $\gamma$. The other terms are different as they involve the norm of the iterates $\|x^k\|^2$ throughout training. However, they also cannot be decreased by diminishing the step size, which is similar to the quadratic case. Thus, we showed that for the more general case of smooth losses, IST can converge to the irreducible neighborhood of the stationary point. The obtained conclusions highlight the generalizability of our insights beyond the quadratic model.

Moreover, there is an additional part representing the heterogeneity of the distributed problem proportional to $\Delta$. In contrast to the term involving $\|x^k\|^2$, it can be eliminated in the homogeneous case when functions have a shared \say{minimizer} $f_i^{\inf} = \finf$, which may hold in the overparametrized regime.

\textbf{Contrasting to prior works.} \citet{khaled2019gradient} analyzed a similar method with unbiased compression in the single node and strongly convex setting. \citet{yuan2022distributed} extended their results to a finite-sum case with random sampling of one client. However, prior works suffer from a very strong condition on the sparsification variance $\omega \lesssim \nicefrac{\mu}{L}$, which is not the case for our analysis. Another important difference is that convergence bounds in \citet{khaled2019gradient} and \citet{yuan2022distributed} depend on the norm of the optimal solution $x^\star = \argmin f(x)$, which may not exist in the non-convex setting.

\begin{proof}
We start with the Descent Lemma \eqref{ist:eq:descent_lemma}

\begin{eqnarray} \label{ist:eq:descent_lemma_}
    f(x^{k+1}) &\leq&
    f(x^k) - \frac{\gamma}{2} \sqN{\nabla f(x^k)}
    - \br{\frac{1}{2\gamma} - \frac{L}{2}} \sqn{x^{k+1} - x^k}
    + \frac{\gamma}{2} \sqN{g^k - \nabla f(x^k)} \notag
    \\ &\leq&
    f(x^k) - \frac{\gamma}{2} \sqN{\nabla f(x^k)}
    + \frac{\gamma}{2} \sqN{g^k - \nabla f(x^k)},
\end{eqnarray}
where the step size is chosen as $\gamma \leq 1/L$.

Next prove an auxiliary result needed to work with the last term of \eqref{ist:eq:descent_lemma_}
\begin{eqnarray} \label{ist:eq:ES_distributed}
    \E \sqN{\frac{1}{n} \sum_{i=1}^n \nabla f_i(\cC_i^k (x^k)) - \nabla f_i(x^k)}
    &\overset{\text{Lemma~\ref{common:lem:average_squared_norm}}}{\leq}&
    \frac{1}{n} \sum_{i=1}^n \E \sqN{\nabla f_i(\cC_i^k (x^k)) - \nabla f_i(x^k)} \notag
    \\ &\overset{\eqref{ist:eq:smoothness}}{\leq}&
    \frac{1}{n} \sum_{i=1}^n L_i^2 \E \sqN{\cC_i^k(x^k) - x^k} \notag
    \\&\overset{\eqref{ist:eq:unbiased_compressor}}{\leq}&
    \frac{1}{n} \sum_{i=1}^n L_i^2 \omega_i \sqn{x^k} \notag
    \\ &=&
    \overline{L^2_\omega} \sqn{x^k}, %
\end{eqnarray}
for $\overline{L^2_\omega} \eqdef \frac{1}{n} \sum_{i=1}^n L_i^2 \omega_i$.

Recall the expression for $g^k = \frac{1}{n} \sum_{i=1}^n \cQ_i^k \br{\nabla f_i(w_i^k)}$ and $w_i^k \eqdef \cC_i^k (x^k)$. Now we can upper bound the last term of \eqref{ist:eq:descent_lemma_}
\begin{align*}
    \E \sqN{g^k - \nabla f(x^k)}
    &=
    \E \sqN{\frac{1}{n} \sum_{i=1}^n \cQ_i^k\br{\nabla f_i(w_i^k)} - \nabla f_i(x^k)} \\
    &\overset{\text{Lemma~\ref{common:lem:young}}}{\leq}
    2 \E \sqN{\frac{1}{n} \sum_{i=1}^n \cQ_i^k\br{\nabla f_i(w_i^k)} - \nabla f_i(w_i^k)}
    \\
    &\qquad
    +2 \E \sqN{\frac{1}{n} \sum_{i=1}^n \nabla f_i(w_i^k) - \nabla f_i(x^k)} \\
    &\overset{\eqref{ist:eq:AB_inequality}}{\leq}
    2 \sbr{A \frac{1}{n} \sum_{i=1}^n \E \sqN{ \nabla f_i(w_i^k)} - B \E \sqN{\frac{1}{n} \sum_{i=1}^n \nabla f_i(w_i^k)}}
    \\ &\qquad +
    2 \E \sqN{\frac{1}{n} \sum_{i=1}^n \nabla f_i(\cC_i^k (x^k)) - \nabla f_i(x^k)} \\
    &\overset{\eqref{ist:eq:ES_distributed}}{\leq}
    2 A \frac{1}{n} \sum_{i=1}^n
    \E \sqN{\nabla f_i(\cC_i^k (x^k)) \pm \nabla f_i(x^k)}
    \\
    &\qquad
    + 2 \frac{1}{n} \sum_{i=1}^n L_i^2 \omega_i \sqn{x^k}
    \\&\overset{\text{Lemma~\ref{common:lem:young}}}{\leq}
    2 A \frac{1}{n} \sum_{i=1}^n
    \sbr{\begin{aligned}[t]
        &\br{1 + \beta^{-1}} \E \sqN{\nabla f_i(\cC_i^k (x^k)) - \nabla f_i(x^k)}
        \\
        &+ \br{1 + \beta} \sqN{\nabla f_i(x^k)}
    \end{aligned}}
    \\
    &\qquad +
    2 \overline{L^2_\omega} \sqn{x^k}
    \\&\overset{\eqref{ist:eq:ES_distributed}}{\leq}
    2 A \frac{1}{n} \sum_{i=1}^n
    \sbr{\begin{aligned}[t]
        &\br{1 + \beta^{-1}} L_i^2 \omega_i \sqn{x^k}
        \\
        &+ \br{1 + \beta} \sqN{\nabla f_i(x^k)}
    \end{aligned}}
    \\
    &\qquad
    + 2 \overline{L^2_\omega} \sqn{x^k}
\end{align*}
\begin{align*}
    &\overset{\eqref{ist:eq:grad_f_lemma}}{\leq}
    2 \br{1 + A \br{1 + \beta^{-1}}} \overline{L^2_\omega} \sqn{x^k}
    \\
    &\qquad
    + 2 A \br{1 + \beta} \frac{1}{n} \sum_{i=1}^n 2 L_i
    \sbr{f_i(x^k) - f_i^{\inf}} \\
    &=
    2 \br{1 + A \br{1 + \beta^{-1}}} \overline{L^2_\omega} \sqn{x^k}
    \\
    &\qquad
    + 4 A \br{1 + \beta} \frac{1}{n} \sum_{i=1}^n L_i
    \sbr{f_i(x^k) - \finf - f_i^{\inf} + \finf} \\
    &\leq
    2 \br{1 + A \br{1 + \beta^{-1}}} \overline{L^2_\omega} \sqn{x^k}
    \\
    &\qquad
    + 4 A \br{1 + \beta} L_{\max} \frac{1}{n} \sum_{i=1}^n
    \sbr{f_i(x^k) - \finf + \finf - f_i^{\inf}} \\
    &=
    2 \br{1 + A \br{1 + \beta^{-1}}} \overline{L^2_\omega} \sqn{x^k}
    + 4 A \br{1 + \beta} L_m
    \\&\qquad\cdot
    \Big[f(x^k) - \finf
    + \underbrace{\frac{1}{n} \sum_{i=1}^n \br{\finf - f_i^{\inf}}}_{\Delta}\Big],
\end{align*}
where $L_{m} = L_{\max} = \max L_i$.

Combined with \eqref{ist:eq:descent_lemma_} %
and denoting $\delta^k \eqdef f(x^k) - \finf$ it leads to

\begin{equation}
\begin{aligned}
    \delta^{k+1}
    &\leq
    \delta^k - \frac{\gamma}{2} \sqN{\nabla f(x^k)} + 2 \gamma A \br{1 + \beta} L_m \br{f(x^k) - \finf} \\
    &\qquad +
    \frac{\gamma}{2} \sbr{2 \br{1 + A \br{1 + \beta^{-1}}} \overline{L^2_\omega} \sqn{x^k} + 4 A \br{1 + \beta} L_m \Delta} \\
    &=
    \br{1 + 2 \gamma A L_m \br{1 + \beta}} \delta^k - \frac{\gamma}{2} \sqN{\nabla f(x^k)}
    \\
    &\qquad +
    \frac{\gamma}{2} \underbrace{\sbr{2 \br{1 + A \br{1 + \beta^{-1}}} \overline{L^2_\omega} \sqn{x^k} + 4 A \br{1 + \beta} L_m \Delta}}_{C^k}.
\end{aligned}
\end{equation}

After rearranging we obtain
\begin{equation} \label{ist:eq:init_recursion}
\begin{aligned}
    \sqN{\nabla f(x^k)} \leq
    \frac{2}{\gamma} \big(1 + \underbrace{2 \gamma A \br{1 + \beta} L_m}_{D} \big) \delta^k - \frac{2}{\gamma} \delta^{k+1} + C^k.
\end{aligned}
\end{equation}

Next by following technique by \citet{stich2019unified} we introduce an exponentially decaying weighting sequence %
\begin{equation*}
    w^k = \frac{w^{k-1}}{1 + D} \leq \dots \leq w^{-1}.
\end{equation*}

Multiplying recursion \eqref{ist:eq:init_recursion} by $w^k$, we get
\begin{equation*}
\begin{aligned}
    w^k  \sqN{\nabla f(x^k)} & \leq
    \frac{2w^k\left(1 + D\right)}{\gamma} \delta^{k} - \frac{2w^k}{\gamma} \delta^{k+1} + w^k C^k \\ & =
    \frac{2 w^{k-1}}{\gamma} \delta^{k} - \frac{2 w^k}{\gamma} \delta^{k+1} + w^k C^k
\end{aligned}
\end{equation*}
After summing up both sides for $k$ from 0 to $K-1$, and telescoping terms
\begin{equation} \label{ist:eq:sum_recursion}
    \sum_{k=0}^{K-1} w^k \sqN{\nabla f(x^k)} \leq \frac{2 w^{-1}}{\gamma} \delta^{0} - \frac{2 w^{K-1}}{\gamma} \delta^{K} + \sum_{k=0}^{K-1} w^k C^k.
\end{equation}
Now define $W^K \eqdef \sum_{k=0}^{K-1} w^k$ and divide both sides of \eqref{ist:eq:sum_recursion} by $W^K$
\begin{equation} \label{ist:eq:weighted_sum}
    \min_{0\leq k \leq K-1} \sqN{\nabla f(x^k)} \leq
    \frac{1}{W^k} \sum_{k=0}^{K-1} w^k \sqN{\nabla f(x^k)} \leq \frac{2 w^{-1}}{W^k\gamma} \delta^{0} - \frac{2 w^{K-1}}{W^k\gamma} \delta^{K} + \frac{1}{W^k} \sum_{k=0}^{K-1} w^k C^k.
\end{equation}
By using the fact that
\begin{equation*}
    W^k = \sum_{k=0}^{K-1} w^k \geq \sum_{k=0}^{K-1} \min_{0 \leq i \leq K-1} w^i = K w^{K-1} = \frac{K w^{-1}}{(1 + D)^K},
\end{equation*}
we obtain
\begin{equation} \label{ist:eq:exponential_bound}
\begin{aligned}
    \min_k \sqN{\nabla f(x^k)}
    &\leq
    2\frac{\br{1 + 2 \gamma A \br{1 + \beta} L_m}^K}{\gamma K} \delta^0 + \max_k C^k.
\end{aligned}
\end{equation}
Next to simplify the obtained upper bound we use the fact that $1 + x \leq \exp (x)$
\begin{align*}
\left(1 + 2 \gamma A \br{1 + \beta} L_m\right)^{K}
&\leq\left(\exp \left(2 \gamma A \br{1 + \beta} L_m\right)\right)^{K} \\
&= \exp \left(2 \gamma A \br{1 + \beta} L_m K\right)
\leq \exp (1) \leq 3,
\end{align*}
where the second inequality for $\beta \leq \gamma$ holds if
$$\underbrace{2 A L_m K}_{A'_K} \gamma \br{1 + \gamma} \leq 1.$$
Then condition $A'_K \gamma + A'_K \gamma^2 -1 \leq 0$ holds for
\begin{equation*}
    \gamma \leq \frac{\sqrt{1 + 4/A'_K} - 1}{2} =
    \frac{\sqrt{1 + \nicefrac{2}{(A L_m K)}} - 1}{2}.
\end{equation*}

As a result \eqref{ist:eq:exponential_bound} leads to
\begin{equation*}
\begin{aligned}
    \min_k \sqN{\nabla f(x^k)}
    &\leq
    \frac{6 \br{f(x^0) - \finf}}{\gamma K} +
    2 \br{1 + A \br{1+\beta^{-1}}} \overline{L^2_\omega} \max_k \sqn{x^k}
    \\&\qquad
    + 4 A \br{1 + \beta} L_m \Delta,
\end{aligned}
\end{equation*}
where $\beta \leq \gamma$, $\Delta = \frac{1}{n} \sum_{i=1}^n \br{\finf - f_i^{\inf}}$, and $\overline{L^2_\omega} = \frac{1}{n} \sum_{i=1}^n L_i^2 \omega_i$.

\end{proof}

\section{Comparison to related work} \label{ist:sec:app_comparison}

\paragraph{Overview of theory provided in the original IST work \citep{yuan2022distributed}.}

The authors consider the following method
\begin{equation} \label{ist:eq:orig_IST_method}
    x^{k+1} = \cC (x^k) - \gamma \nabla f_{i_k} (\cC(x^k)),
\end{equation}

where $[\cC(x)]_i = x_i \cdot \mathcal{B}e (p)$\footnote{$\mathcal{B}_p (x) \eqdef \left\{\begin{array}{ll} x/p & \text { with probability } p \\ 0 & \text { with probability } 1-p\end{array}\right.$} is a Bernoulli sparsifier and $i_k$ is sampled uniformly at random from $[n]$.

The analysis in \citet{yuan2022distributed} relies on the assumptions

\begin{enumerate}
    \item $L_i$-smoothness of individual losses $f_i$;
    \item $Q$-Lipschitz continuity of $f$: $|f(x) - f(y)| \leq Q \norm{x - y}$;
    \item Error bound (or PŁ-condition): $\norm{\nabla f(x)} \geq \mu \norm{x^\star - x}$, where $x^\star$ is the global optimum;
    \item Stochastic gradient variance: $\Exp{\sqN{\nabla f_{i_k} (x)}} \leq M + M_f \sqN{\nabla f(x)}$;
    \item $\Exp{\nabla f_{i_k}(\cC(x^k)) \,|\, x^k} = \nabla f (x^k) + \eps, \quad \norm{\eps} \leq B$.
\end{enumerate}

Convergence result from Theorem 1 \citep{yuan2022distributed} for step size $\gamma=\nicefrac{1}{(2 L_{\max})}$:
\begin{equation}
    \min_{k \in \{1,\dots,K\}} \Exp{\sqN{\nabla f (x^k)}} \leq \frac{f(x^0) - f(x^\star)}{\alpha (K+1)} + \frac{1}{\alpha} \cdot \br{\frac{B Q}{2 L_{\max}} + \frac{5 L_{\max} \omega}{2} \sqn{x^\star} + \frac{M}{4 L_{\max}}},
\end{equation}
where $\alpha \eqdef \frac{1}{2 L_{\max}} \br{1 - \frac{M_f}{2}} - \frac{5 \omega L_{\max}}{2 \mu^2}$, $\omega \eqdef \frac{1}{p} - 1 < \frac{\mu^2}{10 L_{\max}^2}$, and $L_{\max} \eqdef \max_i L_i$.

If Lipschitzness and Assumption 5 are replaced with \emph{norm condition}:
\begin{equation} \label{ist:eq:norm_cond}
    \norm{\Exp{\nabla f_{i_k}(\cC(x^k)) \,|\, x^k} - \nabla f(x^k)} \leq \theta \norm{\nabla f(x^k)}
\end{equation}
they obtain the following (for step size $\gamma=\nicefrac{1}{2 L_{\max}}$)
\begin{equation}
    \min_{k \in \{1,\dots,K\}} \Exp{\sqN{\nabla f (x^k)}} \leq \frac{f(x^0) - f(x^\star)}{\alpha (K+1)} + \frac{1}{\alpha} \cdot \br{\frac{5 L_{\max} \omega}{2} \sqn{x^\star} + \frac{M}{4 L_{\max}}},
\end{equation}
where $\alpha = \frac{1}{2 L_{\max}} \br{\frac12 - \theta - \frac{M_f}{2}} - \frac{5 \omega L_{\max}}{2 \mu^2}$ and $\omega = \frac{1}{p} - 1 < \frac{\mu^2}{5 L_{\max}^2 \br{\frac12 - \theta - \frac{M_f}{2}}}$.

\begin{remark}
The original method \eqref{ist:eq:orig_IST_method} does not incorporate gradient sparsification, which can create a significant disparity between theory and practice. This is because the gradient computed at the compressed model, denoted by $\nabla f(\cC(x))$, is not guaranteed to be sparse and representative of the submodel computations. This modification of the method also significantly simplifies the theoretical analysis, as using a single sketch (instead of $\mC\mL\mC$) allows for an unbiased gradient estimator.

Through our analysis of the IST gradient estimator in Equation \eqref{ist:eq:hom_gen_estimator}, we discover that conditions---such as Assumption 5 and Inequality \eqref{ist:eq:norm_cond}---are not satisfied, even in the homogeneous setting for a simple quadratic problem. Furthermore, it is evident that such conditions are also not met for logistic loss. At the same time, in general, it is expected that insightful theory for general \text{(non-)convex} functions should yield appropriate results for quadratic problems. Additionally, it remains unclear whether the norm condition \eqref{ist:eq:norm_cond} is satisfied in practical scenarios. The situation is not straightforward---even for quadratic problems---as we show in the expression for $\sigma^2$ in Equation \eqref{ist:eq:hetero_bound}.
\end{remark}

\paragraph{Masked Training \citep{mohtashami2022masked}.}

The authors consider the following \say{Partial SGD} method

\begin{equation} \label{ist:eq:partial_sgd}
\begin{aligned}
   \hat{x}^k &= x^k + \delta x^k= x^k - (\mathrm{1} - p) \odot x^k\\
   x^{k+1} &= x^k - \gamma p \odot \nabla f(\hat{x}^k, \xi^k),
\end{aligned}
\end{equation}

where $\nabla f(x, \xi)$ is an unbiased stochastic gradient estimator of a $L$-smooth loss function $f$, $\odot$ is an element-wise product, and $p$ is a binary sparsification mask.

\citet{mohtashami2022masked} make the following \say{bounded perturbation} assumption
\begin{equation} \label{ist:eq:bound_perturb}
    \max_{k} \frac{\norm{\delta x^k}}{\max \left\{\norm{p^k \odot \nabla f(x^k)}, \norm{p^k \odot \nabla f(\hat{x}^k)}\right\}} \leq \frac{1}{2L}.
\end{equation}

This inequality may not hold for a simple convex case. Consider a function $f(x) = \frac{1}{2} x^\top A x$, for
\begin{equation}
    A = \begin{pmatrix}
        a & 0 \\
        0 & c
        \end{pmatrix},
    \qquad
    x^0 = \begin{pmatrix}
        x_1 \\
        x_2
        \end{pmatrix},
    \qquad
    p^0 = \begin{pmatrix}
        0 \\
        1
        \end{pmatrix}.
\end{equation}

Then condition \eqref{ist:eq:bound_perturb} (at iteration $k=0$) will be equivalent to
\begin{equation*}
    \frac{x_1}{c x_2} \leq \frac{1}{2a} \Leftrightarrow 2 \leq \frac{2a}{c} \leq \frac{x_2}{x_1},
\end{equation*}
which clearly does not hold for an arbitrary initialization $x^0$.

In addition, the convergence bound in Theorem 1 \citep{mohtashami2022masked} suggests choosing the step size as $\gamma_0 \alpha^k$, where
\begin{equation}
    \alpha^k = \min \left\{1, \frac{\left\langle p^k \odot \nabla f(x^k), p^k \odot \nabla f(\hat{x}^k)\right\rangle}{\left\|p^k \odot \nabla f(\hat{x}^k)\right\|^2}\right\}
\end{equation}
is not guaranteed to be positive to the inner product $\left\langle p^k \odot \nabla f(x^k), p^k \odot \nabla f(\hat{x}^k)\right\rangle$, which may lead to non-convergence of the method.

\paragraph{Optimization with access to auxiliary information framework \citep{chayti2022optimization}}
suggests modeling training with compressed models via performing gradient steps with respect to function $h(x) \eqdef \ExpSub{\mathcal{M}}{f(1_{\mathcal{M}} \odot x)}$. This function allows access to a sparse version of the original model $f(x)$. They impose the following bounded Hessian dissimilarity assumption on $h$ and $f$

\begin{equation}
    \left\|\nabla^2 f(x) - \ExpSub{\mathcal{M}}{\mD_{\mathcal{M}} \nabla^{2} f(1_{\mathcal{M}} \odot x) \mD_{\mathcal{M}}}\right\|_{2} \leq \delta,
\end{equation}
where $1_{\mathcal{M}}$ and $\mD_{\mathcal{M}} = \mathrm{Diag}(1_{\mathcal{M}})$ refer to a binary vector and matrix sparsification masks.

This approach relies on variance-reduction and requires gradient computations on the full model $x$, and thus it is not suitable for our problem setting.

\paragraph{Comparison to the work of \citet{liao2022on}.}
Next, we try our best to briefly and accurately represent some of the previous work’s findings and comment on the differences.
The authors provide a \textit{high probability} convergence analysis of a \say{Single Hidden-Layer Neural Network with ReLU activations} based on the Neural Tangent Kernel (NTK) framework. The network’s first layer weights are initialized based on $\mathcal{N} (0, \kappa^2 \mathbf{I})$ and the weight vector of the second layer is initialized uniformly at random from $\{-1, 1\}$.
In contrast, we do not make any assumptions on the initialized parameters $x$ (in our notation).

The second difference concerns the assumptions on the data. \citet{liao2022on} assume that for every data point $(a_j, y_j)$, it holds that $||a_j||^2 = 1$ and $|y_j| \leq C-1$ for some constant $C \geq 1$. Moreover, for any $j \neq l$, the points $a_j$ and $a_l$ are not co-aligned, i.e., $a_j \neq \xi a_l$ for any $\xi \in \mathbb{R}$. In contrast, we do not make any assumptions about the data apart from those on the matrices $\mathbf{L}_i$.
In addition, analysis by \citet{liao2022on} assumes that the number of hidden nodes is greater than a certain quantity and that NN’s weights distance from initialization is uniformly bounded.

\citet{liao2022on} consider a regression (MSE) loss function, a special case of quadratic loss and full gradients computation. They provide guarantees for IST under a \say{simplified assumption that every worker has full data access}, which corresponds to the homogeneous setting in our terminology.

\paragraph{Comparison of IST and 3D Parallelism \citep{shoeybi2019megatron}.}

IST and 3D parallelism were introduced independently and concurrently in 2019. While sharing some conceptual similarities (combination of data and model parallelism), they have a different focus. The key distinction is the way how model parallelism is implemented. Namely, 3D parallelism combines \textit{Pipeline} and \textit{Tensor} parallelism.

Pipeline parallelism suggests splitting the model’s layers across computing nodes which requires additional transmission of tensors for every forward and backward step. Tensor parallelism breaks the layers (MLP blocks and attention heads) into parts, which creates a need for additional synchronization between GPUs. Thus, 3D parallelism makes computation on nodes dependent on mutual communications, unlike IST.

Independent Subnetwork Training decomposes the model into smaller subnets independently trained in parallel. IST does not need synchronization during local updates and requires the transmission of fewer parameters, which decreases per-step communication costs.

3D parallelism introduces significant communication overhead, which greatly increases computational cost \citep{bian2023does}. The problem described is especially relevant for public compute clouds (such as Amazon EC2), which often suffer from slow interconnects. At the same time, IST is most beneficial for such a setup as it improves communication efficiency by design. Moreover, 3D parallelism is incompatible with a standard federated learning setting. At the same time, the IST-like approach is a viable technique, as every network is independent and can be trained on resource-constrained devices \citep{dun2023efficient}. In summary, both IST and 3D parallelism are viable approaches with pros and cons and are best suited for different scenarios.

\endgroup

\chapter{Supplementary Material for Chapter 3 (MAST)}
\label{appendix:chapter3_mast}

\begingroup
\input{KAUST_Thesis/imported/03_mast/import_setup}

\section{Sampling and sketching facts}
We use the generic inner-product, norm, and smoothness facts collected in Appendix~\ref{appendix:common_facts}. The remaining facts in this section are specific to sampling, sketching, or scalar step-size conditions used in this chapter.

\begin{lemma}[Lemma 1 from \citet{mishchenko2020random}]
\label{mast:lemma:RR}
     Let $X_1, \ldots, X_n \in \mathbb{R}^d$ be fixed vectors, $\overline{X} \stackrel{\text { def }}{=} \frac{1}{n} \sum_{i=1}^n X_i$ be their average. Fix any $k \in\{1, \ldots, n\}$, let $X_{\pi_1}, \ldots X_{\pi_k}$ be sampled uniformly without replacement from $\left\{X_1, \ldots, X_n\right\}$ and $\overline{X}_\pi$ be their average. Then, the sample average and variance are given by
$$
\mathbb{E}\left[\overline{X}_\pi\right]=\overline{X}
$$
$$
\mathbb{E}\left[\left\|\overline{X}_\pi-\overline{X}\right\|^2\right]=\frac{n-k}{k(n-1)} \frac{1}{n} \sum_{i=1}^n\left\|X_i-\overline{X}\right\|^2
$$
\end{lemma}

\begin{lemma} (Lemma 5 from \citet{EF21}).
\label{mast:lemma:peter}
Let $a, b>0$. If $0 \leq \gamma \leq \frac{1}{\sqrt{a}+b}$, then $a \gamma^2+b \gamma \leq 1$. The bound is tight up to the factor of 2 since $\frac{1}{\sqrt{a}+b} \leq \min \left\{\frac{1}{\sqrt{a}}, \frac{1}{b}\right\} \leq \frac{2}{\sqrt{a}+b}$.
\end{lemma}

\begin{proposition}\label{mast:proposition:eigenvalues}
   Nonzero eigenvalues of $\;\mS\mS^{\top}$ and $\;\mS^{\top}\mS$ coincide.
\end{proposition}
\begin{proof}
    Indeed, suppose $\lambda\neq 0$ is an eigenvalue of $\mS^{\top}\mS$ with an eigenvector $v\in\mathbb{R}^d,$ then $\lambda$ is an eigenvalue of $\mS\mS^{\top}$ with an eigenvector $\mS v.$
\end{proof}

\section{\texorpdfstring{Auxiliary facts about functions $\tf(x)$ and $f_{\mS}(x)$}{Auxiliary facts about functions f\textunderscore D(x) and f\textunderscore S(x)}}

For a differentiable function $f:\mathbb{R}^d\to\mathbb{R}$ and $x,y\in\mathbb{R}^d$, the Bregman divergence associated with $f$ is $D_f(x,y)\eqdef f(x)-f(y)-\langle\nabla f(y),x-y\rangle.$
\begin{lemma}[Bregman divergence]
If $f$ is continuously differentiable, then $D_{\tf}(x,y) = \Exp{D_{f_{\mS}}(x,y)}.$
\end{lemma}
\begin{proof}
    Since $f$ is continuously differentiable, we can interchange integration and differentiation. The result follows from the linearity of expectation.
\end{proof}

\subsection{\texorpdfstring{Consequences of $L_f$-smoothness}{Consequences of L\textunderscore f-smoothness}}
Recall the $L_f$-smoothness assumption.\\
\noindent\textbf{Assumption 2.} Function $f$ is differentiable and $L_f$-\textbf{smooth}, i.e., there is $L_f>0$ such that $\forall x, h \in \R^d$
	\[f(x+h) \leq f(x) + \ve{\nabla f(x)}{h} + \frac{L_f}{2}\norm{h}^2.\]
We also require $f$ to be lower bounded by $f^{\inf} \in \mathbb{R}$.
\begin{lemma}[Consequences of $L_f$-smoothness]\label{mast:lemma:smoothness_appendix} If $f$ is $L_f$-smooth, then
	\begin{itemize}
		\item[(i)] $f_{\mS}$ is $L_{f_\mS}$-smooth with $L_{f_\mS} \leq L_{\mS} L_f.$ That is,
		\[ f_{\mS}(x+h) \leq f_{\mS}(x) + \ve{\nabla f_{\mS}(x)}{h} + \frac{L_{\mS} L_f }{2} \norm{h}^2, \quad \forall x, h \in \R^{d}. \]

		\item[(ii)]
		$\tf$ is $L_{\tf}$-smooth with $L_{\tf}\leq L_{\mathcal{D}} L_f.$ That is,
		\[\tf(x+h) \leq \tf(x) + \ve{\nabla \tf(x)}{h} + \frac{L_{\mathcal{D}} L_f }{2} \norm{h}^2, \quad \forall x, h \in \R^{d}. \]

		\item[(iii)]
		\begin{equation}\label{mast:eq:L_f-smooth-iii_appendix}
			\tf(x) \leq f(x) + \frac{(L_{\mathcal{D}}-1) L_f}{2} \norm{x-\shift}^2,\quad \forall x\in \R^d.
		\end{equation}

	\end{itemize}

\end{lemma}

\begin{proof}
	\begin{itemize}
		\item[(i)]  For any $x, h\in \R^d$, we have
		\begin{align*}
			f_{\mS}(x+h)
			&= f(\shift + \mS (x + h -\shift))  \\
			&= f(\shift +\mS (x  - \shift) + \mS h)  \\
			&\stackrel{\mathrm{Asn.}~\ref{mast:ass:l_f_smooth}}{\leq}
			f( \shift +\mS (x  - \shift) )
			+ \ve{\nabla f(\shift +\mS ( x  -\shift ) )}{\mS h}
			+ \frac{L_f}{2}\norm{\mS h}^2 \\
			&=  f( \shift + \mS (x  -\shift) )
			+ \ve{\mS^\top \nabla f(\shift +\mS (x  - \shift) )}{h}
			+ \frac{L_f}{2} \ve{ \mS^\top \mS h}{h}  \\
			&\leq f_{\mS}(x) + \ve{\nabla f_{\mS}(x)}{h} +  \frac{L_{\mS} L_f }{2} \norm{h}^2.
		\end{align*}

		\item[(ii)] For any $x, h\in \R^d$, we have
		\begin{align*}
			\tf(x+h)
			&= \Exp{ f(\shift +\mS (x + h - \shift)) } \\
			&= \Exp{ f(\shift +\mS (x  -\shift) + \mS h) } \\
			&\stackrel{\mathrm{Asn.}~\ref{mast:ass:l_f_smooth}}{\leq}
			\Exp{ f(\shift +\mS (x  -\shift) )
			+ \ve{\nabla f(\shift +\mS (x  -\shift) )}{\mS h}
			+ \frac{L_f}{2}\norm{\mS h}^2} \\
			&= \Exp{ f(\shift +\mS (x - \shift) ) }
			+ \ve{\Exp{\mS^\top \nabla f(\shift +\mS (x  - \shift) )}}{h}\\
                &\qquad+\frac{L_f}{2}\Exp{\norm{\mS h}^2} \\
			&= \tf(x) + \ve{\nabla \tf(x)}{h} + \frac{L_f}{2} \ve{\Exp{ \mS^\top \mS} h}{h} \\
			&\leq \tf(x) + \ve{\nabla \tf(x)}{h} + \frac{L_{\mathcal{D}} L_f }{2} \norm{h}^2.
		\end{align*}

		\item[(iii)]
		For any $x\in \R^d$, we have
		\begin{align*}
			\tf(x)
			&= \Exp{ f(\shift +\mS (x-\shift)) }\\
			&\stackrel{\mathrm{Asn.}~\ref{mast:ass:l_f_smooth}}{\leq}
			\Exp{ f(x) + \ve{ \nabla f(x) }{  \mS( x - \shift) - (x-\shift) }
			+ \frac{L_f}{2}\norm{\mS (x-\shift) - (x-\shift)}^2 } \\
			&= f(x) +  \ve{ \nabla f(x) }{ \Exp{ \mS (x-\shift) - (x-\shift) } }\\
                &\qquad+\frac{L_f}{2} \Exp{\norm{\mS (x-\shift) - (x-\shift)}^2} \\
			&\overset{\eqref{mast:eq:sketch_def}}{=}
			f(x) +  \ve{ \nabla f(x) }{ 0 }
			+ \frac{L_f}{2} (x-\shift)^\top \left(\Exp{\mS^\top \mS} -\mI \right) (x-\shift) \\
			&= f(x) + \frac{ (L_{\mathcal{D}}-1) L_f}{2} \norm{x-\shift}^2.
		\end{align*}

	\end{itemize}
\end{proof}
\subsection{Consequences of convexity}

We do not assume differentiability of $f$ here. Recall that function $f$ is convex if, for all $x,y\in\mathbb{R}^d$ and $\alpha\in[0,1],$ we have that $f(\alpha x + (1-\alpha)y) \leq \alpha f(x) + (1-\alpha)f(y).$
\begin{lemma}\label{mast:lem:convex-only_appendix} If $f$ is convex, then $\tf$ is convex and $\tf(x)\geq f(x)$ for all $x\in \R^d$.
\end{lemma}
\begin{proof}

	\begin{itemize}
		\item[(i)] Let $x, y\in \R^d$ and $\alpha \in [0,1]$. Then
		\begin{eqnarray*} \tf(\alpha x + (1-\alpha)y) &\overset{\eqref{mast:eq:pretrained_compressed_problem}}{=}& \Exp{ f(\shift + \mS (\alpha x + (1-\alpha)y - \shift)) } \\
			&=&\Exp{ f( \alpha \left( \shift + \mS (x-\shift) \right) + (1-\alpha) \left( \shift + \mS (y - \shift)\right) ) } \\
			&\leq &\Exp{ \alpha f( \shift + \mS (x-\shift)) + (1-\alpha) f( \shift + \mS (y - \shift)) } \\
			&= & \alpha \Exp{ f(\shift +  \mS (x-\shift))} + (1-\alpha) \Exp{ f( \shift + \mS (y - \shift)) } \\
			&\overset{\eqref{mast:eq:pretrained_compressed_problem}}{=} & \alpha \tf(x) + (1-\alpha) \tf(y).
		\end{eqnarray*}
  Alternative proof: Each $f_{\mS}$ is obviously convex, and expectation of convex functions is a convex function.
		\item[(ii)]
		Fix $x\in \R^d$ and let $g \in \partial f(x)$ be a subgradient of $f$ at $x$. Then
		\begin{eqnarray*} \tf(x) &\overset{\eqref{mast:eq:pretrained_compressed_problem}}{=}& \Exp{ f( \shift +\mS (x-\shift)) }\\
			&\geq & \Exp{ f(x) + \ve{ g }{  \mS (x-\shift) - (x-\shift) } } \\
			& = & f(x) +  \ve{ g }{ \Exp{ \mS (x-\shift) - (x-\shift) } } \\
			& \overset{\eqref{mast:eq:sketch_def}}{=} & f(x) +  \ve{ g }{ 0 }  \\
			& = & f(x).
		\end{eqnarray*}
    Alternative proof: Using Jensen's inequality, $\tf(x) = \Exp{f(\shift + \mS(x-\shift))} \geq f(\Exp{\shift + \mS(x-\shift)}) = f(x)$.
	\end{itemize}
\end{proof}
\subsection{\texorpdfstring{Consequences of $\mu_f$-convexity}{Consequences of mu\textunderscore f-convexity}}
Recall the $\mu_f$-strong convexity (or, for simplicity, $\mu_f$-convexity) assumption.\\
\noindent\textbf{Assumption~3.} Function $f$ is differentiable and $\mu_f$-\textbf{strongly convex}, i.e., there is $\mu_f>0$ such that $\forall x, h \in \R^d$
    \[f(x+h) \geq f(x) + \ve{\nabla f(x)}{h} + \frac{\mu_f}{2}\norm{h}^2.\]

\begin{lemma}[Consequences of $\mu_f$-convexity]\label{mast:lemma:consequences_strongly_convex_appendix} If $f$ is $\mu_f$-convex, then
	\begin{itemize}
		\item[(i)] $f_{\mS}$ is $\mu_{f_\mS}$-convex with $\mu_{f_\mS} \geq \mu_{\mS} \mu_f.$ That is,
		\[ f_{\mS}(x+h) \geq f_{\mS}(x) + \ve{\nabla f_{\mS}(x)}{h} + \frac{\mu_{\mS} \mu_f }{2} \norm{h}^2, \quad \forall x, h \in \R^{d}. \]

		\item[(ii)]
		$\tf$ is $\mu_{\tf}$-convex with $\mu_{\tf} \geq \mu_{\mathcal{D}} \mu_f.$ That is,
		\[\tf(x+h) \geq \tf(x) + \ve{\nabla \tf(x)}{h} + \frac{\mu_{\mathcal{D}} \mu_f }{2} \norm{h}^2, \quad \forall x, h \in \R^{d}. \]

		\item[(iii)]
		\begin{equation}\label{mast:eq:mu_f-convex-iii_appendix}\tf(x) \geq f(x) + \frac{ (\mu_{\mathcal{D}}-1) \mu_f}{2} \norm{x-\shift}^2,\quad \forall x\in \R^d.\end{equation}
	\end{itemize}

\end{lemma}

\begin{proof}
	\begin{itemize}
		\item[(i)]  For any $x, h\in \R^d$, we have
		\begin{align*}
			f_{\mS}(x+h)
			&= f(\shift +\mS (x + h -\shift))  \\
			&= f(\shift +\mS (x  -\shift) + \mS h)  \\
			&\stackrel{\mathrm{Asn.}~\ref{mast:ass:mu_f-convex}}{\geq}
			f(\shift +\mS (x  -\shift) )
			+ \ve{\nabla f(\shift +\mS (x  -\shift) )}{\mS h}
			+ \frac{\mu_f}{2}\norm{\mS h}^2 \\
			&= f(\shift +\mS (x  -\shift) )
			+ \ve{\mS^\top \nabla f(\shift +\mS (x  -\shift) )}{h}
			+ \frac{\mu_f}{2} \ve{ \mS^\top \mS h}{h}  \\
			&\geq f_{\mS}(x) + \ve{\nabla f_{\mS}(x)}{h} +  \frac{\mu_{\mS} \mu_f }{2} \norm{h}^2.
		\end{align*}

		\item[(ii)] For any $x, h\in \R^d$, we have
			\begin{align*}
				\tf(x+h)
				&= \Exp{ f(\shift +\mS (x + h -\shift)) } \\
				&= \Exp{ f(\shift +\mS (x  -\shift) + \mS h) } \\
				&\stackrel{\mathrm{Asn.}~\ref{mast:ass:mu_f-convex}}{\geq}
				\Exp{ f(\shift +\mS (x  -\shift) )
				+ \ve{\nabla f(\shift +\mS (x  -\shift) )}{\mS h}
				+ \frac{\mu_f}{2}\norm{\mS h}^2} \\
				&= \Exp{ f(\shift +\mS (x  -\shift) )}
				+ \ve{\Exp{\mS^\top \nabla f(\shift +\mS (x  -\shift) )}}{h}\\
                &\qquad+ \frac{\mu_f}{2}\Exp{\norm{\mS h}^2} \\
				&= \tf(x) + \ve{\nabla \tf(x)}{h}
				+ \frac{\mu_f}{2} \ve{\Exp{ \mS^\top \mS} h}{h} \\
				&\geq \tf(x) + \ve{\nabla \tf(x)}{h} + \frac{\mu_{\mathcal{D}} \mu_f }{2} \norm{h}^2.
			\end{align*}

		\item[(iii)]
		For any $x\in \R^d$, we have
			\begin{align*}
				\tf(x)
				&= \Exp{ f(\shift +\mS (x-\shift)) }\\
				&\stackrel{\mathrm{Asn.}~\ref{mast:ass:mu_f-convex}}{\geq}
				\Exp{ f(x) + \ve{ \nabla f(x) }{  \mS( x - \shift) - (x-\shift) }
				+ \frac{\mu_f}{2}\norm{\mS (x-\shift) - (x-\shift)}^2 } \\
				&= f(x) +  \ve{ \nabla f(x) }{ \Exp{ \mS (x-\shift) - (x-\shift) } }\\
                &\qquad+ \frac{\mu_f}{2} \Exp{\norm{\mS (x-\shift) - (x-\shift)}^2} \\
				&\overset{\eqref{mast:eq:sketch_def}}{=}
				f(x) +  \ve{ \nabla f(x) }{ 0 }
				+ \frac{\mu_f}{2} (x-\shift)^\top \left(\Exp{\mS^\top \mS} -\mI \right) (x-\shift) \\
				&= f(x) + \frac{ (\mu_{\mathcal{D}}-1) \mu_f}{2} \norm{x-\shift}^2.
			\end{align*}
	\end{itemize}
\end{proof}

\section{\texorpdfstring{Relation between minima of $f$ and $\tf$}{Relation between minima of f and f\textunderscore D}}
\begin{theorem} \label{mast:thm:approx_appendix} Let Assumptions~\ref{mast:ass:l_f_smooth}~and~\ref{mast:ass:mu_f-convex} hold, and let $x_\cD^\star\in \tilde{\cX}$ and $x^\star \in \cX^\star$. Then
\begin{equation*}
    f(x^\star) \leq f(x_\cD^\star) \leq f(x^\star)  + \frac{(L_{\mathcal{D}}-1)L_f}{2}\norm{x^\star-\shift}^2 - \frac{(\mu_{\mathcal{D}}-1)\mu_f}{2}\norm{x_\cD^\star-\shift}^2,
\end{equation*}
\begin{equation*}
    f(x^\star) + \frac{(\mu_{\mathcal{D}}-1)\mu_f}{2}\norm{x_\cD^\star-\shift}^2 \leq \tf(x_\cD^\star) \leq f(x^\star)  + \frac{(L_{\mathcal{D}}-1)L_f}{2}\norm{x^\star-\shift}^2.
\end{equation*}
\end{theorem}
\begin{proof} To obtain the result, combine inequalities \eqref{mast:eq:L_f-smooth-iii_appendix} and \eqref{mast:eq:mu_f-convex-iii_appendix}:
\begin{equation*}
\begin{split}
    f(x_\cD^\star) + \frac{(\mu_{\mathcal{D}}-1)\mu_f}{2}\norm{x_\cD^\star-\shift}^2 \overset{\eqref{mast:eq:mu_f-convex-iii_appendix}}{\leq}   \tf(x_\cD^\star) \leq \tf(x^\star)  \overset{\eqref{mast:eq:L_f-smooth-iii_appendix}}{\leq}   f(x^\star) + \frac{(L_{\mathcal{D}}-1)L_f}{2}\norm{x^\star-\shift}^2.
\end{split}
\end{equation*}
\end{proof}

\begin{theorem}\label{mast:thm:approx_noncvx_appendix}
     Let Assumption~\ref{mast:ass:l_f_smooth} hold, and let $x_\cD^\star\in \cX^\star_{\cD}$ and $x^\star \in \cX^\star$. Then
\begin{equation*}
    f(x^\star) \leq \tf(x_\cD^\star) \leq f(x^\star)  + \frac{(L_{\mathcal{D}}-1)L_f}{2}\norm{x^\star-\shift}^2.
\end{equation*}
\end{theorem}
\begin{proof}
    To obtain the result, use inequality \eqref{mast:eq:L_f-smooth-iii_appendix}. Also, note that, since for every $\mS\sim \mathcal{D},$ we have $f_{\mS}(x) = f\left(\shift + \mS\left(x-\shift\right)\right)\geq f(x^\star),$ for all $x\in\mathbb{R}^d,$ we can conclude that $\tf(x_\cD^\star) = \Exp{f_{\mS}(x_\cD^\star)}\geq \Exp{f(x^\star)}= f(x^{\star}).$
\end{proof}

\subsection{Consequences of Lipschitz continuity of the gradient}
The gradient of $f(x)$ is $L_f$-Lipschitz if, for all $x,y\in\mathbb{R}^d$ we have that $\left\|\nabla f(x) - \nabla f(y)\right\|\leq L_f\left\|x-y\right\|.$
\begin{lemma}  If $\nabla f$ is $L_f$-Lipschitz, then $\nabla \tf$ is $L_{\tf}$-Lipschitz with
	$$L_{\tf} \leq L_f \Exp{ \norm{\mS^\top} \norm{\mS}}. $$
\end{lemma}
\begin{proof}
	We have that
	\begin{eqnarray*} \norm{\nabla \tf(x) - \nabla \tf(y)}  &=& \norm{ \nabla \Exp{  f( \shift +\mS (x-\shift) )} - \nabla \Exp{ f(\shift +\mS (y - \shift))}} \\
		&=& \norm{\Exp{ \mS^\top \nabla f(\shift + \mS (x - \shift))} - \Exp{ \mS^\top \nabla f(\shift + \mS (y-\shift))}} \\
		&= &\norm{\Exp{ \mS^\top \nabla f(\shift +\mS (x-\shift)) -  \mS^\top \nabla f(\shift +\mS (y - \shift)) }} \\
		&\leq &\Exp{ \norm{  \mS^\top \nabla f(\shift +\mS (x-\shift)) -  \mS^\top \nabla f(\shift +\mS (y - \shift)) } } \\
		&\leq &\Exp{ \norm{\mS^\top} \norm{   \nabla f(\shift +\mS (x-\shift)) -  \nabla f(\shift +\mS (y - \shift)) } } \\
		&\leq &\Exp{ \norm{\mS^\top} L_f \norm{  \mS x - \mS y } } \\
		&\leq &L_f \Exp{ \norm{\mS^\top} \norm{\mS}}\norm{x-y}.
	\end{eqnarray*}
\end{proof}

\section{Double sketched $\mathsf{GD}$}
Recall that $L_{\mS}^{\max} = \sup_{\mS}\left\lbrace\lambda_{\max}\left(\mS^{\top}\mS\right)\right\rbrace=\sup_{\mS}\left\lbrace\lambda_{\max}\left(\mS\mS^{\top}\right)\right\rbrace$ (we used Proposition~\ref{mast:proposition:eigenvalues}).
\subsection{Nonconvex analysis: proof of theorem \ref{mast:thm:sgd_nonconvex}}
The following lemma restates Lemma~\ref{mast:lemma:ac} from Chapter~\ref{chapter3:mast}.
\begin{lemma}\label{mast:lemma:ac_appendix}
	For all $x\in\mathbb{R}^d,$ we have that
	\begin{equation*}
		\begin{split}
			\Exp{\left\|\nabla f_{\mS}(x)\right\|^2} & \leq 2L_fL_{\mS}^{\max} \left(\tf(x) - f^{\inf}\right).\\
		\end{split}
	\end{equation*}
	where the expectation is taken with respect to $\mS.$
\end{lemma}
\begin{proof}
	Due to $L_f$-smoothness of $f,$ we have that
	\begin{equation*}
	\begin{split}
	\Exp{\left\|\nabla f_{\mS}(x)\right\|^2}
    & = \Exp{\left\|\mS^\top \nabla f(y)|_{y=\shift+\mS(x - \shift)}\right\|^2} \\
    & = \Exp{\lin{\mS^\top \nabla f(y), \mS^\top \nabla f(y)}|_{y=\shift+\mS(x - \shift)}} \\
    & = \Exp{\lin{\mS \mS^{\top} \nabla f(y), \nabla f(y)}|_{y=\shift+\mS(x - \shift)}} \\
    & \leq  \Exp{\lambda_{\max}\left(\mS \mS^{\top}\right) \left\|\nabla f(y)|_{y=\shift+\mS(x - \shift)}\right\|^2} \\
    & \leq L_{\mS}^{\max} \Exp{\left\|\nabla f(y)|_{y=\shift+\mS(x - \shift)}\right\|^2} \\
	& \stackrel{\text{Lemma~\ref{common:lem:smooth_lower_gradient}}}{\leq} 2L_fL_{\mS}^{\max} \Exp{f\left(\shift+\mS(x-\shift)\right) - f^{\inf}} \\
	& = 2L_fL_{\mS}^{\max} \left(\tf(x) - f^{\inf}\right). \\
	\end{split}
	\end{equation*}
\end{proof}
All convergence results in the nonconvex scenarios rely on the following key lemma:
\begin{lemma} \label{mast:lemma:noncvx_weighted}
 The iterates $\{x^t\}_{t\geq 0}$ of $\mathsf{SGD}$ satisfy
	\begin{equation}\label{mast:eq:initial_recursion}
		\begin{split}
			\gamma r^t \leq \left(1 + \gamma^2 M_1\right)\delta^t - \delta^{t+1} + \gamma^2 M_2,
		\end{split}
	\end{equation}
	where $M_1$ and $M_2$ are non-negative constants, $\delta^t\eqdef\mathbb{E}\left[\tf(x^t)-\tf^{\inf}\right]$ and $r^t\eqdef\mathbb{E}\left[\left\|\nabla \tf(x^t)\right\|^2\right].$ Fix $w_{-1}>0$ and, for all $t\geq 0,$ define $w_t=\frac{w_{t-1}}{1+\gamma^2M_1}.$ Then, for any $T\geq 1,$ the iterates $\{x^t\}_{t\geq 0}$ satisfy
	\begin{equation*}
		\sum_{t=0}^{T-1}w_tr^t \leq \frac{w_1}{\gamma}\delta^0 - \frac{w_{T-1}}{\gamma}\delta^T + \gamma M_2\sum_{t=0}^{T-1}w_t.
	\end{equation*}
\end{lemma}
\begin{proof}
	Multiplying both sides of~\eqref{mast:eq:initial_recursion} by $\frac{w_t}{\gamma},$ we obtain
	\begin{equation*}
		w_t r^t\leq\frac{w_{t-1}}{\gamma}\delta^t - \frac{w_t}{\gamma}\delta^{t+1} + \gamma w_tM_2.
	\end{equation*}
	For every $0\leq t\leq T-1,$ sum these inequalities. We arrive at
	\begin{equation*}
		\sum_{t=0}^{T-1}w_tr^t\leq \frac{w_{-1}}{\gamma}\delta^0 - \frac{w_{T-1}}{\gamma}\delta^{T} + \gamma M_2\sum_{t=0}^{T-1}w_t.
	\end{equation*}
\end{proof}
Recall that $D \eqdef L_f \sqrt{L_{\mathcal{D}}L_{\mS}^{\max} }.$
\begin{theorem}
	Let Assumptions~\ref{mast:ass:C}~and~\ref{mast:ass:l_f_smooth} hold. For every $t\geq 0,$ put $\delta^t\eqdef\mathbb{E}\left[\tf(x^t)-\tf^{\inf}\right]$ and $r^t\eqdef\mathbb{E}\left[\left\|\nabla \tf(x^t)\right\|^2\right].$ Then, for any $T\geq 1,$ the iterates $\{x^t\}_{t=0}^{T-1}$ of Algorithm~\ref{mast:alg:SGD} satisfy
	\begin{equation}\label{mast:eq:main_noncvx_convergence_basicsgd}
		\min_{0 \leq t < T}r^t\leq \frac{\left(1+D^2\gamma^2\right)^T}{\gamma T}\delta^0 + D^2  \gamma \left(\tf^{\inf} - f^{\inf}\right).
	\end{equation}
\end{theorem}
\begin{proof}
	Due to $L_f$-smoothness of $f,$ we have that
	\begin{equation*}
		\begin{split}
			f(s+\mS^{t+1}(x^{t+1} - s)) &\leq f(s+\mS^{t+1}(x^{t} - s))\\
			& + \bigg{<} \nabla f\left(y\right)|_{y=s+\mS^{t+1}(x^t-s)}, \mS^{t+1}\left(x^{t+1} - s\right) - \mS^{t+1}\left(x^t-s\right) \bigg{>}\\
			& + \frac{L_f}{2}\left\|\mS^{t+1}\left(x^{t+1} - s\right) - \mS^{t+1}\left(x^t-s\right) \right\|^2\\
			& = f(s+\mS^{t+1}(x^{t} - s)) -\gamma \bigg{<} \nabla f\left(y\right)|_{y=s+\mS^{t+1}(x^t-s)}, \mS^{t+1}\nabla f_{\mS^{t}}(x^t) \bigg{>}\\
			& + \frac{L_f\gamma^2}{2}\left\|\mS^{t+1}\nabla f_{\mS^{t}}(x^t) \right\|^2\\
			& \leq f(s+\mS^{t+1}(x^{t} - s))-\gamma \bigg{<} \nabla f_{\mS^{t+1}}\left(x^t\right), \nabla f_{\mS^{t}}(x^t) \bigg{>}\\
			& + \frac{L_f\gamma^2}{2}\bigg{<} \left(\mS^{t+1}\right)^{\top}\mS^{t+1}\nabla f_{\mS^{t}}(x^t), \nabla f_{\mS^{t}}(x^t) \bigg{>}.\\
		\end{split}
	\end{equation*}
	Taking the expectation with respect to $\mS^{t+1}$ yields
	\begin{equation*}
		\begin{split}
			\tf\left(x^{t+1}\right) &\leq \tf\left(x^{t}\right) - \gamma\bigg{<} \nabla \tf\left(x^t\right), \nabla f_{\mS^{t}}(x^t) \bigg{>}\\
            & + \frac{L_f\gamma^2}{2}\bigg{<} \Exp{\left(\mS^{t+1}\right)^{\top}\mS^{t+1}}\nabla f_{\mS^{t}}(x^t), \nabla f_{\mS^{t}}(x^t) \bigg{>}\\
			&\leq \tf\left(x^{t}\right) - \gamma\bigg{<} \nabla \tf\left(x^t\right), \nabla f_{\mS^{t}}(x^t) \bigg{>} + \frac{L_fL_{\mathcal{D}}\gamma^2}{2}\left\|\nabla f_{\mS^{t}}(x^t)\right\|^2.
		\end{split}
	\end{equation*}
	Conditioned on $x^t,$ take expectation with respect to $\mS^{t}:$
	\begin{equation*}
		\begin{split}
			\Exp{\tf\left(x^{t+1}\right)|x^t} \leq \tf\left(x^{t}\right) - \gamma \left\|\nabla \tf\left(x^t\right)\right\|^2 + \frac{L_fL_{\mathcal{D}}\gamma^2}{2}\Exp{\left\|\nabla f_{\mS^{t}}(x^t) \right\|^2}.\\
		\end{split}
	\end{equation*}
	From Lemma~\ref{mast:lemma:ac}, we obtain that
	\begin{equation*}
		\begin{split}
			\Exp{\tf\left(x^{t+1}\right)|x^t} & \leq \tf\left(x^{t}\right) - \gamma \left\|\nabla \tf\left(x^t\right)\right\|^2 \\
            & \qquad + \frac{L_fL_{\mathcal{D}}\gamma^2}{2} \left(2L_fL_{\mS}^{\max} \left(\tf(x) - \tf^{\inf}\right)\right.
            + \left. 2L_fL_{\mS}^{\max} \left(\tf^{\inf} - f^{\inf}\right)\right).
		\end{split}
	\end{equation*}
	Subtract $\tf^{\inf}$ from both sides, take expectations on both sides, and use the tower property:
	\begin{equation*}
		\begin{split}
			\Exp{\tf\left(x^{t+1}\right) - \tf^{\inf}} & \leq \left(1+D^2\gamma^2\right)\Exp{\tf\left(x^{t}\right) - \tf^{\inf}} - \gamma \Exp{\left\|\nabla \tf\left(x^t\right)\right\|^2}\\
			&+ D^2\gamma^2\left(\tf^{\inf} - f^{\inf}\right).\\
		\end{split}
	\end{equation*}
	We obtain that
	\begin{equation*}
		\begin{split}
			\gamma r^t\leq \left(1+D^2\gamma^2\right)\delta^t- \delta^{t+1} + D\gamma^2\left(\tf^{\inf} - f^{\inf}\right).
		\end{split}
	\end{equation*}
	Notice that the iterates $\{x^t\}_{t\geq 0}$ of Algorithm~\ref{mast:alg:SGD} satisfy condition~\eqref{mast:eq:initial_recursion} of Lemma~\ref{mast:lemma:noncvx_weighted} with $M_1=D^2,$
	and
	$M_2=D^2\left(\tf^{\inf} - f^{\inf}\right).$ Therefore, we can conclude that, for any $T\geq 1,$ the iterates $\{x^t\}_{t=0}^{T-1}$ of Algorithm~\ref{mast:alg:SGD} satisfy
	\begin{equation*}
		\sum_{t=0}^{T-1}w_tr^t\leq \frac{w_{-1}}{\gamma}\delta^0 - \frac{w_{T-1}}{\gamma}\delta^{T} + D^2\gamma \left(\tf^{\inf} - f^{\inf}\right)\sum_{t=0}^{T-1}w_t.
	\end{equation*}
	Divide both sides by $\sum_{t=0}^{T-1}w_t.$ From $\sum_{t=0}^{T-1}w_t\geq Tw_{T-1} = \frac{Tw_{-1}}{1+D^2\gamma^2}$, we can conclude that
	\begin{equation*}
		\min_{0 \leq t < T}r^t\leq \frac{\left(1+D^2\gamma^2\right)^T}{\gamma T}\delta^0 + D^2\gamma \left(\tf^{\inf} - f^{\inf}\right).
	\end{equation*}
\end{proof}
\begin{corollary}
	Fix $\varepsilon>0.$ Choose the stepsize $\gamma>0$ as
	\begin{equation*}
		\gamma = \min\left\lbrace \frac{1}{D\sqrt{ T}}, \frac{\varepsilon^2}{2D^2\left(\tf^{\inf} - f^{\inf}\right)} \right\rbrace.
	\end{equation*}
	Then, provided that
	\begin{equation*}
		T\geq \frac{12\delta^0D^2}{\varepsilon^4}\max\left\lbrace 3\delta^0, \tf^{\inf} - f^{\inf}\right\rbrace,
	\end{equation*}
	we have
	\begin{equation*}
		\min_{0 \leq t < T}\Exp{\left\|\nabla \tf(x^t)\right\|^2}\leq\varepsilon^2.
	\end{equation*}
\end{corollary}
\begin{proof}
	Since $\gamma \leq \frac{\varepsilon^2}{2D^2\left(\tf^{\inf} - f^{\inf}\right)},$ we obtain
	\begin{equation*}
		D^{2} \gamma \left(\tf^{\inf} - f^{\inf}\right) \leq \frac{\varepsilon^2}{2}.
	\end{equation*}

	Since $\gamma\leq \frac{1}{D\sqrt{T}},$
	\begin{equation*}
		\left(1+D^2\gamma^2\right)^T\leq \exp\left(TD^2\gamma^2\right)\leq\exp(1)\leq 3.
	\end{equation*}
	If $\gamma = \frac{1}{D\sqrt{T}},$ then, since
	\begin{equation*}
		T\geq \frac{36\left(\delta^0\right)^2D^2}{\varepsilon^4},
	\end{equation*}
	we have $\frac{3\delta^0}{\gamma T}\leq \frac{\varepsilon^2}{2}.$ Further, if $\gamma= \frac{\varepsilon^2}{2D^2\left(\tf^{\inf} - f^{\inf}\right)},$ then, since
	\begin{equation*}
		T\geq \frac{12\delta^0D^2\left(\tf^{\inf} - f^{\inf}\right)}{\varepsilon^4},
	\end{equation*}
	we have $\frac{3\delta^0}{\gamma T}\leq \frac{\varepsilon^2}{2}.$
	Combining it with $\eqref{mast:eq:main_noncvx_convergence_basicsgd},$ we arrive at $\min_{0 \leq t < T}\Exp{\left\|\nabla \tf(x^t)\right\|^2}\leq\varepsilon^2.$
\end{proof}

\subsection{Strongly convex analysis: proof of theorem \ref{mast:thm:str_conv}}
\begin{theorem}
	Let Assumptions~\ref{mast:ass:C},~\ref{mast:ass:l_f_smooth},~and~\ref{mast:ass:mu_f-convex} hold. Let $r^t\eqdef x^t - x_\cD^\star,$ $t\geq 0.$ Choose a stepsize $0<\gamma \leq \frac{1}{L_fL_{\mS}^{\max}}.$ Then the iterates $\{x^t\}_{t\geq 0}$ of Algorithm~\ref{mast:alg:SGD} satisfy
    \begin{equation}\label{mast:eq:main_strcvx_basicsgd}
	\begin{split}
		\Exp{\left\|r^{t+1}\right\|^2} & \leq \left(1-\gamma\mu_{\mathcal{D}}\mu_f\right)\Exp{\left\|r^{t}\right\|^2} + 2\gamma^2L_fL_{\mS}^{\max}\left(\tf^{\inf} - f^{\inf}\right).\\
	\end{split}
    \end{equation}
\end{theorem}
\begin{proof}
Let $r^t\eqdef x^t - x_\cD^\star.$ We get
\begin{equation*}
	\begin{split}
		\left\|r^{t+1}\right\|^2 & = \left\|\left(x^t-\gamma \nabla f_{\mS^{t}}(x^t) \right) - x_\cD^\star\right\|^2  = \left\|x^t-x_\cD^\star - \gamma \nabla f_{\mS^{t}}(x^t)\right\|^2 \\
		& = \left\|r^{t}\right\|^2 - 2\gamma\big{<}r^t, \nabla f_{\mS^{t}}(x^t)\big{>} + \gamma^2\left\|\nabla f_{\mS^{t}}(x^t)\right\|^2.
	\end{split}
\end{equation*}
Now we compute expectation of both sides of the inequality, conditional on $x^t:$
\begin{equation*}
	\mathbb{E}\left[\left\|r^{t+1}\right\|^2|x^t\right] = \left\|r^{t}\right\|^2 - 2\gamma\big{<}r^t, \mathbb{E}[\nabla f_{\mS^{t}}(x^t)|x^t]\big{>} + \gamma^2\mathbb{E}\left[\left\|\nabla f_{\mS^{t}}(x^t)\right\|^2|x^t\right].
\end{equation*}
Taking the expectation with respect to $\mS_t,$ using the fact that $f$ is continuously differentiable and using Lemma~\ref{mast:lemma:ac}, we obtain that
\begin{equation*}
	\begin{split}
		\mathbb{E}\left[\left\|r^{t+1}\right\|^2\right] & \leq \left\|r^{t}\right\|^2 - 2\gamma\big{<}r^t, \nabla \tf(x^t)\big{>} + 2\gamma^2L_fL_{\mS}^{\max}\left(\left(\tf(x^t) - \tf^{\inf}\right)+  \left(\tf^{\inf} - f^{\inf}\right)\right).\\
	\end{split}
\end{equation*}
Since $\tf$ is $\mu_{\mathcal{D}}\mu_f$-convex, we conclude that $\big{<}r^t, \nabla \tf(x^t)\big{>}\geq \tf(x^t) - \tf^{\inf} + \frac{\mu_{\mathcal{D}}\mu_f}{2}\left\|r^t\right\|^2.$ Therefore,
\begin{equation*}
	\begin{split}
		\Exp{\left\|r^{t+1}\right\|^2} & \leq \left(1-\gamma\mu_{\mathcal{D}}\mu_f\right)\left\|r^{t}\right\|^2 - 2\gamma\left(\tf(x^t) - \tf^{\inf}\right)\left(1-\gamma L_fL_{\mS}^{\max} \right)\\
		&+ 2\gamma^2L_fL_{\mS}^{\max}\left(\tf^{\inf} - f^{\inf}\right).\\
	\end{split}
\end{equation*}
Since $\gamma \leq \frac{1}{L_fL_{\mS}^{\max}},$ taking expectation and using the tower property we get
\begin{equation*}
	\begin{split}
		\Exp{\left\|r^{t+1}\right\|^2} & \leq \left(1-\gamma\mu_{\mathcal{D}}\mu_f\right)\Exp{\left\|r^{t}\right\|^2} + 2\gamma^2L_fL_{\mS}^{\max}\left(\tf^{\inf} - f^{\inf}\right).\\
	\end{split}
\end{equation*}
Unrolling the recurrence, we get
\begin{equation*}
	\Exp{\left\|r^{t}\right\|^2} \leq \left(1-\gamma\mu_{\mathcal{D}}\mu_f\right)^t\left\|r^0\right\|^2 + \frac{2\gamma L_fL_{\mS}^{\max}\left(\tf^{\inf} - f^{\inf}\right)}{\mu_{\mathcal{D}}\mu_f}.
\end{equation*}
\end{proof}
\begin{corollary}
	Fix $\delta > 0.$ Choose the stepsize $\gamma>0$ as
	\begin{equation*}
		\gamma = \min\left\lbrace \frac{1}{L_fL_{\mS}^{\max}}, \frac{\mu_{\mathcal{D}}\mu_f\delta\left\|r^0\right\|^2}{2L_fL_{\mS}^{\max}\left(\tf^{\inf} - f^{\inf}\right)}\right\rbrace.
	\end{equation*}
	Then, provided that
	\begin{equation*}
		t \geq \frac{L_fL_{\mS}^{\max}}{\mu_{\mathcal{D}}\mu_f}\left\lbrace 1, \frac{2\left(\tf^{\inf} - f^{\inf}\right)}{\mu_{\mathcal{D}}\mu_f\delta\left\|r^0\right\|^2}\right\rbrace \log\frac{1}{\delta},
	\end{equation*}
	we have $\Exp{\left\|r^{t}\right\|^2} \leq 2\delta\left\|r^0\right\|^2.$
\end{corollary}
\begin{proof}
	Since $\gamma\leq \frac{\mu_{\mathcal{D}}\mu_f\delta\left\|r^0\right\|^2}{2L_fL_{\mS}^{\max}\left(\tf^{\inf} - f^{\inf}\right)},$
	we have that
	\begin{equation*}
		\frac{2\gamma L_fL_{\mS}^{\max}\left(\tf^{\inf} - f^{\inf}\right)}{\mu_{\mathcal{D}}\mu_f} \leq\delta\left\|r^0\right\|^2.
	\end{equation*}
	If $\gamma= \frac{\mu_{\mathcal{D}}\mu_f\delta\left\|r^0\right\|^2}{2L_fL_{\mS}^{\max}\left(\tf^{\inf} - f^{\inf}\right)},$ then, since
	\begin{equation*}
		t\geq \frac{2L_fL_{\mS}^{\max}\left(\tf^{\inf} - f^{\inf}\right)}{\mu_{\mathcal{D}}^2\mu_f^2\delta\left\|r^0\right\|^2}\log\frac{1}{\delta},
	\end{equation*}
	we obtain that
	\begin{equation*}
		\left(1-\gamma\mu_{\mathcal{D}}\mu_f\right)^t\leq\exp\left(-\gamma\mu_{\mathcal{D}}\mu_ft\right)\leq\delta.
	\end{equation*}
	Further, if $\gamma=\frac{1}{L_fL_{\mS}^{\max}},$ then, since
	\begin{equation*}
		t\geq \frac{L_fL_{\mS}^{\max}}{\mu_{\mathcal{D}}\mu_f}\log\frac{1}{\delta},
	\end{equation*}
	we obtain that
	\begin{equation*}
		\left(1-\gamma\mu_{\mathcal{D}}\mu_f\right)^t\leq\exp\left(-\gamma\mu_{\mathcal{D}}\mu_ft\right)\leq\delta.
	\end{equation*}
	Thus, combining it with~\eqref{mast:eq:main_strcvx_basicsgd}, we arrive at $\Exp{\left\|r^{t}\right\|^2} \leq 2\delta\left\|r^0\right\|^2.$
\end{proof}
\subsection{Convex analysis}
\begin{assumption}\label{mast:ass:minimizers_exist}
	A set $\tilde{\cX}=\{x_\cD^\star\;|\;\tf\left(x_\cD^\star\right)\leq\tf\left(x\right)\;\forall x\in\mathbb{R}^d\}$ is nonempty.
\end{assumption}
\begin{theorem}
	Let $r^t\eqdef x^t - x_\cD^\star.$ Let Assumptions~\ref{mast:ass:C},~\ref{mast:ass:l_f_smooth}~and~\ref{mast:ass:minimizers_exist} hold. Let $f$ be convex. Choose a stepsize $0<\gamma\leq\frac{1}{2L_fL_{\mS}^{\max}}.$ Fix $T\geq 1$ and let $\bar{x}^T$ be chosen uniformly from the iterates $x^0,\ldots,x^{T-1}$ of Algorithm~\ref{mast:alg:SGD}. Then
	\begin{equation}\label{mast:eq:main_convergence_cvx_basicsgd}
		\begin{split}
			\Exp{\tf(\bar{x}^t) - \tf^{\inf}} &\leq \frac{\norm{r^0}^2}{\gamma T} + 2\gamma L_fL_{\mS}^{\max}\left(\tf^{\inf} - f^{\inf}\right),\\
		\end{split}
	\end{equation}
	where $r^t\eqdef x^t - x_\cD^\star,$ $t\in\{0,\ldots,T-1\}.$
\end{theorem}
\begin{proof}
Let us start by analyzing the behavior of $\norm{x^t - x_\cD^\star}^2$. By developing the squares, we obtain
\begin{align*}
	\norm{x^{t+1} - x_\cD^\star}^2 &= \norm{x^t - x_\cD^\star}^2 - 2\gamma\langle \nabla f_{\mS}(x^t), x^t - x_\cD^\star\rangle + \gamma^2\norm{\nabla f_{\mS}(x^t)}^2
\end{align*}
Hence, after taking the expectation with respect to $\mS$ conditioned on $x^t$, we can use the convexity of $f$ and Lemma~\ref{mast:lemma:ac} to obtain:
\begin{equation*}
	\begin{split}
		\Exp{\norm{x^{t+1} - x_\cD^\star}^2|x^t} & = \norm{x^t - x_\cD^\star}^2 + 2\gamma\langle \nabla \tf(x^t), x_\cD^\star - x^t\rangle + \gamma^2\Exp{\norm{\nabla f_{\mS}(x^t)}^2}\\
		&\leq \norm{x^t - x_\cD^\star}^2 + 2\gamma\left(\gamma L_fL_{\mS}^{\max}  - 1\right)\left(\tf(x^t) - \tf^{\inf}\right)\\
		& + 2\gamma^2L_fL_{\mS}^{\max}\left(\tf^{\inf} - f^{\inf}\right).
	\end{split}
\end{equation*}
Rearranging, taking expectation and taking into account the condition on the stepsize, we have
\begin{equation*}
	\begin{split}
		\gamma\Exp{\tf(x^t) - \tf^{\inf}} &\leq \Exp{\norm{r^t}^2} - \Exp{\norm{r^{t+1}}^2}
		+ 2\gamma^2L_fL_{\mS}^{\max}\left(\tf^{\inf} - f^{\inf}\right).\\
	\end{split}
\end{equation*}
Summing over $t = 0,\ldots, T-1$ and using telescopic cancellation gives:
\begin{equation*}
	\begin{split}
		\gamma\sum_{t=0}^{T-1}\left(\Exp{\tf(x^t) - \tf^{\inf}}\right) &\leq \norm{r^0}^2 - \Exp{\norm{r^{T}}^2}
		+ 2T\gamma^2L_fL_{\mS}^{\max}\left(\tf^{\inf} - f^{\inf}\right).\\
	\end{split}
\end{equation*}
Since $\Exp{\norm{r^{T}}^2} \geq 0$, dividing both sides by $\gamma T$ gives:
\begin{equation*}
	\begin{split}
		\frac{1}{T}\sum_{t=0}^{T-1}\left(\Exp{\tf(x^t) - \tf^{\inf}}\right) &\leq \frac{\norm{r^0}^2}{\gamma T} + 2\gamma L_fL_{\mS}^{\max}\left(\tf^{\inf} - f^{\inf}\right).\\
	\end{split}
\end{equation*}
We treat the $\left(\frac{1}{T},\ldots,\frac{1}{T}\right)$ as if it is a probability vector.
Indeed, using that $\tf$ is convex together with Jensen's inequality gives
\begin{equation*}
	\begin{split}
		\Exp{\tf(\bar{x}^t) - \tf^{\inf}} &\leq \frac{\norm{r^0}^2}{\gamma T} + 2\gamma L_fL_{\mS}^{\max}\left(\tf^{\inf} - f^{\inf}\right).\\
	\end{split}
\end{equation*}
\end{proof}
\begin{corollary}
	Fix $\delta>0.$ Choose the stepsize $\gamma>0$ as
	\begin{equation*}
		\gamma = \min\left\lbrace \frac{1}{2L_f\lambda_m^{\mS}}, \frac{\delta\norm{r^0}^2}{2L_f\lambda_m^{\mS}\left(\tf^{\inf} - f^{\inf}\right)} \right\rbrace.
	\end{equation*}
	Then, provided that
	\begin{equation*}
		T\geq \frac{2L_f\lambda_m^{\mS}}{\delta}\max\left\lbrace 1, \frac{\tf^{\inf} - f^{\inf}}{\delta\norm{r^0}^2}\right\rbrace,
	\end{equation*}
	we have $\Exp{\tf(\bar{x}^t) - \tf^{\inf}}\leq 2\delta\norm{r^0}^2.$
\end{corollary}
\begin{proof}
	Since $\gamma\leq\frac{\delta\norm{r^0}^2}{2L_f\lambda_m^{\mS}\left(\tf^{\inf} - f^{\inf}\right)},$ we have that
	\begin{equation*}
		2\gamma L_fL_{\mS}^{\max}\left(\tf^{\inf} - f^{\inf}\right)\leq\delta\norm{r^0}^2.
	\end{equation*}
	If $\gamma = \frac{\delta\norm{r^0}^2}{2L_f\lambda_m^{\mS}\left(\tf^{\inf} - f^{\inf}\right)},$ then, since
	\begin{equation*}
		T \geq \frac{2L_f\lambda_m^{\mS}\left(\tf^{\inf} - f^{\inf}\right)}{\delta^2\norm{r^0}^2},
	\end{equation*}
	we obtain that $\frac{\norm{r^0}^2}{\gamma T}\leq \delta\norm{r^0}^2.$ Further, if $\gamma = \frac{1}{2L_f\lambda_m^{\mS}},$ then, since $T\geq \frac{2L_f\lambda_m^{\mS}}{\delta},$ we have $\frac{\norm{r^0}^2}{\gamma T}\leq\delta\norm{r^0}^2.$ Thus, combining it with~\eqref{mast:eq:main_convergence_cvx_basicsgd}, we arrive at $\Exp{\tf(\bar{x}^t) - \tf^{\inf}}\leq 2\delta\left\|r^0\right\|^2.$
\end{proof}

\section{(Stochastic) inexact gradient}
\subsection{Nonconvex analysis: proof of theorem \ref{mast:thm:abc_ncvx}}
We solve the problem~\eqref{mast:eq:pretrained_compressed_problem} with the method
\begin{equation}\label{mast:eq:abc_sgd_appendix}
	x^{t+1} = x^t - \gamma g^t,
\end{equation}
where $g^t := g(x^t)$ is the gradient estimator that satisfies
\begin{equation}\label{mast:eq:unbiased_abc_appendix}
	\Exp{g(x)} = \nabla f_{\mS}(x), \quad\forall x \in\mathbb{R}^d,
\end{equation}
\begin{equation}\label{mast:eq:abc_appendix}
	\Exp{\left\|g(x)\right\|^2}\leq 2A\left(f_{\mS}(x) - f^{\inf}_{\mS}\right) + B\left\|\nabla f_{\mS}(x)\right\|^2 + C, \quad\forall x \in\mathbb{R}^d.
\end{equation}
Recall that $D_{A,B}=A+BL_fL_{\mS}^{\max}.$
\begin{theorem}
	Let Assumptions~\ref{mast:ass:C},~\ref{mast:ass:l_f_smooth},~\ref{mast:eq:unbiased_abc}~and~\ref{mast:eq:abc} hold. For every $t\geq 0,$ put $\delta^t\eqdef\mathbb{E}\left[\tf(x^t)-\tf^{\inf}\right]$ and $r^t\eqdef\mathbb{E}\left[\left\|\nabla \tf(x^t)\right\|^2\right].$ Then, for any $T\geq 1,$ the iterates $\{x^t\}_{t=0}^{T-1}$ of Algorithm~\ref{mast:eq:abc_sgd} satisfy
	\begin{equation}\label{mast:eq:main_noncvx_abc}
		\begin{split}
			\min_{0 \leq t < T}r^t &\leq \frac{\left(1+D_{A,B}L_fL_{\mathcal{D}}\gamma^2\right)^T}{\gamma T}\delta^0
			+ \frac{L_fL_{\mathcal{D}}\gamma}{2}\left(2\left(\tf^{\inf} - 	f^{\inf}\right)D_{A,B} + C\right).
		\end{split}
	\end{equation}
\end{theorem}
\begin{proof}
	Due to $L_f$-smoothness of $f,$ we have that
	\begin{equation*}
		\begin{split}
			f(s+\mS^{t+1}(x^{t+1} - s)) &\leq f(s+\mS^{t+1}(x^{t} - s))\\
			& + \bigg{<} \nabla f\left(y\right)|_{y=s+\mS^{t+1}(x^t-s)}, \mS^{t+1}\left(x^{t+1} - s\right) - \mS^{t+1}\left(x^t-s\right) \bigg{>}\\
			& + \frac{L_f}{2}\left\|\mS^{t+1}\left(x^{t+1} - s\right) - \mS^{t+1}\left(x^t-s\right) \right\|^2\\
			& = f(s+\mS^{t+1}(x^{t} - s)) -\gamma \bigg{<} \nabla f\left(y\right)|_{y=s+\mS^{t+1}(x^t-s)}, \mS^{t+1}g^t \bigg{>}\\
                & + \frac{L_f\gamma^2}{2}\left\|\mS^{t+1}g^t \right\|^2\\
			& \leq f(s+\mS^{t+1}(x^{t} - s))-\gamma \bigg{<} \nabla f_{\mS^{t+1}}\left(x^t\right), g^t \bigg{>}\\
                & + \frac{L_f\gamma^2}{2}\bigg{<} \left(\mS^{t+1}\right)^{\top}\mS^{t+1}g^t, g^t \bigg{>}.\\
		\end{split}
	\end{equation*}
	Taking the expectation with respect to $\mS^{t+1},$ we obtain that
	\begin{equation*}
		\begin{split}
			\tf(x^{t+1}) & \leq \tf(x^t)-\gamma  \bigg{<} \nabla\tf(x^t), g^t \bigg{>} + \frac{L_f\gamma^2}{2}\bigg{<} \Exp{\left(\mS^{t+1}\right)^{\top}\mS^{t+1}}g^t, g^t \bigg{>}\\
			& \leq \tf(x^t)-\gamma  \bigg{<} \nabla\tf(x^t), g^t \bigg{>} + \frac{L_fL_{\mathcal{D}}\gamma^2}{2}\left\|g^t\right\|^2.\\
		\end{split}
	\end{equation*}
	Taking the expectation with respect to $\mS^{t},$ conditional on $x^t,$ using Lemma~\ref{mast:lemma:ac} and \eqref{mast:eq:abc} we have that
	\begin{equation*}
		\begin{split}
			\Exp{\tf(x^{t+1})|x^t} & \leq \tf(x^t)-\gamma  \bigg{<} \nabla\tf(x^t), \Exp{g^t|x^t} \bigg{>} + \frac{L_fL\gamma^2}{2}\Exp{\left\|g^t\right\|^2|x^t}\\
			& = \tf(x^t)-\gamma  \left\|\nabla\tf(x^t)\right\|^2\\
                & + \frac{L_fL_{\mathcal{D}}\gamma^2}{2}\Exp{2A\left(f_{\mS_t}(x^t) - f^{\inf}_{\mS_t}\right) + B\left\|\nabla f_{\mS_t}(x^t)\right\|^2 + C}\\
			& \leq \tf(x^t)-\gamma  \left\|\nabla\tf(x^t)\right\|^2 + AL_fL_{\mathcal{D}}\gamma^2\left(\tf(x^t) - \tf^{\inf}\right)\\
			& + \frac{L_fL_{\mathcal{D}}B\gamma^2}{2}\left(2L_fL_{\mS}^{\max} \left(\tf(x^t) - \tf^{\inf}\right)+ 2L_fL_{\mS}^{\max} \left(\tf^{\inf} - f^{\inf}\right)\right)\\
			& + \frac{L_fL_{\mathcal{D}}\gamma^2C}{2} + AL_fL_{\mathcal{D}}\gamma^2\left(\tf^{\inf} - \Exp{f_{\mS^{t}}^{\inf}}\right).\\
		\end{split}
	\end{equation*}
	Substitute $\tf^{\inf}$ from both sides, take expectation on both sides and use the tower property:
	\begin{equation*}
		\begin{split}
			\Exp{\tf(x^{t+1}) - \tf^{\inf}} & \leq \Exp{\tf(x^{t}) - \tf^{\inf}}\left(1+D_{A,B}L_fL_{\mathcal{D}}\gamma^2\right) - \gamma  \Exp{\left\|\nabla\tf(x^t)\right\|^2} \\
			& + \frac{L_fL_{\mathcal{D}}\gamma^2}{2}\left(2\left(\tf^{\inf} - f^{\inf}\right)D_{A,B} + C\right).
		\end{split}
	\end{equation*}
	We obtain that
	\begin{equation*}
		\begin{split}
			\gamma r^t &\leq \left(1+D_{A,B}L_fL_{\mathcal{D}}\gamma^2\right)\delta^t- \delta^{t+1}
			+ \frac{L_fL_{\mathcal{D}}\gamma^2}{2}\left(2\left(\tf^{\inf} - f^{\inf}\right)D_{A,B} + C\right).
		\end{split}
	\end{equation*}
	Notice that the iterates $\{x^t\}_{t\geq 0}$ of Algorithm~\ref{mast:eq:abc_sgd} satisfy condition~\eqref{mast:eq:initial_recursion} of Lemma~\ref{mast:lemma:noncvx_weighted} with $M_1=D_{A,B}L_fL_{\mathcal{D}},$
	$$
	M_2=\frac{L_fL_{\mathcal{D}}}{2}\left(2\left(\tf^{\inf} - f^{\inf}\right)D_{A,B} + C\right)
	.
	$$
	Therefore, for any $T\geq 1,$ the iterates $\{x^t\}_{t=0}^{T-1}$ of Algorithm~\ref{mast:eq:abc_sgd} satisfy
	\begin{equation*}
		\begin{split}
			\sum_{t=0}^{T-1}w_tr^t & \leq \frac{w_{-1}}{\gamma}\delta^0 - \frac{w_{T-1}}{\gamma}\delta^{T}\\
			& + \frac{L_fL_{\mathcal{D}}\gamma}{2}\left(2A\left(\tf^{\inf} - \Exp{f_{\mS^{t}}^{\inf}}\right)+2BL_fL_{\mS}^{\max} \left(\tf^{\inf} - f^{\inf}\right) + C\right)\sum_{t=0}^{T-1}w_t.
		\end{split}
	\end{equation*}
	Divide both sides by $\sum_{t=0}^{T-1}w_t.$ From $\sum_{t=0}^{T-1}w_t\geq Tw_{T-1} = \frac{Tw_{-1}}{1+D_{A,B}L_fL_{\mathcal{D}}\gamma^2}$ we can conclude that
	\begin{equation*}
		\begin{split}
			\min_{0 \leq t < T}r^t &\leq \frac{\left(1+D_{A,B}L_fL_{\mathcal{D}}\gamma^2\right)^T}{\gamma T}\delta^0
			+ \frac{L_fL_{\mathcal{D}}\gamma}{2}\left(2\left(\tf^{\inf} - 	f^{\inf}\right)D_{A,B} + C\right).
		\end{split}
	\end{equation*}
\end{proof}
\begin{corollary}
	Fix $\varepsilon>0.$ Choose the stepsize $\gamma>0$ as
	\begin{equation*}
		\begin{split}
			\gamma &= \min\left\lbrace \frac{1}{\sqrt{L_fL_{\mathcal{D}}D_{A,B} T}}, \frac{\varepsilon^2}{L_fL_{\mathcal{D}}\left(2\left(\tf^{\inf} - 	f^{\inf}\right)D_{A,B} + C\right)} \right\rbrace.
		\end{split}
	\end{equation*}
	Then, provided that
	\begin{equation*}
		\begin{split}
			T & \geq \frac{6\delta^0L_fL_{\mathcal{D}}}{\varepsilon^4}\max\left\lbrace 6\delta^0D_{A,B},\;\;2\left(\tf^{\inf} - 	f^{\inf}\right)D_{A,B} + C\right\rbrace,
		\end{split}
	\end{equation*}
	we have
	\begin{equation*}
		\min_{0 \leq t < T}\Exp{\left\|\nabla \tf(x^t)\right\|^2}\leq\varepsilon^2.
	\end{equation*}
\end{corollary}
\begin{proof}
	Since $\gamma \leq \frac{\varepsilon^2}{L_fL_{\mathcal{D}}\left(2\left(\tf^{\inf} - 	f^{\inf}\right)D_{A,B} + C\right)},$ we obtain
	\begin{equation*}
		\frac{L_fL_{\mathcal{D}}\gamma}{2}\left(2\left(\tf^{\inf} - 	f^{\inf}\right)D_{A,B} + C\right) \leq \frac{\varepsilon^2}{2}.
	\end{equation*}

	Since $\gamma\leq \frac{1}{\sqrt{L_fL_{\mathcal{D}}D_{A,B} T}},$ we deduce that
	\begin{equation*}
		\begin{split}
			\left(1+D_{A,B}L_fL_{\mathcal{D}}\gamma^2\right)^T &\leq \exp\left(TL_fL_{\mathcal{D}}\gamma^2D_{A,B}\right)\\
			&\leq\exp(1)\leq 3.
		\end{split}
	\end{equation*}
	If $\gamma = \frac{1}{\sqrt{L_fL_{\mathcal{D}}D_{A,B} T}},$ then, since
	\begin{equation*}
		T\geq \frac{36\left(\delta^0\right)^2L_fL_{\mathcal{D}}D_{A,B}}{\varepsilon^4},
	\end{equation*}
	we have $\frac{3\delta^0}{\gamma T}\leq \frac{\varepsilon^2}{2}.$ Further, if $\gamma= \frac{\varepsilon^2}{L_fL_{\mathcal{D}}\left(2\left(\tf^{\inf} - 	f^{\inf}\right)D_{A,B} + C\right)},$ then, since
	\begin{equation*}
		T\geq \frac{6\delta^0L_fL_{\mathcal{D}}\left(2\left(\tf^{\inf} - 	f^{\inf}\right)D_{A,B} + C\right)}{\varepsilon^4},
	\end{equation*}
	we have $\frac{3\delta^0}{\gamma T}\leq \frac{\varepsilon^2}{2}.$
	Combining it with $\eqref{mast:eq:main_noncvx_abc},$ we arrive at $\min_{0 \leq t < T}\Exp{\left\|\nabla \tf(x^t)\right\|^2}\leq\varepsilon^2.$
\end{proof}

\subsection{Strongly convex analysis}
\begin{theorem} \label{mast:eq:ABC_strcvx_res}
Let Assumptions~\ref{mast:ass:C},~\ref{mast:ass:l_f_smooth},~\ref{mast:ass:mu_f-convex},~\ref{mast:eq:unbiased_abc}~and~\ref{mast:eq:abc} hold. Let $r^t\eqdef x^t - x_\cD^\star,$ $t\geq 0.$ Choose a stepsize $0<\gamma \leq \frac{1}{D_{A,B}}.$ Then the iterates $\{x^t\}_{t\geq 0}$ of Algorithm~\ref{mast:eq:abc_sgd} satisfy
	\begin{equation}\label{mast:eq:main_stroncvx_abc}
		\Exp{\left\|r^{t}\right\|^2} \leq \left(1-\gamma\mu_{\mathcal{D}}\mu_f\right)^t\left\|r^0\right\|^2 + \frac{\gamma\left(2\left(\tf^{\inf} - f^{\inf}\right)D_{A,B}+C\right)}{\mu_{\mathcal{D}}\mu_f}.
	\end{equation}
\end{theorem}
\begin{proof}
We get
\begin{equation*}
	\begin{split}
		\left\|r^{t+1}\right\|^2 & = \left\|\left(x^t-\gamma g^t \right) - x_\cD^\star\right\|^2  = \left\|x^t-x_\cD^\star - \gamma g^t\right\|^2 \\
		& = \left\|r^{t}\right\|^2 - 2\gamma\big{<}r^t, g^t\big{>} + \gamma^2\left\|g^t\right\|^2.
	\end{split}
\end{equation*}
Now we compute expectation of both sides of the inequality with respect to $\mS^{t}$, conditioned on $x^t,$ use the fact that $f$ is continuously differentiable and use \eqref{mast:eq:abc}:
\begin{equation*}
	\begin{split}
		\mathbb{E}\left[\left\|r^{t+1}\right\|^2|x^t\right] &= \left\|r^{t}\right\|^2 - 2\gamma\big{<}r^t, \Exp{g^t|x^t}\big{>} + \gamma^2\Exp{\left\|g^t\right\|^2|x^t}\\
		& \leq \left\|r^{t}\right\|^2 - 2\gamma\big{<}r^t, \Exp{\nabla f_{\mS^{t}}(x^t)}\big{>}\\
  &+\gamma^2\Exp{\left(2A\left(f_{\mS_t}(x^t) - f^{\inf}_{\mS_t}\right) + B\left\|\nabla f_{\mS_t}(x^t)\right\|^2 + C\right)}\\
		& = \left\|r^{t}\right\|^2 - 2\gamma\big{<}r^t, \nabla \tf(x^t)\big{>} + 2A\gamma^2\left(\tf(x^t) - \tf^{\inf}\right)+\gamma^2B\Exp{\left\|\nabla f_{\mS_t}(x^t)\right\|^2}\\
		& +\gamma^2\left(C + 2A\left(\tf^{\inf} - \Exp{f_{\mS_t}^{\inf}}\right)\right).\\
	\end{split}
\end{equation*}
Since $\tf$ is $\mu_{\mathcal{D}}\mu_f$-convex, we conclude that $\big{<}r^t, \nabla \tf(x^t)\big{>}\geq \tf(x^t) - \tf^{\inf} + \frac{\mu_{\mathcal{D}}\mu_f}{2}\left\|r^t\right\|^2.$ Therefore, taking the expectation and using the tower property, we obtain
\begin{equation*}
	\begin{split}
		\Exp{\left\|r^{t+1}\right\|^2} & \leq \left(1-\gamma\mu_{\mathcal{D}}\mu_f\right)\left\|r^{t}\right\|^2 - 2\gamma\left(\tf(x) - \tf^{\inf}\right)\left(1 - \gamma D_{A,B}\right)\\ & + \gamma^2\left(2\left(\tf^{\inf} - f^{\inf}\right)D_{A,B}+C\right).\\
	\end{split}
\end{equation*}
Since $\gamma \leq \frac{1}{D_{A,B}},$ we get
\begin{equation*}
	\begin{split}
		\Exp{\left\|r^{t+1}\right\|^2} & \leq \left(1-\gamma\mu_{\mathcal{D}}\mu_f\right)\Exp{\left\|r^{t}\right\|^2}
		+ \gamma^2\left(2\left(\tf^{\inf} - f^{\inf}\right)D_{A,B}+C\right).\\
	\end{split}
\end{equation*}
Unrolling the recurrence, we get
\begin{equation*}
	\Exp{\left\|r^{t}\right\|^2} \leq \left(1-\gamma\mu_{\mathcal{D}}\mu_f\right)^t\left\|r^0\right\|^2 + \frac{\gamma\left(2\left(\tf^{\inf} - f^{\inf}\right)D_{A,B}+C\right)}{\mu_{\mathcal{D}}\mu_f}.
\end{equation*}
\end{proof}
\begin{corollary}
	Fix $\delta > 0.$ Choose the stepsize $\gamma>0$ as
	\begin{equation*}
		\gamma = \min\left\lbrace \frac{1}{D_{A,B}}, \frac{\mu_{\mathcal{D}}\mu_f\delta\left\|r^0\right\|^2}{2\left(\tf^{\inf} - f^{\inf}\right)D_{A,B}+C}\right\rbrace.
	\end{equation*}
	Then, provided that
	\begin{equation*}
		t \geq \frac{1}{\mu_{\mathcal{D}}\mu_f}\left\lbrace D_{A,B}, \frac{2\left(\tf^{\inf} - f^{\inf}\right)D_{A,B}+C}{\mu_{\mathcal{D}}\mu_f\delta\left\|r^0\right\|^2}\right\rbrace \log\frac{1}{\delta},
	\end{equation*}
	we have $\Exp{\left\|r^{t}\right\|^2} \leq 2\delta\left\|r^0\right\|^2.$
\end{corollary}
\begin{proof}
	Since $\gamma\leq \frac{\mu_{\mathcal{D}}\mu_f\delta\left\|r^0\right\|^2}{2\left(\tf^{\inf} - f^{\inf}\right)D_{A,B}+C},$
	we have that
	\begin{equation*}
		\frac{\gamma\left(2\left(\tf^{\inf} - f^{\inf}\right)D_{A,B}+C\right)}{\mu_{\mathcal{D}}\mu_f} \leq\delta\left\|r^0\right\|^2.
	\end{equation*}
	If $\gamma= \frac{\mu_{\mathcal{D}}\mu_f\delta\left\|r^0\right\|^2}{2\left(\tf^{\inf} - f^{\inf}\right)D_{A,B}+C},$ then, since
	\begin{equation*}
		t\geq \frac{2\left(\tf^{\inf} - f^{\inf}\right)D_{A,B}+C}{\mu_{\mathcal{D}}^2\mu_f^2\delta\left\|r^0\right\|^2}\log\frac{1}{\delta},
	\end{equation*}
	we obtain that
	\begin{equation*}
		\left(1-\gamma\mu_{\mathcal{D}}\mu_f\right)^t\leq\exp\left(-\gamma\mu_{\mathcal{D}}\mu_ft\right)\leq\delta.
	\end{equation*}
	Further, if $\gamma=\frac{1}{D_{A,B}},$ then, since
	\begin{equation*}
		t\geq \frac{D_{A,B}}{\mu_{\mathcal{D}}\mu_f}\log\frac{1}{\delta},
	\end{equation*}
	we obtain that
	\begin{equation*}
		\left(1-\gamma\mu_{\mathcal{D}}\mu_f\right)^t\leq\exp\left(-\gamma\mu_{\mathcal{D}}\mu_ft\right)\leq\delta.
	\end{equation*}
	Thus, combining it with~\eqref{mast:eq:main_stroncvx_abc}, we arrive at $\Exp{\left\|r^{t}\right\|^2} \leq 2\delta\left\|r^0\right\|^2.$
\end{proof}
\subsection{Convex analysis}
\begin{theorem}
	Let Assumptions~\ref{mast:ass:C},~\ref{mast:ass:l_f_smooth},~\ref{mast:ass:minimizers_exist},~\ref{mast:eq:unbiased_abc}~and~\ref{mast:eq:abc} hold. Let $f$ be convex. Choose a stepsize $0<\gamma\leq\frac{1}{2D_{A,B}}.$ Fix $T\geq 1$ and let $\bar{x}^T$ be chosen uniformly from the iterates $x^0,\ldots,x^{T-1}$ of Algorithm~\ref{mast:eq:abc_sgd}. Then
	\begin{equation}\label{mast:eq:main_cvx_abc}
		\begin{split}
			\Exp{\tf(\bar{x}^t) - \tf^{\inf}} \leq \frac{\norm{r^0}^2}{\gamma T} + 2\gamma\left(\tf^{\inf} - f^{\inf}\right)D_{A,B}+\gamma C,
		\end{split}
	\end{equation}
	where $r^t\eqdef x^t - x_\cD^\star,$ $t\in\{0,\ldots,T-1\}.$
\end{theorem}
\begin{proof}
 We get
\begin{equation*}
	\begin{split}
		\left\|r^{t+1}\right\|^2 & = \left\|\left(x^t-\gamma g^t \right) - x_\cD^\star\right\|^2  = \left\|x^t-x_\cD^\star - \gamma g^t\right\|^2 \\
		& = \left\|r^{t}\right\|^2 - 2\gamma\big{<}r^t, g^t\big{>} + \gamma^2\left\|g^t\right\|^2.
	\end{split}
\end{equation*}
Now we compute expectation of both sides of the inequality with respect to $\mS^{t}$, conditioned on $x^t,$ use the fact that $f$ is continuously differentiable and use~\eqref{mast:eq:abc}:
\begin{equation*}
	\begin{split}
		\mathbb{E}\left[\left\|r^{t+1}\right\|^2|x^t\right] &= \left\|r^{t}\right\|^2 - 2\gamma\big{<}r^t, \Exp{g^t|x^t}\big{>} + \gamma^2\Exp{\left\|g^t\right\|^2|x^t}\\
		& \leq \left\|r^{t}\right\|^2 - 2\gamma\big{<}r^t, \Exp{\nabla f_{\mS^{t}}(x^t)}\big{>}\\ &+\gamma^2\Exp{\left(2A\left(f_{\mS_t}(x^t) - f^{\inf}_{\mS_t}\right) + B\left\|\nabla f_{\mS_t}(x^t)\right\|^2 + C\right)}\\
		& = \left\|r^{t}\right\|^2 - 2\gamma\big{<}r^t, \nabla \tf(x^t)\big{>} + 2A\gamma^2\left(\tf(x^t) - \tf^{\inf}\right)+\gamma^2B\Exp{\left\|\nabla f_{\mS_t}(x^t)\right\|^2}\\
		& +\gamma^2\left(C + 2A\left(\tf^{\inf} - \Exp{f_{\mS_t}^{\inf}}\right)\right).\\
	\end{split}
\end{equation*}
We can use the convexity of $f$ and Lemma~\ref{mast:lemma:ac} to obtain:
\begin{equation*}
	\begin{split}
		\Exp{\norm{r^{t+1}}^2} & = \norm{r^t}^2 - 2\gamma\left(\tf(x^t) - \tf^{\inf}\right)\left(1-A\gamma\right)+ 2\gamma^2BL_fL_{\mS}^{\max} \left(\tf(x^t) - \tf^{\inf}\right)\\
		& + 2\gamma^2BL_fL_{\mS}^{\max} \left(\tf^{\inf} - f^{\inf}\right) + \gamma^2\left(C + 2A\left(\tf^{\inf} - \Exp{f_{\mS_t}^{\inf}}\right)\right)\\
		& \leq \norm{r^t}^2 - 2\gamma\left(\tf(x^t) - \tf^{\inf}\right)\left(1 - \gamma D_{A,B} \right)\\
		& + 2\gamma^2BL_fL_{\mS}^{\max} \left(\tf^{\inf} - f^{\inf}\right) + \gamma^2\left(C + 2A\left(\tf^{\inf} - f^{\inf}\right)\right).\\
	\end{split}
\end{equation*}
Rearranging and taking expectation, taking into account the condition on the stepsize, we have
\begin{equation*}
	\begin{split}
		\gamma\left(\tf(x^t) - \tf^{\inf}\right) &\leq \Exp{\norm{r^t}^2} - \Exp{\norm{r^{t+1}}^2}
		+ 2\gamma^2\left(\tf^{\inf} - f^{\inf}\right)D_{A,B} + \gamma^2C.\\
	\end{split}
\end{equation*}
Summing over $t = 0,\ldots, T-1$ and using telescopic cancellation gives
\begin{equation*}
	\begin{split}
		\gamma\sum_{t=0}^{T-1}\left(\Exp{\tf(x^t) - \tf^{\inf}}\right) &\leq \norm{r^0}^2 - \Exp{\norm{r^{T}}^2}
		+ 2\gamma^2T\left(\tf^{\inf} - f^{\inf}\right)D_{A,B}+\gamma^2T C.\\
	\end{split}
\end{equation*}
Since $\Exp{\norm{r^{T}}^2} \geq 0$, dividing both sides by $\gamma T$ gives:
\begin{equation*}
	\begin{split}
		\frac{1}{T}\sum_{t=0}^{T-1}\left(\Exp{\tf(x^t) - \tf^{\inf}}\right) &\leq \frac{\norm{r^0}^2}{\gamma T} + 2\gamma\left(\tf^{\inf} - f^{\inf}\right)D_{A,B}+\gamma C.\\
	\end{split}
\end{equation*}
We treat the $\left(\frac{1}{T},\ldots,\frac{1}{T}\right)$ as if it is a probability vector.
Indeed, using that $\tf$ is convex together with Jensen's inequality gives
\begin{equation*}
	\begin{split}
		\Exp{\tf(\bar{x}^t) - \tf^{\inf}} \leq \frac{\norm{r^0}^2}{\gamma T} + \gamma\left(2\left(\tf^{\inf} - f^{\inf}\right)D_{A,B}+ C\right).\\
	\end{split}
\end{equation*}
\end{proof}
\begin{corollary}
	Fix $\delta>0.$ Choose the stepsize $\gamma>0$ as
	\begin{equation*}
		\gamma = \min\left\lbrace \frac{1}{2D_{A,B}}, \frac{\delta\norm{r^0}^2}{2\left(\tf^{\inf} - f^{\inf}\right)D_{A,B}+ C} \right\rbrace.
	\end{equation*}
	Then, provided that
	\begin{equation*}
		T\geq \frac{2L_f\lambda_m^{\mS}}{\delta}\max\left\lbrace 1, \frac{\tf^{\inf} - f^{\inf}}{\delta\norm{r^0}^2}\right\rbrace,
	\end{equation*}
	we have $\Exp{\tf(\bar{x}^t) - \tf^{\inf}}\leq 2\delta\norm{r^0}^2.$
\end{corollary}
\begin{proof}
	Since $\gamma\leq\frac{\delta\norm{r^0}^2}{2\left(\tf^{\inf} - f^{\inf}\right)D_{A,B}+ C},$ we have that
	\begin{equation*}
		\gamma\left(2\left(\tf^{\inf} - f^{\inf}\right)D_{A,B}+ C\right)\leq\delta\norm{r^0}^2.
	\end{equation*}
	If $\gamma = \frac{\delta\norm{r^0}^2}{2\left(\tf^{\inf} - f^{\inf}\right)D_{A,B}+ C},$ then, since
	\begin{equation*}
		T \geq \frac{2\left(\tf^{\inf} - f^{\inf}\right)D_{A,B}+ C}{\delta^2\norm{r^0}^2},
	\end{equation*}
	we obtain that $\frac{\norm{r^0}^2}{\gamma T}\leq \delta\norm{r^0}^2.$ Further, if $\gamma = \frac{1}{2D_{A,B}},$ then, since $T\geq \frac{2D_{A,B}}{\delta},$ we have $\frac{\norm{r^0}^2}{\gamma T}\leq\delta\norm{r^0}^2.$ Thus, combining it with~\eqref{mast:eq:main_cvx_abc}, we arrive at $\Exp{\tf(\bar{x}^t) - \tf^{\inf}}\leq 2\delta\left\|r^0\right\|^2.$
\end{proof}

\section{Distributed setting}
Consider $f$ being a finite sum over a number of machines, i.e., we consider the distributed setup:
\[ \min_{x\in \R^d} \left[f_{\cD}(x) =  \frac{1}{n}\sum_{i=1}^n f_{i, \cD_i} (x)\right],\]
where $f_{i, \cD_i} \eqdef \Exp{f_{i, \mS_i}(x)} = \Exp{f_i(\shift + \mS_i (x-\shift))}$.

Recall that $D_{\max} = \max_{i}\left\lbrace L_{f_i}^2L_{D_i}L_{\mS_i}^{\max}\right\rbrace.$
\subsection{Nonconvex analysis: proof of theorem \ref{mast:thm:distributed_ncvx}}
We solve the problem~\eqref{mast:eq:pretrained_compressed_problem} with the method
\begin{equation}\label{mast:alg:finsum_noncvx_appendix}
	x^{t+1} = x^t - \frac{\gamma}{n} \sum_{i=1}^{n}\left(\mS^{t}_i\right)^{\top} \nabla f_i(y^t)|_{y^t=\shift + \mS^{t}_i (x^t - \shift)}.
\end{equation}

\begin{theorem}
	Assume that each $f_i,$ $i\in[n],$ is differentiable, $L_{f_i}$-smooth and bounded from below by $f_i^{\inf}.$ For every $t\geq 0,$ put $\delta^t\eqdef\mathbb{E}\left[\tf(x^t)-\tf^{\inf}\right]$ and $r^t\eqdef\mathbb{E}\left[\left\|\nabla \tf(x^t)\right\|^2\right].$ Fix $T\geq 1.$ Then the iterates $\{x^t\}_{t=0}^{T-1}$ of Algorithm~\ref{mast:alg:finsum_noncvx} satisfy
	\begin{equation}\label{mast:eq:main_noncvx_finite_sum}
		\begin{split}
			\min_{0 \leq t < T}r^t &\leq \frac{\left(1+\gamma^2D_{\max}\right)^T}{\gamma T}\delta^0 + \gamma D_{\max}\left(\tf^{\inf} - \frac{1}{n}\sum_{i=1}^{n}f^{\inf}_i\right).
		\end{split}
	\end{equation}
\end{theorem}
\begin{proof}
	For $i\in[n],$ due to $L_{f_i}$-smoothness of $f_i,$ we have that
	\begin{equation*}
		\begin{split}
			f_i(s+\mS^{t+1}_i(x^{t+1} - s)) & \leq f_i(s+\mS^{t+1}_i(x^{t} - s))-\gamma \bigg{<} \nabla f_{i,\mS^{t+1}_i}\left(x^t\right), \nabla f_{i,\mS^{t}_i}(x^t) \bigg{>}\\
			& + \frac{L_{f_i}\gamma^2}{2}\bigg{<} \left(\mS^{t+1}_i\right)^{\top}\mS_i^{t+1}\nabla f_{i,\mS^{t}_i}(x^t), \nabla f_{i,\mS^{t}_i}(x^t) \bigg{>}.\\
		\end{split}
	\end{equation*}
	Taking the expectation with respect to $\mS_i^{t+1}$ yields
	\begin{equation*}
		\begin{split}
			f_{i, \cD_i}\left(x^{t+1}\right) &\leq f_{i, \cD_i}\left(x^{t}\right) - \gamma\bigg{<} \nabla f_{i, \cD_i}\left(x^t\right), \nabla f_{\mS_i^{t}}(x^t) \bigg{>} + \frac{L_{f_i}L_{\cD_i}\gamma^2}{2}\left\|\nabla f_{i,\mS_i^{t}}(x^t)\right\|^2.
		\end{split}
	\end{equation*}
	Conditioned on $x^t,$ take expectation with respect to $\mS^{t}_i:$
	\begin{equation*}
		\begin{split}
			\Exp{f_{i, \cD_i}\left(x^{t+1}\right)|x^t} \leq f_{i, \cD_i}\left(x^{t}\right) - \gamma \left\|\nabla f_{i, \cD_i}\left(x^t\right)\right\|^2 + \frac{L_{f_i}L_{\cD_i}\gamma^2}{2}\Exp{\left\|\nabla f_{i,\mS^{t}_i}(x^t) \right\|^2}.\\
		\end{split}
	\end{equation*}
	From Lemma~\ref{mast:lemma:ac} we obtain that
	\begin{equation*}
		\begin{split}
			\Exp{f_{i, \cD_i}\left(x^{t+1}\right)|x^t} & \leq f_{i, \cD_i}\left(x^{t}\right) - \gamma \left\|\nabla f_{i, \cD_i}\left(x^t\right)\right\|^2 + L_{f_i}^2L_{\cD_i}L^{\max}_{\mS_i}\gamma^2 \left(f_{i, \cD_i}(x) - f_i^{\inf}\right).\\
		\end{split}
	\end{equation*}
	For every $i\in[n],$ sum these inequalities, divide by $n:$
	\begin{equation*}
		\begin{split}
			\Exp{\tf\left(x^{t+1}\right)|x^t} & \leq \tf\left(x^t\right) - \frac{\gamma}{n}\sum_{i=1}^{n}\left\|\nabla f_{i, \cD_i}\left(x^t\right)\right\|^2 + \frac{\gamma^2D_{\max}}{n}\sum_{i=1}^{n}\left(f_{i, \cD_i}(x^t) - f_i^{\inf}\right)\\
			& = \tf\left(x^t\right) - \frac{\gamma}{n}\sum_{i=1}^{n}\left\|\nabla f_{i, \cD_i}\left(x^t\right)\right\|^2
			+\gamma^2D_{\max}\left(\tf(x^t) - \frac{1}{n}\sum_{i=1}^{n}f_i^{\inf}\right)\\
			& = \tf\left(x^t\right) - \frac{\gamma}{n}\sum_{i=1}^{n}\left\|\nabla f_{i, \cD_i}\left(x^t\right)\right\|^2
			+\gamma^2D_{\max}\left(\tf(x^t) - \tf^{\inf}\right)\\
			& + \gamma^2D_{\max}\left(\tf^{\inf} - \frac{1}{n}\sum_{i=1}^{n}f^{\inf}_i\right).\\
		\end{split}
	\end{equation*}
	Notice that by Jensen's inequality
	\begin{equation*}
		\frac{1}{n}\sum_{i=1}^{n}\left\|\nabla f_{i, \cD_i}\left(x^t\right)\right\|^2\geq \left\|\frac{1}{n}\sum_{i=1}^{n}\nabla f_{i, \cD_i}\left(x^t\right)\right\|^2=\left\|\nabla \tf(x^t)\right\|^2.
	\end{equation*}
	Subtract $\tf^{\inf}$ from both sides, take expectation on both sides and use the tower property:
	\begin{equation*}
		\begin{split}
			\Exp{\tf\left(x^{t+1}\right) - \tf^{\inf}} & \leq \left(1 + \gamma^2D_{\max}\right)\Exp{\tf\left(x^{t}\right) - \tf^{\inf}} - \gamma \Exp{\left\|\nabla \tf\left(x^t\right)\right\|^2}\\
            &+ \gamma^2D_{\max}\left(\tf^{\inf} - \frac{1}{n}\sum_{i=1}^{n}f^{\inf}_i\right).\\
		\end{split}
	\end{equation*}
	We obtain that
	\begin{equation*}
		\begin{split}
			\gamma r^t & \leq \left(1 + \gamma^2D_{\max}\right)\delta^t- \delta^{t+1}
			+ \gamma^2D_{\max}\left(\tf^{\inf} - \frac{1}{n}\sum_{i=1}^{n}f^{\inf}_i\right).
		\end{split}
	\end{equation*}
	Notice that the iterates $\{x^t\}_{t\geq 0}$ of Algorithm~\ref{mast:eq:abc_sgd} satisfy condition~\eqref{mast:eq:initial_recursion} of Lemma~\ref{mast:lemma:noncvx_weighted} with $M_1=D_{\max},$
	$$
	M_2=D_{\max}\left(\tf^{\inf} - \frac{1}{n}\sum_{i=1}^{n}f^{\inf}_i\right).
	$$
	Divide both sides by $\sum_{t=0}^{T-1}w_t.$ From $\sum_{t=0}^{T-1}w_t\geq Tw_{T-1} = \frac{Tw_{-1}}{1+\gamma^2D_{\max}}$ we can conclude that
	\begin{equation*}
		\begin{split}
			\min_{0 \leq t < T}r^t &\leq \frac{\left(1+\gamma^2 D_{\max}\right)^T}{\gamma T}\delta^0
			+ \gamma D_{\max}\left(\tf^{\inf} - \frac{1}{n}\sum_{i=1}^{n}f^{\inf}_i\right).
		\end{split}
	\end{equation*}
\end{proof}
\begin{corollary}
	Fix $\varepsilon>0.$ Choose the stepsize $\gamma>0$ as
	\begin{equation*}
		\begin{split}
			\gamma &= \min\left\lbrace \frac{1}{\sqrt{D_{\max} T}},\frac{\varepsilon^2}{2D_{\max}\left(\tf^{\inf} - \frac{1}{n}\sum_{i=1}^{n}f^{\inf}_i\right)} \right\rbrace.
		\end{split}
	\end{equation*}
	Then, provided that
	\begin{equation*}
		T\geq \frac{12\delta^0D_{\max}}{\varepsilon^4}\max\left\lbrace 3\delta^0, \tf^{\inf} - \frac{1}{n}\sum_{i=1}^{n}f^{\inf}_i\right\rbrace,
	\end{equation*}
	we have
	\begin{equation*}
		\min_{0 \leq t < T}\Exp{\left\|\nabla \tf(x^t)\right\|^2}\leq\varepsilon^2.
	\end{equation*}
\end{corollary}
\begin{proof}
	Since $\gamma \leq \frac{\varepsilon^2}{2D_{\max}\left(\tf^{\inf} - \frac{1}{n}\sum_{i=1}^{n}f^{\inf}_i\right)},$ we obtain
	\begin{equation*}
		\gamma D_{\max}\left(\tf^{\inf} - \frac{1}{n}\sum_{i=1}^{n}f^{\inf}_i\right) \leq \frac{\varepsilon^2}{2}.
	\end{equation*}

	Since $\gamma\leq \frac{1}{\sqrt{D_{\max} T}},$ we deduce that
	\begin{equation*}
		\begin{split}
			\left(1+\gamma^2D_{\max}\right)^T&\leq \exp\left(T\gamma^2D_{\max}\right)\\
			&\leq\exp(1)\leq 3.
		\end{split}
	\end{equation*}

	If $\gamma = \frac{1}{\sqrt{D_{\max} T}},$ then, since
	\begin{equation*}
		T\geq \frac{36\left(\delta^0\right)^2D_{\max}}{\varepsilon^4},
	\end{equation*}
	we have $\frac{3\delta^0}{\gamma T}\leq \frac{\varepsilon^2}{2}.$ Further, if $\gamma= \frac{\varepsilon^2}{2D_{\max}\left(\tf^{\inf} - \frac{1}{n}\sum_{i=1}^{n}f^{\inf}_i\right)},$ then, since
	\begin{equation*}
		T\geq \frac{12\delta^0D_{\max}\left(\tf^{\inf} - \frac{1}{n}\sum_{i=1}^{n}f^{\inf}_i\right)}{\varepsilon^4},
	\end{equation*}
	we have $\frac{3\delta^0}{\gamma T}\leq \frac{\varepsilon^2}{2}.$
	Combining it with $\eqref{mast:eq:main_noncvx_finite_sum},$ we arrive at $\min_{0 \leq t < T}\Exp{\left\|\nabla \tf(x^t)\right\|^2}\leq\varepsilon^2.$
\end{proof}
\subsection{Strongly convex analysis}
\begin{theorem}
	Assume that each $f_i,$ $i\in[n],$ is differentiable, $L_{f_i}$-smooth and bounded from below by $f_i^{\inf},$ $f$ is $\mu_f$-convex (Assumption~\ref{mast:ass:mu_f-convex} holds). Choose a stepsize $0<\gamma \leq \frac{1}{\max_i\left\lbrace L_{f_i}L_{\mS_i}^{\max} \right\rbrace}.$ Then the iterations $\{x^t\}_{t\geq 0}$ of Algorithm~\ref{mast:alg:finsum_noncvx} satisfy
		\begin{equation*}
		\Exp{\left\|r^{t}\right\|^2} \leq \left(1-\gamma\mu_{\mathcal{D}}\mu_f\right)^t\left\|r^0\right\|^2 + \frac{2\gamma\max_i\left\lbrace L_{f_i}L_{\mS_i}^{\max} \right\rbrace \left(\tf^{\inf} - \frac{1}{n}\sum_{i=1}^{n}f_i^{\inf}\right)}{\mu\mu_f},
	\end{equation*}
	where $r^t\eqdef x^t - x_\cD^\star.$
\end{theorem}
\begin{proof}
	We get that
	\begin{equation*}
		\begin{split}
			\left\|r^{t+1}\right\|^2 & = \left\|\left(x^t-\frac{\gamma}{n}\sum_{i=1}^{n}\nabla f_{i,\mS_i^{t}}(x^t) \right) - x_\cD^\star\right\|^2  = \left\|x^t-x_\cD^\star - \frac{\gamma}{n}\sum_{i=1}^{n}\nabla f_{i,\mS_i^{t}}(x^t)\right\|^2 \\
			& = \left\|r^{t}\right\|^2 - 2\gamma\bigg{<}r^t, \frac{1}{n}\sum_{i=1}^{n}\nabla f_{i,\mS_i^{t}}(x^t)\bigg{>} + \gamma^2\left\|\frac{1}{n}\sum_{i=1}^{n}\nabla f_{i,\mS_i^{t}}(x^t)\right\|^2\\
			& \leq \left\|r^{t}\right\|^2 - 2\gamma\bigg{<}r^t, \frac{1}{n}\sum_{i=1}^{n}\nabla f_{i,\mS_i^{t}}(x^t)\bigg{>} + \gamma^2\frac{1}{n}\sum_{i=1}^{n}\left\|\nabla f_{i,\mS_i^{t}}(x^t)\right\|^2.\\
		\end{split}
	\end{equation*}
	Conditioned on $x^t,$ take expectation with respect to $\mS^{t}_i,$ $i\in[n]:$
	\begin{equation*}
		\begin{split}
			\Exp{\left\|r^{t+1}\right\|^2|x^t} & \leq \left\|r^{t}\right\|^2 - 2\gamma\bigg{<}r^t, \nabla \tf(x^t)\bigg{>} + \gamma^2\frac{1}{n}\sum_{i=1}^{n}\Exp{\left\|\nabla f_{i,\mS_i^{t}}(x^t)\right\|^2|x^t}.\\
		\end{split}
	\end{equation*}
	From Lemma~\ref{mast:lemma:ac} we obtain that
	\begin{equation*}
		\begin{split}
			\Exp{\left\|r^{t+1}\right\|^2|x^t} & \leq \left\|r^{t}\right\|^2 - 2\gamma\bigg{<}r^t, \nabla \tf(x^t)\bigg{>}\\
                & + \gamma^2\frac{2}{n}\sum_{i=1}^{n}L_{f_i}L_{\mS_i}^{\max} \Exp{\left(f_i\left(s+\mS_{i}^t(x^t-s)\right) - f_i^{\inf}\right)}\\
			& = \left\|r^{t}\right\|^2 - 2\gamma\bigg{<}r^t, \nabla \tf(x^t)\bigg{>} + \gamma^2\frac{2}{n}\sum_{i=1}^{n}L_{f_i}L_{\mS_i}^{\max}\left(f_{\mathcal{D}_i}(x^t)-f_i^{\inf}\right)\\
			& \leq \left\|r^{t}\right\|^2 - 2\gamma\bigg{<}r^t, \nabla \tf(x^t)\bigg{>} + \frac{2\gamma^2\max_i\left\lbrace L_{f_i}L_{\mS_i}^{\max} \right\rbrace }{n}\sum_{i=1}^{n}\left(f_{\mathcal{D}_i}(x^t)-f_i^{\inf}\right)\\
			& = \left\|r^{t}\right\|^2 - 2\gamma\bigg{<}r^t, \nabla \tf(x^t)\bigg{>} + 2\gamma^2\max_i\left\lbrace L_{f_i}L_{\mS_i}^{\max} \right\rbrace\left(\tf(x^t) - \tf^{\inf}\right)\\
			& + 2\gamma^2\max_i\left\lbrace L_{f_i}L_{\mS_i}^{\max} \right\rbrace \left(\tf^{\inf} - \frac{1}{n}\sum_{i=1}^{n}f_i^{\inf}\right).\\
		\end{split}
	\end{equation*}
	Since $\tf$ is $\mu_{\mathcal{D}}\mu_f$-convex, we conclude that $\big{<}r^t, \nabla \tf(x^t)\big{>}\geq \tf(x^t) - \tf^{\inf} + \frac{\mu_{\mathcal{D}}\mu_f}{2}\left\|r^t\right\|^2.$ Therefore,
	\begin{equation*}
		\begin{split}
			\Exp{\left\|r^{t+1}\right\|^2|x^t} & \leq \left(1-\gamma\mu_{\mathcal{D}}\mu_f\right)\left\|r^t\right\|^2 - 2\gamma\left(1-\gamma\max_i\left\lbrace L_{f_i}L_{\mS_i}^{\max} \right\rbrace\right)\left(\tf(x^t) - \tf^{\inf}\right)\\
			& + 2\gamma^2\max_i\left\lbrace L_{f_i}L_{\mS_i}^{\max} \right\rbrace \left(\tf^{\inf} - \frac{1}{n}\sum_{i=1}^{n}f_i^{\inf}\right).\\
		\end{split}
	\end{equation*}
	Since $\gamma \leq \frac{1}{\max_i\left\lbrace L_{f_i}L_{\mS_i}^{\max} \right\rbrace},$ taking expectation and using the tower property we get
	\begin{equation*}
		\begin{split}
			\Exp{\left\|r^{t+1}\right\|^2} & \leq \left(1-\gamma\mu_{\mathcal{D}}\mu_f\right)\Exp{\left\|r^t\right\|^2} + 2\gamma^2\max_i\left\lbrace L_{f_i}L_{\mS_i}^{\max} \right\rbrace \left(\tf^{\inf} - \frac{1}{n}\sum_{i=1}^{n}f_i^{\inf}\right).\\
		\end{split}
	\end{equation*}
	Unrolling the recurrence, we get
	\begin{equation*}
		\Exp{\left\|r^{t}\right\|^2} \leq \left(1-\gamma\mu_{\mathcal{D}}\mu_f\right)^t\left\|r^0\right\|^2 + \frac{2\gamma\max_i\left\lbrace L_{f_i}L_{\mS_i}^{\max} \right\rbrace \left(\tf^{\inf} - \frac{1}{n}\sum_{i=1}^{n}f_i^{\inf}\right)}{\mu_{\mathcal{D}}\mu_f}.
	\end{equation*}
\end{proof}
\begin{corollary}
	Fix $\delta > 0.$ Choose the stepsize $\gamma>0$ as
	\begin{equation*}
		\gamma = \min\left\lbrace \frac{1}{\max_i\left\lbrace L_{f_i}L_{\mS_i}^{\max} \right\rbrace}, \frac{\mu_{\mathcal{D}}\mu_f\delta\left\|r^0\right\|^2}{2\max_i\left\lbrace L_{f_i}L_{\mS_i}^{\max} \right\rbrace \left(\tf^{\inf} - \frac{1}{n}\sum_{i=1}^{n}f_i^{\inf}\right)}\right\rbrace.
	\end{equation*}
	Then, provided that
	\begin{equation*}
		t \geq \frac{L_fL_{\mS}^{\max}}{\mu_{\mathcal{D}}\mu_f}\left\lbrace 1, \frac{2\left(\tf^{\inf} - f^{\inf}\right)}{\mu_{\mathcal{D}}\mu_f\delta\left\|r^0\right\|^2}\right\rbrace \log\frac{1}{\delta},
	\end{equation*}
	we have $\Exp{\left\|r^{t}\right\|^2} \leq 2\delta\left\|r^0\right\|^2.$
\end{corollary}
\begin{proof}
	Since $\gamma\leq\frac{\mu_{\mathcal{D}}\mu_f\delta\left\|r^0\right\|^2}{2\max_i\left\lbrace L_{f_i}L_{\mS_i}^{\max} \right\rbrace \left(\tf^{\inf} - \frac{1}{n}\sum_{i=1}^{n}f_i^{\inf}\right)},$
	we have that
	\begin{equation*}
		\frac{2\gamma\max_i\left\lbrace L_{f_i}L_{\mS_i}^{\max} \right\rbrace \left(\tf^{\inf} - \frac{1}{n}\sum_{i=1}^{n}f_i^{\inf}\right)}{\mu_{\mathcal{D}}\mu_f} \leq\delta\left\|r^0\right\|^2.
	\end{equation*}
	If $\gamma=\frac{\mu_{\mathcal{D}}\mu_f\delta\left\|r^0\right\|^2}{2\max_i\left\lbrace L_{f_i}L_{\mS_i}^{\max} \right\rbrace \left(\tf^{\inf} - \frac{1}{n}\sum_{i=1}^{n}f_i^{\inf}\right)},$ then, since
	\begin{equation*}
		t\geq \frac{2\max_i\left\lbrace L_{f_i}L_{\mS_i}^{\max} \right\rbrace \left(\tf^{\inf} - \frac{1}{n}\sum_{i=1}^{n}f_i^{\inf}\right)}{\mu_{\mathcal{D}}^2\mu_f^2\delta\left\|r^0\right\|^2}\log\frac{1}{\delta},
	\end{equation*}
	we obtain that
	\begin{equation*}
		\left(1-\gamma\mu_{\mathcal{D}}\mu_f\right)^t\leq\exp\left(-\gamma\mu_{\mathcal{D}}\mu_ft\right)\leq\delta.
	\end{equation*}
	Further, if $\gamma=\frac{1}{\max_i\left\lbrace L_{f_i}L_{\mS_i}^{\max} \right\rbrace},$ then, since
	\begin{equation*}
		t\geq \frac{\max_i\left\lbrace L_{f_i}L_{\mS_i}^{\max} \right\rbrace}{\mu_{\mathcal{D}}\mu_f}\log\frac{1}{\delta},
	\end{equation*}
	we obtain that
	\begin{equation*}
		\left(1-\gamma\mu_{\mathcal{D}}\mu_f\right)^t\leq\exp\left(-\gamma\mu_{\mathcal{D}}\mu_ft\right)\leq\delta.
	\end{equation*}
	Thus, we arrive at $\Exp{\left\|r^{t}\right\|^2} \leq 2\delta\left\|r^0\right\|^2.$
\end{proof}
\subsection{Convex analysis}
\begin{theorem}
	Assume that each $f_i,$ $i\in[n],$ is differentiable, $L_{f_i}$-smooth and bounded from below by $f_i^{\inf},$ $f$ is convex. Let Assumptions~\ref{mast:ass:C}~and~\ref{mast:ass:minimizers_exist} hold. Let $f$ be convex. Choose a stepsize $0<\gamma\leq\frac{1}{2\max_i\left\lbrace L_{f_i}L_{\mS_i}^{\max} \right\rbrace}.$ Fix $T\geq 1$ and let $\bar{x}^T$ be chosen uniformly from the iterates $x^0,\ldots,x^{T-1}.$ Then
	\begin{equation*}
		\begin{split}
			\Exp{\tf(\bar{x}^T) - \tf^{\inf}}&\leq \frac{\norm{r^0}^2}{\gamma T} + 2\gamma\max_i\left\lbrace L_{f_i}L_{\mS_i}^{\max} \right\rbrace,\\
		\end{split}
	\end{equation*}
	where $r^t\eqdef x^t - x_\cD^\star,$ $t=\{0,\ldots,T-1\}.$
\end{theorem}
\begin{proof}
	We get that
	\begin{equation*}
		\begin{split}
			\left\|r^{t+1}\right\|^2 & = \left\|\left(x^t-\frac{\gamma}{n}\sum_{i=1}^{n}\nabla f_{i,\mS_i^{t}}(x^t) \right) - x_\cD^\star\right\|^2  = \left\|x^t-x_\cD^\star - \frac{\gamma}{n}\sum_{i=1}^{n}\nabla f_{i,\mS_i^{t}}(x^t)\right\|^2 \\
			& = \left\|r^{t}\right\|^2 - 2\gamma\bigg{<}r^t, \frac{1}{n}\sum_{i=1}^{n}\nabla f_{i,\mS_i^{t}}(x^t)\bigg{>} + \gamma^2\left\|\frac{1}{n}\sum_{i=1}^{n}\nabla f_{i,\mS_i^{t}}(x^t)\right\|^2\\
			& \leq \left\|r^{t}\right\|^2 - 2\gamma\bigg{<}r^t, \frac{1}{n}\sum_{i=1}^{n}\nabla f_{i,\mS_i^{t}}(x^t)\bigg{>} + \gamma^2\frac{1}{n}\sum_{i=1}^{n}\left\|\nabla f_{i,\mS_i^{t}}(x^t)\right\|^2.\\
		\end{split}
	\end{equation*}
	Conditioned on $x^t,$ take expectation with respect to $\mS^{t}_i,$ $i\in[n]:$
	\begin{equation*}
		\begin{split}
			\Exp{\left\|r^{t+1}\right\|^2|x^t} & \leq \left\|r^{t}\right\|^2 - 2\gamma\bigg{<}r^t, \nabla \tf(x^t)\bigg{>} + \gamma^2\frac{1}{n}\sum_{i=1}^{n}\Exp{\left\|\nabla f_{i,\mS_i^{t}}(x^t)\right\|^2|x^t}.\\
		\end{split}
	\end{equation*}
	From Lemma~\ref{mast:lemma:ac} and from convexity of $\tf$ we obtain that
	\begin{equation*}
		\begin{split}
			\Exp{\left\|r^{t+1}\right\|^2|x^t} & \leq \left\|r^{t}\right\|^2 - 2\gamma\left( \tf(x^t) - \tf^{\inf}\right)\\ & + \gamma^2\frac{2}{n}\sum_{i=1}^{n}L_{f_i}L_{\mS_i}^{\max} \Exp{\left(f_i\left(s+\mS_{i,t}(x^t-s)\right) - f_i^{\inf}\right)}\\
			& = \left\|r^{t}\right\|^2 - 2\gamma\left( \tf(x^t) - \tf^{\inf}\right) + \gamma^2\frac{2}{n}\sum_{i=1}^{n}L_{f_i}L_{\mS_i}^{\max}\left(f_{\mathcal{D_i}}(x^t)-f_i^{\inf}\right)\\
			& \leq \left\|r^{t}\right\|^2 - 2\gamma\left( \tf(x^t) - \tf^{\inf}\right) + \frac{2\gamma^2\max_i\left\lbrace L_{f_i}L_{\mS_i}^{\max} \right\rbrace }{n}\sum_{i=1}^{n}\left(f_{\mathcal{D_i}}(x^t)-f_i^{\inf}\right)\\
			& = \left\|r^{t}\right\|^2 - 2\gamma\left( \tf(x^t) - \tf^{\inf}\right) + 2\gamma^2\max_i\left\lbrace L_{f_i}L_{\mS_i}^{\max} \right\rbrace\left(\tf(x^t) - \tf^{\inf}\right)\\
			& + 2\gamma^2\max_i\left\lbrace L_{f_i}L_{\mS_i}^{\max} \right\rbrace \left(\tf^{\inf} - \frac{1}{n}\sum_{i=1}^{n}f_i^{\inf}\right)\\
			& = \left\|r^{t}\right\|^2 - 2\gamma\left( \tf(x^t) - \tf^{\inf}\right)\left(1 - \gamma\max_i\left\lbrace L_{f_i}L_{\mS_i}^{\max} \right\rbrace\right)\\
			& + 2\gamma^2\max_i\left\lbrace L_{f_i}L_{\mS_i}^{\max} \right\rbrace \left(\tf^{\inf} - \frac{1}{n}\sum_{i=1}^{n}f_i^{\inf}\right).\\
		\end{split}
	\end{equation*}

	Rearranging and taking expectation, taking into account the condition on the stepsize, we have
	\begin{equation*}
		\begin{split}
			\gamma\left( \tf(x^t) - \tf^{\inf}\right) &\leq \Exp{\norm{r^t}^2} - \Exp{\norm{r^{t+1}}^2} + 2\gamma^2\max_i\left\lbrace L_{f_i}L_{\mS_i}^{\max} \right\rbrace\left(\tf^{\inf} - \frac{1}{n}\sum_{i=1}^{n}f_i^{\inf}\right).\\
		\end{split}
	\end{equation*}
	Summing over $t = 0,\ldots, T-1$ and using telescopic cancellation gives
	\begin{equation*}
		\begin{split}
			\gamma\sum_{t=0}^{T-1}\left(\Exp{\tf(x^t) - \tf^{\inf}}\right) &\leq \norm{r^0}^2 - \Exp{\norm{r^{T}}^2}\\ & + 2T\gamma^2\max_i\left\lbrace L_{f_i}L_{\mS_i}^{\max} \right\rbrace\left(\tf^{\inf} - \frac{1}{n}\sum_{i=1}^{n}f_i^{\inf}\right).\\
		\end{split}
	\end{equation*}
	Since $\Exp{\norm{r^{T}}^2} \geq 0$, dividing both sides by $\gamma T$ gives:
	\begin{equation*}
		\begin{split}
			\frac{1}{T}\sum_{t=0}^{T-1}\left(\Exp{\tf(x^t) - \tf^{\inf}}\right) &\leq \frac{\norm{r^0}^2}{\gamma T} + 2\gamma\max_i\left\lbrace L_{f_i}L_{\mS_i}^{\max} \right\rbrace\left(\tf^{\inf} - \frac{1}{n}\sum_{i=1}^{n}f_i^{\inf}\right).\\
		\end{split}
	\end{equation*}
	We treat the $\left(\frac{1}{T},\ldots,\frac{1}{T}\right)$ as if it is a probability vector.
	Indeed, using that $\tf$ is convex together with Jensen's inequality gives
	\begin{equation*}
		\begin{split}
			\Exp{\tf(\bar{x}^T) - \tf^{\inf}}&\leq \frac{\norm{r^0}^2}{\gamma T} + 2\gamma\max_i\left\lbrace L_{f_i}L_{\mS_i}^{\max} \right\rbrace\left(\tf^{\inf} - \frac{1}{n}\sum_{i=1}^{n}f_i^{\inf}\right).\\
		\end{split}
	\end{equation*}
\end{proof}
\begin{corollary}
	Fix $\delta>0.$ Choose the stepsize $\gamma>0$ as
	\begin{equation*}
		\gamma = \frac{1}{2\max_i\left\lbrace L_{f_i}L_{\mS_i}^{\max} \right\rbrace}\min\left\lbrace 1, \frac{\delta\norm{r^0}^2}{\tf^{\inf} - \frac{1}{n}\sum_{i=1}^{n}f_i^{\inf}} \right\rbrace.
	\end{equation*}
	Then, provided that
	\begin{equation*}
		T\geq \frac{2L_f\lambda_m^{\mS}}{\delta}\max\left\lbrace 1, \frac{\tf^{\inf} - f^{\inf}}{\delta\norm{r^0}^2}\right\rbrace,
	\end{equation*}
	we have $\Exp{\tf(\bar{x}^t) - \tf^{\inf}}\leq 2\delta\norm{r^0}^2.$
\end{corollary}
\begin{proof}
	Since $\gamma\leq\frac{\delta\norm{r^0}^2}{2\max_i\left\lbrace L_{f_i}L_{\mS_i}^{\max} \right\rbrace\left(\tf^{\inf} - \frac{1}{n}\sum_{i=1}^{n}f_i^{\inf}\right)},$ we have that
	\begin{equation*}
		2\gamma\max_i\left\lbrace L_{f_i}L_{\mS_i}^{\max} \right\rbrace\left(\tf^{\inf} - \frac{1}{n}\sum_{i=1}^{n}f_i^{\inf}\right)\leq\delta\norm{r^0}^2.
	\end{equation*}
	If $\gamma = \frac{\delta\norm{r^0}^2}{2\max_i\left\lbrace L_{f_i}L_{\mS_i}^{\max} \right\rbrace\left(\tf^{\inf} - \frac{1}{n}\sum_{i=1}^{n}f_i^{\inf}\right)},$ then, since
	\begin{equation*}
		T \geq \frac{2\max_i\left\lbrace L_{f_i}L_{\mS_i}^{\max} \right\rbrace\left(\tf^{\inf} - \frac{1}{n}\sum_{i=1}^{n}f_i^{\inf}\right)}{\delta^2\norm{r^0}^2},
	\end{equation*}
	we obtain that $\frac{\norm{r^0}^2}{\gamma T}\leq \delta\norm{r^0}^2.$ Further, if $\gamma = \frac{1}{2\max_i\left\lbrace L_{f_i}L_{\mS_i}^{\max} \right\rbrace},$ then, since $T\geq \frac{2\max_i\left\lbrace L_{f_i}L_{\mS_i}^{\max} \right\rbrace}{\delta},$ we have $\frac{\norm{r^0}^2}{\gamma T}\leq\delta\norm{r^0}^2.$ Thus, we arrive at $\Exp{\tf(\bar{x}^t) - \tf^{\inf}}\leq 2\delta\left\|r^0\right\|^2.$
\end{proof}

\section{Variance reduction} \label{mast:sec:variance_reduction}
\begin{algorithm}[t]
\begin{algorithmic}[1]
\caption{Loopless Stochastic Variance Reduced Double Sketched Gradient ($\mathsf{L}$-$\mathsf{SVRDSG}$)} \label{mast:alg:L-SVRSG}
    \STATE \textbf{Parameters:} learning rate $\gamma > 0$; probability $p$; sketches $\mS_1, \ldots, \mS_{N}$; initial model and shift $x^0, \shift \in \R^d$, sketch minibatch size $b$; initial sketch minibatch $\mathcal{B}^0 \subset [N]$.
    \STATE \textbf{Initialization: } $w^0=x^0$, $\hat{h}^0 = \frac{1}{b}\sum \limits_{i \in \mathcal{B}^0}\nabla f_{\mS_i}(w^0)$.
    \FOR{$t = 0, 1, 2 \ldots$}
    \STATE Sample a sketch: $\mS^t$ from $\{\mS_1, \ldots, \mS_{N}\}$
    \STATE Form a gradient estimator: $h^t = \nabla f_{\mS^t}(x^t) - \nabla f_{\mS^t}(w^t) + \hat{h}^t$.
    \STATE Perform a gradient-type step: $x^{t+1} = x^t - \gamma h^t$
    \STATE Sample a Bernoulli random variable $\beta_p$
    \IF{$\beta_p = 1$}
        \STATE Sample $\mathcal{B}^{t}$ uniformly without replacement
       \STATE $w^{t+1} = x^t, \quad \hat{h}^{t+1} = \frac{1}{b}\sum \limits_{i \in \mathcal{B}^{t}}\nabla f_{\mS^t_i}(x^{t}) $
    \ELSE
        \STATE $w^{t+1} = w^t , \quad \hat{h}^{t+1} = \hat{h}^{t}$
    \ENDIF
    \ENDFOR
\end{algorithmic}
\end{algorithm}
In Theorem~\ref{mast:thm:str_conv}, we established linear convergence toward a neighborhood of the solution $x_\cD^\star$ for Algorithm \ref{mast:alg:SGD} (I). To reach the exact solution, the stepsize must decrease to zero, resulting in slower sublinear convergence. The neighborhood's size is linked to the variance of gradient estimator at the solution. Various Variance Reduction (VR) techniques have been proposed to address this issue \citep{gower2020variance}.
Consider the case in which the distribution $\cD$ is uniform and has finite support, i.e., $\{\mS_1, \dots, \mS_N\}$, leading to a finite-sum modification of the MAST problem \eqref{mast:eq:pretrained_compressed_problem}
\begin{equation} \label{mast:eq:fin_sum}
    \tf(x) = \frac{1}{N} \sum_{i=1}^N \mathbb{E}\left[f(\shift + \mS_i(x - \shift))\right].
\end{equation}
In this situation, VR-methods can eliminate the neighborhood enabling linear convergence to the solution.
We utilize the \lsvrg \citep{kovalev2020don,hofmann2015variance} approach, which requires computing the full gradient with probability $p$.
For our formulation, calculating $\nabla f_{\mS}$ for all possible $\mS$ is rarely feasible. For instance, for \rnd{\text{$K$}}, there are
$N = d!/(K!(d-K)!)$ possible operators $\mS_i$.
Therefore, in our Algorithm \ref{mast:alg:L-SVRSG}, we employ a sketch \textit{minibatch} estimator $\hat{h}^t$ computed for a subset $\mathcal{B} \subset [N]$ (sampled uniformly without replacement) of sketches instead of the full gradient.
Finally, we present the convergence results for the strongly convex case.

\begin{theorem} \label{mast:thm:str_cvx-vr}
Assume that $f$ is $L_f$-smooth (\ref{mast:ass:l_f_smooth}), $\mu_f$-strongly convex \eqref{mast:ass:mu_f-convex}, and $\mS$ is sampled from finite set $\{\mS_1,\ldots, \mS_{N}\}$.
Then, for stepsize $\gamma \leq 1/(20 L_f L_{\mS}^{\max})$ and sketch minibatch size $b \in (0, N]$, the iterates of Algorithm~\ref{mast:alg:L-SVRSG} satisfy
\begin{equation*}
\begin{split}
	\mathbb{E} \left[\Psi^{T}\right]&\leq \left(1 - \rho \right)^T\Psi^0
    + \frac{8\gamma^2L_fL_{\mS}^{\max}\left(N - b\right) }{\rho  \max\left\lbrace  1, N - 1\right\rbrace b} \left(\tf^{\inf}- f^{\inf}\right),
\end{split}
\end{equation*}
where $\rho \eqdef \max\left\lbrace\gamma \mu_{\mathcal{D}} \mu_f, \sfrac{p}{2}\right\rbrace$ and Lyapunov function
\[
\begin{aligned}
\Psi^t &\eqdef \|x^t - x_\cD^\star\|^2 \\
&\quad + \frac{16 \gamma^2}{pN} \sum\limits^{N}_{i = 1}\left\|\nabla f_{\mS_i}(w^t) - \nabla f_{\mS_i}(x_\cD^\star)\right\|^2.
\end{aligned}
\]
\end{theorem}
Note that the achieved result demonstrates linear convergence towards the solution's neighborhood. However, this neighborhood is roughly reduced by a factor of $1/b$ compared to Theorem~\ref{mast:thm:str_conv}, and it scales with $N-b$. Thus, when employing a full gradient for $\hat{h}^t$ with $b=N$, the neighborhood shrinks to zero, resulting in a linear convergence rate to the exact solution.

\subsection{$\mathsf{L}$-$\mathsf{SVRDSG}$: strongly convex analysis} %

The proof of Theorem \ref{mast:thm:str_cvx-vr} relies on the following Lemma:
\begin{lemma}
\label{mast:lemma:contr-lsvrg}
	Let Assumptions \ref{mast:ass:C} and \ref{mast:ass:mu_f-convex} hold. Then the following inequality holds:
 \begin{align*}
   \Exp{\left\|x^{t+1} - x_\cD^\star\right\|^2}   = \left(1 - \gamma\mu_{\mathcal{D}}\mu_f\right)\left\|x^t - x_\cD^\star\right\|^2 -2\gamma\left(\tf(x^t) - f(x_\cD^\star)\right) + \gamma^2\Exp{\left\|h^t\right\|^2}.
 \end{align*}
\end{lemma}
\begin{proof}
We start from expanding squared norm:
	\begin{equation*}
		\begin{split}
			\Exp{\left\|x^{t+1} - x_\cD^\star\right\|^2} & = \Exp{\left\|x^t - \gamma h^t - x_\cD^\star\right\|^2}\\
			& = \Exp{\left\|x^t - x_\cD^\star\right\|^2} - 2\gamma \langle h^t, x^t - x_\cD^\star\rangle + \gamma^2\Exp{\left\|h^t\right\|^2}\\
			& = \Exp{\left\|x^t - x_\cD^\star\right\|^2} - 2\gamma \langle \nabla\tf(x^t) , x^t - x_\cD^\star\rangle + \gamma^2\Exp{\left\|h^t\right\|^2}\\
			& = \left(1 - \gamma\mu\mu_f\right)\left\|x^t - x_\cD^\star\right\|^2 -2\gamma\left(\tf(x^t) - f(x_\cD^\star)\right) + \gamma^2\Exp{\left\|h^t\right\|^2}.
		\end{split}
	\end{equation*}
\end{proof}

\begin{lemma}
\label{mast:lemma:moment-lsvrg}
	Let Assumptions \ref{mast:ass:C} and \ref{mast:ass:l_f_smooth} hold. Then the following inequality holds:
 \begin{equation*}
 \begin{split}
     \Exp{\left\|h^t\right\|^2} &\leq 8L_fL_{\mS^t}^{\max}\left(\tf(x^t) - \tf(x_\cD^\star)\right) + 8 \frac{1}{N}\sum_{i = 1}^{N}\left\|\nabla f_{\mS^t_i}(w^t) - \nabla f_{\mS^t_i}(x_\cD^\star)\right\|^2\\
     &+\frac{8(N-b)}{(N-1)b} L_fL_{\mS^t}^{\max}\left(\tf^{\inf} - f^{\inf}\right).
    \end{split}
 \end{equation*}
\end{lemma}
\begin{proof}
We start the proof from the definition of $h^t$:
	\begin{equation*}
		\begin{split}
			\Exp{\left\|h^t\right\|^2} &=\Exp{\left\|\nabla f_{\mS^t}(x^t) - \nabla f_{\mS^t}(w^t) + \frac{1}{b}\sum_{i\in\mathcal{B}^t}\nabla f_{\mS^t_i}(w^t)\right\|^2}\\
			& = \mathbb{E}\left[\left\|\nabla f_{\mS^t}(x^t) - \nabla f_{\mS^t}(w^t) + \frac{1}{b}\sum_{i\in\mathcal{B}^t}\nabla f_{\mS^t_i}(w^t) - \nabla f_{\mS^t}(x_\cD^\star) + \nabla f_{\mS^t}(x_\cD^\star)\right.\right.\\
			&\left.\left. - \frac{1}{b}\sum_{i\in\mathcal{B}^t}\nabla f_{\mS^t_i}(x_\cD^\star) + \frac{1}{b}\sum_{i\in\mathcal{B}^t}\nabla f_{\mS^t_i}(x_\cD^\star)\right\|^2\right]\\
			& \leq 4\Exp{\left\|\nabla f_{\mS^t}(x^t) - \nabla f_{\mS^t}(x_\cD^\star)\right\|^2} + 4\Exp{\left\|\nabla f_{\mS^t}(w^t) - \nabla f_{\mS^t}(x_\cD^\star)\right\|^2}\\
			& + 4\Exp{\left\|\frac{1}{b}\sum_{i\in\mathcal{B}^t}\left(\nabla f_{\mS^t_i}(w^t) - \nabla f_{\mS^t_i}(x_\cD^\star)\right)\right\|^2} + 4\Exp{\left\|\frac{1}{b}\sum_{i\in\mathcal{B}^t}\nabla f_{\mS^t_i}(x_\cD^\star)\right\|^2}\\
			& \stackrel{\text{Lemma~\ref{common:lem:smooth_lower_gradient}}}{\leq} 8L_fL_{\mS^t}^{\max}\left(\tf(x^t) - \tf(x_\cD^\star)\right) + 8 \frac{1}{N}\sum_{i = 1}^{N}\left\|\nabla f_{\mS^t_i}(w^t) - \nabla f_{\mS^t_i}(x_\cD^\star)\right\|^2\\ & + 4\Exp{\left\|\frac{1}{b}\sum_{i\in\mathcal{B}^t}\nabla f_{\mS^t_i}(x_\cD^\star)\right\|^2}.
		\end{split}
	\end{equation*}
Using Lemma \ref{mast:lemma:RR} we obtain
\begin{align*}
    \Exp{\left\|h^t\right\|^2} &\leq 8L_fL_{\mS^t}^{\max}\left(\tf(x^t) - \tf(x_\cD^\star)\right) + 8 \frac{1}{N}\sum_{i = 1}^{N}\left\|\nabla f_{\mS^t_i}(w^t) - \nabla f_{\mS^t_i}(x_\cD^\star)\right\|^2\\
    &+ 4\frac{N-b}{\max\{1,N-1\}b} \frac{1}{N} \sum_{i\in\mathcal{B}^t}\left\|\nabla f_{\mS^t_i}(x_\cD^\star)\right\|^2\\
    &\leq 8L_fL_{\mS^t}^{\max}\left(\tf(x^t) - \tf(x_\cD^\star)\right) + 8 \frac{1}{N}\sum_{i = 1}^{N}\left\|\nabla f_{\mS^t_i}(w^t) - \nabla f_{\mS^t_i}(x_\cD^\star)\right\|^2\\
    &+ \frac{8(N-b)}{\max\{1,N-1\}b} L_fL_{\mS^t}^{\max}\left(\tf^{\inf} - f^{\inf}\right).
\end{align*}
\end{proof}

\begin{lemma}
\label{mast:lemma:prob-lsvrg}
	Let Assumptions \ref{mast:ass:C} and \ref{mast:ass:l_f_smooth} hold. Let
    \[
    D^t = \frac{16\gamma^2}{pN}\sum \limits_{i=1}^N\left\|\nabla f_{\mS^t_i}(x_\cD^\star) - \nabla f_{\mS^t_i}(w^t)\right\|^2.
    \]
    Then the following inequality holds:
    \begin{equation*}
	\begin{split}
		\Exp{D^{t+1}} \leq (1-p)D^t + 32\gamma^2L_fL_{\mS^t}^{\max}\left(\tf(w^t) - \tf(x_\cD^\star)\right).
	\end{split}
\end{equation*}
\end{lemma}
\begin{proof}
Using the smoothness property, we obtain
    \begin{equation*}
	\begin{split}
		\Exp{D^{t+1}} & = (1-p)D^t + p\frac{16\gamma^2}{pN}\sum_{i=1}^{N}\Exp{\left\|\nabla f_{\mS^t_i}(x_\cD^\star) - \nabla f_{\mS^t_i}(w^t)\right\|^2}\\
		& \leq (1-p)D^t + 32\gamma^2L_fL_{\mS^t}^{\max}\left(\tf(w^t) - \tf(x_\cD^\star)\right).
	\end{split}
\end{equation*}
\end{proof}

\begin{theorem}
\label{mast:thm:strong-lsvrg}
Assume that $f$ is $L_f$-smooth (\ref{mast:ass:l_f_smooth}), $\mu_f$-strongly convex \eqref{mast:ass:mu_f-convex}, and $\mS$ is sampled from finite set $\{\mS_1,\ldots, \mS_{N}\}$.
Then, for stepsize $\gamma \leq 1/(20 L_f L_{\mS}^{\max})$ and sketch minibatch size $b \in (0, N]$, the iterates of Algorithm~\ref{mast:alg:L-SVRSG} satisfy
  	\begin{equation*}
	\begin{split}
		\mathbb{E} \left[\Psi^{T}\right]&\leq \left(1 - \rho \right)^T\Psi^0  + \frac{8\gamma^2L_fL_{\mS}^{\max}\left(N - b\right) }{\rho  \max\left\lbrace  1, N - 1\right\rbrace b} \left(\tf^{\inf}- f^{\inf}\right),
	\end{split}
    \end{equation*}
    where
    \[
    \rho \eqdef \max\left\lbrace\gamma \mu_{\mathcal{D}} \mu_f, \sfrac{p}{2}\right\rbrace
    \]
    and the Lyapunov function is
    \[
    \Psi^t \eqdef \|x^t - x_\cD^\star\|^2 + \frac{16 \gamma^2}{pN} \sum\limits^{N}_{i = 1}\left\|\nabla f_{\mS_i}(w^t) - \nabla f_{\mS_i}(x_\cD^\star)\right\|^2.
    \]
\end{theorem}
\begin{proof} We combine three previous lemmas \ref{mast:lemma:contr-lsvrg},\ref{mast:lemma:moment-lsvrg}, \ref{mast:lemma:prob-lsvrg}:
	\begin{equation*}
		\begin{split}
			\Exp{\left\|x^{t+1} - x_\cD^\star\right\|^2 + D^{t+1}} & \leq \left(1-\gamma\mu_{\mathcal{D}}\mu_f\right)\left\|x^{t}- x_\cD^\star\right\|^2 - 2\gamma\left(\tf(x^t) - f(x_\cD^\star)\right) + \gamma^2\Exp{\left\|g^t\right\|^2}\\
			& + (1-p)D^t + 32\gamma^2L_fL_{\mS}^{\max} \left(\tf(w^t) - \tf(x_\cD^\star)\right)\\
			& \leq \left(1-\gamma\mu_{\mathcal{D}}\mu_f\right)\left\|x^{t}- x_\cD^\star\right\|^2 - 2\gamma\left(\tf(x^t) - f(x_\cD^\star)\right) + (1-p)D^t\\
			& + 32\gamma^2L_fL_{\mS}^{\max} \left(\tf(w^t) - \tf(x_\cD^\star)\right)\\
			& + \gamma^2\left(8L_fL_{\mS}^{\max} \left(\tf(w^t) - \tf(x_\cD^\star)\right) + \frac{p}{2\gamma^2}D^t\right.\\
                & \left.+  \frac{8(N-b)}{\max\{1,N-1\}b} L_fL_{\mS^t}^{\max}\left(\tf^{\inf} - f^{\inf}\right)\right)\\
			& = \left(1-\gamma\mu_{\mathcal{D}}\mu_f\right)\left\|x^{t}- x_\cD^\star\right\|^2\\
                &- 2\gamma\left(\tf(x^t) - f(x_\cD^\star)\right)\left(1 - 20\gamma L_fL_{\mS}^{\max}\right)\\
			& + \left(1-\frac{p}{2}\right)D^t +  \frac{8(N-b)}{\max\{1,N-1\}b} \gamma^2 L_fL_{\mS}^{\max}\left(\tf^{\inf} - f^{\inf}\right).
		\end{split}
	\end{equation*}
	Since $\gamma\leq \frac{1}{20L_fL_{\mS}^{\max}},$ we get that
	\begin{equation*}
		\begin{split}
			\Psi^{t+1} \leq \left(1 - \max\left\lbrace\gamma\mu_{\mathcal{D}}\mu_f, \frac{p}{2}\right\rbrace\right)\Psi^t + \frac{8(N-b)}{\max\{1,N-1\}b} \gamma^2 L_fL_{\mS}^{\max}\left(\tf^{\inf} - f^{\inf}\right).
		\end{split}
	\end{equation*}

Unrolling the recursion and using $\rho = \max\left\lbrace\gamma\mu_{\mathcal{D}}\mu_f, \frac{p}{2}\right\rbrace$, we obtain
  	\begin{equation*}
	\begin{split}
		\mathbb{E} \left[\Psi^{T}\right]&\leq \left(1 - \rho \right)^T\Psi^0  + \frac{8\gamma^2L_fL_{\mS}^{\max}\left(N - b\right) }{\rho  \max\left\lbrace  1, N - 1\right\rbrace b} \left(\tf^{\inf}- f^{\inf}\right),
	\end{split}
    \end{equation*}

\end{proof}

\subsubsection{Convex analysis}

Now we formulate and prove theorem for the general (non-strongly) convex regime:
\begin{theorem}
Assume that $f$ is $L_f$-smooth (\ref{mast:ass:l_f_smooth}), convex, and $\mS$ is sampled from finite set $\{\mS_1,\ldots, \mS_{N}\}$.
Then, for stepsize $\gamma \leq 1/(40 L_f L_{\mS}^{\max})$ and sketch minibatch size $b \in (0, N]$ the iterates of Algorithm~\ref{mast:alg:L-SVRSG} satisfy
    \begin{equation*}
    \Exp{\tf(\bar{x}^t)} - f(x_\cD^\star)
    \leq \frac{\Psi^0}{\gamma T}
    + \frac{8(N-b)}{\max\{1,N-1\}b} \gamma L_fL_{\mS}^{\max}\left(\tf^{\inf} - f^{\inf}\right).
    \end{equation*}
    where
    \[
    \bar{x}^T = \frac{1}{T}\sum^{T-1}_{t=0}x^t
    \]
    and the Lyapunov function is
    \[
    \begin{aligned}
    \Psi^t &\eqdef \|x^t - x_\cD^\star\|^2 \\
    &\quad + \frac{16 \gamma^2}{pN} \sum\limits^{N}_{i = 1}\left\|\nabla f_{\mS_i}(w^t) - \nabla f_{\mS_i}(x_\cD^\star)\right\|^2.
    \end{aligned}
    \]
\end{theorem}
\begin{proof}
    We start from the recursion in Theorem \ref{mast:thm:strong-lsvrg}:
    	\begin{equation*}
		\begin{split}
			\Exp{\left\|x^{t+1} - x_\cD^\star\right\|^2 + D^{t+1}} & \leq
			 \left(1-\gamma\mu_{\mathcal{D}}\mu_f\right)\left\|x^{t}- x_\cD^\star\right\|^2\\& - 2\gamma\left(\tf(x^t) - f(x_\cD^\star)\right)\left(1 - 20\gamma L_fL_{\mS}^{\max}\right)\\
			& + \left(1-\frac{p}{2}\right)D^t +  \frac{8(N-b)}{\max\{1,N-1\}b} \gamma^2 L_fL_{\mS}^{\max}\left(\tf^{\inf} - f^{\inf}\right).
		\end{split}
	\end{equation*}
 Using $\mu_f = 0$ and $\left( 1- \frac{p}{2} \right) \leq 1$ we have
     	\begin{equation*}
		\begin{split}
			\Exp{\left\|x^{t+1} - x_\cD^\star\right\|^2 + D^{t+1}} & \leq
			 \left\|x^{t}- x_\cD^\star\right\|^2 - 2\gamma\left(\tf(x^t) - f(x_\cD^\star)\right)\left(1 - 20\gamma L_fL_{\mS}^{\max}\right)\\
			& + D^t +  \frac{8(N-b)}{\max\{1,N-1\}b} \gamma^2 L_fL_{\mS}^{\max}\left(\tf^{\inf} - f^{\inf}\right).
		\end{split}
	\end{equation*}
 Since $\gamma\leq \frac{1}{40L_fL_{\mS}^{\max}},$ we have $\left(1 - 20\gamma L_fL_{\mS}^{\max}\right) \geq \frac{1}{2}$ and it leads to
      	\begin{equation*}
		\begin{split}
			\Exp{\left\|x^{t+1} - x_\cD^\star\right\|^2 + D^{t+1}} & \leq
			 \left\|x^{t}- x_\cD^\star\right\|^2 - \gamma\left(\tf(x^t) - f(x_\cD^\star)\right)\\
			& + D^t +  \frac{8(N-b)}{\max\{1,N-1\}b} \gamma^2 L_fL_{\mS}^{\max}\left(\tf^{\inf} - f^{\inf}\right).
		\end{split}
	\end{equation*}
 Using the tower property, we have
       	\begin{equation*}
		\begin{split}
			\Exp{\Psi^{t+1}} & \leq
			 \Exp{\Psi^t}-\gamma\Exp{\left(\tf(x^t) - f(x_\cD^\star)\right)}\\ & + \frac{8(N-b)}{\max\{1,N-1\}b} \gamma^2 L_fL_{\mS}^{\max}\left(\tf^{\inf} - f^{\inf}\right)\\
    \gamma\Exp{\left(\tf(x^t) - f(x_\cD^\star)\right)} &\leq \Exp{\Psi^t} - \Exp{\Psi^{t+1}}\\&+ \frac{8(N-b)}{\max\{1,N-1\}b} \gamma^2 L_fL_{\mS}^{\max}\left(\tf^{\inf} - f^{\inf}\right)\\
        \gamma\frac{1}{T}\sum^{T-1}_{t=0}\Exp{\left(\tf(x^t) - f(x_\cD^\star)\right)} &\leq \frac{1}{T}\sum^{T-1}_{t=0}\left( \Exp{\Psi^t} - \Exp{\Psi^{t+1}}\right)\\&+ \frac{8(N-b)}{\max\{1,N-1\}b} \gamma^2 L_fL_{\mS}^{\max}\left(\tf^{\inf} - f^{\inf}\right)\\
        \Exp{\tf(\bar{x}^T)} - f(x_\cD^\star) &\leq \frac{\Psi^0}{\gamma T} + \frac{8(N-b)}{\max\{1,N-1\}b} \gamma L_fL_{\mS}^{\max}\left(\tf^{\inf} - f^{\inf}\right).
		\end{split}
	\end{equation*}
\end{proof}

\subsection{$\mathsf{S}$-$\mathsf{PAGE}$: nonconvex analysis}
In this section, we introduce a variant of the Probabilistic Gradient Estimator ($\mathsf{PAGE}$) algorithm applied to the MAST formulation as defined in Equation \ref{mast:eq:pretrained_compressed_problem} for non-convex setting. Li et al.~\citeyearpar{li2021page} showed that this method is optimal in the non-convex regime. We refer to this method as the Sketched Probabilistic Gradient Estimator ($\mathsf{S}$-$\mathsf{PAGE}$). Calculating the full gradient is not efficient, as the number of possible sketches when considering \rnd{$K$} is given by $\frac{n!}{(n-k)!k!}$. Consequently, we employ a minibatch estimator to achieve partial variance reduction, using a large minibatch size where $b > b'$. For the purpose of analysis, we assume that the variance of the sketch gradient is bounded.
\begin{algorithm}[t]
\begin{algorithmic}[1]
\caption{Sketched ProbAbilistic Gradient Estimator ($\mathsf{S}$-$\mathsf{PAGE}$)} \label{mast:alg:S-PAGE}
    \STATE \textbf{Parameters:} learning rate $\gamma > 0$; probability $p$; sketches $\mS_1, \ldots, \mS_{N}$; initial model and shift $x^0, \shift \in \R^d$, sketch minibatch sizes $b$ and $b^\prime < b $; initial sketch minibatch $\mathcal{B}^0 \subset [N]$.
    \STATE \textbf{Initialization: }  $h^0 = \frac{1}{b}\sum \limits_{i \in \mathcal{B}^0}\nabla f_{\mS_i}(x^0)$.
    \FOR{$t = 0, 1, 2 \ldots$}
    \STATE  Perform a gradient-type step: %
    $x^{t+1}=x^t-\gamma h^t$
    \STATE Sample a Bernoulli random variable $\beta_p$
    \IF{$\beta_p = 1$}
        \STATE Sample minibatch $\mathcal{B}^{t}$ with size $b$ uniformly without replacement
        \STATE Form a gradient estimator:
       $h^{t+1} = \frac{1}{b}\sum \limits_{i \in \mathcal{B}^{t}}\nabla f_{\mS^t_i}(x^{t+1}) $
    \ELSE
    \STATE Sample minibatch $\left(\mathcal{B}^{t}\right)^\prime$ with size $b^\prime$ uniformly without replacement
    \STATE Form a gradient estimator:
        $h^{t+1} = h^t+\frac{1}{b^{\prime}} \sum \limits_{i \in \left(\mathcal{B}^{t}\right)^\prime}\left( \nabla f_{\mS^t_i}(x^{t+1})-\nabla f_{\mS^t_i}(x^{t+1})\right)$
    \ENDIF
    \ENDFOR
\end{algorithmic}
\end{algorithm}

\begin{assumption}[Bounded sketch variance]
\label{mast:ass:bounded-variance}
    The sketched gradient has bounded variance if exists $\sigma_{\mathcal{D}} > 0$, such that
   $$ \mathbb{E}\left[\left\|\nabla f_{\mathbf{S}}\left(x\right)-\nabla f_{\mathcal{D}}\left(x\right)\right\|^2\right] \leq \sigma_{\mathcal{D}}^2 \quad \forall x \in \mathbb{R}^d.$$
\end{assumption}

\begin{lemma}[Lemma 2 from \citet{li2021page}]
Suppose that function $f$ is  $L$-smooth and let $x^{t+1}:=x^t-\gamma g^t.$ Then for any $g^t \in \mathbb{R}^d$ and $\gamma>0$, we have
   \begin{equation}
       f\left(x^{t+1}\right) \leq f\left(x^t\right)-\frac{\gamma}{2}\left\|\nabla f(x^t)\right\|^2-\left(\frac{1}{2 \gamma}-\frac{L}{2}\right)\left\|x^{t+1}-x^t\right\|^2+\frac{\gamma}{2}\left\|h^t-\nabla f (x^t)\right\|^2
   \end{equation}
\end{lemma}

\begin{lemma}
\label{mast:lemma:Page}
Suppose that function $f$ is $L_f$-smooth and $\mS$ satisfies Assumption~\ref{mast:ass:C} and let $x^{t+1} = x^t - \gamma h^t$. Then for any $h^t \in \mathbb{R}^d$ and $\gamma>0$, we have
       \begin{equation}
       \begin{aligned}
       \tf\left(x^{t+1}\right)
       &\leq  \tf\left(x^t\right)-\frac{\gamma}{2}\left\|\nabla  \tf(x^t)\right\|^2
       -\left(\frac{1}{2 \gamma}-\frac{L_{f}L_{\mathcal{D}}}{2}\right)\left\|x^{t+1}-x^t\right\|^2\\
       &\qquad+\frac{\gamma}{2}\left\|h^t-\nabla  \tf (x^t)\right\|^2.
       \end{aligned}
   \end{equation}
\end{lemma}
\begin{proof}
    Since $f$ is $L_f$-smooth and $\mS$ satisfies Assumption~\ref{mast:ass:C} the function $f_\mathcal{D}$ is $L_{\tf}$. Then using Lemma~\ref{mast:lemma:Page}, we obtain
           \begin{equation*}
       \begin{aligned}
       \tf\left(x^{t+1}\right)
       &\leq  \tf\left(x^t\right)-\frac{\gamma}{2}\left\|\nabla  \tf(x^t)\right\|^2
       -\left(\frac{1}{2 \gamma}-\frac{L_{\tf}}{2}\right)\left\|x^{t+1}-x^t\right\|^2\\
       &\qquad+\frac{\gamma}{2}\left\|h^t-\nabla  \tf (x^t)\right\|^2.
       \end{aligned}
   \end{equation*}
   Using Lemma~\ref{mast:lemma:smoothness} we have $L_{\tf} \leq L_{\mathcal{D}} L_f $. Plugging this into the inequality, we obtain
          \begin{equation}
       \begin{aligned}
       \tf\left(x^{t+1}\right)
       &\leq  \tf\left(x^t\right)-\frac{\gamma}{2}\left\|\nabla  \tf(x^t)\right\|^2
       -\left(\frac{1}{2 \gamma}-\frac{L_{f}L_{\mathcal{D}}}{2}\right)\left\|x^{t+1}-x^t\right\|^2\\
       &\qquad+\frac{\gamma}{2}\left\|h^t-\nabla  \tf (x^t)\right\|^2.
       \end{aligned}
   \end{equation}
\end{proof}
\begin{lemma}
Suppose that Assumptions \ref{mast:ass:C}, \ref{mast:ass:l_f_smooth} and \ref{mast:ass:bounded-variance} hold. If the gradient estimator $h^{t+1}$ is defined according to Algorithm \ref{mast:alg:S-PAGE}, then we have
  \begin{align*}
\mathbb{E}\left[\left\|h^{t+1}-\nabla \tf(x^{t+1})\right\|^2\right] &\leq\left(1-p\right)\left\|h^t-\nabla \tf(x^t)\right\|^2\\&+\frac{N-b^{\prime}}{(N-1)b^{\prime}}\left(1-p\right) L_f^2 L^{\max}_{\mS} L_{\mathcal{D}}\left\|x^{t+1}-x^t\right\|^2\\
&+ p\frac{N-b}{(N-1)b} \sigma^2_{\mathcal{D}}
  \end{align*}

\begin{proof}
We start by considering two events:
   \begin{align*}
         H &=   \mathbb{E}\left[\left\|h^{t+1}-\nabla \tf(x^{t+1})\right\|^2\right] =  p \mathbb{E}\left[\left\|\frac{1}{b}\sum \limits_{i \in \mathcal{B}^{t}}\nabla f_{\mS^t_i}(x^{t+1})-\nabla \tf(x^{t+1})\right\|^2\right]\\
           &+\left(1-p\right) \mathbb{E}\left[\left\|h^t+\frac{1}{b^{\prime}} \sum \limits_{i \in \left(\mathcal{B}^{t}\right)^\prime}\left( \nabla f_{\mS^t_i}(x^{t+1})-\nabla f_{\mS^t_i}(x^{t})\right)-\nabla \tf(x^{t+1})\right\|^2\right].
   \end{align*}
   Using Assumption~\ref{mast:ass:bounded-variance} and Lemma \ref{mast:lemma:RR}, we obtain
   \begin{equation*}
   \begin{split}
           H &=   \mathbb{E}\left[\left\|h^{t+1}-\nabla \tf(x^{t+1})\right\|^2\right]\\
           &\leq   p\frac{N-b}{(N-1)b} \sigma^2_{\mathcal{D}} +\left(1-p\right) \mathbb{E}\left[\left\|h^t+\frac{1}{b^{\prime}} \sum_{i \in \left(\mathcal{B}^{t}\right)^\prime}\left(\nabla f_{\mS^t_i}(x^{t+1})-\nabla f_{\mS^t_i}(x^{t})\right)-\nabla f_{\mS^t_i}(x^{t+1})\right\|^2\right]\\
              &\leq p\frac{N-b}{(N-1)b} \sigma^2_{\mathcal{D}}\\
              &+ \left(1-p\right) \\
              & \times\mathbb{E}\left[\left\|h^t-\nabla f_{\mathcal{D}}\left(x^t\right)+\frac{1}{b^{\prime}} \sum_{i \in \left(\mathcal{B}^{t}\right)^\prime}\left(\nabla f_{\mS^t_i}(x^{t+1})-\nabla f_{\mS^t_i}(x^{t})\right)-\nabla f_{\mathcal{D}}\left(x^{t+1}\right)+\nabla f_{\mathcal{D}}\left(x^t\right)\right\|^2\right]\\
              &\leq p\frac{N-b}{(N-1)b} \sigma^2_{\mathcal{D}}+\left(1-p\right)\left\|h^t-\nabla f_{\mathcal{D}}\left(x^t\right)\right\|^2\\
              &+\left(1-p\right) \mathbb{E}\left[\left\|\frac{1}{b^{\prime}} \sum_{i \in \left(\mathcal{B}^{t}\right)^\prime}\left(\nabla f_{\mS^t_i}\left(x^{t+1}\right)-\nabla f_{\mS^t_i}\left(x^t\right)\right)-\nabla f_{\mathcal{D}}\left(x^{t+1}\right)+\nabla f_{\mathcal{D}}\left(x^t\right)\right\|^2\right]\\
              & \leq p\frac{N-b}{(N-1)b} \sigma^2_{\mathcal{D}}+\left(1-p\right)\left\|h^t-\nabla f_{\mathcal{D}}\left(x^t\right)\right\|^2\\
              &+p\frac{N-b^{\prime}}{(N-1)b^{\prime}} \mathbb{E}\left[\frac{1}{N}\sum_{i \in \left(\mathcal{B}^{t}\right)^\prime }\left\|\left(\nabla f_{\mS^t_i}\left(x^{t+1}\right)-\nabla f_{\mS^t_i}\left(x^t\right)\right)-\left(\nabla f\left(x^{t+1}\right)-\nabla f\left(x^t\right)\right)\right\|^2\right]\\
             & \leq p\frac{N-b}{(N-1)b} \sigma^2_{\mathcal{D}}+\left(1-p\right)\left\|h^t-\nabla f_{\mathcal{D}}\left(x^t\right)\right\|^2\\
             &+p\frac{N-b^{\prime}}{(N-1)b^{\prime}} \mathbb{E}\left[\frac{1}{N}\sum_{i \in \left(\mathcal{B}^{t}\right)^\prime }\left\|\left(\nabla f_{\mS^t_i}\left(x^{t+1}\right)-\nabla f_{\mS^t_i}\left(x^t\right)\right)\right\|^2\right].
    \end{split}
   \end{equation*}
   Let us consider the last term:
   \begin{align*}
&\mathbb{E}\left[\left\|\nabla f_{\mS^t_i}\left(x^{t+1}\right)-\nabla f_{\mS^t_i}\left(x^t\right)\right\|^2 \right] \\& = \mathbb{E}\left[\left\|\left( (\mS^t_i)^{\top} \nabla f\left( \shift+ \mS^t_i(x^{t+1} - \shift)\right)-(\mS^t_i)^{\top} \nabla f\left( \shift+ \mS^t_i(x^{t} - \shift)\right)\right)\right\|^2 \right] \\
&=\mathbb{E}\left[\left\| (\mS^t_i)^{\top} \left(\nabla f\left( \shift+ \mS^t_i(x^{t+1} - \shift)\right)-\nabla f\left( \shift+ \mS^t_i(x^{t} - \shift)\right)\right)\right\|^2 \right] \\
&\leq\mathbb{E}\left[ \lambda_{\max}\left[(\mS^t_i)(\mS^t_i)^{\top} \right]\left\|  \left(\nabla f\left( \shift+ \mS^t_i(x^{t+1} - \shift)\right)-\nabla f\left( \shift+ \mS^t_i(x^{t} - \shift)\right)\right)\right\|^2 \right] \\
&\leq L^{\max}_{\mS}\mathbb{E} \left[\left\|  \left(\nabla f\left( \shift+ \mS^t_i(x^{t+1} - \shift)\right)-\nabla f\left( \shift+ \mS^t_i(x^{t} - \shift)\right)\right)\right\|^2 \right].
   \end{align*}
Using Lipschitz continuity of gradient of function $f$, we have
\begin{align*}
    \mathbb{E}\left[\left\|\nabla f_{\mS^t_i}\left(x^{t+1}\right)-\nabla f_{\mS^t_i}\left(x^t\right)\right\|^2 \right] & \leq  L^{\max}_{\mS} L_f^2 \mathbb{E} \left[ \left\| \mS^t_i (x^{t+1} - x^t) \right\|^2 \right]\\
    &\leq L^{\max}_{\mS} L_f^2 \lambda_{\max}\left[ \mathbb{E}\left[\mS^t_i (\mS^t_i)^{\top} \right] \right]\left\| x^{t+1} - x^t\right\|^2\\
    &=L^{\max}_{\mS} L_f^2 L_{\mathcal{D}} \left\| x^{t+1} - x^t\right\|^2.
\end{align*}
Plugging this inequality leads us to the final result:

  \begin{align*}
\mathbb{E}\left[\left\|h^{t+1}-\nabla \tf(x^{t+1})\right\|^2\right] &\leq\left(1-p\right)\left\|h^t-\nabla \tf(x^t)\right\|^2\\
&+\frac{N-b^{\prime}}{(N-1)b^{\prime}}\left(1-p\right) L_f^2 L^{\max}_{\mS} L_{\mathcal{D}}\left\|x^{t+1}-x^t\right\|^2\\
&+ p\frac{N-b}{(N-1)b} \sigma^2_{\mathcal{D}}.
  \end{align*}

\end{proof}

\end{lemma}

\begin{theorem}
Assume that $f$ is $L_f$-smooth \eqref{mast:ass:l_f_smooth} and $\mS$ satisfy Assumptions \ref{mast:ass:C} and \ref{mast:ass:bounded-variance}.
    Then, for stepsize $\gamma \leq \frac{1}{\sqrt{\frac{1-p}{p b^\prime}}L_f\left( L_{\mathcal{D}}+\sqrt{L^{\max}_{\mS}L_{\mathcal{D}}}\right)}$, the iterates of Algorithm~\ref{mast:alg:S-PAGE} satisfy

       \begin{align*}
       \mathbb{E}\left[\left\|\nabla \tf\left(\widehat{x}_T\right)\right\|^2\right] \leq \frac{2 \mathbb{E}\left[\Psi_0\right]}{\gamma T} +  \frac{N-b}{(N-1)b}\sigma^2_{\mathcal{D}}.
   \end{align*}
\end{theorem}
\begin{proof}
    \begin{align*}
     &\mathbb{E}\left[\Psi^{t+1}\right]\\ &=    \mathbb{E}\left[\tf(x^{t+1})-f^{\inf}_{\mathcal{D}}+\frac{\gamma}{2 p}\left\|g^{t+1}-\nabla \tf(x^{t+1})\right\|^2\right] \\
     &\leq \mathbb{E}\left[\tf(x^t) - \tf^{\inf}-\frac{\gamma}{2}\left\|\nabla  \tf(x^t)\right\|^2-\left(\frac{1}{2 \gamma}-\frac{L_{\tf}}{2}\right)\left\|x^{t+1}-x^t\right\|^2+\frac{\gamma}{2}\left\|g^t-\nabla  \tf (x^t)\right\|^2\right]\\
     &+\mathbb{E}\left[\frac{\gamma}{2p} \left( \left(1-p\right)\left\|g^t-\nabla \tf(x^t)\right\|^2+\frac{\left(1-p\right) L^{\max}_{\mS} L_{\mathcal{D}}}{b^{\prime}}\left\|x^{t+1}-x^t\right\|^2+ p\frac{N-b}{(N-1)b} \sigma^2_{\mathcal{D}}  \right)\right]\\
     & = \mathbb{E} \left[ \tf(x^t) - \tf^{\inf} -\frac{\gamma}{2}\left\|\nabla  \tf(x^t)\right\|^2 + \frac{\gamma}{2p}\left( \left(1-p\right)\left\|g^t-\nabla \tf(x^t)\right\|^2 + p \left\|g^t-\nabla \tf(x^t)\right\|^2 \right)\right]\\
     &+\mathbb{E}\left[  \frac{\gamma}{2}\frac{N-b}{(N-1)b} \sigma^2_{\mathcal{D}} - \left( \frac{1}{2 \gamma}-\frac{L_{\tf}}{2} - \frac{\left(1-p\right) \gamma L^{\max}_{\mS} L_{\mathcal{D}}}{2pb^{\prime}} \right)\left\|x^{t+1}-x^t\right\|^2\right].
    \end{align*}
Using stepsize $\gamma\leq \frac{1}{\sqrt{\frac{1-p}{p b^\prime}}L_f\left( L_{\mathcal{D}}+\sqrt{L^{\max}_{\mS}L_{\mathcal{D}}}\right)}$ and Lemma, we get
    \begin{align}
     \mathbb{E}\left[\Psi^{t+1}\right] & \leq \mathbb{E}\left[ \Psi^t - \frac{\gamma}{2}\left\|\nabla  \tf(x^t)\right\|^2 + \frac{\gamma}{2}\frac{N-b}{(N-1)b} \sigma^2_{\mathcal{D}} \right]\\
     \frac{\gamma}{2}\left\|\nabla  \tf(x^t)\right\|^2 &\leq \mathbb{E}\left[ \Psi^t \right] - \mathbb{E}\left[ \Psi^{t+1} \right] + \frac{\gamma}{2} \frac{N-b}{(N-1)b}\sigma^2_{\mathcal{D}} \\
     \frac{\gamma}{2}
     \sum^{T-1}_{t=0}\left\|\nabla  \tf(x^t)\right\|^2 &\leq \mathbb{E}\left[ \Psi^T \right] - \mathbb{E}\left[ \Psi^{0} \right] + \frac{\gamma}{2} \frac{N-b}{(N-1)b}\sigma^2_{\mathcal{D}}.
     \end{align}
   Let  $\widehat{x}_T$ be randomly chosen from $\left\{x^t\right\}_{t \in[T]},$ we have
   \begin{align*}
       \mathbb{E}\left[\left\|\nabla \tf\left(\widehat{x}_T\right)\right\|^2\right] \leq \frac{2 \mathbb{E}\left[\Psi_0\right]}{\gamma T} +  \frac{N-b}{(N-1)b}\sigma^2_{\mathcal{D}}.
   \end{align*}
\end{proof}

\section{Additional experiments and details} \label{mast:apdx:experiments}

First, we provide additional details on the experimental settings from Section \ref{mast:sec:experiments}.

\subsection{Experimental details}
In Section \ref{mast:sec:experiments} and for all further experiments, the following problem is considered:
\begin{equation} \label{mast:eq:logreg_problem}
    f(x) \eqdef
    \frac{1}{n} \sum_{i=1}^{n} \log \left(1+\exp (-\mathbf{A}_{i}^{\top} x \cdot b_{i})\right) + \frac{\lambda}{2}\|x\|^{2},
\end{equation}
where $\mathbf{A}_{i} \in \mathbb{R}^{d}$ and $b_{i} \in\{-1,1\}$ are the feature vector and label of the $i$th data point, respectively. %
The approximate \say{optimal} $f^\star$ solution of optimization problem \eqref{mast:eq:logreg_problem} is obtained by running Accelerated Gradient Descent \citep{nesterov1983method} until $\|\nabla f^\star\|^2 \leq 10^{-30}$. Our implementation is based on the public Github \href{https://github.com/konstmish/opt_methods}{repository} of Konstantin Mishchenko. Simulations were performed on a machine with $24 \operatorname{Intel}(\mathrm{R}) \operatorname{Xeon}(\mathrm{R})$ Gold 6246 $\mathrm{CPU}$ @ 3.30 $\mathrm{GHz}$.

\textbf{Sketches.} Problem \eqref{mast:eq:pretrained_compressed_problem} may not be easily solvable precisely for the most general sketches. Therefore, we consider the scenario in which $\cD$ is uniform and has finite support, as in Section \ref{mast:sec:variance_reduction}. We use a special class of diagonal permutation sparsifiers formally introduced in the following example: %
\begin{example}%
Assume\footnote{This is done for simplicity of presentation and can be easily generalized (see Appendix I.1 from Szlendak et al.~\citeyearpar{szlendak2022permutation}).} that $K$ divides $d$, let $q \eqdef d/K $ and $\pi=(\pi_1, \ldots, \pi_d)$ be a random permutation of $[d]$. Then for all $i \in [K]$, we define \textbf{Permutation sparsification} (in short $\texttt{Perm-K}$) operator as
\begin{equation} \label{mast:eq:perm_sketch}
    \mS_i \eqdef K \cdot \sum \limits_{j=q(i-1)+1}^{q i} e_{\pi_j} e_{\pi_j}^\top,
\end{equation}
where $e_1, \dots, e_d \in \R^d$ are standard unit basis vectors. %
\end{example}
In simple words $\texttt{Perm-K}$ sparsifiers require random shuffling and then dividing\footnote{We use \href{https://numpy.org/doc/stable/reference/generated/numpy.array_split.html}{array\underline{{ }{ }}split} method from NumPy (version 1.26.2) package \citep{harris2020array}.} the set of indices $[d] = \{1, \dots, d\}$ into $K$ non-overlapping subsets $C_i \subset \mathcal{G}$ of equal size such that $\cup_{C_i \in \mathcal{G}} C_i = [d]$. Then, every sketch $\mS_i$ is formed from $e_l$ vectors based on indexes $l \in C_i$, resulting in $\cD = \{\mS_1, \dots, \mS_K\}$. To ensure unbiasedness, sketches are sampled uniformly with probability $1/K$.
This class of sketches satisfies $L_\cD = \mu_\cD = K$ and $L_\mS^{\max} = K^2$.

\textbf{Details for Figure \ref{mast:fig:test_acc_box} (and \ref{mast:fig:test_acc_box_appendix}, \ref{mast:fig:test_acc_swarm_appendix}).}
Regularization parameter $\lambda$ in \eqref{mast:eq:logreg_problem} is set to guarantee that the condition number of the loss function $\kappa_f$ is equal to $10^2$. The dataset is shuffled and split to train and test in 0.75 to 0.25 proportions. Initial model weights $x^0 \in \R^d$ are generated from a standard Gaussian distribution $\mathcal{N}(0, 1)$. For every sparsity level $\texttt{Perm-K}$ sketches $\cD = \{\mS_1, \dots, \mS_K\}$ are generated leading to different MAST problem formulations. This process is repeated 10 times with various permutations $\pi$ for changing random seeds. After ERM and MAST models $x$ are obtained the accuracy of sparsified model $\mS_i x$ is calculated for every $i \in [K]$ on the test set. This results in distributions of accuracies for every sparsity level.

\begin{figure*}[h]%
\centering
\subfigure[\textit{a5a} dataset]{%
\includegraphics[width=0.32\linewidth]{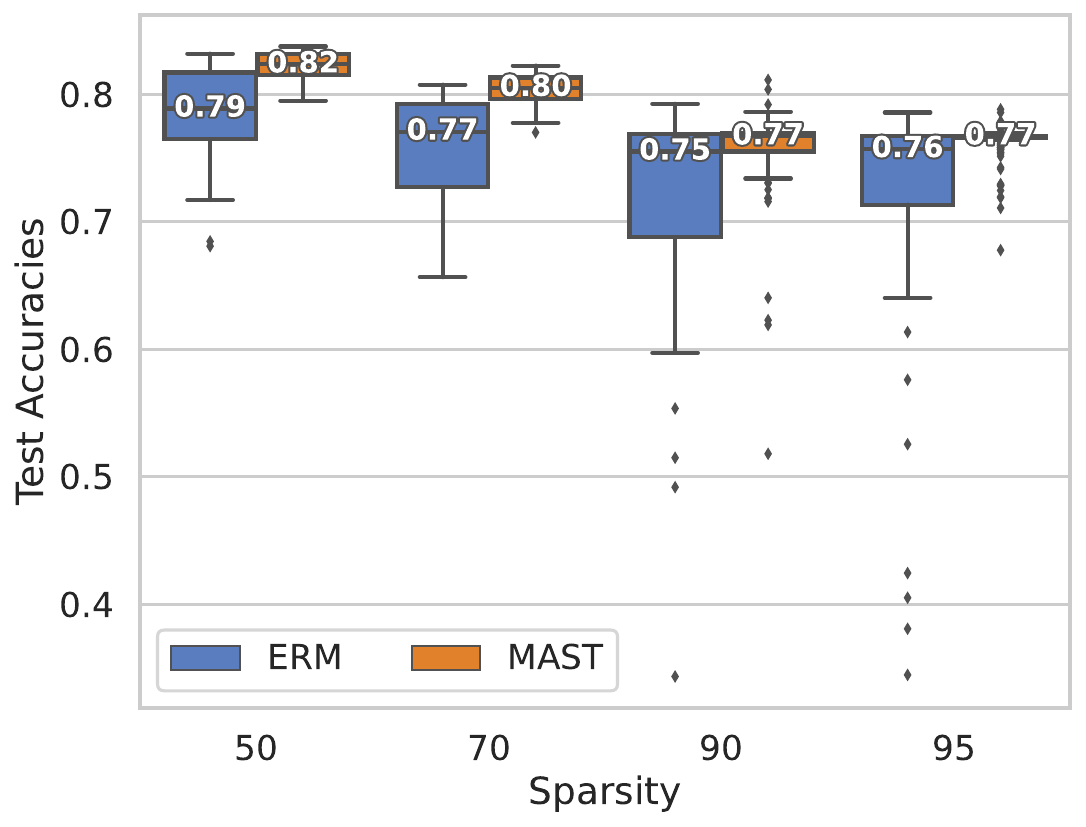}}%
\ %
\subfigure[\textit{a1a} dataset]{%
\includegraphics[width=0.32\linewidth]{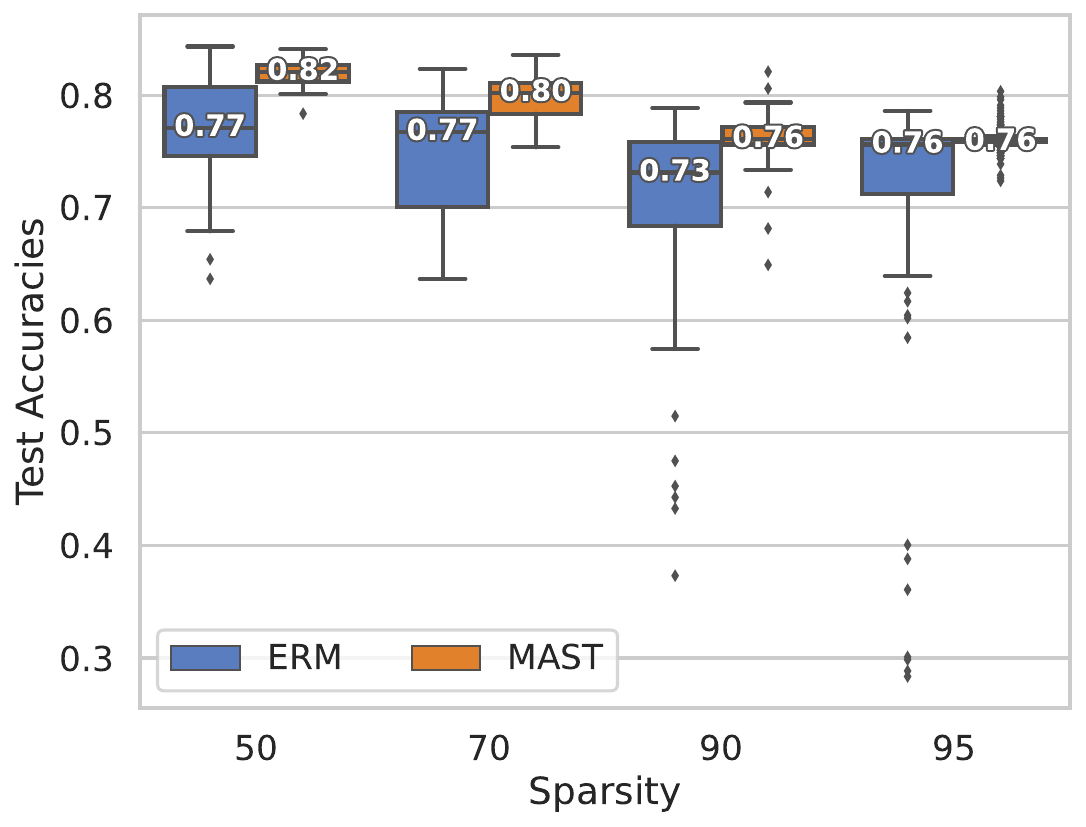}}%
\ %
\subfigure[\textit{w8a} dataset]{%
\includegraphics[width=0.32\linewidth]{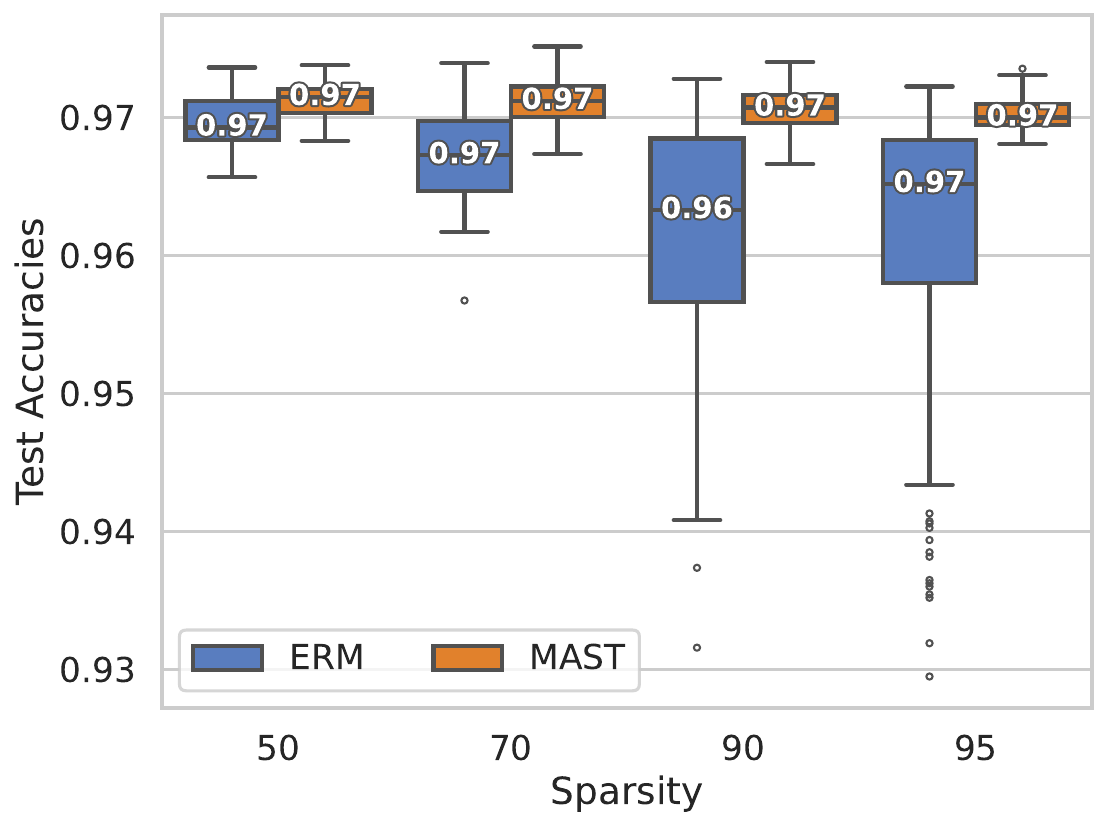}}%
\caption{Distributions of accuracies of sparsified solutions for the ERM \eqref{mast:eq:main} and MAST \eqref{mast:eq:pretrained_compressed_problem} formulations.}
\label{mast:fig:test_acc_box_appendix}
\end{figure*}

\begin{figure*}[h]%
\centering
\subfigure[\textit{a5a} dataset]{%
\includegraphics[width=0.32\linewidth]{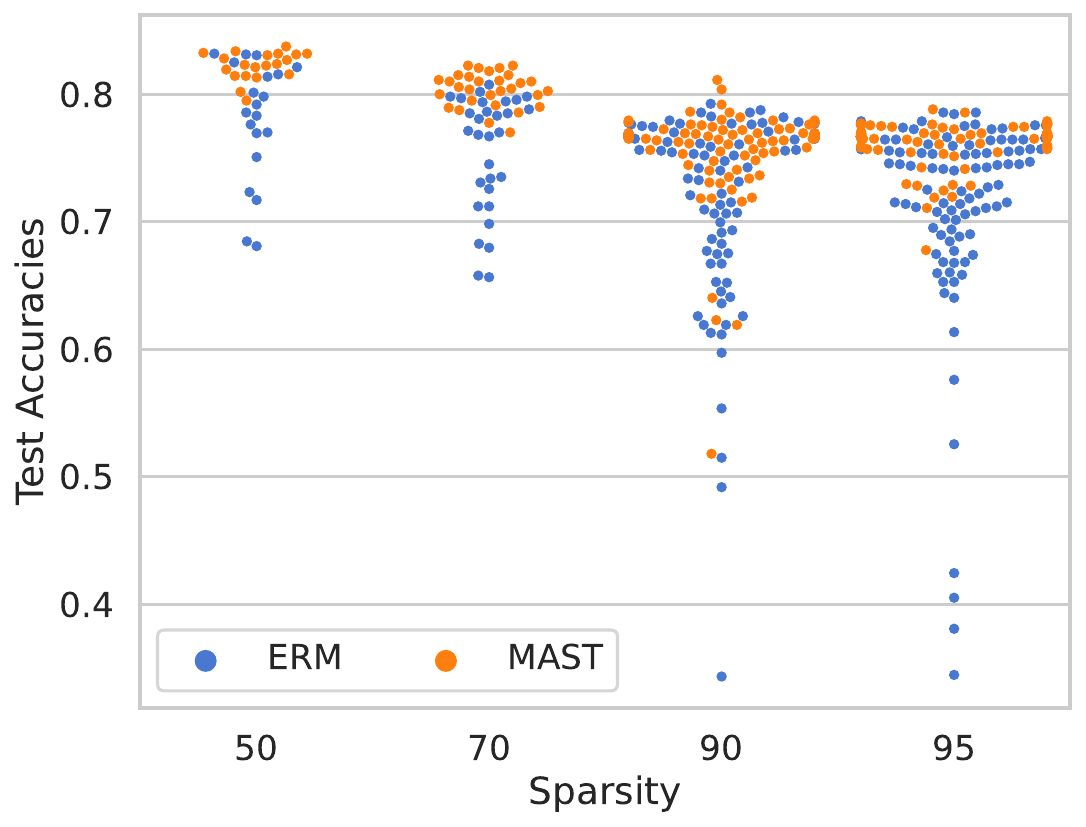}}%
\ %
\subfigure[\textit{a1a} dataset]{%
\includegraphics[width=0.32\linewidth]{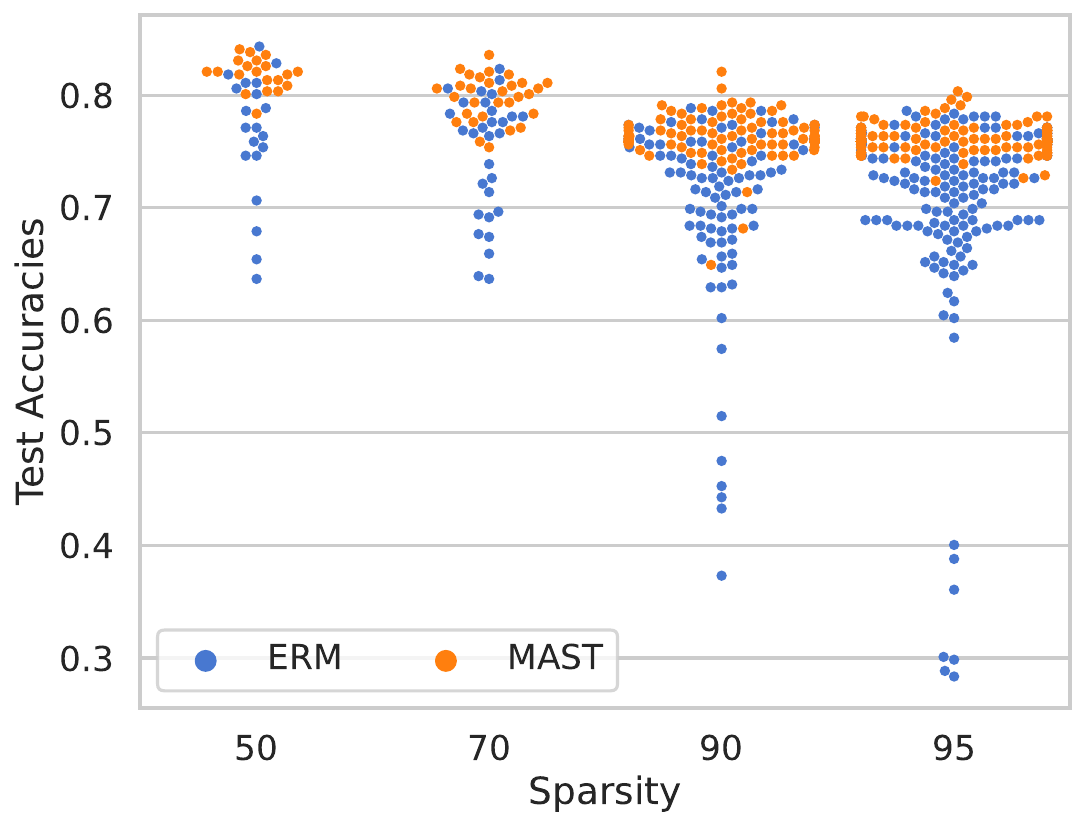}}%
\ %
\subfigure[\textit{w8a} dataset]{%
\includegraphics[width=0.32\linewidth]{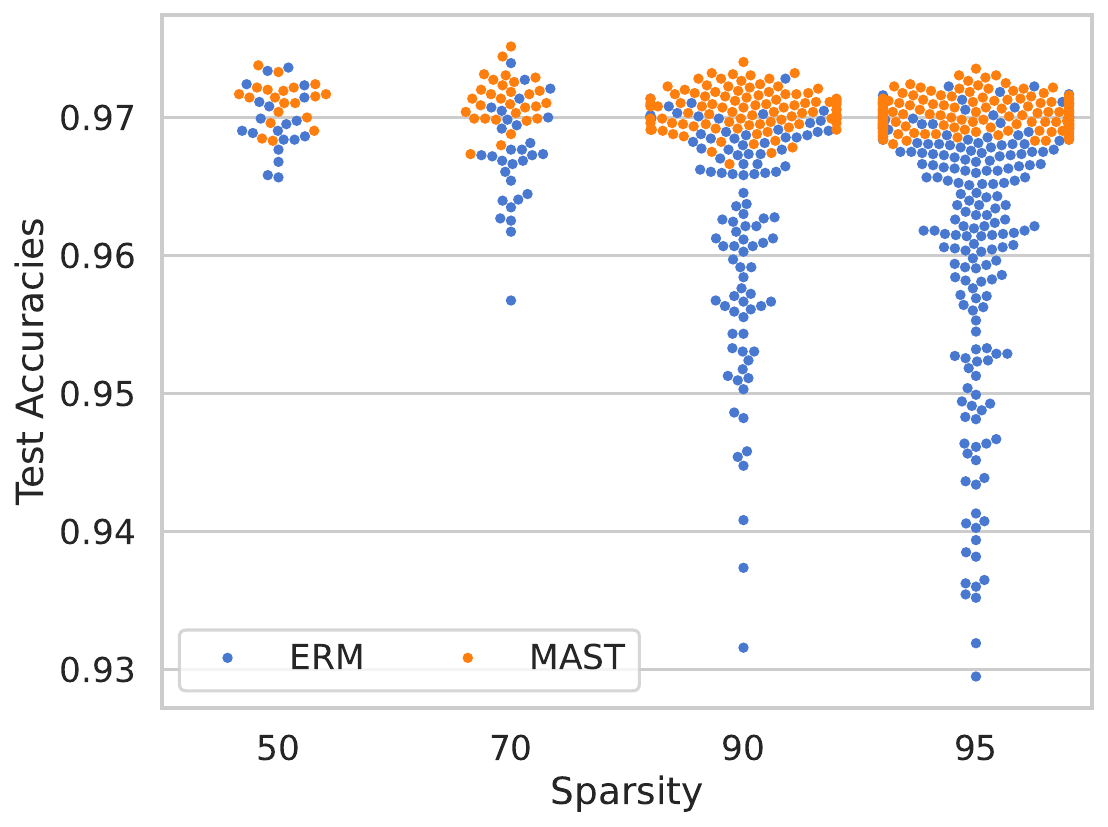}}%
\caption{Test accuracies of sparsified solutions for the ERM formulation \eqref{mast:eq:main} and MAST problem \eqref{mast:eq:pretrained_compressed_problem}.}
\label{mast:fig:test_acc_swarm_appendix}
\end{figure*}

In Figure \ref{mast:fig:test_acc_box_appendix}, we display complete results, including the accuracies lower than 0.5 (unlike in Section \ref{mast:sec:experiments}), which occurred only for the most aggressive sparsification of ERM models. In addition, the median of accuracies (in white) is shown for every (ERM/MAST) approach and sparsification level. Additional results for \textit{a1a} and \textit{w8a} datasets are consistent with those in Section \ref{mast:sec:experiments} as MAST models' performance is higher, less variable, and more resistant to sparsification. Figure \ref{mast:fig:test_acc_swarm_appendix} also shows the swarmplots for the same experiments to represent accuracy values' distributions better.

\subsection{Additional experiments}\label{mast:sec:add_experiments}

\subsubsection{MAST loss trajectory}
In the next experiment, we investigate the optimization efficacy of Double Sketched Gradient Descent (Algorithm \ref{mast:alg:SGD} (II)) with \texttt{Perm-K} sketches \eqref{mast:eq:perm_sketch} for MAST formulation \eqref{mast:eq:pretrained_compressed_problem} with $f$ chosen as \eqref{mast:eq:logreg_problem} for several datasets. The model weights are divided into $K=10$ groups, which allows us to solve the MAST problem, find $\tf^{\inf}$, and evaluate $\tf$, $\nabla \tf$ precisely. Moreover, unlike the previous experiment, the inexactness of the stochastic gradient estimator is introduced via uniform (single element) random subsampling of data $f_i$ as the problem enjoys finite-sum representation \eqref{mast:eq:sketch_fin_sum}.

Figure \ref{mast:fig:mast_loss_perm} shows the trajectory of
a method which averages across all sketches $\mS$ which results in exact (w.r.t. $\cD$) gradient estimator $\nabla f_{i, \cD}$ (right column on legend) and the same algorithm but with uniform sketch subsampling $\nabla f_{i, \mS}$ (left column). The methods are run with 3 different step sizes $\gamma$.
Subsampling of data introduces oscillations preventing the algorithms from converging to the exact optimum. Sketch subsampling leads to additional variance highlighted by the curves corresponding to the same step size. Moreover, our results clearly indicate the necessity of decreasing the learning rate $\gamma$ for sparse/dropout training with SGD. The method with sketch subsampling and the standard SGD step size (dotted \textcolor{blue}{blue} curve) fluctuates around initialization, while full averaging across sketches (dotted \textcolor{cyan}{cyan} curve) partially fixes the issue: the error floor is lower, but there is still almost no convergence. Scaling $\gamma$ inversely proportional to the sparsity level $L_\cD = K$ results in clear linear convergence to a neighborhood of the solution for the $\nabla f_{i, \cD}$ estimator. However, the $\nabla f_{i, \mS}$ estimator requires decreasing the step size by $L_\mS^{\max} = K^2$, which agrees well with the conclusions of Theorems \ref{mast:eq:ABC_noncvx_res} and \ref{mast:eq:ABC_strcvx_res}. %

\begin{figure*}[!h]%
\centering
\subfigure{%
\includegraphics[width=0.81\linewidth]{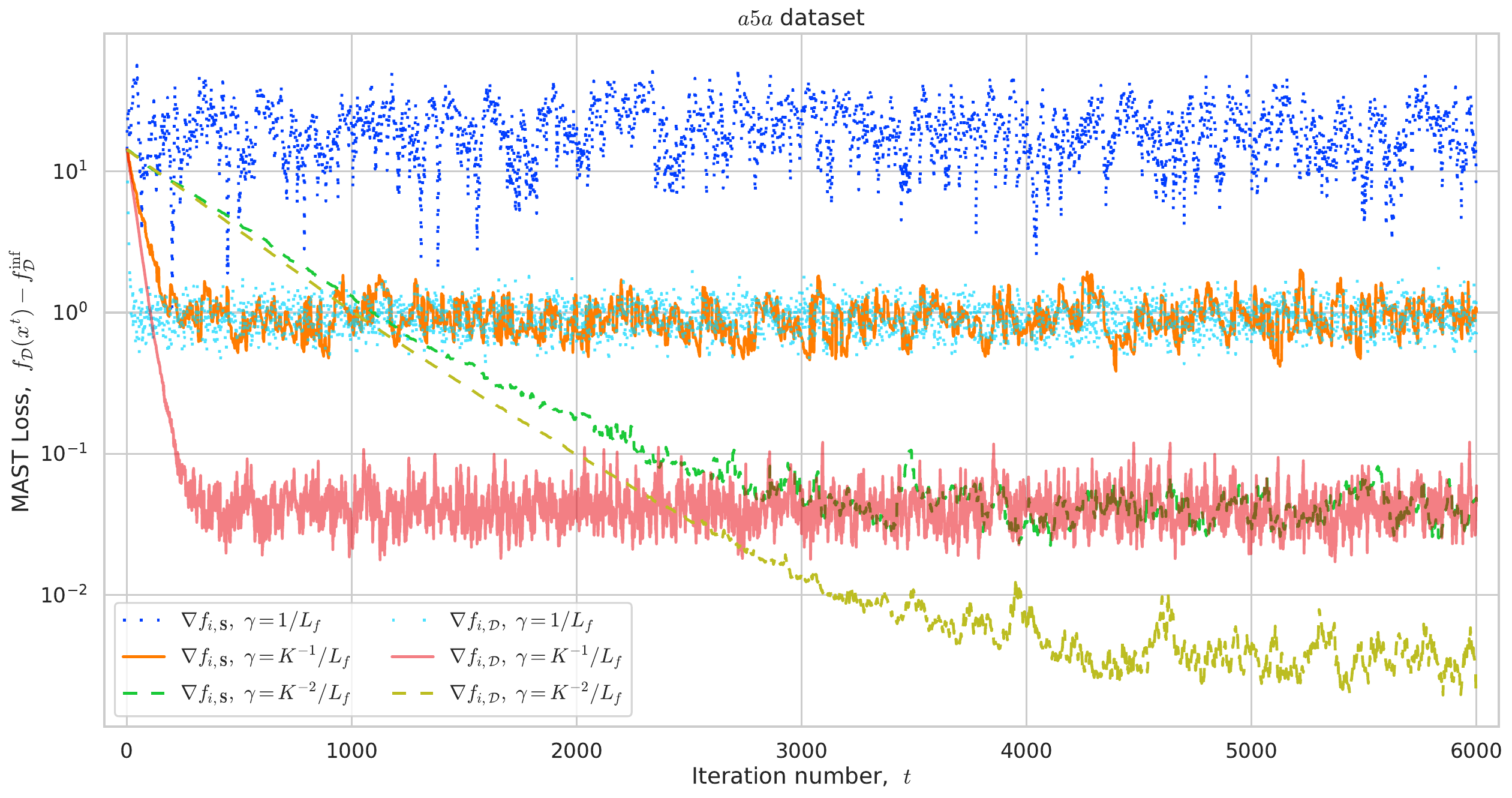}}%
\ %
\subfigure{%
\includegraphics[width=0.81\linewidth]{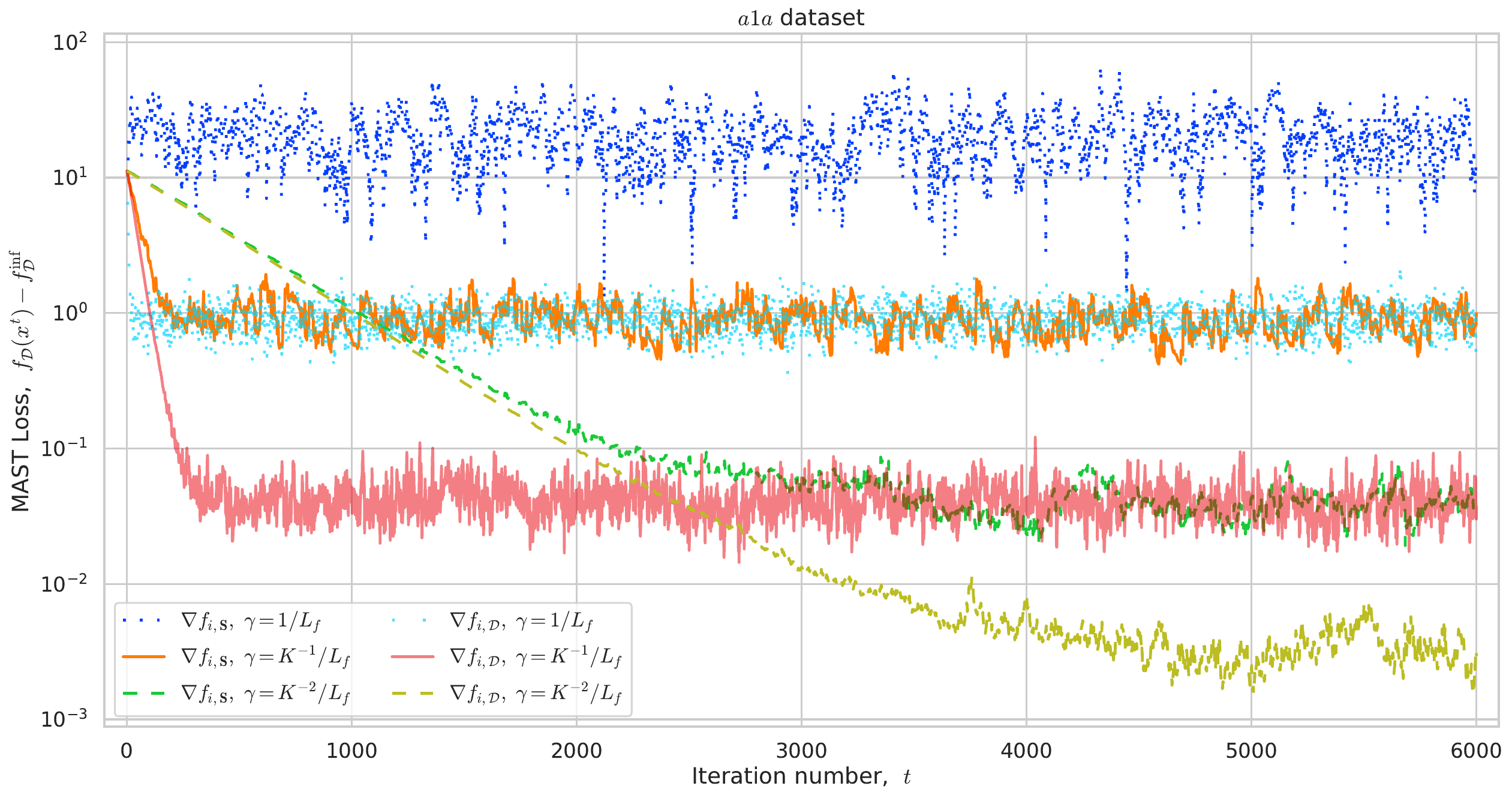}}%
\ %
\subfigure{%
\includegraphics[width=0.81\linewidth]{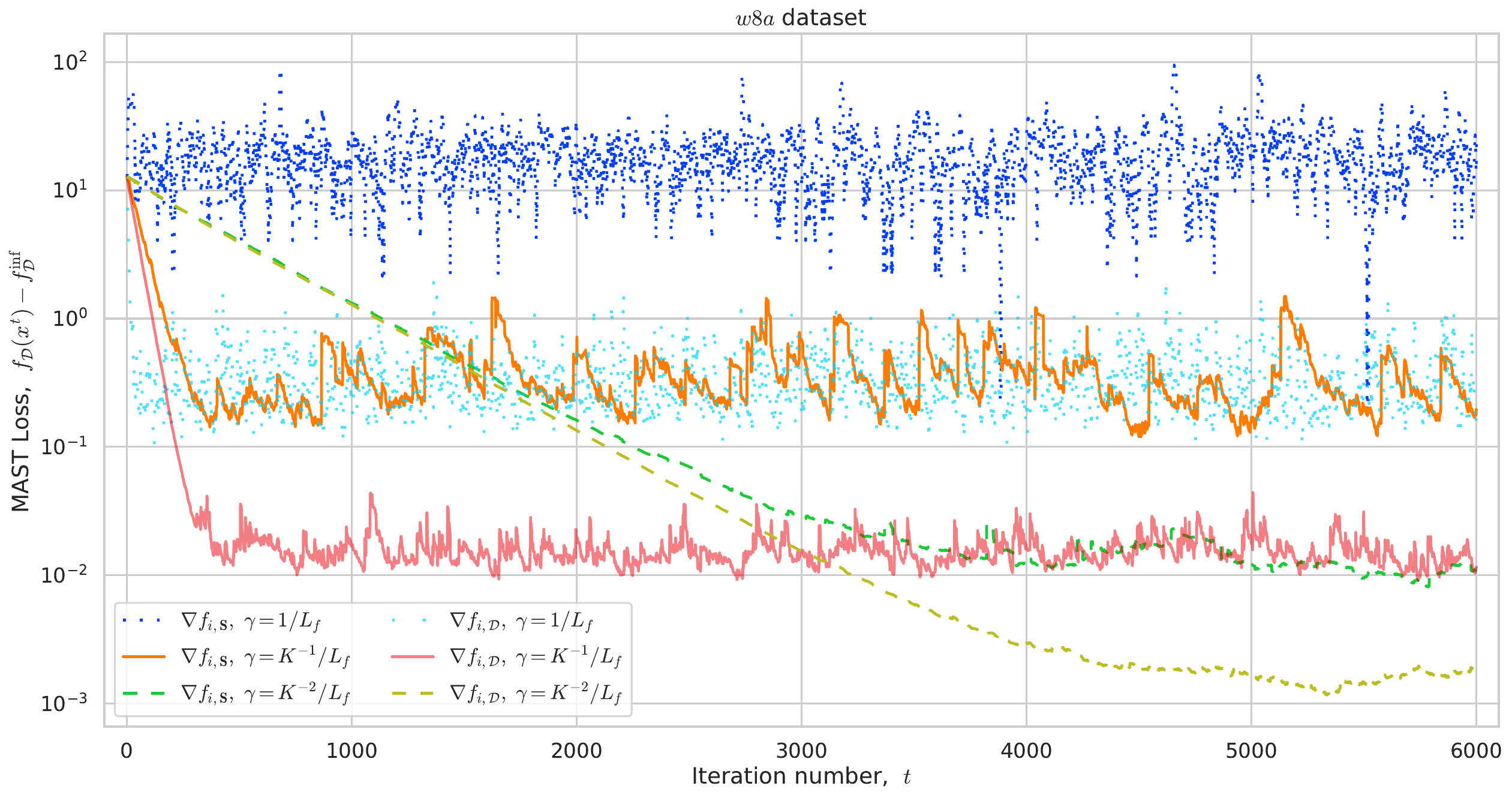}}%
\caption{Finite-sum MAST loss \eqref{mast:eq:sketch_fin_sum} convergence for Algorithm \ref{mast:alg:SGD} (II) with subsampling.}
\label{mast:fig:mast_loss_perm}
\end{figure*}

\subsubsection{\rnd{K} sketches}
\begin{figure}[tbp]
\centering
\begin{minipage}{0.62\textwidth}
\centering
    \includegraphics[width=\linewidth]{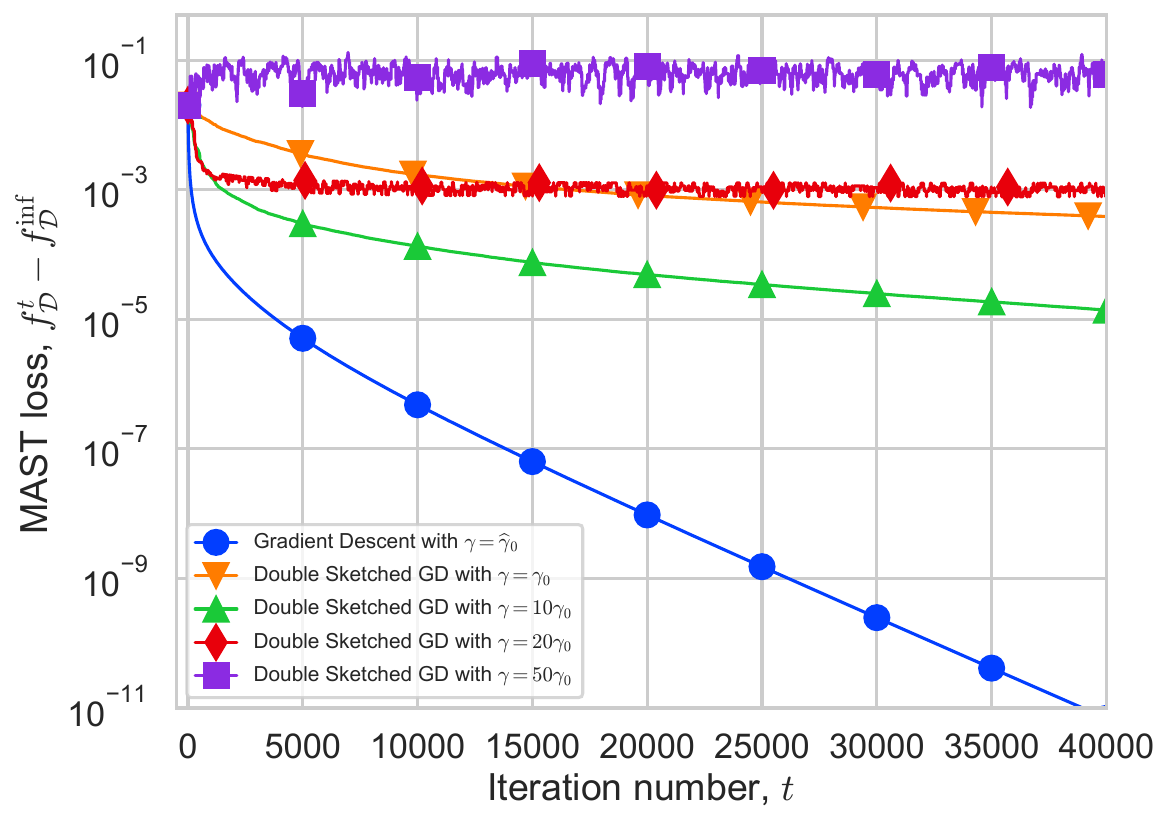}
    \caption{MAST loss \eqref{mast:eq:pretrained_compressed_problem} convergence for Algorithm \ref{mast:alg:SGD} (I) with varying step size and \rnd{$1$} sketches.
    }
\label{mast:fig:mast_loss}
\end{minipage}
\end{figure}
In this experiment depicted in Figure \ref{mast:fig:mast_loss}, we validate the claims of Theorem \ref{mast:thm:str_conv}.
We consider the same logistic regression optimization problem \eqref{mast:eq:logreg_problem} with $\lambda$ set to make sure $\kappa_f = 10^3$. We use the whole dataset \textit{a1a} and initialization $x^0 = 0$.
Consider $\cD$ as a uniform distribution over \rnd{$K$} sketches for $K=1$. Then, MAST stochastic optimization formulation \eqref{mast:eq:pretrained_compressed_problem} leads to a finite-sum problem over sketches $\mS_i$, as defined in \eqref{mast:eq:sketch_fin_sum}. This allows us to evaluate the performance of Algorithm \ref{mast:alg:SGD} (I), which converges linearly for the exact MAST loss \eqref{mast:eq:pretrained_compressed_problem}.
Note that applying Gradient Descent requires computing double sketched gradient for all possible ($N=d$) sketches.
Following the theory, we set the step sizes to $\widehat{\gamma}_0 = 1/(L_f L_{\mathcal{D}})$ and $\gamma_0 = 1/(L_f L^{\max}_{\mS})$.

Our findings accentuate the pivotal role of the appropriate step size $\gamma$ in steering the trajectory of sparse training. Guided by the proposed theoretical framework, this step size must be adjusted in proportion to $K^2/d^2$ for \rnd{$K$}. Notably, for this particular problem at hand, a larger $\gamma$ (e.g., $\gamma = 10 \gamma_0$) can accelerate convergence. Yet, surpassing a delineated boundary can result in stagnation of the progress (e.g., $\gamma = 20 \gamma_0$) and, in specific scenarios, even derail the convergence altogether (e.g., $\gamma = 50 \gamma_0$). Such observations underscore the imperative of modulating the learning rate, especially within the realms of sparse and Dropout training.

\subsubsection{Neural networks} \label{mast:apdx:nn_exp}

This section presents our deep learning experimental results. Figure \ref{mast:fig:lr_p} illustrates the loss behavior of the \emph{distributed} Algorithm \ref{mast:alg:distributed_GD} (for $M=10$ clients) using Bernoulli sketches \eqref{mast:eq:ind_sparse} for $p_i \equiv p$ on the standard loss \eqref{mast:eq:fin_sketch_sum} (for $\mS_i \equiv \mI$).
Our experimental setup closely follows that of \citet{liao2022on}, and for completeness, we reiterate key details.
We employ a ResNet-50 model \citep{he2016deep} pre-trained on ImageNet as a feature extractor, concatenated with two fully connected layers. This combined model is then fine-tuned on the CIFAR-10 dataset \citep{krizhevsky2009cifar}. The outputs of the re-trained ResNet-50 serve as input embeddings, %
while the logit outputs of the combined model are used as labels.

The first column of Figure \ref{mast:fig:lr_p} shows how the method’s performance is affected by the sparsity level ($p$) and step size ($\gamma$). Specifically, for high sparsity $p=0.5$, Figure \ref{mast:fig:1a} illustrates that an excessively large step size ($\gamma=1$) may even lead to divergence of the method for ERM loss. Across all sparsity levels, we observe a \say{sweet spot} for the step size, beyond which increasing $\gamma$ results in slower convergence. Furthermore, training with high sparsity ($p=0.5$) leads to a quick stagnation of the loss, unlike for $p \in\{0.7, 0.9\}$. The second column of Figure \ref{mast:fig:lr_p} displays the subsequent divergence (for $p=0.5$), indicating that high sparsity significantly alters the minimized loss and confirming that Sparse/Dropout training indeed optimizes a different formulation than standard ERM.

In general, larger step sizes and more aggressive sparsification (lower $p$) result in increased loss variance, aligning with our theoretical predictions from Sections \ref{mast:sec:sketches} and \ref{mast:sec:convergence}.
Interestingly, \cref{mast:fig:2b} and~\cref{mast:fig:3b} %
reveal that Dropout training can outperform non-sparse optimization for small step sizes ($\gamma=0.01$) or initial iterations (up to $\sim$ 1200 for $\gamma=0.1$). However, the largest step size ($\gamma=0.5$) is the most efficient in terms of minimizing canonical loss. %

One of the key practical insights derived from our theoretical analysis is that the step size $\gamma$ (learning rate) must be decreased for sparse optimization and Dropout training. Our neural network training results demonstrate that this insight extends to a broader range of models.

\begin{figure*}[h]
\centering
\setlength{\subfigcapskip}{-3pt}  %
\subfigure[High sparsity]{%
\label{mast:fig:1a}%
\includegraphics[width=0.5\linewidth]{KAUST_Thesis/imported/03_mast/plots/stepsize_sparsity/lr_p=0.5.pdf}}%
\hfill
\subfigure[Large step size]{%
\label{mast:fig:2a}%
\includegraphics[width=0.5\linewidth]{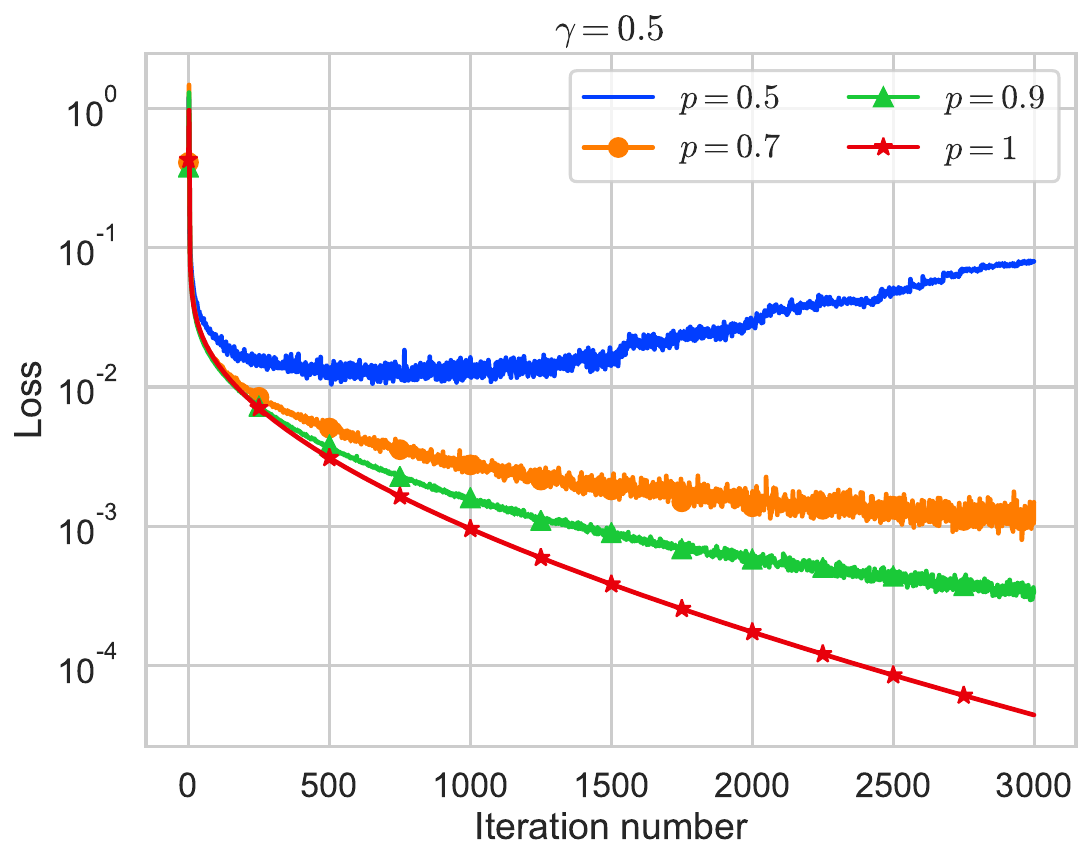}}%
\\[-0.5em]  %
\subfigure[Mid sparsity]{%
\label{mast:fig:1b}%
\includegraphics[width=0.5\linewidth]{KAUST_Thesis/imported/03_mast/plots/stepsize_sparsity/lr_p=0.7.pdf}}%
\hfill
\subfigure[Medium step size]{%
\label{mast:fig:2b}%
\includegraphics[width=0.5\linewidth]{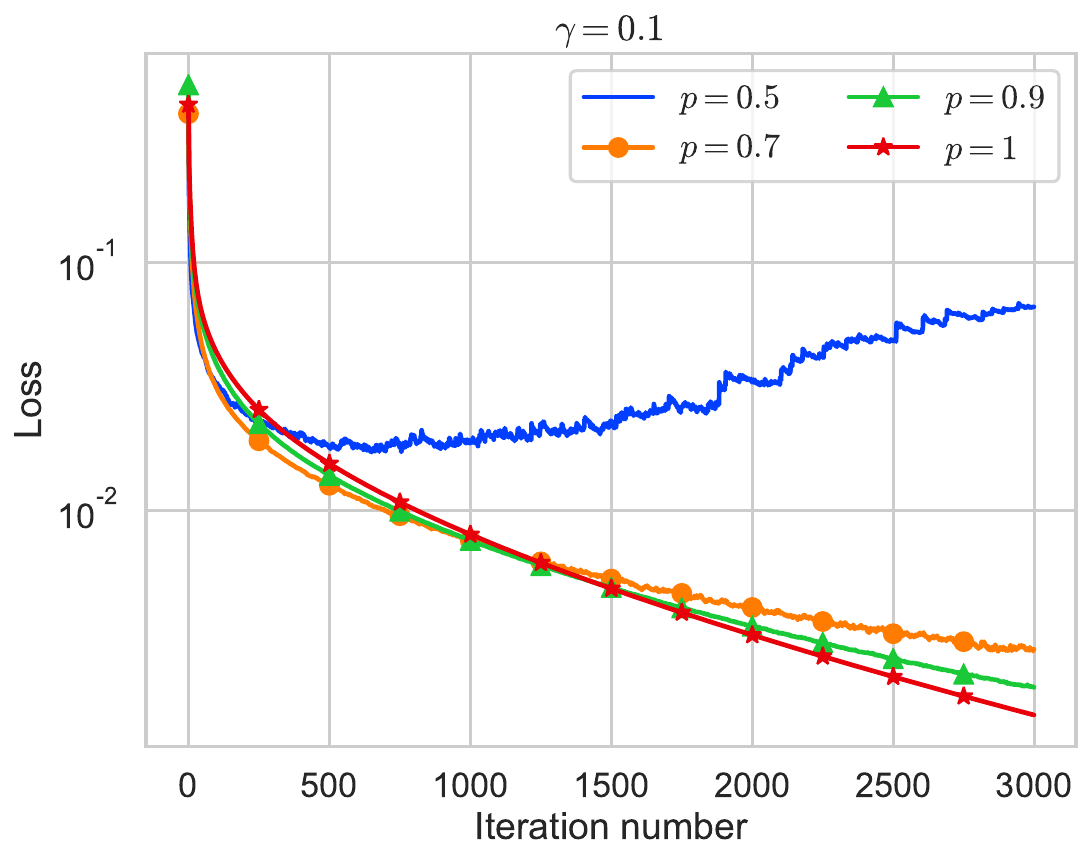}}%
\\[-0.5em]  %
\subfigure[Low sparsity]{%
\label{mast:fig:3a}%
\includegraphics[width=0.5\linewidth]{KAUST_Thesis/imported/03_mast/plots/stepsize_sparsity/lr_p=0.9.pdf}}%
\hfill
\subfigure[Small step size]{%
\label{mast:fig:3b}%
\includegraphics[width=0.5\linewidth]{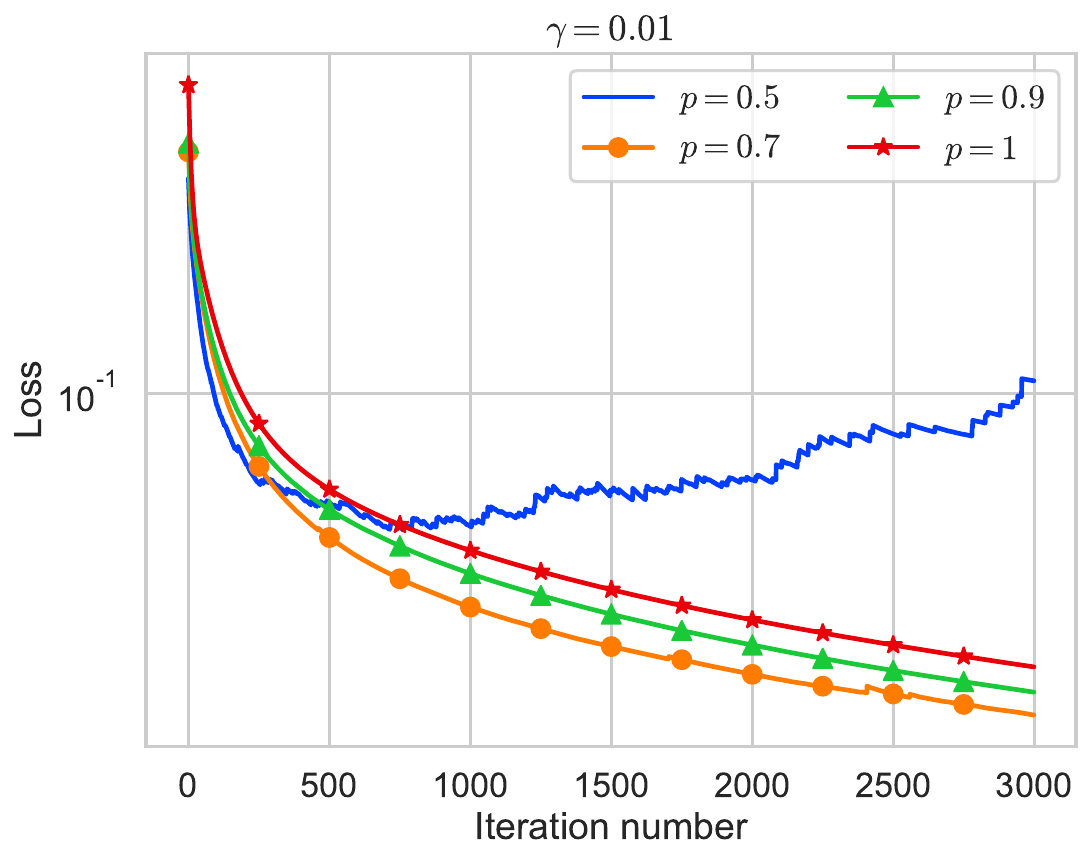}}%

\caption[Performance of Algorithm \ref{mast:alg:distributed_GD} with Bernoulli sketches]{
 Performance of Algorithm \ref{mast:alg:distributed_GD} with Bernoulli sketches \eqref{mast:eq:ind_sparse} on standard loss \eqref{mast:eq:fin_sketch_sum} (for $\mS_i \equiv \mI$).
}
\label{mast:fig:lr_p}
\end{figure*}

\endgroup

\chapter{Supplementary Material for Chapter 4 (Adaptive Clipping)}
\label{appendix:chapter4_adaptive_clipping}

\begingroup
\input{KAUST_Thesis/imported/04_adaptive_clipping/import_setup}

\section{Proofs}
\subsection{Proof of lemma \ref{adaptive:lemma:key_lemma}}
We provide the proof adapted from \citet[Lemma 2]{merad2023robust} for completeness.
Note that
\begin{equation}
    \II\cbr{\norm{\gf(x)} \leq \tau(x)} + \II\cbr{\norm{\gf(x)} > \tau(x)} = 1.
\end{equation}
\textbf{1.} As a reminder $\oalpha(x) \eqdef \Exp{\alpha_{\xi}(x)}$. Then
\begin{align*}
&\Exp{\alpha_{\xi}(x) \gf(x)} - \oalpha(x) \nabla f(x)
    = \Exp{\br{\alpha_{\xi}(x) - \oalpha(x)} \br{\gf(x) - \nabla f(x)}}\\
&= \Exp{\br{\alpha_{\xi}(x) - \oalpha(x)} \br{\gf(x) - \nabla f(x)} \II\cbr{\norm{\gf(x)} \leq \tau(x)}}\\
&\quad{}+
    \Exp{\br{\alpha_{\xi}(x) - \oalpha(x)} \br{\gf(x) - \nabla f(x)} \II\cbr{\norm{\gf(x)} > \tau(x)}}\\
&\overset{\eqref{adaptive:eq:alpha}}{=}
    \Exp{\br{1 - \oalpha(x)} \br{\gf(x) - \nabla f(x)} \br{1 - \II\cbr{\norm{\gf(x)} > \tau(x)}}}\\
&\quad{}-
    \oalpha(x)\Exp{ \br{\gf(x) - \nabla f(x)} \II\cbr{\norm{\gf(x)} > \tau(x)}}\\
&\quad{}+
    \Exp{\frac{\tau(x)}{\norm{\gf(x)}} \br{\gf(x) - \nabla f(x)} \II\cbr{\norm{\gf(x)} > \tau(x)}}\\
&= \Exp{\br{1 - \oalpha(x)} \br{\gf(x) - \nabla f(x)}}\\
&\quad{}+
    \Exp{\br{\oalpha(x) - 1} \br{\gf(x) - \nabla f(x)} \II\cbr{\norm{\gf(x)} > \tau(x)}}\\
&\quad{}-
    \oalpha(x)\Exp{ \br{\gf(x) - \nabla f(x)} \II\cbr{\norm{\gf(x)} > \tau(x)}}\\
&\quad{}+
    \Exp{ \frac{\tau(x)}{\norm{\gf(x)}} \br{\gf(x) - \nabla f(x)} \II\cbr{\norm{\gf(x)} > \tau(x)}}\\
&= \Exp{\br{\frac{\tau(x)}{\norm{\gf(x)}} - 1} \br{\gf(x) - \nabla f(x)} \II\cbr{\norm{\gf(x)} > \tau(x)}}.
\end{align*}
Next, we use the fact that if $\frac{\tau(x)}{\norm{\gf(x)}} \in (0, 1)$ then $|\tau(x)/\norm{\gf(x)} - 1| \in (0, 1)$. Thus,
\begin{align*}
    \norm{\Exp{\alpha_{\xi}(x) \gf(x)} - \oalpha(x) \nabla f(x)}
    &\leq
    \Exp{\left|\frac{\tau(x)}{\norm{\gf(x)}} - 1\right| \II\cbr{\norm{\gf(x)} > \tau(x)} \norm{\gf(x) - \nabla f(x)}}
    \\&\leq
    \Exp{\II\cbr{\norm{\gf(x)} > \tau(x)} \norm{\gf(x) - \nabla f(x)}}
    \\&\overset{(*)}{\leq}
    \br{1-p}^{1-1/q} \Exp{\norm{\gf(x) - \nabla f(x)}^q}^{1/q}
    \\&\overset{\eqref{adaptive:eq:bound_var}}{\leq}
    \br{1-p}^{1-1/q} \sigma_q,
\end{align*}
where $(*)$ is due to Hölder's inequality.

\textbf{2.} Denote by $\cQ_p\br{\norm{\gf(x)}} = \tau(x)$ the $p$-th quantile of $\norm{\gf(x)}$ distribution. Then
\begin{equation}
\begin{aligned}
    \tau(x)
    &= \cQ_p\br{\norm{\gf(x) - \nabla f(x) + \nabla f(x)}}\\
    &\overset{\text{Lemma~\ref{common:lem:triangle}}}{\leq}
    \norm{\nabla f(x)} + \cQ_p\Bigl(\bigl\|\underbrace{\gf(x) - \nabla f(x)}_{\delta_x}\bigr\|\Bigr).
\end{aligned}
\end{equation}
By using quantile definition and Markov's inequality
\begin{equation}
    1 - p =
    \mathrm{Prob} \cbr{\norm{\delta_x} > \cQ_p(\norm{\delta_x})} \leq
    \br{\frac{\Exp{\norm{\delta_x}}}{\cQ_p(\norm{\delta_x})}}^q \leq
    \br{\frac{\sigma_q}{\cQ_p(\norm{\delta_x})}}^q,
\end{equation}
where the last inequality holds for $q>1$ as $\br{\Exp{\norm{\delta_x}}}^q \leq \Exp{\norm{\delta_x}^q}$.
Therefore $\cQ_p(\norm{\delta_x}) \leq \frac{\sigma_q}{\br{1-p}^{1/q}}$.

\subsection{Proof of lemma \ref{adaptive:lemma:second_moment_qc}}
Fix $x\in\RR^d$ and denote $\delta \eqdef \nabla f_{\xi}(x)-\nabla f(x)$ and $h\eqdef 1-p$.
Let $\tau\eqdef \tau(x)$ and $g\eqdef g(x)=\alpha_{\xi}(x)\nabla f_{\xi}(x)$.
By definition of clipping, $\|g\|=\min\{\|\nabla f(x)+\delta\|,\tau\}$, hence
\begin{align*}
    \Exp{\|g\|^2}
    &=
    \Exp{\|\nabla f(x)+\delta\|^2\II\{\|\nabla f(x)+\delta\|\le \tau\}}
    + \tau^2\,\Prob(\|\nabla f(x)+\delta\|>\tau)
    \\&=
    \Exp{\|\nabla f(x)+\delta\|^2\II\{\cdot\}} + \tau^2 h.
\end{align*}
Using Lemma~\ref{common:lem:young} and $\II\{\cdot\}\le 1$ we get
\begin{equation*}
    \Exp{\|\nabla f(x)+\delta\|^2\II\{\cdot\}}
    \le 2\|\nabla f(x)\|^2 + 2\Exp{\|\delta\|^2\II\{\|\nabla f(x)+\delta\|\le \tau\}}.
\end{equation*}
On the event $\{\|\nabla f(x)+\delta\|\le \tau\}$ we have $\|\delta\|\le \tau+\|\nabla f(x)\|$.
Since $q\le 2$, we can write $\|\delta\|^2=\|\delta\|^q\|\delta\|^{2-q}\le \|\delta\|^q(\tau+\|\nabla f(x)\|)^{2-q}$ on this event.
Therefore,
\begin{equation*}
    \Exp{\|\delta\|^2\II\{\|\nabla f(x)+\delta\|\le \tau\}}
    \le (\tau+\|\nabla f(x)\|)^{2-q}\Exp{\|\delta\|^q}
    \le (\tau+\|\nabla f(x)\|)^{2-q}\sigma_q^q.
\end{equation*}
Combining the last two displays yields
\begin{equation*}
    \Exp{\|g\|^2}
    \le 2\|\nabla f(x)\|^2 + 2(\tau+\|\nabla f(x)\|)^{2-q}\sigma_q^q + \tau^2 h.
\end{equation*}
Using Lemma~\ref{adaptive:lemma:key_lemma} we have $\tau\le \|\nabla f(x)\| + \sigma_q h^{-1/q}$, hence
$
\tau+\|\nabla f(x)\|\le 2\|\nabla f(x)\| + \sigma_q h^{-1/q}.
$
Since $2-q\in[0,1]$, $(a+b)^{2-q}\le a^{2-q}+b^{2-q}$ for all $a,b\ge 0$, so
\begin{equation*}
    (\tau+\|\nabla f(x)\|)^{2-q}
    \le (2\|\nabla f(x)\|)^{2-q} + (\sigma_q h^{-1/q})^{2-q}
    = 2^{2-q}\|\nabla f(x)\|^{2-q} + \sigma_q^{2-q}h^{-(2-q)/q}.
\end{equation*}
Moreover, by Lemma~\ref{common:lem:young} and Lemma~\ref{adaptive:lemma:key_lemma},
\begin{equation*}
    \tau^2 h \le 2h\|\nabla f(x)\|^2 + 2\sigma_q^2 h^{1-2/q} = 2h\|\nabla f(x)\|^2 + 2\sigma_q^2 h^{-(2-q)/q}.
\end{equation*}
Finally, the mixed term can be bounded by weighted AM--GM:
\begin{equation*}
\|\nabla f(x)\|^{2-q}\sigma_q^q
= (\|\nabla f(x)\|^2)^{(2-q)/2}(\sigma_q^2)^{q/2}
\le \frac{2-q}{2}\|\nabla f(x)\|^2 + \frac{q}{2}\sigma_q^2.
\end{equation*}

Collecting bounds and using $h\le 1$ and $h^{-(2-q)/q}\ge 1$ gives
$
\Exp{\|g\|^2} \le 6\|\nabla f(x)\|^2 + 8\sigma_q^2 h^{-(2-q)/q}.
$
\qed

\subsection{Proof of lemma \ref{adaptive:lemma:recursion}}
By $L$-smoothness (Assumption \ref{adaptive:ass:l_smooth}) and the update
\begin{equation} \label{adaptive:eq:sgd_qc_full}
    x^{t+1} = x^t - \gamma_t g^t, \qquad g^t = \alpha_{\xi^t}(x^t) \nabla f_{\xi}(x^t) =  \min \cbr{1, \frac{\tau(x^t)}{\norm{\nabla f_{\xi}({x^t})}}} \nabla f_{\xi}(x^t)
\end{equation}
we have
\begin{align}
\Exp{f(x^{t+1})\mid x^t}
&\le f(x^t) - \gamma_t \left\langle \nabla f(x^t), \Exp{g^t\mid x^t}\right\rangle
+ \frac{L\gamma_t^2}{2}\Exp{\|g^t\|^2\mid x^t}.
\end{align}
Add and subtract $\alpha(x^t)\nabla f(x^t)$ inside the inner product and denote
$G_t \eqdef \bigl\|\Exp{g^t\mid x^t}-\alpha(x^t)\nabla f(x^t)\bigr\|$. Then
\begin{align}
\Exp{f(x^{t+1})\mid x^t} \notag
&\le f(x^t) - \gamma_t \alpha(x^t)\|\nabla f(x^t)\|^2
+ \gamma_t \|\nabla f(x^t)\|\,G_t
+ \frac{L\gamma_t^2}{2}\Exp{\|g^t\|^2\mid x^t} \\
&\le f(x^t) - \gamma_t\Bigl(\alpha(x^t)-\frac{\beta}{2}\Bigr)\|\nabla f(x^t)\|^2
+ \frac{\gamma_t}{2}\beta^{-1} G_t^2
+ \frac{L\gamma_t^2}{2}\Exp{\|g^t\|^2\mid x^t},
\end{align}
where the last inequality uses Fenchel--Young with parameter $\beta>0$.

By Lemma \ref{adaptive:lemma:key_lemma} (bias bound), $G_t \le \sigma_q (1-p)^{1-1/q}$, hence
\begin{equation}
\frac{\gamma_t}{2}\beta^{-1}G_t^2 \le \frac{\gamma_t}{2}\beta^{-1}\sigma_q^2(1-p)^{2-2/q}.
\end{equation}
By Lemma \ref{adaptive:lemma:second_moment_qc} (second moment bound) and $h=1-p$,
\begin{equation}
\Exp{\|g^t\|^2\mid x^t} \le 6\|\nabla f(x^t)\|^2 + 8\sigma_q^2 h^{-(2-q)/q}.
\end{equation}
Therefore,
\begin{align}
\Exp{f(x^{t+1})\mid x^t} \notag
&\le f(x^t)
- \gamma_t\Bigl(\alpha(x^t)-\frac{\beta}{2}\Bigr)\|\nabla f(x^t)\|^2
+ \frac{\gamma_t}{2}\beta^{-1}\sigma_q^2 h^{2-2/q} \\
&{}
+ \frac{L\gamma_t^2}{2}\Bigl(6\|\nabla f(x^t)\|^2 + 8\sigma_q^2 h^{-(2-q)/q}\Bigr) \\
&= f(x^t)
- \gamma_t\Bigl(\alpha(x^t)-\frac{\beta}{2}-3L\gamma_t\Bigr)\|\nabla f(x^t)\|^2
\\&{}
+ \frac{\gamma_t}{2}\beta^{-1}\sigma_q^2 h^{2-2/q}
+ 4L\gamma_t^2\sigma_q^2 h^{-(2-q)/q}. \notag
\end{align}
Rearranging the terms yields \eqref{adaptive:eq:core_recursion}.
\qed

\subsection{\texorpdfstring{Proof of theorem \ref{adaptive:thm:non_cvx_gen} (\alg)}{Proof of Theorem \ref{adaptive:thm:non_cvx_gen} (QC-SGD)}}
Let $h\eqdef 1-p$. Taking full expectation in Lemma \ref{adaptive:lemma:recursion} and using $\alpha(x^t)\ge p$ gives
\begin{align}
\Exp{f(x^{t+1})}
&\le \Exp{f(x^t)}
- \gamma_t\Bigl(p-\frac{\beta}{2}-3L\gamma_t\Bigr)\Exp{\|\nabla f(x^t)\|^2}
\\&{}
+ \frac{\gamma_t}{2}\beta^{-1}\sigma_q^2 h^{2-2/q}
+ 4L\gamma_t^2\sigma_q^2 h^{-(2-q)/q}.
\end{align}
Assume $0<\gamma_t \le (2p-\beta-c)/(6L)$ so that
$p-\beta/2-3L\gamma_t \ge c/2$. Then
\begin{equation}
\Exp{f(x^{t+1})}
\le \Exp{f(x^t)}
- \frac{c}{2}\gamma_t\Exp{\|\nabla f(x^t)\|^2}
+ \frac{\gamma_t}{2}\beta^{-1}\sigma_q^2 h^{2-2/q}
+ 4L\gamma_t^2\sigma_q^2 h^{-(2-q)/q}.
\end{equation}
Rearranging yields
\begin{equation}
c\gamma_t\Exp{\|\nabla f(x^t)\|^2}
\le
2\Bigl(\Exp{f(x^t)}-\Exp{f(x^{t+1})}\Bigr)
+ \sigma_q^2\Bigl(\beta^{-1}\gamma_t h^{2-2/q} + 8L\gamma_t^2 h^{-(2-q)/q}\Bigr).
\end{equation}
Summing for $t=0,\dots,T-1$ and using $f(x^T)\ge f^{\inf}$ gives
\begin{equation}
c\sum_{t=0}^{T-1}\gamma_t\Exp{\|\nabla f(x^t)\|^2}
\le
2\bigl(f(x^0)-f^{\inf}\bigr)
+ \sigma_q^2\sum_{t=0}^{T-1}\Bigl(\beta^{-1}\gamma_t h^{2-2/q} + 8L\gamma_t^2 h^{-(2-q)/q}\Bigr).
\end{equation}
Dividing by $\Gamma_T=\sum_{t=0}^{T-1}\gamma_t$ completes the proof of \eqref{adaptive:eq:gen_res}. \qed

%
%
%
%
%
%
%
%
%
%
%
%
%
%
%
%
%
%
%

%
%
%
%
%
%
%
%
%
%
%
%
%
%
%
%
%
%
%
%
%
%
%
%
%
%
%
%
%
%

%
%
%
%
%
%
%
%
%
%
%
%
%
%
%
%
%
%
%
%
%
%
%
%
%

\subsection{\texorpdfstring{Proof of theorem \ref{adaptive:thm:dp_sgd} (\algname{DP-QC-SGD})}{Proof of Theorem \ref{adaptive:thm:dp_sgd} (DP-QC-SGD)}}
Recall the \algname{DP-QC-SGD} update
\begin{equation}
    x^{t+1} = x^t - \gamma_t \underbrace{\frac{1}{B} \br{\sum_{j=1}^B g^t_j + z^t}}_{\tilde{g}^t}, \qquad
    g^t_j = \underbrace{\min \biggl\{1, \frac{\tau(x^t)}{\bigl\|\nabla f_{\xi^t_j}(x^t)\bigr\|}\biggr\}}_{\alpha_{\xi_j^t}(x^t)} \nabla f_{\xi^{t}_j}(x^t),
\end{equation}
where, conditionally on $x^t$, $z^t \sim \mathcal{N}(0, (\tau(x^t))^2 \sigma_{\mathrm{DP}}^2 \mathbf{I})$,
and the noises $\{z^t\}$ are independent across $t$.

By $L$-smoothness,
\begin{equation}
\Exp{f(x^{t+1})\mid x^t}
\le
f(x^t) - \gamma_t\left\langle \nabla f(x^t), \Exp{\tilde g^t\mid x^t}\right\rangle
+ \frac{L\gamma_t^2}{2}\Exp{\|\tilde g^t\|^2\mid x^t}.
\end{equation}
Since $\Exp{z^t}=0$ and the $\xi_j^t$ are i.i.d., we have
$\Exp{\tilde g^t\mid x^t}=\Exp{g(x^t)}$ (the same mean as in the non-private case).
Thus, proceeding exactly as in the proof of Lemma \ref{adaptive:lemma:recursion} (Cauchy--Schwarz and Fenchel--Young),
for any $\beta>0$,
\begin{equation}
\begin{aligned}
\Exp{f(x^{t+1})\mid x^t}
\le
f(x^t)
-\gamma_t\Bigl(\alpha(x^t)-\frac{\beta}{2}\Bigr)\|\nabla f(x^t)\|^2
\\+\frac{\gamma_t}{2}\beta^{-1}\sigma_q^2 h^{2-2/q}
+\frac{L\gamma_t^2}{2}\Exp{\|\tilde g^t\|^2\mid x^t},
\end{aligned}
\end{equation}
where we used Lemma \ref{adaptive:lemma:key_lemma} (bias bound) with $h=1-p$.

Next, we bound the second moment. Conditioned on $x^t$, the noise $z^t$ is independent of
$\{g_j^t\}_{j=1}^B$ and has mean zero, hence the cross term vanishes and
\[
\Exp{\|\tilde g^t\|^2\mid x^t}
=
\Exp{\Bigl\|\frac{1}{B}\sum_{j=1}^B g_j^t\Bigr\|^2\mid x^t}
+
\Exp{\Bigl\|\frac{1}{B}z^t\Bigr\|^2\mid x^t}.
\]
By Jensen and i.i.d.-ness,
\[
\Exp{\Bigl\|\frac{1}{B}\sum_{j=1}^B g_j^t\Bigr\|^2\mid x^t}
\le
\Exp{\|g(x^t)\|^2\mid x^t}
\le
6\|\nabla f(x^t)\|^2 + 8\sigma_q^2 h^{-(2-q)/q},
\]
where the last inequality is Lemma \ref{adaptive:lemma:second_moment_qc}.
Moreover,
\[
\Exp{\Bigl\|\frac{1}{B}z^t\Bigr\|^2\mid x^t}
=
\frac{1}{B^2}\Exp{\|z^t\|^2\mid x^t}
=
\frac{d\sigma_{\mathrm{DP}}^2}{B^2}\,\tau(x^t)^2.
\]
Using Lemma \ref{adaptive:lemma:key_lemma} (quantile bound) and $\|a+b\|^2\le 2\|a\|^2+2\|b\|^2$, we have
\[
\tau(x^t)^2
\le
\Bigl(\|\nabla f(x^t)\|+\sigma_q h^{-1/q}\Bigr)^2
\le
2\|\nabla f(x^t)\|^2 + 2\sigma_q^2 h^{-2/q}.
\]
Therefore, with $\mathfrak{S}\eqdef 1 + d\sigma_{\mathrm{DP}}^2/B^2$,
\begin{align*}
\Exp{\|\tilde g^t\|^2\mid x^t}
&\le
\Bigl(6 + 2(\mathfrak{S}-1)\Bigr)\|\nabla f(x^t)\|^2
+ 8\sigma_q^2 h^{-(2-q)/q}
+ 2(\mathfrak{S}-1)\sigma_q^2 h^{-2/q}.
\end{align*}
Substituting into the descent inequality yields
\begin{align*}
\Exp{f(x^{t+1})\mid x^t}
&\le f(x^t)
-\gamma_t\Bigl(\alpha(x^t)-\frac{\beta}{2}-(2+\mathfrak{S})L\gamma_t\Bigr)\|\nabla f(x^t)\|^2 \\
&{}
+\frac{\gamma_t}{2}\beta^{-1}\sigma_q^2 h^{2-2/q}
+4L\gamma_t^2\sigma_q^2 h^{-(2-q)/q}
+(\mathfrak{S}-1)L\gamma_t^2\sigma_q^2 h^{-2/q}.
\end{align*}
Taking full expectation and using $\alpha(x^t)\ge p$, we get
\begin{align*}
\Exp{f(x^{t+1})}
\le \Exp{f(x^t)}
&-\gamma_t\Bigl(p-\frac{\beta}{2}-(2+\mathfrak{S})L\gamma_t\Bigr)\Exp{\|\nabla f(x^t)\|^2}
+\frac{\gamma_t}{2}\beta^{-1}\sigma_q^2 h^{2-2/q}
\\&{}+4L\gamma_t^2\sigma_q^2 h^{-(2-q)/q}
+(\mathfrak{S}-1)L\gamma_t^2\sigma_q^2 h^{-2/q}.
\end{align*}
Assume $\gamma_t \le (2p-\beta-c)/(2L(2+\mathfrak{S}))$, so that
$p-\beta/2-(2+\mathfrak{S})L\gamma_t\ge c/2$. Rearranging yields
\[
\begin{aligned}
c\gamma_t\Exp{\|\nabla f(x^t)\|^2}
&\leq
2\bigl(\Exp{f(x^t)}-\Exp{f(x^{t+1})}\bigr)\\
&{}+\sigma_q^2\Bigl(
\beta^{-1}\gamma_t h^{2-2/q}
+8L\gamma_t^2 h^{-(2-q)/q}
+2L(\mathfrak{S}-1)\gamma_t^2 h^{-2/q}
\Bigr).
\end{aligned}
\]
Summing over $t=0,\dots,T-1$, using $f(x^T)\ge f^{\inf}$, and dividing by $\Gamma_T$
gives \eqref{adaptive:eq:dp_sgd_res}. \qed

\endgroup

\chapter{Supplementary Material for Chapter 5 (Smoothed Normalization)}
\label{appendix:chapter5_smoothed_normalization}

\begingroup
\input{KAUST_Thesis/imported/05_smoothed_normalization/import_setup}

\section{Preliminaries}

\subsection{Proof of lemma~\ref{smooth:lemma:prop_norm_vr2}}
We prove the first statement by taking the Euclidean norm. Next, we prove the second statement. From the definition of the Euclidean norm,
 \begin{eqnarray*}
     \sqnorm{g -   \beta\Normalize{\alpha}{g}}
     & \overset{\eqref{smooth:eqn:normalize}}{=} & \sqnorm{g} + \frac{\beta^2}{(\alpha+\norm{g})^2}\sqnorm{g} - 2\beta \frac{\sqnorm{g}}{\alpha+\norm{g}} \\
     & = & \left( 1 - \frac{\beta}{\alpha + \norm{g}} \right)^2 \sqnorm{g}.
 \end{eqnarray*}

\subsection{Comparison of EF21 with clipping and smoothed normalization}\label{smooth:app:ef21_clip_norm}

We compare the modified \algname{EF21} mechanism, where a contractive compressor is replaced with clipping in \algname{Clip21} and with smoothed normalization in \alg.
To compare these modified updates, given the optimal vector $g^\star\in\R^d$,  we consider the single-node  \algname{EF21} mechanism, which computes the memory vector $g^k\in\R^d$ according to
\begin{eqnarray}
    g^{k+1} = g^k + \Psi(g^\star - g^k),
\end{eqnarray}
where $\Psi:\R^d\rightarrow\R^d$ is the biased gradient estimator, and $g^0\in\R^d$ is the initial memory vector.

If $\Psi(g) = \Clip{\tau}{g}$, then from Theorem 4.3 of \citet{khirirat2023clip21}
\begin{eqnarray*}
    \norm{g^{k}-g^\star} \leq \max(0, \norm{g^0-g^\star}-k\tau).
\end{eqnarray*}

If $\Psi(g) = \Normalize{\alpha}{g}$, then from Lemma~\ref{smooth:lemma:prop_norm_vr2}
\begin{eqnarray*}
    \sqnorm{g^\star-g^k}
    & = & \sqnorm{g^\star-g^{k-1} -  \beta\Normalize{\alpha}{g^\star - g^{k-1}}}  \\
    & = & \left( 1- \frac{\beta}{\alpha + \norm{g^\star-g^{k-1}}} \right)^2 \sqnorm{g^\star-g^{k-1}}  \\
    & \vdots &   \\
    & = & \sqnorm{g^\star-g^{0}} \cdot \prod_{l=1}^k \left( 1- \frac{\beta}{\alpha + \norm{g^\star-g^{l-1}}} \right)^2.
\end{eqnarray*}

In conclusion, while the \algname{EF21} mechanism with clipping ensures that the memory $g^k$ will reach $g^\star$ within a finite number of iterations $k$ (when $k \geq \norm{g^0-g^\star}/\tau$), the \algname{EF21} mechanism with smoothed normalization guarantees that $g^k$ will eventually reach $g^\star$ (provided that $\beta/\alpha < 1$).

\section{Non-private results}

\subsection{Proof of theorem~\ref{smooth:thm:norm_ef21_nondp}}

\textbf{Proof outline.}
By the $L$-smoothness of the objective function $f$, and by the update for $x^{k+1}$ in \alg, we obtain
\begin{align*}
    V^{k+1}
     \leq   V^k - \gamma \norm{\nabla f(x^k)}+ \frac{L\gamma^2}{2}  +  2\gamma W^k,
\end{align*}
where $V^k \eqdef f(x^k)-f^{\inf}$, and $W^k \eqdef \frac{1}{n}\sum_{i=1}^n \norm{\nabla f_i(x^k)-g_i^{k+1}}$.
The key step to establish the convergence is to bound $\norm{\nabla f_i(x^k) - g_i^{k+1}}$.
Using Lemma~\ref{smooth:lemma:norm21_approach2} and appropriate choices of the tuning parameters $\beta$, $\alpha$, and $\gamma$, we get
\begin{align*}
    \norm{\nabla f_i(x^k)-g_i^{k+1}} \leq \max_{i\in[1,n]}\norm{\nabla f_i(x^0)-g_i^{0}}, \quad \forall k \geq 0.
\end{align*}
Finally, substituting this bound into the previous descent inequality yields the convergence bound in $\min_{k \in [0, K]} \norm{\nabla f(x^k)}$.
Deriving the bound on $\norm{\nabla f_i(x^k) - g_i^{k+1}}$ for \alg by induction is similar to but simpler than \algname{Clip21}. This simplified proof is possible because smoothed normalization possesses a contractive property similar to the contractive compressor used in \algname{EF21}.

We prove Theorem~\ref{smooth:thm:norm_ef21_nondp} using Lemma~\ref{smooth:lemma:norm21_approach2}, which states $\norm{\nabla f_i(x^{k+1})-g_i^{k+1}} \leq R$ for some positive scalar $R$, given that $\norm{\nabla f_i(x^{k})-g_i^{k}} \leq R$ and the hyperparameters $\alpha,\beta,\gamma$ are properly tuned.

\begin{lemma}[Non-private setting]\label{smooth:lemma:norm21_approach2}
Consider \alg (Algorithm~\ref{smooth:alg:norm_norm21_dp_v1})  for solving Problem~\eqref{smooth:eqn:problem}, where Assumption~\ref{smooth:assum:lowebound_f} holds.
If $\norm{\nabla f_i(x^k)-g_i^{k}} \leq R$, $\frac{\beta}{\alpha + R} < 1$, and $\gamma \leq \frac{\beta R}{\alpha + R} \frac{1}{L_{\max}}$ with $L_{\max}=\max_{i\in[1,n]} L_i$, then $  \norm{\nabla f_i(x^{k+1})-g_i^{k+1}}  \leq R$.
\end{lemma}
\begin{proof}
From the definition of the Euclidean norm,
\begin{align*}
    \norm{\nabla f_i(x^{k+1})-g_i^{k+1}}
    &\overset{\text{Lemma~\ref{common:lem:triangle}}}{\leq}
    \norm{\nabla f_i(x^{k+1})-\nabla f_i(x^k)}
    + \norm{\nabla f_i(x^k)-g_i^{k+1}} \\
    &\overset{g_i^{k+1}}{=}
    \norm{\nabla f_i(x^{k+1})-\nabla f_i(x^k)} \\
    &\qquad + \norm{\nabla f_i(x^k)-g_i^{k}
    -\beta\Normalize{\alpha}{\nabla f_i(x^k)-g_i^{k}}} \\
    &\overset{\text{Lemma~\ref{smooth:lemma:prop_norm_vr2}}}{\leq}
    \norm{\nabla f_i(x^{k+1})-\nabla f_i(x^k)} \\
    &\qquad +  \left\vert 1 - \frac{\beta}{\alpha + \norm{\nabla f_i(x^k)-g_i^{k}}} \right\vert
    \norm{\nabla f_i(x^k)-g_i^{k}} \\
    &\overset{\substack{\text{Assumption~\ref{smooth:assum:lowebound_f}}\\\text{and }x^{k+1}}}{\leq}
    L_{\max}\gamma
    +  \left\vert 1 - \frac{\beta}{\alpha + \norm{\nabla f_i(x^k)-g_i^{k}}} \right\vert
    \norm{\nabla f_i(x^k)-g_i^{k}}.
\end{align*}

If $\norm{\nabla f_i(x^k)-g_i^{k}} \leq R$ and $\frac{\beta}{\alpha + R} < 1$, then
$\norm{\nabla f_i(x^{k+1})-g_i^{k+1}}  \leq R$ when
\begin{align*}
    \gamma \leq \frac{\beta R}{\alpha + R} \frac{1}{L_{\max}}.
\end{align*}
\end{proof}

Now, we are ready to prove the result in Theorem~\ref{smooth:thm:norm_ef21_nondp} in four steps.

\paragraph{Step 1) Prove by induction that}
\[
\norm{\nabla f_i(x^k)-g_i^{k}} \leq R
\quad\text{for}\quad
R = \max_{i\in[1,n]}\norm{\nabla f_i(x^0)-g_i^{0}}.
\]
For $k=0$, this is obvious. Next, let
\[
\norm{\nabla f_i(x^l)-g_i^{l}} \leq R
\quad\text{for}\quad
R = \max_{i\in[1,n]}\norm{\nabla f_i(x^0)-g_i^{0}},
\quad l=0,1,\ldots,k.
\]
Then, if $\beta/(\alpha + R) < 1$, and $\gamma \leq \frac{\beta R}{\alpha + R}\frac{1}{L_{\max}}$, then
from Lemma~\ref{smooth:lemma:norm21_approach2} $\norm{\nabla f_i(x^{k+1})-g_i^{k+1}} \leq R$.

\paragraph{Step 2) Bound $\norm{\nabla f_i(x^k)-g_i^{k+1}}$.}

From the definition of the Euclidean norm,
\begin{eqnarray*}
\norm{\nabla f_i(x^k)-g_i^{k+1}}
& \overset{g_i^{k+1}}{=} & \norm{\nabla f_i(x^k) - g_i^{k} -  \beta\Normalize{\alpha}{\nabla f_i(x^k)-g_i^{k} }} \\
& \overset{\text{Lemma~\ref{smooth:lemma:prop_norm_vr2}}}{\leq} & \left\vert 1 - \frac{\beta}{\alpha + \norm{\nabla f_i(x^k)-g_i^{k}}} \right\vert \norm{\nabla f_i(x^k)-g_i^{k}} \\
& \overset{(*)}{\leq} & \left( 1 - \frac{\beta}{\alpha + R } \right) R \leq R,
\end{eqnarray*}
where we reach $(*)$ by the fact that $\norm{\nabla f_i(x^k)-g_i^{k}} \leq R$, $\frac{\beta}{\alpha +R} < 1$, and $\gamma \leq \frac{\beta R}{\alpha+R}\frac{1}{L_{\max}}$.

\paragraph{Step 3) Derive the descent inequality.}
By the $L$-smoothness of $f$, by the definition of $x^{k+1}$, and by the fact that $\hat g^{k+1} = g^{k+1}$,
\begin{align*}
     f(x^{k+1})-f^{\inf}
    &\leq f(x^k)-f^{\inf}
    - \frac{\gamma}{\norm{ g^{k+1}}} \inp{\nabla f(x^k)}{ g^{k+1}}
    + \frac{L\gamma^2}{2} \\
    &= f(x^k)-f^{\inf} - \gamma \norm{ g^{k+1}}
    + \frac{\gamma}{\norm{ g^{k+1}}} \inp{\nabla f(x^k)-g^{k+1}}{g^{k+1}}
    + \frac{L\gamma^2}{2} \\
    &\overset{\text{Lemma~\ref{common:lem:cauchy_schwarz}}}{\leq}
    f(x^k)-f^{\inf} - \gamma \norm{g^{k+1}}
    + \gamma\norm{\nabla f(x^k)-g^{k+1}} + \frac{L\gamma^2}{2} \\
    &\overset{\text{Lemma~\ref{common:lem:triangle}}}{\leq}
    f(x^k)-f^{\inf} - \gamma \norm{\nabla f(x^k)}
    + 2\gamma\norm{\nabla f(x^k)-g^{k+1}} + \frac{L\gamma^2}{2} \\
    &\overset{\text{Lemma~\ref{common:lem:average_norm}}}{\leq}
    f(x^k)-f^{\inf} - \gamma \norm{\nabla f(x^k)}
    + 2\gamma \frac{1}{n}\sum_{i=1}^n \norm{\nabla f_i(x^k)-g_i^{k+1}}
    + \frac{L\gamma^2}{2}.
\end{align*}
Next, since $\norm{\nabla f_i(x^k)-g_i^{k+1}} \leq R$ with $R = \max_{i\in[1,n]}\norm{\nabla f_i(x^0)-g_i^{0}}$, we have
\begin{align*}
    f(x^{k+1})-f^{\inf}
    \leq f(x^k)-f^{\inf} - \gamma \norm{\nabla f(x^k)}
    + 2\gamma \max_{i\in[1,n]} \norm{\nabla f_i(x^0)-g_i^{0}}
    + \frac{L\gamma^2}{2}.
\end{align*}

\paragraph{Step 4) Derive the convergence bound.}
Finally, by rearranging the terms of the inequality,
\begin{align*}
    \min_{k\in[0,K]} \norm{\nabla f(x^k)}
    & \leq  \frac{1}{K+1} \sum_{k=0}^K \norm{\nabla f(x^k)} \\
    & \leq  \frac{[f(x^0)-f^{\inf}]-[f(x^{K+1})-f^{\inf}]}{\gamma(K+1)} + 2\max_{i\in[1,n]} \norm{\nabla f_i(x^0)-g_i^{0}} + \frac{L}{2}\gamma \\
    & \overset{(\dagger)}{\leq} \frac{f(x^0)-f^{\inf}}{\gamma(K+1)} + 2\max_{i\in[1,n]} \norm{\nabla f_i(x^0)-g_i^{0}} + \frac{L}{2}\gamma,
\end{align*}
where $(\dagger)$ follows from $f(x^{K+1}) \geq f^{\inf}$.

\paragraph{Step 5) Finalize the $\cO(1/\sqrt{K})$  rate.}
If $g_i^{0}\in\R^d$ is chosen such that
\[
\max_{i\in[1,n]} \norm{\nabla f_i(x^0)-g_i^{0}}  = \frac{D}{(K+1)^{1/2}}
\]
with any $D>0$, and
\[
\gamma \leq \frac{\beta}{L_{\max}} \frac{D}{\alpha+D} \frac{1}{(K+1)^{1/2}},
\]
then from Theorem~\ref{smooth:thm:norm_ef21_nondp}, we obtain $\beta < \alpha + R$ and $\gamma \leq \frac{\beta R}{\alpha + R}\frac{1}{L_{\max}}$ with $R=\max_{i\in[1,n]}\norm{\nabla f_i(x^0)-g_i^{0}}$, and
\begin{align*}
     \min_{k\in[0,K]} \norm{\nabla f(x^k)}
     &\leq
     \frac{L_{\max} (\alpha + D)}{\beta D}
     \frac{f(x^0)-f^{\inf}}{(K+1)^{1/2}}
     + 2 \frac{D}{(K+1)^{1/2}} \\
     &\qquad
     + \frac{L}{2} \frac{\beta D}{L_{\max}(\alpha+D)} \frac{1}{(K+1)^{1/2}}.
\end{align*}

\subsection{\texorpdfstring{The $\cO(1/\sqrt{K})$ convergence of \alg in the non-private setting}{The O(1/sqrt(K)) convergence of alpha-NormEC in the non-private setting}}\label{smooth:app:derivation_sublinear_nonprivate}

From~\Cref{smooth:thm:norm_ef21_nondp}, we show that \alg achieves an $\cO(1/\sqrt{K})$ convergence rate for the gradient norm, which almost matches the convergence bound for classical gradient descent.
We show this by further choosing $\alpha>1$, and
$
    \beta = \frac{L_{\max}(\alpha+D)}{D}\sqrt{\frac{2(f(x^0)-f^{\inf})}{L}},
$
which ensures $\frac{L_{\max} (\alpha + D)}{\beta D} (f(x^0)-f^{\inf}) = \frac{L}{2} \frac{\beta D}{L_{\max}(\alpha+D)}$,
the associated convergence bound from~\Cref{smooth:thm:norm_ef21_nondp} becomes
\begin{eqnarray}\label{smooth:a_eqn:final_bound_norm21}
     \min_{k\in[0,K]} \norm{\nabla f(x^k)}
     \leq  \frac{\sqrt{2L (f(x^0)-f^{\inf})} + 2D}{(K+1)^{1/2}}.
\end{eqnarray}
This bound comprises two terms. The first term $\sqrt{2L (f(x^0)-f^{\inf})}(K+1)^{-1/2}$ is the convergence bound obtained by classical gradient descent \citep{carmon2020lower}, while the second term $2D(K+1)^{-1/2}$ comes from the initialized memory vectors $g_i^{0}$ for running the error-feedback mechanism.

\subsection{Comparison to \algname{Clip21} }\label{smooth:app:compare_norm21_clip21}

We show that the convergence bound of \alg~\eqref{smooth:a_eqn:final_bound_norm21} has a smaller factor than that of \algname{Clip21} from Theorem 5.6 of \citet{khirirat2023clip21}.

To show this, let $\hat x^K$ be selected uniformly at random from a set $\{x^0,x^1,\ldots,x^K\}$. Then, from Theorem 5.6 of \citet{khirirat2023clip21}, \algname{Clip21} achieves the following convergence bound:
\begin{align*}
    \min_{k\in[0,K]} \norm{\nabla f(x^k)} & \leq& \Exp{\norm{\nabla f(\hat x^K)}}  \\
    & \leq \sqrt{\Exp{\sqnorm{\nabla f(\hat x^K)}}} \\
    & \leq \frac{L_{\max}(f(x^0)-f^{\inf})}{\tau (K+1)^{1/2}}
    + \frac{\sqrt{(1+C_1/\tau)C_2}}{(K+1)^{1/2}},
\end{align*}
where $\tau>0$ is a clipping threshold, and
\[
C_1 = \max_{i\in[1,n]}\norm{\nabla f_i(x^0)},\qquad
C_2 = \max(\max(L,L_{\max})(f(x^0)-f^{\inf})), C_1^2).
\]

If $\tau = \frac{L_{\max}}{\sqrt{2L}}\sqrt{f(x^0)-f^{\inf}}$, then
\begin{eqnarray*}
    \min_{k\in[0,K]} \norm{\nabla f(x^k)}
    & \leq & \sqrt{\frac{2L (f(x^0)-f^{\inf})}{K+1}}  + \frac{\sqrt{ \left(1+ \frac{C_1\sqrt{2L}}{L_{\max}\sqrt{f(x^0)-f^{\inf}}} \right)C_2}}{(K+1)^{1/2}} \\
    & \leq & \sqrt{\frac{2L (f(x^0)-f^{\inf})}{K+1}} \\
    &&+ \frac{\sqrt{C_2} + \cO\left( {\max}(\sqrt{C_1}\sqrt[4]{f(x^0)-f^{\inf}}, C_1^3/\sqrt{f(x^0)-f^{\inf}}) \right) }{(K+1)^{1/2}}.
\end{eqnarray*}
The first term in the convergence bound of \algname{Clip21} matches that of \alg as given in \eqref{smooth:eqn:final_bound_norm21}. However, the second term in the convergence bound of \alg is ${D}/{\sqrt{K+1}}$, where $ D > 0 $ can be made arbitrarily small. In contrast, the corresponding term for \algname{Clip21} is $ C/\sqrt{K+1} $, where $ C > 0 $ may become significantly larger than $D$ if $ x^0 \in \mathbb{R}^d $ is far from the stationary point, leading to a large value of $ C_1 =\max_{i\in[1,n]}\norm{\nabla f_i(x^0)} $.

\subsection{Comparison to \algname{EF21}} \label{smooth:app:alg_and_EF21}
\alg modifies \algname{EF21} to support distributed optimization algorithms for training machine learning models under differential privacy.
However, \alg is not a special case of \algname{EF21}, as the two algorithms rely on fundamentally different biased operator conditions  and convergence analyses.
Specifically, \alg replaces the contractive compression operators used in \algname{EF21} with smoothed normalization. While this modification enables practical advantages in private learning, it introduces significant analytical challenges. The convergence analysis of \algname{EF21} crucially depends on the assumption that the compression operators are contractive, satisfying the condition:
\begin{eqnarray*}
    \psi^{k+1} \leq (1-q) \psi^k, \quad \forall q \in (0, 1],
\end{eqnarray*}
where $\psi^k = \norm{\nabla f_i(x^k)-g_i^k}$ is the gradient estimation error. Nonetheless, this strong contractivity condition does not hold in general for smoothed normalization. Rather, we prove that smoothed normalization satisfies only a contractive-like property, which is insufficient for directly applying the convergence analysis of \algname{EF21}.
To overcome this, we develop a new induction-based proof technique that establishes the weaker monotonicity condition:
\begin{eqnarray*}
    \psi^{k+1} \leq \psi^k,
\end{eqnarray*}
which is sufficient to guarantee  convergence within our novel analysis framework for \alg.

{\color{black}
\subsection{Analysis without server normalization}\label{smooth:app:convg_norm21_no_server_norm}

In this section, we prove convergence in the non-private setting for \alg without server normalization---specifically, in the variant of \Cref{smooth:alg:norm_norm21_dp_v1} where the server update in Step 11 becomes $x^{k+1} = x^k - \gamma g^{k+1}$.

To facilitate our analysis, we impose one additional assumption on the objective function.
\begin{assumption}\label{smooth:assum:bounded_function}
Let $f:\R^d\rightarrow\R$ satisfy $f(x)-f^{\inf} \leq \Delta$ for some $\Delta >0$ and for all $x\in\R^d$.
\end{assumption}

From \Cref{smooth:assum:bounded_function}, we can bound the gradient error norm in the next lemma.
\begin{lemma}\label{smooth:lemma:norm21_no_server_norm}
Consider \alg (Algorithm~\ref{smooth:alg:norm_norm21_dp_v1}) without the server normalization for solving Problem~\eqref{smooth:eqn:problem}, where Assumptions~\ref{smooth:assum:lowebound_f} and~\ref{smooth:assum:bounded_function} hold.
If $\norm{\nabla f_i(x^k)-g_i^{k}} \leq R$, $\frac{\beta}{\alpha + R} < 1$, and $\gamma \leq \frac{\beta R}{\alpha + R} \frac{1}{R + \sqrt{2 L\Delta}} \frac{1}{L_{\max}}$ with $L_{\max}=\max_{i\in[1,n]} L_i$, then $  \norm{\nabla f_i(x^{k+1})-g_i^{k+1}}  \leq R$.
\end{lemma}
\begin{proof}
From the definition of the Euclidean norm,
\begin{align*}
    \norm{\nabla f_i(x^{k+1})-g_i^{k+1}}
    &\overset{\text{Lemma~\ref{common:lem:triangle}}}{\leq}
    \norm{\nabla f_i(x^{k+1})-\nabla f_i(x^k)}
    + \norm{\nabla f_i(x^k)-g_i^{k+1}} \\
    &\overset{g_i^{k+1}}{=}
    \norm{\nabla f_i(x^{k+1})-\nabla f_i(x^k)} \\
    &\qquad + \norm{\nabla f_i(x^k)-g_i^{k}
    -\beta\Normalize{\alpha}{\nabla f_i(x^k)-g_i^{k}}} \\
    &\overset{\text{Lemma~\ref{smooth:lemma:prop_norm_vr2}}}{\leq}
    \norm{\nabla f_i(x^{k+1})-\nabla f_i(x^k)} \\
    &\qquad + \left\vert 1 - \frac{\beta}{\alpha + \norm{\nabla f_i(x^k)-g_i^{k}}} \right\vert
    \norm{\nabla f_i(x^k)-g_i^{k}} \\
    &\overset{\substack{\text{Assumption~\ref{smooth:assum:lowebound_f}}\\\text{and }x^{k+1}}}{\leq}
    L_{\max}\gamma \norm{\hat g^k} \\
    &\qquad +  \left\vert 1 - \frac{\beta}{\alpha + \norm{\nabla f_i(x^k)-g_i^{k}}} \right\vert
    \norm{\nabla f_i(x^k)-g_i^{k}}.
\end{align*}

Next, since
\begin{eqnarray*}
    \norm{\hat g^k}
    & \overset{\text{Lemma~\ref{common:lem:triangle}}}{\leq} & \norm{\nabla f(x^k)-\hat g^k} + \norm{\nabla f(x^k)} \\
    & \overset{\text{Lemma~\ref{common:lem:triangle}}}{\leq} & \frac{1}{n}\sum_{i=1}^n \norm{\nabla f_i(x^k)-\hat g_i^k} + \norm{\nabla f(x^k)} \\
    & \overset{\text{\Cref{smooth:assum:lowebound_f}}}{\leq} & \frac{1}{n}\sum_{i=1}^n \norm{\nabla f_i(x^k)-\hat g_i^k} + \sqrt{2L [f(x^k)-f^{\inf}]} \\
    & \overset{\text{\Cref{smooth:assum:bounded_function}}}{\leq} & \frac{1}{n}\sum_{i=1}^n \norm{\nabla f_i(x^k)-\hat g_i^k} + \sqrt{2L \Delta},
\end{eqnarray*}
by the fact that $\hat g^k = g^k$, we obtain
\begin{eqnarray*}
   \norm{\nabla f_i(x^{k+1})-g_i^{k+1}}
   & \leq & \frac{L_{\max}\gamma}{n}\sum_{i=1}^n \norm{\nabla f_i(x^k)-g_i^k} + L_{\max}\gamma\sqrt{2L \Delta} \\
   && +  \left\vert 1 - \frac{\beta}{\alpha + \norm{\nabla f_i(x^k)-g_i^{k}}} \right\vert \norm{\nabla f_i(x^k)-g_i^{k}}.
\end{eqnarray*}

If $\norm{\nabla f_i(x^k)-g_i^{k}} \leq R$ for all $i$, and $\frac{\beta}{\alpha + R} < 1$, then
$\norm{\nabla f_i(x^{k+1})-g_i^{k+1}}  \leq R$ when
\begin{eqnarray*}
    \gamma \leq \frac{\beta R}{\alpha + R} \frac{1}{R + \sqrt{2 L\Delta}} \frac{1}{L_{\max}}.
\end{eqnarray*}
\end{proof}

From~\Cref{smooth:lemma:norm21_no_server_norm}, we can establish the convergence theorem for \alg without server-side normalization in the non-private setting, similar to the one for \alg with server-side normalization in~\Cref{smooth:thm:norm_ef21_nondp}.

\begin{theorem}\label{smooth:thm:norm21_no_server_norm}
 Consider  \alg (Algorithm~\ref{smooth:alg:norm_norm21_dp_v1}) without server normalization for solving Problem~\eqref{smooth:eqn:problem}, where Assumptions~\ref{smooth:assum:lowebound_f} and~\ref{smooth:assum:bounded_function} hold.
Let $\beta,\alpha,\gamma>0$ be chosen such that
\begin{align*}
 \frac{\beta}{\alpha + R} < 1, \quad \text{and} \quad  \gamma \leq \frac{\beta R}{\alpha + R} \frac{1}{R +\sqrt{2L\Delta}}\frac{1}{L_{\max}},
\end{align*}
where $R = \max_{i\in[1,n]}\norm{\nabla f_i(x^0)-g_i^{0}}$ and $L_{\max}=\max_{i\in[1,n]} L_i$.  Then,
\begin{eqnarray*}
    \min_{k\in[0,K]} \norm{\nabla f(x^k)}
    \leq  \frac{f(x^0)-f^{\inf}}{\gamma(K+1)} + 2R + \frac{L}{2}\gamma.
\end{eqnarray*}
\end{theorem}
\begin{proof}
We prove the result in~\Cref{smooth:thm:norm21_no_server_norm} in four steps.

\paragraph{Step 1) Prove by induction that $\norm{\nabla f_i(x^k)-g_i^{k}} \leq R$ for $R = \max_{i\in[1,n]}\norm{\nabla f_i(x^0)-g_i^{0}}$.}
For $k=0$, this is obvious. Next, let $\norm{\nabla f_i(x^l)-g_i^{l}} \leq R$ for $R = \max_{i\in[1,n]}\norm{\nabla f_i(x^0)-g_i^{0}}$ for $l=0,1,\ldots,k$. Then, if $\beta/(\alpha + R) < 1$, and $\gamma \leq \frac{\beta R}{\alpha + R} \frac{1}{R + \sqrt{2 L\Delta}} \frac{1}{L_{\max}}$, then
from Lemma~\ref{smooth:lemma:norm21_no_server_norm} $\norm{\nabla f_i(x^{k+1})-g_i^{k+1}} \leq R$.

\paragraph{Step 2) Bound $\norm{\nabla f_i(x^k)-g_i^{k+1}}$.}
From the definition of the Euclidean norm,
\begin{eqnarray*}
\norm{\nabla f_i(x^k)-g_i^{k+1}}
& \overset{g_i^{k+1}}{=} & \norm{\nabla f_i(x^k) - g_i^{k} -  \beta\Normalize{\alpha}{\nabla f_i(x^k)-g_i^{k} }} \\
& \overset{\text{Lemma~\ref{smooth:lemma:prop_norm_vr2}}}{\leq} & \left\vert 1 - \frac{\beta}{\alpha + \norm{\nabla f_i(x^k)-g_i^{k}}} \right\vert \norm{\nabla f_i(x^k)-g_i^{k}} \\
& \overset{(*)}{\leq} & \left( 1 - \frac{\beta}{\alpha + R } \right) R \leq R,
\end{eqnarray*}
where we reach $(*)$ by the fact that $\norm{\nabla f_i(x^k)-g_i^{k}} \leq R$, $\frac{\beta}{\alpha +R} < 1$, and $\gamma \leq \frac{\beta R}{\alpha + R} \frac{1}{R + \sqrt{2 L\Delta}} \frac{1}{L_{\max}}$.

Finally, we follow Step 3) and Step 4) by following the proof arguments in~\Cref{smooth:thm:norm_ef21_nondp} to obtain the final result.

\end{proof}
}

{\color{black}
\subsection{Extension to stochastic gradients}\label{smooth:app:norm21_stochastic}

We can leverage our analysis to study \alg, where clients compute their local stochastic gradients $\nabla f_i(x^k;\xi_{i,[b]}^k)= \frac{1}{b}\sum_{j=1}^b\nabla f_i(x^k;\xi_{i,j}^k)$ with the mini-batch size $b$, instead of full local gradients $\nabla f_i(x^k)$, in Step 4 of~\Cref{smooth:alg:norm_norm21_dp_v1}.

We make the following assumption on the clients' local stochastic gradients, which is stronger than the unbiased and variance-bounded conditions of local stochastic gradients.
\begin{assumption}\label{smooth:assum:stoc_grad_noise}
Let $\frac{1}{b}\sum_{j=1}^b \nabla f_i(x;\xi_{i,j})$ be the mini-batch stochastic gradient estimator of the local gradient $\nabla f_i(x)$ at client $i$ such that almost surely for some $\sigma >0$ and for any $b > 0$,
\begin{eqnarray*}
    \norm{ \frac{1}{b}\sum_{j=1}^b \nabla f_i(x;\xi_{i,j}) - \nabla f_i(x) } \leq \frac{\sigma}{b}.%
\end{eqnarray*}
Here, $\xi_{i,1},\ldots, \xi_{i,b}$ are independent and identically distributed random variables drawn from the data distribution $\cD_i$ at client $i$.
\end{assumption}

From~\Cref{smooth:assum:stoc_grad_noise}, we can bound the gradient error norm of \alg for stochastic optimization in the following lemma.
\begin{lemma}\label{smooth:lemma:norm21_stochastic}
Consider \alg (Algorithm~\ref{smooth:alg:norm_norm21_dp_v1}) that uses  stochastic gradients $\nabla f_i(x^k;\xi_{i,[b]}^k)= \frac{1}{b}\sum_{j=1}^b\nabla f_i(x^k;\xi_{i,j}^k)$  for solving Problem~\eqref{smooth:eqn:problem}, where Assumptions~\ref{smooth:assum:lowebound_f} and~\ref{smooth:assum:stoc_grad_noise} hold.
If $\norm{\nabla f_i(x^k;\xi_{i,[b]}^k)-g_i^{k}} \leq R$, $\frac{\beta}{\alpha + R} < 1$, $b \geq \frac{4\sigma (\alpha + R)}{R}$, and $\gamma \leq \frac{1}{L_{\max}} \frac{\beta R}{2(\alpha +R)}$, then $  \norm{\nabla f_i(x^{k+1};\xi_i^{k+1})-g_i^{k+1}}  \leq R$.
\end{lemma}
\begin{proof}
From the definition of the Euclidean norm,
\begin{align*}
    \norm{\nabla f_i(x^{k+1};\xi_{i,[b]}^{k+1})-g_i^{k+1}}
    &\overset{\text{Lemma~\ref{common:lem:triangle}}}{\leq}
    \norm{\nabla f_i(x^{k+1};\xi_{i,[b]}^{k+1})-\nabla f_i(x^{k+1})}  \\
    &\qquad + \norm{\nabla f_i(x^{k};\xi_{i,[b]}^{k})-\nabla f_i(x^{k})} \\
    &\qquad + \norm{\nabla f_i(x^{k+1}) - \nabla f_i(x^k)}
    + \norm{\nabla f_i(x^k;\xi_{i,[b]}^k)-g_i^{k+1}} \\
   &\overset{g_i^{k+1}}{\leq}
    \norm{\nabla f_i(x^{k+1};\xi_{i,[b]}^{k+1})-\nabla f_i(x^{k+1})}  \\
    &\qquad + \norm{\nabla f_i(x^{k};\xi_{i,[b]}^{k})-\nabla f_i(x^{k})} \\
    &\qquad + \norm{\nabla f_i(x^{k+1}) - \nabla f_i(x^k)} \\
    &\qquad + \norm{\nabla f_i(x^k;\xi_{i,[b]}^k)-g_i^k
    - \beta \Normalize{\alpha}{ \nabla f_i(x^k;\xi_{i,[b]}^k)-g_i^k}}.
\end{align*}

From~\Cref{smooth:assum:stoc_grad_noise},
\begin{align*}
    \norm{\nabla f_i(x^{k+1};\xi_{i,[b]}^{k+1})-g_i^{k+1}}
   &\leq \frac{2\sigma}{b} + \norm{\nabla f_i(x^{k+1})-\nabla f_i(x^k)}  \\
   &\qquad + \left\vert 1 - \frac{\beta}{\alpha + \norm{ \nabla f_i(x^k; \xi_{i,[b]}^k) - g_i^k }} \right\vert
   \norm{ \nabla f_i(x^k; \xi_{i,[b]}^k) - g_i^k  } \\
    &\overset{x^{k+1} + \text{\Cref{smooth:assum:lowebound_f}}}{\leq}
    \frac{2\sigma}{b} + L_i \gamma  \\
   &\qquad + \left\vert 1 - \frac{\beta}{\alpha + \norm{ \nabla f_i(x^k; \xi_{i,[b]}^k) - g_i^k }} \right\vert
   \norm{ \nabla f_i(x^k; \xi_{i,[b]}^k) - g_i^k  }.
\end{align*}

If $\beta/(\alpha + R) < 1$, $b \geq \frac{4\sigma (\alpha + R)}{R}$, and $\gamma \leq \frac{1}{L_{\max}} \frac{\beta R}{2(\alpha +R)}$, then  we can prove that $ \norm{\nabla f_i(x^{k+1};\xi_i^{k+1})-g_i^{k+1}}  \leq R$.
\end{proof}

From~\Cref{smooth:lemma:norm21_stochastic}, we can establish the convergence theorem for \alg that uses local stochastic gradients.
\begin{theorem}\label{smooth:thm:norm21_stochastic}
 Consider  \alg (Algorithm~\ref{smooth:alg:norm_norm21_dp_v1}) that uses  stochastic gradients $\nabla f_i(x^k;\xi_{i,[b]}^k)= \frac{1}{b}\sum_{j=1}^b\nabla f_i(x^k;\xi_{i,j}^k)$  for solving Problem~\eqref{smooth:eqn:problem}, where Assumptions~\ref{smooth:assum:lowebound_f} and~\ref{smooth:assum:stoc_grad_noise} hold.
Let $\beta,\alpha,\gamma>0$ be chosen such that
\begin{align*}
 b \geq \frac{4\sigma (\alpha + R)}{R}, \quad \frac{\beta}{\alpha+R} <1 \quad \text{and} \quad \gamma \leq \frac{\beta R}{2(\alpha + R)} \frac{1}{L_{\max}},
\end{align*}
where $R = \max_{i\in[1,n]}\norm{\nabla f_i(x^0;\xi_{i,[b]}^0)-g_i^{0}}$ and $L_{\max}=\max_{i\in[1,n]} L_i$.  Then,  almost surely,
\begin{eqnarray*}
    \min_{k\in[0,K]} \norm{\nabla f(x^k)}
    \leq  \frac{f(x^0)-f^{\inf}}{\gamma(K+1)} + 2\left( \frac{\sigma}{b} + R \right) + \frac{L}{2}\gamma.
\end{eqnarray*}
\end{theorem}
\begin{proof}
We prove the result in~\Cref{smooth:thm:norm21_no_server_norm} in four steps.

\paragraph{Step 1) Prove a uniform bound.}
We prove by induction that
\[
\norm{\nabla f_i(x^k;\xi_{i,[b]}^k)-g_i^{k}} \leq R
\quad
\text{for }
R = \max_{i\in[1,n]}\norm{\nabla f_i(x^0;\xi_{i,[b]}^0)-g_i^{0}}.
\]
For $k=0$, this is obvious. Next, let
\[
\norm{\nabla f_i(x^l;\xi_{i,[b]}^l)-g_i^{l}} \leq R
\quad
\text{for }
R = \max_{i\in[1,n]}\norm{\nabla f_i(x^0;\xi_{i,[b]}^0)-g_i^{0}}
\text{ and } l=0,1,\ldots,k.
\]
Then, from Lemma~\ref{smooth:lemma:norm21_stochastic} $\norm{\nabla f_i(x^{k+1};\xi_{i,[b]}^{k+1})-g_i^{k+1}} \leq R$.

\paragraph{Step 2) Bound $\norm{\nabla f_i(x^k)-g_i^{k+1}}$.}
From the definition of the Euclidean norm,
\begin{align*}
\norm{\nabla f_i(x^k)-g_i^{k+1}}
&\overset{g_i^{k+1}}{=}
\norm{\nabla f_i(x^k) - g_i^{k}
-  \beta\Normalize{\alpha}{\nabla f_i(x^k;\xi_{i,[b]}^k)-g_i^{k} }} \\
&\overset{\text{Lemma~\ref{common:lem:triangle}}}{\leq}
\norm{\nabla f_i(x^k;\xi_{i,[b]}^k) - \nabla f_i(x^k)}  \\
&\qquad + \norm{\nabla f_i(x^k;\xi_{i,[b]}^k) - g_i^k
- \beta\Normalize{\alpha}{\nabla f_i(x^k;\xi_{i,[b]}^k)-g_i^{k} } } \\
&\overset{\text{Lemma~\ref{smooth:lemma:prop_norm_vr2}}}{\leq}
\norm{\nabla f_i(x^k;\xi_{i,[b]}^k) - \nabla f_i(x^k)}   \\
&\qquad + \left\vert 1 - \frac{\beta}{\alpha + \norm{\nabla f_i(x^k;\xi_{i,[b]}^k)-g_i^{k}}} \right\vert
\norm{\nabla f_i(x^k;\xi_{i,[b]}^k)-g_i^{k}}.
\end{align*}

From~\Cref{smooth:assum:stoc_grad_noise},
\begin{eqnarray*}
\norm{\nabla f_i(x^k)-g_i^{k+1}}
& \leq & \frac{\sigma}{b}   + \left\vert 1 - \frac{\beta}{\alpha + \norm{\nabla f_i(x^k;\xi_{i,[b]}^k)-g_i^{k}}} \right\vert \norm{\nabla f_i(x^k;\xi_{i,[b]}^k)-g_i^{k}}.
\end{eqnarray*}

By the fact that $\norm{\nabla f_i(x^k;\xi_{i,[b]}^k)-g_i^{k}} \leq R$, and $\beta/(\alpha +R) < 1$ from Step 1) of the proof,
\begin{eqnarray*}
\norm{\nabla f_i(x^k)-g_i^{k+1}}
& \leq & \frac{\sigma }{b}  + \left( 1 - \frac{\beta}{\alpha +R} \right)R \leq \frac{\sigma}{b} +R.
\end{eqnarray*}

Finally, we follow Step 3) and Step 4) by following the proof arguments in~\Cref{smooth:thm:norm_ef21_nondp} to obtain the final result.
\end{proof}

\Cref{smooth:thm:norm21_stochastic} states that \alg using local stochastic gradients achieves the sublinear, almost sure convergence up to the additive constants $2(\sigma/b+ R)+\gamma L/2$. Here, the mini-batch size $b$ must be sufficiently large to ensure the convergence and to improve the accuracy of the solution obtained from running the method.
}

\section{Private results}

\subsection{Convergence theorem in the DP setting}

\begin{theorem}[DP setting]\label{smooth:app:thm:dp_n_norm21}
Consider \algname{DP-}\alg (Algorithm~\ref{smooth:alg:norm_norm21_dp_v1})  for solving Problem~\eqref{smooth:eqn:problem}, where Assumption~\ref{smooth:assum:lowebound_f} holds.
Let %
$\beta,\alpha,\gamma > 0$ be chosen such that
\begin{eqnarray*}
    \frac{\beta}{\alpha + R } < 1, \quad \text{and} \quad \gamma \leq \frac{\beta R}{\alpha+R}\frac{1}{L_{\max}},
\end{eqnarray*}
where $R = \max_{i\in[1,n]}\norm{\nabla f_i(x^0) - g_i^{0}}$, and $L_{\max} = \max_{i\in[1,n]} L_i$. Then,
\begin{align*}
    \min_{k\in[0,K]} \Exp{\norm{\nabla f(x^k)}} \leq  \frac{f(x^0)-f^{\inf}}{\gamma(K+1)} + 2R + \frac{L}{2}\gamma
     + 2 \sqrt{\frac{\beta^2 (K+1)\sigma_{\rm DP}^2}{n}}.
\end{align*}
\end{theorem}

\subsection{Proof of theorem~\ref{smooth:app:thm:dp_n_norm21}}

We prove Theorem~\ref{smooth:app:thm:dp_n_norm21} by  two useful lemmas:
\begin{enumerate}
    \item Lemma~\ref{smooth:lemma:norm21_approach2}, which states  $\norm{\nabla f_i(x^{k+1})-g_i^{k+1}} \leq R$  for some positive scalars $R$, given that $\norm{\nabla f_i(x^{k})-g_i^{k}} \leq R$ and the hyperparameters $\gamma,\beta,\alpha$ are properly tuned, and
    \item Lemma~\ref{smooth:lemma:dp_norm21_weight_tricks}, which bounds the difference in expectation between the memory vectors maintained by the central server and clients.
\end{enumerate}

\begin{lemma}[DP setting]\label{smooth:lemma:dp_norm21_weight_tricks}
Consider \algname{DP-}\alg (Algorithm~\ref{smooth:alg:norm_norm21_dp_v1})  for solving Problem~\eqref{smooth:eqn:problem}, where Assumption~\ref{smooth:assum:lowebound_f} holds.
If $\hat g^0 = \frac{1}{n}\sum_{i=1}^n g_i^0$, then
\begin{eqnarray*}
\Exp{\norm{\hat g^{k+1} - \frac{1}{n}\sum_{i=1}^n g^{k+1}_i}}  \leq \sqrt{{\color{black}\frac{{\beta^2(K+1) \sigma_{\rm DP}^2}}{n}} }.
\end{eqnarray*}
\end{lemma}
\begin{proof}
From the definition of $g^k_i$ and $\hat g^k$,
\begin{eqnarray*}
    e^{k+1} = e^{k} + \beta z^{k+1},
\end{eqnarray*}
where $e^k = \hat g^k - \frac{1}{n}\sum_{i=1}^n g^k_i$, and $z^k = \frac{1}{n}\sum_{i=1}^n z_i^k$.
By applying the equation recursively,
\begin{eqnarray*}
    e^{k+1} = e^{0} + \beta \sum_{l=1}^{k+1} z^l.
\end{eqnarray*}
Therefore, by the triangle inequality,
\begin{eqnarray*}
    \norm{e^{k+1}} \leq \norm{e^{0}} +  \norm{\beta \sum_{l=1}^{k+1} z^l}.
\end{eqnarray*}

If $\hat g^{0}= \frac{1}{n}\sum_{i=1}^n g_i^{0}$, then $e^0=0$ and therefore
\begin{eqnarray*}
    \norm{e^{k+1}} \leq   \norm{\beta \sum_{l=1}^{k+1} z^l}.
\end{eqnarray*}
By taking the expectation,
\begin{eqnarray*}
    \Exp{\norm{e^{k+1}}} &\leq& \Exp{\norm{\beta \sum_{l=1}^{k+1} z^l}} \\
    & = & \Exp{\sqrt{ \sqnorm{\beta \sum_{l=1}^{k+1} z^l}  }} \\
    & {\leq} & \sqrt{\Exp{ \sqnorm{\beta \sum_{l=1}^{k+1} z^l}  }},
\end{eqnarray*}
where we reach the last inequality by Jensen's inequality. Next, by expanding the terms,
\begin{eqnarray*}
    \Exp{\norm{e^{k+1}}}
    & \leq & \sqrt{ \beta^2 \sum_{l=1}^{k+1} \Exp{\sqnorm{z^l}} + \beta^2 \sum_{j\neq i} \Exp{\inp{z^i}{z^j}}  } \\
    & \overset{(*)}{=} &\sqrt{ \beta^2 \sum_{l=1}^{k+1} \Exp{\sqnorm{z^l}}} \\
    & \overset{(\ddagger)}{\leq} & \sqrt{\frac{\beta^2}{n} \sum_{l=1}^{k+1}\sigma_{\rm DP}^2 },
\end{eqnarray*}
where we reach $(*)$ by the fact that  $\Exp{\langle z^j , z^i\rangle}=0$ for $i \neq j$, and $(\ddagger)$ by the fact that $\Exp{\sqnorm{z^k}} = {\sigma_{\rm DP}^2}/{n}$ (as $z_i^k$ is independent of $z_j^k$ for $i\neq j$).
Therefore,
\begin{eqnarray*}
\Exp{\norm{e^{k+1}}}
& \leq & \sqrt{\frac{\beta^2 (k+1)\sigma_{\rm DP}^2}{n}} \\
& \overset{k \leq K}{\leq} & \sqrt{ \frac{\beta^2(K+1) \sigma_{\rm DP}^2}{n} }.
\end{eqnarray*}
\end{proof}

Now, we prove~\Cref{smooth:app:thm:dp_n_norm21} in three steps.

\paragraph{Step 1) Prove by induction that $\norm{\nabla f_i(x^k)-g_i^{k}} \leq R$ for $R = \max_{i\in[1,n]}\norm{\nabla f_i(x^0)-g_i^{0}}$.} For $k=0$, this is obvious. Next, let $\norm{\nabla f_i(x^l)-g_i^{l}} \leq R$ for $R = \max_{i\in[1,n]}\norm{\nabla f_i(x^0)-g_i^{0}}$ for $l=0,1,\ldots,k$. Then, if $\beta/(\alpha + R) < 1$, and $\gamma \leq \frac{\beta R}{\alpha + R}\frac{1}{L_{\max}}$, then
from Lemma~\ref{smooth:lemma:norm21_approach2} $\norm{\nabla f_i(x^{k+1})-g_i^{k+1}} \leq R$.

\paragraph{Step 2) Bound $\norm{\nabla f_i(x^k)-g_i^{k+1}}$.}

From the definition of the Euclidean norm,
\begin{eqnarray*}
\norm{\nabla f_i(x^k)-g_i^{k+1}}
& \overset{g_i^{k+1}}{=} & \norm{\nabla f_i(x^k) - g_i^{k} -  \beta\Normalize{\alpha}{\nabla f_i(x^k)-g_i^{k} }} \\
& \overset{\text{Lemma~\ref{smooth:lemma:norm21_approach2}}}{\leq} & \left\vert 1 - \frac{\beta}{\alpha + \norm{\nabla f_i(x^k)-g_i^{k}}} \right\vert \norm{\nabla f_i(x^k)-g_i^{k}}.
\end{eqnarray*}

\paragraph{Step 3) Derive the descent inequality in $\Exp{f(x^k)-f^{\inf}}$.}
Denote $g^k = \frac{1}{n}\sum_{i=1}^n g_i^k$. By the $L$-smoothness of $f$, and by the definition of $x^{k+1}$,
\begin{align*}
     f(x^{k+1})-f^{\inf}
    &\leq f(x^k)-f^{\inf}
    - \frac{\gamma}{\norm{\hat g^{k+1}}} \inp{\nabla f(x^k)}{\hat g^{k+1}}
    + \frac{L\gamma^2}{2} \\
    &= f(x^k)-f^{\inf} - \gamma \norm{\hat g^{k+1}}
    + \frac{\gamma}{\norm{\hat g^{k+1}}} \inp{\nabla f(x^k)-\hat g^{k+1}}{\hat g^{k+1}}
    + \frac{L\gamma^2}{2} \\
    &\overset{\text{Lemma~\ref{common:lem:cauchy_schwarz}}}{\leq}
    f(x^k)-f^{\inf} - \gamma \norm{\hat g^{k+1}}
    + \gamma\norm{\nabla f(x^k)-\hat g^{k+1}} + \frac{L\gamma^2}{2} \\
    &\overset{\text{Lemma~\ref{common:lem:triangle}}}{\leq}
    f(x^k)-f^{\inf} - \gamma \norm{\nabla f(x^k)}
    + 2\gamma\norm{\nabla f(x^k)-\hat g^{k+1}} + \frac{L\gamma^2}{2} \\
    &\overset{\text{Lemma~\ref{common:lem:average_norm}}}{\leq}
    f(x^k)-f^{\inf} - \gamma \norm{\nabla f(x^k)}
    + 2\gamma \frac{1}{n}\sum_{i=1}^n \norm{\nabla f_i(x^k)-g_i^{k+1}} \\
    &\qquad + 2\gamma\norm{\hat g^{k+1}- g^{k+1}} + \frac{L\gamma^2}{2}.
\end{align*}
Next, let $\cF^k$ be the history up to iteration $k$, i.e. $\cF^k \eqdef \{x^0,z_1^0,\ldots,z_n^0,\ldots,x^k,z_1^k,\ldots,z_n^k\}$. Then,
\begin{align*}
    \ExpCond{f(x^{k+1})-f^{\inf}}{\cF^k}
    &\leq f(x^k)-f^{\inf} - \gamma \norm{\nabla f(x^k)}
    + 2\gamma \frac{1}{n}\sum_{i=1}^n
    \ExpCond{\norm{\nabla f_i(x^k)-g_i^{k+1}}}{\cF^k} \\
   &\qquad + 2\gamma \ExpCond{\norm{\hat g^{k+1}- g^{k+1}}}{\cF^k}
   + \frac{L\gamma^2}{2}.
\end{align*}
Next, by the upper-bound for $\norm{\nabla f_i(x^k)-g_i^{k+1}}$,
\begin{eqnarray*}
    \ExpCond{\norm{\nabla f_i(x^k)-g_i^{k+1}}}{\cF^k}
    & \leq &  \ExpCond{\left\vert 1 - \frac{\beta}{\alpha + \norm{\nabla f_i(x^k)-g_i^{k}}} \right\vert \norm{\nabla f_i(x^k)-g_i^{k}}}{\cF^k}  \\
    & = & \left\vert 1 - \frac{\beta}{\alpha + \norm{\nabla f_i(x^k)-g_i^{k}}} \right\vert \norm{\nabla f_i(x^k)-g_i^{k}} \\
    & \leq & \left(1- \frac{\beta}{\alpha+R} \right) R \leq R,
\end{eqnarray*}
where we reach the second inequality by the fact that $\norm{\nabla f_i(x^k)-g_i^{k}} \leq R$, $\frac{\beta}{\alpha +R} < 1$, and $\gamma \leq \frac{\beta R}{\alpha+R}\frac{1}{L_{\max}}$. Thus,
\begin{eqnarray*}
    \ExpCond{f(x^{k+1})-f^{\inf}}{\cF^k}
    & \leq & f(x^k)-f^{\inf} - \gamma \norm{\nabla f(x^k)} + 2\gamma R \\
   && + 2\gamma \ExpCond{\norm{\hat g^{k+1}- g^{k+1}}}{\cF^k} + \frac{L\gamma^2}{2}.
\end{eqnarray*}
By taking the expectation, and by the tower property $\Exp{\ExpCond{X}{Y}}=\Exp{X}$,
\begin{eqnarray*}
    \Exp{f(x^{k+1})-f^{\inf}}
    & = & \Exp{ \ExpCond{f(x^{k+1})-f^{\inf}}{\cF^k} } \\
    & \leq & \Exp{f(x^k)-f^{\inf}} - \gamma \Exp{\norm{\nabla f(x^k)}} + 2\gamma R  \\
    && + 2\gamma \Exp{\norm{\hat g^{k+1}- g^{k+1}}} + \frac{L\gamma^2}{2}.
\end{eqnarray*}
Next, by using Lemma~\ref{smooth:lemma:dp_norm21_weight_tricks},
\begin{eqnarray*}
  \Exp{ f(x^{k+1})-f^{\inf} } & \leq &   \Exp{f(x^k)-f^{\inf}} - \gamma \Exp{\norm{\nabla f(x^k)}} + 2\gamma R \\
  && + 2\gamma {\color{black}\sqrt{ \frac{\beta^2(K+1) \sigma_{\rm DP}^2}{n} }}+ \frac{L\gamma^2}{2}.
\end{eqnarray*}
Therefore,
\begin{eqnarray*}
   && \min_{k\in[0,K]} \Exp{\norm{\nabla f(x^k)}}
     \leq  \frac{1}{K+1} \sum_{k=0}^K \Exp{\norm{\nabla f(x^k)}} \\
    && \leq  \frac{\Exp{f(x^0)-f^{\inf}}-\Exp{f(x^{K+1})-f^{\inf}}}{\gamma(K+1)}  + 2 R + 2 {\color{black}\sqrt{ \frac{\beta^2(K+1) \sigma_{\rm DP}^2}{n} }}+ \frac{L}{2}\gamma \\
    && \leq  \frac{f(x^0)-f^{\inf}}{\gamma(K+1)}   + 2 R+ 2 {\color{black}\sqrt{ \frac{\beta^2(K+1) \sigma_{\rm DP}^2}{n} }} + \frac{L}{2}\gamma,
\end{eqnarray*}
where the last inequality follows from $f(x^{K+1}) \geq f^{\inf}$.

\subsection{\texorpdfstring{Discussion of~\Cref{smooth:app:thm:dp_n_norm21}}{Discussion of Theorem~\ref{smooth:app:thm:dp_n_norm21}}} \label{smooth:app:dp_convergence}

By choosing $g_i^{0}$ such that $R = \frac{D}{(K+1)^{1/6}}$ with any $D > 0$, $\beta = \frac{\beta_0}{(K+1)^{2/3}}$ with $\beta_0\in (0,1]$, $\alpha > 1$, and $\gamma \leq \frac{A}{(K+1)^{5/6}}$ with $A=\frac{\beta_0 D}{L_{\max}(\alpha + D)}$, then the conditions for $\beta,\alpha,\gamma$ in Theorem~\ref{smooth:app:thm:dp_n_norm21} are satisfied, and from Theorem~\ref{smooth:app:thm:dp_n_norm21},~\algname{DP-}\alg attains the $\cO(1/K^{1/6})$ convergence rate in the gradient norm:
\begin{eqnarray*}
     \min_{k\in[0,K]} \Exp{\norm{\nabla f(x^k)}} \leq   \frac{C}{(K+1)^{1/6}} + \frac{LA}{2 (K+1)^{5/6}},
\end{eqnarray*}
where $C_1 = \frac{f(x^0)-f^{\inf}}{A} + 2D + 2\beta_0 \sigma_{\rm DP}$.

\subsection{\texorpdfstring{Utility guarantee of~\algname{DP-}\alg (\Cref{smooth:thm:dp_n_norm21})}{Utility guarantee of DP-alpha-NormEC (Theorem~\ref{smooth:thm:dp_n_norm21})}}\label{smooth:app:dp_n_norm21}

In this section, we present the utility guarantee of ~\algname{DP-}\alg (\Cref{smooth:thm:dp_n_norm21}).

\begin{corollary}[Utility guarantee in DP setting]\label{smooth:corr:dp_n_norm21}
Consider \algname{DP-}\alg (Algorithm~\ref{smooth:alg:norm_norm21_dp_v1})  for solving Problem~\eqref{smooth:eqn:problem} under the same setting as  Theorem~\ref{smooth:thm:dp_n_norm21}. If $\sigma_{\rm DP} = \cO(\sqrt{(K+1)\log(1/\delta)}\epsilon^{-1})$,  $\beta = \frac{\beta_0}{K+1}$
{\color{black}
with $\beta_0 = \cO\left( \sqrt{\frac{\Delta}{A}} \right)$ and $\alpha = R = \cO\left( \sqrt[4]{d}\sqrt{  \Delta A} \right)$ for $\Delta = \sqrt{ L_{\max} (f(x^0)-f^{\inf})}$ and $A = \frac{\sqrt{\log(1/\delta)}}{\sqrt{n}\epsilon}$, then Algorithm~\ref{smooth:alg:norm_norm21_dp_v1} satisfies $(\epsilon,\delta)$-DP and attains the bound
\begin{align*}
  \min_{k\in[0,K]} \Exp{\norm{\nabla f(x^k)}}
 \leq \cO\left(  \Delta \sqrt[4]{d\frac{\log(1/\delta)}{n \epsilon^2}} \right).
\end{align*}
}
\end{corollary}

\textbf{Proof:}
Let $\sigma_{\rm DP} = \cO\left( \frac{\sqrt{(K+1)\log(1/\delta)}}{\epsilon} \right)$,
and  $\beta = \frac{\beta_0}{K+1}$ with $0<\beta_0<\alpha + R$.
Then, from Theorem~\ref{smooth:thm:dp_n_norm21}, we get $\gamma \leq \frac{\beta_0 R}{\alpha + R}\frac{1}{L_{\max}} \frac{1}{K+1}$ with $R = \max_{i\in[1,n]}\norm{\nabla f_i(x^0)-g_i^{0}}$, and
\begin{eqnarray*}
  \min_{k\in[0,K]} \Exp{\norm{\nabla f(x^k)}}
  &\leq & \frac{L_{\max}(\alpha +R)(f(x^0)-f^{\inf})}{\beta_0 R}  + 2 R + 2\frac{\beta_0  \sqrt{\log(1/\delta)}}{ {\color{black}\sqrt{n}} \epsilon} \\
  && + \frac{L \beta_0 R}{2(\alpha + R)L_{\max}} \frac{1}{K+1}.
\end{eqnarray*}

{\color{black}
If $\beta_0 = \cO\left( \sqrt{\frac{L_{\max} (f(x^0)-f^{\inf})}{A}} \right)$ and $\alpha = R = \cO\left( \sqrt[4]{d}\sqrt{  L_{\max} (f(x^0) - f^{\inf})A} \right)$ for $A = \frac{\sqrt{\log(1/\delta)}}{\sqrt{n}\epsilon}$, then
\begin{eqnarray*}
  \min_{k\in[0,K]} \Exp{\norm{\nabla f(x^k)}}
  &\leq & \cO\left(  \sqrt[4]{d}\sqrt{ L_{\max} (f(x^0)-f^{\inf}) A} \right).
\end{eqnarray*}
}

\subsection{Private initialization of the memory vectors}\label{smooth:app:private_initial}

Secure aggregation can be used in \alg to initialize the server's aggregated memory vector, $\hat g^0 = \frac{1}{n}\sum_{i=1}^n g_i^0$, while keeping the clients' individual memory vectors, $g_i^0$, private.
For example, clients that share a random seed can add and subtract identical cryptographic noise, $h$, from their respective vectors (e.g., $g_1^0 + h$ and $g_2^0 - h$).
When these are aggregated by the server, the noise cancels out, preserving the accuracy of the average: $\frac{1}{2}(g_1^0 + h) + \frac{1}{2}(g_2^0 - h) = \frac{1}{2}(g_1^0 + g_2^0)$.
This method protects the individual vectors from the server without compromising the accuracy of the overall average.

Furthermore,
we can extend this initialization for $\hat g^0$ to include general DP noise. Specifically, we can set $\hat g^0 = \frac{1}{n}\sum_{i=1}^n (g_i^0 + e_i^0)$, where $e_i^0$ is the DP noise that Client $i$ adds to its local memory vector before communicating it to the server. Our analysis, especially in~\Cref{smooth:lemma:dp_norm21_weight_tricks}, shows that this generalized initialization can be accommodated. The worst-case bound for the error term, $\Exp{\norm{\hat g^{k+1} - \frac{1}{n}\sum_{i=1}^n g^{k+1}_i}}$, will simply include an additional term, $e = \frac{1}{n}\sum_{i=1}^n e_i^0$, i.e.
\begin{eqnarray*}
    \Exp{\norm{\hat g^{k+1} - \frac{1}{n}\sum_{i=1}^n g^{k+1}_i}} \leq \sqrt{\frac{\beta^2 (K+1)}{n}\sigma_{\rm DP}^2} + \sqrt{ \frac{1}{n}\sum_{i=1}^n \norm{e_i^0}}.
\end{eqnarray*}

{
\subsection{Discussion of concurrent work}
A concurrent work by~\citet{islamov2025double} introduces \algname{Clip21-SGD2M}, which utilizes double momentum to address the convergence issues of \algname{Clip21} in the stochastic setting.
However, \alg and \algname{Clip21-SGD2M} are built upon fundamentally different operators:
\alg employs smoothed normalization, while \algname{Clip21-SGD2M} uses traditional clipping.
Thanks to its state-dependent contractive-like property \eqref{smooth:eq:norm_2}, smoothed normalization allows for significantly simpler analysis than method with clipping.
In addition, \algname{DP-}\alg attains the utility bound in expectation, which is more standard than the high-probability utility bound of \algname{DP-Clip21-SGD2M} of \citet{islamov2025double}.
}

\section{Experimental details and additional results} \label{smooth:app:experiments}

We include details on experimental setups and additional results in the non-private and private training for the ResNet20 model on the CIFAR-10 dataset.

\subsection{Additional experimental details}
The dataset was split into train (90\%) and test (10\%) sets. The train samples were randomly shuffled and distributed across 10 workers. Every worker computed gradients with batch size 32. The training was performed for 300 communication rounds. The random seed was fixed to 42 for reproducibility.

All the methods were run with a constant step size (learning rate) without other techniques, such as schedulers, warm-up, or weight decay.
They were evaluated across the following hyperparameter combinations:
\begin{itemize}
    \item step size $\gamma$: $\{0.001, 0.01, 0.1, 1.0\}$,
    \item Sensitivity/clip threshold $\beta$: $\{0.01, 0.1, 1.0, 10.0\}$,
    \item Smoothed normalization value $\alpha$: $\{0.01, 0.1, 1.0\}$.
\end{itemize}

Our implementation is based on the public GitHub repository \citep{idelbayev18a}. Experiments were performed on a machine with a single GPU: NVIDIA GeForce RTX 3090.

\subsection{Non-private training}

\subsubsection{Sensitivity of \alg to parameters $\beta, \alpha$} \label{smooth:app:exp_sensitivity}
\begin{figure}[h]
\centering
\begin{minipage}[c]{0.32\textwidth}
    \centering
    \includegraphics[width=\linewidth]{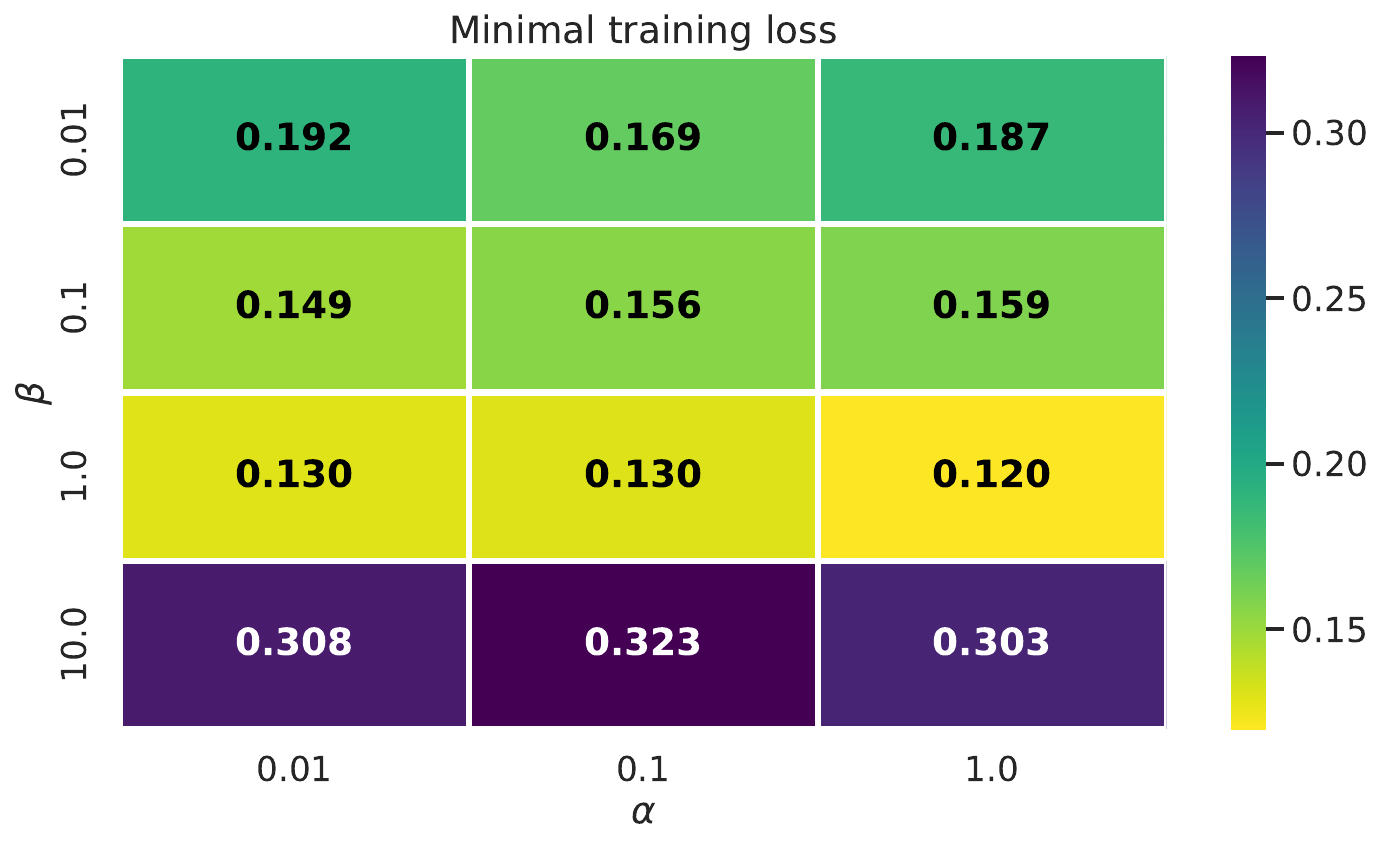}
    \label{smooth:fig:loss_heatmap}
\end{minipage}
\hfill
\begin{minipage}[c]{0.32\textwidth}
    \centering
    \includegraphics[width=\linewidth]{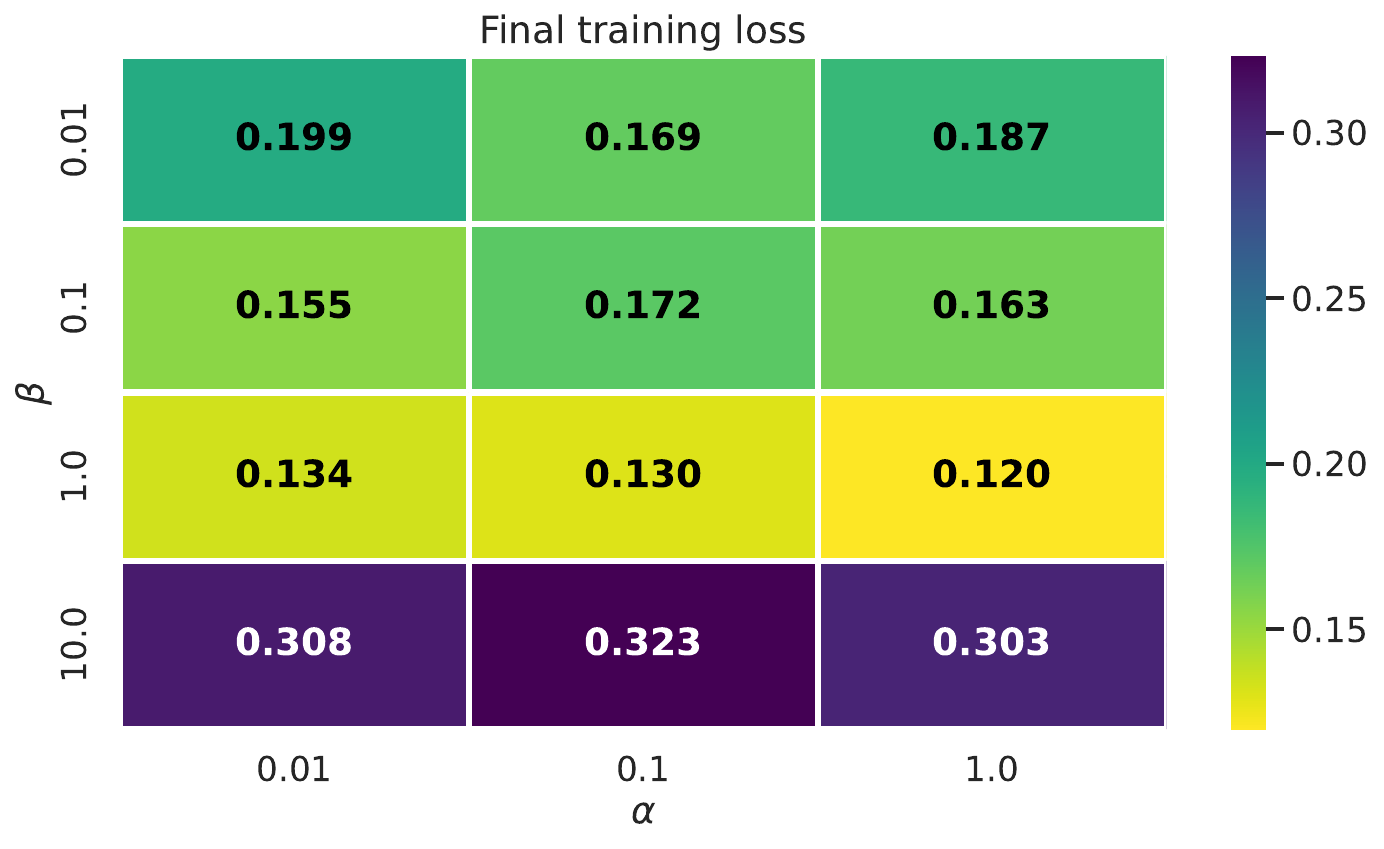}
    \label{smooth:fig:final_accuracy_heatmap}
\end{minipage}
\hfill
\begin{minipage}[c]{0.32\textwidth}
    \centering
    \includegraphics[width=\linewidth]{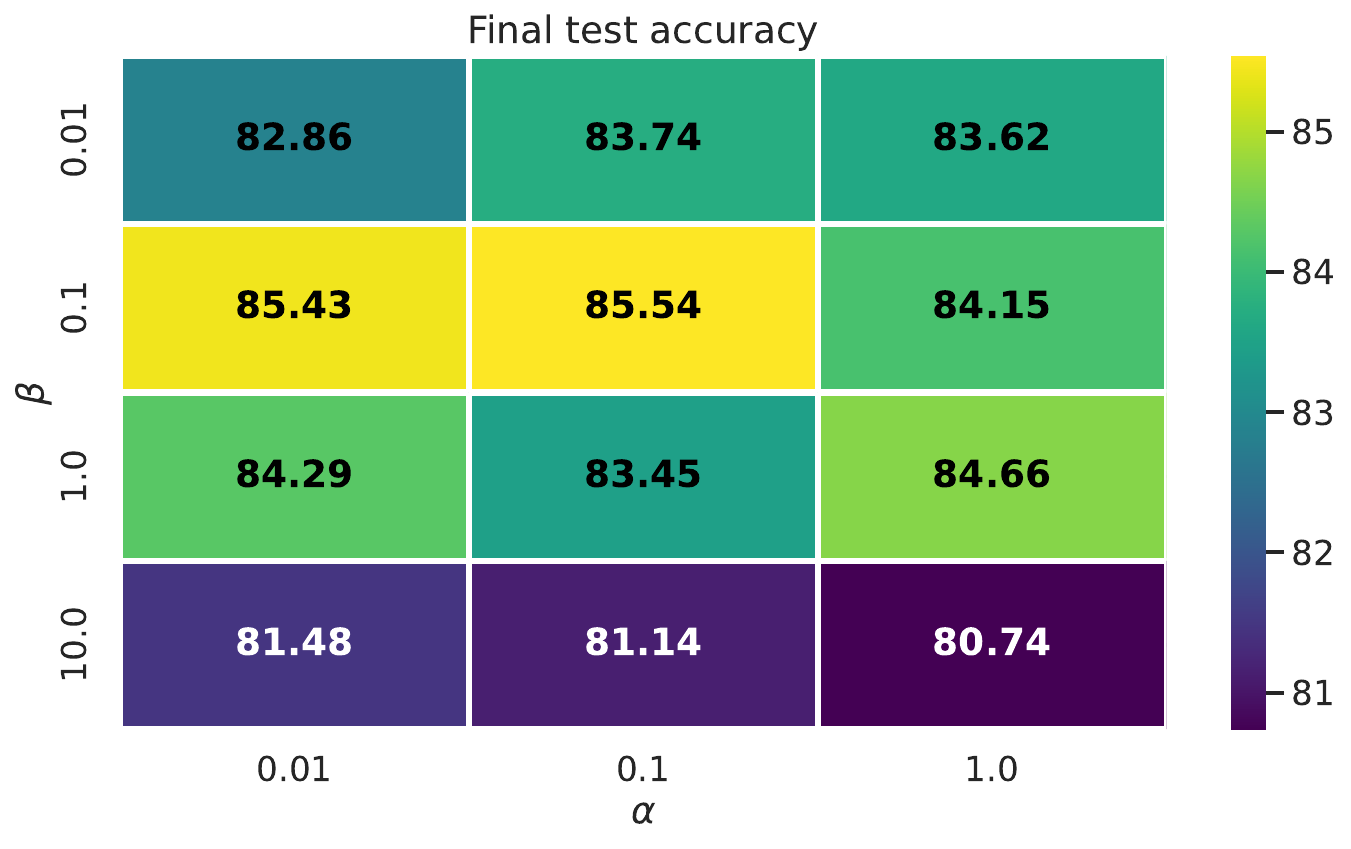}
    \label{smooth:fig:final_loss_heatmap}
\end{minipage}
\caption{\textbf{Minimal} train loss (left), \textbf{final} train loss (middle), and \textbf{final} test accuracy (right) achieved by non-private \alg, after 300 communication rounds using a fine-tuned constant step size $\gamma$.}
\label{smooth:fig:heatmap_non_private_other}
\end{figure}
Figure~\ref{smooth:fig:heatmap_non_private_other} supplements the results in Figure \ref{smooth:fig:accuracy_heatmap} with other metrics.
Figure \ref{smooth:fig:norm21_param_convergence} displays convergence curves across different combinations of $\alpha, \beta$ parameters with optimally selected step sizes $\gamma$.
\begin{figure}[h]
\begin{center}
\centerline{\includegraphics[width=0.9\linewidth]{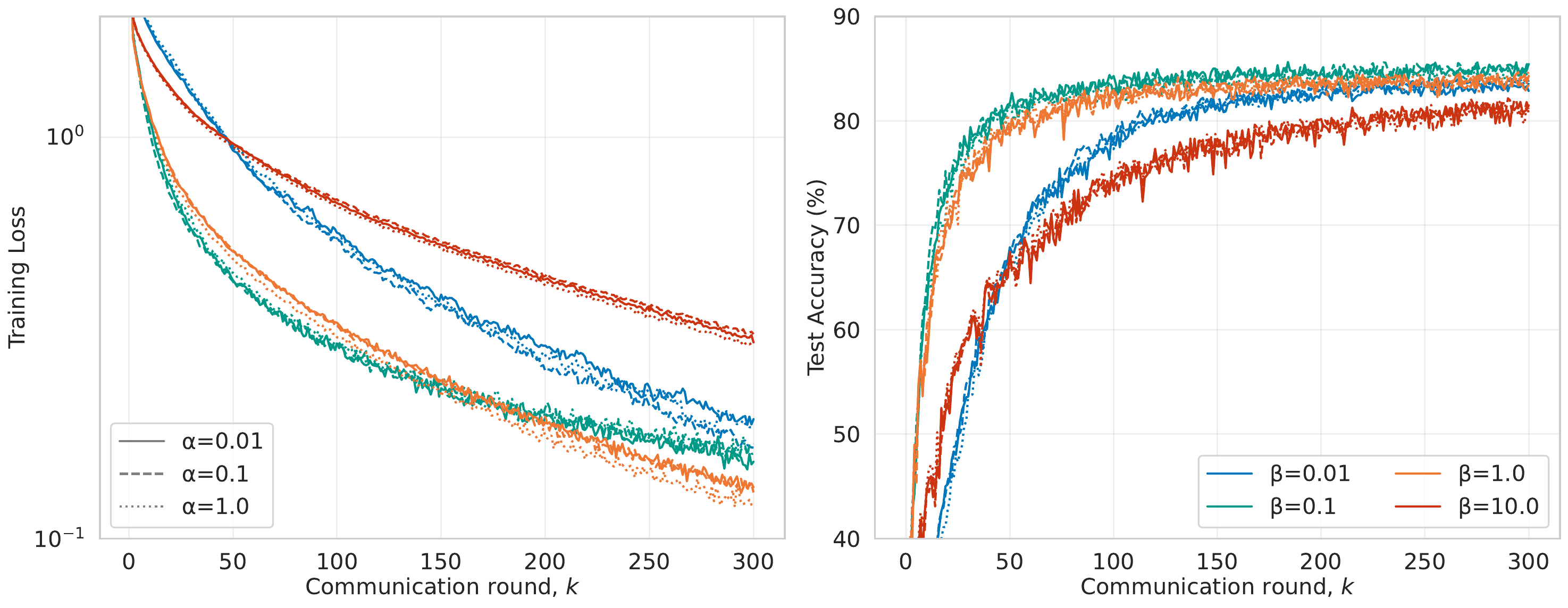}}
\caption{Training loss and test accuracy of non-private \alg with $\alpha=0.01$ [solid], $0.1$ [dashed], and $1.0$ [dotted], and $\beta=0.01$ [blue], $0.1$ [green], $1.0$ [orange], and $10.0$ [red].}
\label{smooth:fig:norm21_param_convergence}
\end{center}
\end{figure}

\subsubsection{Benefits of error compensation} \label{smooth:app:exp_ef}

Leveraging error compensation (EC), \alg without server normalization achieves superior performance compared to \algname{DP-SGD} with direct smoothed normalization across a range of $\beta$ and $\gamma$ hyperparameter settings (where $\alpha=0.01$), in terms of the final test accuracy reported in Figure \ref{smooth:fig:ef21_effect_acc} and Table \ref{smooth:tab:norm_comparison}.
From Table \ref{smooth:tab:norm_comparison}, \alg without server normalization consistently outperforms \algname{DP-SGD} across most combinations. This trend is particularly evident for small $\beta$ values ($\beta=0.01$), where \algname{DP-SGD} achieves only 51.10\% accuracy while \alg reaches 84.04\%.
The only exception is $\beta=10.0$, where \algname{DP-SGD} outperforms \alg. However, this combination is less practical in the private setting, as too high $\beta$ values imply high private noise, thus leading to slow algorithmic convergence.
\begin{figure}[!htbp]
\begin{minipage}[t]{0.53\textwidth}
\vspace{0pt}
\begin{center}
\centerline{\includegraphics[width=\linewidth]{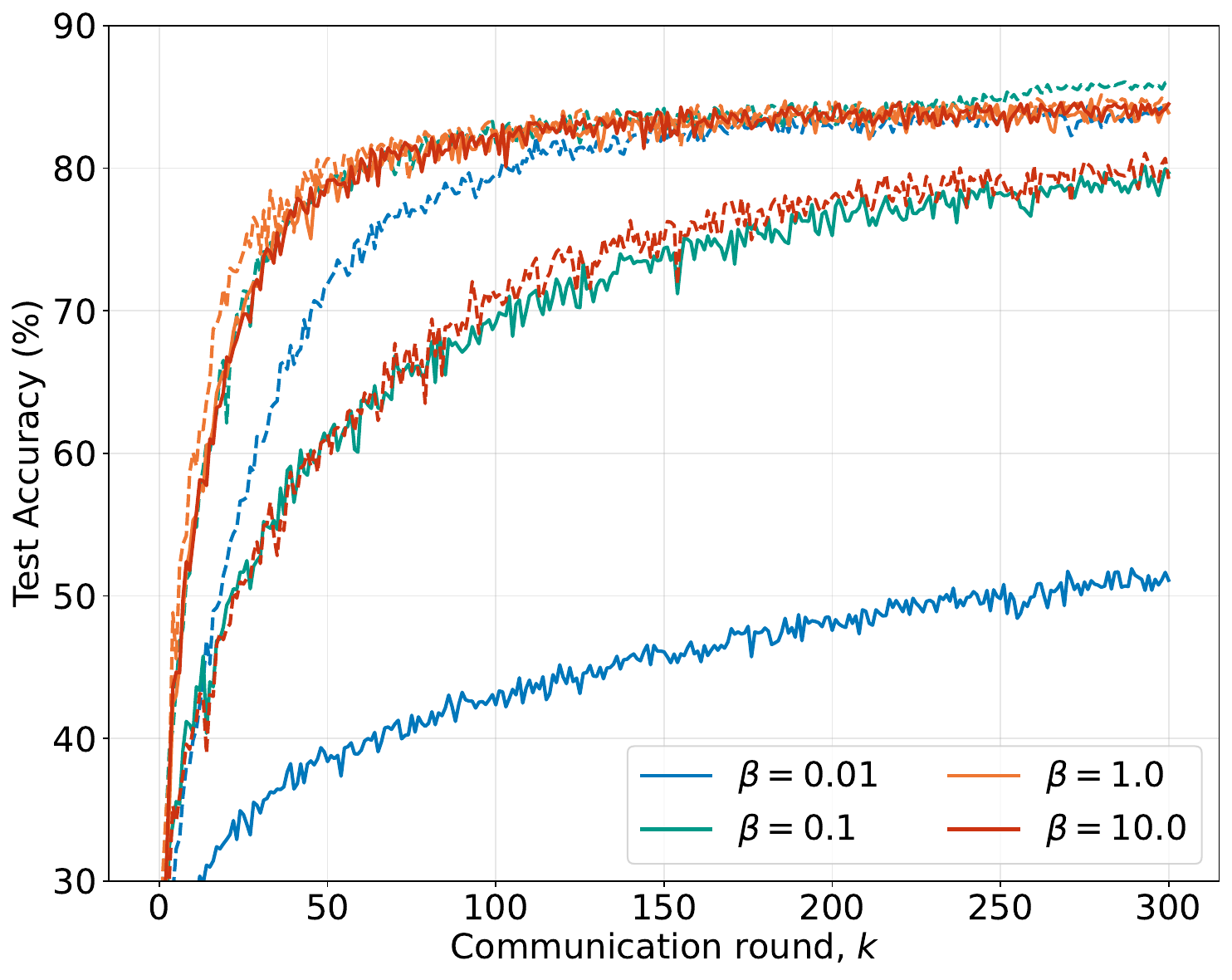}}
\caption{Comparison of \algname{DP-SGD} \eqref{smooth:eqn:DP_biased_GD} [solid] and \alg (\ref{smooth:alg:norm_norm21_dp_v1}) [dashed] without server normalization.} %
\label{smooth:fig:ef21_effect_acc}
\end{center}
\end{minipage}
\hfill
\begin{minipage}[t]{0.45\textwidth}
\vspace{0pt}
\begin{center}
\resizebox{\linewidth}{!}{%
\begin{tabular}{cccc}
\toprule
Method & $\beta$ & $\gamma$ & Final Accuracy \\
\midrule
\alg & 0.01 & 0.1 & 84.04\% \\
 & 0.1 & 0.1 & \textbf{86.09}\% \\
 & 1.0 & 0.1 & 84.80\% \\
 & 10.0 & 0.01 & 79.25\% \\
\midrule
\algname{DP-SGD} \eqref{smooth:eqn:DP_biased_GD} & 0.01 & 1.0 & 51.10\% \\
 & 0.1 & 1.0 & 79.68\% \\
 & 1.0 & 1.0 & 83.89\% \\
 & 10.0 & 0.1 & 84.50\% \\
\bottomrule
\end{tabular}%
}
\caption{Best configurations and final test accuracies.}
\label{smooth:tab:norm_comparison}
\end{center}
\end{minipage}
\end{figure}

\subsubsection{Effect of server normalization} \label{smooth:app:exp_serv_norm}

We investigate the impact of server-side normalization (Line 11 in Algorithm \ref{smooth:alg:norm_norm21_dp_v1}) on the convergence performance of \alg. We reported training loss and test accuracy of \alg without and with server normalization in Figure \ref{smooth:fig:serv_norm} while summarizing their final test accuracy in Table \ref{smooth:tab:serv_norm_}.

\alg without server normalization generally achieves faster convergence in training loss and higher test accuracy than \alg with server normalization across varying $\beta$ values.
Notably, at $\beta=0.1$, \alg without server normalization achieves the highest test accuracy of \textbf{86.09\%}.
Only at the large value of $\beta=10.0$ does server normalization improve the test accuracy of \alg without server normalization by approximately $2.2\%$.
\begin{figure}[h]
\begin{center}
\centerline{\includegraphics[width=0.9\linewidth]{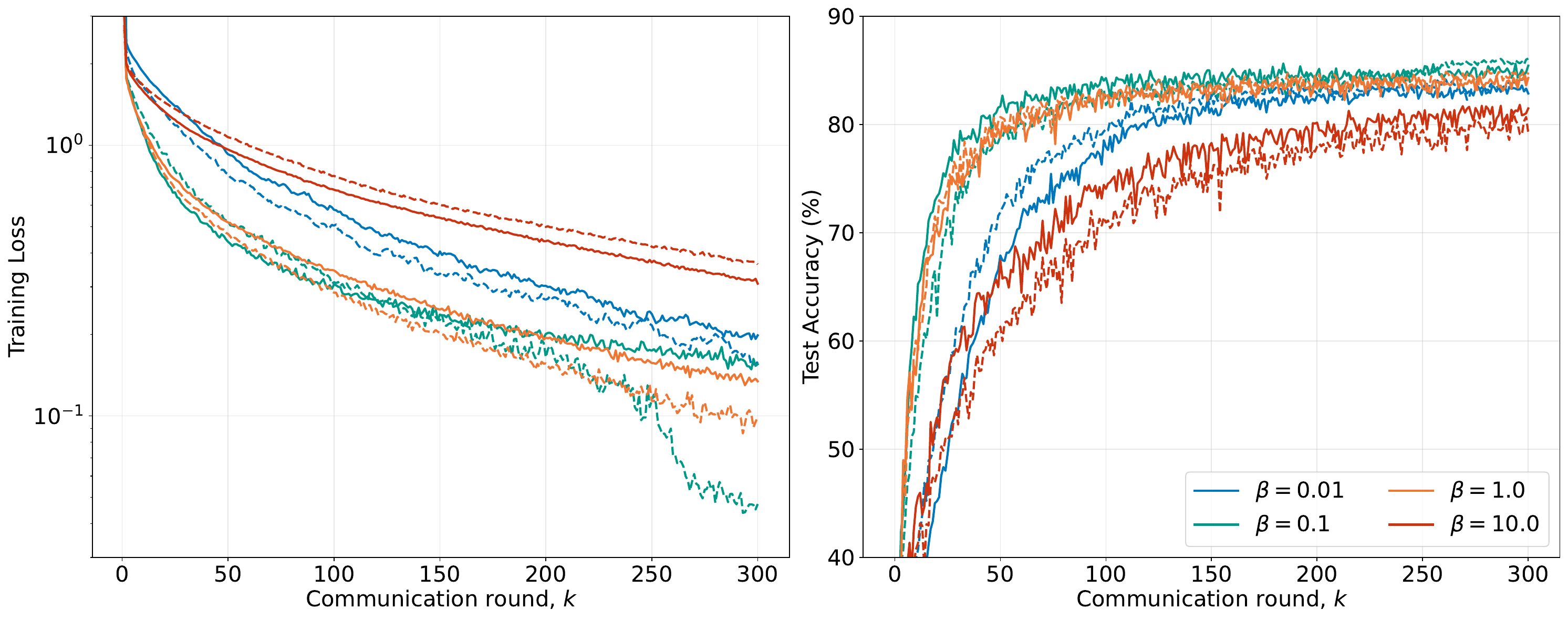}}
\caption{Training loss and test accuracy of \alg with [solid] and without [dashed] server normalization.}
\label{smooth:fig:serv_norm}
\end{center}
\vspace{-35pt}
\end{figure}

\begin{table}[h]
\centering
\begin{tabular}{cccc}
\toprule
Method: \alg & $\beta$ & $\gamma$ & Final Accuracy \\
\midrule
With server normalization & 0.01 & 0.01 & 82.86\% \\
 & 0.1 & 0.1 & 85.43\% \\
 & 1.0 & 0.1 & 84.29\% \\
 & 10.0 & 0.1 & 81.48\% \\
\midrule
Without server normalization & 0.01 & 0.1 & 84.04\% \\
 & 0.1 & 0.1 & \textbf{86.09}\% \\
 & 1.0 & 0.1 & 84.80\% \\
 & 10.0 & 0.01 & 79.25\% \\
\bottomrule
\end{tabular}
\caption{Best configurations and final test accuracies of \alg with and without server normalization.}
\label{smooth:tab:serv_norm_}
\end{table}

\subsubsection{Comparison of \algname{Clip21} and \alg} \label{smooth:app:exp_clip_norm}

Figure \ref{smooth:fig:clip_norm_comparison} and Table \ref{smooth:tab:clip_norm_comparison} show that \alg without server normalization\footnote{We ran \alg without server normalization because it showed better performance than \alg with server normalization according to Appendix \ref{smooth:app:exp_serv_norm}.} achieves comparable convergence performance to \algname{Clip21} for most $\beta$ values.
At small $\beta$ values ($0.01,0.1$), \alg without server normalization attains slightly lower final test accuracy. However, at high $\beta=10.0$, \algname{Clip21} maintains the higher test accuracy, as the large clipping threshold effectively disables clipping.
Furthermore, in most cases, both methods achieve their best performance with $\gamma=0.1$, except for \alg at $\beta=10.0$, where a smaller learning rate ($\gamma=0.01$) was optimal.

\begin{figure}[h]
\begin{center}
\centerline{\includegraphics[width=0.9\linewidth]{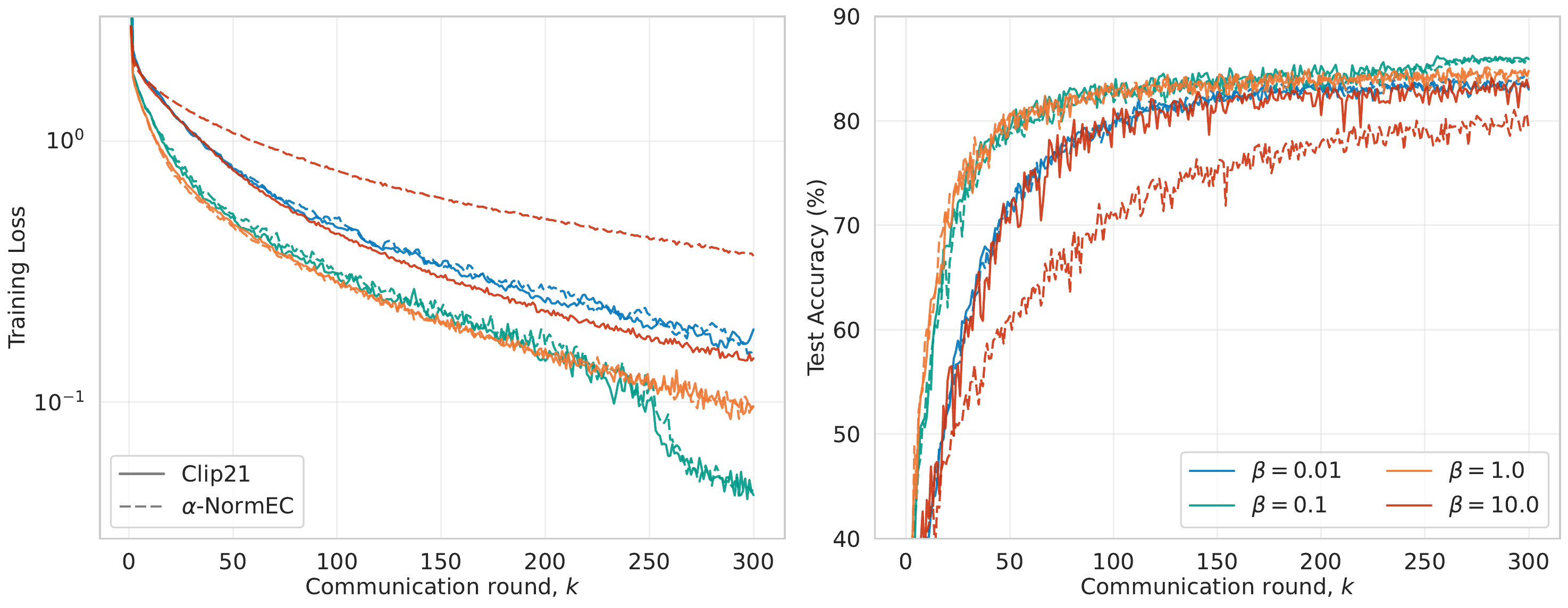}}
\caption{Training loss and test accuracy of \algname{Clip21} [solid] and \alg [dashed] without server normalization in non-private training.}
\label{smooth:fig:clip_norm_comparison}
\end{center}
\end{figure}

\begin{figure}[!htbp]
    \centering
    \begin{minipage}[c]{0.45\textwidth}
    \centering
    \begin{tabular}{cccc}
        \toprule
        Method & $\beta$ & $\gamma$ & Final Accuracy \\
        \midrule
        \algname{Clip21} & 0.01 & 0.1 & 83.00\% \\
        & 0.1 & 0.1 & 85.91\% \\
        & 1.0 & 0.1 & 84.78\% \\
        & 10.0 & 0.1 & 83.19\% \\
        \midrule
        \alg & 0.01 & 0.1 & 84.04\% \\
        & 0.1 & 0.1 & \textbf{86.09}\% \\
        & 1.0 & 0.1 & 84.80\% \\
        & 10.0 & 0.01 & 79.25\% \\
        \bottomrule
    \end{tabular}
    \caption{Best configurations and final test accuracies.}
    \label{smooth:tab:clip_norm_comparison}
    \end{minipage}\hfill
    \begin{minipage}[c]{0.50\textwidth}
    \centering
    \includegraphics[width=\linewidth]{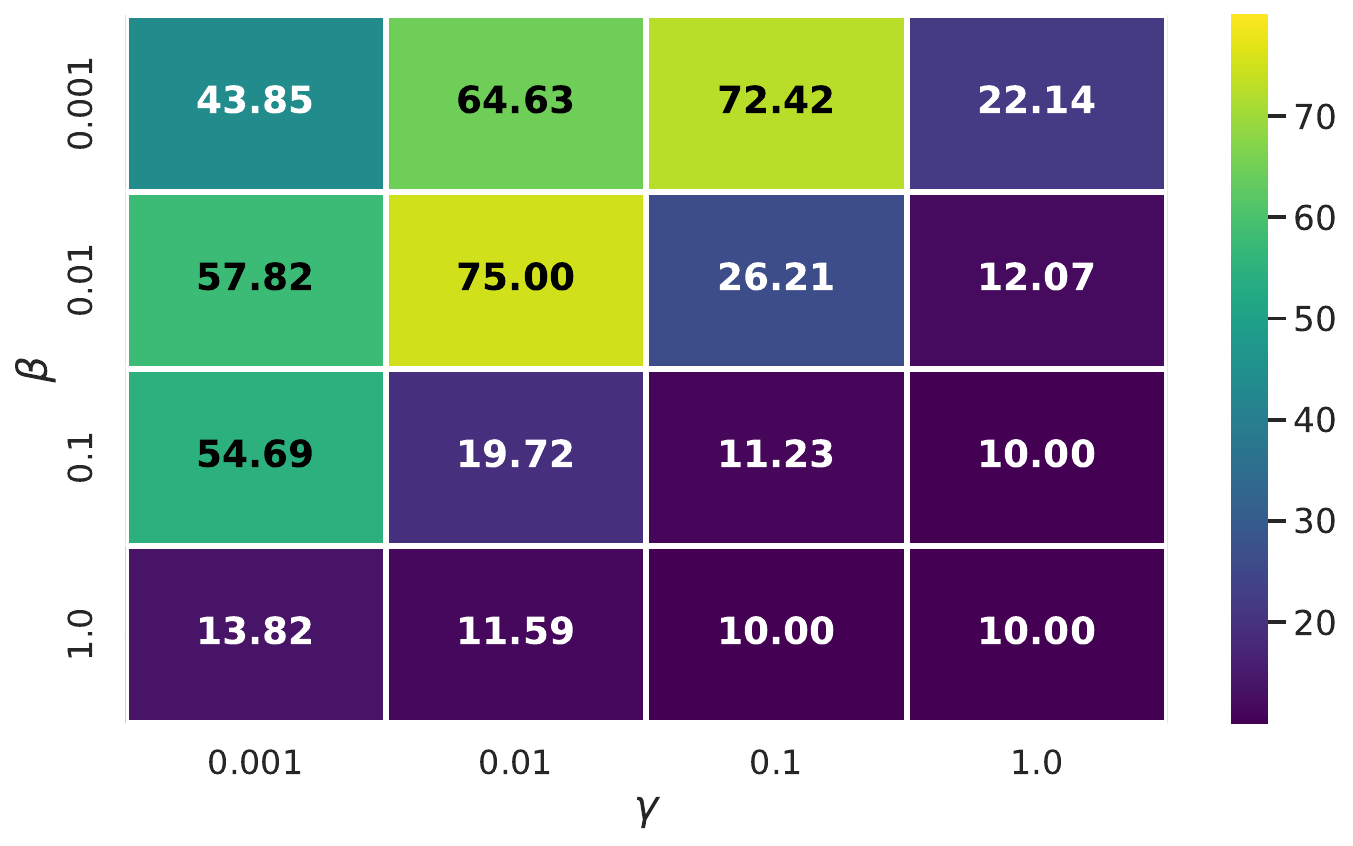}
    \caption{The highest test accuracy of \algname{DP-Clip21}.}
    \label{smooth:fig:clip_accuracy}
    \end{minipage}
\end{figure}

\subsection{Private training} \label{smooth:app:exp_dp}
We complement the results in Section \ref{smooth:sec:private_exp} with test accuracy convergence curves in Figure \ref{smooth:fig:dp_methods} (right). Additionally, Figures \ref{smooth:fig:dp_methods_accuracy} and \ref{smooth:fig:clip_accuracy} present a comprehensive heatmap analysis of the highest test accuracy achieved by \algname{DP-}\alg with and without server normalization (SN) and \algname{Clip21} across different privacy levels ($\beta$) and learning rates ($\gamma$).

The heatmaps reveal that without server normalization, performance is highly sensitive to hyperparameter selection, with accuracy ranging from 10\% to 77.56\% depending on the specific $\beta$-$\gamma$ combination. With server normalization, this sensitivity is significantly reduced, with performance varying more gradually across the parameter space. The rightmost heatmap quantifies this difference, showing that server normalization provides substantial benefits (up to +53.49\%) at high privacy levels ($\beta=1.0$) and larger step sizes, while the non-normalized version can perform better at lower privacy levels with specific step sizes.

\begin{figure}[ht]
\includegraphics[width=\linewidth]{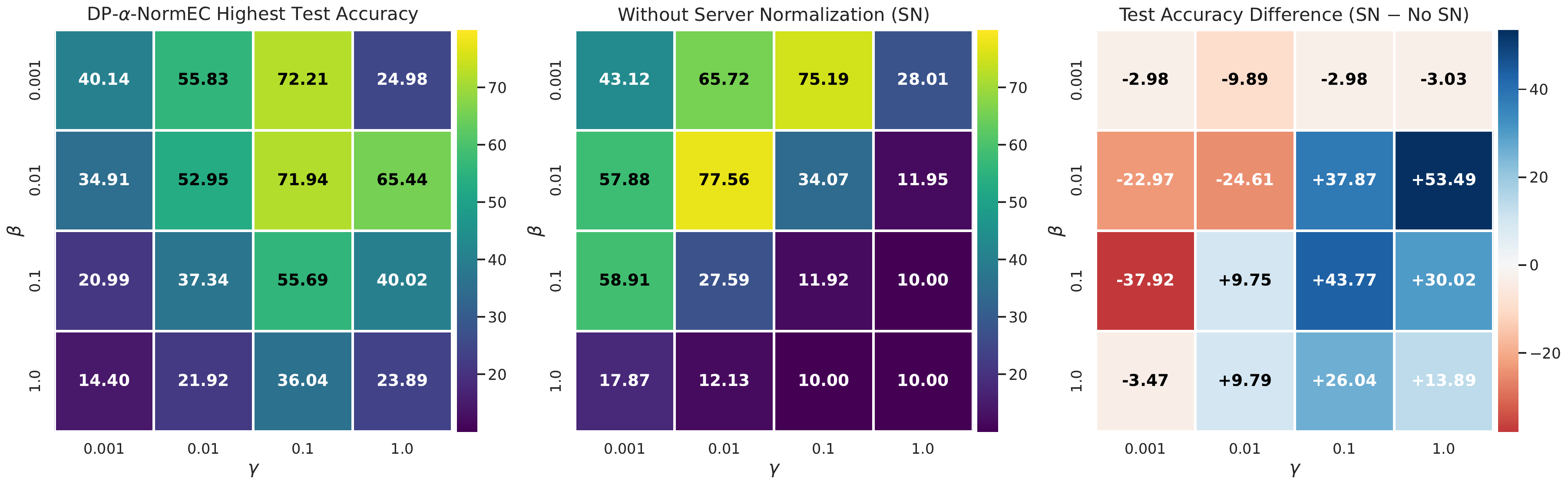}
\caption{The highest test accuracy of \algname{DP}-\alg with [left] and without [center] Server Normalization (SN), and their difference [right].}
\label{smooth:fig:dp_methods_accuracy}
\vspace{-5pt}
\end{figure}

\begin{figure}[H]
\begin{center}
\centerline{\includegraphics[width=\linewidth]{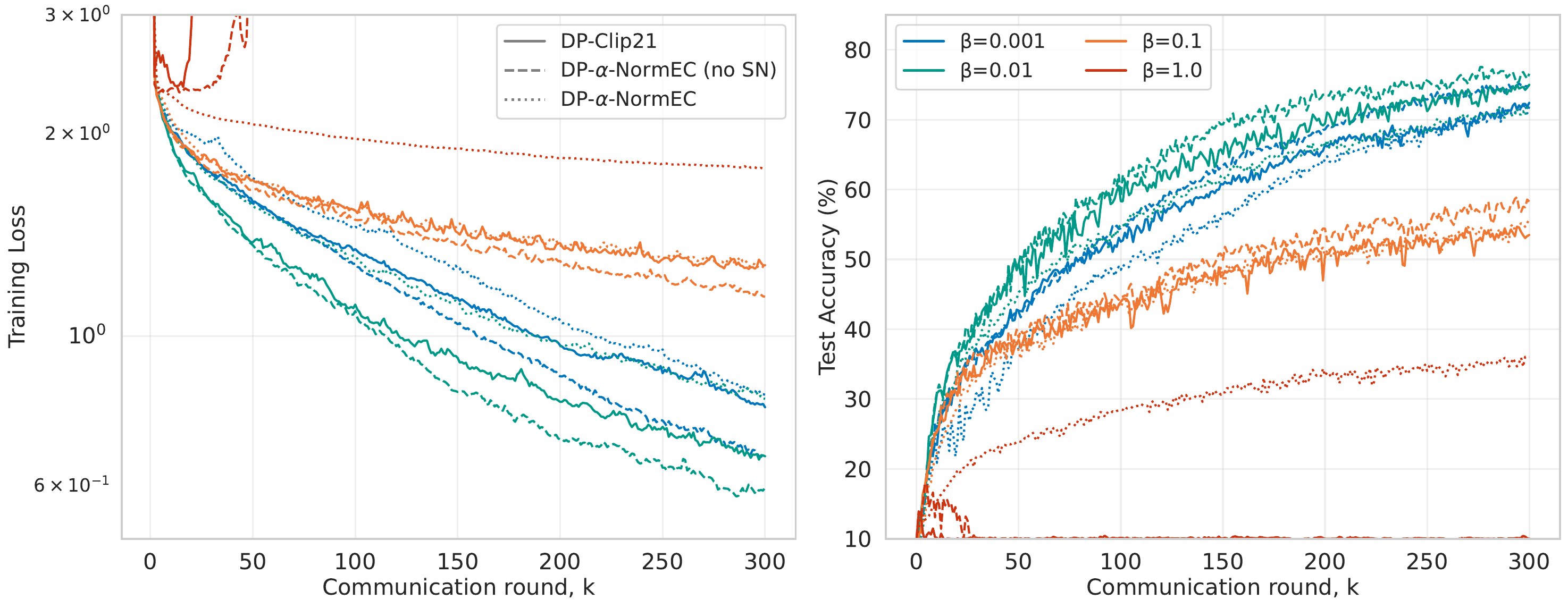}}
\caption{
Training loss and test accuracy of \algname{DP-Clip21} [solid], and \algname{DP-}\alg with [dotted] and without [dashed] server normalization (SN) across different $\beta$ values (with fine-tuned step sizes).}
\label{smooth:fig:dp_methods}
\end{center}
\vspace{-15pt}
\end{figure}

\subsubsection{Stricter privacy budget ($\epsilon=1$)} \label{smooth:app:small_eps}

To validate our method under a stricter privacy guarantee, we present additional results for $\epsilon=1$ in Figure \ref{smooth:fig:dp_epsilon1}. As expected, the increased DP noise in this high-privacy regime reduces the overall performance for both methods compared to the $\epsilon=8$ setting. However, the results demonstrate a consistent performance advantage and smaller variability for \algname{DP-}\alg over \algname{DP-Clip21}. For each corresponding hyperparameter setting for $\beta$, \algname{DP-}\alg achieves both a lower training loss and higher test accuracy, confirming its effectiveness in this practical setting.

\begin{figure}[H]
\begin{center}
\includegraphics[width=\linewidth]{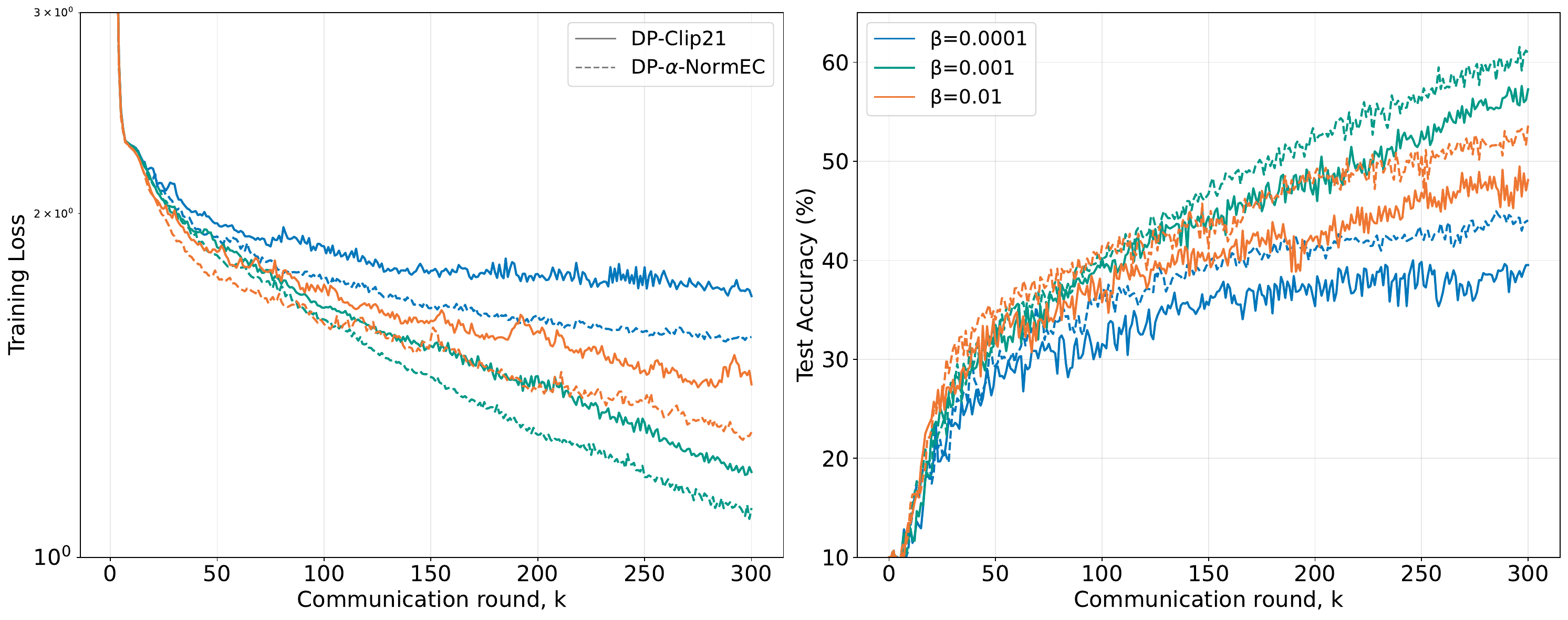}
\caption{Training loss and test accuracy of \algname{DP-Clip21} [solid] and \algname{DP-}\alg [dashed] across different $\beta$ values for $\epsilon=1$. \algname{DP-}\alg demonstrates both faster convergence and higher final accuracy for each $\beta$.}
\label{smooth:fig:dp_epsilon1}
\end{center}
\vspace{-10pt}
\end{figure}

\subsubsection{Shorter training}

\begin{wrapfigure}{r}{0.4\textwidth}
\vspace{-1.0\baselineskip}
\centering
\includegraphics[width=\linewidth]{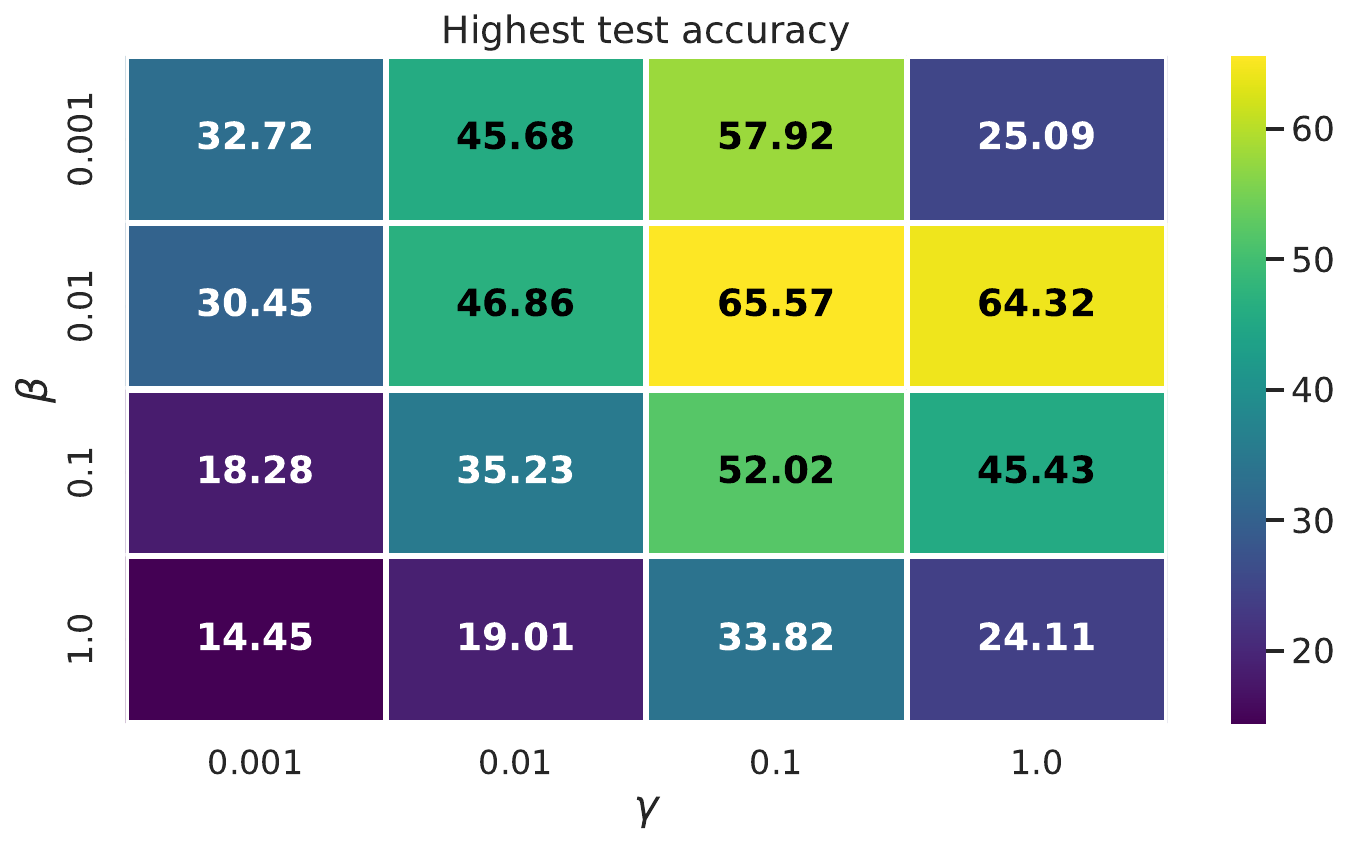}
\caption{Best test accuracy of \algname{DP-}\alg across different $\beta, \gamma$ values.}
\label{smooth:fig:dp_150}
\end{wrapfigure}

We present additional results in Figures \ref{smooth:fig:dp_150}, \ref{smooth:fig:dp_loss} by running \algname{DP-}\alg for \textbf{150 communication rounds}. The step size $\gamma$ is tuned for every parameter $\beta$. In the non-private setting, reasonably longer training is almost always beneficial. However, this may not hold in the private setting because the noise variance scales with the number of iterations. Interestingly, we observe that for $\beta=1$, the highest achieved accuracy after 150 iterations is almost the same as after a doubled communication budget of 300 iterations.

\begin{figure}[h]
\begin{minipage}[c]{0.49\textwidth}
\centerline{\includegraphics[width=\linewidth]{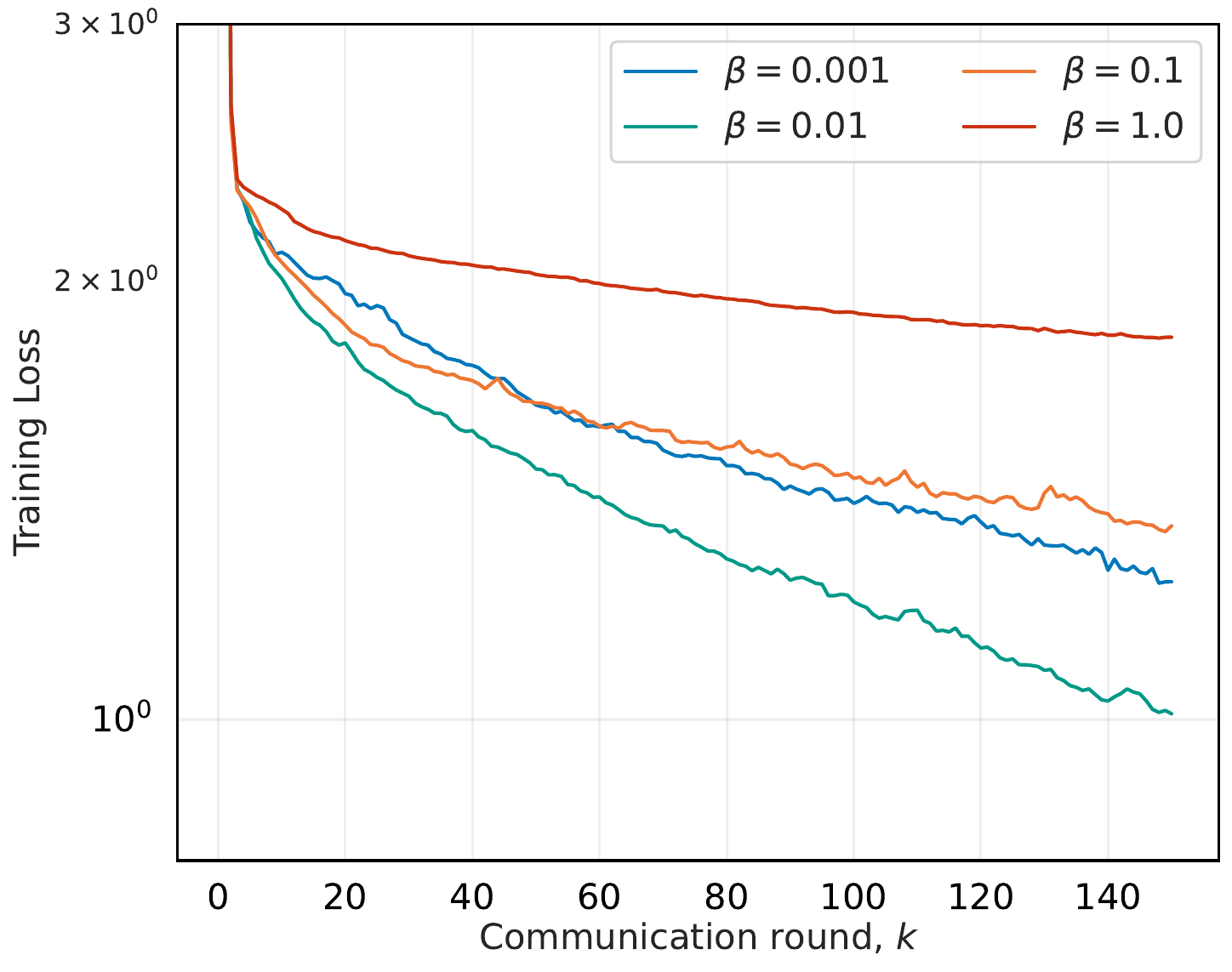}}
\end{minipage}
\hfill
\begin{minipage}[c]{0.49\textwidth}
\centerline{\includegraphics[width=\linewidth]{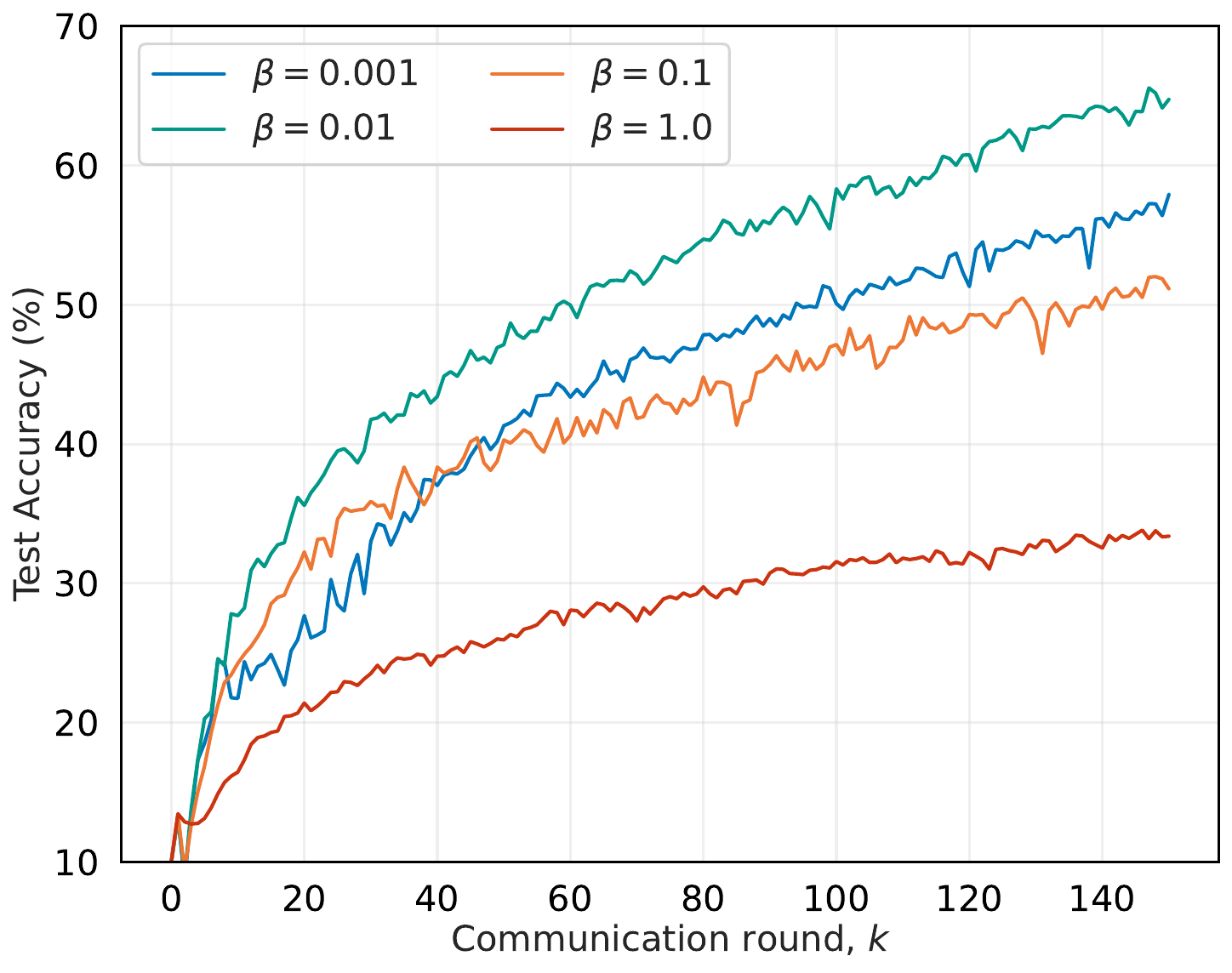}}
\end{minipage}
\caption{Training loss and test accuracy of \algname{DP-}\alg across different $\beta$ values.}
\label{smooth:fig:dp_loss}
\end{figure}

\subsection{Logistic regression} \label{smooth:sec:log_reg}

We consider distributed binary logistic regression, where worker $i\in\{1,\dots,n\}$ holds a shard $\mathcal{D}_i$ of the training set and
\[
f_i(x) \eqdef \frac{1}{|\mathcal{D}_i|}\sum_{(a,b)\in \mathcal{D}_i}\ell(a^\top x,b),
\qquad
\ell(s,b)=-b\log\sigma(s)-(1-b)\log(1-\sigma(s)),
\]
with labels $b\in\{0,1\}$ and $\sigma(\cdot)$ the logistic sigmoid. We use the \texttt{breast\_cancer} dataset (from scikit-learn) with a stratified train/test split (test fraction $0.2$, seed $0$); features are standardized using training-set mean/std and a constant bias feature is appended. We run with $n=10$ workers and $(\varepsilon,\delta)$-DP using $\varepsilon=3$ and $\delta=n^{-1.1}$. For each method we tune $(\beta,\gamma)$ by grid search (single seed $0$) over
\[
\begin{aligned}
\beta &\in \{10^{-7},10^{-6},10^{-5},10^{-4},10^{-3},10^{-2},10^{-1},1,10\},\\
\gamma &\in \{10^{-6},10^{-5},10^{-4},10^{-3},10^{-2},10^{-1},1,10,100\},
\end{aligned}
\]
with $\alpha=0.01$ fixed for \algname{DP-}\alg, selecting the configuration that maximizes final test accuracy, and then rerunning the selected configuration with $5$ seeds to report mean trajectories.

\begin{figure}[H]
\begin{center}
\includegraphics[width=\linewidth]{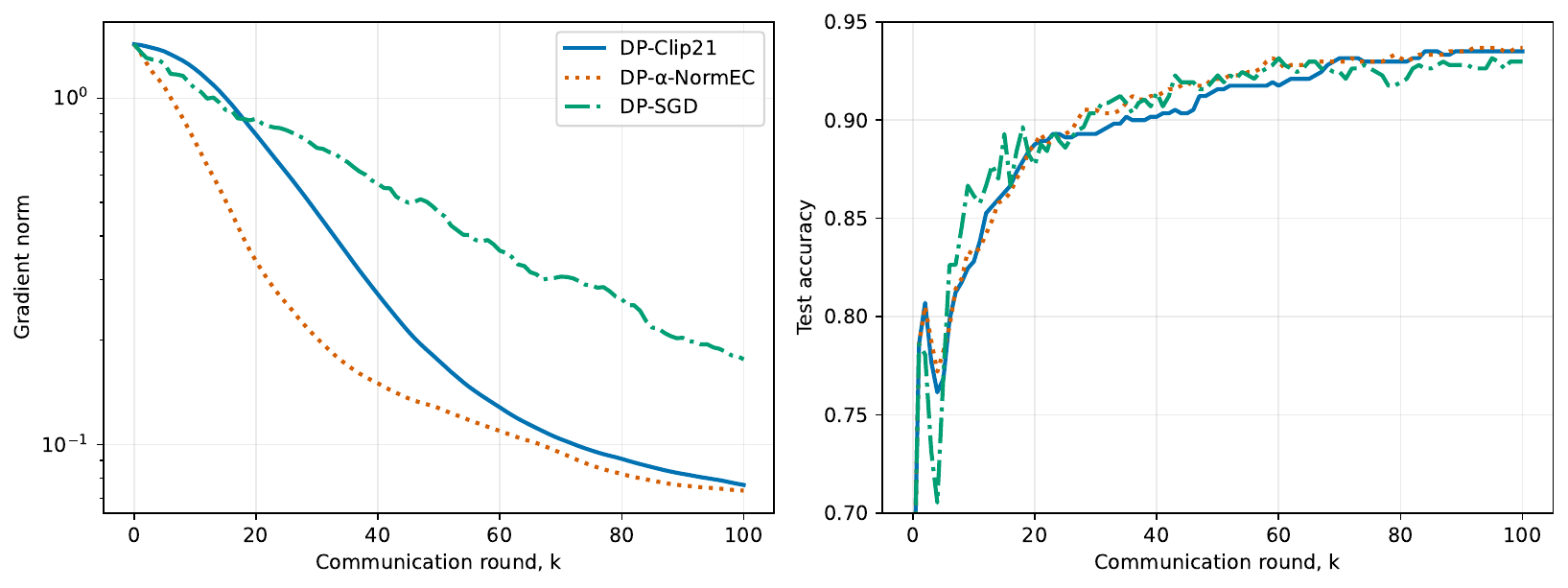}
\caption{Gradient norm and test accuracy of \algname{DP-Clip21} [solid], \algname{DP-}\alg [dotted], and \algname{DP-SGD} [dash-dotted] on Logistic Regression using the best hyperparameters $(\beta,\gamma)$ selected by final test accuracy and then averaged over $5$ seeds.} %
\label{smooth:fig:log_reg_full}
\end{center}
\end{figure}

As shown in Figure~\ref{smooth:fig:log_reg_full}, the error-feedback methods exhibit substantially faster reduction of the gradient norm than \algname{DP-SGD}: \algname{DP-}\alg decreases the gradient norm most rapidly in the initial phase, while \algname{DP-Clip21} reduces it more steadily. In contrast, \algname{DP-SGD} makes slower optimization progress and maintains a noticeably larger gradient norm throughout.

These trends translate into test performance: Figure \ref{smooth:fig:log_reg_full} shows that all methods reach high accuracy, but \algname{DP-}\alg achieves the highest final test accuracy, with \algname{DP-Clip21} very close behind and \algname{DP-SGD} slightly lower. The final numbers and the corresponding optimal hyperparameters $(\beta, \gamma)$ are reported in Table \ref{smooth:tab:reg_table}, confirming that \algname{DP-}\alg attains the best overall performance in this setting.

\begin{center}
\begin{table}%
\centering
\begin{tabular}{lccc}
\toprule
Method & Best $(\beta,\gamma)$ & Final test acc (\%) & Final grad norm \\
\midrule
\algname{DP-}\alg        & $(10^{-7},\,0.1)$ & $\mathbf{93.68} \pm 0.656$ & $\mathbf{0.074} \pm 0.023$ \\
\algname{DP-Clip21}      & $(10^{-4},\,10)$  & $93.509 \pm 0.895$           & $0.077 \pm 0.027$ \\
\algname{DP-SGD}         & $(10^{-1},\,0.1)$ & $92.982 \pm 0.961$           & $0.176 \pm 0.015$ \\
\bottomrule
\end{tabular}
\caption{Final performance (mean $\pm$ std over $5$ seeds; selected by final test accuracy).}
\label{smooth:tab:reg_table}
\end{table}
\end{center}

\endgroup

\chapter{Supplementary Material for Chapter 6 (Gluon)}
\label{appendix:chapter6_gluon}

\begingroup
\input{KAUST_Thesis/imported/06_gluon/import_setup}

\section{Related work}

\paragraph{Generalized Smoothness.}
The classical $L$-smoothness assumption, where the gradient is Lipschitz continuous with a global constant $L$, often fails to accurately capture the complex geometry of loss landscapes in deep learning \citep{crawshaw2022robustness,zhang2020why}. To address this, \citet{zhang2020why} introduced the $(L^0, L^1)$-smoothness condition, empirically observing in language model experiments that a bound of the form $\|\nabla^2 f(x)\| \leq L_0 + L_1\|\nabla f(x)\|$ better described the Hessian norm behavior.
Subsequent works have analyzed standard optimization algorithms under this generalized smoothness framework. For instance, \citet{gorbunov2024methods} and \citet{vankov2025optimizing} provided convergence analyses for the Gradient Method. \citet{hubler2024parameter} analyzed Normalized SGD with momentum in a parameter-agnostic setting under $(L^0, L^1)$-smoothness. \citet{yu2025mirror} proposed non-Euclidean generalized smoothness and established convergence rates for mirror-descent-type methods. Our work extends this line by incorporating $(L^0, L^1)$-smoothness into a \emph{layer-wise} context using arbitrary norms, an approach that is particularly well-suited for the LMO-based optimizers we study.

\paragraph{Anisotropic Smoothness.}
Recognizing the heterogeneous nature of parameters in large models, researchers have explored anisotropic smoothness conditions, where smoothness constants can vary across different dimensions or parameter blocks. Early work in this direction includes coordinate-wise Lipschitz continuity for coordinate descent methods \citep{nesterov2012efficiency,richtarik2014iteration}. More recently, \citet{bernstein2018signsgd} analyzed \algname{signSGD} under a weaker notion of coordinate-wise smoothness. \citet{crawshaw2022robustness} further developed this by analyzing Generalized \algname{signSGD} under a generalized coordinate-wise smoothness assumption, highlighting that different parameter groups can exhibit vastly different geometries. \citet{jiang2024convergence} focused on \algname{Adagrad}'s analysis under coordinate-wise smoothness and established lower bounds for SGD, underscoring the benefits of adaptivity.
\citet{liu2024adagradanisotropicsmoothness} proposed ``Anisotropic $(L^0, L^1)$-smoothness'' (a vector version of $(L_0,L_1)$-smoothness applied coordinate-wise) and demonstrated \algname{Adagrad}'s provable advantages over \algname{SGD} in this setting. \citet{xie2024adam} also leveraged anisotropic smoothness concepts in their convergence analysis of \algname{Adam}.
Our work contributes by defining and analyzing \emph{layer-wise $(L^0, L^1)$-smoothness}, which combines the benefits of the generalized smoothness model with a structured, anisotropic perspective tailored to the layer-block architecture of neural networks and compatible with arbitrary layer-specific norms. This framework is essential for understanding LMO-based methods like \algname{Muon} and \algname{Scion}.

\paragraph{LMO-based Optimizers.}
The optimizers \algname{Muon} \citep{jordan2024muon} and \algname{Scion} \citep{pethick2025training} represent a recent class of methods that have shown strong empirical performance in deep learning. \algname{Muon} was initially introduced as an effective empirical method, with its update rule for hidden layers inspired by ideas from \citet{bernstein2024old}. Subsequently, \citet{pethick2025training} (authors of \algname{Scion}) explicitly connected these types of updates to the Frank-Wolfe (FW) framework \citep{frank1956algorithm,jaggi2013revisiting}, proposing the use of layer-specific norms within an LMO-based update rule. These methods perform updates by solving, for each layer, a linear minimization problem over a norm ball centered at the current iterate. Prior theoretical analyses of these optimizers \citep{kovalev2025muon,li2025noteconvergencemuon,pethick2025training} have relied on standard $L$-smoothness and analyzed a simplified global update. Our work provides the first convergence guarantees for these methods under the more realistic layer-wise $(L_0,L_1)$-smoothness, directly addressing their practical layer-wise nature and leveraging the geometric insights offered by LMOs over general norms.

\section{Auxiliary lemmas}

\begin{lemma}\label{gluon:lemma:generalized-smoothness}
Let $f: \cS \to \mathbb{R}$ satisfy Assumption~\ref{gluon:ass:generalized-smoothness}. Then, for any $X, Y \in \cS$, we have
\begin{align*}
\left|f(Y)-f(X)-\left\langle \nabla f(X), Y-X\right\rangle\right| \leq \sum_{i=1}^p \frac{L^0_i + L^1_i \| \nabla _i f(X)\|_{(i) \star}}{2}\|Y_i-X_i\|_{(i)}^2.
\end{align*}
\end{lemma}

\begin{proof}
For all $X, Y \in \cS$ we have
\begin{align*}
f(Y) & =f(X)+\int_0^1\left\langle \nabla f(X+\tau(Y-X)), Y-X\right\rangle d \tau \\
&=f(X)+\left\langle \nabla f(X), Y-X\right\rangle+\int_0^1\left\langle \nabla f(X+\tau(Y-X))-\nabla f(X), Y-X\right\rangle d \tau.
\end{align*}

Therefore, using the Cauchy-Schwarz inequality and Assumption~\ref{gluon:ass:generalized-smoothness}, we obtain
\begin{eqnarray*}
&&\hspace{-1cm}\left|f(Y)-f(X)-\left\langle \nabla f(X), Y-X\right\rangle\right| \\
&\leq&\left|\int_0^1 \sum_{i=1}^p \left\langle \nabla _i f(X+\tau(Y-X))-\nabla _i f(X), Y_i-X_i\right\rangle_{(i)} d \tau \right| \\
&\leq&\int_0^1 \sum_{i=1}^p \left| \left\langle \nabla _i f(X+\tau(Y-X))-\nabla _i f(X), Y_i-X_i\right\rangle_{(i)} \right| d \tau \\
& \leq& \int_0^1 \sum_{i=1}^p \left\|\nabla _i f(X+\tau(Y-X))-\nabla _i f(X)\right\|_{(i) \star} \|Y_i-X_i\|_{(i)} d \tau \\
& \leq& \int_0^1 \sum_{i=1}^p \tau \left( L^0_i + L^1_i \| \nabla _i f(X) \|_{(i) \star} \right) \|Y_i-X_i\|_{(i)}^2 d \tau\\
&=&\sum_{i=1}^p \frac{L^0_i + L^1_i \| \nabla _i f(X)\|_{(i) \star}}{2}\|Y_i-X_i\|_{(i)}^2.
\end{eqnarray*}
\end{proof}

\begin{lemma}\label{gluon:lemma:L0vsL}
    Suppose that $f$ is $L$-smooth with respect to the norm defined in \eqref{gluon:eq:max_norm}, i.e.,
    \begin{align*}
        \norm{\nabla f(X) - \nabla f(Y)}_{\max\star} \leq L \norm{X-Y}_{\max},
    \end{align*}
    where $X = [X_1, \dots, X_p]$ and $Y = [Y_1, \dots, Y_p]$ with $X_i, Y_i \in \cS_i$.
    Then \Cref{gluon:ass:generalized-smoothness} holds with $L^0_i \leq L$ and $L^1_i = 0$ for all $i=1,\ldots,p$.
\end{lemma}
\begin{proof}
    $L$-smoothness and the definition of the norm give
    \begin{align*}
        \sum_{i=1}^p \|\nabla_i f(X) - \nabla_i f(Y)\|_{(i) \star} \leq L \max\brac{\norm{X_1 - Y_1}_{(1)}, \ldots, \norm{X_p - Y_p}_{(p)}}
    \end{align*}
    for all $X, Y \in \cS$. In particular, choosing
    \[
    X = [X_1, \dots, X_p],
    \qquad
    Y = [X_1, \dots, X_{j-1}, Y_j, X_{j+1}, \ldots X_p],
    \]
    we have
    \begin{align*}
        \|\nabla_j f(X) - \nabla_j f(Y)\|_{(j) \star} \leq \sum_{i=1}^p \|\nabla_i f(X) - \nabla_i f(Y)\|_{(i) \star} \leq L \norm{X_j - Y_j}_{(j)}
    \end{align*}
    for any $j\in\{1,\ldots,p\}$, proving the claim.
\end{proof}

\begin{lemma}\label{gluon:lemma:ineq}
    Suppose that $x_1, \ldots, x_p, y_1, \ldots, y_p \in \R$, $\max_{i\in[p]} |x_i| > 0$ and $z_1, \ldots, z_p > 0$. Then
    \begin{align*}
        \sum_{i=1}^p \frac{y_i^2}{z_i} \geq \frac{\parens{\sum_{i=1}^p x_i y_i}^2}{\sum_{i=1}^p z_i x_i^2}.
    \end{align*}
\end{lemma}
\begin{proof}
    Cauchy-Schwarz inequality gives
    \begin{align*}
        \parens{\sum_{i=1}^p x_i y_i}^2
        = \parens{\sum_{i=1}^p \frac{y_i}{\sqrt{z_i}} \sqrt{z_i} x_i}^2
        \leq \parens{\sum_{i=1}^p \frac{y_i^2}{z_i}} \parens{\sum_{i=1}^p z_i x_i^2}.
    \end{align*}
    Rearranging, we obtain the result.
\end{proof}

\begin{lemma}[Technical Lemma 10 by \citet{hubler2024parameter}]\label{gluon:lemma:tech1}
    Let $q \in(0,1)$, $p \geq 0$, and $p \geq q$. Further, let $a, b \in \mathbb{N}_{\geq 2}$ with $a \leq b$. Then
    \begin{align*}
    \sum_{k=a-1}^{b-1} (1+k)^{-p} \prod_{\tau=a-1}^k\left(1-(\tau+1)^{-q}\right) \leq (a-1)^{q-p} \exp \left(\frac{a^{1-q}-(a-1)^{1-q}}{1-q}\right).
    \end{align*}
\end{lemma}

\begin{lemma}[Technical Lemma 11 by \citet{hubler2024parameter}]\label{gluon:lemma:tech2}
    Let $t>0$ and for $k \in \mathbb{N}_{\geq 0}$, set $\beta^k =1-(k+1)^{-1 / 2}$, $t^k =t (k+1)^{-3 / 4}$, $t>0$. Then, for all $K \in \mathbb{N}_{\geq 1}$ the following inequalities hold:
    \begin{enumerate}[label=(\roman*)]
        \item $\sum_{k=0}^{K-1} t^k \sqrt{\sum_{\tau=0}^k (1-\beta^{\tau})^2 \prod_{\kappa=\tau+1}^k (\beta^\kappa)^2} \leq t\left(\frac{7}{2}+\sqrt{2 e^2} \log (K)\right)$,
        \item $\sum_{k=0}^{K-1} t^k \sum_{\tau=1}^k t^\tau \prod_{\kappa=\tau}^k \beta^\kappa \leq 7 t^2\left(3 +\log (K)\right)$.
    \end{enumerate}
\end{lemma}
\begin{proof}
    This is a direct consequence of Lemma~11 by~\citet{hubler2024parameter}. To obtain \textit{(ii)}, it suffices to take the limit as $ L^1 \to 0 $ in statement \textit{(ii)} of part~(b).
\end{proof}

\section{Remarks on the theoretical results}

\subsection{Note on radii and stepsizes}\label{gluon:sec:step_note}

It is known (see, e.g., \citet[Theorem D.1]{BPM}, who establish this for $\cS = \R^d$ under Euclidean norms; the extension to general normed vector spaces is entirely analogous) that if $g$ is a convex function, then the solution to the problem
\begin{align*}
    \argmin_{X\in\mathcal{B}^k} g(X)
\end{align*}
lies on the boundary of the ball $\mathcal{B}^k \eqdef \{ X \in \cS : \|X - X^k\| \leq t^k \}$ (unless $\cB^k \cap \argmin_{X\in\cS} g(X) \neq \emptyset$, that is, the ball intersects the set of minimizers of $g$).

This applies directly to the LMO subproblem solved at each iteration of \algname{Gluon} in \eqref{gluon:eq:update_rule_stoch}, since the objective $\langle M_i^k, X_i \rangle_{(i)}$ is linear in $X_i$, and hence convex. In other words, each LMO step moves the iterate from the center of the ball $X_i^k$ to a new point $X_i^{k+1}$ located on the boundary of $\mathcal{B}_i^k$, effectively traversing a distance of $t_i^k$ at each step. For this reason, we use the terms \emph{radius}, \emph{stepsize}, and \emph{learning rate} interchangeably.

\subsection{Note on prior analyses}

As presented, prior convergence results do not directly apply to the algorithms used in practice. However, there is a workaround. Specifically, some of the existing convergence guarantees \citep{kovalev2025muon,pethick2025training} expressed in terms of the flat vector $x$ are transferable to the structured parameters $X=[X_1, \ldots, X_l]\in\cS$ by employing the max-norm \citep{bernstein2024modular,larges2024calable}, defined as
\begin{align}\label{gluon:eq:max_norm}
    \norm{X}_{\max} \eqdef \max\brac{\norm{X_1}_{(1)}, \ldots, \norm{X_p}_{(p)}},
\end{align}
with corresponding dual norm $\|Y\|_{\max \star} = \sup_{\|X\|_{\text{max}} \leq 1} \langle X, Y \rangle = \sum_{i=1}^p \|Y_i\|_{(i) \star}$. Nevertheless, these works do not make this connection explicit, and an additional layer of analysis is required to ensure the guarantees meaningfully extend to the structured practical setting. Even if such a translation was attempted, the global treatment introduces serious practical limitations.
For example, real-world training pipelines tune parameters on a per-layer basis, reflecting the heterogeneous structure of deep networks. Max-norm-based guarantees overlook this variability and offer no mechanism for per-layer control in hyperparameter selection.

\section{Deterministic case}

We begin by considering the deterministic counterpart of \algname{Gluon}, as formalized in \Cref{gluon:algo:deterministic}. We first review several existing algorithms that fall within this framework (\Cref{gluon:sec:lmo_examples}), followed by a proof of \Cref{gluon:theorem:1} (\Cref{gluon:sec:det_proof}). Finally, we present an additional convergence guarantee under the layer-wise Polyak-Łojasiewicz (PŁ) condition (\Cref{gluon:sec:pl}).

\begin{algorithm}[t]
    \caption{Deterministic Adaptive Layer-Wise LMO-based Optimizer}
    \label{gluon:algo:deterministic}
    \begin{algorithmic}[1]
    \STATE \textbf{Input:} Initial model parameters $X^0 = [X_1^0, \dots, X_p^0] \in \cS$
    \FOR{$k = 0, 1, \dots, K-1$}
        \FOR{$i = 1, 2, \dots, p$}
            \STATE Choose adaptive stepsize/radius $t_i^k>0$ for layer $i$
            \STATE Update parameters for layer $i$ via LMO over $\mathcal{B}_i^k \eqdef \{ X_i \in \cS_i : \|X_i - X_i^k\|_{(i)} \leq t_i^k \}$:
            \begin{equation}
                \label{gluon:eq:update_rule}
                X_i^{k+1} = \mathrm{LMO}_{\mathcal{B}_i^k} \left( \nabla_i f(X^k) \right)
                \eqdef \underset{X_i \in \mathcal{B}_i^k}{\arg\min} \ \langle \nabla_i f(X^k), X_i \rangle_{(i)}
            \end{equation}
        \ENDFOR
        \STATE Update full vector: \
        $
        X^{k+1} = [X_1^{k+1}, \dots, X_p^{k+1}]
        $
    \ENDFOR
    \end{algorithmic}
\end{algorithm}

\subsection{Special cases of the LMO framework}\label{gluon:sec:lmo_examples}

As outlined in \Cref{gluon:sec:det_examples}, deterministic \algname{Gluon} encompasses a general class of algorithms, parameterized by the choice of norms $\|\cdot\|_{(i)}$ in the LMO. We now provide a more detailed discussion of the most notable special cases.

\paragraph{Layer-wise normalized GD \citep{wei2018blocknormalized}.}  Let $ \|\cdot\|_{(i)}= \|\cdot\|_{2 \rightarrow 2} $ for each parameter group and assume that $n_i = 1$ for all $i=1,\ldots,p$. In this case, the spectral norm reduces to the standard Euclidean norm $ \|\cdot\|_2 $, yielding the update rule
\begin{equation*}
    X_i^{k+1} = X_i^k - t^k_i \frac{\nabla _i f(X^k)}{\|\nabla _i f(X^k)\|_2}, \quad i = 1, \dots, p,
\end{equation*}
which corresponds to layer-wise normalized \algname{GD}. With a suitable choice of $t_i^k$ (see Theorem~\ref{gluon:theorem:1}), the method can also recover the Gradient Method for $(L^0, L^1)$-smooth functions \citep{vankov2025optimizing}.

\paragraph{Layer-wise signGD \citep{balles2020geometrysigngradientdescent}.} Suppose that $\|\cdot\|_{(i)}=\|\cdot\|_{1 \rightarrow \infty}$ for each parameter group, with $ n_i = 1 $ for all $i=1,\ldots,p$. Then, $ \|\cdot\|_{1 \rightarrow \infty} $ reduces to $ \|\cdot\|_{\infty} $, and the update becomes
\begin{equation*}
    X_i^{k+1} = X_i^k - t^k_i \text{sign}\parens{\nabla _i f(X^k)}, \quad i = 1, \dots, p,
\end{equation*}
where the sign function is applied element-wise. This is equivalent to layer-wise \algname{signGD}.

\paragraph{Muon \citep{jordan2024muon}.} Here, the spectral norm $ \|\cdot\|_{2 \rightarrow 2} $ is used for all parameter groups, without restrictions on $n_i$. In this case, it can be shown that \eqref{gluon:eq:update_rule} is equivalent to
\begin{equation}\label{gluon:eq:muon}
    X_i^{k+1} = X_i^k - t^k_i U_i^k \left(V_i^k\right)^{\top}, \quad i = 1, \dots, p,
\end{equation}
where $\nabla _i f(X^k) = U^k_i \Sigma^k_i \left(V^k_i\right)^{\top}$ is the singular value decomposition \citep{bernstein2024old}. This is exactly the per-layer deterministic version of the \algname{Muon} optimizer. In practical LLM training, a more general variant of \eqref{gluon:eq:muon} incorporating stochasticity and momentum is applied to the intermediate layers, while the input and output layers are optimized using other methods.

\paragraph{Unconstrained Scion \citep{pethick2025training}.} We can also recover two variants of \algname{unScion} introduced by~\citet{pethick2025training}: one for training LLMs on next-token prediction, and another for training CNNs for image classification.

\begin{itemize}
    \item \textbf{Training LLMs.} Define the norms $ \|\cdot\|_{(i)} $ as follows: for $ i = 1, \dots, p-1 $, corresponding to weight matrices of transformer blocks, set $\|\cdot\|_{(i)} = \sqrt{\nicefrac{n_i}{m_i}} \|\cdot\|_{2 \to 2},$
    and for the last group~$ X_p $, representing the embedding and output layers (which coincide under the weight sharing regime considered here), let $\|\cdot\|_{(p)} = n_p \|\cdot\|_{1 \to \infty}$.
    In this case, \eqref{gluon:eq:update_rule} becomes
    \begin{equation}\label{gluon:eq:scion_llm}
    \begin{aligned}
        X_i^{k+1} &= X_i^k - t^k_i \sqrt{\frac{m_i}{n_i}} U_i^k \left(V_i^k\right)^{\top}, \quad i = 1, \dots, p-1, \\
        X_p^{k+1} &= X_p^k - \frac{t^k_p}{n_p} \text{sign}\left(\nabla _p f(X^k)\right),
    \end{aligned}
    \end{equation}
    where $\nabla _i f(X^k) = U^k_i \Sigma^k_i \left(V^k_i\right)^{\top}$ is the singular value decomposition. This is equivalent to deterministic layer-wise \algname{unScion} optimizer without momentum.
    A more general variant, incorporating stochasticity and momentum and applied to all layers, was shown by \citet{pethick2025training} to outperform \algname{Muon} on LLM training tasks.

    \item \textbf{Training CNNs.} The main difference in the CNN setting is the presence of not only 2D weight matrices but also 1D bias vectors and parameters of 4D convolutional kernels. Biases are 1D tensors of shape $\mathbb{R}^{C_i^{out}}$, for which we use scaled Euclidean norms. Convolutional parameters (conv) are 4D tensors with shapes $\mathbb{R}^{C_i^{out} \times C_i^{in} \times k \times k}$, where $C_i^{out}$ and $C_i^{in}$ denote the number of output and input channels, and $k$ is the kernel size. To compute norms, we reshape each 4D tensor to a 2D matrix of shape $\mathbb{R}^{C_i^{out} \times C_i^{in}k^2}$ and then apply a scaled $ \|\cdot\|_{2 \rightarrow 2} $ norm. This yields the norm choices $\|\cdot\|_{(i)} = \sqrt{\nicefrac{1}{C_i^{out}}}\|\cdot\|_2$ for biases, $ \|\cdot\|_{(i)} = k^2 \sqrt{\nicefrac{C_i^{in}}{C_i^{out}}} \|\cdot\|_{2 \to 2} $ for conv, and $\|\cdot\|_{(p)} = n_p \|\cdot\|_{1 \to \infty}$ for the last group $ X_p $, associated with classification-head weights. Then, it can be shown that (\ref{gluon:eq:update_rule}) is equivalent to
    \begin{equation}\label{gluon:eq:scion_cnn}
    \begin{aligned}
        X_i^{k+1} &= X_i^k - t^k_i \sqrt{C_i^{out}} \frac{\nabla _i f(X^k)}{\|\nabla _i f(X^k)\|_2}, \quad\quad\, \text{(for biases)}, \\
        X_i^{k+1} &= X_i^k - t^k_i \frac{1}{k^2} \sqrt{\frac{C_i^{out}}{C_i^{in}}} U_i^k \left(V_i^k\right)^{\top}, \quad \text{(for conv)}, \\
        X_p^{k+1} &= X_p^k - \frac{t^k_p}{n_p} \text{sign}\left(\nabla _p f(X^k)\right), \quad\quad\, \text{(for head)}
    \end{aligned}
    \end{equation}
    where $\nabla _i f(X^k) = U^k_i \Sigma^k_i \left(V^k_i\right)^{\top}$ is the singular value decomposition. This corresponds to the deterministic layer-wise \algname{unScion} optimizer without momentum.
\end{itemize}

\subsection{Proof of theorem \ref{gluon:theorem:1}}\label{gluon:sec:det_proof}

We now state and prove a generalization of \Cref{gluon:theorem:1}.
\begin{theorem}\label{gluon:theorem:1_full}
    Let \Cref{gluon:ass:generalized-smoothness} hold and fix $\varepsilon>0$. Let $X^0,\ldots,X^{K-1}$ be the iterates of deterministic \algname{Gluon} (Algorithm \ref{gluon:algo:deterministic}) run with stepsizes $t^k_i = \frac{\|\nabla _i f(X^k)\|_{(i) \star}}{L^0_i + L^1_i\|\nabla _i f(X^k)\|_{(i) \star}}$.
    Then,
    \begin{enumerate}
        \item In order to reach the precision
        \begin{align*}
            \min_{k=0,\ldots,K-1} \sum_{i=1}^p \left\| \nabla _i f(X^{k}) \right\|_{(i) \star} \leq \epsilon,
        \end{align*}
        it suffices to run the algorithm for
        \begin{align}\label{gluon:eq:det_rate1}
            K = \left\lceil \frac{2 \Delta^0 \sum_{i=1}^p L^0_i}{\epsilon^2} + \frac{2 \Delta^0 L^1_{\max}}{\epsilon} \right\rceil
        \end{align}
        iterations;

        \item In order to reach the precision
        \begin{align}\label{gluon:eq:det_precision2}
            \min\limits_{k=0,\ldots,K-1} \sum\limits_{i=1}^p \left[\frac{\frac{1}{L_i^1}}{\frac{1}{p} \sum_{j=1}^p \frac{1}{L_j^1}} \norm{\nabla _i f(X^k)}_{(i) \star}\right] \leq \varepsilon,
        \end{align}
        it suffices to run the algorithm for
        \begin{align}\label{gluon:eq:det_rate2}
            K = \left\lceil \frac{2 \Delta^0 \parens{\sum_{i=1}^p \frac{L^0_i}{(L_i^1)^2}}}{\varepsilon^2 \parens{\frac{1}{p} \sum_{j=1}^p \frac{1}{L_j^1}}^2}
            + \frac{2 \Delta^0}{\varepsilon \parens{\frac{1}{p} \sum_{j=1}^p \frac{1}{L_j^1}}} \right\rceil
        \end{align}
        iterations,
    \end{enumerate}
    where $\Delta^0\eqdef f(X^0) - \inf_{X \in \cS} f(X)$ and $L^1_{\max}\eqdef\max_{i=1,\dots,p} L^1_i$.
\end{theorem}

\begin{remark}\label{gluon:rem:det}
    Let us compare bounds \eqref{gluon:eq:det_rate1} and \eqref{gluon:eq:det_rate2}. Due to the reweighting of the gradient component norms in \eqref{gluon:eq:det_precision2}, the rates are not exactly equivalent. Nevertheless, both use weights that sum to $p$, ensuring a fair comparison. Obviously, $(\nicefrac{1}{p} \sum_{j=1}^p \nicefrac{1}{L_j^1})^{-1} \leq L^1_{\max}$,
    so the second term in~\eqref{gluon:eq:det_rate2} is always no worse than its counterpart in~\eqref{gluon:eq:det_rate1}. The comparison of the first terms, however, depends on how the sequences $\{L^0_i\}$ and $\{L^1_i\}$ relate: if larger values of $L^0_i$ tend to be attached to smaller values of $L^1_i$, then the first term in~\eqref{gluon:eq:det_rate1} improves over that in~\eqref{gluon:eq:det_rate2}, while for a positive correlation the opposite is true. Indeed, in the extreme case in which $L^0_1 \geq \ldots \geq L^0_p$ and $L^1_1 \leq \ldots \leq L^1_p$ (or the reverse ordering), Chebyshev's sum inequality implies that
    \begin{align*}
        \frac{\sum\limits_{i=1}^p \frac{L^0_i}{(L_i^1)^2}}{\parens{\frac{1}{p} \sum\limits_{j=1}^p \frac{1}{L_j^1}}^2}
        \geq \frac{\parens{\frac{1}{p} \sum\limits_{i=1}^p \frac{L^0_i}{L_i^1}} \parens{\frac{1}{p} \sum\limits_{i=1}^p \frac{1}{L_i^1}}}{\frac{1}{p} \parens{\frac{1}{p} \sum\limits_{j=1}^p \frac{1}{L_j^1}}^2}
        \geq \frac{\parens{\frac{1}{p} \sum\limits_{i=1}^p L^0_i} \parens{\frac{1}{p} \sum\limits_{i=1}^p \frac{1}{L_i^1}}}{\frac{1}{p} \parens{\frac{1}{p} \sum\limits_{j=1}^p \frac{1}{L_j^1}}}
        = \sum\limits_{i=1}^p L^0_i.
    \end{align*}
    Conversely, if both sequences $\{L^0_i\}$ and $\{L^1_i\}$ are sorted in the same order (either increasing or decreasing), the inequality reverses, and the first term of \eqref{gluon:eq:det_rate2} may be tighter.
    That said, empirical evidence we provide in \Cref{gluon:sec:experiments} indicates that in practice $L^0_i \approx 0$ across all layers, in which case the first terms in \eqref{gluon:eq:det_rate1} and \eqref{gluon:eq:det_rate2} effectively vanish. Then, \eqref{gluon:eq:det_rate2} is clearly superior, replacing the worst-case constant $L^1_{\max}$ by the harmonic mean.
\end{remark}

\begin{proof}
We start with the result obtained in Lemma~\ref{gluon:lemma:generalized-smoothness} with $X=X^k$ and $Y=X^{k+1}$
\begin{align*}
f(X^{k+1})
&\leq f(X^k) + \left\langle \nabla f(X^k), X^{k+1}-X^k\right\rangle \\
&\qquad + \sum_{i=1}^p
\frac{L^0_i + L^1_i\|\nabla _i f(X^k)\|_{(i) \star}}{2}
\|X^k_i-X^{k+1}_i\|^2_{(i)}\\
&= f(X^k) + \sum_{i=1}^p
\left[ \left\langle \nabla _i f(X^k), X_i^{k+1}-X_i^k\right\rangle_{(i)}
\right.\\&\qquad\qquad\left.
+ \frac{L^0_i + L^1_i\|\nabla _i f(X^k)\|_{(i) \star}}{2}
\|X^k_i-X^{k+1}_i\|^2_{(i)}\right].
\end{align*}

The update rule (\ref{gluon:eq:update_rule}) and the definition of the dual norm $\|\cdot\|_{(i) \star}$ give
\begin{align*}
\|X^k_i-X^{k+1}_i\|^2_{(i)}\leq \left(t^k_i\right)^2
\end{align*}
and
\begin{align*}
    \left\langle \nabla _i f(X^k), X_i^{k+1}-X_i^k\right\rangle_{(i)}
    &= \left\langle \nabla _i f(X^k), \mathrm{LMO}_{\mathcal{B}_i^k} \left( \nabla _i f(X^k) \right)-X_i^k\right\rangle_{(i)} \\
    &= - t_i^k \max_{\|X_i\|_{(i)} \leq 1} \left\langle \nabla _i f(X^k), X_i\right\rangle_{(i)} \\
    &= - t_i^k \| \nabla _i f(X^k) \|_{(i) \star}.
\end{align*}
Consequently,
\begin{align*}
f(X^{k+1}) \leq f(X^k) + \sum_{i=1}^p \left[ - t_i^k \| \nabla _i f(X^k) \|_{(i) \star} + \frac{L^0_i + L^1_i\|\nabla _i f(X^k)\|_{(i) \star}}{2} \left(t^k_i\right)^2 \right].
\end{align*}
Now, choosing
\begin{align*}
t^k_i = \frac{\|\nabla _i f(X^k)\|_{(i) \star}}{L^0_i + L^1_i\|\nabla _i f(X^k)\|_{(i) \star}},
\end{align*}
which minimizes the right-hand side of the last inequality, yields the descent inequality
\begin{align}\label{gluon:eq:aousghr}
f(X^{k+1}) \leq f(X^k) - \sum_{i=1}^p \frac{\|\nabla _i f(X^k)\|^2_{(i) \star}}{2\left(L^0_i + L^1_i\|\nabla _i f(X^k)\|_{(i) \star}\right)}.
\end{align}

Summing the terms, we obtain
\begin{equation}
\begin{aligned}
\sum_{k=0}^{K-1} \sum_{i=1}^p \frac{\|\nabla _i f(X^k)\|^2_{(i) \star}}{2\left(L^0_i + L^1_i\|\nabla _i f(X^k)\|_{(i) \star}\right)} &\leq \sum_{k=0}^{K-1} \left( f(X^k) - f(X^{k+1}) \right) \\
&= f(X^0) - f(X^{K}) \\
&\leq f(X^0) - \inf_{X \in \cS} f(X) =: \Delta^0.
\end{aligned}
\label{gluon:eq:telescoping}
\end{equation}

Now, the analysis can proceed in two ways:

\begin{enumerate}
    \item Upper-bounding $L^1_i$ by $L^1_{\max}:=\max_{i=1,\dots,p} L^1_i$ in \eqref{gluon:eq:telescoping}, we obtain
    \begin{align}\label{gluon:eq:anmzdgzn}
    \sum_{k=0}^{K-1} \sum_{i=1}^p \frac{\|\nabla _i f(X^k)\|^2_{(i) \star}}{2\left(L^0_i + L^1_{\max}\|\nabla _i f(X^k)\|_{(i) \star}\right)}
    \leq \Delta^0.
    \end{align}
    Now, applying \Cref{gluon:lemma:ineq} with
    \[
    \begin{aligned}
    x_i &= 1,\\
    y_i &= \|\nabla _i f(X^{k})\|_{(i) \star},\\
    z_i &= 2 \parens{L^0_i + L^1_{\max} \norm{\nabla _i f(X^k)}_{(i) \star}},
    \end{aligned}
    \]
    gives
    \begin{align*}
    \phi \left( \sum_{i=1}^p \|\nabla _i f(X^{k})\|_{(i) \star}\right)
    &= \frac{\left( \sum_{i=1}^p \|\nabla _i f(X^{k})\|_{(i) \star} \right)^2}{2 \left(\sum_{i=1}^p L^0_i + L^1_{\max} \sum_{i=1}^p \|\nabla _i f(X^{k})\|_{(i) \star}\right)} \\
    &\leq \sum_{i=1}^p \frac{\|\nabla _i f(X^{k})\|^2_{(i) \star}}{2 \left(L^0_i + L^1_{\max} \|\nabla _i f(X^{k})\|_{(i) \star}\right)},
    \end{align*}
    where $\phi(t)\eqdef \frac{t^2}{2(\sum_{i=1}^p L^0_i+L^1_{\max} t)}$.
    Combining the last inequality with (\ref{gluon:eq:anmzdgzn}) and using the fact that $\phi$ is increasing, we obtain
    \begin{align}
    K \phi \left(\min_{k=0,\ldots,K-1} \sum_{i=1}^p \|\nabla _i f(X^{k})\|_{(i) \star}\right) \leq \sum_{k=0}^{K-1} \phi\left(\sum_{i=1}^p \|\nabla _i f(X^k)\|_{(i) \star}\right) \leq \Delta^0,
    \label{gluon:eq:for_the_proof_th2_part1}
    \end{align}
    and hence
    \begin{align*}
    \min_{k=0,\ldots,K-1} \sum_{i=1}^p \|\nabla _i f(X^{k})\|_{(i) \star} \leq \phi^{-1}\left( \frac{\Delta^0}{K} \right),
    \end{align*}
    where $\phi^{-1}$ is the inverse function (which exists since $\phi$ is increasing).
    Therefore, to reach the precision $\min_{k=0,\ldots,K-1} \sum_{i=1}^p \left\| \nabla _i f(X^{k}) \right\|_{(i) \star} \leq \epsilon$, it
    is sufficient to choose the number of iterations to be
    \begin{align*}
    K = \left\lceil \frac{\Delta^0}{\phi(\epsilon)} \right\rceil = \left\lceil \frac{2 \sum_{i=1}^p L^0_i \Delta^0}{\epsilon^2} + \frac{2L^1_{\max} \Delta^0}{\epsilon} \right\rceil.
    \end{align*}

    \item Alternatively, we can start from the inequality \eqref{gluon:eq:telescoping} and apply \Cref{gluon:lemma:ineq} with $x_i = \nicefrac{1}{L_i^1}$, $y_i = \norm{\nabla _i f(X^k)}_{(i) \star}$ and $z_i = 2 (L^0_i + L^1_i \norm{\nabla _i f(X^k)}_{(i) \star})$
    to obtain
    \begin{eqnarray*}
        \Delta^0 &\geq& \sum_{k=0}^{K-1} \sum_{i=1}^p \frac{\|\nabla _i f(X^k)\|^2_{(i) \star}}{2\left(L^0_i + L^1_i\|\nabla _i f(X^k)\|_{(i) \star}\right)} \\
        &\geq& \sum_{k=0}^{K-1} \frac{\parens{\sum_{i=1}^p \frac{1}{L_i^1} \norm{\nabla _i f(X^k)}_{(i) \star}}^2}{2 \parens{\sum_{i=1}^p \frac{1}{(L_i^1)^2} \parens{L^0_i + L^1_i \norm{\nabla _i f(X^k)}_{(i) \star}}}} \\
        &=& \sum_{k=0}^{K-1} \frac{\parens{\sum_{i=1}^p \frac{1}{L_i^1} \norm{\nabla _i f(X^k)}_{(i) \star}}^2}{2 \parens{\sum_{i=1}^p \frac{L^0_i}{(L_i^1)^2} + \sum_{i=1}^p \frac{1}{L_i^1} \norm{\nabla _i f(X^k)}_{(i) \star}}} \\
        &=& \sum_{t=0}^{K-1} \psi{\parens{\sum_{i=1}^p \frac{1}{L_i^1} \norm{\nabla _i f(X^k)}_{(i) \star}}},
    \end{eqnarray*}
    where $\psi(t) \eqdef \frac{t^2}{2 \parens{\sum_{i=1}^p \frac{L^0_i}{(L_i^1)^2} + t}}$. Since the function $\psi$ is increasing for $t>0$, $\psi^{-1}$ exists. It follows that
    \begin{eqnarray*}
        \Delta^0 &\geq& \sum_{k=0}^{K-1} \psi{\parens{\sum_{i=1}^p \frac{1}{L_i^1} \norm{\nabla _i f(X^k)}_{(i) \star}}} \\
        &\geq& K \psi{\parens{\min_{k=0,\ldots,K-1} \sum_{i=1}^p \frac{1}{L_i^1} \norm{\nabla _i f(X^k)}_{(i) \star}}},
    \end{eqnarray*}
    and hence
    \begin{eqnarray*}
        \min_{k=0,\ldots,K-1} \sum_{i=1}^p \frac{1}{L_i^1} \norm{\nabla _i f(X^k)}_{(i) \star}
        &\leq& \psi^{-1}\parens{\frac{\Delta^0}{K}}.
    \end{eqnarray*}
    This in turn means that to reach the precision
    \begin{eqnarray*}
        \min_{k=0,\ldots,K-1} \sum_{i=1}^p \left[\frac{\frac{1}{L_i^1}}{\frac{1}{p} \sum_{j=1}^p \frac{1}{L_j^1}} \norm{\nabla _i f(X^k)}_{(i) \star}\right] \leq \varepsilon,
    \end{eqnarray*}
    it suffices to run the algorithm for
    \begin{align*}
        K = \left\lceil \frac{\Delta^0}{\psi\parens{\varepsilon \parens{\frac{1}{p} \sum_{j=1}^p \frac{1}{L_j^1}}}} \right\rceil
        = \left\lceil \frac{2 \Delta^0 \parens{\sum_{i=1}^p \frac{L^0_i}{(L_i^1)^2}}}{\varepsilon^2 \parens{\frac{1}{p} \sum_{j=1}^p \frac{1}{L_j^1}}^2}
        + \frac{2 \Delta^0}{\varepsilon \parens{\frac{1}{p} \sum_{j=1}^p \frac{1}{L_j^1}}} \right\rceil
    \end{align*}
    iterations.
\end{enumerate}

\end{proof}

\subsection{Convergence under the PŁ condition}\label{gluon:sec:pl}

We now establish convergence rates under the layer-wise Polyak–Łojasiewicz (PŁ) condition, introduced in Assumption~\ref{gluon:ass:pl}. This property is especially relevant for heavily over-parameterized neural networks, as it has been shown to capture the properties of their loss landscapes \citep{overparameterized-loss-landscapes}.

\begin{assumption}[Layer-wise Polyak-Łojasiewicz condition]\label{gluon:ass:pl}
    The function $f: \cS \to \mathbb{R}$ satisfies the layer-wise Polyak-Łojasiewicz (PŁ) condition with a constant $\mu>0$, i.e., for any $X\in \cS$
    \begin{equation*}
       \sum_{i=1}^p \|\nabla _i f(X)\|^2_{(i) \star} \geq 2\mu\left(f(X)-f^{\star}\right),
    \end{equation*}
    where $f^{\star}:=\inf_{X \in \cS} f(X) > - \infty$.
\end{assumption}

Assumption~\ref{gluon:ass:pl} reduces to the standard PŁ condition \citep{karimi2020linearconvergencegradientproximalgradient} by vectorizing the parameters and adopting the Euclidean norm $\|\cdot\|_{2}$.

\begin{theorem}\label{gluon:theorem:1-pl}
    Let Assumptions \ref{gluon:ass:generalized-smoothness} and \ref{gluon:ass:pl} hold and fix $\varepsilon>0$. Let $X^0,\ldots,X^{K-1}$ be the iterates of deterministic \algname{Gluon} (Algorithm \ref{gluon:algo:deterministic}) run with $ t^k_i = \frac{\|\nabla _i f(X^k)\|_{(i) \star}}{L^0_i + L^1_i\|\nabla _i f(X^k)\|_{(i) \star}}$.

    \begin{enumerate}
        \item If $L^1_i\geq0$, then to reach the precision $\min_{k=0,\ldots,K-1} f(X^{k})-f^{\star} \leq \epsilon$, it suffices to run the algorithm for
        \begin{align*}
        K = \left\lceil \frac{\sum^p_{i=1} L^0_i \Delta^0}{\mu\epsilon} + \frac{\sqrt{2} L^1_{\max} \Delta^0}{\sqrt{\mu\epsilon}} \right\rceil
        \end{align*}
        iterations,

        \item If $L^1_i=0$ for all $i = 1, \dots, p$, then to reach the precision $f(X^{K})-f^{\star} \leq \epsilon$, it suffices to run the algorithm for
        \begin{align*}
        K = \left\lceil \frac{L^0_{\max}}{\mu} \log{\frac{\Delta^0}{\epsilon}} \right\rceil,
        \end{align*}
    \end{enumerate}
    where $L^0_{\max}:=\max_{i=1,\dots,p}L^0_i$, $L^1_{\max}:=\max_{i=1,\dots,p}L^1_i$, $\Delta^0 := f(X^0) - f^{\star}$ and $f^{\star} := \inf_{X \in \cS} f(X)$.
\end{theorem}

\begin{proof}
We consider two scenarios: (1) the general case with arbitrary $L^1_i\geq0$ and (2) $L^1_i=0$ for all $i = 1, \dots, p$.

\paragraph{Case 1: $L^1_i\geq0$.}

We start by following the same steps as in the proof of Theorem~\ref{gluon:theorem:1}. From~\eqref{gluon:eq:for_the_proof_th2_part1}, we have

\begin{align*}
\sum_{k=0}^{K-1} \phi\left(\sum_{i=1}^p \|\nabla _i f(X^k)\|_{(i) \star}\right) \leq \Delta^0,
\end{align*}
where $\phi(t):= \frac{t^2}{2(\sum_{i=1}^p L^0_i+L^1_{\max} t)}$.
Now, using Assumption~\ref{gluon:ass:pl}, we get
\begin{align*}
\left(\sum_{i=1}^p \|\nabla_i f(X^{k})\|_{(i)\star}\right)^2
\geq \sum_{i=1}^p \|\nabla_i f(X^{k})\|^2_{(i)\star}
\geq 2\mu \left(f(X^{k})-f^{\star}\right).
\end{align*}

Consequently, since $\phi$ is an increasing function,
\begin{align*}
K \phi \left(\sqrt{2\mu} \sqrt{f(X^{k^{\star}})-f^{\star}} \right) &\leq \sum_{k=0}^{K-1} \phi \left( \sqrt{2\mu} \sqrt{f(X^{k})-f^{\star}} \right) \\ &\leq \sum_{k=0}^{K-1} \phi\left(\sum_{i=1}^p \|\nabla _i f(X^k)\|_{(i) \star}\right) \leq \Delta^0,
\end{align*}
where $k^{\star}:=\operatorname{argmin}_{k=0,\ldots,K-1} f(X^{k})-f^{\star}$.
Denoting the corresponding inverse function (which exists since $\phi$ is increasing) by $\phi^{-1}$, it follows that
\begin{align*}
\sqrt{2\mu} \sqrt{f(X^{k^{\star}})-f^{\star}} \leq \phi^{-1}\left( \frac{\Delta^0}{K} \right) \leq \sqrt{2\mu \epsilon}.
\end{align*}

Therefore, to reach the precision $f(X^{k^{\star}})-f^{\star} \leq \epsilon$, it
is sufficient to choose the number of iterations
\begin{align*}
K = \left\lceil \frac{\Delta^0}{\phi\left( \sqrt{2\mu\epsilon}\right)} \right\rceil = \left\lceil \frac{\sum^p_{i=1} L^0_i \Delta^0}{\mu\epsilon} + \frac{\sqrt{2} L^1_{\max} \Delta^0}{\sqrt{\mu\epsilon}} \right\rceil.
\end{align*}

\paragraph{Case 2: $L^1_i = 0$.}

Inequality \eqref{gluon:eq:aousghr} from the proof of Theorem~\ref{gluon:theorem:1} with $L^1_i=0$ gives
\begin{align*}
f(X^{k+1}) \leq f(X^k) - \sum_{i=1}^p \frac{\|\nabla _i f(X^k)\|^2_{(i) \star}}{2L^0_i}.
\end{align*}

Using the fact that
\begin{align*}
\sum_{i=1}^p \frac{\|\nabla_i f(X^k)\|^2_{(i)\star}}{2L^0_i}
&\geq \min_{j=1,\dots,p} \frac{1}{2L^0_j}
\sum_{i=1}^p \|\nabla_i f(X^k)\|^2_{(i)\star} \\
&= \frac{1}{2 \max_{j=1,\dots,p} L^0_j}
\sum_{i=1}^p \|\nabla f(X^k)\|^2_{(i)\star}
\end{align*}
along with Assumption~\ref{gluon:ass:pl}, we obtain
\begin{align*}
f(X^{k+1}) \leq f(X^k) - \frac{\mu}{L^0_{\max}} \left(f(X^k)-f^{\star}\right).
\end{align*}
The remaining part of the proof follows from the simple observation
\begin{align*}
\log \left( \frac{\Delta_0}{\epsilon} \right) \leq k \frac{\mu}{L^0_{\max}} \leq k \log \left( \frac{1}{1 - \frac{\mu}{L^0_{\max}}} \right).
\end{align*}

\end{proof}

\section{Stochastic case}

\subsection{Adaptive stepsizes}

Before proving the main result from \Cref{gluon:sec:convergence_stochastic}, we first present an attempt to formulate an adaptive stepsize strategy for the stochastic setting. This requires the following assumption:

\begin{assumption}\label{gluon:ass:grad_stochastic}
    The stochastic gradient estimator $\nabla f_{\xi}: \cS \to \cS$ is unbiased and has bounded relative variance. That is, $\mathbb{E}[\nabla f_{\xi}(X)]=\nabla f(X)$ for all $X \in \cS$ and there exists $0\leq\zeta<1$ such that
    \begin{align*}
        \|\nabla_i f_{\xi}(X)-\nabla_i f(X)\|_{(i) \star} \leq \zeta \|\nabla_i f_{\xi}(X)\|_{(i) \star}, \quad \ i = 1, \dots, p
    \end{align*}
    holds almost surely for all $X \in \cS$.
\end{assumption}

This assumption is somewhat unconventional due to the presence of the stochastic gradients on the right-hand side of the inequality. It does not follow from standard conditions and does not fall within known frameworks for modeling stochasticity, such as the ABC inequality of \citet{khaled2020better}. Instead, it introduces a novel structure with parallels to the literature on contractive compression \citep{beznosikov2023biased,demidovich2023guide}.

To elaborate, recall the definition of a contractive compressor:
\begin{definition}[Contractive compressor]\label{gluon:def:contractive_compr}
	A stochastic mapping $\cC:\cS \to \cS$ is called a \emph{contractive compressor} if there exists  $\alpha \in [0,1)$ such that
	\begin{align}\label{gluon:eq:contractive_compr}
		\Exp{\|\cC(X) - X\|^2} \leq (1-\alpha)\|X\|^2
	\end{align}
    for any $X\in\cS$.
\end{definition}
There is a conceptual similarity between \Cref{gluon:ass:grad_stochastic} and the contractive property in~\eqref{gluon:eq:contractive_compr}. \Cref{gluon:ass:grad_stochastic} can be interpreted as asserting that the true gradient $\nabla f(X)$ is effectively a contraction of the stochastic gradient $\nabla f_{\xi}(X)$, with contraction factor $1 - \zeta$. Unlike contractive compressors, there is no explicit mapping from $\nabla f_{\xi}(X)$ to $\nabla f(X)$, and the uniform bound implies the same contraction-like behavior across all stochastic gradients.

Although \Cref{gluon:ass:grad_stochastic} is admittedly strong, it allows us to establish a convergence theorem using an adaptive stepsize strategy similar to the one employed in the deterministic case in \Cref{gluon:theorem:1_full}.

\begin{theorem}\label{gluon:theorem:2}
    Let Assumptions \ref{gluon:ass:generalized-smoothness} and \ref{gluon:ass:grad_stochastic} hold and fix $\varepsilon>0$. Let $X^0,\ldots,X^{K-1}$ be the iterates of \algname{Gluon} (Algorithm \ref{gluon:algo:stochastic}) run with $\beta^k=0$ and $ t^k_i = \frac{(1-\zeta)\|\nabla _i f_{\xi^k}(X^k)\|_{(i) \star}}{L^0_i + (1+\zeta)L^1_i\|\nabla _i f_{\xi^k}(X^k)\|_{(i) \star}}$.
    Then,
    \begin{enumerate}
        \item In order to reach the precision
        \begin{align*}
            \min_{k=0,\ldots,K-1} \sum_{i=1}^p \mathbb{E} \left[\left\| \nabla _i f(X^{k}) \right\|_{(i) \star}\right] \leq \epsilon,
        \end{align*}
        it suffices to run the algorithm for
        \begin{align*}
            K = \left\lceil \frac{2 \sum_{i=1}^p L^0_i \Delta^0}{\left(1 - \zeta \right)^2\epsilon^2} + \frac{2(1+\zeta)L^1_{\max} \Delta^0}{\left(1 - \zeta \right)^2\epsilon} \right\rceil
        \end{align*}
        iterations.

        \item In order to reach the precision
        \begin{align*}
            \min_{k=0,\ldots,K-1} \sum_{i=1}^p \left[\frac{\frac{1}{L_i^1}}{\frac{1}{p} \sum_{j=1}^p \frac{1}{L_j^1}} \norm{\nabla _i f(X^k)}_{(i) \star}\right] \leq \varepsilon,
        \end{align*}
        it suffices to run the algorithm for
        \begin{align*}
            K = \left\lceil \frac{2 \Delta^0 \sum_{i=1}^p \frac{L^0_i}{(L_i^1)^2}}{\varepsilon^2 (1-\zeta)^2 \parens{\frac{1}{p} \sum_{j=1}^p \frac{1}{L_j^1}}^2}
            + \frac{2 \Delta^0 (1+\zeta)}{\varepsilon (1-\zeta)^2 \parens{\frac{1}{p} \sum_{j=1}^p \frac{1}{L_j^1}}} \right\rceil
        \end{align*}
        iterations,
    \end{enumerate}
    where $\Delta^0\eqdef f(X^0) - \inf_{X \in \cS} f(X)$ and $L^1_{\max}\eqdef\max_{i=1,\dots,p} L^1_i$.
\end{theorem}

\begin{proof}
Lemma~\ref{gluon:lemma:generalized-smoothness} with $X=X^k$ and $Y=X^{k+1}$ gives
\begin{align*}
    f(X^{k+1})
    &\leq f(X^k) + \left\langle \nabla f(X^k), X^{k+1}-X^k\right\rangle
    + \sum_{i=1}^p\frac{L^0_i + L^1_i\|\nabla _i f(X^k)\|_{(i) \star}}{2}
    \|X^k_i-X^{k+1}_i\|^2_{(i)} \\
    &= f(X^k) + \sum_{i=1}^p
    \left[ \left\langle \nabla _i f(X^k), X_i^{k+1}-X_i^k\right\rangle_{(i)}
    \right.\\
    &\qquad\qquad\left.
    + \frac{L^0_i + L^1_i\|\nabla _i f(X^k)\|_{(i) \star}}{2}
    \|X^k_i-X^{k+1}_i\|^2_{(i)}\right]\\
    &= f(X^k) + \sum_{i=1}^p \Big[
    \left\langle \nabla_i f_{\xi^k}(X^k), X_i^{k+1}-X_i^k\right\rangle_{(i)}
    \Big.\\
    &\qquad\qquad\Big.
    + \left\langle \nabla _i f(X^k) - \nabla_i f_{\xi^k}(X^k), X_i^{k+1}-X_i^k\right\rangle_{(i)}\Big]\\
    &\qquad + \sum_{i=1}^p
    \frac{L^0_i + L^1_i\|\nabla _i f(X^k)\|_{(i) \star}}{2}
    \|X^k_i-X^{k+1}_i\|^2_{(i)},
\end{align*}
and applying the Cauchy-Schwarz inequality, we get
\begin{align*}
    f(X^{k+1}) \leq f(X^k) + \sum_{i=1}^p \Bigg[ &\left\langle \nabla_i f_{\xi^k}(X^k), X_i^{k+1}-X_i^k\right\rangle_{(i)} \\
    &+ \| \nabla _i f(X^k) - \nabla_i f_{\xi^k}(X^k) \|_{(i) \star} \| X_i^{k+1}-X_i^k \|_{(i)}\\
    &+ \frac{L^0_i + L^1_i\|\nabla _i f(X^k)\|_{(i) \star}}{2}\|X^k_i-X^{k+1}_i\|^2_{(i)}\Bigg].
\end{align*}

The update rule (\ref{gluon:eq:update_rule_stoch}) and the definition of the dual norm $\|\cdot\|_{(i) \star}$ give
\begin{align*}
\|X^k_i-X^{k+1}_i\|^2_{(i)}\leq \left(t^k_i\right)^2
\end{align*}
and
\begin{align*}
\left\langle \nabla_i f_{\xi^k}(X^k), X_i^{k+1}-X_i^k\right\rangle_{(i)} &= \left\langle \nabla_i f_{\xi^k}(X^k), \mathrm{LMO}_{\mathcal{B}_i^k} \left( \nabla_i f_{\xi^k}(X^k) \right)-X_i^k\right\rangle_{(i)} \\
&= - t_i^k \max_{\|X_i\|_{(i)} \leq 1} \left\langle \nabla_i f_{\xi^k}(X^k), X_i\right\rangle_{(i)} \\
&= - t_i^k \| \nabla_i f_{\xi^k}(X^k) \|_{(i) \star}.
\end{align*}

Consequently, using Assumption~\ref{gluon:ass:grad_stochastic}, we obtain
\begin{align*}
    f(X^{k+1}) &\leq f(X^k) + \sum_{i=1}^p \Bigg[ - t_i^k \| \nabla_i f_{\xi^k}(X^k) \|_{(i) \star} + t^k_i \| \nabla _i f(X^k) - \nabla_i f_{\xi^k}(X^k) \|_{(i) \star} \\
    &\qquad\qquad\qquad\qquad+ \frac{L^0_i + L^1_i\|\nabla _i f(X^k)\|_{(i) \star}}{2} \left(t^k_i\right)^2 \Bigg]\\
    &\leq f(X^k) + \sum_{i=1}^p \Bigg[ - (1-\zeta) t_i^k \| \nabla _i f_{\xi^k}(X^k) \|_{(i) \star} \\
    &\qquad\qquad\qquad\qquad+ \frac{L^0_i + (1+\zeta) L^1_i\|\nabla _i f_{\xi^k}(X^k)\|_{(i) \star}}{2} \left(t^k_i\right)^2 \Bigg].
\end{align*}

Minimizing the right-hand side of the last inequality with respect to $ t^k_i $ yields
\begin{align*}
t^k_i = \frac{(1-\zeta)\|\nabla _i f_{\xi^k}(X^k)\|_{(i) \star}}{L^0_i + (1+\zeta)L^1_i\|\nabla _i f_{\xi^k}(X^k)\|_{(i) \star}}.
\end{align*}

This greedy approach for choosing $ t^k_i $ gives the descent inequality
\begin{align*}
f(X^{k+1}) \leq f(X^k) - \sum_{i=1}^p \frac{(1-\zeta)^2\|\nabla _i f_{\xi^k}(X^k)\|^2_{(i) \star}}{2\left(L^0_i + (1+\zeta)L^1_i\|\nabla _i f_{\xi^k}(X^k)\|_{(i) \star}\right)}.
\end{align*}

Taking expectations, we have
\begin{align}\label{gluon:eq:aoisbhr5}
\mathbb{E}[f(X^{k+1})] \leq \mathbb{E}[f(X^k)] - \sum_{i=1}^p \mathbb{E} \left[ \frac{(1-\zeta)^2\|\nabla _i f_{\xi^k}(X^k)\|^2_{(i) \star}}{2\left(L^0_i + (1+\zeta)L^1_i\|\nabla _i f_{\xi^k}(X^k)\|_{(i) \star}\right)}\right].
\end{align}

Now, let us define the function $\phi_i(t):= \frac{(1 - \zeta)^2 t^2}{2(L^0_i+(1+\zeta)L^1_i t)}$. Since $\phi_i(t)$ is convex, Jensen's inequality gives
\begin{align*}
    \mathbb{E}[f(X^k)] - \mathbb{E}[f(X^{k+1})]
    &\geq \sum_{i=1}^p \mathbb{E} \left[ \frac{(1-\zeta)^2\|\nabla _i f_{\xi^k}(X^k)\|^2_{(i) \star}}{2\left(L^0_i + (1+\zeta)L^1_i\|\nabla _i f_{\xi^k}(X^k)\|_{(i) \star}\right)}\right] \\
    &\geq \sum_{i=1}^p \frac{(1-\zeta)^2\left(\mathbb{E} \left[ \|\nabla _i f_{\xi^k}(X^k)\|_{(i) \star}\right]\right)^2}{2\left(L^0_i + (1+\zeta)L^1_i\mathbb{E} \left[ \|\nabla _i f_{\xi^k}(X^k)\|_{(i) \star}\right]\right)}.
\end{align*}
By Jensen's inequality and \Cref{gluon:ass:grad_stochastic}
\begin{align*}
    \mathbb{E} \left[ \left\| \nabla_i f(X^k) \right\|_{(i)\star} \right]
    &= \Exp{\norm{\ExpCond{\nabla_i f_{\xi_k}(X^k)}{X^k}}_{(i)\star}} \\
    &\leq \Exp{\ExpCond{\norm{\nabla_i f_{\xi_k}(X^k)}_{(i)\star}}{X^k}} \\
    &= \mathbb{E} \left[ \left\| \nabla_i f_{\xi_k}(X^k) \right\|_{(i)\star} \right],
\end{align*}
and hence, using the fact that $\phi_i$ is increasing, we get
\begin{align*}
    \mathbb{E}[f(X^k)] - \mathbb{E}[f(X^{k+1})]
    &\geq \sum_{i=1}^p \frac{(1-\zeta)^2\left(\Exp{\norm{\nabla _i f(X^k)}_{(i) \star}}\right)^2}{2\left(L^0_i + (1+\zeta)L^1_i \Exp{\norm{\nabla _i f(X^k)}_{(i) \star}}\right)}.
\end{align*}
Summing the terms gives
\begin{equation}
    \begin{aligned}
    \sum_{k=0}^{K-1} \sum_{i=1}^p \frac{(1-\zeta)^2\left(\Exp{\norm{\nabla _i f(X^k)}_{(i) \star}}\right)^2}{2\left(L^0_i + (1+\zeta)L^1_i \Exp{\norm{\nabla _i f(X^k)}_{(i) \star}}\right)}
    &\leq \sum_{k=0}^{K-1} \parens{\mathbb{E}[f(X^k)] - \mathbb{E}[f(X^{k+1})]} \\
    &= \mathbb{E}[f(X^0)] - \mathbb{E}[f(X^K)] \\
    &\leq f(X^0) - \inf_{X \in \cS} f(X) =: \Delta^0,
    \end{aligned}
    \label{gluon:eq:artjnsd32}
\end{equation}
The remaining part of the proof closely follows the proof of Theorem~\ref{gluon:theorem:1_full}.
We can proceed in two ways:
\begin{enumerate}
    \item Upper-bounding $L^1_i$ by $L^1_{\max}:=\max_{i=1,\dots,p} L^1_i$ in \eqref{gluon:eq:artjnsd32}, we obtain
    \begin{align}\label{gluon:eq:sarja}
        \sum_{k=0}^{K-1} \sum_{i=1}^p \frac{(1-\zeta)^2\left(\Exp{\norm{\nabla _i f(X^k)}_{(i) \star}}\right)^2}{2\left(L^0_i + (1+\zeta) L^1_{\max} \Exp{\norm{\nabla _i f(X^k)}_{(i) \star}}\right)}
        \leq \Delta^0.
    \end{align}
    Now, \Cref{gluon:lemma:ineq} with
    \[
    \begin{aligned}
    x_i &= 1,\\
    y_i &= (1-\zeta) \Exp{\|\nabla _i f(X^{k})\|_{(i) \star}},\\
    z_i &= 2 \parens{L^0_i + (1+\zeta) L^1_{\max} \Exp{\norm{\nabla _i f(X^k)}_{(i) \star}}},
    \end{aligned}
    \]
    gives
    \begin{align*}
        \phi\parens{\sum_{i=1}^p \Exp{\|\nabla _i f(X^{k})\|_{(i) \star}}}
        &= \frac{\parens{(1-\zeta) \sum_{i=1}^p \Exp{\|\nabla _i f(X^{k})\|_{(i) \star}}}^2}{2 \sum_{i=1}^p \parens{L^0_i + (1+\zeta) L^1_{\max} \Exp{\norm{\nabla _i f(X^k)}_{(i) \star}}}} \\
        &\leq \sum_{i=1}^p \frac{(1-\zeta)^2 \Exp{\|\nabla _i f(X^{k})\|_{(i) \star}}^2}{2 \parens{L^0_i + (1+\zeta) L^1_{\max} \Exp{\norm{\nabla _i f(X^k)}_{(i) \star}}}}
    \end{align*}
    where $\phi(t)\eqdef \frac{(1-\zeta)^2 t^2}{2(\sum_{i=1}^p L^0_i + (1+\zeta) L^1_{\max} t)}$.
    Combining the last inequality with \eqref{gluon:eq:sarja} and using the fact that $\phi$ is increasing, we get
    \begin{align*}
        K \phi\parens{\min_{k=0,\ldots,K-1} \sum_{i=1}^p \Exp{\|\nabla _i f(X^{k})\|_{(i) \star}}}
        \leq \sum_{k=0}^{K-1} \phi\parens{\sum_{i=1}^p \Exp{\|\nabla _i f(X^{k})\|_{(i) \star}}}
        \leq \Delta^0.
    \end{align*}
    and hence
    \begin{align*}
        \min_{k=0,\ldots,K-1} \sum_{i=1}^p \Exp{\|\nabla _i f(X^{k})\|_{(i) \star}} \leq \phi^{-1}\left( \frac{\Delta^0}{K} \right),
    \end{align*}
    where $\phi^{-1}$ denotes the inverse function (which exists since $\phi$ is increasing).
    Therefore, to reach the precision $\min_{k=0,\ldots,K-1} \sum_{i=1}^p \Exp{\|\nabla _i f(X^{k})\|_{(i) \star}} \leq \epsilon$, it suffices to run the algorithm for
    \begin{align*}
        K = \left\lceil \frac{\Delta^0}{\phi(\epsilon)} \right\rceil
        = \left\lceil \frac{2 \Delta^0 \sum_{i=1}^p L^0_i}{(1-\zeta)^2 \epsilon^2} + \frac{2 \Delta^0 (1+\zeta) L^1_{\max}}{(1-\zeta)^2 \epsilon}\right\rceil
    \end{align*}
    iterations.

    \item Alternatively, we can start from inequality \eqref{gluon:eq:artjnsd32} and apply \Cref{gluon:lemma:ineq} with $x_i = \nicefrac{1}{L_i^1}$, $y_i = (1-\zeta) \Exp{\norm{\nabla _i f(X^k)}_{(i) \star}}$ and $z_i = 2 \parens{L^0_i + (1+\zeta) L^1_i \Exp{\norm{\nabla _i f(X^k)}_{(i) \star}}}$ to obtain
    \begin{eqnarray*}
        \Delta^0 &\geq& \sum_{k=0}^{K-1} \sum_{i=1}^p \frac{(1-\zeta)^2 \Exp{\norm{\nabla _i f(X^k)}_{(i) \star}}^2}{2 \parens{L^0_i + (1+\zeta) L^1_i \Exp{\norm{\nabla _i f(X^k)}_{(i) \star}}}} \\
        &\geq& \sum_{k=0}^{K-1} \frac{\parens{\sum_{i=1}^p \frac{1}{L_i^1} (1-\zeta) \Exp{\norm{\nabla _i f(X^k)}_{(i) \star}}}^2}{2 \sum_{i=1}^p \parens{\frac{L^0_i}{(L_i^1)^2} + (1+\zeta) \frac{1}{L_i^1} \Exp{\norm{\nabla _i f(X^k)}_{(i) \star}}}} \\
        &=& \sum_{t=0}^{K-1} \psi{\parens{\sum_{i=1}^p \frac{1}{L_i^1} \Exp{\norm{\nabla _i f(X^k)}_{(i) \star}}}},
    \end{eqnarray*}
    where $\psi(t) \eqdef \frac{(1-\zeta)^2 t^2}{2 \parens{\sum_{i=1}^p \frac{L^0_i}{(L_i^1)^2} + (1+\zeta) t}}$. Since the function $\psi$ is increasing for $t>0$, $\psi^{-1}$ exists. It follows that
    \begin{eqnarray*}
        \Delta^0 &\geq& \sum_{k=0}^{K-1} \psi{\parens{\sum_{i=1}^p \frac{1}{L_i^1} \Exp{\norm{\nabla _i f(X^k)}_{(i) \star}}}} \\
        &\geq& K \psi{\parens{\min_{k=0,\ldots,K-1} \sum_{i=1}^p \frac{1}{L_i^1} \Exp{\norm{\nabla _i f(X^k)}_{(i) \star}}}},
    \end{eqnarray*}
    and hence
    \begin{eqnarray*}
        \min_{k=0,\ldots,K-1} \sum_{i=1}^p \frac{1}{L_i^1} \Exp{\norm{\nabla _i f(X^k)}_{(i) \star}}
        \leq \psi^{-1}\parens{\frac{\Delta^0}{K}}.
    \end{eqnarray*}
    This in turn means that to reach the precision
    \begin{eqnarray*}
        \min_{k=0,\ldots,K-1} \sum_{i=1}^p \left[\frac{\frac{1}{L_i^1}}{\frac{1}{p} \sum_{j=1}^p \frac{1}{L_j^1}} \norm{\nabla _i f(X^k)}_{(i) \star}\right] \leq \varepsilon,
    \end{eqnarray*}
    it suffices to run the algorithm for
    \begin{align*}
        K &= \left\lceil \frac{\Delta^0}{\psi\parens{\varepsilon \parens{\frac{1}{p} \sum_{j=1}^p \frac{1}{L_j^1}}}} \right\rceil \\
        &= \left\lceil \frac{2 \Delta^0 \sum_{i=1}^p \frac{L^0_i}{(L_i^1)^2}}{(1-\zeta)^2 \varepsilon^2 \parens{\frac{1}{p} \sum_{j=1}^p \frac{1}{L_j^1}}^2}
        + \frac{2 \Delta^0 (1+\zeta)}{(1-\zeta)^2 \varepsilon \parens{\frac{1}{p} \sum_{j=1}^p \frac{1}{L_j^1}}} \right\rceil
    \end{align*}
    iterations.
\end{enumerate}
\end{proof}

\subsection{Proof of theorem \ref{gluon:theorem:3}}\label{gluon:sec:thm_stoch_proof}

We now establish the main result of \Cref{gluon:sec:convergence_stochastic}. The guarantees in \Cref{gluon:theorem:3} follow from the more general result below.
Here, $\rho_i>0$ for $i \in [p]$ denote the norm equivalence constants, i.e., $\norm{X_i}_{(i)\star} \leq \rho_i \norm{X_i}_2$ for all $X_i\in\cS_i$.

\begin{theorem}\label{gluon:theorem:3_appendix}
    Let Assumptions \ref{gluon:ass:generalized-smoothness} and \ref{gluon:ass:bounded_var} hold and fix $\varepsilon>0$. Let $X^0,\ldots,X^{K-1}$ be the iterates of \algname{Gluon} (Algorithm \ref{gluon:algo:stochastic}) run with $ \beta^k = 1 - (k+1)^{-1/2} $, $ t_i^k = t_i (k+1)^{-3/4} $ for some $ t_i>0$, and $M^0_i=\nabla_i f_{\xi^0}(X^0)$.
    \begin{enumerate}
        \item If $L^1_i=0$, then
	    \begin{align*}
	        &\min_{k=0,\ldots,K-1} \sum_{i=1}^p t_i \mathbb{E}\left[\| \nabla _i f(X^k) \|_{(i) \star}\right] \\
	        &\hspace{1cm}\leq \frac{\Delta^0}{K^{1/4}}
	        + \frac{1}{K^{1/4}} \sum_{i=1}^p
	        \Bigg[ \sigma \rho_i t_i \left(7+2\sqrt{2 e^2} \log (K)\right)
	        + L^0_i t^2_i \left(\frac{87}{2} + 14\log (K)\right)\Bigg],
        \end{align*}
        \item If $L^1_i\neq0$, then for $ t_i = \frac{1}{12L^1_i} $, we have
	    \begin{align*}
	        &\min_{k=0,\ldots,K-1} \sum_{i=1}^p \frac{1}{12L^1_i} \mathbb{E}\left[\| \nabla _i f(X^k) \|_{(i) \star}\right] \\
	        &\hspace{1cm}\leq \frac{2\Delta^0}{K^{1/4}}
	        + \frac{1}{K^{1/4}} \sum_{i=1}^p
	        \Bigg[ \frac{\sigma \rho_i}{6L^1_i}\left(7+2\sqrt{2 e^2} \log (K)\right) \\
	        &\hspace{3cm}
	        + \frac{L^0_i}{144(L^1_i)^2}\left(87 +28\log (K)\right)\Bigg],
        \end{align*}
    \end{enumerate}
    where $\Delta^0:=f(X^0) - \inf_{X\in\cS} f(X)$.
\end{theorem}

\begin{proof}
We again start with the result in Lemma~\ref{gluon:lemma:generalized-smoothness} with $X=X^k$ and $Y=X^{k+1}$, obtaining
\begin{align*}
    f(X^{k+1})
    &\leq f(X^k) + \left\langle \nabla f(X^k), X^{k+1}-X^k\right\rangle\\
    &\qquad + \sum_{i=1}^p \frac{L^0_i + L^1_i\|\nabla _i f(X^k)\|_{(i) \star}}{2}
    \cdot \|X^k_i-X^{k+1}_i\|^2_{(i)}\\
    &= f(X^k) + \sum_{i=1}^p \Bigg[
    \left\langle \nabla _i f(X^k), X_i^{k+1}-X_i^k\right\rangle_{(i)}\\
    &\qquad\qquad
    + \frac{L^0_i + L^1_i\|\nabla _i f(X^k)\|_{(i) \star}}{2}
    \|X^k_i-X^{k+1}_i\|^2_{(i)}\Bigg]\\
    &= f(X^k) + \sum_{i=1}^p \Bigg[
    \left\langle M_i^k, X_i^{k+1}-X_i^k\right\rangle_{(i)}
    \Bigg.\\
    &\qquad\qquad\Bigg.
    + \left\langle \nabla _i f(X^k) - M_i^k, X_i^{k+1}-X_i^k\right\rangle_{(i)}\Bigg]\\
    &\qquad+ \sum_{i=1}^p
    \frac{L^0_i + L^1_i\|\nabla _i f(X^k)\|_{(i) \star}}{2}
    \cdot \|X^k_i-X^{k+1}_i\|^2_{(i)}.
\end{align*}

Applying the Cauchy-Schwarz inequality, we have
\begin{align*}
    f(X^{k+1})
    &\leq f(X^k) + \sum_{i=1}^p \Bigg[
    \left\langle M_i^k, X_i^{k+1}-X_i^k\right\rangle_{(i)}
    + \| \nabla _i f(X^k) - M_i^k \|_{(i) \star}
    \| X_i^{k+1}-X_i^k \|_{(i)}\Bigg]\\
    &\qquad+\sum_{i=1}^p
    \frac{L^0_i + L^1_i\|\nabla _i f(X^k)\|_{(i) \star}}{2}
    \|X^k_i-X^{k+1}_i\|^2_{(i)}.
\end{align*}
Now, the update rule (\ref{gluon:eq:update_rule_stoch}) and the definition of the dual norm $\|\cdot\|_{(i) \star}$ give
\begin{align*}
\|X^k_i-X^{k+1}_i\|^2_{(i)}\leq \left(t^k_i\right)^2
\end{align*}
and
\begin{align*}
\left\langle M_i^k, X_i^{k+1}-X_i^k\right\rangle
&= \left\langle M_i^k, \mathrm{LMO}_{\mathcal{B}_i^k} \left( M_i^k \right)-X_i^k\right\rangle \\
&= - t_i^k \max_{\|X_i\|_{(i)} \leq 1} \left\langle M_i^k, X_i\right\rangle
= - t_i^k \| M_i^k \|_{(i) \star}.
\end{align*}

Consequently,
\begin{align*}
f(X^{k+1})
&\leq f(X^k) + \sum_{i=1}^p \Bigg[
- t_i^k \| M_i^k \|_{(i) \star}
+ t^k_i \| \nabla _i f(X^k) - M_i^k \|_{(i) \star}\\
&\qquad\qquad\qquad\qquad
+ \frac{L^0_i + L^1_i\|\nabla _i f(X^k)\|_{(i) \star}}{2} \left(t^k_i\right)^2
\Bigg]\\
&= f(X^k) + \sum_{i=1}^p \Bigg[
- t_i^k \| M_i^k - \nabla _i f(X^k) + \nabla _i f(X^k) \|_{(i) \star}\\
&\qquad\qquad\qquad\qquad
+ t^k_i \|M_i^k - \nabla _i f(X^k)\|_{(i) \star} \Bigg]\\
&\qquad + \sum_{i=1}^p
    \frac{L^0_i + L^1_i\|\nabla _i f(X^k)\|_{(i) \star}}{2} \left(t^k_i\right)^2 \\
&\leq f(X^k) + \sum_{i=1}^p \left[
- t_i^k \| \nabla _i f(X^k) \|_{(i) \star}
+ 2t^k_i \|M_i^k - \nabla _i f(X^k)\|_{(i) \star}\right] \\
&\qquad + \sum_{i=1}^p
\frac{L^0_i + L^1_i\|\nabla _i f(X^k)\|_{(i) \star}}{2} \left(t^k_i\right)^2 .
\end{align*}

Taking expectations, we obtain
\begin{align*}
\mathbb{E}[f(X^{k+1})]
&\leq \mathbb{E}[f(X^k)] + \sum_{i=1}^p \Bigg[
- t_i^k \mathbb{E}[\| \nabla _i f(X^k) \|_{(i) \star}]\\
&\qquad + 2t^k_i \mathbb{E}\left[\left\|M_i^k - \nabla_i f(X^k)\right\|_{(i) \star}\right]
+ \frac{L^0_i + L^1_i\mathbb{E}[\|\nabla _i f(X^k)\|_{(i) \star}]}{2} \left(t^k_i\right)^2 \Bigg].
\end{align*}

Telescoping the last inequality gives
\begin{equation}\label{gluon:eq:main_eq_momentum}
\begin{aligned}
\sum_{i=1}^p \sum_{k=0}^{K-1} t_i^k \mathbb{E}[\| \nabla _i f(X^k) \|_{(i) \star}]
&\leq \Delta^0 + \sum_{i=1}^p \Bigg[
2 \sum_{k=0}^{K-1} t^k_i \mathbb{E}\left[\left\|M_i^k - \nabla_i f(X^k)\right\|_{(i) \star}\right]\\
&\qquad + \sum_{k=0}^{K-1} \frac{L^0_i}{2}\left(t^k_i\right)^2\\
&\qquad + \sum_{k=0}^{K-1} \frac{L^1_i}{2} \mathbb{E}[\|\nabla _i f(X^k)\|_{(i) \star}]\left(t^k_i\right)^2 \Bigg].
\end{aligned}
\end{equation}
where $\Delta^0:=f(X^0) - \inf_{X \in \cS} f(X)$.

Now, inspired by the analysis in \citet{hubler2024parameter}, we introduce the following notation: $\mu_i^k:=M_i^k - \nabla_i f(X^k)$, $\gamma_i^k:=\nabla_i f_{\xi^k}(X^k) - \nabla_i f(X^k)$, $\alpha^k=1-\beta^k$, $\beta^{a: b}:=\prod_{k=a}^b \beta^k$ and $S_i^k:=\nabla_i f(X^{k-1}) - \nabla_i f(X^{k})$. Then, we can rewrite the algorithm's momentum update rule as
\begin{align*}
M_i^k &= \beta^k M^{k-1}_i+(1-\beta^k)\nabla_i f_{\xi^k}(X^k)\\
&=\beta^k \left(\mu_i^{k-1}+\nabla_i f(X^{k-1})\right)+(1-\beta^k)\left(\gamma_i^k+\nabla_i f(X^k)\right) \\
&= \nabla_i f\left(X^k\right)+\alpha^k \gamma_i^k+\beta^k S_i^k+\beta^k \mu_i^{k-1}.
\end{align*}

This yields
\begin{align*}
\mu_i^k &=M_i^k-\nabla_i f\left(X^k\right) \\
&=\alpha^k \gamma_i^k+\beta^k S_i^k+\beta^k \mu_i^{k-1} \\
&=\sum_{\tau=1}^{k} \beta^{(\tau+1): k} \alpha^{\tau} \gamma_i^{\tau}+\sum_{\tau=1}^k \beta^{\tau: k} S_i^\tau+\beta^{1: k} \mu^0_i\\
&=\sum_{\tau=0}^{k} \beta^{(\tau+1): k} \alpha^{\tau} \gamma_i^{\tau}+\sum_{\tau=1}^k \beta^{\tau: k} S_i^\tau,
\end{align*}
where the last line follows from the fact that $M^0_i=\nabla_i f_{\xi^0}(X^0)$ and $\beta^0=0$.
Thus,
\begin{align*}
\mathbb{E}\left[\left\|M_i^k - \nabla_i f(X^k)\right\|_{(i) \star}\right]
&=\mathbb{E}\left[\left\|\mu_i^k\right\|_{(i) \star}\right] \\
&\leq \mathbb{E}\left[\left\|\sum_{\tau=0}^{k} \beta^{(\tau+1): k} \alpha^{\tau} \gamma_i^{\tau}\right\|_{(i) \star}\right]+\sum_{\tau=1}^k \beta^{\tau: k} \mathbb{E}\left[\left\|S_i^\tau\right\|_{(i) \star}\right]\\
&\leq \rho_i \mathbb{E}\left[\left\|\sum_{\tau=0}^{k} \beta^{(\tau+1): k} \alpha^{\tau} \gamma_i^{\tau}\right\|_{2}\right]+\sum_{\tau=1}^k \beta^{\tau: k} \mathbb{E}\left[\left\|S_i^\tau\right\|_{(i) \star}\right]\\
&\leq \rho_i\sqrt{\sum_{\tau=0}^{k} \left(\beta^{(\tau+1): k} \alpha^{\tau}\right)^2 \mathbb{E}\left[\left\|\gamma_i^{\tau}\right\|_{2}^2\right]}+\sum_{\tau=1}^k \beta^{\tau: k} \mathbb{E}\left[\left\|S_i^\tau\right\|_{(i) \star}\right],
\end{align*}
where in the last line we used Jensen's inequality and the fact that for all $ q < l $
\begin{align*}
    \mathbb{E} \left[ ({\gamma^l_i})^\top \gamma^q_i \right] &= \mathbb{E} \left[ \mathbb{E} \left[ ({\gamma^l_i})^\top \gamma^q_i \mid X^l_i \right] \right] = \mathbb{E} \left[ \mathbb{E} \left[ \gamma^l_i \mid X^l_i \right]^\top \gamma^q_i \right] \\
    &= \mathbb{E} \left[ \left( \mathbb{E} \left[ \nabla_i f_{\xi^l}(X^l) - \nabla_i f(X^l) \mid X^l_i \right] \right)^\top \gamma^q_i \right] = 0.
\end{align*}
Using Assumptions~\ref{gluon:ass:generalized-smoothness} and \ref{gluon:ass:bounded_var}, we get
\begin{align*}
\mathbb{E}\left[\left\|\gamma_i^{\tau}\right\|_2^2\right] = \mathbb{E} \Bigg[ \underbrace{\mathbb{E}\left[\left\|\gamma_i^{\tau}\right\|_2^2 \mid X^{\tau}_i \right]}_{\leq \sigma^2}\Bigg] \leq \sigma^2
\end{align*}
and
\begin{align*}
\left\|S_i^\tau\right\|_{(i) \star} \leq \left( L^0_i + L^1_i \| \nabla _i f(X^{\tau}) \|_{(i) \star} \right) \|X^{\tau+1}_i - X^{\tau}_i \|_{(i)} \leq \left( L^0_i + L^1_i \| \nabla _i f(X^{\tau}) \|_{(i) \star} \right) t^{\tau}_i.
\end{align*}

Therefore,
\begin{eqnarray*}
\mathbb{E}\left[\left\|M_i^k - \nabla_i f(X^k)\right\|_{(i) \star}\right] &\leq& \sigma \rho_i \sqrt{\sum_{\tau=0}^{k} \left(\beta^{(\tau+1): k} \alpha^{\tau} \right)^2} + L^0_i \sum_{\tau=1}^k \beta^{\tau: k} t^{\tau}_i \\
&&+ L^1_i \sum_{\tau=1}^k \beta^{\tau: k} t^{\tau}_i \mathbb{E} \left[\| \nabla _i f(X^{\tau}) \|_{(i) \star}\right].
\end{eqnarray*}

Combining the last inequality with (\ref{gluon:eq:main_eq_momentum}) gives
\begin{align}\label{gluon:eq:main_eq_momentum_2}
\sum_{i=1}^p \sum_{k=0}^{K-1} t_i^k \mathbb{E}[\| \nabla _i& f(X^k) \|_{(i) \star}] \nonumber \\
\leq \Delta^0 +\sum_{i=1}^p \Bigg[ & \underbrace{2 \sigma \rho_i \sum_{k=0}^{K-1} t^k_i \sqrt{\sum_{\tau=0}^{k} \left(\beta^{(\tau+1): k} \alpha^{\tau}\right)^2}}_{=:I_1} + \underbrace{2 L^0_i \sum_{k=0}^{K-1} t^k_i \sum_{\tau=1}^k \beta^{\tau: k} t^{\tau}_i}_{=:I_2}\notag\\
&+ \underbrace{2 L^1_i \sum_{k=0}^{K-1} t^k_i \sum_{\tau=1}^k \beta^{\tau: k} t^{\tau}_i \mathbb{E} \left[\| \nabla _i f(X^{\tau}) \|_{(i) \star}\right]}_{=:I_3}\notag\\
&+ \underbrace{\frac{L^0_i}{2} \sum_{k=0}^{K-1} \left(t^k_i\right)^2}_{=:I_4} + \frac{L^1_i}{2} \sum_{k=0}^{K-1} \left(t^k_i\right)^2 \mathbb{E}\left[\|\nabla _i f(X^k)\|_{(i) \star}\right]\Bigg].
\end{align}

Let us now upper-bound each term $ I_i $, $ i = 1, 2, 3, 4$.

$ I_1 $: using Lemma~\ref{gluon:lemma:tech2}, we obtain
\begin{align*}
I_1 \leq \sigma \rho_i t_i\left(7+2\sqrt{2 e^2} \log (K)\right).
\end{align*}

$ I_2 $: using Lemma~\ref{gluon:lemma:tech2}, we obtain
\begin{align*}
I_2 \leq 14 L^0_i t_i^2\left(3 +\log (K)\right).
\end{align*}

$ I_3 $: rearranging the sums and using Lemma~\ref{gluon:lemma:tech1} with $a=\tau+1$, $b=K$, $p=3/4$ and $q=1/2$, we have
\begin{align*}
I_3 &= 2 L^1_i \sum_{k=0}^{K-1} t^k_i \sum_{\tau=1}^k \beta^{\tau: k} t^{\tau}_i \mathbb{E} \left[\| \nabla _i f(X^{\tau}) \|_{(i) \star}\right] \\
&= 2 L^1_i \sum_{\tau=1}^{K-1} t^\tau_i \left(\sum_{k=\tau}^{K-1} t_i^k \beta^{\tau: k}\right) \mathbb{E} \left[\| \nabla _i f(X^{\tau}) \|_{(i) \star}\right]\\
&= 2 L^1_i \sum_{\tau=1}^{K-1} t^\tau_i t_i \left(\sum_{k=\tau}^{K-1} (k+1)^{-3/4} \beta^{\tau: k}\right) \mathbb{E} \left[\| \nabla _i f(X^{\tau}) \|_{(i) \star}\right]\\
&\leq 2 L^1_i \sum_{\tau=1}^{K-1} t^\tau_i t_i \tau^{-1/4} \underbrace{e^{2\left((\tau+1)^{1/2}-\tau^{1/2}\right)}}_{\leq e^{2(\sqrt{2}-1)} \ \text{for} \ \tau \geq 1} \mathbb{E} \left[\| \nabla _i f(X^{\tau}) \|_{(i) \star}\right]\\
&\leq 2 e^{2(\sqrt{2}-1)} L^1_i \sum_{\tau=1}^{K-1} t^\tau_i t_i \tau^{-1/4} \mathbb{E} \left[\| \nabla _i f(X^{\tau}) \|_{(i) \star}\right]\\
&\leq 2 e^{2(\sqrt{2}-1)} L^1_i \sum_{k=0}^{K-1} t^k_i t_i \mathbb{E} \left[\| \nabla _i f(X^k) \|_{(i) \star}\right].
\end{align*}

$ I_4 $:
\begin{align*}
I_4 &= \frac{L^0_i}{2} \sum_{k=0}^{K-1} \left(t^k_i\right)^2 \leq \frac{L^0_i}{2} \sum_{k=0}^{\infty} \left(t^k_i\right)^2 = \frac{L^0_i}{2} t_i^2 \sum_{k=0}^{\infty} (1+k)^{-3/2} \\
&\leq \frac{L^0_i}{2} t_i^2 \left( 1 + \int_1^{\infty} \frac{1}{z^{3/2}} \, dz \right) = \frac{3L^0_i}{2} t_i^2.
\end{align*}

Combining the upper-bounds for $ I_i $, $ i = 1, 2, 3, 4$ with (\ref{gluon:eq:main_eq_momentum_2}) gives
\begin{align*}
\sum_{i=1}^p \sum_{k=0}^{K-1} t_i^k \mathbb{E}[\| \nabla _i f(X^k) \|_{(i) \star}]
&\leq \Delta^0 +\sum_{i=1}^p \Bigg[ \sigma \rho_i t_i\left(7+2\sqrt{2 e^2} \log (K)\right) \\
&\qquad + 14 L^0_i t_i^2\left(3 +\log (K)\right)\\
&+ 2 e^{2(\sqrt{2}-1)} L^1_i \sum_{k=0}^{K-1} t^k_i t_i \mathbb{E} \left[\| \nabla _i f(X^k) \|_{(i) \star}\right]\\
&+ \frac{3L^0_i}{2} t_i^2 + \frac{L^1_i}{2} \sum_{k=0}^{K-1} \left(t^k_i\right)^2 \mathbb{E}[\|\nabla _i f(X^k)\|_{(i) \star}]\Bigg].
\end{align*}

Using the fact that $ t^k_i = t_i (1+k)^{-3/4} \leq t_i $, and denoting $ C := 2e^{2(\sqrt{2} - 1)} + \frac{1}{2} \leq 5.1$, we get
\begin{align*}
\sum_{i=1}^p \sum_{k=0}^{K-1} t_i^k \mathbb{E}[\| \nabla _i f(X^k) \|_{(i) \star}]
&\leq \Delta^0 +\sum_{i=1}^p \Bigg[ \sigma \rho_i t_i\left(7+2\sqrt{2 e^2} \log (K)\right) \\
&\qquad + 14 L^0_i t_i^2\left(\frac{87}{28} +\log (K)\right)\\
&+ C L^1_i t_i \sum_{k=0}^{K-1} t^k_i \mathbb{E} \left[\| \nabla _i f(X^k) \|_{(i) \star}\right]\Bigg].
\end{align*}

Now, let us consider two options: (1) $L^1_i = 0$ for all $i \in \{1, \dots, p\}$ and (2) $L^1_i \neq 0$, for all $i \in \{1, \dots, p\}$.

\paragraph{Case 1: $L^1_i = 0$, $i = 1, \dots, p$.}

In this case,
\begin{align*}
\sum_{i=1}^p \sum_{k=0}^{K-1} t_i^k \mathbb{E}[\| \nabla _i f(X^k) \|_{(i) \star}]
&\leq \Delta^0 +\sum_{i=1}^p \Bigg[
\sigma \rho_i t_i\left(7+2\sqrt{2 e^2} \log (K)\right)\\
&\qquad + 14 L^0_i t_i^2\left(\frac{87}{28} +\log (K)\right)\Bigg],
\end{align*}
and therefore,
\begin{eqnarray*}
&&\hspace{-2cm}\min_{k=0,\ldots,K-1} \sum_{i=1}^p t_i \mathbb{E}[\| \nabla _i f(X^k) \|_{(i) \star}] \\
&\leq& \frac{1}{K} \sum_{k=0}^{K-1} \sum_{i=1}^p t_i \mathbb{E}[\| \nabla _i f(X^k) \|_{(i) \star}]\\
&\leq& \frac{1}{K^{1/4}} \sum_{k=0}^{K-1} \sum_{i=1}^p t_i (1+k)^{-3/4} \mathbb{E}[\| \nabla _i f(X^k) \|_{(i) \star}]\\
&=& \frac{1}{K^{1/4}} \sum_{k=0}^{K-1} \sum_{i=1}^p t^k_i \mathbb{E}[\| \nabla _i f(X^k) \|_{(i) \star}]\\
&\leq& \frac{\Delta^0}{K^{1/4}} + \frac{1}{K^{1/4}} \sum_{i=1}^p \Bigg[ \sigma \rho_i t_i \left(7+2\sqrt{2 e^2} \log (K)\right) + L^0_i t^2_i \left(\frac{87}{2} + 14\log (K)\right)\Bigg].
\end{eqnarray*}

\paragraph{Case 2:  $L^1_i \neq 0$, $i = 1, \dots, p$.}

Let us choose $t_i=\frac{1}{12L^1_i}$. Then
\begin{align*}
\sum_{i=1}^p \sum_{k=0}^{K-1} t_i^k \mathbb{E}[\| \nabla _i f(X^k) \|_{(i) \star}]
&\leq 2\Delta^0 \\
&\quad +\sum_{i=1}^p \Bigg[ 2\sigma \rho_i t_i\left(7+2\sqrt{2 e^2} \log (K)\right) + L^0_i t_i^2\left(87 +28\log (K)\right)\Bigg],
\end{align*}
and hence
\begin{eqnarray*}
&&\hspace{-1.1cm}\min_{k=0,\ldots,K-1} \sum_{i=1}^p \frac{1}{12L^1_i} \mathbb{E}[\| \nabla _i f(X^k) \|_{(i) \star}] \\
&\leq& \frac{1}{K} \sum_{k=0}^{K-1} \sum_{i=1}^p t_i \mathbb{E}[\| \nabla _i f(X^k) \|_{(i) \star}]\\
&\leq& \frac{1}{K^{1/4}} \sum_{k=0}^{K-1} \sum_{i=1}^p t_i (1+k)^{-3/4} \mathbb{E}[\| \nabla _i f(X^k) \|_{(i) \star}]\\
&=& \frac{1}{K^{1/4}} \sum_{i=1}^p \sum_{k=0}^{K-1} t_i^k \mathbb{E}[\| \nabla _i f(X^k) \|_{(i) \star}]\\
&\leq& \frac{2\Delta^0}{K^{1/4}} + \frac{1}{K^{1/4}} \sum_{i=1}^p \Bigg[ \frac{\sigma \rho_i}{6L^1_i}\left(7+2\sqrt{2 e^2} \log (K)\right) + \frac{L^0_i}{144(L^1_i)^2}\left(87 +28\log (K)\right)\Bigg].
\end{eqnarray*}

\end{proof}

\section{Additional experimental results and details} \label{gluon:appendix:more_exp}

\subsection{Experimental details}\label{gluon:sec:exp_details}

All experiments for the \texttt{NanoGPT} model are conducted using PyTorch\footnote{PyTorch Documentation. Available at: \url{https://pytorch.org/docs/stable/index.html}} with Distributed Data Parallel (DDP)\footnote{Distributed Data Parallel (DDP) in PyTorch. Available at: \url{https://pytorch.org/docs/stable/notes/ddp.html}} across 4 NVIDIA A100 GPUs (40GB each). For the \texttt{CNN} experiments, training is performed on a single NVIDIA A100 GPU (40GB). The training and evaluation pipelines are implemented using open-source codebases \citep{jordan2024cifar10,jordan2024moddednanogpt,code_scion}, with all modifications clearly documented and properly referenced where applicable.

For LMO-based methods, we compute inexact LMOs using the Newton–Schulz iteration when an analytical solution is unavailable (e.g., for SVD-type updates), following the approach proposed by \citet{jordan2024muon}. This method provides a computationally efficient approximation of the required orthogonalization while preserving the convergence behavior of the overall algorithm.

\subsection{\texorpdfstring{Fitting $L^0_i$ and $L^1_i$}{Fitting L\textunderscore i\textasciicircum 0 and L\textunderscore i\textasciicircum 1}}\label{gluon:sec:nanogpt_l0l1}
To minimize the Euclidean error between the true value $ \hat{L}_i[k] $ and its approximation $ \hat{L}_i^{\text{approx}}[k] $, while penalizing underestimation, we incorporate a hinge-like penalty term. Specifically, we fit $L^0_i$ and $L^1_i$ by minimizing the loss function
\begin{align}\label{gluon:eq:l0l1_loss}
    \mathcal{L}_i\parens{L^0_i, L^1_i} := \sum_{k=0}^{K-1} \left( \hat{L}_i[k] - \hat{L}_i^{\text{approx}}[k] \right)^2 + \lambda \sum_{k=0}^{K-1} \max\left(0, \hat{L}_i[k] - \hat{L}_i^{\text{approx}}[k]\right)^2.
\end{align}

The first term of $\mathcal{L}_i$ captures the standard Euclidean (squared) error, while the second term introduces an additional penalty proportional to the amount of underestimation (i.e., when $ \hat{L}_i[k] > \hat{L}_i^{\text{approx}}[k] $). The hyperparameter $ \lambda \geq 0 $ controls the strength of this penalty.

\subsection{Training NanoGPT on FineWeb}\label{gluon:sec:nanogpt_app}

In this section, we present additional results and experimental details for the experiment described in the main text, which involves training a \texttt{NanoGPT} model on the \texttt{FineWeb} dataset using the \algname{unScion} optimizer.

\subsubsection{Empirical validation of \Cref{gluon:ass:generalized-smoothness}}

We begin by presenting additional results for the experiments described in \Cref{gluon:sec:experiments}, aimed at empirically validating \Cref{gluon:ass:generalized-smoothness}. We plot the estimated \emph{trajectory smoothness}
\begin{align*}
    \hat{L}_i[k] \eqdef \frac{\|\nabla_i f_{\xi^{k+1}} (X^{k+1}) - \nabla_i f_{\xi^k} (X^{k}) \|_{(i)\star}}{\|X_i^{k+1} - X_i^{k}\|_{(i)}}
\end{align*}
and its approximation
$$\hat{L}_i^{\text{approx}}[k] \eqdef L_i^0 + L_i^1 \|\nabla_i f_{\xi^{k+1}}(X^{k+1})\|_{(i) \star}$$
as functions of the iteration index $k$, where $L^0_i, L^1_i\geq0$ are fitted using the procedure described in \Cref{gluon:sec:nanogpt_l0l1}.

Figures~\ref{gluon:fig:1}, \ref{gluon:fig:2}, and \ref{gluon:fig:3_appndx} show results for parameter groups from the embedding layer and from the 4th and~8th transformer blocks. Similar patterns are observed across all layers. In each case, we see a strong agreement between $\hat{L}_i[k]$ and $\hat{L}_i^{\text{approx}}[k]$, suggesting that \Cref{gluon:ass:generalized-smoothness} holds approximately along the optimization trajectory.
\begin{figure}[h]
    \centering
    \begin{minipage}{0.32\textwidth}
        \includegraphics[width=\textwidth]{KAUST_Thesis/imported/06_gluon/plots/nanogpt/main_exp/plot7_0.pdf}
    \end{minipage}
    \begin{minipage}{0.32\textwidth}
        \includegraphics[width=\textwidth]{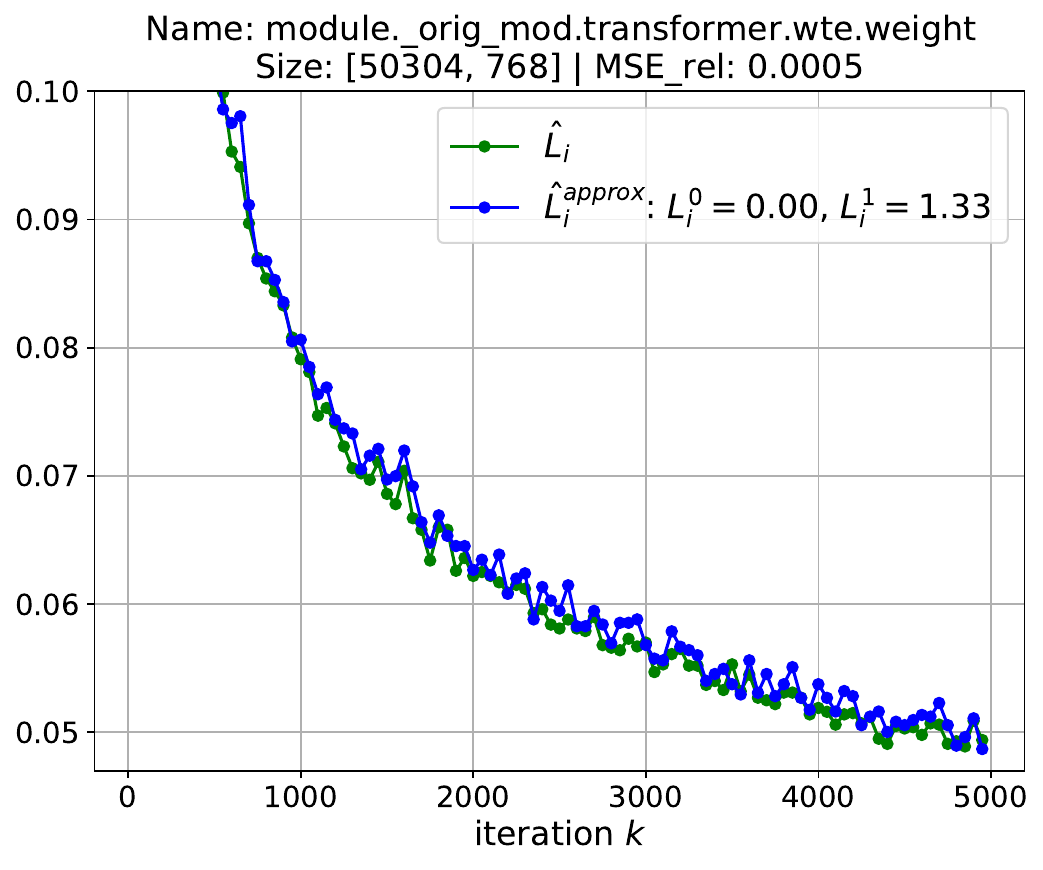}
    \end{minipage}
    \caption{\small {\bf Validation of layer-wise $(L^0, L^1)$-smoothness} for the group of parameters from the embedding layer of \texttt{NanoGPT-124M} along \algname{unScion} training trajectories. The group norm is $\|\cdot\|_{(p)} = n_p \|\cdot\|_{1 \to \infty} $, with fitted values $L_p^0 \approx 0$, $L_p^1 \approx 1.3$. The same plot is shown twice with different $y$-axis limits.}
    \label{gluon:fig:1}
\end{figure}
\begin{figure}[p]
    \centering
    \begin{minipage}{0.32\textwidth}
        \includegraphics[width=\textwidth]{KAUST_Thesis/imported/06_gluon/plots/nanogpt/main_exp/plot1.pdf}
    \end{minipage}
    \hfill
    \begin{minipage}{0.32\textwidth}
        \includegraphics[width=\textwidth]{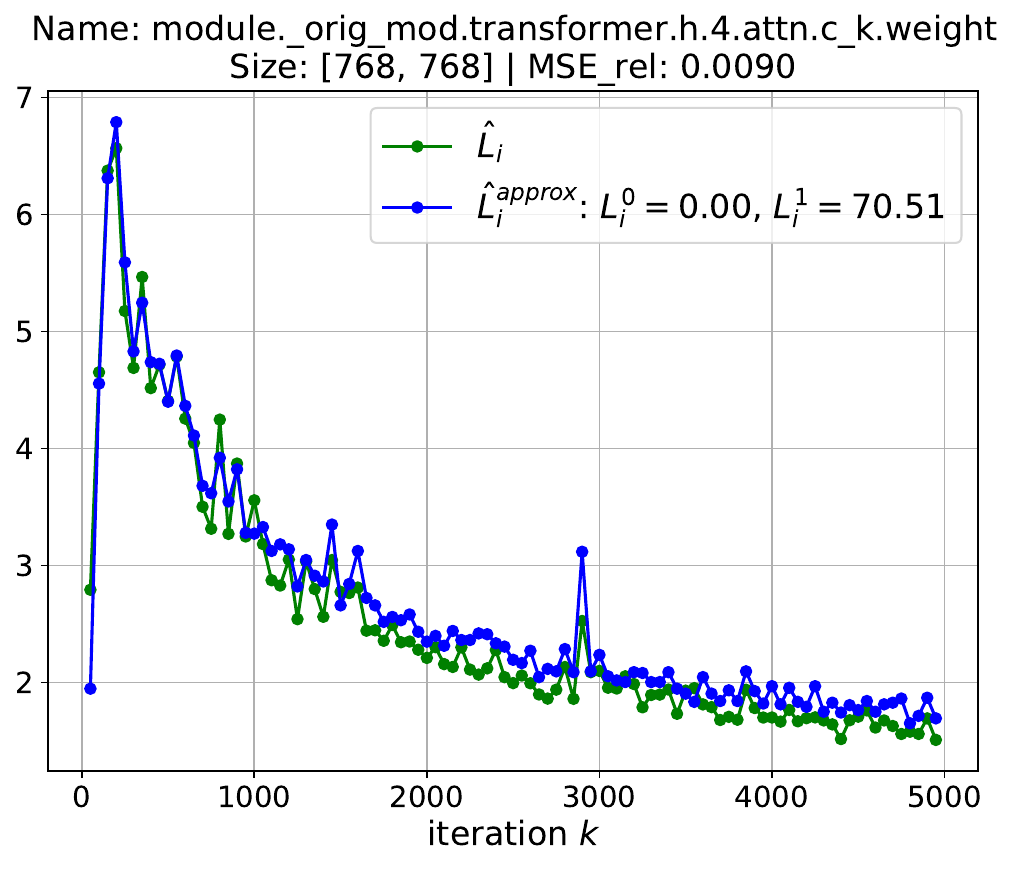}
    \end{minipage}
    \hfill
    \begin{minipage}{0.32\textwidth}
        \includegraphics[width=\textwidth]{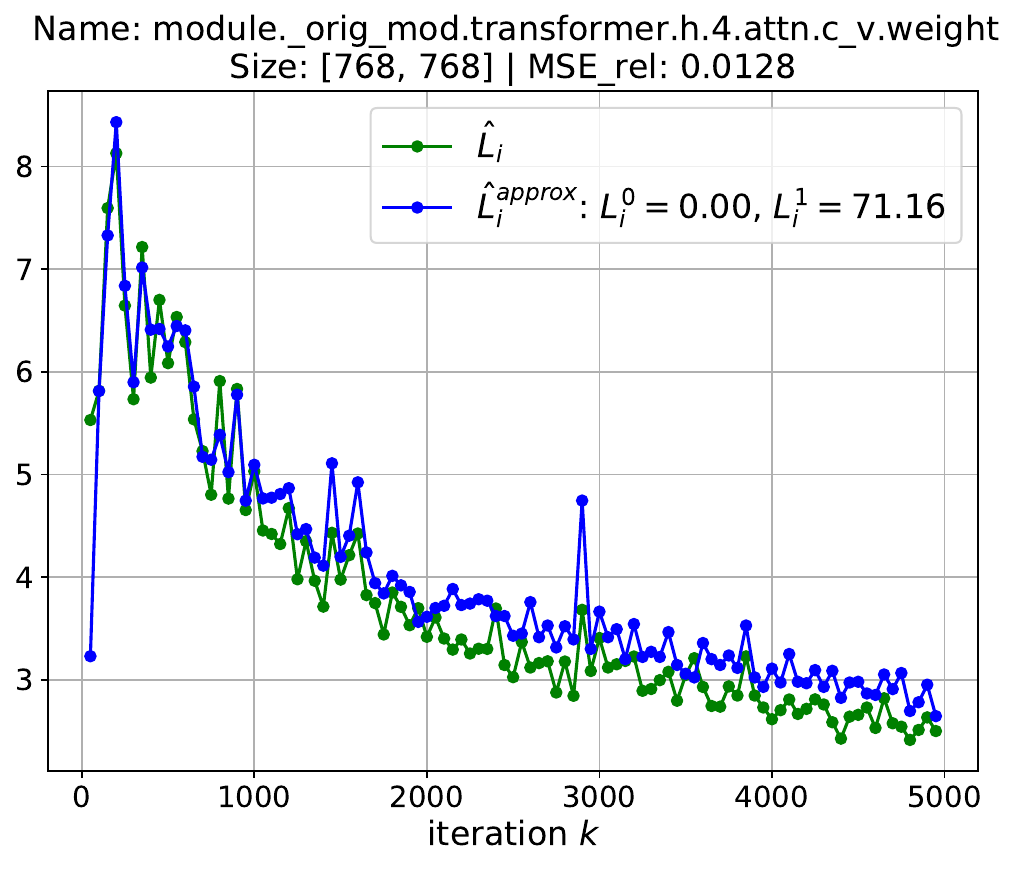}
    \end{minipage}

    \begin{minipage}{0.32\textwidth}
        \includegraphics[width=\textwidth]{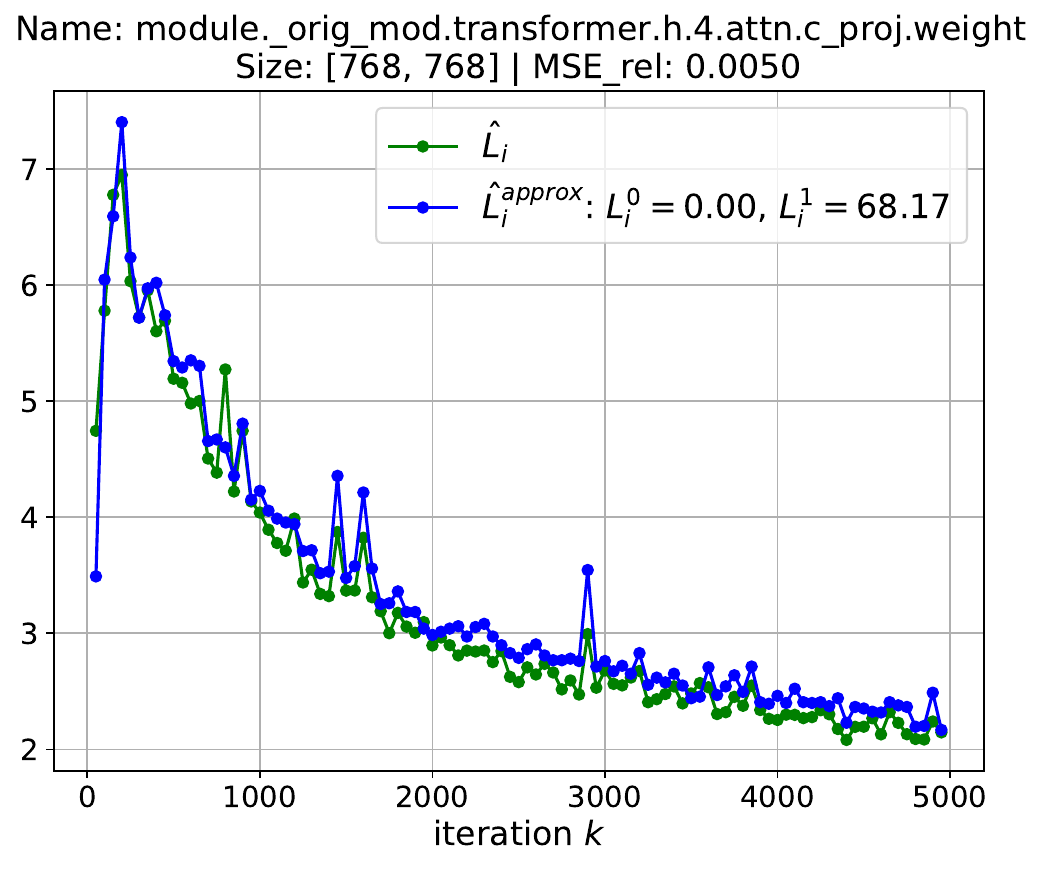}
    \end{minipage}
    \hfill
    \begin{minipage}{0.32\textwidth}
        \includegraphics[width=\textwidth]{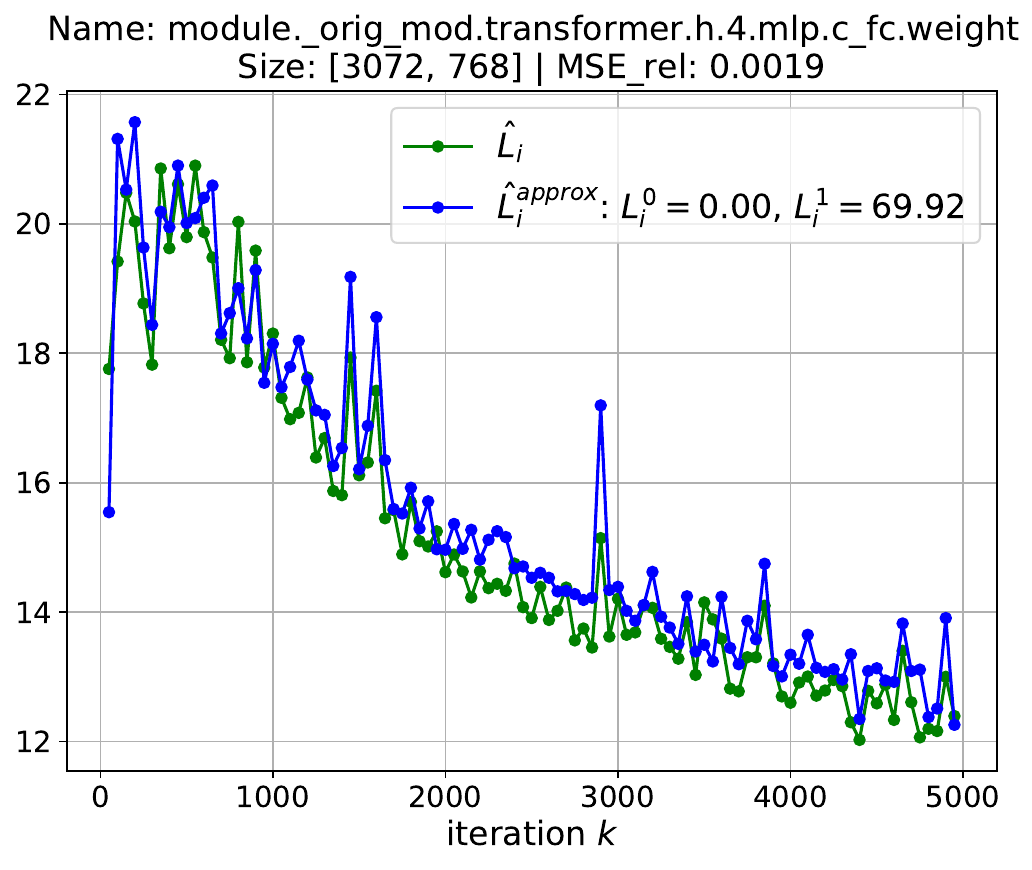}
    \end{minipage}
    \hfill
    \begin{minipage}{0.32\textwidth}
        \includegraphics[width=\textwidth]{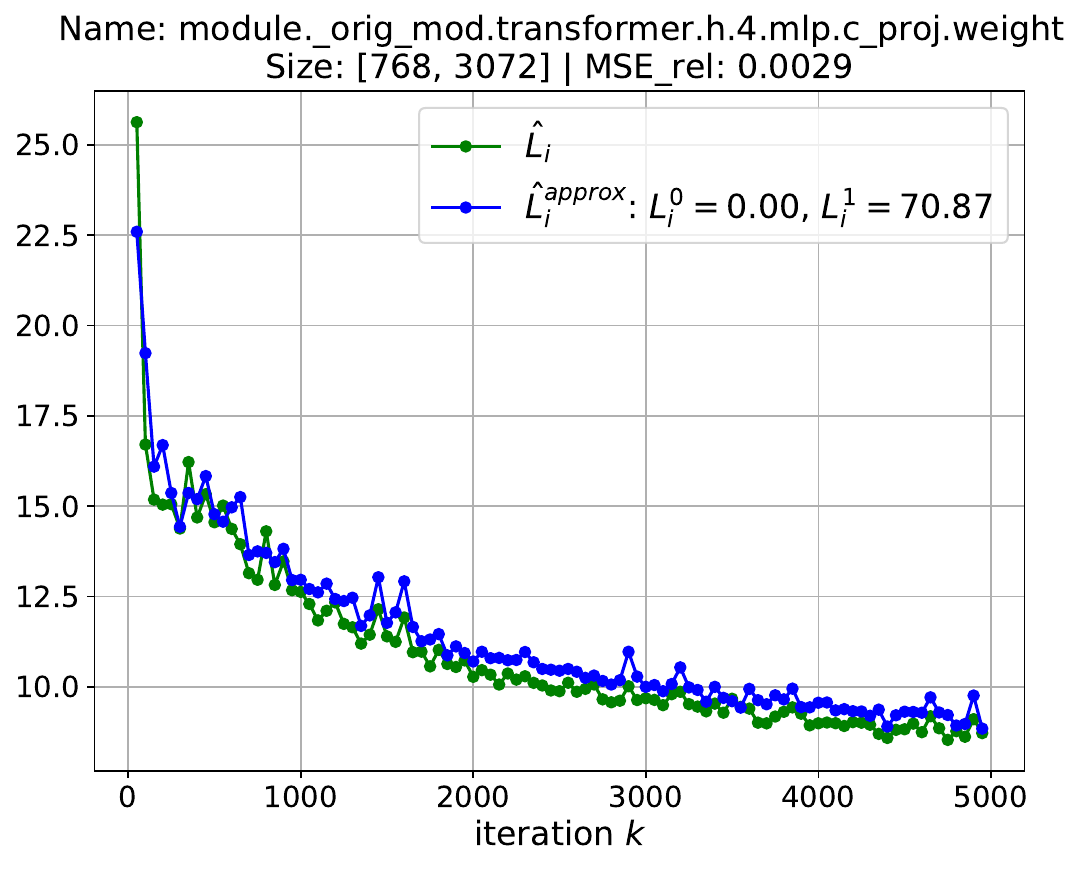}
    \end{minipage}
    \caption{\small {\bf Validation of layer-wise $(L^0, L^1)$-smoothness} for the group of parameters from the 4th transformer block of \texttt{NanoGPT-124M} along \algname{unScion} training trajectories. The group norms are $ \|\cdot\|_{(i)} = \sqrt{\nicefrac{n_i}{m_i}} \|\cdot\|_{2 \to 2} $, with fitted values $L_i^0 \approx 0$, $L_i^1 \approx 70$.}
    \label{gluon:fig:2}
\end{figure}

\begin{figure}[h]
    \centering
    \begin{minipage}{0.32\textwidth}
        \includegraphics[width=\textwidth]{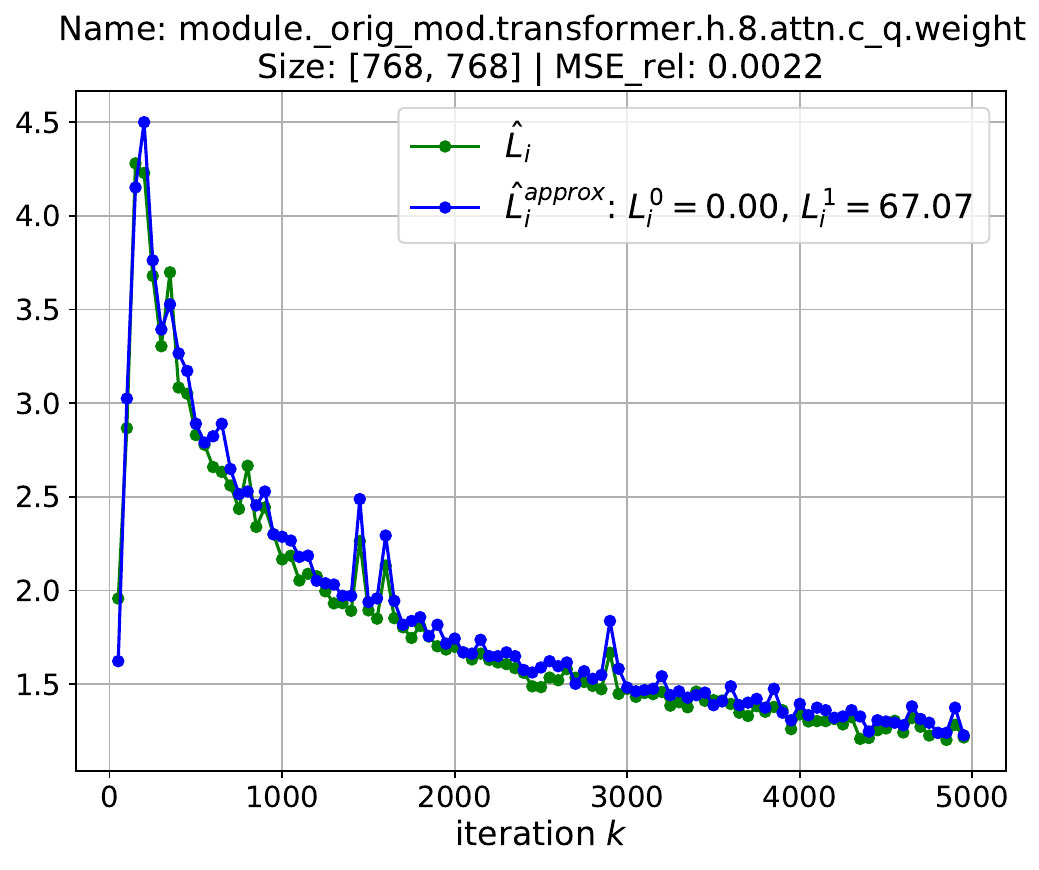}
    \end{minipage}
    \hfill
    \begin{minipage}{0.32\textwidth}
        \includegraphics[width=\textwidth]{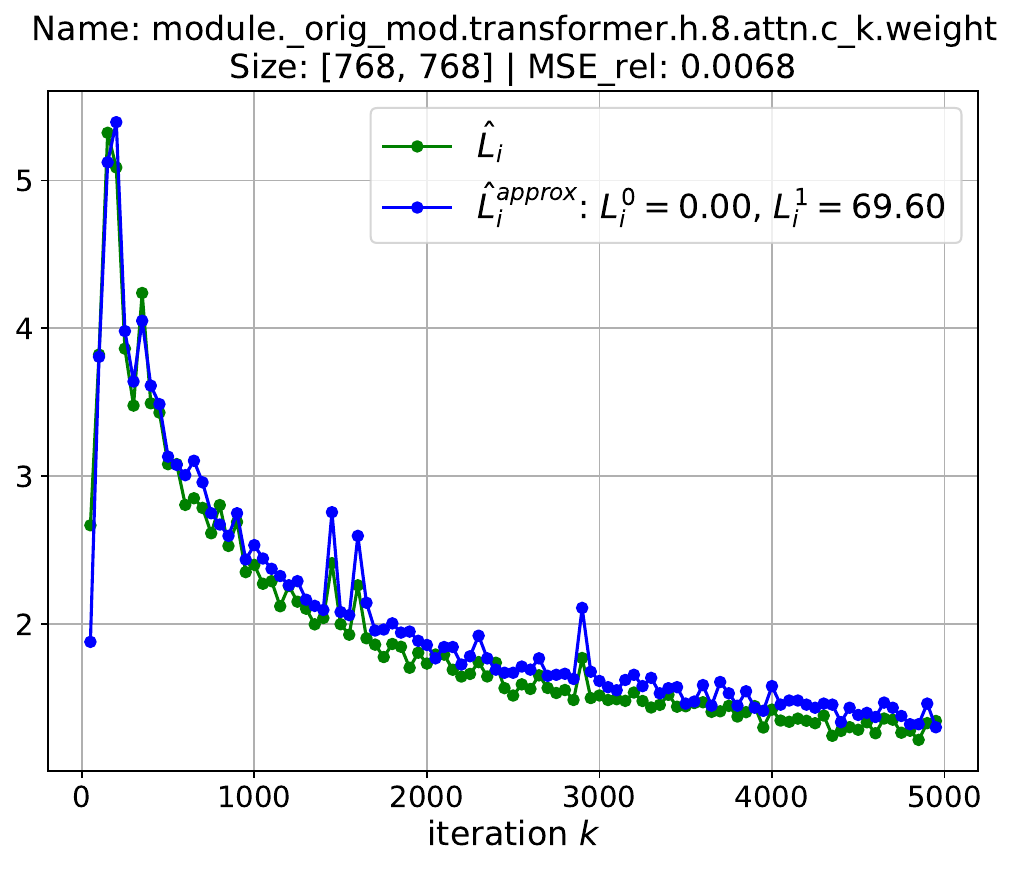}
    \end{minipage}
    \hfill
    \begin{minipage}{0.32\textwidth}
        \includegraphics[width=\textwidth]{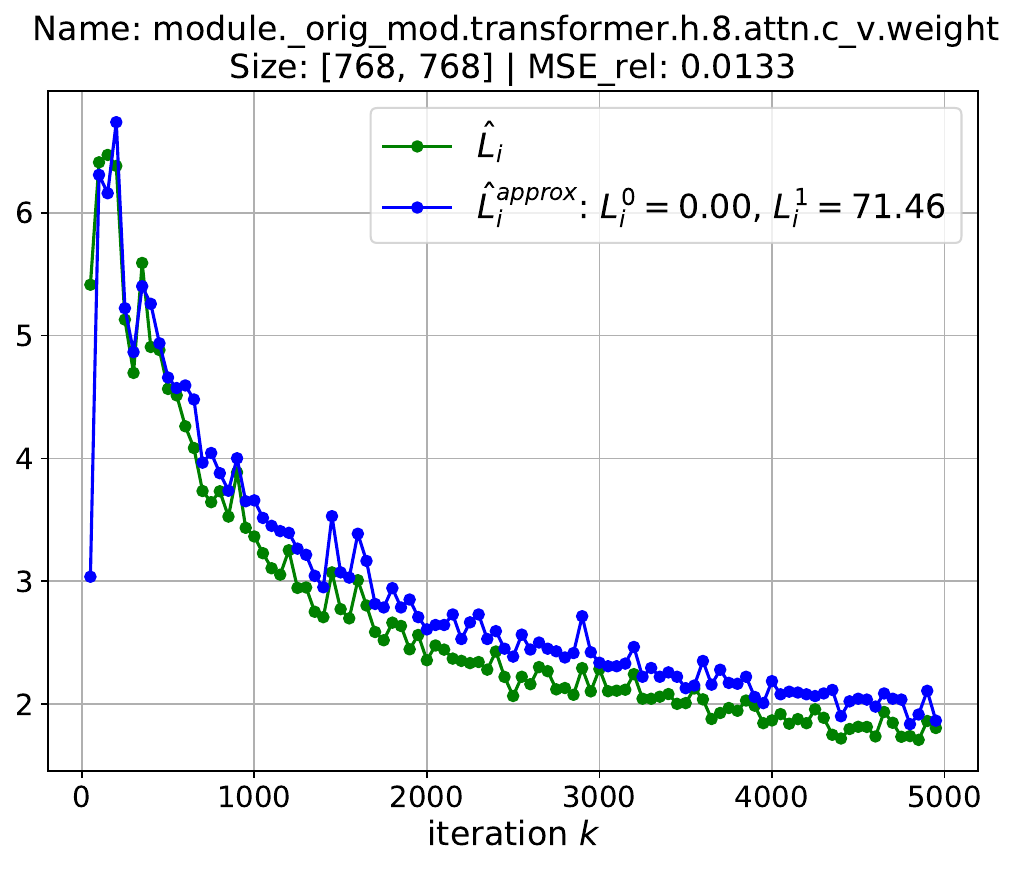}
    \end{minipage}

    \begin{minipage}{0.32\textwidth}
        \includegraphics[width=\textwidth]{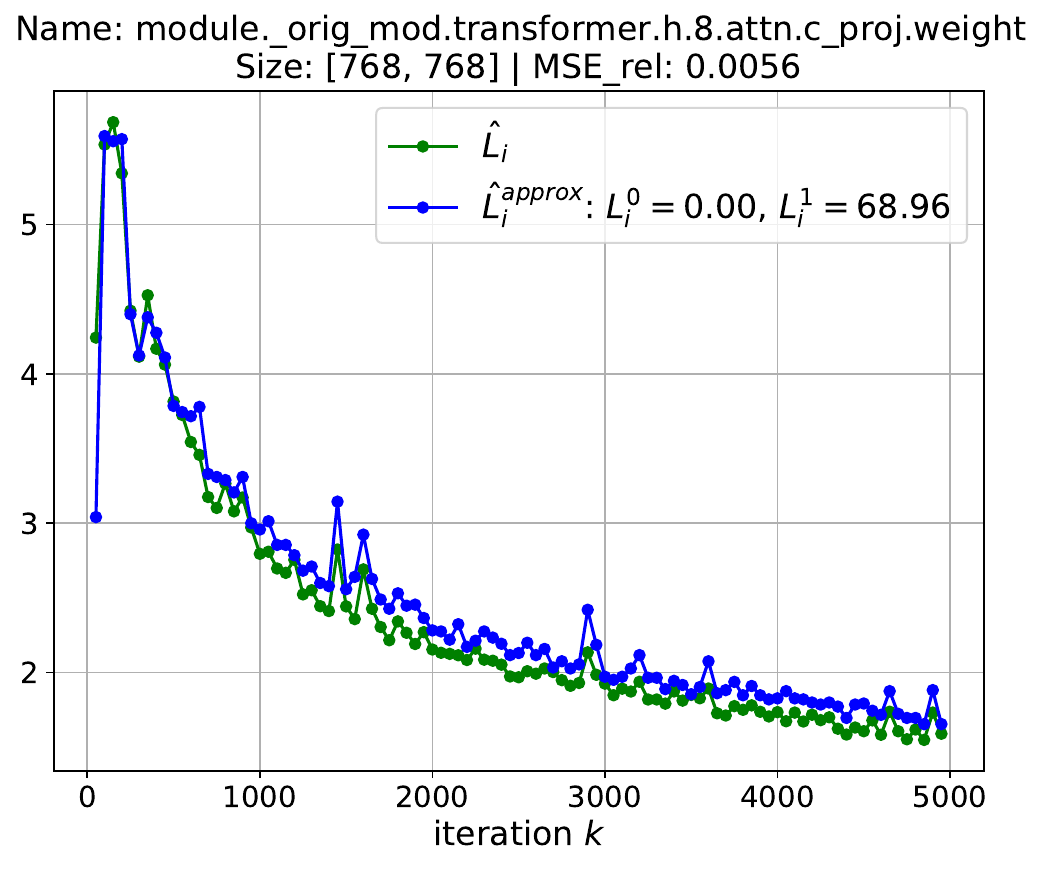}
    \end{minipage}
    \hfill
    \begin{minipage}{0.32\textwidth}
        \includegraphics[width=\textwidth]{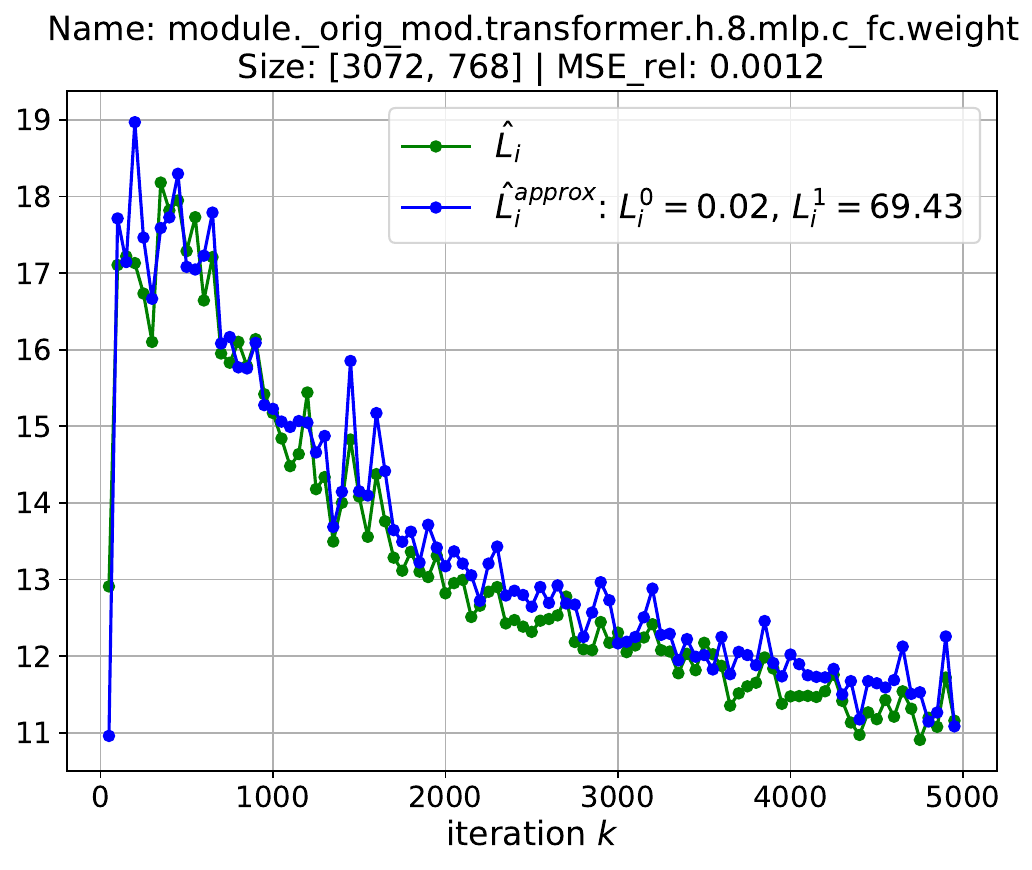}
    \end{minipage}
    \hfill
    \begin{minipage}{0.32\textwidth}
        \includegraphics[width=\textwidth]{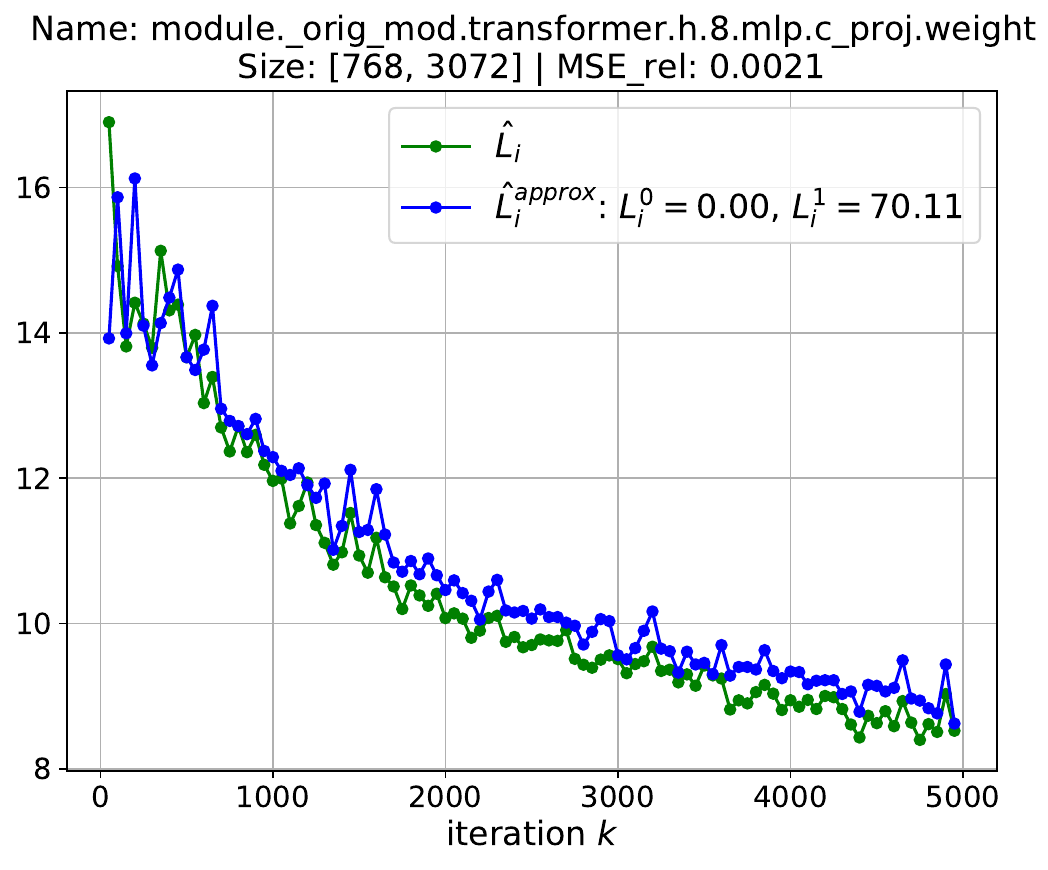}
    \end{minipage}

    \caption{{\bf Validation of layer-wise $(L^0, L^1)$-smoothness} for the group of parameters from the 8th transformer block of \texttt{NanoGPT-124M} along \algname{unScion} training trajectories. The group norms are $ \|\cdot\|_{(i)} = \sqrt{\nicefrac{n_i}{m_i}} \|\cdot\|_{2 \to 2} $, with fitted values $L_i^0 \approx 0$, $L_i^1 \approx 70$.}
    \label{gluon:fig:3_appndx}
\end{figure}

\subsubsection{Generalized smoothness under Euclidean vs. specialized norms} \label{gluon:appendix:gensmooth_euclid}

In this experiment, we compare how well the layer-wise $(L^0, L^1)$-smoothness assumption is satisfied under the standard Euclidean norms $ \|\cdot\|_2 $ for each parameter block, as opposed to the specialized norms described in~(\ref{gluon:eq:scion_llm}). We adopt the same training setup as in Section~\ref{gluon:sec:experiments}, plotting the estimated trajectory smoothness $ \hat{L}_i $ and its approximation $ \hat{L}_i^{\text{approx}} $ along the training trajectories across several parameter groups. Unlike previous sections, here we do not penalize instances where $ \hat{L}_i > \hat{L}_i^{\text{approx}} $ in order to find the best approximation (i.e., $\lambda = 0$ in \eqref{gluon:eq:l0l1_loss}). Additionally, when using the standard Euclidean norm $ \|\cdot\|_2 $ for approximation, we exclude the first point, as it could distort the result.

We evaluate the quality of each approximation using the relative mean squared error ($\mathrm{MSE}_i^{\mathrm{rel}}$, denoted MSE\_rel in the figures), defined as
\begin{align*}
    \text{MSE}_i^{\text{rel}}:= \frac{1}{K} \sum_{i=1}^{K} \left( \frac{\hat{L}_i[k] - \hat{L}_i^{\text{approx}}[k]}{\hat{L}_i[k]} \right)^2,
\end{align*}
where a lower value indicates a better fit.

As shown in Figures~\ref{gluon:fig:specific_norms_mse} and~\ref{gluon:fig:euclideum_norms_mse}, both visually and in terms of $\mathrm{MSE}_i^{\mathrm{rel}}$, using specialized norms for each group of parameters provides a better approximation than the standard Euclidean norm~$ \|\cdot\|_2 $. Notably, the relative mean squared error $\mathrm{MSE}_i^{\mathrm{rel}}$ is consistently an order of magnitude lower under specialized norms.

\begin{figure}[h]
    \centering
    \subfigure[$\mathrm{MSE}_i^{\mathrm{rel}} = 0.0023$]{%
        \includegraphics[width=0.32\textwidth]{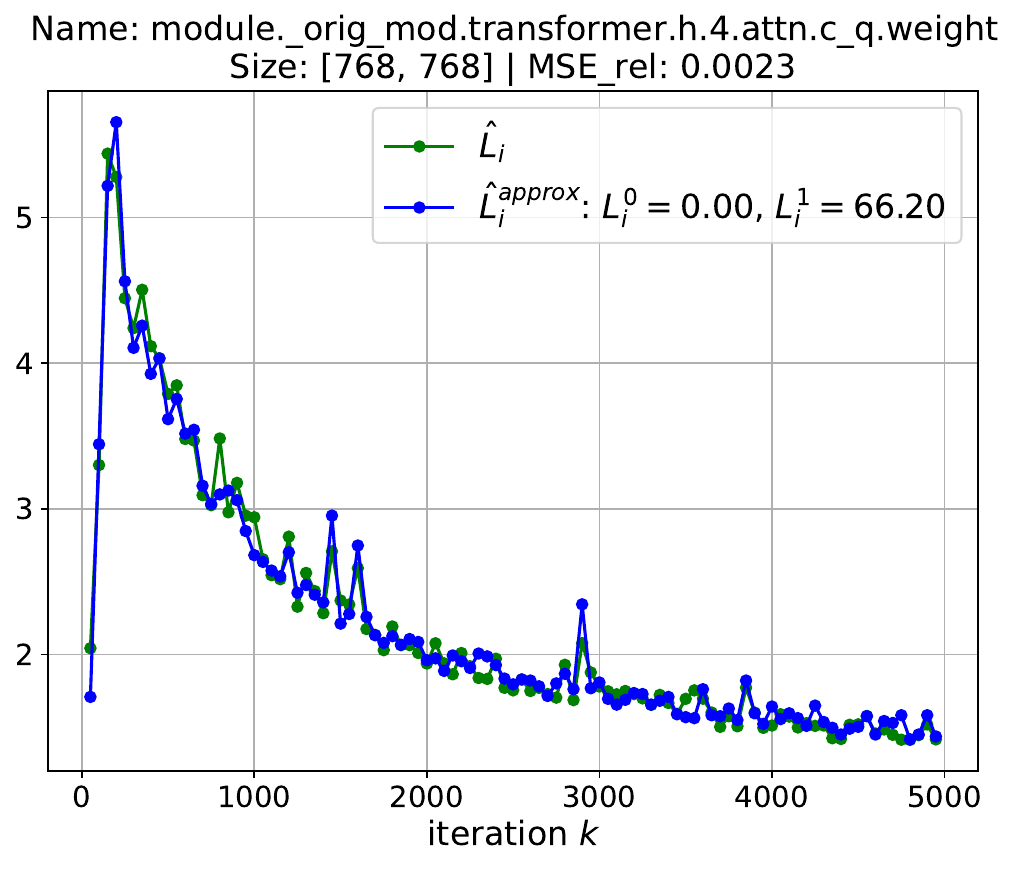}}%
    \hfill
    \subfigure[$\mathrm{MSE}_i^{\mathrm{rel}} = 0.0048$]{%
        \includegraphics[width=0.32\textwidth]{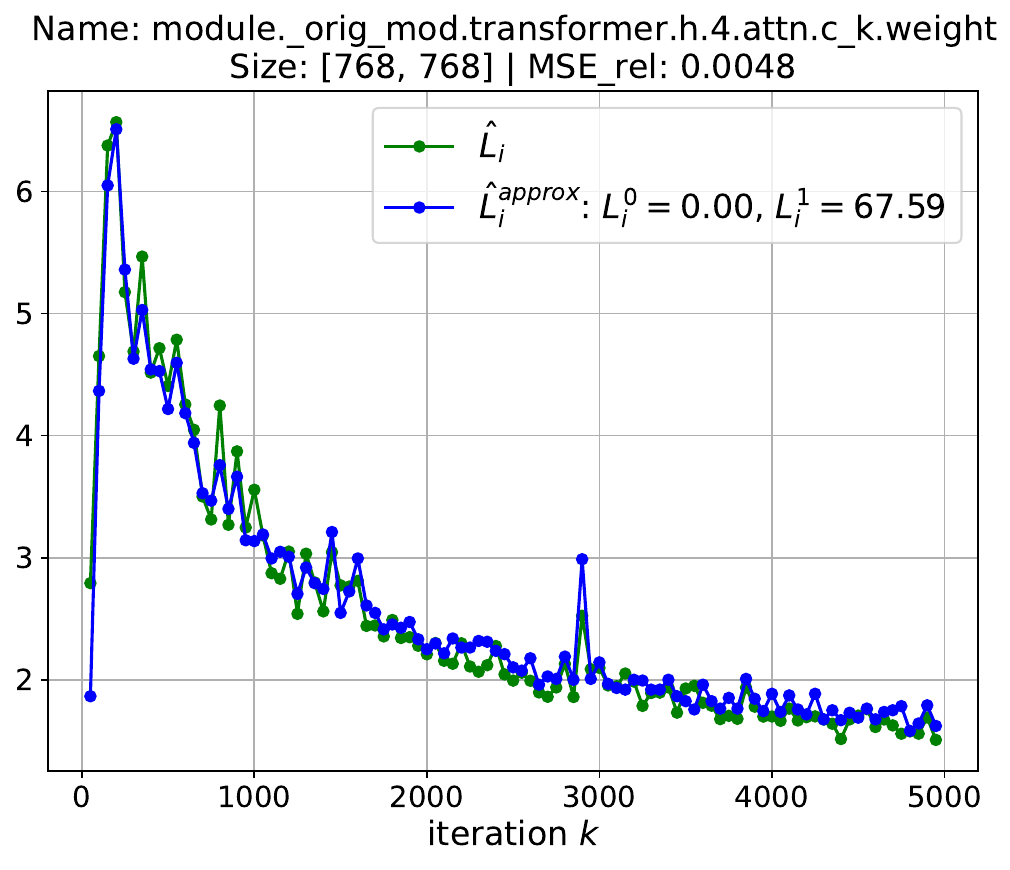}}%
    \hfill
    \subfigure[$\mathrm{MSE}_i^{\mathrm{rel}} = 0.0065$]{%
        \includegraphics[width=0.32\textwidth]{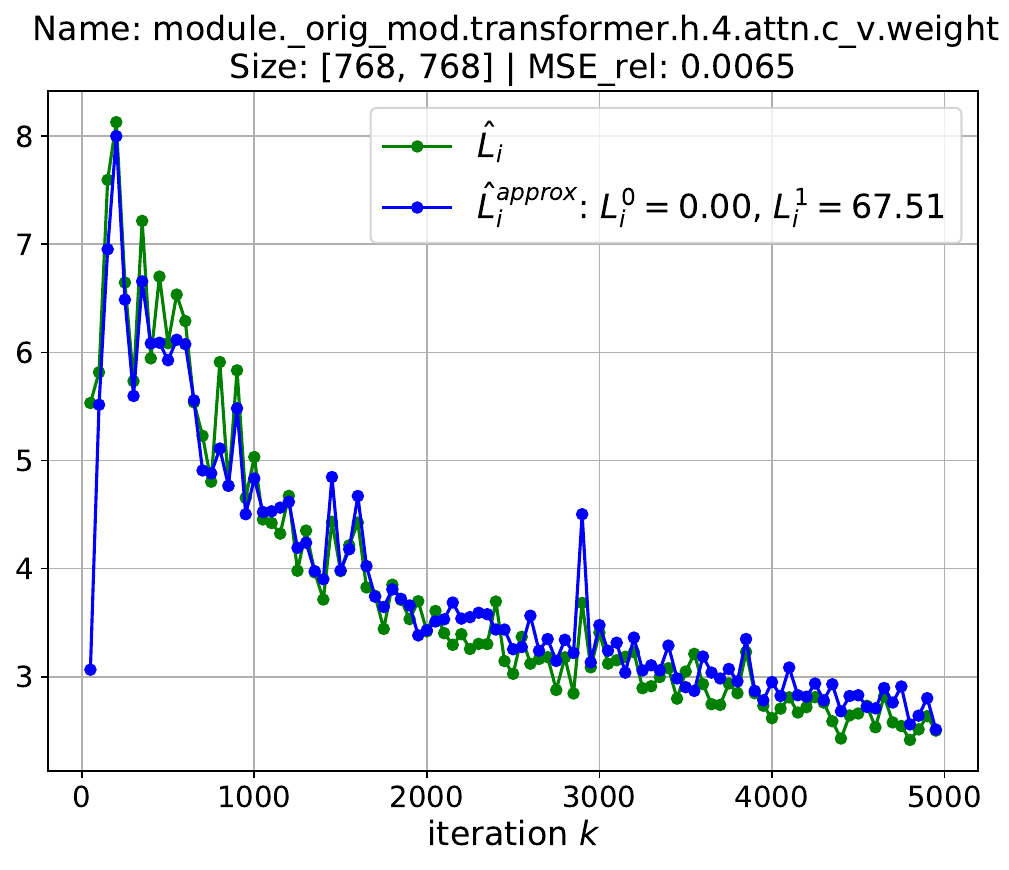}}%

    \subfigure[$\mathrm{MSE}_i^{\mathrm{rel}} = 0.0025$]{%
        \includegraphics[width=0.32\textwidth]{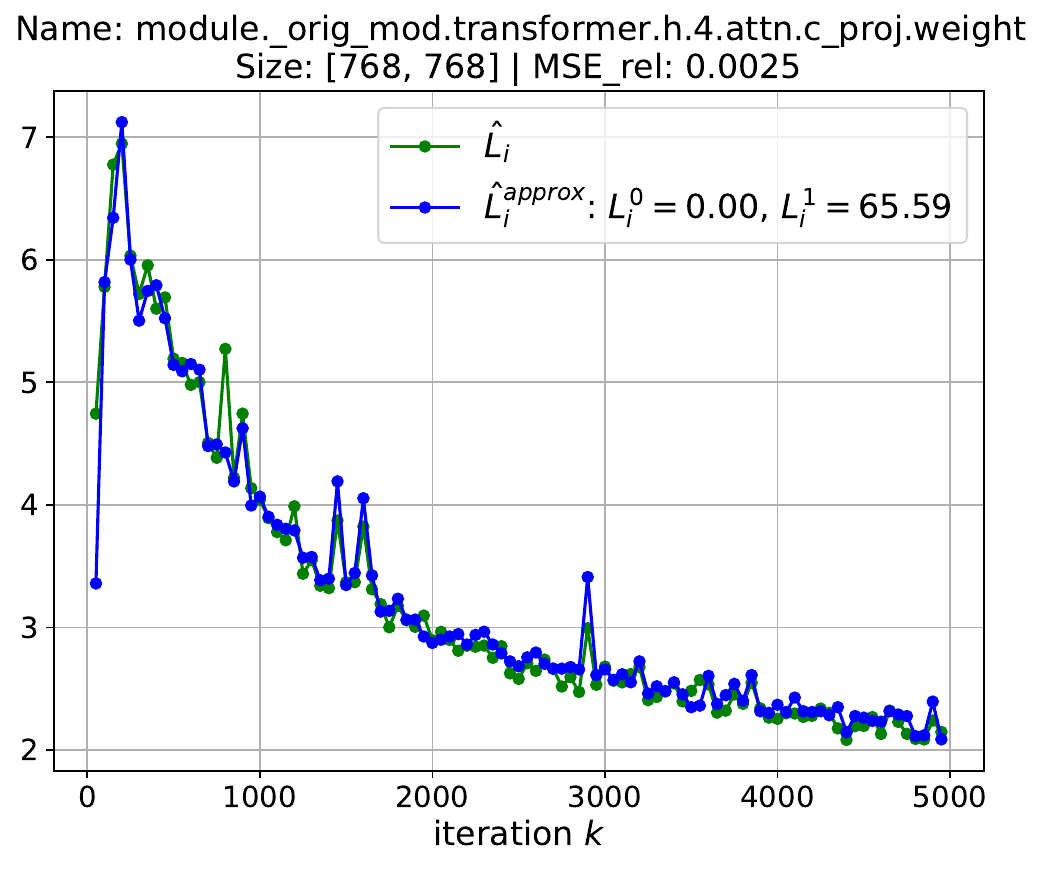}}%
    \hfill
    \subfigure[$\mathrm{MSE}_i^{\mathrm{rel}} = 0.001$]{%
        \includegraphics[width=0.32\textwidth]{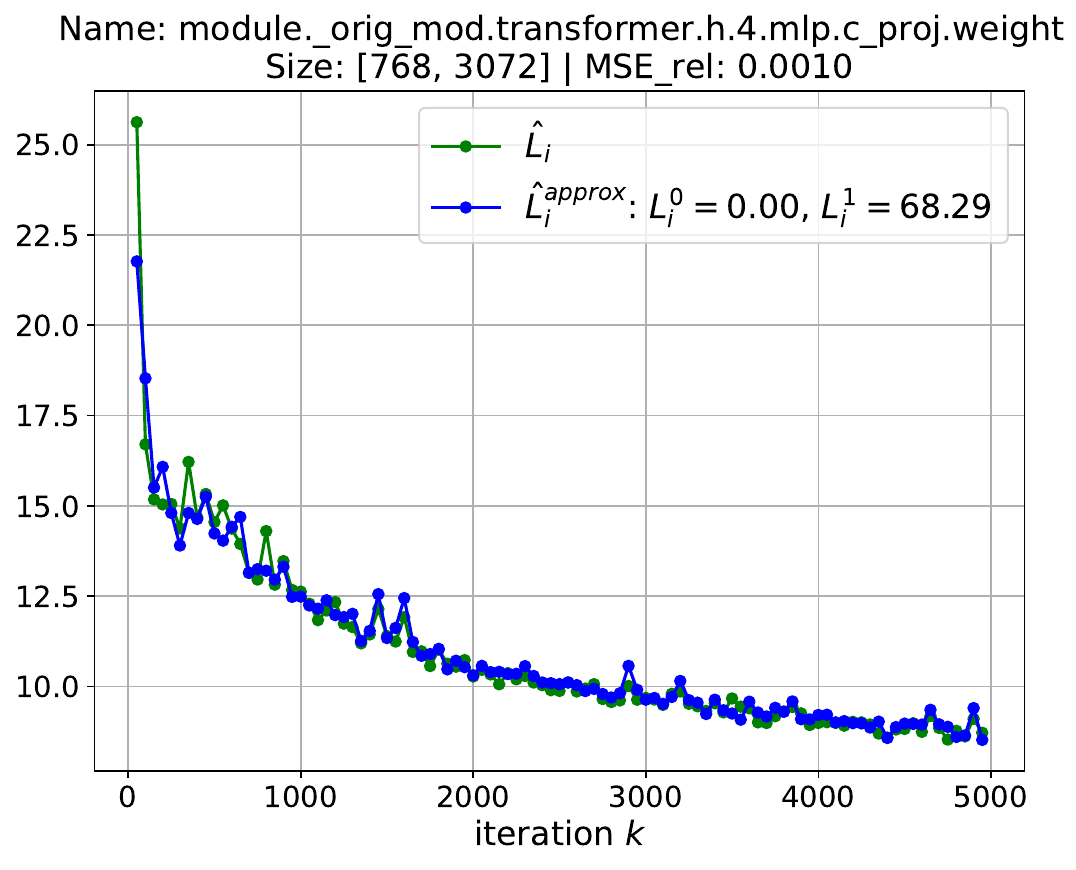}}%
    \hfill
    \subfigure[$\mathrm{MSE}_i^{\mathrm{rel}} = 0.0003$]{%
        \includegraphics[width=0.32\textwidth]{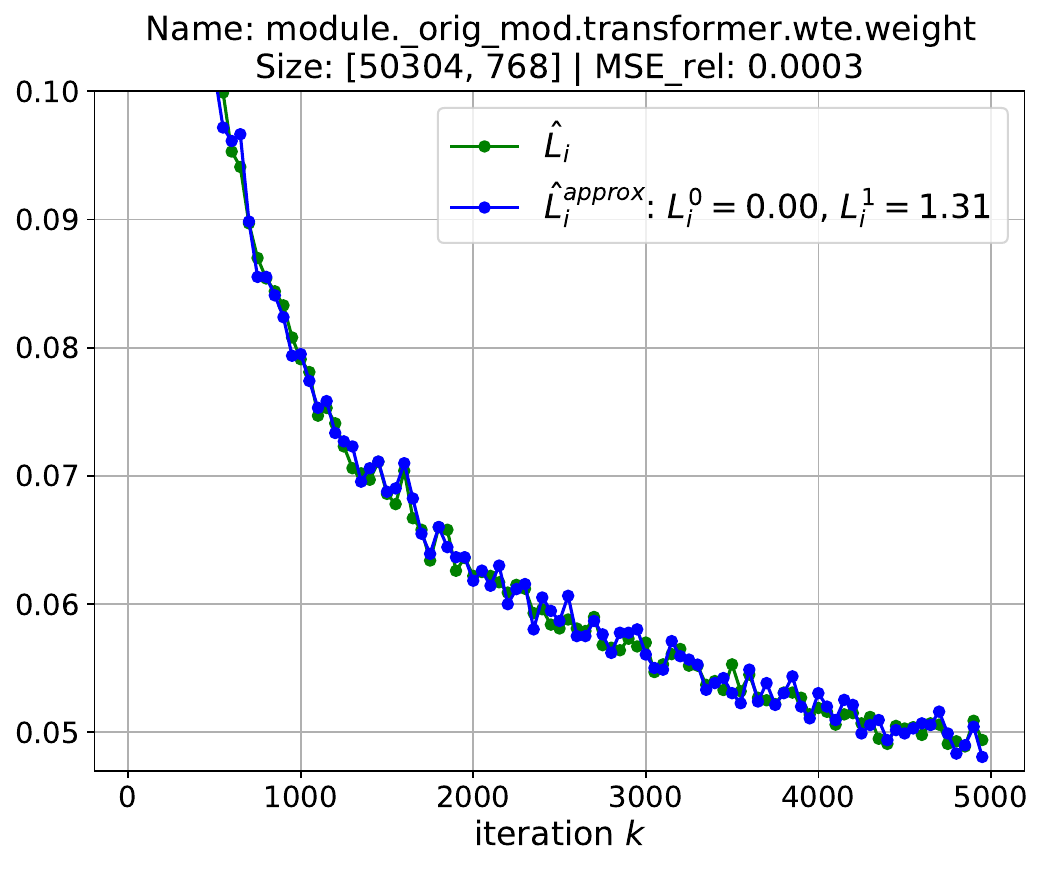}}%

    \caption{\small {\bf Validation of layer-wise $(L^0, L^1)$-smoothness} for different groups of parameters in \texttt{NanoGPT-124M} along training trajectories of \algname{unScion} using the specialized norm choices defined in~(\ref{gluon:eq:scion_llm}).}
    \label{gluon:fig:specific_norms_mse}
\end{figure}

\begin{figure}[H]
    \centering
    \begin{minipage}{0.32\textwidth}
        \includegraphics[width=\textwidth]{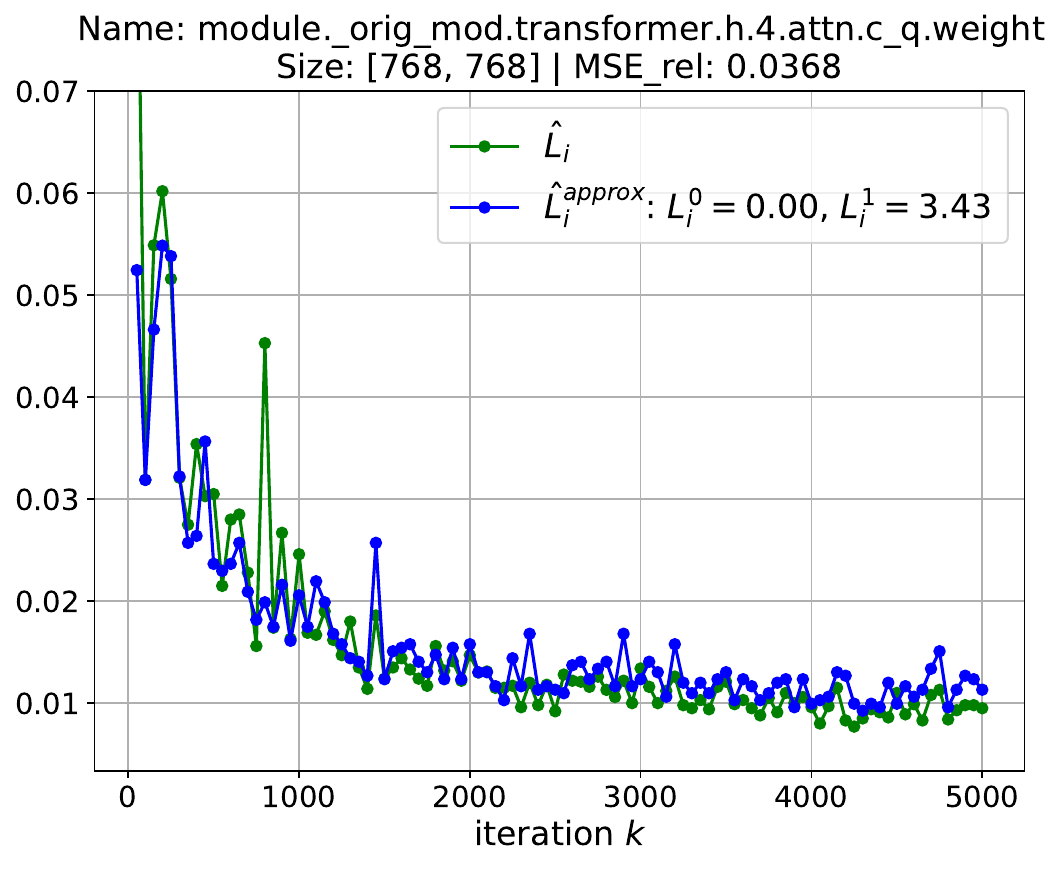}
        \centering{\small $\mathrm{MSE}_i^{\mathrm{rel}} = 0.0368$}
    \end{minipage}
    \hfill
    \begin{minipage}{0.32\textwidth}
        \includegraphics[width=\textwidth]{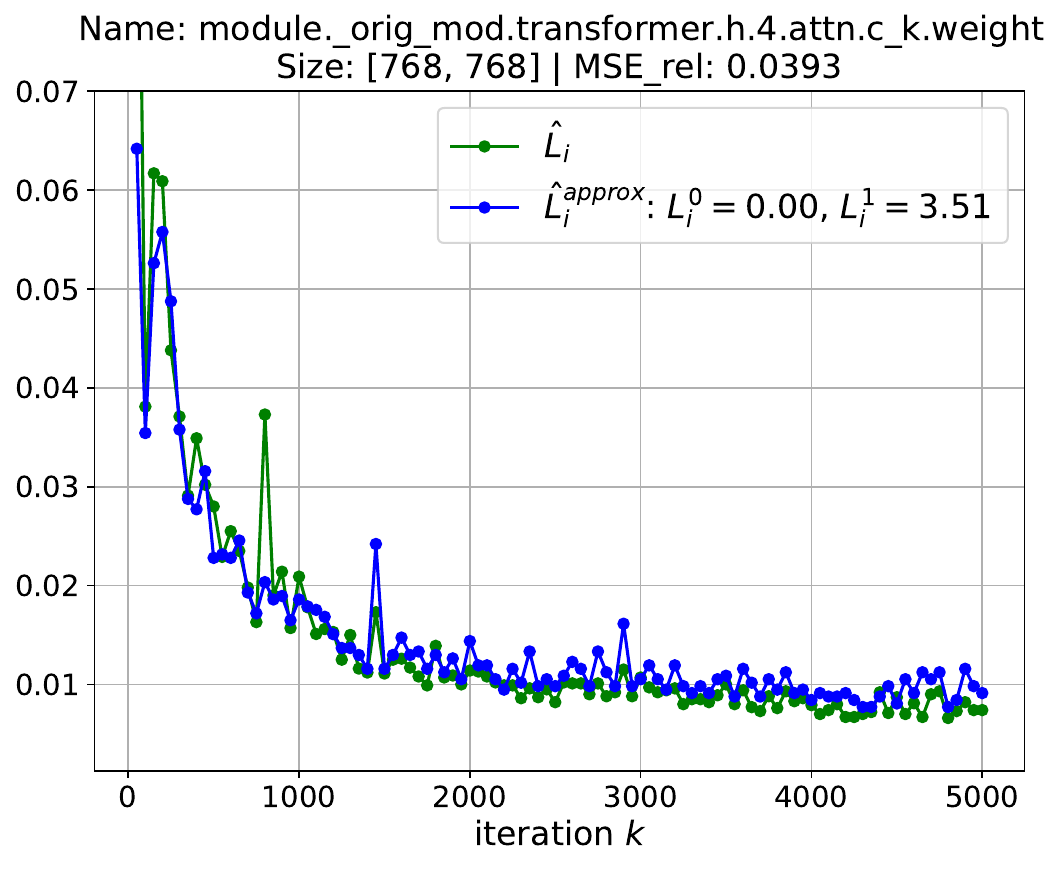}
        \centering{\small $\mathrm{MSE}_i^{\mathrm{rel}} = 0.0393$}
    \end{minipage}
    \hfill
    \begin{minipage}{0.32\textwidth}
        \includegraphics[width=\textwidth]{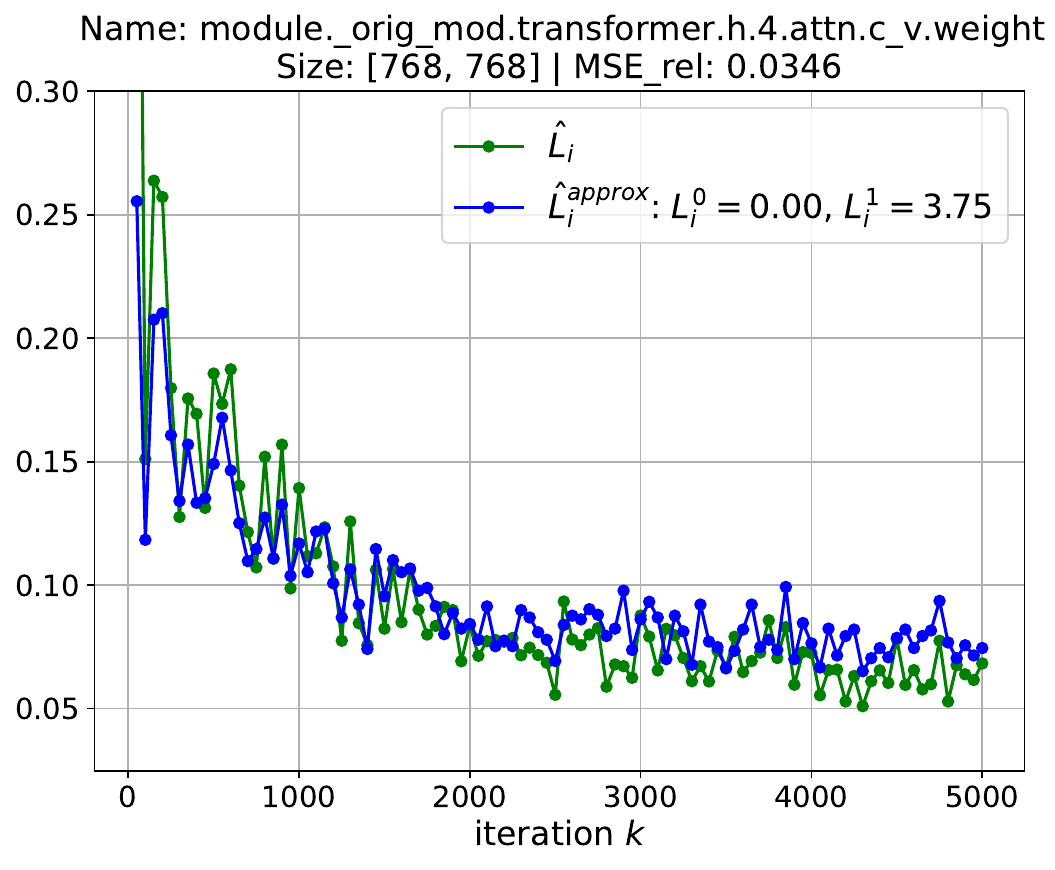}
        \centering{\small $\mathrm{MSE}_i^{\mathrm{rel}} = 0.0346$}
    \end{minipage}

    \begin{minipage}{0.32\textwidth}
        \includegraphics[width=\textwidth]{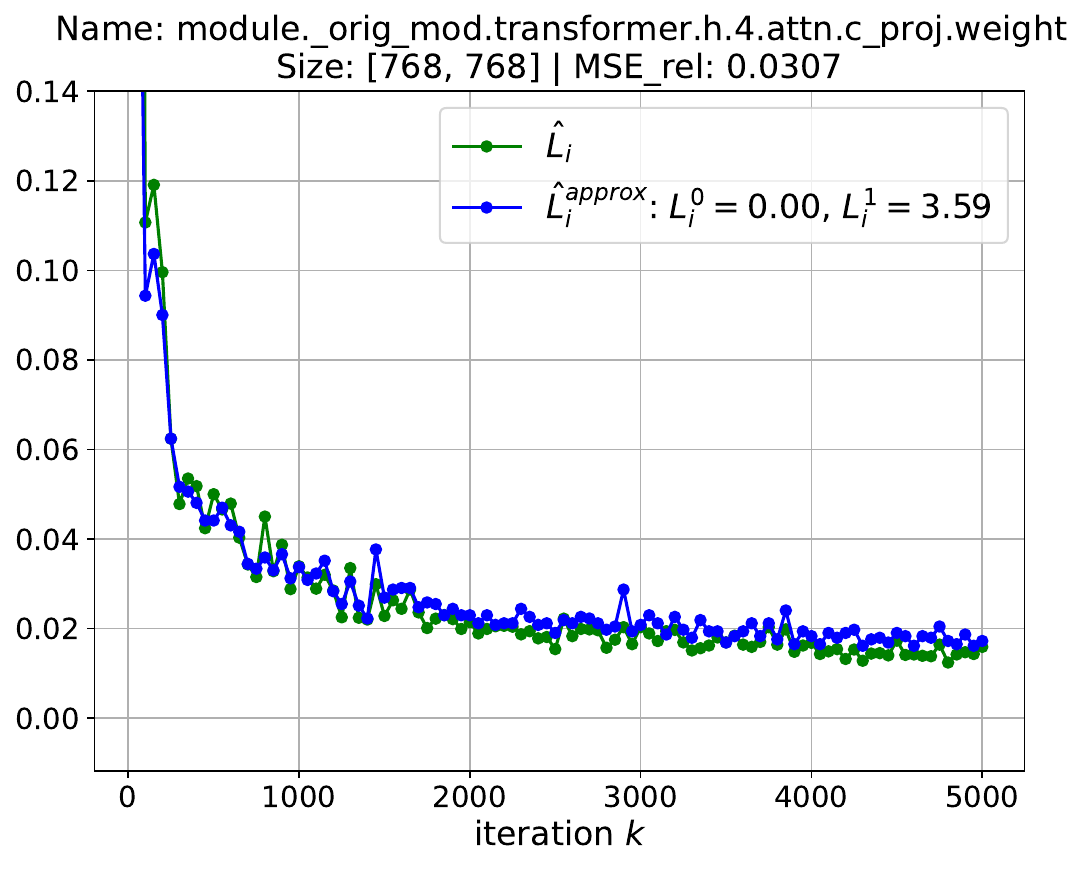}
        \centering{\small $\mathrm{MSE}_i^{\mathrm{rel}} = 0.0307$}
    \end{minipage}
    \hfill
    \begin{minipage}{0.32\textwidth}
        \includegraphics[width=\textwidth]{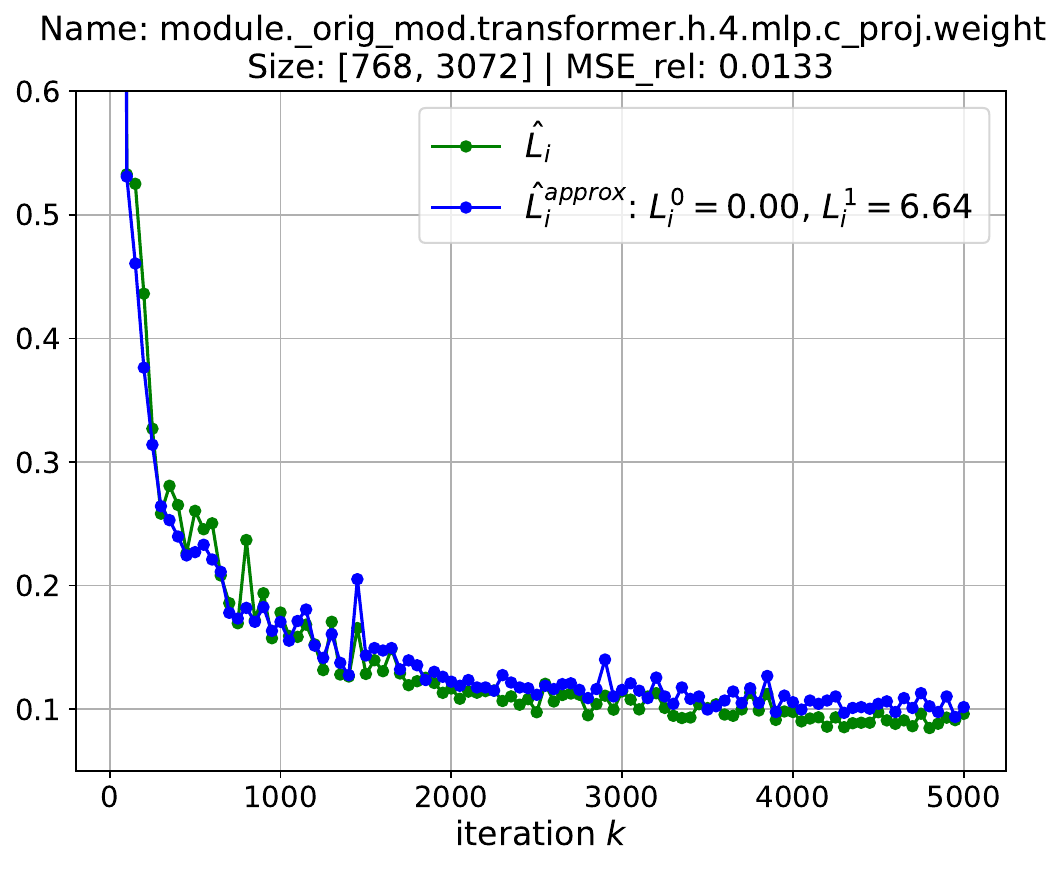}
        \centering{\small $\mathrm{MSE}_i^{\mathrm{rel}} = 0.0133$}
    \end{minipage}
    \hfill
    \begin{minipage}{0.32\textwidth}
        \includegraphics[width=\textwidth]{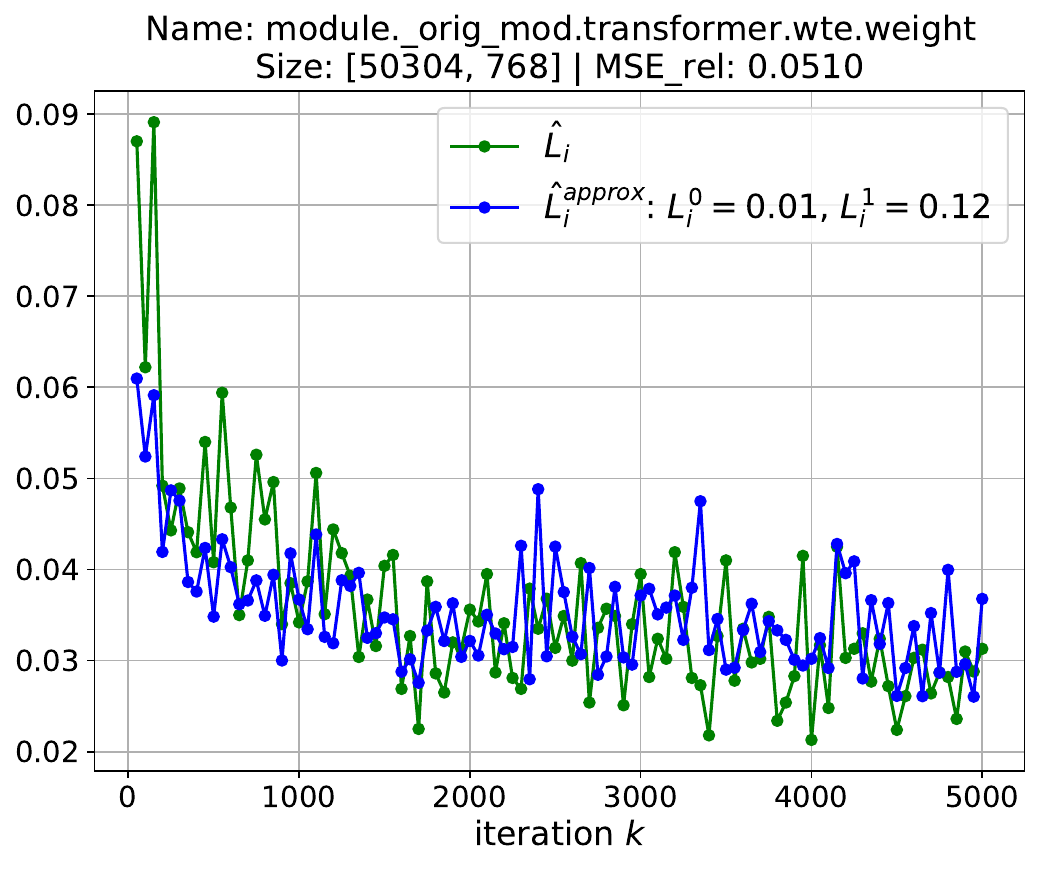}
        \centering{\small $\mathrm{MSE}_i^{\mathrm{rel}} = 0.051$}
    \end{minipage}

    \caption{\small {\bf Validation of layer-wise $(L^0, L^1)$-smoothness} for different groups of parameters in \texttt{NanoGPT-124M} along training trajectories of \algname{unScion} using the standard Euclidean norm $ \|\cdot\|_2 $.}
    \label{gluon:fig:euclideum_norms_mse}
\end{figure}

\subsubsection{Learning rate transfer from AdamW}
\label{gluon:sec:Adam_exp}

In this set of experiments, we aim to verify layer-wise $(L^0, L^1)$-smoothness following the approach used in Section~\ref{gluon:sec:experiments}, but employing the \algname{AdamW} optimizer. We use hyperparameters specified in \citet[Table 7]{pethick2025training}.

In Figure~\ref{gluon:fig:Adam_app}, we present the results for the estimated trajectory smoothness $\hat{L}_i$ and its approximation $\hat{L}_i^{\text{approx}}$ across several parameter groups along the training trajectories. Notably, for the group of parameters from the embedding layer $X_p$ (the last plot in Figure~\ref{gluon:fig:Adam_app}), the fitted value of $L^1_p$ is approximately 20–30 times smaller than in other groups. Since in all plots we observe that $L^0_i \ll L^1_i \|\nabla_i f_{\xi^k}(X^k)\|_{(i) \star}$, \Cref{gluon:theorem:1} implies that $t^k_i \approx \nicefrac{1}{L^k_i}$. Thus, $ t^k_p $ should be 20–30 times larger than $ t^k_i $ for $ i = 1, \dots, p-1 $, which is consistent with the tuned parameters from \citet[Table 7]{pethick2025training}.

\Cref{gluon:fig:all_layers_scales-124M-Adam} shows the estimated trajectory smoothness across all layers and blocks. We again observe substantial variation across layers within each block, reinforcing the need for the per-layer treatment introduced in our framework.

This insight provides an efficient and principled method for initializing learning rates in \algname{Scion}. Smoothness statistics collected during standard \algname{AdamW} training (which is commonly used for training LLMs) can serve as a strong prior, allowing practitioners to directly incorporate structure-aware choices, such as larger stepsizes for embedding layers, into their tuning process.
Importantly, computing these statistics is computationally inexpensive, introducing minimal additional cost.

\begin{figure}[h]
    \centering
    \begin{minipage}{0.32\textwidth}
        \includegraphics[width=\textwidth]{KAUST_Thesis/imported/06_gluon/plots/nanogpt/AdamW/plot1.pdf}
    \end{minipage}
    \hfill
    \begin{minipage}{0.32\textwidth}
        \includegraphics[width=\textwidth]{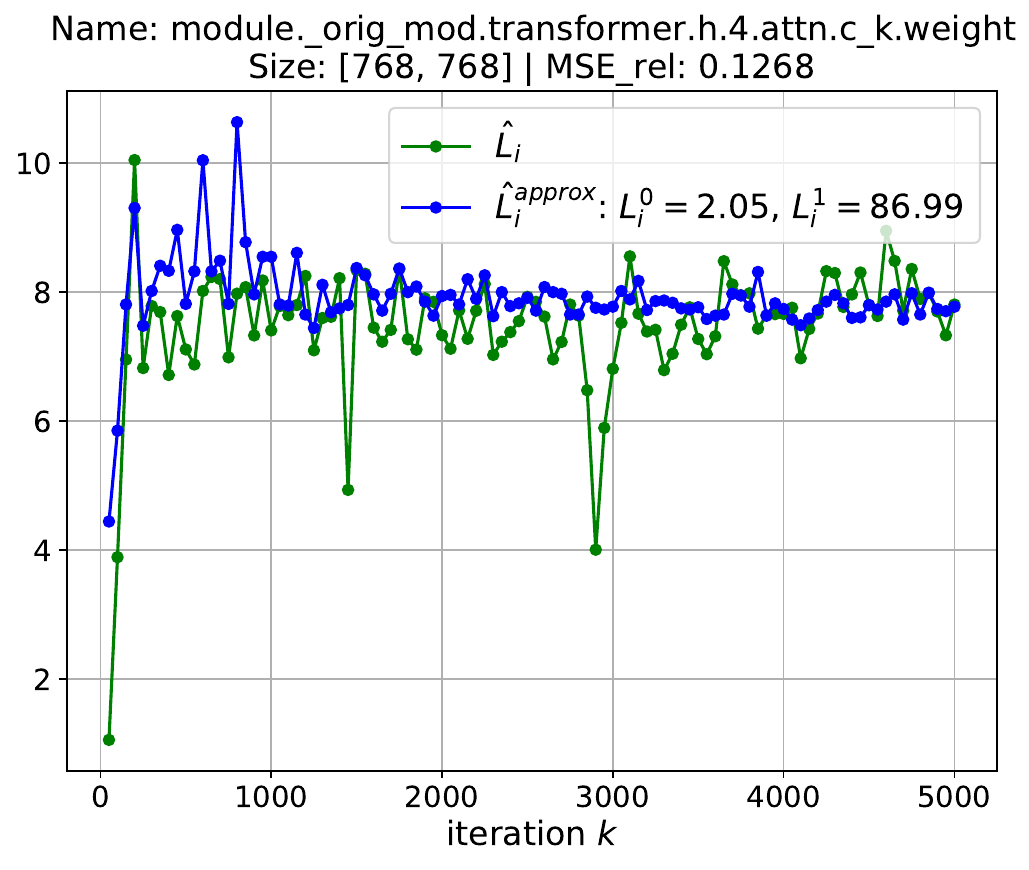}
    \end{minipage}
    \hfill
    \begin{minipage}{0.32\textwidth}
        \includegraphics[width=\textwidth]{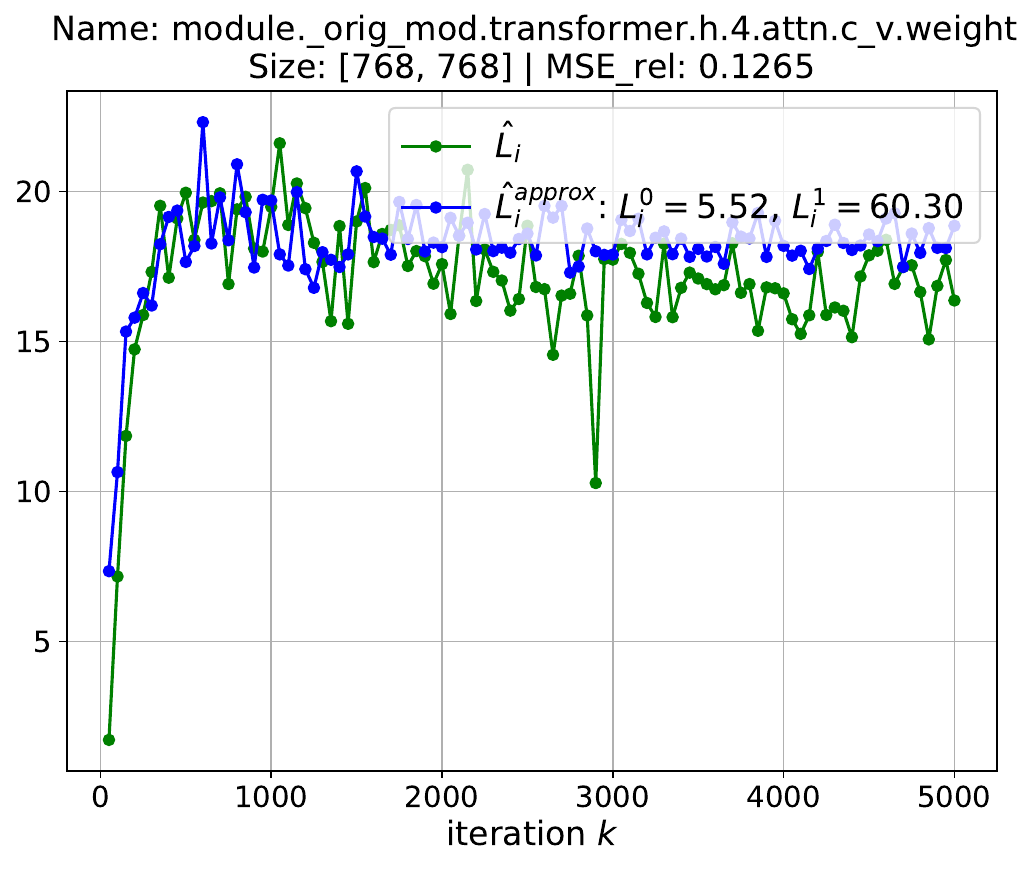}
    \end{minipage}

    \vspace{0.5cm}

    \begin{minipage}{0.32\textwidth}
        \includegraphics[width=\textwidth]{KAUST_Thesis/imported/06_gluon/plots/nanogpt/AdamW/plot5.pdf}
    \end{minipage}
    \hfill
    \begin{minipage}{0.32\textwidth}
        \includegraphics[width=\textwidth]{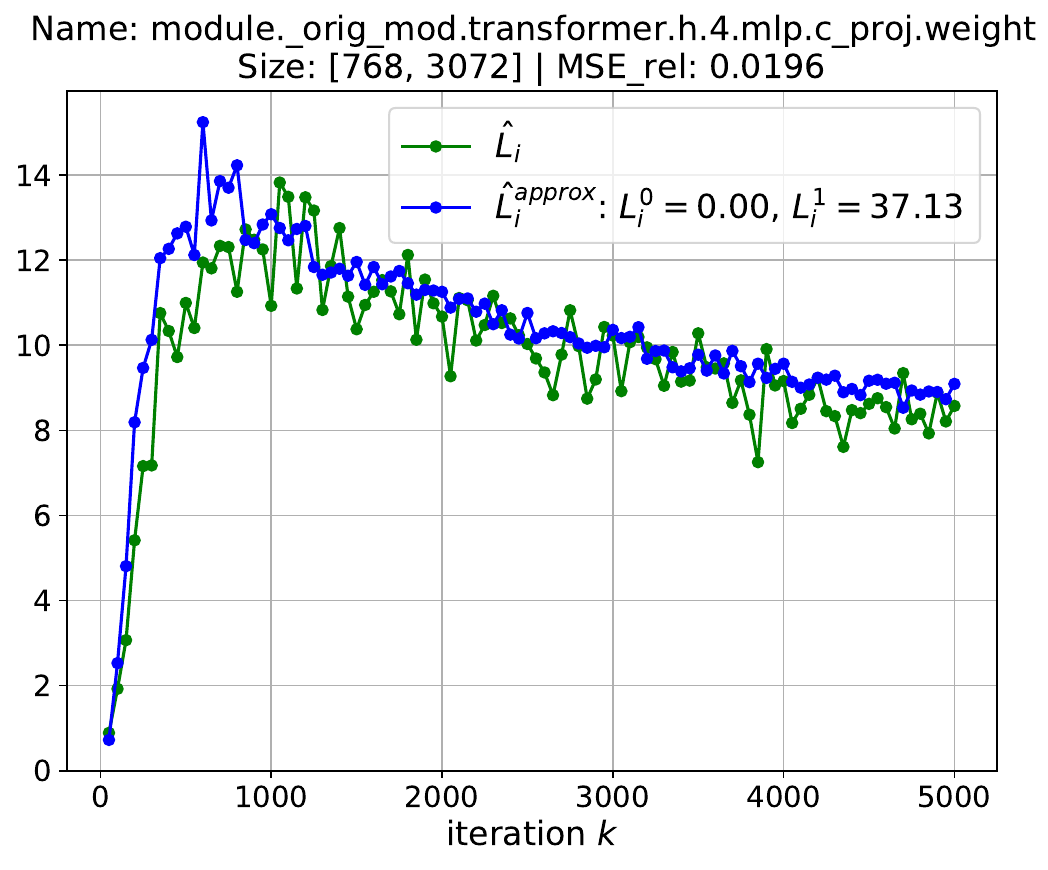}
    \end{minipage}
    \hfill
    \begin{minipage}{0.32\textwidth}
        \includegraphics[width=\textwidth]{KAUST_Thesis/imported/06_gluon/plots/nanogpt/AdamW/plot7.pdf}
    \end{minipage}

    \caption{\small {\bf Validation of layer-wise $(L^0, L^1)$-smoothness} for different groups of parameters in \texttt{NanoGPT-124M} along \algname{AdamW} training trajectories.}
    \label{gluon:fig:Adam_app}
\end{figure}

\begin{figure}[h]
    \centering
    \includegraphics[width=\textwidth]{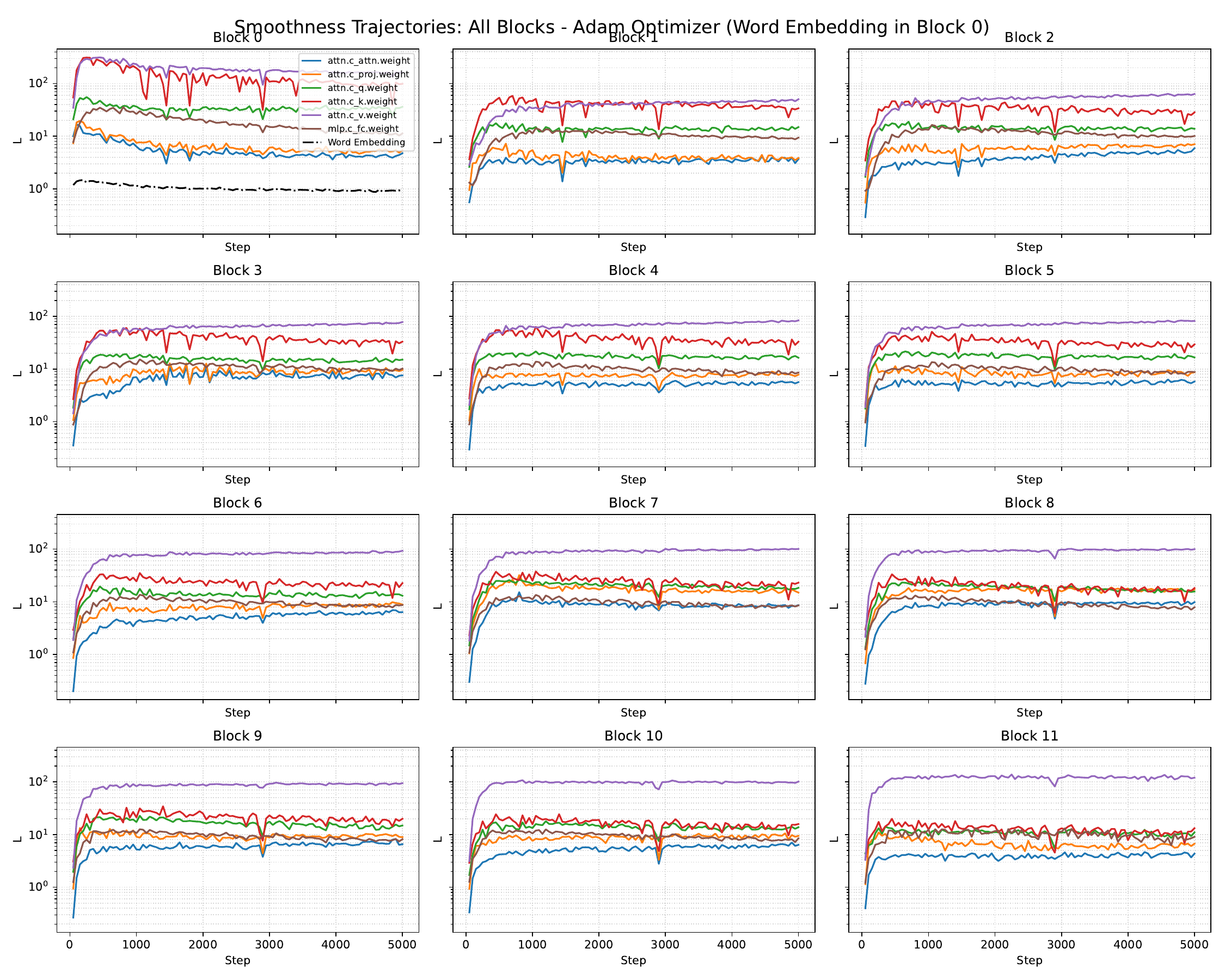}
    \caption{{\bf Estimated trajectory smoothness for \texttt{NanoGPT-124M} trained with \algname{AdamW}.}
    The same cross-layer heterogeneity pattern persists, indicating that layer-wise $(L^0_i, L^1_i)$-smoothness is not specific to \algname{unScion}.}
    \label{gluon:fig:all_layers_scales-124M-Adam}
\end{figure}

\subsubsection{Best vs.\ worst fits and aggregate fit quality} \label{gluon:appendix:fit_quality}

Figure~\ref{gluon:fig:best_worst_fits} illustrates the best and worst per-layer fits of \Cref{gluon:ass:generalized-smoothness}, comparing the measured trajectory smoothness $L_i[k]$ with its approximation $L_i^{\text{approx}}[k]$ along the \algname{unScion} training trajectory on \texttt{NanoGPT-124M}. For comparison, we also report fits obtained under the classical (though layer-wise) smoothness model, corresponding to \Cref{gluon:ass:generalized-smoothness} with $L_i^1 = 0$ for all layers.

We observe very tight fits for many layers (e.g., embeddings and several attention V matrices--see also \Cref{gluon:fig:fit_quality_by_matrix_type}), while a few layers show looser, yet still bounded, fits.
Quantitatively, the error of our model, measured by the relative mean squared error, is more than an order of magnitude smaller than that of classical smoothness (i.e., constant-$L_i$ fits) for most layers, and in some cases smaller by several orders of magnitude. These results further demonstrate that \Cref{gluon:ass:generalized-smoothness} provides an accurate description of the loss landscape, whereas classical smoothness, even when applied in a layer-wise manner, yields a substantially poorer fit.

\begin{figure}[h] \centering
  \begin{minipage}{0.32\textwidth}
    \includegraphics[width=\textwidth]{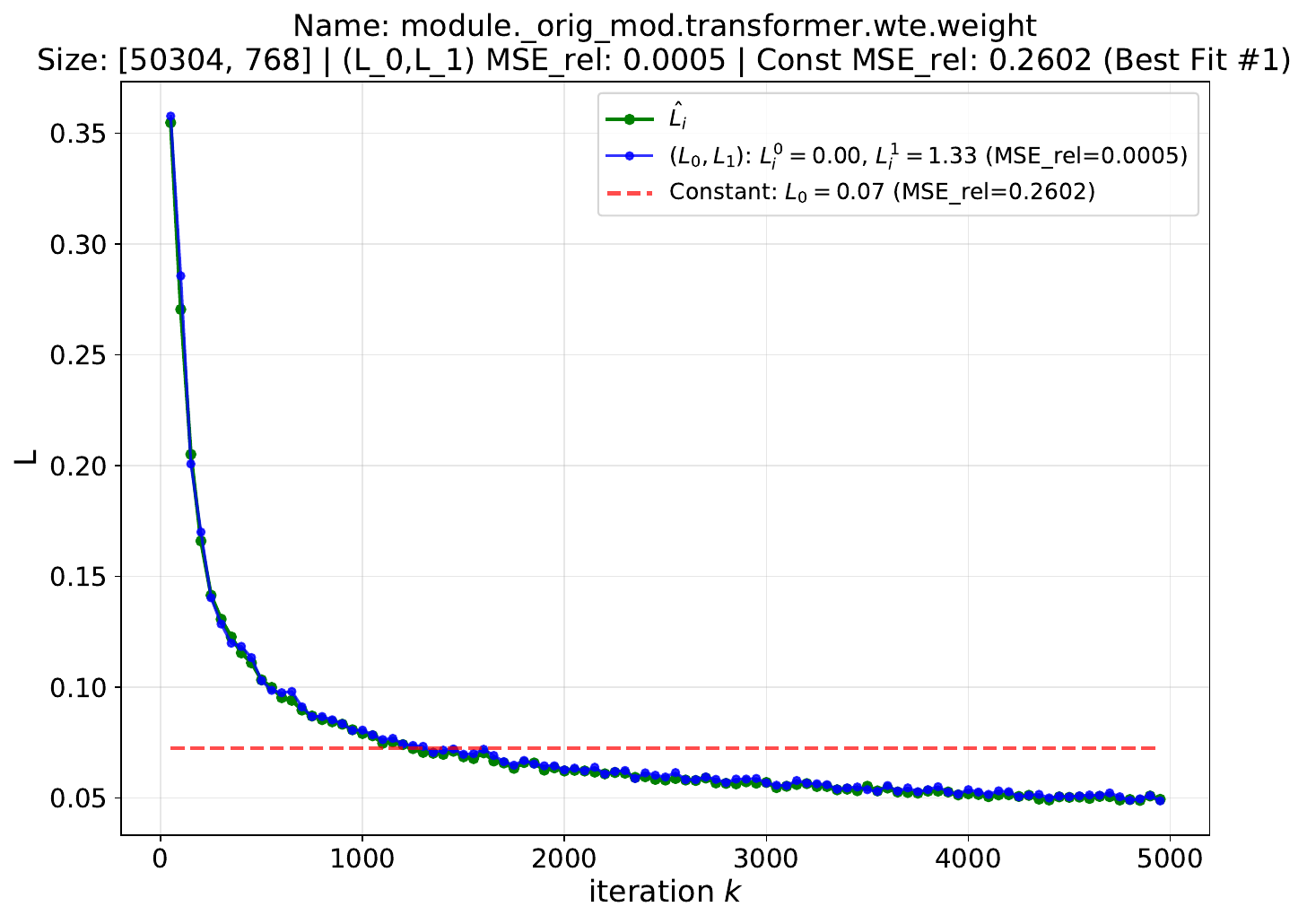}
  \end{minipage}\hfill
  \begin{minipage}{0.32\textwidth}
    \includegraphics[width=\textwidth]{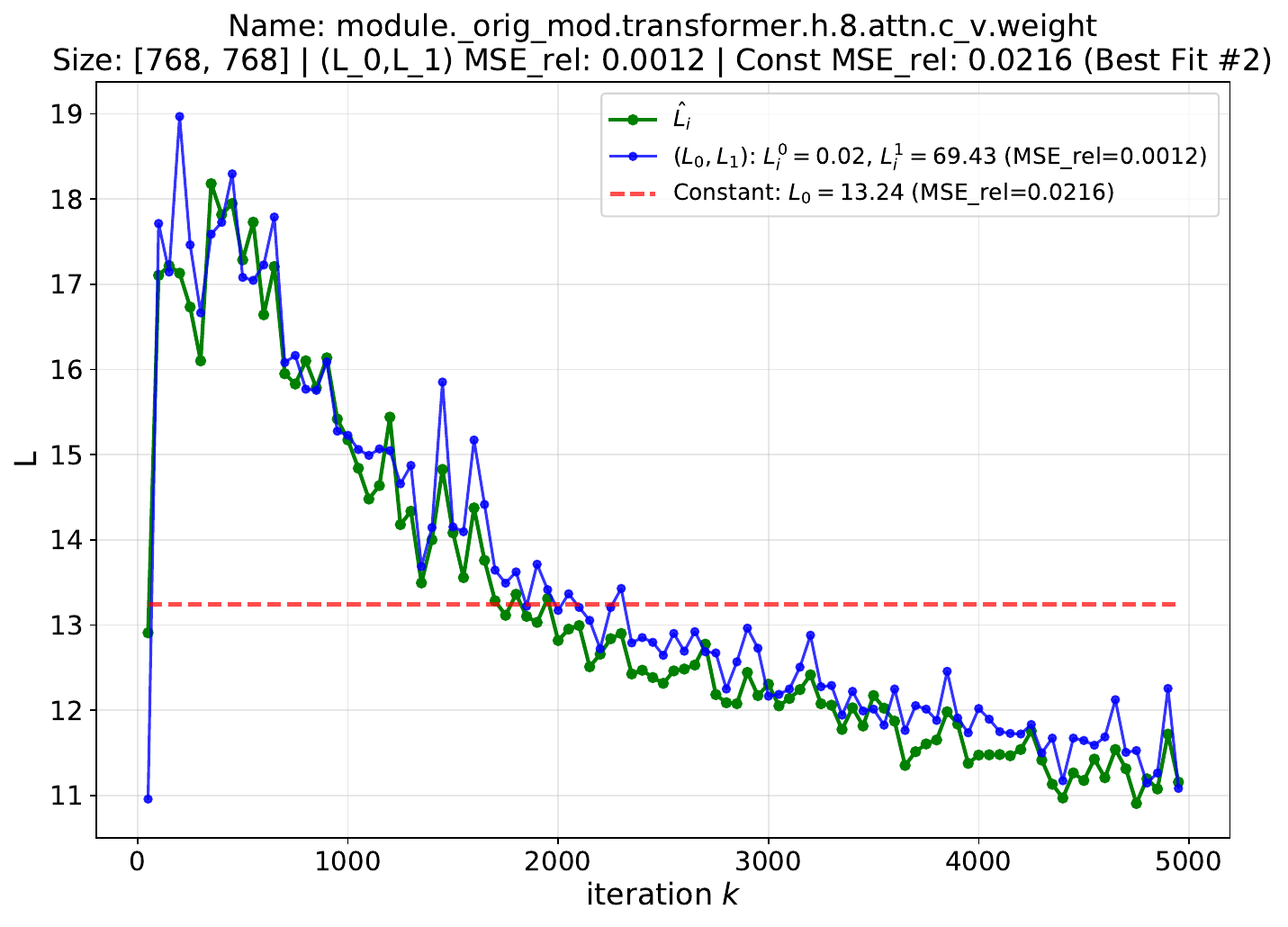}
  \end{minipage}\hfill
  \begin{minipage}{0.32\textwidth}
    \includegraphics[width=\textwidth]{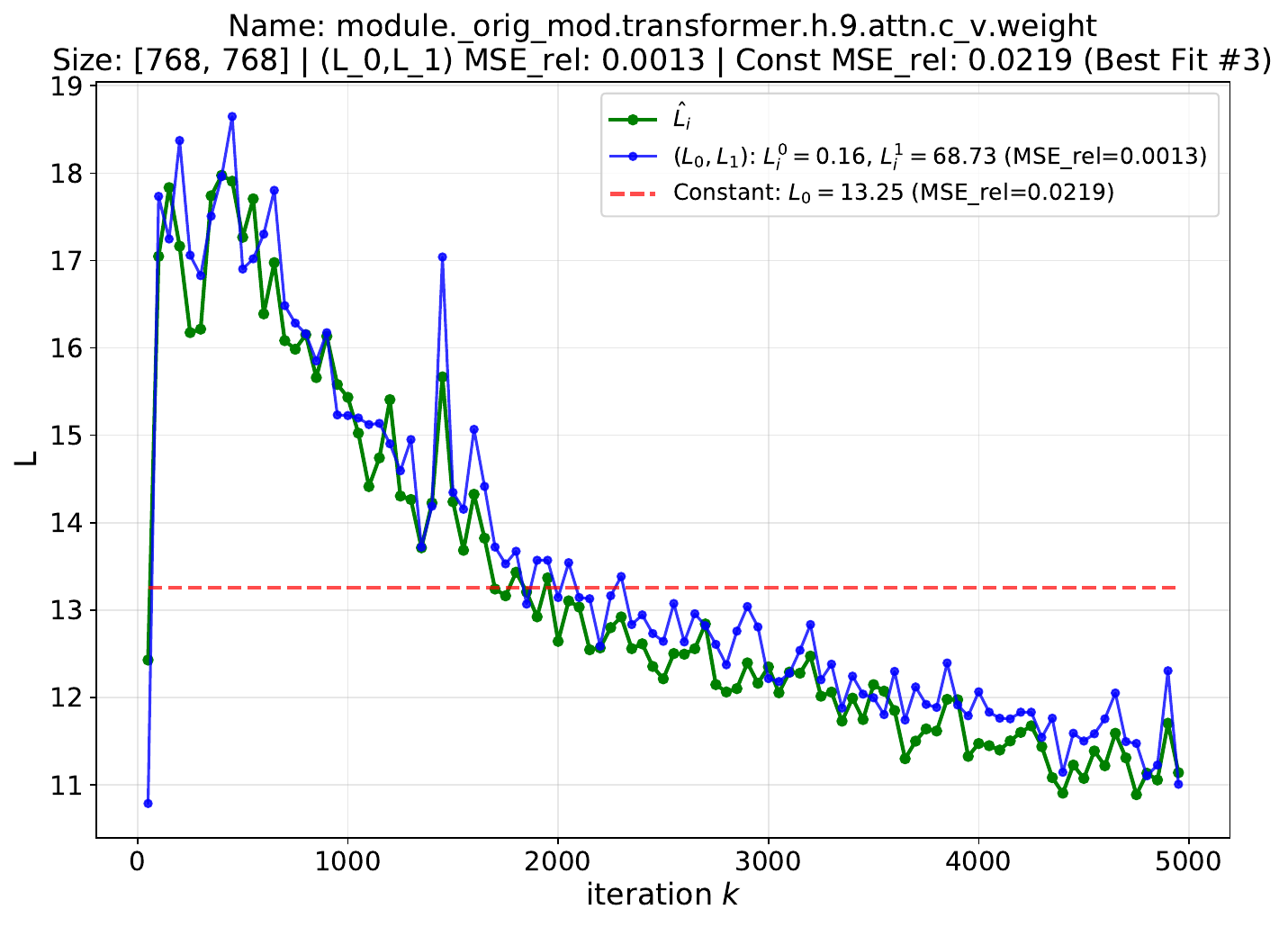}
  \end{minipage}

  \vspace{0.25em}

  \begin{minipage}{0.32\textwidth}
    \includegraphics[width=\textwidth]{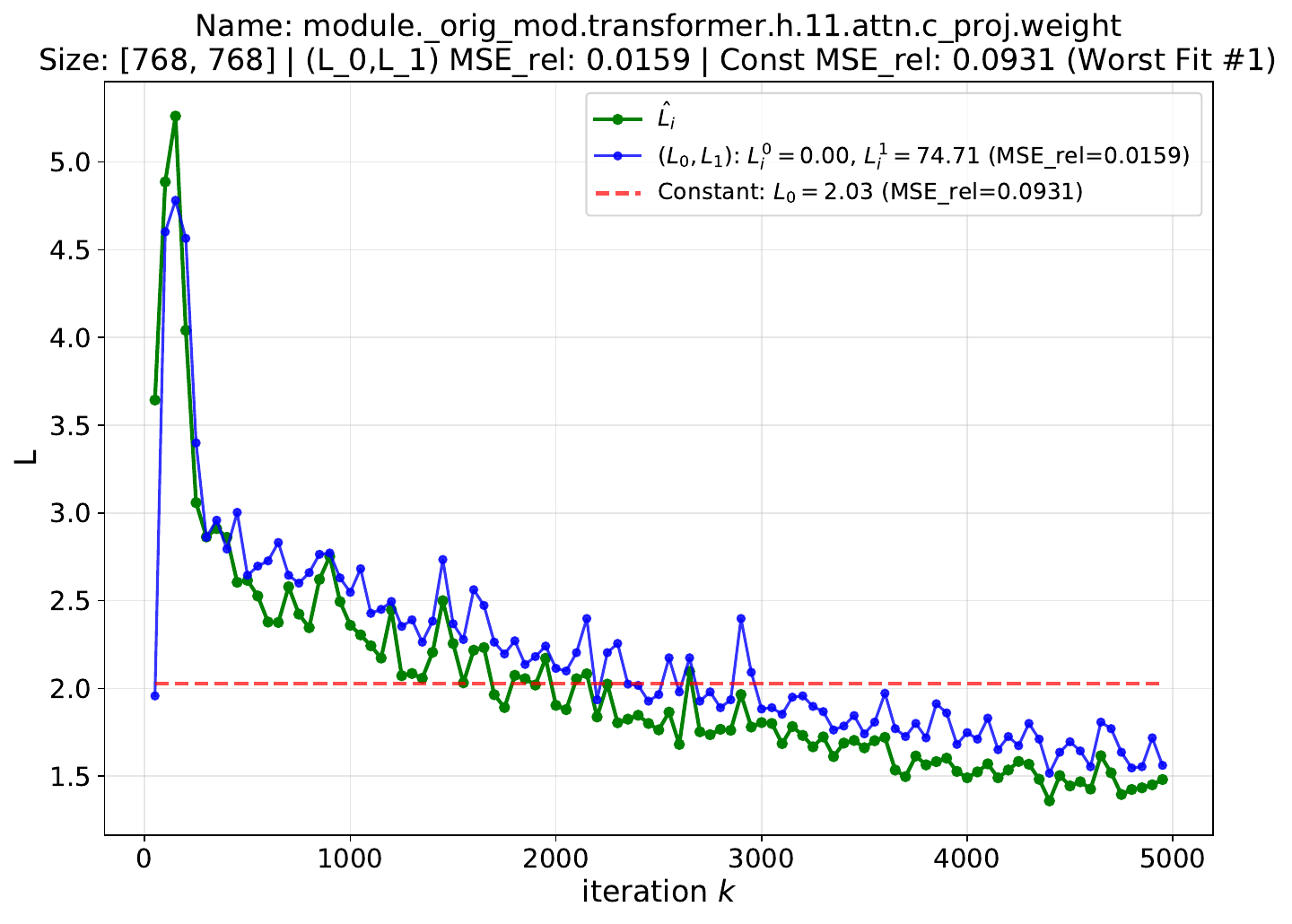}
  \end{minipage}\hfill
  \begin{minipage}{0.32\textwidth}
    \includegraphics[width=\textwidth]{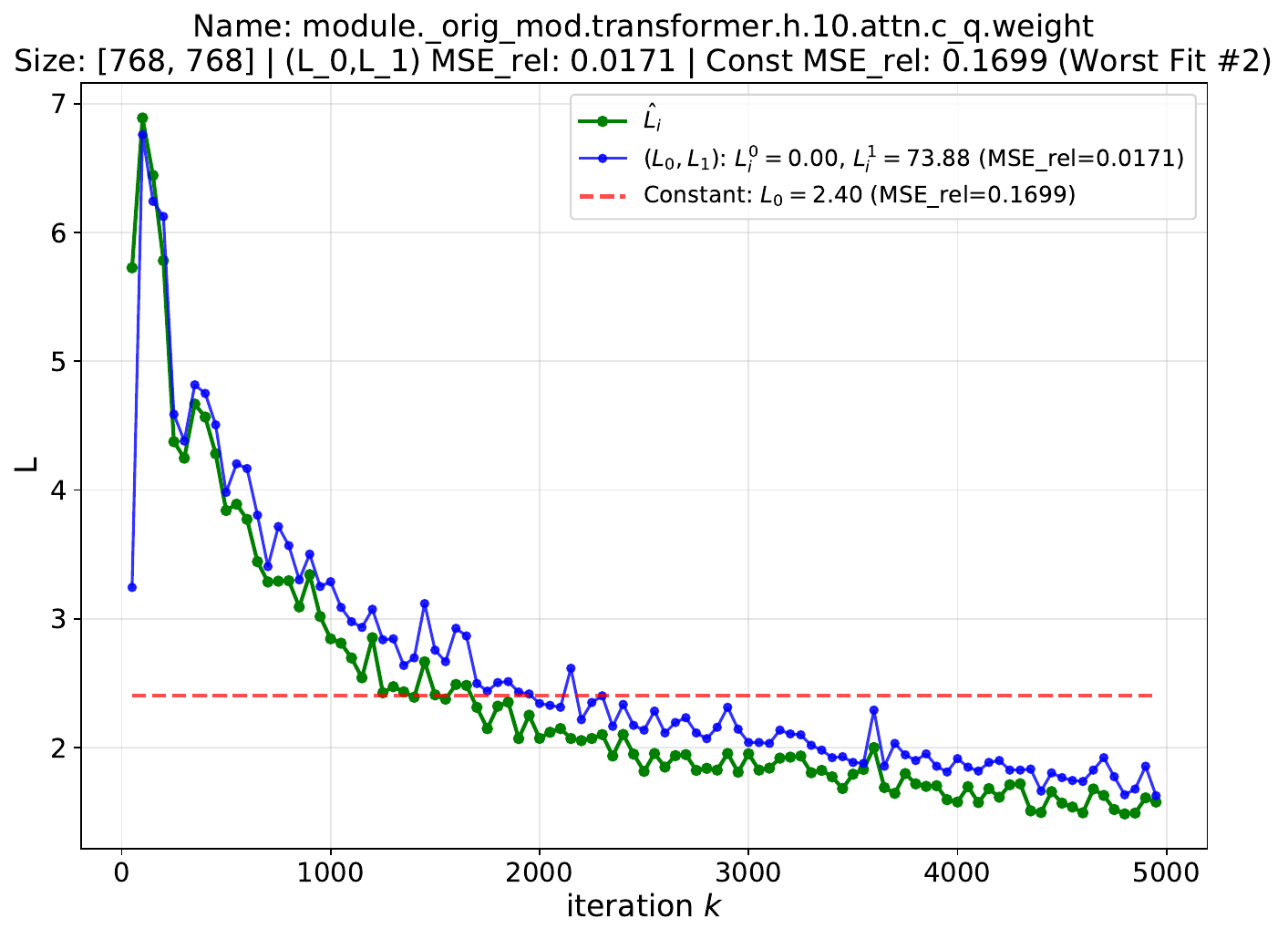}
  \end{minipage}\hfill
  \begin{minipage}{0.32\textwidth}
    \includegraphics[width=\textwidth]{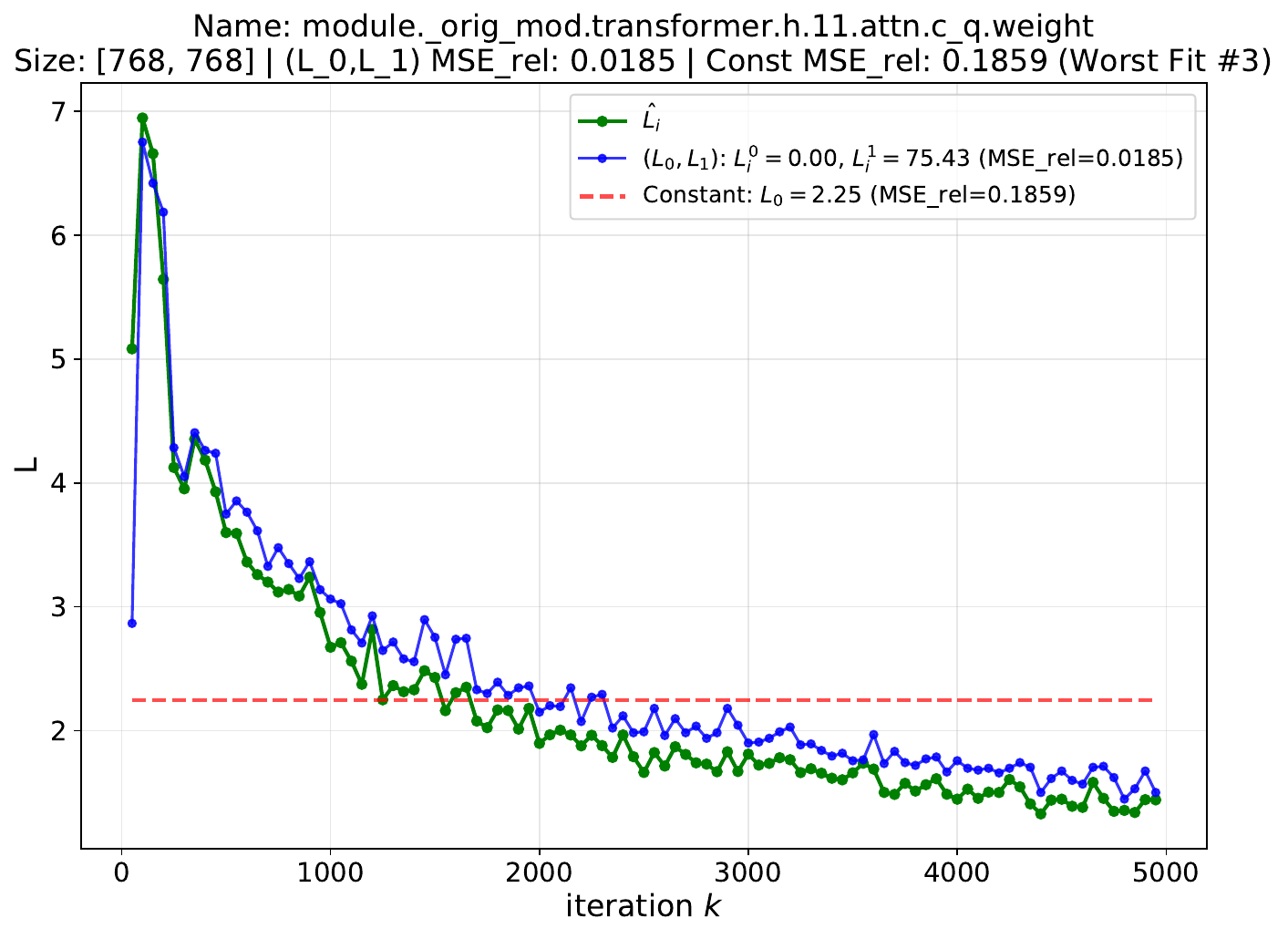}
  \end{minipage}

  \caption{Illustrative \emph{best} (top row) and \emph{worst} (bottom row) per‑layer fits of Assumption~1 (measured $L_i[k]$ vs.\ $L_i^{\text{approx}}[k]$) along \texttt{NanoGPT-124M} training.}%
  \label{gluon:fig:best_worst_fits}
\end{figure}

\begin{figure}[h]
  \centering
  \includegraphics[width=\textwidth]{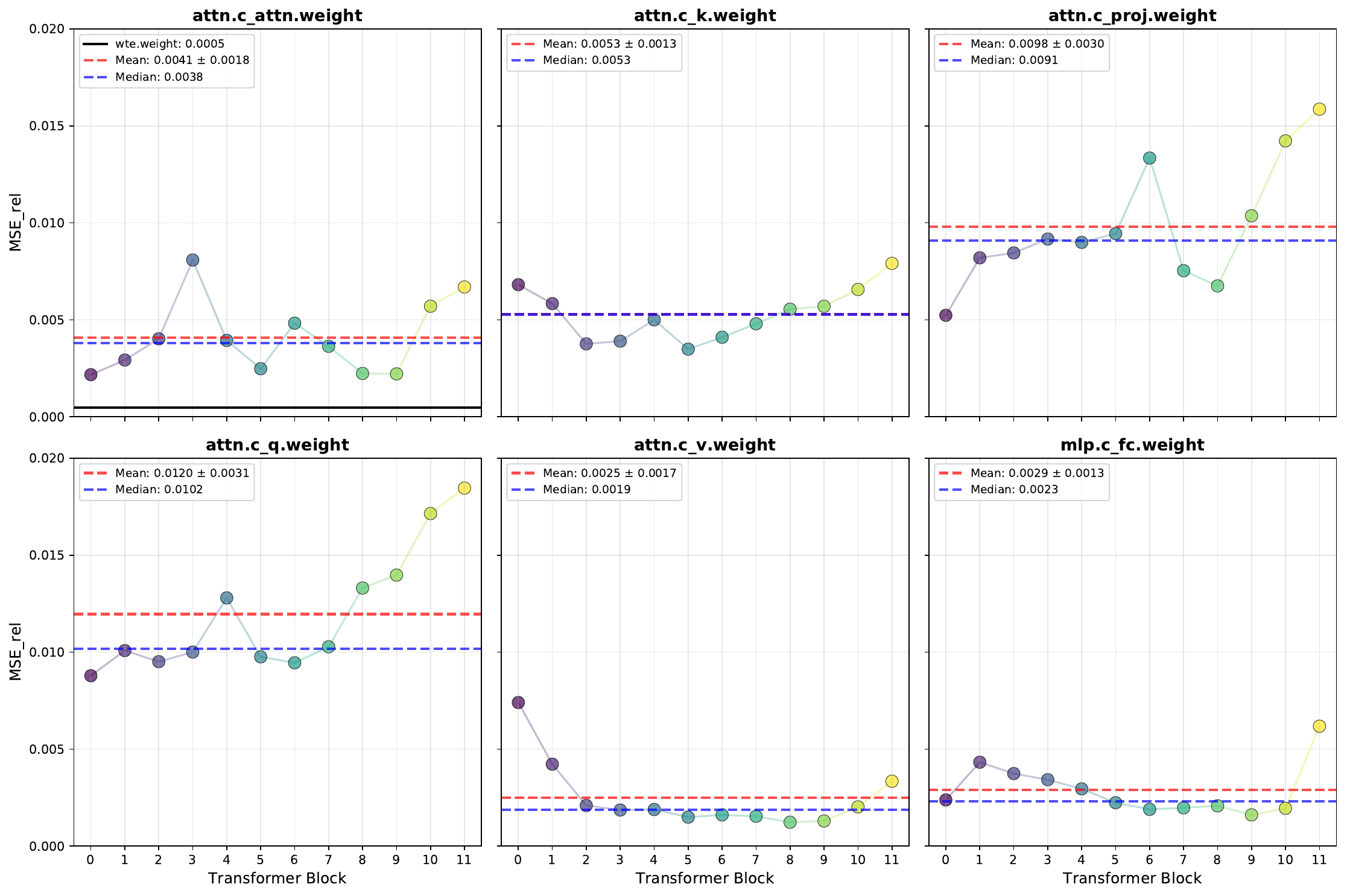}
  \caption{{\bf Relative fit $\mathrm{MSE}_{\mathrm{rel}}$ error of the layer-wise $(L^0_i,L^1_i)$ model} across transformer blocks in \texttt{NanoGPT-124M} for each matrix type. Dashed lines indicate the mean and median over blocks (embedding matrix: $\mathrm{MSE}_{\mathrm{rel}}=0.0005$).}
  \label{gluon:fig:fit_quality_by_matrix_type}
\end{figure}

\subsubsection{Layer-wise smoothness across layers and model scales}\label{gluon:appendix:layers_scales}

The final set of language modeling experiments examines whether the observations made for \texttt{NanoGPT-124M} extend to larger model scales. To this end, we additionally estimate trajectory smoothness for models trained with the \algname{unScion} optimizer at two larger scales: \texttt{GPT-2 Medium} ($\sim$355M parameters) and \texttt{GPT-2 Large} ($\sim$774M parameters). The estimation procedure follows the same methodology used in the preceding \texttt{NanoGPT-124M} experiments.

Figures~\ref{gluon:fig:all_layers_scales-124M}, \ref{gluon:fig:all_layers_scales-355M}, and \ref{gluon:fig:all_layers_scales-774M} report the aggregate layer-wise trajectory smoothness across all layers and blocks for the three model scales: \texttt{NanoGPT} ($\sim$124M), \texttt{GPT-2 Medium} ($\sim$355M), and \texttt{GPT-2 Large} ($\sim$774M). Across all scales, we consistently observe pronounced cross-layer heterogeneity and the empirical trend $L_i^0 \approx 0$, indicating that these phenomena persist from 124M up to 774M parameters.

\begin{figure}[h]
    \centering
    \includegraphics[width=\linewidth]{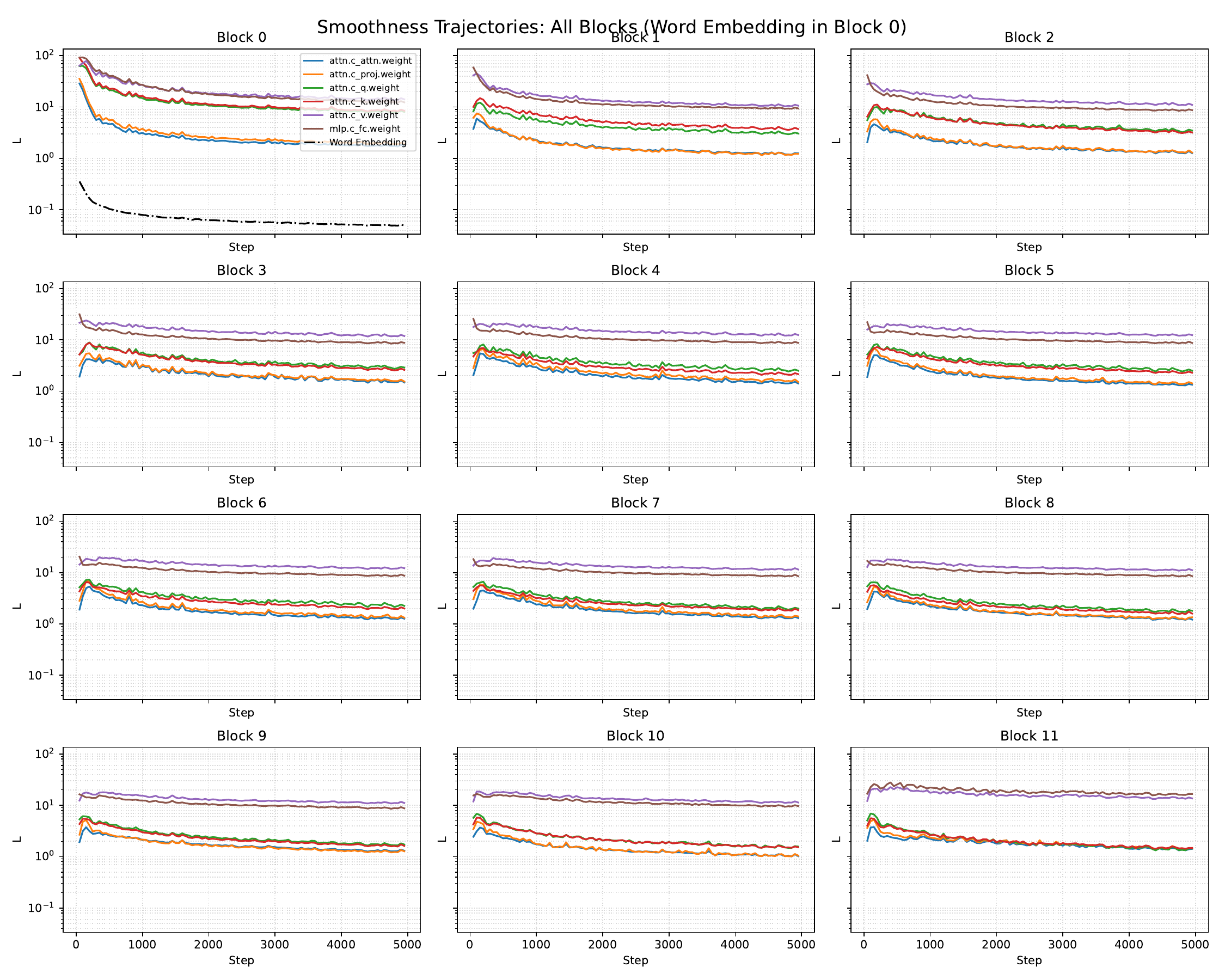}
    \caption{Estimated trajectory smoothness for \texttt{NanoGPT-124M} trained with \algname{unScion}.}
    \label{gluon:fig:all_layers_scales-124M}
\end{figure}

\begin{figure}[h]
    \centering
    \includegraphics[width=0.84\textwidth]{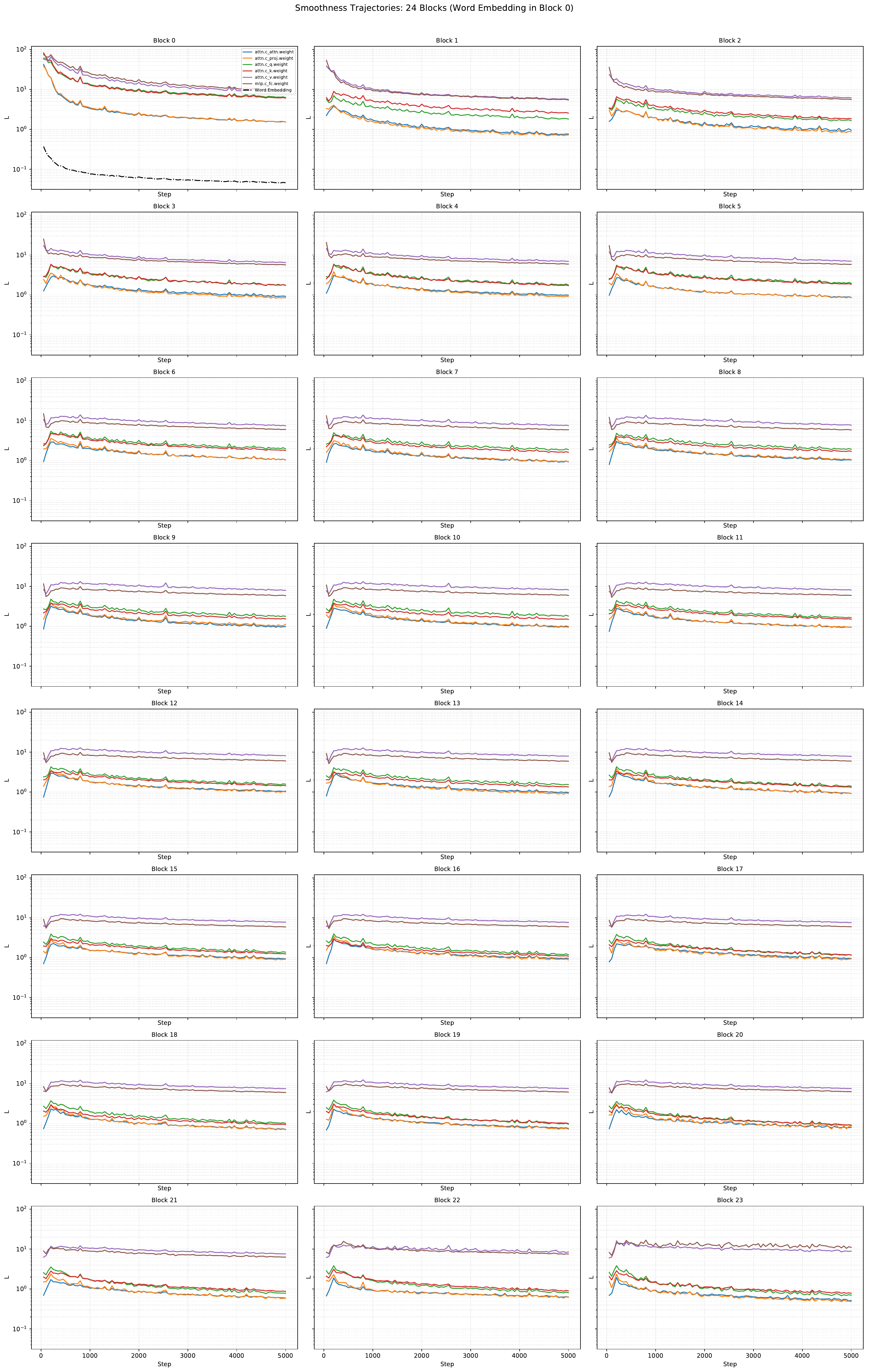}
    \caption{Estimated trajectory smoothness for \texttt{GPT-2 Medium} ($\sim$355M parameters) trained with \algname{unScion}.}
    \label{gluon:fig:all_layers_scales-355M}
\end{figure}

\begin{figure}[p]
    \centering
    \includegraphics[width=0.78\textwidth]{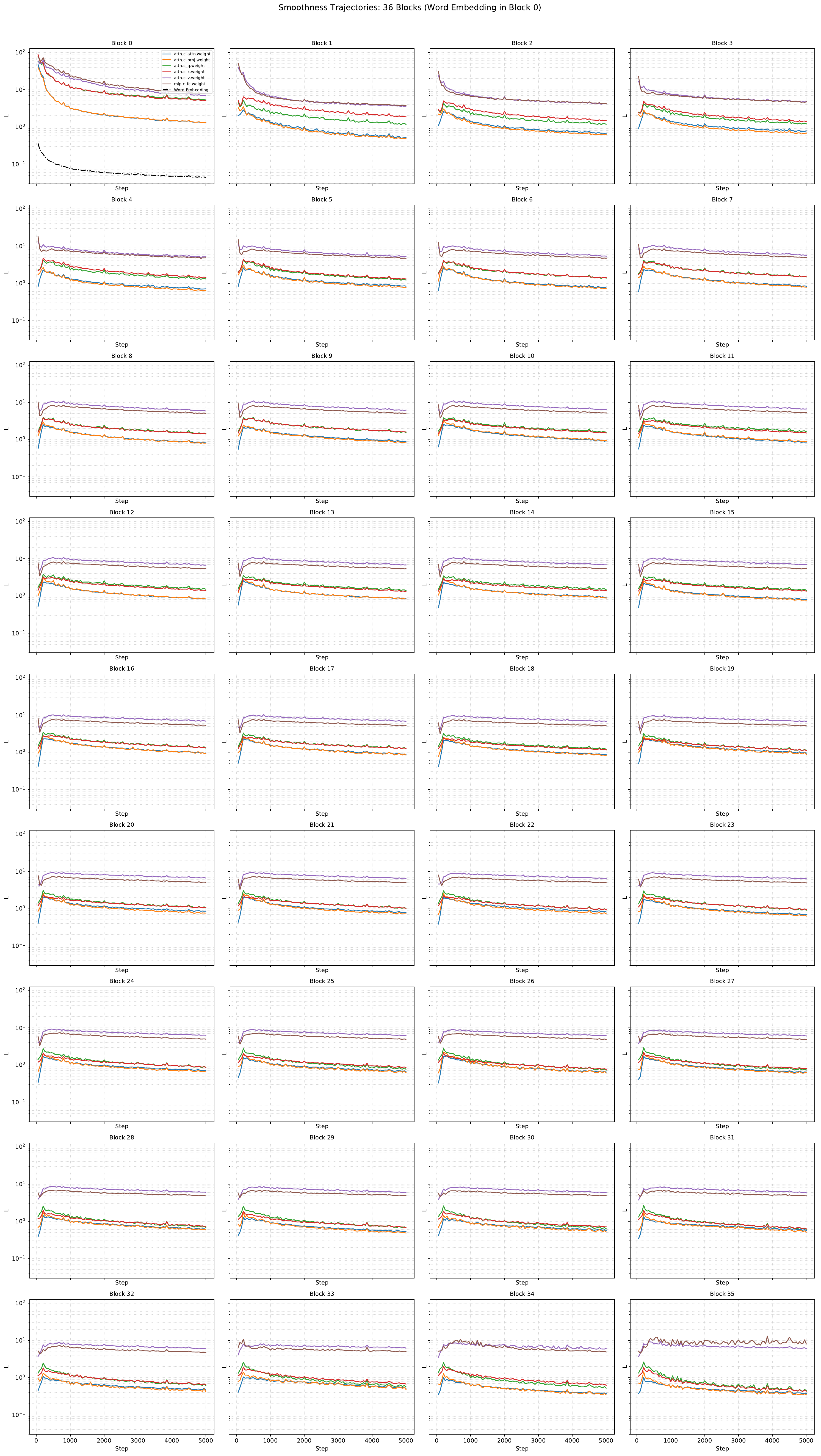}
    \caption{Estimated trajectory smoothness for \texttt{GPT-2 Large} ($\sim$774M parameters) trained with \algname{unScion}.}
    \label{gluon:fig:all_layers_scales-774M}
\end{figure}

\subsection{Training CNN on CIFAR-10} \label{gluon:appendix:cifar}

In this experiment, we further validate layer-wise $(L^0, L^1)$-smoothness by training a \texttt{CNN} model on the \texttt{CIFAR-10} dataset, following implementations from two open-source GitHub repositories~\citep{jordan2024cifar10,code_scion}.

\paragraph{Full-batch (deterministic) gradients.}

\begin{figure}[!b]
    \centering
    \begin{minipage}{0.32\textwidth}
        \includegraphics[width=\textwidth]{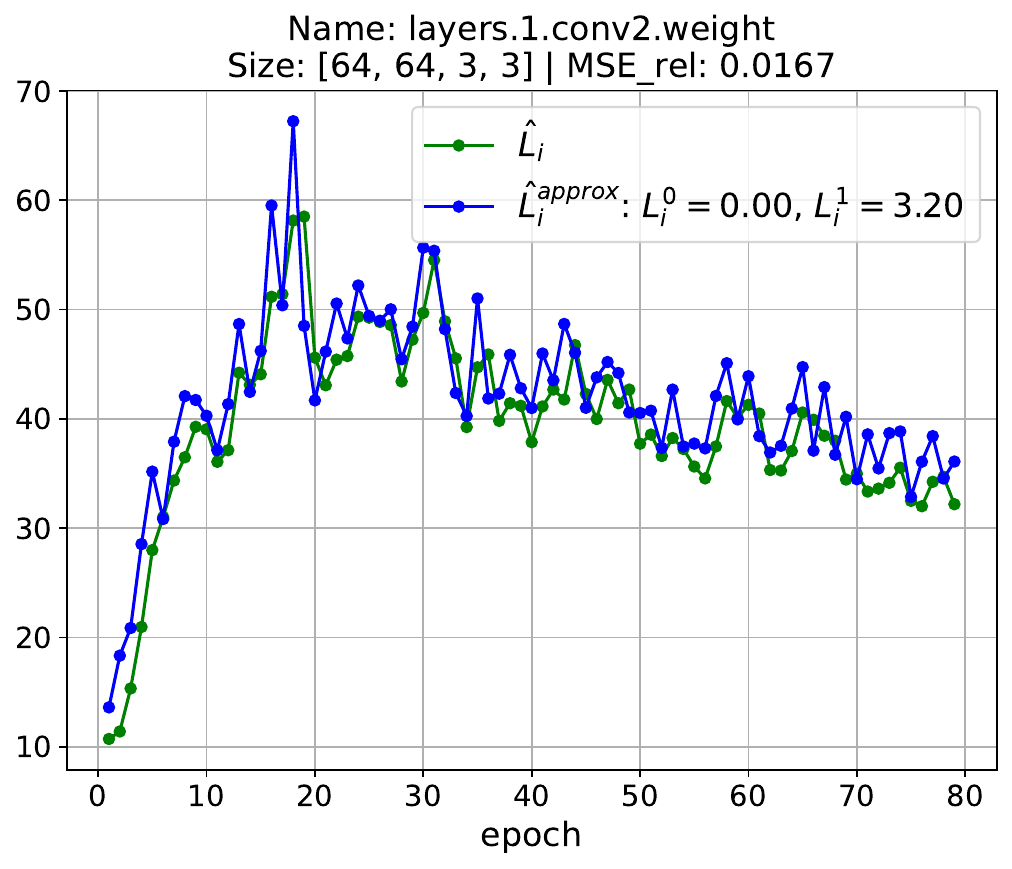}
    \end{minipage}
    \hfill
    \begin{minipage}{0.32\textwidth}
        \includegraphics[width=\textwidth]{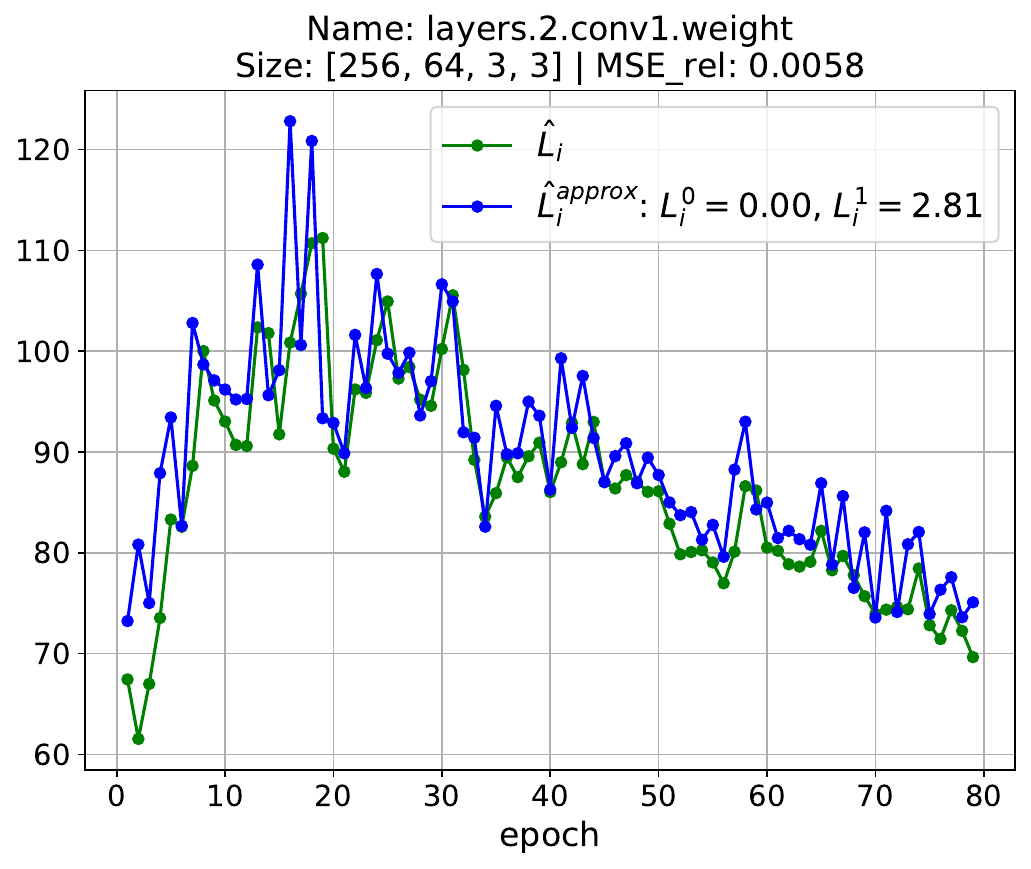}
    \end{minipage}
    \hfill
    \begin{minipage}{0.32\textwidth}
        \includegraphics[width=\textwidth]{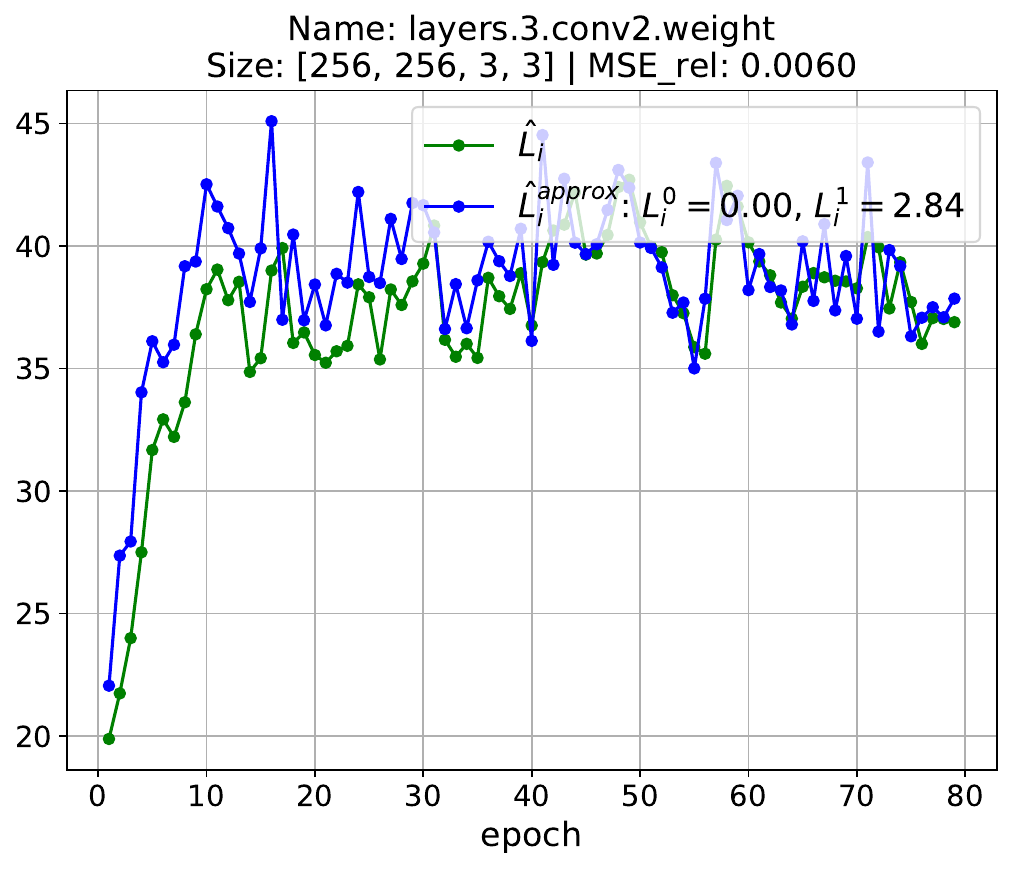}
    \end{minipage}

    \vspace{0.5cm}

    \begin{minipage}{0.32\textwidth}
        \includegraphics[width=\textwidth]{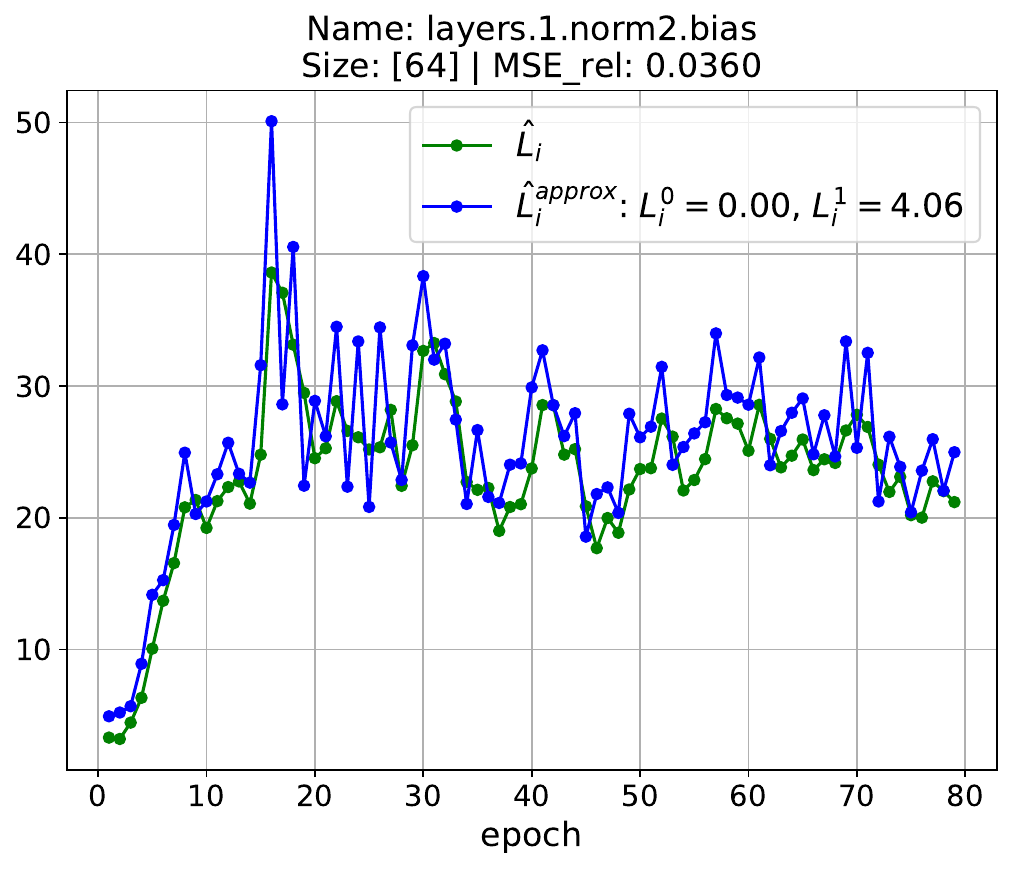}
    \end{minipage}
    \hfill
    \begin{minipage}{0.32\textwidth}
        \includegraphics[width=\textwidth]{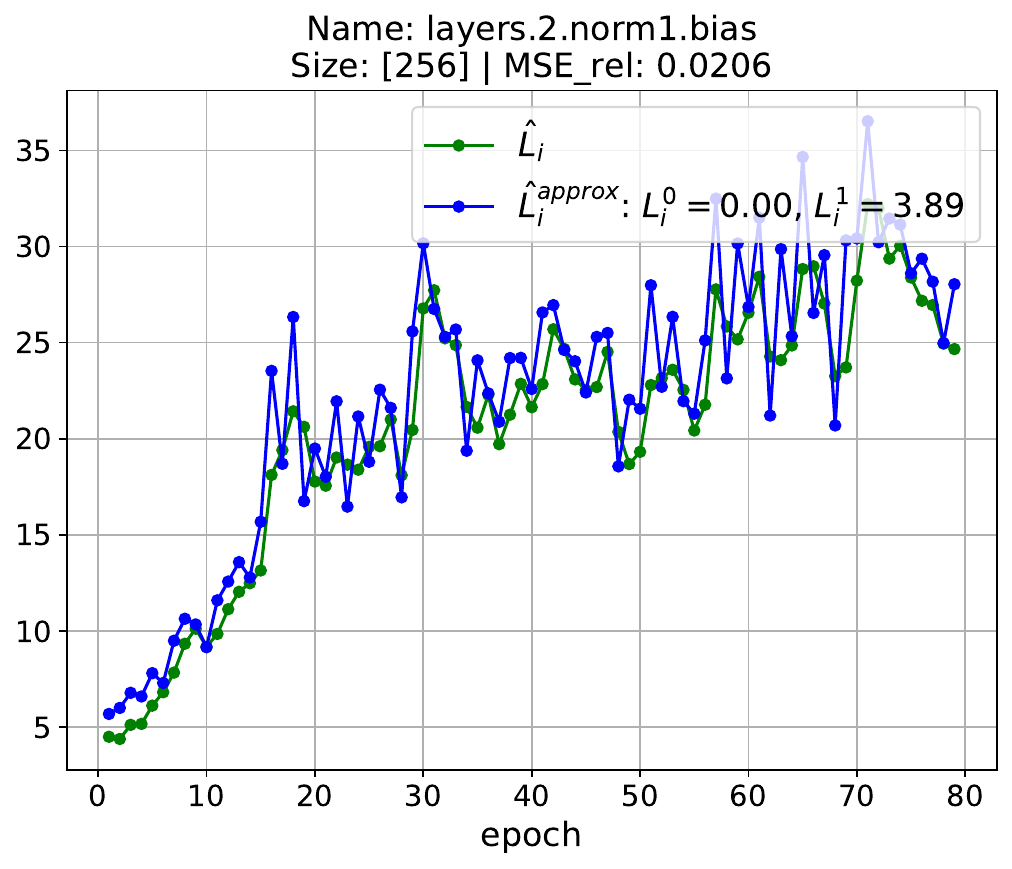}
    \end{minipage}
    \hfill
    \begin{minipage}{0.32\textwidth}
        \includegraphics[width=\textwidth]{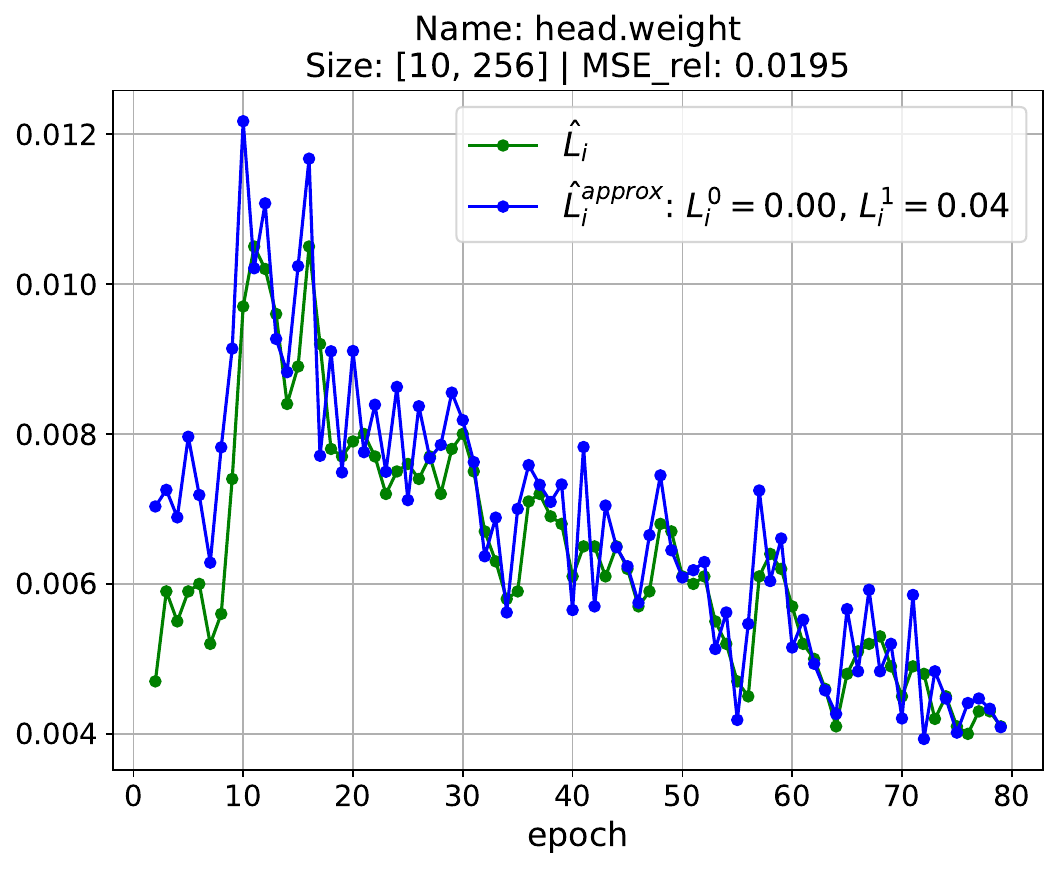}
    \end{minipage}

    \caption{\small {\bf Validation of layer-wise $(L^0, L^1)$-smoothness} for different groups of parameters of a \texttt{CNN} model along the training trajectories of \algname{unScion} with \textbf{full-batch gradients}. The norms used for each group are as follows: $\|\cdot\|_{(i)} = \sqrt{\nicefrac{1}{C_i^{out}}}\|\cdot\|_2$ for biases, $ \|\cdot\|_{(i)} = k^2 \sqrt{\nicefrac{C_i^{in}}{C_i^{out}}} \|\cdot\|_{2 \to 2} $ for conv, and $\|\cdot\|_{(p)} = n_p \|\cdot\|_{1 \to \infty}$ for the last group $ X_p $, associated with classification head weights.}
    \label{gluon:fig:5_appendix}
\end{figure}

We begin by presenting results in the deterministic setting, where the model is trained using the \algname{unScion} optimizer~\eqref{gluon:eq:scion_cnn} with full-batch gradients $ \nabla_i f $, no momentum and no learning rate decay. Other hyperparameters are as in \citet[Table 10]{pethick2025training}, except that we train for more epochs.
Similar to the \texttt{NanoGPT} experiments discussed in \Cref{gluon:sec:experiments}, we plot the estimated (non-stochastic) trajectory smoothness
\begin{align*}
    \hat{L}_i[k] \eqdef \frac{\|\nabla_i f (X^{k+1}) - \nabla_i f (X^{k}) \|_{(i)\star}}{\|X_i^{k+1} - X_i^{k}\|_{(i)}}
\end{align*}
alongside its approximation
\begin{align*}
    \hat{L}_i^{\text{approx}}[k] \eqdef L_i^1 \|\nabla_i f (X^{k+1})\|_{(i)\star}
\end{align*}
for selected parameter groups. In this experiment, we consider a simplified variant of \Cref{gluon:ass:generalized-smoothness}, setting $L^0_i = 0$, and estimate $L^1_i\geq0$ using the same procedure as in \Cref{gluon:sec:experiments}.
Figure~\ref{gluon:fig:5_appendix} presents the results, demonstrating that \Cref{gluon:ass:generalized-smoothness} is approximately satisfied along the training trajectory.
When this condition holds with $L^0_i=0$, \Cref{gluon:theorem:1} guarantees convergence under the stepsize choice $t^k_i \equiv t_i =\nicefrac{1}{L^1_i}$.
In this setting, the estimated $L^1_i$ values (shown in Figure~\ref{gluon:fig:5_appendix}) are $L^1_i \approx 3$-$4$ for all parameter groups except for the classification head weights $X_p$, where $L^1_p \approx 0.04$.
This roughly two-orders-of-magnitude difference justifies the much larger radius $t_p^k$ used for the head weights in the tuned configuration reported in \citet[Table 10]{pethick2025training}.

\paragraph{Stochastic gradients.}

\begin{figure}[h]
    \centering
    \begin{minipage}{0.32\textwidth}
        \includegraphics[width=\textwidth]{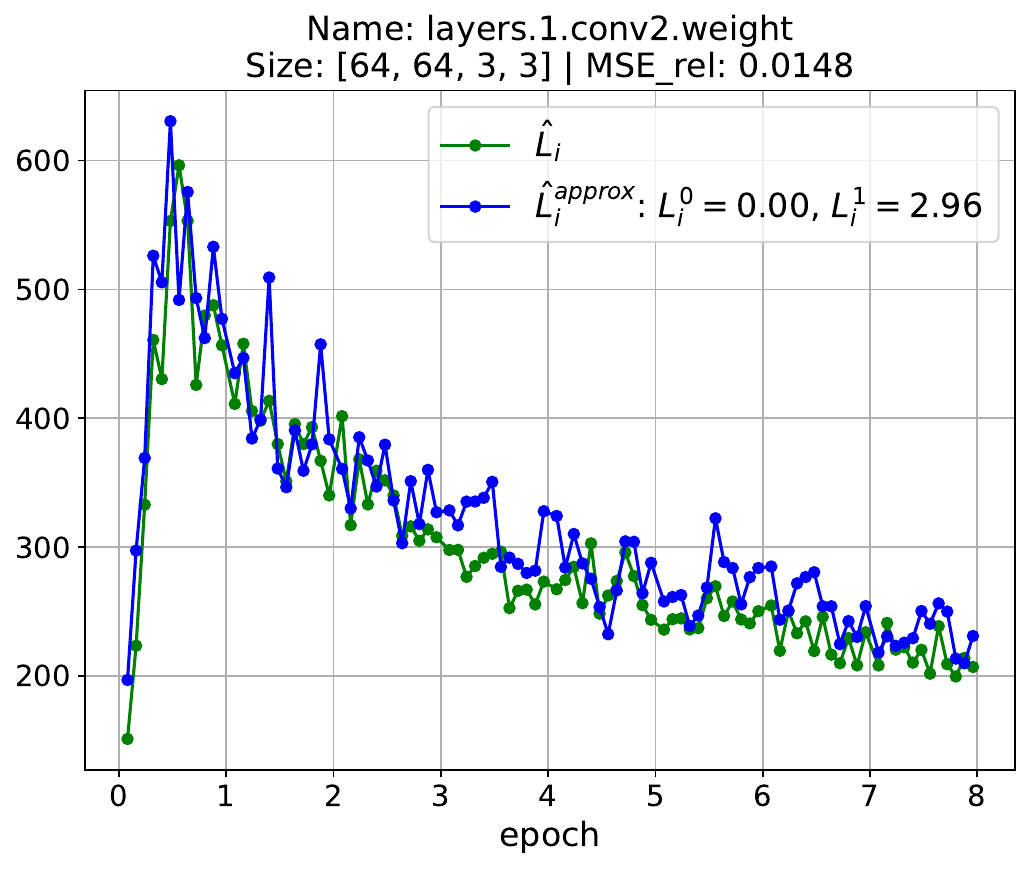}
    \end{minipage}
    \hfill
    \begin{minipage}{0.32\textwidth}
        \includegraphics[width=\textwidth]{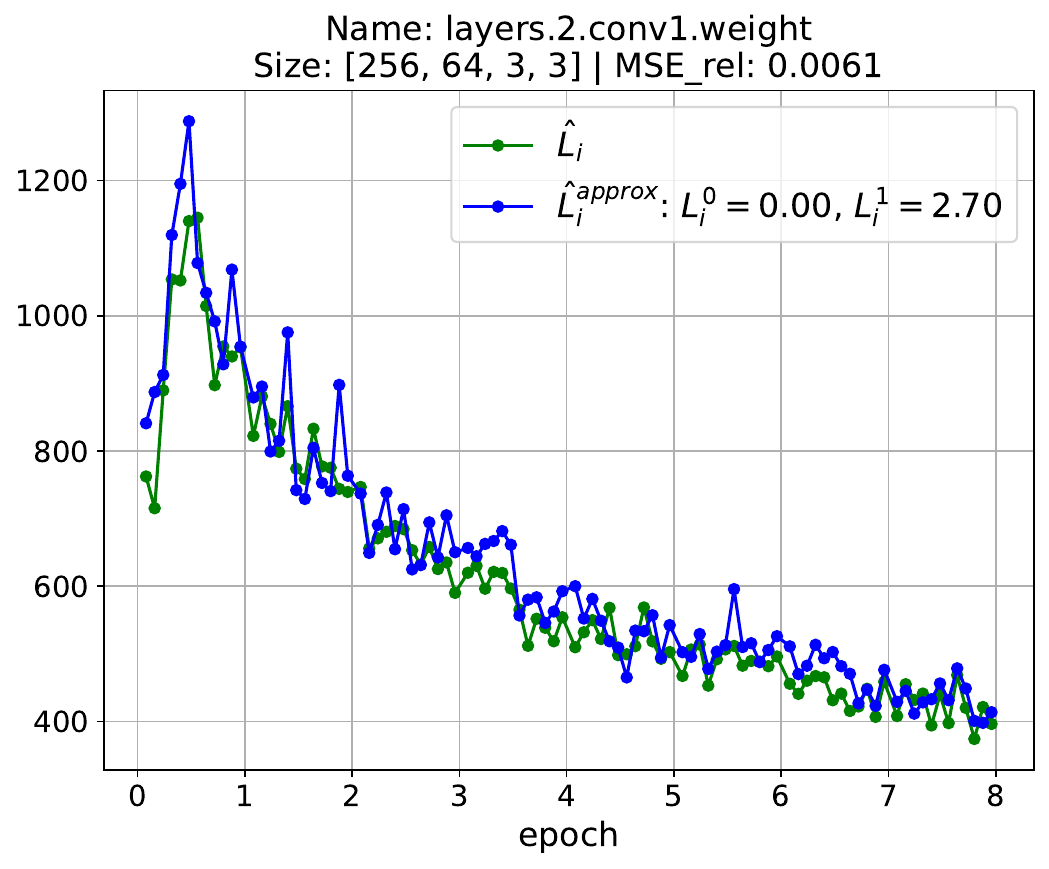}
    \end{minipage}
    \hfill
    \begin{minipage}{0.32\textwidth}
        \includegraphics[width=\textwidth]{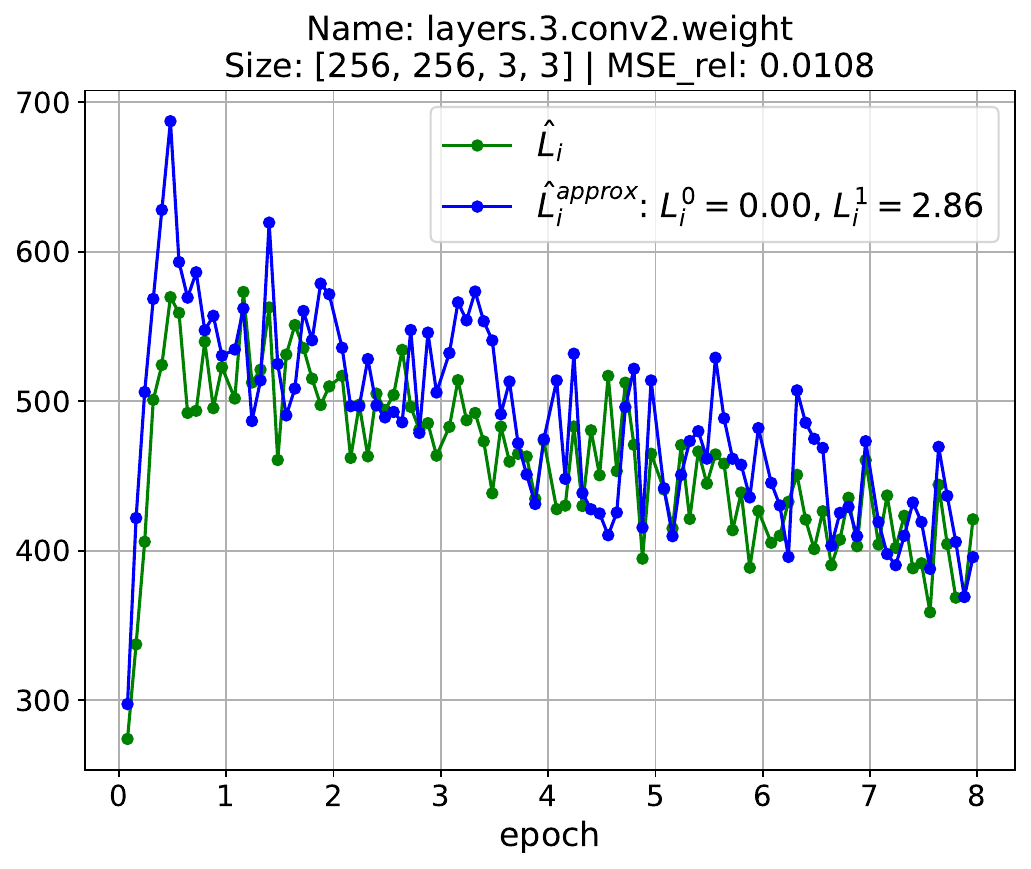}
    \end{minipage}

    \vspace{0.5cm}

    \begin{minipage}{0.32\textwidth}
        \includegraphics[width=\textwidth]{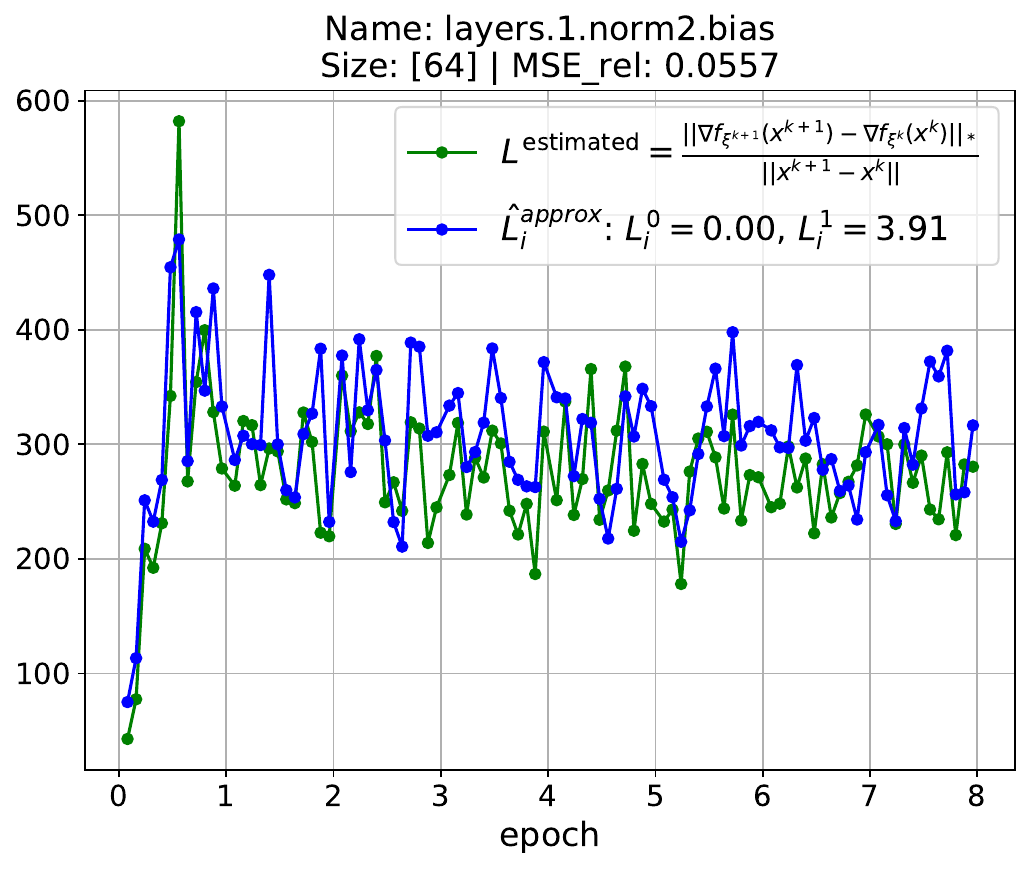}
    \end{minipage}
    \hfill
    \begin{minipage}{0.32\textwidth}
        \includegraphics[width=\textwidth]{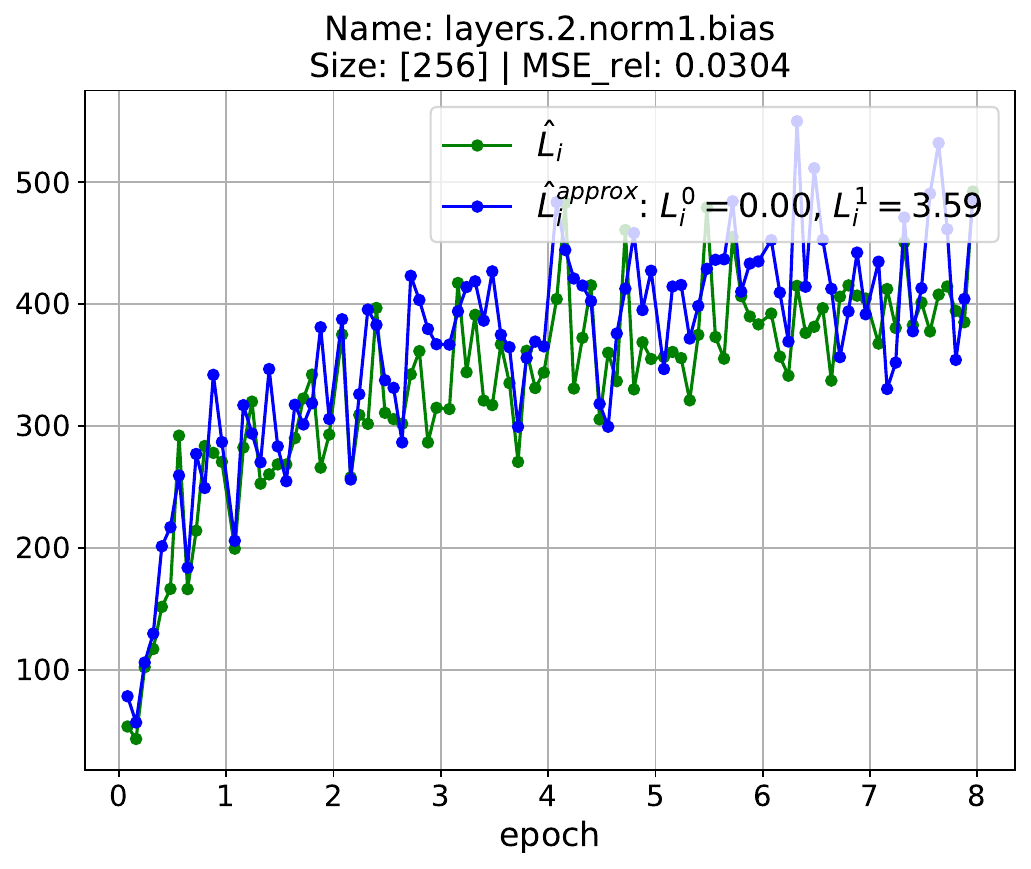}
    \end{minipage}
    \hfill
    \begin{minipage}{0.32\textwidth}
        \includegraphics[width=\textwidth]{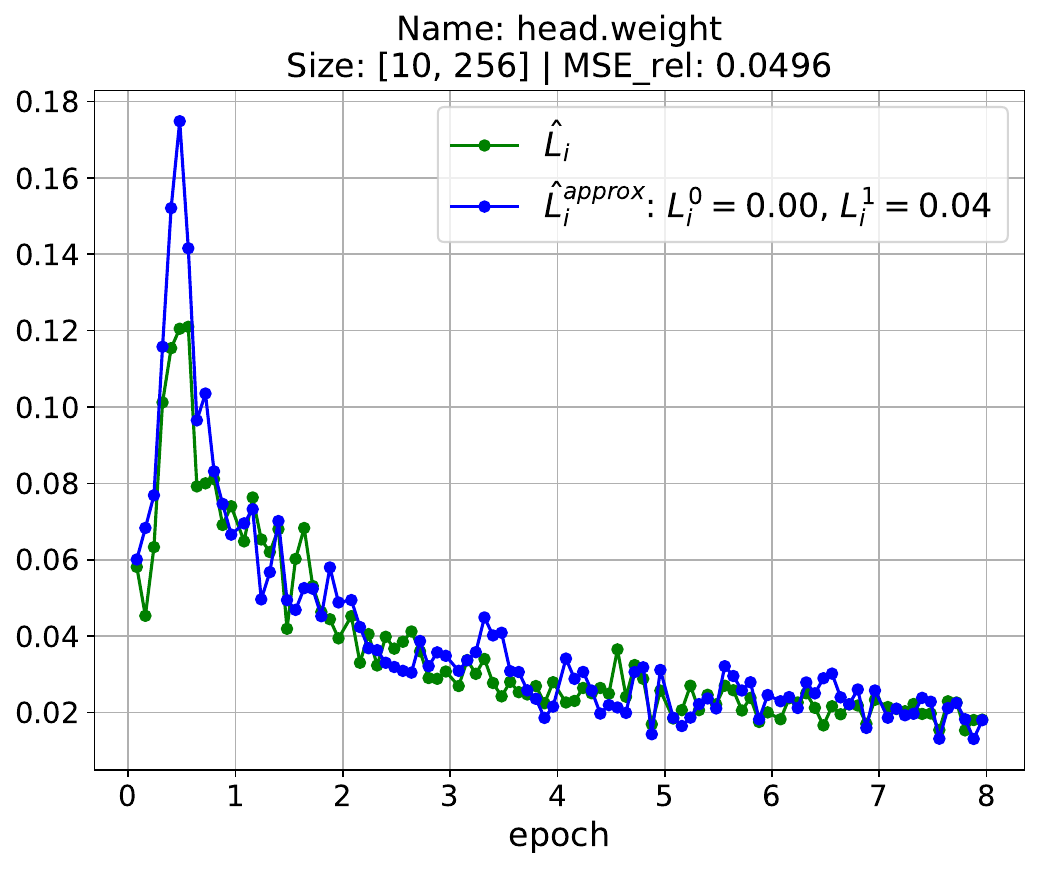}
    \end{minipage}

    \caption{\small {\bf Validation of layer-wise $(L^0, L^1)$-smoothness} for different groups of parameters of a \texttt{CNN} model along the training trajectories of \algname{unScion} with \textbf{stochastic gradients}. The norms used for each group are as follows: $\|\cdot\|_{(i)} = \sqrt{\nicefrac{1}{C_i^{out}}}\|\cdot\|_2$ for biases, $ \|\cdot\|_{(i)} = k^2 \sqrt{\nicefrac{C_i^{in}}{C_i^{out}}} \|\cdot\|_{2 \to 2} $ for conv, and $\|\cdot\|_{(p)} = n_p \|\cdot\|_{1 \to \infty}$ for the last group $ X_p $, associated with classification head weights.}
    \label{gluon:fig:4}
\end{figure}

Here, we report results for analogous experiments in the stochastic setting, using noisy gradients $ \nabla_i f_{\xi^k} $. We use momentum as in \citet[Table 10]{pethick2025training}, but do not apply a linear decay schedule.
In Figure~\ref{gluon:fig:4}, we plot
\[
\hat{L}_i[k] = \frac{\|\nabla_i f_{\xi^{k+1}}(X^{k+1}) - \nabla_i f_{\xi^k}(X^k)\|_{(i)\star}}{\|X_i^{k+1} - X_i^k\|_{(i)}}, \qquad \hat{L}_i^{\text{approx}}[k] = L^1_i \|\nabla_i f_{\xi^{k+1}}(X^{k+1})\|_{\star},
\]
again setting $L^0_i = 0$. Despite the added variance, we still observe that the stochastic trajectory roughly adheres to Assumption~\ref{gluon:ass:generalized-smoothness}.

Altogether, the results further support the validity of Assumption~\ref{gluon:ass:generalized-smoothness} with $L^0_i = 0$.

\endgroup

\chapter{Supplementary Material for Chapter 7 (Beyond the Ideal)}
\label{appendix:chapter7_beyond_the_ideal}

\begingroup
\input{KAUST_Thesis/imported/07_beyond_the_ideal/import_setup}

\section{Justification of the inexact LMO error model}
\label{beyond:app:error_model}

In this section, we provide additional details on how practical approximation schemes for the orthogonalized update satisfy Assumption~\ref{beyond:assump:inexact_lmo_main}.

The \approxalg{PolarExpress} algorithm~\citep{amsel2025polar} is designed to approximate the polar factor $U = \operatorname{polar}(M)$ of a matrix $M$ through iterative refinement. Theorem 4.3 in~\citet{amsel2025polar} establishes that for a matrix $M$ with singular values normalized to lie in $[\ell, 1]$, the \approxalg{PolarExpress} approximation after $r = 2q+1$ (odd) iterations satisfies:
\begin{equation*}
    \|\operatorname{polar}(M) - X_p\|_2 \le |1-\ell^2|^{(q+1)^p},
\end{equation*}
where $\|\cdot\|_2$ denotes the spectral norm (largest singular value). For $r=3$ this yields quadratic convergence, and for $r=5$ cubic convergence. We refer to~\citet{amsel2025polar} for a comprehensive analysis including extensions to rectangular matrices and detailed convergence properties.

The classical \approxalg{Newton-Schulz} iteration~\citep{higham2008functions} exhibits similar convergence properties, with the approximation error decreasing rapidly with the number of iterations.

Since these methods are always run for a small, fixed number of steps in practice (e.g., $p=5$ in \algname{Muon}), their error is bounded by a constant $\err$. Different iteration counts correspond to different values of $\err$, with more iterations producing smaller error. This directly justifies the additive error model in Assumption~\ref{beyond:assump:inexact_lmo_main}.

\subsection{Additional discussion of recent closely related work}
\label{beyond:app:additional_related_work}

Several recent papers reinforce the same broad message as ours, that the quality of the orthogonalization step matters, but from more specialized angles. In contrast, our paper studies a general additive inexact-LMO model, allows infeasible approximate directions, covers both deterministic and stochastic settings, and derives explicit $\delta$-dependent prescriptions for the step size and momentum.

\paragraph{SUMO.}
\citet{refael2025sumo} analyze the approximation error of the commonly used Newton--Schulz5 orthogonalization and show that it worsens with the condition number of the moment matrix. They then compare Muon using Newton--Schulz5 to exact SVD-based orthogonalization and advocate replacing the former by exact orthogonalization in an adaptive low-rank subspace. Thus, their treatment of inexactness is algorithm-specific and condition-number-driven, and the remedy is to modify the optimizer itself. We view this as complementary to our work: they provide a concrete mechanism explaining \emph{why} approximation quality can deteriorate, whereas our analysis isolates how a generic inexactness level $\delta$ propagates into convergence rates and tuning rules.

\paragraph{Low-rank orthogonalization.}
\citet{he2025lowrankorth} propose a different inner solver based on low-rank orthogonalization via sketching, and then build low-rank matrix-signed gradient descent and a low-rank variant of Muon on top of it. Their theory is tailored to this modified direction computation, giving deterministic stationarity guarantees for the low-rank signed method and stochastic guarantees for low-rank Muon under heavy-tailed noise. In particular, they do not analyze the original practical Muon update under a generic approximation model; instead, they replace the orthogonalization step by a low-rank surrogate. The complementary insight is that orthogonalization error may be reduced not only by running more inner iterations, but also by restricting attention to a more informative low-rank subspace.

\paragraph{FedMuon.}
\citet{takezawa2025fedmuon} study inexact Newton--Schulz orthogonalization in \algname{FedMuon}. The most relevant point for our paper is that, in the spectral-norm/Newton--Schulz setting, they prove a solver-specific result that is in some sense sharper than a generic perturbation bound: the method can converge for \emph{any} number of Newton--Schulz iterations, with the approximate-LMO guarantee stated in a $q$-dependent Schatten-$p$ norm that improves toward the trace-norm guarantee as $q$ increases. This differs substantially from our abstraction, which is solver-agnostic and based on an additive direction error $\|\hat d-d\|\le \delta$. We therefore view \citet{takezawa2025fedmuon} as complementary evidence that specialized analyses can exploit the structure of a particular solver more sharply than a general inexact-LMO framework.

\paragraph{Convergence of Muon with Newton--Schulz.}
The closest concurrent work is \citet{kim2026muonns}, which analyzes the original Muon algorithm with finitely many Newton--Schulz steps. Their results are specialized to operator-nuclear geometry and show that Muon with Newton--Schulz matches the exact SVD-polar rate up to a $q$-dependent constant factor, which approaches $1$ doubly exponentially fast in the number of Newton--Schulz steps. Compared to our paper, this is narrower but sharper: their analysis is explicitly tied to Newton--Schulz and yields a direct dependence on the inner-iteration count $q$, while our framework covers multiple approximation schemes (e.g., Newton--Schulz and \approxalg{PolarExpress}), allows infeasible approximate directions, and yields generic $\delta$-dependent guidance for learning rate and momentum. In this sense, \citet{kim2026muonns} refine the Newton--Schulz-specific part of the story that our broader inexact-LMO model abstracts.

\section{Proofs}
\label{beyond:sec:proofs}

\subsection{Deterministic case}

\subsubsection{Proof of theorem \ref{beyond:thm:general_deterministic} (general result)}

The proof establishes a general convergence guarantee for the deterministic inexact LMO method. It proceeds by deriving a per-iteration descent inequality and then summing it over all iterations to obtain a global bound.

\begin{proof}

\textbf{Per-Iteration Descent Inequality.}
We begin with the standard $L$-smoothness inequality for $f$, as stated in Assumption~\ref{beyond:assump:smoothness}:
\begin{align*}
    f(x^{k+1}) \le f(x^k) + \inp{\nabla f(x^k)}{x^{k+1}-x^k} + \frac{L}{2}\|x^{k+1}-x^k\|^2.
\end{align*}
Let $g^k \eqdef \nabla f(x^k)$. We substitute the update rule from Equation~\eqref{beyond:eq:deterministic_method}, $x^{k+1}-x^k = \gamma^k \hat{d}^k$, into the inequality:
\begin{align} \label{beyond:eq:appendix_smoothness_substituted}
    f(x^{k+1}) \le f(x^k) + \gamma^k \inp{g^k}{\hat{d}^k} + \frac{L}{2} \|\gamma^k \hat{d}^k\|^2.
\end{align}
We now bound the two rightmost terms involving the inexact direction $\hat{d}^k$.

\textbf{Bounding the Squared Norm Term.}
Let $d^k$ be the exact LMO solution for the gradient $g^k$. We use the triangle inequality to bound the norm of the inexact direction $\|\hat{d}^k\|$:
\begin{align*}
    \|\hat{d}^k\| = \|\hat{d}^k - d^k + d^k\| \le \|\hat{d}^k - d^k\| + \|d^k\|.
\end{align*}
Using the inexactness from Assumption~\ref{beyond:assump:inexact_lmo_main} ($\|\hat{d}^k - d^k\| \le \err_k$) and the property of the exact LMO solution ($\|d^k\| \le 1$), we obtain $\|\hat{d}^k\| \le \err_k + 1$. Therefore, the squared norm term is bounded as:
\begin{align*}
    \|\gamma^k \hat{d}^k\|^2 = (\gamma^k)^2 \|\hat{d}^k\|^2 \le (\gamma^k)^2 (1+\err_k)^2.
\end{align*}

\textbf{Bounding the Inner Product Term.}
We decompose the inner product using the exact direction $d^k$:
\begin{align*}
    \inp{g^k}{\hat{d}^k} = \inp{g^k}{d^k + (\hat{d}^k - d^k)} = \inp{g^k}{d^k} + \inp{g^k}{\hat{d}^k - d^k}.
\end{align*}
By the definition of the LMO and the dual norm, the first term is exactly $- \|g^k\|_{\star}$. The second term is bounded using the definition of dual norm:
\begin{align*}
    \inp{g^k}{\hat{d}^k - d^k} \le \left|\inp{g^k}{\hat{d}^k - d^k}\right| \le \|g^k\|_{\star} \|\hat{d}^k - d^k\| \le \|g^k\|_{\star} \err_k.
\end{align*}
Combining these results gives an upper bound on the inner product:
\begin{align*}
    \inp{g^k}{\hat{d}^k} \le -\|g^k\|_{\star} + \|g^k\|_{\star} \err_k = -\|g^k\|_{\star}(1-\err_k).
\end{align*}

\textbf{Combining the Bounds.}
Substituting these two bounds back into the smoothness inequality \eqref{beyond:eq:appendix_smoothness_substituted}, we obtain the per-iteration descent guarantee:
\begin{align*}
    f(x^{k+1}) \le f(x^k) - \gamma^k \|g^k\|_{\star}(1-\err_k) + \frac{L}{2} (\gamma^k)^2 (1+\err_k)^2.
\end{align*}

\textbf{Global Convergence Bound.}
For the descent property to hold, we require $1-\err_k > 0$. Rearranging the per-iteration inequality to isolate the stationarity measure yields:
\begin{align*}
    \gamma^k \|\nabla f(x^k)\|_{\star}(1-\err_k) \le f(x^k) - f(x^{k+1}) + \frac{L}{2} (\gamma^k)^2 (1+\err_k)^2.
\end{align*}
Summing this inequality from $k=0$ to $K-1$:
\begin{align*}
    \sum_{k=0}^{K-1} \gamma^k \|\nabla f(x^k)\|_{\star}(1-\err_k) &\le \sum_{k=0}^{K-1} \left( f(x^k) - f(x^{k+1}) \right) + \frac{L}{2} \sum_{k=0}^{K-1} (\gamma^k)^2 (1+\err_k)^2 \\
    &= f(x^0) - f(x^K) + \frac{L}{2} \sum_{k=0}^{K-1} (\gamma^k)^2 (1+\err_k)^2.
\end{align*}
Using the fact that $f(x^K) \ge f^*$ and letting $\Delta^0 = f(x^0) - f^*$, we have $f(x^0) - f(x^K) \le \Delta^0$. This gives the general bound
\begin{align*}
    \sum_{k=0}^{K-1} \gamma^k \|\nabla f(x^k)\|_{\star}(1-\err_k) \le \Delta^0 + \frac{L}{2} \sum_{k=0}^{K-1} (\gamma^k)^2 (1+\err_k)^2.
\end{align*}
To bound the minimum gradient norm, we note that for any non-negative sequence $\{a_k\}$ and positive sequence $\{w_k\}$, we have
\[
\min_{0 \le j < K} a_j \cdot \sum_{k=0}^{K-1} w_k \le \sum_{k=0}^{K-1} w_k a_k.
\]
Applying this with $a_k = \|\nabla f(x^k)\|_{\star}$ and $w_k = \gamma^k(1-\err_k)$, we get:
\begin{align*}
    \left( \min_{0 \le j < K} \|\nabla f(x^j)\|_{\star} \right) \sum_{k=0}^{K-1} \gamma^k (1-\err_k) \le \sum_{k=0}^{K-1} \gamma^k \|\nabla f(x^k)\|_{\star}(1-\err_k).
\end{align*}
Combining the inequalities and dividing by the sum $\sum \gamma^k(1-\err_k)$ yields the final result stated in Theorem~\ref{beyond:thm:general_deterministic}.
\end{proof}

\subsubsection{Proof of corollary \ref{beyond:cor:constant_deterministic} (constant parameters)}
\begin{proof}
The proof follows by simplifying the general bound from Theorem~\ref{beyond:thm:general_deterministic} under the specific assumptions of the corollary. We are given that the step size is constant, $\gamma^k = \gamma > 0$, and the LMO error is constant, $\err_k = \err < 1$, for all $k=0, \dots, K-1$.

We start with the summed inequality from the proof of Theorem 1:
\begin{align*}
    \sum_{k=0}^{K-1} \gamma (1-\err) \|\nabla f(x^k)\|_{\star} \le \Delta^0 + \frac{L}{2} \sum_{k=0}^{K-1} \gamma^2 (1+\err)^2.
\end{align*}
The constants can be factored out of the summations:
\begin{align*}
    \gamma (1-\err) \sum_{k=0}^{K-1} \|\nabla f(x^k)\|_{\star} \le \Delta^0 + \frac{L K \gamma^2 (1+\err)^2}{2}.
\end{align*}
To obtain a bound on the average gradient norm, we divide the entire inequality by $K\gamma(1-\err)$:
\begin{align*}
\frac{1}{K} \sum_{k=0}^{K-1} \|\nabla f(x^k)\|_{\star} \le \frac{\Delta^0}{K\gamma(1-\err)} + \frac{L \gamma (1+\err)^2}{2(1-\err)},
\end{align*}
which gives the final result stated in Corollary~\ref{beyond:cor:constant_deterministic}.
\end{proof}

\subsubsection{Proof of corollary \ref{beyond:cor:optimal_deterministic} (optimal parameters)}
\begin{proof}
The goal is to find the constant step size $\gamma$ that minimizes the upper bound on the average gradient norm derived in Corollary~\ref{beyond:cor:constant_deterministic}. Let the upper bound be denoted by the function $\mathcal{E}(\gamma)$:
\begin{align*}
    \mathcal{E}(\gamma) = \frac{\Delta^0}{K\gamma(1-\err)} + \frac{L \gamma (1+\err)^2}{2(1-\err)}.
\end{align*}
This expression is of the form $A/\gamma + B\gamma$, where $A = \frac{\Delta^0}{K(1-\err)}$ and $B = \frac{L(1+\err)^2}{2(1-\err)}$ are positive constants. This function is convex for $\gamma > 0$. To find the minimizer $\gamma^*$, we take the derivative with respect to $\gamma$ and set it to zero:
\begin{align*}
    \frac{\partial \mathcal{E}}{\partial \gamma} = -\frac{A}{\gamma^2} + B = 0 \implies (\gamma^*)^2 = \frac{A}{B}.
\end{align*}
Substituting the expressions for $A$ and $B$, we obtain
\begin{align*}
    (\gamma^*)^2 = \frac{\frac{\Delta^0}{K(1-\err)}}{\frac{L(1+\err)^2}{2(1-\err)}} = \frac{\Delta^0}{K(1-\err)} \cdot \frac{2(1-\err)}{L(1+\err)^2} = \frac{2\Delta^0}{K L (1+\err)^2}.
\end{align*}
Taking the square root gives the optimal constant step size:
\begin{align*}
    \gamma^* = \sqrt{\frac{2\Delta^0}{K L (1+\err)^2}} = \frac{1}{1+\err} \sqrt{\frac{2\Delta^0}{K L}}.
\end{align*}
To find the best achievable convergence rate, we substitute this optimal step size $\gamma^*$ back into the bound $\mathcal{E}(\gamma)$. At the minimum, the two terms $A/\gamma^*$ and $B\gamma^*$ are equal, so the total error is $2\sqrt{AB}$:
\begin{align*}
    \mathcal{E}(\gamma^*) = 2\sqrt{AB} &= 2 \sqrt{\frac{\Delta^0}{K(1-\err)} \cdot \frac{L(1+\err)^2}{2(1-\err)}} \\
    &= 2 \frac{\sqrt{\Delta^0 L} (1+\err)}{\sqrt{2K}(1-\err)} \\
    &= \frac{\sqrt{2\Delta^0 L}}{ \sqrt{K}} \frac{1+\err}{1-\err}.
\end{align*}
This confirms the final optimized convergence rate stated in Corollary~\ref{beyond:cor:optimal_deterministic}.
\end{proof}

\begin{corollary}[Iteration Complexity with Constant Step Size]
Under the conditions of Corollary~\ref{beyond:cor:constant_deterministic}, to guarantee that the average stationarity measure satisfies $\frac{1}{K}\sum_{k=0}^{K-1} \|\nabla f(x^k)\|_{\star} \le \varepsilon$ for a target precision $\varepsilon > 0$, it is sufficient to run the algorithm for a number of iterations $K$ of the order of
\begin{align*}
    \mathcal{O}\left( \frac{L\Delta^0}{\varepsilon^2} \cdot \frac{(1+\err)^2}{(1-\err)^2} \right).
\end{align*}
\end{corollary}
\begin{proof}
The result is obtained by setting the optimized bound from Corollary~\ref{beyond:cor:optimal_deterministic}, $\frac{1+\err}{1-\err} \sqrt{\frac{2\Delta^0 L}{K}}$, to be less than or equal to $\varepsilon$ and solving for $K$.
\end{proof}

\subsection{Analysis with an adaptive step size}
\label{beyond:app:adaptive_step_size}

In the main text, we focus on the analysis with constant or pre-defined diminishing step sizes. Here, we present a complementary result for the deterministic case that considers an adaptive step size, chosen at each iteration to maximize the guaranteed descent. This analysis further highlights the impact of the inexactness level $\err_k$.

\begin{theorem}
\label{beyond:thm:optimal_rate_time_varying}
    Let Assumption~\ref{beyond:assump:smoothness} hold and let the LMO errors be a sequence $\{\err_k\}$ with $\err_k \in [0, 1)$. If the algorithm is run with the time-varying adaptive step size
    \begin{align*}
        \gamma^k = \frac{\norm{\nabla f(x^k)}_{\star}(1-\err_k)}{L(1+\err_k)^2},
    \end{align*}
    then the minimum gradient norm after $K$ iterations is bounded by:
    \begin{align}
    \label{beyond:eq:rate_adaptive_time_varying}
        \min_{0 \leq j < K} \ \norm{\nabla f(x^j)}^2_{\star} \leq \frac{2 L \Delta^0}{\sum_{k=0}^{K-1} \frac{(1-\err_k)^2}{(1+\err_k)^2}}.
    \end{align}
\end{theorem}
\begin{proof}
    The proof starts from the per-iteration descent inequality derived in the proof of Theorem~\ref{beyond:thm:general_deterministic}:
    \begin{align*}
    f(x^{k+1}) \le f(x^k) - \gamma^k \norm{\nabla f(x^k)}_{\star}(1-\err_k) + \frac{L (\gamma^k)^2 (1+\err_k)^2}{2}.
    \end{align*}
    The adaptive step size $\gamma^k$ is chosen to minimize the right-hand side of this inequality, which is a quadratic in $\gamma^k$. Its minimizer is precisely the step size given in the theorem statement. Substituting this optimal choice of $\gamma^k$ back into the inequality yields
    \begin{align*}
    f(x^{k+1}) &\leq f(x^k) - \frac{\norm{\nabla f(x^k)}_{\star}^2(1-\err_k)^2}{L(1+\err_k)^2} + \frac{L}{2} \left( \frac{\norm{\nabla f(x^k)}_{\star}(1-\err_k)}{L(1+\err_k)^2} \right)^2 (1+\err_k)^2 \\
    &= f(x^k) - \frac{\norm{\nabla f(x^k)}_{\star}^2(1-\err_k)^2}{2L(1+\err_k)^2}.
    \end{align*}
    Rearranging gives a lower bound on the progress at step $k$:
    \begin{align*}
    \frac{(1-\err_k)^2}{(1+\err_k)^2} \norm{\nabla f(x^k)}_{\star}^2 \leq 2L \left( f(x^k) - f(x^{k+1}) \right).
    \end{align*}
    Summing from $k=0$ to $K-1$, we get
    \begin{align*}
    \sum_{k=0}^{K-1} \frac{(1-\err_k)^2}{(1+\err_k)^2} \norm{\nabla f(x^k)}_{\star}^2 \leq 2L \sum_{k=0}^{K-1} \left( f(x^k) - f(x^{k+1}) \right) \leq 2L \Delta^0.
    \end{align*}
    Let $w_k = \frac{(1-\err_k)^2}{(1+\err_k)^2}$. We have $\sum_{k=0}^{K-1} w_k \norm{\nabla f(x^k)}_{\star}^2 \leq 2L\Delta^0$. Since $\norm{\nabla f(x^j)}_{\star}^2 \geq \min_{0 \leq i < K} \norm{\nabla f(x^i)}_{\star}^2$ for any $j$, we have
    \begin{align*}
    \left( \min_{0 \leq i < K} \ \norm{\nabla f(x^i)}_{\star}^2 \right) \sum_{k=0}^{K-1} w_k \leq \sum_{k=0}^{K-1} w_k \norm{\nabla f(x^k)}_{\star}^2 \leq 2L \Delta^0.
    \end{align*}
    Solving for the minimum squared gradient norm gives the desired inequality:
    \[
        \min_{0 \leq j < K} \ \norm{\nabla f(x^j)}_{\star}^2 \leq \frac{2 L \Delta^0}{\sum_{k=0}^{K-1} \frac{(1-\err_k)^2}{(1+\err_k)^2}}. \qedhere
    \]
\end{proof}

Theorem~\ref{beyond:thm:optimal_rate_time_varying} provides a general expression for the convergence rate, which is governed by the growth of the sum $\sum_{k=0}^{K-1} \frac{(1-\err_k)^2}{(1+\err_k)^2}$ in the denominator of the bound. This allows us to analyze the impact of specific schedules for the inexactness level $\{\err_k\}$. The following corollary establishes a simple and practical sufficient condition for achieving the optimal rate.

\begin{corollary}
\label{beyond:cor:rate_conditions}
    Under the assumptions of Theorem~\ref{beyond:thm:optimal_rate_time_varying}, the optimal convergence rate of $\mathcal{O}(1/\sqrt{K})$ for the minimum gradient norm is achieved if the error sequence $\{\err_k\}$ is uniformly bounded away from 1. Specifically, if there exists a $\err_{\max} = \err \in [0, 1)$ such that $\err_k \le \err_{\max}$ for all $k$, then
    \begin{align}
    \label{beyond:eq:rate_bounded_delta}
        \min_{0 \le j < K} \norm{\nabla f(x^j)}_{\star} \le \frac{1+\err}{1-\err} \sqrt{\frac{2 L \Delta^0}{K}}.
    \end{align}
\end{corollary}
\begin{proof}
    If $\err_k \le \err$, then the term $w_k \eqdef \frac{(1-\err_k)^2}{(1+\err_k)^2}$ is bounded below by a positive constant:
    \begin{align*}
        w_k \ge \frac{(1-\err)^2}{(1+\err)^2} > 0.
    \end{align*}
    Therefore, the sum in the denominator of \eqref{beyond:eq:rate_adaptive_time_varying} grows at least linearly with $K$:
    \begin{align*}
        \sum_{k=0}^{K-1} w_k \ge \sum_{k=0}^{K-1} \frac{(1-\err)^2}{(1+\err)^2} = K \frac{(1-\err)^2}{(1+\err)^2}.
    \end{align*}
    Substituting this lower bound into the inequality \eqref{beyond:eq:rate_adaptive_time_varying} and taking the square root of both sides yields the desired result.
\end{proof}

\begin{corollary}[Iteration Complexity with Adaptive Step Size]
Under the conditions of Corollary~\ref{beyond:cor:rate_conditions}, to guarantee that the minimum gradient norm satisfies $\min_{0 \le j < K} \|\nabla f(x^j)\|_{\star} \le \varepsilon$ for a target precision $\varepsilon > 0$, it is sufficient to run the algorithm for a number of iterations $K$ of the order of
\begin{align*}
    \mathcal{O}\left( \frac{L\Delta^0}{\varepsilon^2} \cdot \frac{(1+\err)^2}{(1-\err)^2} \right).
\end{align*}
\end{corollary}
\begin{proof}
The result is obtained by setting the bound from Corollary~\ref{beyond:cor:rate_conditions} to be less than or equal to $\varepsilon$ and solving for $K$.
\end{proof}

\subsection{Stochastic case}
\label{beyond:app:stochastic_proofs}

This section provides the full proofs for the main theoretical results in the stochastic setting. The analysis proceeds in three main stages. First, in Lemma~\ref{beyond:lemma:inexact_descent_appendix}, we establish a crucial per-iteration descent guarantee that holds deterministically for any single step of the algorithm. Second, in Lemma~\ref{beyond:lemma:inexact_momentum_error_main}, we derive a bound on the expected error of the momentum term. Finally, in the proof of Theorem~\ref{beyond:thm:stochastic}, we combine these results to derive the global convergence guarantee.

\begin{lemma}[Inexact Descent Lemma]
\label{beyond:lemma:inexact_descent_appendix}
Let Assumption~\ref{beyond:assump:smoothness} hold. For any iteration $k$ of Algorithm~\ref{beyond:alg:stochastic_inaxact_lmo}, the following inequality holds:
\begin{align*}
    f(x^{k+1})
    &\le f(x^k) - \gamma^k \|m^{k+1}\|_{\star}(1-\err_k) \\
    &\quad + \gamma^k(1+\err_k)\|\nabla f(x^{k+1}) - m^{k+1}\|_{\star}
    + \frac{3L}{2}(\gamma^k)^2(1+\err_k)^2.
\end{align*}
\end{lemma}
\begin{proof}
The proof establishes a bound on the progress made in a single step. We start with the $L$-smoothness inequality of $f$ from Assumption~\ref{beyond:assump:smoothness}:
\begin{align*}
    f(x^{k+1}) \le f(x^k) + \inp{\nabla f(x^k)}{x^{k+1}-x^k} + \frac{L}{2}\|x^{k+1}-x^k\|^2.
\end{align*}
Substituting the update rule $x^{k+1}-x^k = \gamma^k \hat{d}^k$ gives
\begin{align} \label{beyond:eq:app_inexact_lemma_start}
    f(x^{k+1}) \le f(x^k) + \gamma^k \inp{\nabla f(x^k)}{\hat{d}^k} + \frac{L}{2}\|\gamma^k \hat{d}^k\|^2.
\end{align}
We decompose the inner product term by introducing the momentum vector $m^{k+1}$:
\begin{align*}
    \gamma^k \inp{\nabla f(x^k)}{\hat{d}^k} = \gamma^k \inp{m^{k+1}}{\hat{d}^k} + \gamma^k \inp{\nabla f(x^k) - m^{k+1}}{\hat{d}^k}.
\end{align*}
The proof proceeds by bounding the key terms involving the inexact direction $\hat{d}^k$.

\textbf{1. Bounding the norm of the inexact direction.}
Let $d^k \eqdef \operatorname{argmin}_{\|d\| \le 1} \inp{m^{k+1}}{d}$ be the exact LMO solution. We use the triangle inequality to bound $\|\hat{d}^k\|$:
\begin{align*}
    \|\hat{d}^k\| = \|\hat{d}^k - d^k + d^k\| \le \|\hat{d}^k - d^k\| + \|d^k\|.
\end{align*}
Using Assumption~\ref{beyond:assump:inexact_lmo_main} ($\|\hat{d}^k - d^k\| \le \err_k$) and $\|d^k\| \le 1$, we have $\|\hat{d}^k\| \le \err_k + 1$. This provides a bound for the quadratic term from smoothness:
\begin{align*}
    \|\gamma^k \hat{d}^k\|^2 = (\gamma^k)^2 \|\hat{d}^k\|^2 \le (\gamma^k)^2 (1+\err_k)^2.
\end{align*}

\textbf{2. Bounding the LMO inner product term.}
We explicitly account for the LMO error $\err_k$:
\begin{align*}
    \inp{m^{k+1}}{\hat{d}^k} &= \inp{m^{k+1}}{d^k} + \inp{m^{k+1}}{\hat{d}^k - d^k} \\
    &\le -\|m^{k+1}\|_{\star} + |\inp{m^{k+1}}{\hat{d}^k - d^k}| \\
    &\le -\|m^{k+1}\|_{\star} + \|m^{k+1}\|_{\star} \|\hat{d}^k - d^k\| \quad \text{(by definition of dual norm)} \\
    &\le -\|m^{k+1}\|_{\star}(1-\err_k).
\end{align*}

\textbf{3. Bounding the momentum error term.}
We introduce $\nabla f(x^{k+1})$ to align the error term with the structure needed for subsequent analysis:
\begin{align*}
    \inp{\nabla f(x^k) - m^{k+1}}{\hat{d}^k} &= \inp{\nabla f(x^k) - \nabla f(x^{k+1})}{\hat{d}^k} + \inp{\nabla f(x^{k+1}) - m^{k+1}}{\hat{d}^k} \\
    &\le \|\nabla f(x^k) - \nabla f(x^{k+1})\|_{\star} \|\hat{d}^k\| + \|\nabla f(x^{k+1}) - m^{k+1}\|_{\star} \|\hat{d}^k\|.
\end{align*}
By Assumption~\ref{beyond:assump:smoothness}, we have $\|\nabla f(x^k) - \nabla f(x^{k+1})\|_{\star} \le L\|x^k - x^{k+1}\| = L\gamma^k\|\hat{d}^k\|$. Substituting this in yields
\begin{align*}
    \inp{\nabla f(x^k) - m^{k+1}}{\hat{d}^k} &\le (L\gamma^k\|\hat{d}^k\|)\|\hat{d}^k\| + \|\hat{d}^k\| \cdot \|\nabla f(x^{k+1}) - m^{k+1}\|_{\star} \\
    &= L\gamma^k\|\hat{d}^k\|^2 + \|\hat{d}^k\| \cdot \|\nabla f(x^{k+1}) - m^{k+1}\|_{\star}.
\end{align*}
Using our bound $\|\hat{d}^k\| \le 1+\err_k$, this becomes
\begin{align*}
    \inp{\nabla f(x^k) - m^{k+1}}{\hat{d}^k} \le L\gamma^k(1+\err_k)^2 + (1+\err_k)\|\nabla f(x^{k+1}) - m^{k+1}\|_{\star}.
\end{align*}

\textbf{4. Assembling the Final Inequality.}
We now substitute all the bounded components back into inequality \eqref{beyond:eq:app_inexact_lemma_start}:
\begin{align*}
f(x^{k+1}) &\le f(x^k) + \gamma^k \left( -\|m^{k+1}\|_{\star}(1-\err_k) \right) \\
&\quad + \gamma^k \left( L\gamma^k(1+\err_k)^2 + (1+\err_k)\|\nabla f(x^{k+1}) - m^{k+1}\|_{\star} \right) \\
&\quad + \frac{L}{2}(\gamma^k)^2(1+\err_k)^2.
\end{align*}
Finally, collecting the terms proportional to $L(\gamma^k)^2(1+\err_k)^2$ gives the result stated in the lemma.
\end{proof}

\subsubsection{Proof of lemma \ref{beyond:lemma:inexact_momentum_error_main}}
\begin{proof}
The proof bounds the expected deviation of the momentum vector $m^{k+1}$ from the true gradient $\nabla f(x^k)$. We remind the reader that the update rule for the momentum is $m^{k+1} = (1-\alpha_k)m^k + \alpha_k g^k$. We start by expanding the error term recursively, obtaining
\begin{align*}
    m^{k+1} - \nabla f(x^k) &= (1-\alpha_k)m^k + \alpha_k g^k - \nabla f(x^k) \\
    &= (1-\alpha_k)(m^k - \nabla f(x^{k-1})) + \alpha_k(g^k - \nabla f(x^k))\\
    &\quad + (1-\alpha_k)(\nabla f(x^{k-1}) - \nabla f(x^k)).
\end{align*}
This decomposition separates the error into three distinct components: (1) the decayed error from the previous step, (2) the new stochastic noise, and (3) the gradient drift. By unrolling this recursion from $k$ down to $0$, we can express the total error as a sum of its historical components:
\begin{align*}
    m^{k+1} - \nabla f(x^k)
    &= \left(\prod_{i=0}^k (1-\alpha_i)\right)(m^0 - \nabla f(x^0))\\
    &\quad+ \sum_{i=0}^{k} \left(\prod_{j=i+1}^k (1-\alpha_j)\right) \alpha_i(g^i - \nabla f(x^i)) \\
    &+ \sum_{i=1}^{k} \left(\prod_{j=i}^k (1-\alpha_j)\right)(\nabla f(x^{i-1}) - \nabla f(x^i)).
\end{align*}
We now take the expectation of the norm and apply the triangle inequality, to get
\begin{align*}
    \mathbb{E}[\|m^{k+1} - \nabla f(x^k)\|_{\star}] &\le \left(\prod_{i=0}^k (1-\alpha_i)\right)\mathbb{E}[\|m^0 - \nabla f(x^0)\|_{\star}] \\
    &\quad + \mathbb{E}\left[\left\| \sum_{i=0}^{k} \left(\prod_{j=i+1}^k (1-\alpha_j)\right) \alpha_i(g^i - \nabla f(x^i)) \right\|_{\star}\right] \\
    &\quad + \sum_{i=1}^{k} \left(\prod_{j=i}^k (1-\alpha_j)\right)\mathbb{E}[\|\nabla f(x^{i-1}) - \nabla f(x^i)\|_{\star}].
\end{align*}
To obtain the result in the statement of the lemma, we analyze the bound for constant parameters $\alpha, \gamma, \err$.

\textbf{1. Bounding the Initial Error.}
With the standard initialization $m^0 = g^0$, and using the tower property of expectation along with Assumption~\ref{beyond:assump:stochastic_oracle} and the norm compatibility condition, we have
\begin{align*}
    \mathbb{E}[\|m^0 - \nabla f(x^0)\|_{\star}] = \mathbb{E}[\|g^0 - \nabla f(x^0)\|_{\star}] \le \rho\,\mathbb{E}[\|g^0 - \nabla f(x^0)\|_2].
\end{align*}
By Jensen's inequality, $\mathbb{E}[X] \le \sqrt{\mathbb{E}[X^2]}$, this is further bounded by
\begin{align*}
    \rho\sqrt{\mathbb{E}[\|g^0 - \nabla f(x^0)\|_2^2]} \le \rho\sigma.
\end{align*}

\textbf{2. Bounding the Accumulated Drift.}
This is the term modified by our inexact LMO. Using Assumption~\ref{beyond:assump:smoothness} ($L$-smoothness) and the fact that $\|x^{i-1}-x^i\| = \gamma\|\hat{d}^{i-1}\| \le \gamma(1+\err)$, we can bound the sum as
\begin{align*}
    \sum_{i=1}^{k} (1-\alpha)^{k-i+1} \mathbb{E}[\|\nabla f(x^{i-1}) - \nabla f(x^i)\|_{\star}] &\le \sum_{i=1}^{k} (1-\alpha)^{k-i+1} L\gamma(1+\err) \\
    &= L\gamma(1+\err) \sum_{j=0}^{k-1} (1-\alpha)^{j+1},
\end{align*}
where we changed the summation index to $j = i-1$. This is a geometric series which we can bound by its infinite sum:
\begin{align*}
    L\gamma(1+\err) \sum_{j=0}^{\infty} (1-\alpha)^{j+1} = L\gamma(1+\err) \frac{1-\alpha}{1-(1-\alpha)} \le \frac{L\gamma(1+\err)}{\alpha}.
\end{align*}

\textbf{3. Bounding the Accumulated Noise.}
Let $V_k = \sum_{i=0}^{k} \alpha(1-\alpha)^{k-i}(g^i - \nabla f(x^i))$. The random vectors $(g^i - \nabla f(x^i))$ are zero-mean and independent conditioned on the past. Using the norm compatibility followed by Jensen's inequality, we have
\begin{align*}
    \mathbb{E}[\|V_k\|_{\star}] \le \rho\,\mathbb{E}[\|V_k\|_2] \le \rho \sqrt{\mathbb{E}[\|V_k\|_2^2]}.
\end{align*}
Since the random vectors are independent and zero-mean, the cross-terms vanish in expectation. So, the expected squared norm of the sum is the sum of the expected squared norms:
\begin{align*}
    \mathbb{E}[\|V_k\|_2^2] &= \sum_{i=0}^{k} \alpha^2(1-\alpha)^{2(k-i)} \mathbb{E}[\|g^i - \nabla f(x^i)\|_2^2] \\
    &\le \sum_{i=0}^{k} \alpha^2(1-\alpha)^{2(k-i)} \sigma^2 = \alpha^2\sigma^2 \sum_{j=0}^{k} ((1-\alpha)^2)^j.
\end{align*}
We bound this geometric series by its infinite sum:
\begin{align*}
    \sum_{j=0}^{k} ((1-\alpha)^2)^j \le \sum_{j=0}^{\infty} ((1-\alpha)^2)^j = \frac{1}{1-(1-\alpha)^2} = \frac{1}{2\alpha - \alpha^2} = \frac{1}{\alpha(2-\alpha)}.
\end{align*}
Substituting this back, we get the bound for the accumulated noise:
\begin{align*}
    \mathbb{E}[\|V_k\|_{\star}] \le \rho \sqrt{\alpha^2\sigma^2 \frac{1}{\alpha(2-\alpha)}} = \rho \sqrt{\frac{\alpha\sigma^2}{2-\alpha}} = \frac{\rho\sigma\sqrt{\alpha}}{\sqrt{2-\alpha}}.
\end{align*}

\textbf{4. Combining the Bounds.}
Combining the bounds for the three components gives the final expression. For the simplified steady-state bound (which holds for any $k$), we combine the infinite-sum bounds for the drift and noise terms with the initial error term, which decays exponentially over time due to the $(1-\alpha)^{k+1}$ factor:
\begin{align*}
    \mathbb{E} [\|m^{k+1} - \nabla f(x^k)\|_{\star}] \le (1-\alpha)^{k+1}\rho\sigma + \frac{\rho\sigma\sqrt{\alpha}}{\sqrt{2-\alpha}} + \frac{L\gamma(1+\err)}{\alpha}.
\end{align*}
This is the result stated for constant parameters in Lemma~\ref{beyond:lemma:inexact_momentum_error_main}.
\end{proof}

\subsubsection{Proof of theorem \ref{beyond:thm:stochastic}}
\begin{proof}
The proof derives the final convergence guarantee by combining the per-iteration progress from the \say{Inexact Descent Lemma} (Lemma~\ref{beyond:lemma:inexact_descent_appendix}) with the bound on the momentum error from Lemma~\ref{beyond:lemma:inexact_momentum_error_main}.

We begin with the per-iteration guarantee from Lemma~\ref{beyond:lemma:inexact_descent_appendix}, specialized for constant parameters $\gamma, \alpha, \err$:
\begin{align*}
    f(x^{k+1})
    &\le f(x^k) - \gamma \|m^{k+1}\|_{\star}(1-\err) \\
    &\quad + \gamma(1+\err)\|\nabla f(x^{k+1}) - m^{k+1}\|_{\star}
    + \frac{3L}{2}\gamma^2(1+\err)^2.
\end{align*}
To obtain a guarantee on the true gradient norm, $\|\nabla f(x^{k+1})\|_{\star}$, we apply the reverse triangle inequality to the main descent term:
\begin{align*}
    \|m^{k+1}\|_{\star}
    &= \|\nabla f(x^{k+1}) - (\nabla f(x^{k+1}) - m^{k+1})\|_{\star} \\
    &\ge \|\nabla f(x^{k+1})\|_{\star}
    - \|\nabla f(x^{k+1}) - m^{k+1}\|_{\star}.
\end{align*}
Substituting this back into the descent inequality (noting that the leading minus sign flips the inequality for the error term) gives
\begin{align*}
    -\gamma \|m^{k+1}\|_{\star}(1-\err)
    &\le -\gamma \left( \|\nabla f(x^{k+1})\|_{\star}
    - \|\nabla f(x^{k+1}) - m^{k+1}\|_{\star} \right)(1-\err).
\end{align*}
This yields a new per-iteration guarantee directly in terms of the gradient norm:
\begin{align*}
    f(x^{k+1}) &\le f(x^k) - \gamma\|\nabla f(x^{k+1})\|_{\star}(1-\err) + \gamma(1-\err)\|\nabla f(x^{k+1}) - m^{k+1}\|_{\star} \\
    &\quad + \gamma(1+\err)\|\nabla f(x^{k+1}) - m^{k+1}\|_{\star} + \frac{3L}{2}\gamma^2(1+\err)^2.
\end{align*}
The coefficients for the two momentum error terms sum to $(1-\err) + (1+\err) = 2$. Combining them yields a cleaner expression:
\begin{align*}
    f(x^{k+1})
    &\le f(x^k) - \gamma\|\nabla f(x^{k+1})\|_{\star}(1-\err) \\
    &\quad + 2\gamma\|\nabla f(x^{k+1}) - m^{k+1}\|_{\star}
    + \frac{3L}{2}\gamma^2(1+\err)^2.
\end{align*}
Rearranging to isolate the gradient norm, we get
\begin{align*}
    \gamma(1-\err)\|\nabla f(x^{k+1})\|_{\star}
    &\le f(x^k) - f(x^{k+1})
    + 2\gamma\|\nabla f(x^{k+1}) - m^{k+1}\|_{\star}
    + \frac{3L}{2}\gamma^2(1+\err)^2.
\end{align*}
We now sum this inequality from $k=0$ to $K-1$, take the total expectation, and perform an index shift on the summation of the gradient norm ($j=k+1$), obtaining
\begin{align*}
    \gamma(1-\err)\sum_{j=1}^{K}\mathbb{E}[\|\nabla f(x^j)\|_{\star}] &\le \mathbb{E}\left[\sum_{k=0}^{K-1} (f(x^k) - f(x^{k+1}))\right]
     + 2\gamma \sum_{k=0}^{K-1} \mathbb{E}[\|\nabla f(x^{k+1}) - m^{k+1}\|_{\star}] \\
     &\quad+ K \frac{3L}{2}\gamma^2(1+\err)^2.
\end{align*}
The first term on the right-hand side is a telescoping sum bounded by $\Delta^0 = f(x^0) - f^*$.

\textbf{Bounding the Sum of Momentum Errors.}
The core of the proof is to bound the sum of momentum errors. We use the triangle inequality to bridge the index gap:
\begin{align*}
    \mathbb{E}[\|\nabla f(x^{k+1}) - m^{k+1}\|_{\star}]
    &\le \mathbb{E}[\|\nabla f(x^{k+1}) - \nabla f(x^k)\|_{\star}] \\
    &\quad + \mathbb{E}[\|\nabla f(x^k) - m^{k+1}\|_{\star}].
\end{align*}
The first term (gradient drift) is bounded by $L\gamma(1+\err)$. The second term is bounded by Lemma~\ref{beyond:lemma:inexact_momentum_error_main}. We use the slightly looser but simpler bound $\frac{\rho\sigma\sqrt{\alpha}}{\sqrt{2-\alpha}} \le \rho\sigma\sqrt{\alpha}$ for $\alpha \in (0,1)$. The full error is thus bounded by
\begin{align*}
    \mathbb{E}[\|\nabla f(x^{k+1}) - m^{k+1}\|_{\star}]
    &\le L\gamma(1+\err) + (1-\alpha)^{k+1}\rho\sigma \\
    &\quad + \rho\sigma\sqrt{\alpha} + \frac{L\gamma(1+\err)}{\alpha}.
\end{align*}
We now substitute this bound back into the main summation:
\begin{align*}
    \gamma(1-\err)\sum_{j=1}^{K}\mathbb{E}[\|\nabla f(x^j)\|_{\star}]
    &\le \Delta^0 + K \frac{3L\gamma^2(1+\err)^2}{2} \\
    &\quad + 2\gamma \sum_{k=0}^{K-1}
    \left( L\gamma(1+\err) + (1-\alpha)^{k+1}\rho\sigma \right.\\
    &\qquad\qquad\left. + \rho\sigma\sqrt{\alpha} + \frac{L\gamma(1+\err)}{\alpha} \right).
\end{align*}
Evaluating the summation over $k$ for each component gives
\begin{align*}
    \gamma(1-\err)\sum_{j=1}^{K}\mathbb{E}[\|\nabla f(x^j)\|_{\star}]
    &\le \Delta^0 + K \frac{3L\gamma^2(1+\err)^2}{2}
    + 2KL\gamma^2(1+\err) \\
     &\quad + 2\gamma\left(\sum_{k=0}^{K-1}(1-\alpha)^{k+1}\rho\sigma\right)\\
     &\quad + 2K\gamma\rho\sigma\sqrt{\alpha} + \frac{2KL\gamma^2(1+\err)}{\alpha}.
\end{align*}
For the geometric series, we have $\sum_{k=0}^{K-1}(1-\alpha)^{k+1} \le \sum_{j=1}^{\infty}(1-\alpha)^{j} = \frac{1-\alpha}{\alpha} \le \frac{1}{\alpha}$. Plugging this in gives
\begin{align*}
    \gamma(1-\err)\sum_{j=1}^{K}\mathbb{E}[\|\nabla f(x^j)\|_{\star}] &\le \Delta^0 + K \frac{3L\gamma^2(1+\err)^2}{2} + 2KL\gamma^2(1+\err) \\
    &\quad + \frac{2\gamma\rho\sigma}{\alpha} + 2K\gamma\rho\sigma\sqrt{\alpha} + \frac{2KL\gamma^2(1+\err)}{\alpha}.
\end{align*}
Finally, dividing by $K\gamma(1-\err)$ yields the average bound for the gradient norm. Grouping the terms by their dependency gives the final stated result in Theorem~\ref{beyond:thm:stochastic}.
\end{proof}

\subsubsection{Proof of corollary \ref{beyond:cor:stochastic_optimal_rate}}

The proof consists of two parts. First, we derive the asymptotically optimal choices for the step size $\gamma$ and momentum parameter $\alpha$ by minimizing the convergence upper bound from Theorem~\ref{beyond:thm:stochastic}. Second, we substitute these optimal parameters back into the bound to obtain the final convergence rate.

\paragraph{1. Optimal Parameter Derivation.}
Our goal is to find $\gamma > 0$ and $\alpha \in (0,1)$ that minimize the expression $\mathcal{E}(\gamma, \alpha)$:
\begin{align*}
    \mathcal{E}(\gamma, \alpha)
    &= \frac{\Delta^0}{K\gamma(1-\err)}
    + \frac{2\rho\sigma}{\alpha K(1-\err)}
    + \frac{2\rho\sigma\sqrt{\alpha}}{1-\err} \\
    &\quad + \frac{L\gamma(1+\err)}{1-\err}
    \left( \frac{3(1+\err)}{2} + 2 \right)
    + \frac{2L\gamma(1+\err)}{\alpha(1-\err)}.
\end{align*}
To simplify the algebraic manipulation, we define the following constants:
\begin{itemize}
    \item $C_1 = \frac{\Delta^0}{K(1-\err)}$,
    \item $C_2 = \frac{2\rho\sigma}{K(1-\err)}$,
    \item $C_3 = \frac{2\rho\sigma}{1-\err}$,
    \item $C_4 = \frac{L(1+\err)}{1-\err}\left(\frac{3(1+\err)}{2} + 2\right)$,
    \item $C_5 = \frac{2L(1+\err)}{1-\err}$.
\end{itemize}
Hence, the expression to minimize is $\mathcal{E}(\gamma, \alpha) = \frac{C_1}{\gamma} + \frac{C_2}{\alpha} + C_3\sqrt{\alpha} + C_4\gamma + \frac{C_5\gamma}{\alpha}$. We find the optimal parameters by setting the partial derivatives with respect to $\gamma$ and $\alpha$ to zero, yielding the system of equations
\begin{align}
    \frac{\partial \mathcal{E}}{\partial \gamma} &= -\frac{C_1}{\gamma^2} + C_4 + \frac{C_5}{\alpha} = 0 \label{beyond:eq:partial_gamma}, \\
    \frac{\partial \mathcal{E}}{\partial \alpha} &= -\frac{C_2}{\alpha^2} + \frac{C_3}{2\sqrt{\alpha}} - \frac{C_5\gamma}{\alpha^2} = 0.\label{beyond:eq:partial_alpha}
\end{align}
An exact closed-form solution is intractable. However, we are interested in the asymptotic behavior as $K \to \infty$, where we expect $\gamma^*, \alpha^* \to 0$. The constants scale as:
$C_1 \propto 1/K$, $C_2 \propto 1/K$, while $C_3, C_4, C_5$ are constants with respect to $K$.

We now solve the system in this asymptotic regime. From \eqref{beyond:eq:partial_gamma}, as $\alpha \to 0$, the term $C_5/\alpha$ dominates the constant $C_4$. Thus, the equation asymptotically behaves as
\begin{align}
    \gamma^2 \approx \frac{C_1 \alpha}{C_5}. \label{beyond:eq:gamma_asymptotic}
\end{align}
From \eqref{beyond:eq:partial_alpha}, since $C_2 \propto 1/K$ and we expect $\gamma$ to decay slower than $1/K$, the term $C_2$ is asymptotically negligible compared to $C_5\gamma$. This gives:
\begin{align}
    \frac{C_3}{2}\alpha^{3/2} \approx C_5\gamma. \label{beyond:eq:alpha_asymptotic}
\end{align}
From \eqref{beyond:eq:alpha_asymptotic}, we express $\gamma$ in terms of $\alpha$: $\gamma \approx \frac{C_3}{2 C_5} \alpha^{3/2}$. Substituting this into \eqref{beyond:eq:gamma_asymptotic}:
\begin{align*}
    \left( \frac{C_3}{2 C_5} \alpha^{3/2} \right)^2 \approx \frac{C_1 \alpha}{C_5} \implies \frac{C_3^2}{4 C_5^2} \alpha^3 \approx \frac{C_1 \alpha}{C_5}.
\end{align*}
Solving for $\alpha^2$ (assuming $\alpha \ne 0$) gives $\alpha^2 \approx \frac{4 C_1 C_5}{C_3^2}$. Now, we substitute back the full definitions of the constants:
\begin{align*}
    (\alpha^*)^2 = \frac{4 \cdot \frac{\Delta^0}{K(1-\err)} \cdot \frac{2L(1+\err)}{1-\err}}{\left(\frac{2\rho\sigma}{1-\err}\right)^2} = \frac{\frac{8\Delta^0 L(1+\err)}{K(1-\err)^2}}{\frac{4(\rho\sigma)^2}{(1-\err)^2}} = \frac{2\Delta^0 L(1+\err)}{K(\rho\sigma)^2}.
\end{align*}
This yields the optimal momentum parameter:
\begin{align*}
    \alpha^* = \frac{\sqrt{2\Delta^0 L(1+\err)}}{\sqrt{K}\rho\sigma}.
\end{align*}
To find the optimal step size $\gamma^*$, we use the relation $\gamma^2 \approx (C_1 \alpha) / C_5$:
\begin{align*}
    (\gamma^*)^2
    &= \frac{\Delta^0}{K(1-\err)} \cdot \left( \frac{\sqrt{2\Delta^0 L(1+\err)}}{\sqrt{K}\rho\sigma} \right) \cdot \frac{1-\err}{2L(1+\err)}
    = \frac{\Delta^0 \sqrt{2\Delta^0 L(1+\err)}}{K^{3/2} \rho\sigma \cdot 2L(1+\err)}\\
    &= \frac{(\Delta^0)^{3/2} \sqrt{2L}\sqrt{1+\err}}{K^{3/2} \rho\sigma \cdot 2L(1+\err)} = \frac{(\Delta^0)^{3/2}}{K^{3/2} \sqrt{2L} \rho\sigma\sqrt{1+\err}}.
\end{align*}
Taking the square root gives the optimal step size
\begin{align*}
    \gamma^* = \frac{(\Delta^0)^{3/4}}{2^{1/4} K^{3/4} L^{1/4} (\rho\sigma)^{1/2} (1+\err)^{1/4}}.
\end{align*}

\paragraph{2. Derivation of the Final Convergence Rate.}
We now substitute the asymptotically optimal parameters, $\gamma^*$ and $\alpha^*$, back into the full five-term convergence bound $\mathcal{E}(\gamma, \alpha)$ to determine the best achievable convergence rate. The bound is the sum of five distinct terms:
\begin{align*}
    \mathcal{E}(\gamma^*, \alpha^*) = \underbrace{\frac{C_1}{\gamma^*}}_{\text{Term 1}} + \underbrace{\frac{C_2}{\alpha^*}}_{\text{Term 2}} + \underbrace{C_3\sqrt{\alpha^*}}_{\text{Term 3}} + \underbrace{C_4\gamma^*}_{\text{Term 4}} + \underbrace{\frac{C_5\gamma^*}{\alpha^*}}_{\text{Term 5}}.
\end{align*}
We compute the explicit value of each term to verify the dominant rate and analyze the contribution of the non-dominant, higher-order terms.

\textbf{Dominant Terms ($\mathcal{O}(1/K^{1/4})$).}
These are the terms that dictate the asymptotic convergence rate.

\textit{Term 1: The Initial Gap Component.}
\begin{align*}
    \text{Term 1} = \frac{C_1}{\gamma^*} &= \frac{\Delta^0}{K(1-\err)} \cdot \left( \gamma^* \right)^{-1} = \frac{\Delta^0}{K(1-\err)} \cdot \left( \frac{2^{1/4} K^{3/4} L^{1/4} (\rho\sigma)^{1/2} (1+\err)^{1/4}}{(\Delta^0)^{3/4}} \right) \\
    &= \frac{2^{1/4} (\Delta^0)^{1/4} L^{1/4} (\rho\sigma)^{1/2} (1+\err)^{1/4}}{K^{1/4}(1-\err)}.
\end{align*}

\textit{Term 3: The Steady-State Variance Component.}
\begin{align*}
    \text{Term 3} = C_3\sqrt{\alpha^*} &= \frac{2\rho\sigma}{1-\err} \cdot \left( \frac{\sqrt{2\Delta^0 L(1+\err)}}{\sqrt{K}\rho\sigma} \right)^{1/2} = \frac{2\rho\sigma}{1-\err} \cdot \frac{(2\Delta^0 L(1+\err))^{1/4}}{K^{1/4}(\rho\sigma)^{1/2}} \\
    &= \frac{2 \cdot 2^{1/4} (\rho\sigma)^{1/2} (L\Delta^0(1+\err))^{1/4}}{K^{1/4}(1-\err)} = 2 \cdot (\text{Term 1}).
\end{align*}

\textit{Term 5: The Drift/Interaction Component.}
\begin{align*}
    \text{Term 5} = \frac{C_5\gamma^*}{\alpha^*} &= \frac{2L(1+\err)}{1-\err} \cdot \frac{\gamma^*}{\alpha^*} = \frac{2L(1+\err)}{1-\err} \cdot \frac{C_3}{2C_5} \sqrt{\alpha^*} = \frac{C_3\sqrt{\alpha^*}}{1-\err} \cdot \frac{L(1+\err)}{C_5} \\
    &= \frac{1}{2} C_3\sqrt{\alpha^*} = \text{Term 1}.
\end{align*}
The asymptotic balancing ensures Term 1, Term 3, and Term 5 are all of the same order, $\mathcal{O}(1/K^{1/4})$.

\textbf{Higher-Order Terms.}
The remaining terms decay faster and are asymptotically negligible.

\textit{Term 2: The Decaying Variance Component ($\mathcal{O}(1/K^{1/2})$).}
\begin{align*}
    \text{Term 2} = \frac{C_2}{\alpha^*} = \frac{2\rho\sigma}{K(1-\err)} \cdot \left( \frac{\sqrt{K}\rho\sigma}{\sqrt{2\Delta^0 L(1+\err)}} \right) = \frac{\sqrt{2}(\rho\sigma)^2}{\sqrt{K}(1-\err)\sqrt{\Delta^0 L(1+\err)}}.
\end{align*}

\textit{Term 4: The Step Size Bias Component ($\mathcal{O}(1/K^{3/4})$).}
\begin{align*}
    \text{Term 4}
    &= C_4\gamma^*
    = \frac{L(1+\err)}{1-\err}\left(\frac{3(1+\err)}{2} + 2\right)
    \cdot \frac{(\Delta^0)^{3/4}}{2^{1/4} K^{3/4} L^{1/4} (\rho\sigma)^{1/2} (1+\err)^{1/4}} \\
    &= \frac{(7+3\err)L^{3/4}(\Delta^0)^{3/4}(1+\err)^{3/4}}{2^{5/4}K^{3/4}(1-\err)(\rho\sigma)^{1/2}}.
\end{align*}

\textbf{The Full Bound.}
Combining all terms, the full optimized convergence bound is
\begin{align*}
    \mathcal{E}(\gamma^*, \alpha^*)
    &= \underbrace{\frac{2^{9/4} (\Delta^0)^{1/4} (\rho\sigma)^{1/2} (L(1+\err))^{1/4}}{K^{1/4}(1-\err)}}_{\text{Dominant Part: } \mathcal{O}(K^{-1/4})} \\
    &\quad +
    \underbrace{\frac{\sqrt{2}(\rho\sigma)^2}{K^{1/2}(1-\err)\sqrt{\Delta^0 L(1+\err)}}}_{\text{Higher-Order: } \mathcal{O}(K^{-1/2})} \\
    &\quad +
    \underbrace{\frac{(7+3\err)L^{3/4}(\Delta^0)^{3/4}(1+\err)^{3/4}}{2^{5/4}K^{3/4}(1-\err)(\rho\sigma)^{1/2}}}_{\text{Higher-Order: } \mathcal{O}(K^{-3/4})}.
\end{align*}
This confirms that the convergence rate is $\mathcal{O}(1/K^{1/4})$ with the precise dependence of the dominant part on all problem parameters, matching the result stated in Corollary~\ref{beyond:cor:stochastic_optimal_rate}.

\subsection{Convergence with time-varying parameters}
\label{beyond:app:time_varying}

The analysis with constant parameters provides crucial insights into the fundamental trade-offs of the algorithm. However, the optimal choices for these constant parameters, $\gamma^*$ and $\alpha^*$, depend on problem-specific constants such as the smoothness $L$ and the initial sub-optimality $\Delta^0$, as well as the total iteration budget $K$. In practice, these quantities are typically unknown, making this ``optimal" tuning infeasible.

To develop a more practical and robust guarantee, we now analyze the algorithm with a pre-defined, \textit{time-varying} schedule for the step size and momentum parameters. This approach is ``parameter-agnostic," meaning the schedules are chosen based only on the iteration counter and do not require prior knowledge of the problem's characteristics. The primary goal of this analysis is to prove that the algorithm minimizes the expected gradient (i.e., $\mathbb{E}[\|\nabla f(x^k)\|_{\star}] \to 0$) for any problem satisfying our general assumptions, without any manual tuning of the learning rates with respect to $L$ or $K$.

The cornerstone of this analysis is a general bound on the expected momentum error that holds for any valid time-varying schedule.

\begin{lemma}[Momentum Error Bound for Time-Varying Parameters]
\label{beyond:lemma:aligned_momentum_error}
Let Assumptions \ref{beyond:assump:smoothness},\ref{beyond:assump:stochastic_oracle}, and the norm compatibility condition hold. Let the sequence $\{x^k\}$ be generated by Algorithm~\ref{beyond:alg:stochastic_inaxact_lmo} with LMO errors satisfying Assumption~\ref{beyond:assump:inexact_lmo_main}. The expected momentum error at iteration $T$ is bounded by
\begin{align*}
    E_{T-1} = \mathbb{E}[\|m^T - \nabla f(x^{T-1})\|_{\star}] &\le \left( \prod_{t=0}^{T-1} (1-\alpha_t) \right) \rho\sigma + \left( \sum_{t=0}^{T-1} \alpha_t^2 \prod_{\tau=t+1}^{T-1} (1-\alpha_\tau) \right)^{1/2} \rho\sigma \\
    &\quad + L \sum_{t=0}^{T-1} \gamma^t(1+\err_t) \prod_{\tau=t+1}^{T-1} (1-\alpha_\tau).
\end{align*}
\end{lemma}

\begin{proof}
The proof proceeds by unrolling the one-step recursion for the momentum error. Let $e_k \eqdef m^{k+1} - \nabla f(x^k)$, $v_k \eqdef g^k - \nabla f(x^k)$ be the stochastic noise, and $s_{k-1} \eqdef \nabla f(x^{k-1}) - \nabla f(x^k)$ be the gradient drift. The error recursion is
\begin{align*}
    e_k = (1-\alpha_k)e_{k-1} + \alpha_k v_k + (1-\alpha_k)s_{k-1}.
\end{align*}
Unrolling this relationship from $k=T-1$ down to $0$ gives the expression for the total error:
\begin{align*}
    e_{T-1} = \left(\prod_{t=0}^{T-1}(1-\alpha_t)\right) e_{-1} + \sum_{t=0}^{T-1} \left(\prod_{\tau=t+1}^{T-1}(1-\alpha_\tau)\right) \left( \alpha_t v_t + (1-\alpha_t)s_{t-1} \right),
\end{align*}
where we define $s_{-1}=0$ and $e_{-1} = m^0 - \nabla f(x^0)$. For the initialization $m^0=g^0$, we can bound $\mathbb{E}[\|e_{-1}\|_{\star}] \le \rho\sigma$. Taking the expectation of the norm and applying the triangle inequality yields
\begin{align*}
    \mathbb{E}[\|e_{T-1}\|_{\star}] &\le \left(\prod_{t=0}^{T-1} (1-\alpha_t)\right) \rho\sigma + \mathbb{E}\left[\left\| \sum_{t=0}^{T-1} \alpha_t v_t \prod_{\tau=t+1}^{T-1}(1-\alpha_\tau) \right\|_{\star}\right] \\
    &\quad + \mathbb{E}\left[\left\| \sum_{t=0}^{T-1} (1-\alpha_t) s_{t-1} \prod_{\tau=t+1}^{T-1}(1-\alpha_\tau) \right\|_{\star}\right].
\end{align*}
We now bound the two summation terms separately.

\textbf{1. Bounding the Accumulated Noise Term.}
We apply the norm compatibility condition and then Jensen's inequality, to have
\begin{align*}
    \mathbb{E}\left[\left\| \sum_{t=0}^{T-1} \alpha_t v_t \dots \right\|_{\star}\right] \le \rho \left(\mathbb{E}\left[\left\| \sum_{t=0}^{T-1} \alpha_t v_t \prod_{\tau=t+1}^{T-1}(1-\alpha_\tau) \right\|_{2}^2\right]\right)^{1/2}.
\end{align*}
The stochastic noise vectors $v_t$ are zero-mean and independent conditioned on the past. Therefore, the cross-terms in the squared sum vanish in expectation, yielding
\begin{align*}
    \mathbb{E}\left[\left\| \sum \dots \right\|_{2}^2\right] = \sum_{t=0}^{T-1} \alpha_t^2 \left(\prod_{\tau=t+1}^{T-1}(1-\alpha_\tau)\right)^2 \mathbb{E}[\|v_t\|_2^2] \le \sigma^2 \sum_{t=0}^{T-1} \alpha_t^2 \left(\prod_{\tau=t+1}^{T-1}(1-\alpha_\tau)\right)^2.
\end{align*}
Using the upper bound $(1-x)^2 \le (1-x)$ for any $x \in [0,1]$, we have
\begin{align*}
     \mathbb{E}\left[\left\| \sum \dots \right\|_{2}^2\right] \le \sigma^2 \sum_{t=0}^{T-1} \alpha_t^2 \prod_{\tau=t+1}^{T-1}(1-\alpha_\tau).
\end{align*}
Substituting this back yields the final bound for the noise term:
\begin{align*}
    \text{Noise Term} \le \rho\sigma \left(\sum_{t=0}^{T-1} \alpha_t^2 \prod_{\tau=t+1}^{T-1}(1-\alpha_\tau)\right)^{1/2}.
\end{align*}

\textbf{2. Bounding the Accumulated Drift Term.}
We apply the triangle inequality for sums, Assumption~\ref{beyond:assump:smoothness}, and the step-size bound $\|x^t - x^{t-1}\| \le \gamma^{t-1}(1+\err_{t-1})$, to obtain
\begin{align*}
    \mathbb{E}\left[\left\| \sum (1-\alpha_t) s_{t-1} \dots \right\|_{\star}\right] &\le \sum_{t=0}^{T-1} (1-\alpha_t) \mathbb{E}[\|s_{t-1}\|_{\star}] \prod_{\tau=t+1}^{T-1}(1-\alpha_\tau) \\
    &\le L \sum_{t=0}^{T-1} (1-\alpha_t) \gamma^{t-1}(1+\err_{t-1}) \prod_{\tau=t+1}^{T-1}(1-\alpha_\tau).
\end{align*}
Using the bound $(1-\alpha_t) \le 1$ and re-indexing the summation gives the simplified form
\begin{align*}
    \text{Drift Term} \le L \sum_{t=0}^{T-1} \gamma^t(1+\err_t) \prod_{\tau=t+1}^{T-1}(1-\alpha_\tau).
\end{align*}

\textbf{3. Combining the Bounds.}
Assembling the bounds for the three components gives the inequality stated in the lemma.
\end{proof}

Before analyzing the convergence for a specific time-varying schedule, we first establish a general convergence bound that holds for any choice of sequences $\{\gamma^k\}$ and $\{\alpha_k\}$. This theorem is the stochastic analogue of the deterministic result in Theorem~\ref{beyond:thm:general_deterministic} and serves as the starting point for all subsequent rate analysis.

\begin{lemma}[General Convergence Bound]
\label{beyond:thm:general_stochastic_bound}
Let Assumptions \ref{beyond:assump:inexact_lmo_main}, \ref{beyond:assump:smoothness}, \ref{beyond:assump:stochastic_oracle}, and the norm compatibility condition hold. The sequence of iterates $\{x^k\}$ generated by Algorithm~\ref{beyond:alg:stochastic_inaxact_lmo} with time-varying parameters satisfies the following inequality for any $K \ge 1$:
\begin{align*}
    \sum_{k=0}^{K-1} \gamma^k(1-\err_k)\mathbb{E}[\|\nabla f(x^{k+1})\|_{\star}]
    &\le \Delta^0 + \frac{3L}{2}\sum_{k=0}^{K-1}(\gamma^k)^2(1+\err_k)^2 \\
    &\quad + 2L\sum_{k=0}^{K-1}(\gamma^k)^2(1+\err_k) \\
    &\quad + 2\sum_{k=0}^{K-1} \gamma^k \mathbb{E}[\|\nabla f(x^k) - m^{k+1}\|_{\star}],
\end{align*}
where $\Delta^0 = f(x^0) - f^*$.
\end{lemma}

\begin{proof}
The proof begins with the per-iteration guarantee from the Inexact Descent Lemma (Lemma~\ref{beyond:lemma:inexact_descent_appendix}). To obtain a guarantee on the true gradient norm, we apply the reverse triangle inequality to the main descent term, $\|m^{k+1}\|_{\star} \ge \|\nabla f(x^{k+1})\|_{\star} - \|\nabla f(x^{k+1}) - m^{k+1}\|_{\star}$. Substituting this into the descent lemma's bound gives
\begin{align*}
    f(x^{k+1})
    &\le f(x^k) - \gamma^k\|\nabla f(x^{k+1})\|_{\star}(1-\err_k) \\
    &\quad + \gamma^k(1-\err_k)\|\nabla f(x^{k+1}) - m^{k+1}\|_{\star} \\
    &\quad + \gamma^k(1+\err_k)\|\nabla f(x^{k+1}) - m^{k+1}\|_{\star}
    + \frac{3L}{2}(\gamma^k)^2(1+\err_k)^2.
\end{align*}
Combining the two momentum error terms (whose coefficients sum to 2) and rearranging yields
\begin{align*}
    \gamma^k(1-\err_k)\|\nabla f(x^{k+1})\|_{\star}
    &\le f(x^k) - f(x^{k+1})
    + 2\gamma^k\|\nabla f(x^{k+1}) - m^{k+1}\|_{\star} \\
    &\quad + \frac{3L}{2}(\gamma^k)^2(1+\err_k)^2.
\end{align*}
We now sum this inequality from $k=0$ to $K-1$ and take the total expectation:
\begin{align*}
    \sum_{k=0}^{K-1}\gamma^k(1-\err_k)\mathbb{E}[\|\nabla f(x^{k+1})\|_{\star}]
    &\le \sum_{k=0}^{K-1} \mathbb{E}[f(x^k) - f(x^{k+1})] \\
    &\quad + 2\sum_{k=0}^{K-1}\gamma^k\mathbb{E}[\|\nabla f(x^{k+1}) - m^{k+1}\|_{\star}] \\
    &\quad + \sum_{k=0}^{K-1}\frac{3L}{2}(\gamma^k)^2(1+\err_k)^2.
\end{align*}
The first term on the right-hand side is a telescoping sum bounded by $\Delta^0$. The core of the proof is to decompose the sum of the momentum errors using the triangle inequality to bridge the index gap
\begin{align*}
    2\sum_{k=0}^{K-1}\gamma^k\mathbb{E}[\|\nabla f(x^{k+1}) - m^{k+1}\|_{\star}]
    &\le 2\sum_{k=0}^{K-1}\gamma^k
    \left(\mathbb{E}[\|\nabla f(x^{k+1}) - \nabla f(x^k)\|_{\star}]\right.\\
    &\qquad\qquad\left. + \mathbb{E}[\|\nabla f(x^k) - m^{k+1}\|_{\star}]\right).
\end{align*}
This decomposition separates the error into two distinct components.

\textbf{1. Bounding the Gradient Drift Component.} The first part of this sum is bounded using Assumption~\ref{beyond:assump:smoothness} ($L$-smoothness) and the property of the update step, $\|x^{k+1}-x^k\| = \gamma^k\|\hat{d}^k\| \le \gamma^k(1+\err_k)$, as
\begin{align*}
    2\sum_{k=0}^{K-1}\gamma^k\mathbb{E}[\|\nabla f(x^{k+1}) - \nabla f(x^k)\|_{\star}]
    &\le 2\sum_{k=0}^{K-1}\gamma^k \left( L \gamma^k (1+\err_k) \right) \\
    &= 2L\sum_{k=0}^{K-1}(\gamma^k)^2(1+\err_k).
\end{align*}
This term represents the accumulated error caused by the gradient changing between steps.

\textbf{2. The Momentum Lag Component.} The second part of the sum is the total accumulated momentum error, which is the final term in the theorem statement:
\begin{align*}
    2\sum_{k=0}^{K-1}\gamma^k \mathbb{E}[\|\nabla f(x^k) - m^{k+1}\|_{\star}].
\end{align*}
\textbf{Assembling the Final Bound.}
Combining all parts, we arrive at the final inequality stated in the theorem. This result cleanly separates the total progress (LHS) from the initial sub-optimality gap ($\Delta^0$), the cost from the step size's magnitude (the two $\sum(\gamma^k)^2$ terms), and the accumulated momentum lag error arising from momentum and stochastic noise.
\end{proof}

We now use the \say{General Convergence Bound} to derive an explicit convergence rate for a specific, parameter-agnostic schedule. The choice of the time-varying step size and momentum parameter is critical for ensuring that the accumulated error terms are properly controlled, leading to a guarantee of convergence. We adopt a schedule inspired by the analysis in \citet{yang2023two}, which has been proven effective for this type of recursive error structure.

\begin{theorem}[Convergence with Time-Varying Parameters]
\label{beyond:thm:time_varying_rate_final}
Let Assumptions \ref{beyond:assump:inexact_lmo_main}, \ref{beyond:assump:smoothness}, \ref{beyond:assump:stochastic_oracle}, and the norm compatibility condition hold, with a uniform inexactness bound $\err_k \le \err < 1$. If we choose the time-varying step size and momentum parameters for $k \ge 0$ as
\begin{align*}
    \gamma^k = \frac{\gamma}{(k+1)^{3/4}} \quad \text{and} \quad \alpha_k = \frac{\alpha}{(k+1)^{1/2}},
\end{align*}
for some user-chosen constants $\gamma > 0$ and $\alpha \in (0, 1]$, then after $K$ iterations, the minimum expected gradient norm is bounded as
\begin{align*}
    \min_{0 \le j < K} \mathbb{E}[\|\nabla f(x^{j+1})\|_{\star}] &= \mathcal{O}\left( \frac{\Delta^0 + L\gamma^2((1+\err)^2 + (1+\err))}{(1-\err)\gamma K^{1/4}} \right) \\
    &\quad + \mathcal{O}\left( \frac{\left( \rho\sigma\sqrt{\alpha} + L\gamma(1+\err)\alpha^{-1} \right) \log K}{(1-\err) K^{1/4}} \right).
\end{align*}
\end{theorem}

\begin{proof}
The proof starts from the General Convergence Bound (Theorem~\ref{beyond:thm:general_stochastic_bound}). After rearranging, it provides a bound on the minimum expected gradient norm:
\begin{align*}
    \left(\min_{0 \le j < K} \mathbb{E}[\|\nabla f(x^{j+1})\|_{\star}] \right)
    \sum_{k=0}^{K-1} \gamma^k(1-\err_k)
    &\le \Delta^0 + \frac{3L}{2}\sum_{k=0}^{K-1}(\gamma^k)^2(1+\err_k)^2 \\
    &\quad + 2L\sum_{k=0}^{K-1}(\gamma^k)^2(1+\err_k) \\
    &\quad + 2\sum_{k=0}^{K-1} \gamma^k \mathbb{E}[\|\nabla f(x^k) - m^{k+1}\|_{\star}].
\end{align*}
We analyze each term in this expression under the chosen parameter schedule and the uniform bound $\err_k \le \err$.

\textbf{1. Bounding the Sums of Step Sizes.}
With the schedule $\gamma^k = \gamma/(k+1)^{3/4}$, the sums have the following behavior. The progress-weighting sum on the LHS is bounded below by an integral for $K \ge 1$:
\begin{align*}
    \sum_{k=0}^{K-1} \gamma^k(1-\err_k) \ge \gamma(1-\err) \sum_{k=0}^{K-1} \frac{1}{(k+1)^{3/4}} \ge 4\gamma(1-\err)((K+1)^{1/4}-1).
\end{align*}
The squared step size sums on the RHS correspond to a convergent p-series (since the exponent $p=3/2 > 1$) and are bounded by the constant $\zeta(3/2) = \sum_{k=1}^{\infty} k^{-3/2}$ for any $K$, where $\zeta$ is the Riemann zeta function. Therefore, the two terms are bounded by $\frac{3L}{2}\gamma^2\zeta(3/2)(1+\err)^2$ and $2L\gamma^2\zeta(3/2)(1+\err)$ respectively.

\textbf{2. Bounding the Accumulated Momentum Error.}
The momentum lag component, $2\sum_{k=0}^{K-1}\gamma^k \mathbb{E}[\|\nabla f(x^k) - m^{k+1}\|_{\star}]$, is bounded by applying the results of Lemma~\ref{beyond:lemma:aligned_momentum_error}. The structure of the bound in that lemma is of the form addressed by Lemma 3 in \citet{yang2023two}. Applying their result to the two sum-of-products terms in our bound, with our chosen schedules, yields
\begin{align*}
    \sum_{k=0}^{K-1}\gamma^k \mathbb{E}[\|\nabla f(x^k) - m^{k+1}\|_{\star}]
    &\le \left( C_1 \rho\sigma\gamma\sqrt{\alpha}
    + C_2 L (1+\err) \frac{\gamma^2}{\alpha} \right) \log(K+1),
\end{align*}
where $C_1, C_2$ are universal numerical constants. This shows that the total accumulated error from momentum and noise grows only logarithmically with the number of iterations.

\textbf{3. Assembling the Final Bound.}
Substituting these explicit bounds back into the main inequality:
\begin{align*}
    \left(\min_{j<K}\mathbb{E}[\|\nabla f(x^{j+1})\|_{\star}]\right)
    \cdot 4\gamma(1-\err)(K^{1/4}-1)
    &\le \Delta^0 \\
    &\quad + L\gamma^2\zeta(3/2)\\
    &\qquad\cdot\left(\frac{3}{2}(1+\err)^2 + 2(1+\err)\right) \\
    &\quad + 2\left( C_1 \rho\sigma\gamma\sqrt{\alpha}\right.\\
    &\qquad\left. + C_2 L(1+\err)\frac{\gamma^2}{\alpha} \right) \log(K+1).
\end{align*}
Dividing by $4\gamma(1-\err)(K^{1/4}-1)$ gives the final expression for the convergence rate. For large $K$, the term $(K+1)^{1/4}-1 \approx K^{1/4}$. This leads to the final bound stated in the theorem, where the dependence on all parameters, including the user-chosen schedule constants $\gamma$ and $\alpha$, is made explicit.
\end{proof}

\section{Extensions}

\subsection{\texorpdfstring{$(L^0, L^1)$-smoothness}{(L\textasciicircum 0, L\textasciicircum 1)-smoothness}}\label{beyond:app:l0l1_smooth}

We now extend our analysis from the standard $L$-smooth setting to the more general and realistic $(L^0, L^1)$-smoothness model, as defined in \citet{riabinin2025gluon}.

\begin{assumption}[Local $(L^0, L^1)$-Smoothness]
\label{beyond:assump:l0l1_smoothness}
The function $f$ is continuously differentiable and is bounded below by $f^* > -\infty$. We assume that $f$ is locally $(L^0, L^1)$-smooth along the optimization trajectory. Specifically, for any pair of consecutive iterates $x^k$ and $x^{k+1}$ generated by the algorithm, the following inequality holds:
\begin{align*}
    f(x^{k+1}) \le f(x^k) + \inp{\nabla f(x^k)}{x^{k+1}-x^k} + \frac{L^0 + L^1\|\nabla f(x^k)\|_{\star}}{2}\|x^{k+1}-x^k\|^2.
\end{align*}
\end{assumption}
This local version is sufficient for the analysis of descent methods where the step sizes are bounded, as is the case here. For clarity, we do not explicitly track the radius of the local region, as it is implicitly defined by the maximum possible step size.

\begin{theorem}[Iteration Complexity for $(L^0, L^1)$-Smoothness]
Let the function $f$ satisfy the local $(L^0, L^1)$-smoothness property (Assumption~\ref{beyond:assump:l0l1_smoothness}). Consider the deterministic method with an inexact LMO satisfying a uniform error bound $\err_k \le \err < 1$ (Assumption~\ref{beyond:assump:inexact_lmo_main}), and using the adaptive step size
\begin{align*}
    \gamma^k_\star = \frac{\|\nabla f(x^k)\|_{\star}(1-\err_k)}{(L^0 + L^1\|\nabla f(x^k)\|_{\star})(1+\err_k)^2}.
\end{align*}
To guarantee finding an iterate $x^k$ such that $\|\nabla f(x^k)\|_{\star} \le \varepsilon$ after at most $K$ iterations, it is sufficient to run the algorithm for a number of iterations $K$ satisfying
\begin{align*}
    K \ge \frac{2\Delta^0(1+\err)^2}{(1-\err)^2} \left( \frac{L^0}{\varepsilon^2} + \frac{L^1}{\varepsilon} \right),
\end{align*}
where $\Delta^0 = f(x^0) - f^*$.
\end{theorem}

\begin{proof}
The proof derives the iteration complexity for the deterministic method under $(L^0, L^1)$-smoothness by using a step size that is chosen optimally at each iteration.

\textbf{1. Per-Iteration Descent Inequality.}
Our analysis begins with the generalized smoothness inequality for $(L^0, L^1)$-smooth functions, which holds for any step $x^{k+1} = x^k + \gamma^k \hat{d}^k$:
\begin{align*}
    f(x^{k+1}) \le f(x^k) + \inp{\nabla f(x^k)}{x^{k+1}-x^k} + \frac{L^0 + L^1\|\nabla f(x^k)\|_{\star}}{2}\|x^{k+1}-x^k\|^2.
\end{align*}
As established in the proof of Theorem~\ref{beyond:thm:general_deterministic}, we can bound the terms involving the inexact direction $\hat{d}^k$ as
\begin{itemize}
    \item $\|\gamma^k \hat{d}^k\|^2 \le (\gamma^k)^2 (1+\err_k)^2$.
    \item $\inp{\nabla f(x^k)}{\hat{d}^k} \le -\|\nabla f(x^k)\|_{\star}(1-\err_k)$.
\end{itemize}
Substituting these into the smoothness inequality gives the per-iteration guarantee
\begin{align} \label{beyond:eq:l0l1_per_iter_descent}
    f(x^{k+1}) \le f(x^k) - \gamma^k \|\nabla f(x^k)\|_{\star}(1-\err_k) + \frac{L^0 + L^1\|\nabla f(x^k)\|_{\star}}{2} (\gamma^k)^2 (1+\err_k)^2.
\end{align}

\textbf{2. Optimal Step Size and Resulting Descent.}
The right-hand side of inequality \eqref{beyond:eq:l0l1_per_iter_descent} is a quadratic function of the step size $\gamma^k$. To maximize the guaranteed descent at each step, we choose the $\gamma^k$ that minimizes this expression. The optimal choice is given by
\begin{align*}
    \gamma^k_\star = \frac{\|\nabla f(x^k)\|_{\star}(1-\err_k)}{(L^0 + L^1\|\nabla f(x^k)\|_{\star})(1+\err_k)^2}.
\end{align*}
We now substitute this optimal step size back into the descent inequality \eqref{beyond:eq:l0l1_per_iter_descent}:
\begin{align*}
    f(x^{k+1}) \le  f(x^k) - \frac{(\|\nabla f(x^k)\|_{\star}(1-\err_k))^2}{2(L^0 + L^1\|\nabla f(x^k)\|_{\star})(1+\err_k)^2}.
\end{align*}

\textbf{3. Global Bound and Iteration Complexity.}
Summing the final descent inequality from $k=0$ to $K-1$ gives the global progress:
\begin{align*}
    \sum_{k=0}^{K-1} \frac{(\|\nabla f(x^k)\|_{\star}(1-\err_k))^2}{2(L^0 + L^1\|\nabla f(x^k)\|_{\star})(1+\err_k)^2} \le \sum_{k=0}^{K-1} (f(x^k) - f(x^{k+1})) \le \Delta^0.
\end{align*}
To derive a meaningful convergence bound from this sum, we define a progress function $\phi(t)$ that captures the contribution for a gradient norm of magnitude $t = \|\nabla f(x^k)\|_{\star}$:
\begin{align*}
    \phi(t) \eqdef \frac{t^2}{L^0 + L^1 t}.
\end{align*}
This function is monotonically increasing for $t \ge 0$. Assuming a uniform inexactness bound $\err_k \le \err < 1$, the summed inequality can be rewritten as:
\begin{align*}
    \frac{(1-\err)^2}{2(1+\err)^2} \sum_{k=0}^{K-1} \phi(\|\nabla f(x^k)\|_{\star}) \le \Delta^0.
\end{align*}
Let $g_{\min} = \min_{0 \le j < K} \|\nabla f(x^j)\|_{\star}$. Since $\phi(t)$ is increasing, we can lower-bound the sum by $K \phi(g_{\min})$:
\begin{align*}
    \frac{K (1-\err)^2}{2(1+\err)^2} \phi(g_{\min}) \le \Delta^0 \implies \phi(g_{\min}) \le \frac{2\Delta^0(1+\err)^2}{K(1-\err)^2}.
\end{align*}
To derive the iteration complexity, we require that the algorithm finds an iterate with gradient norm at most $\varepsilon$, i.e., $g_{\min} \le \varepsilon$. This is guaranteed if $\phi(\varepsilon)$ satisfies the above bound:
\begin{align*}
    \phi(\varepsilon) \le \frac{2\Delta^0(1+\err)^2}{K(1-\err)^2}.
\end{align*}
Solving for $K$ gives the required number of iterations:
\begin{align*}
    K \ge \frac{2\Delta^0(1+\err)^2}{\phi(\varepsilon)(1-\err)^2}.
\end{align*}
Substituting the definition of $\phi(\varepsilon) = \frac{\varepsilon^2}{L^0 + L^1\varepsilon}$ yields the final complexity stated in the theorem.
\end{proof}

\subsection{Layer-wise analysis}  \label{beyond:app:layerwise}

We now extend the analysis to the practical, multi-layered setting considered in \citet{riabinin2025gluon}. This framework is essential for modern deep learning, where models are composed of distinct blocks or layers with heterogeneous geometries. We assume that the parameter vector $X$ is a concatenation of $p$ blocks, $X = [X_1, \dots, X_p]$, where each layer $X_i$ belongs to its own vector space $\mathcal{X}_i$ equipped with an inner product $\langle \cdot, \cdot\rangle_{(i)}$ and norm $\|\cdot\|_{(i)}$. Each layer is updated independently via its own LMO.

This framework allows us to model a realistic scenario where not only the smoothness but also the LMO inexactness varies across layers. For example, the LMO for a simple bias vector might be solved almost exactly ($\err_i$ is small), while the LMO for a large transformer attention block might have a larger error ($\err_i$ is large).

\begin{assumption}[Layer-Wise Smoothness]
\label{beyond:assump:layerwise_smoothness}
The function $f$ is smooth with respect to each layer $i$ with a constant $L_i$. Specifically, for any iterates $X^k$ and $X^{k+1}$, the following holds:
\begin{align*}
    f(X^{k+1}) \le f(X^k) + \sum_{i=1}^p \left[ \inp{\nabla_i f(X^k)}{X_i^{k+1}-X_i^k}_{(i)} + \frac{L_i}{2}\|X_i^{k+1}-X_i^k\|^2_{(i)} \right].
\end{align*}
\end{assumption}

We first derive a general bound that holds for any choice of layer-specific, time-varying step sizes $\gamma_{k,i}$ and inexactness levels $\err_{k,i}$.

\begin{theorem}[General Convergence Bound for Layer-Wise Method]
Let Assumption~\ref{beyond:assump:layerwise_smoothness} hold. Let the layer-wise inexact LMO for each layer $i$ satisfy Assumption~\ref{beyond:assump:inexact_lmo_main} with error $\err_{k,i} < 1$. Then, for any sequence of layer-specific step sizes $\gamma_{k,i} > 0$, the following bound holds after $K$ iterations:
\begin{align*}
    \sum_{k=0}^{K-1} \sum_{i=1}^p \gamma_{k,i} \|\nabla_i f(X^k)\|_{(i)\star}(1-\err_{k,i}) \le \Delta^0 + \frac{1}{2} \sum_{k=0}^{K-1} \sum_{i=1}^p L_i (\gamma_{k,i})^2 (1+\err_{k,i})^2.
\end{align*}
\end{theorem}

\begin{proof}
The proof extends the single-layer analysis to the multi-layer setting. We start with the layer-wise smoothness inequality from Assumption~\ref{beyond:assump:layerwise_smoothness}. The update for each layer is $X_i^{k+1}-X_i^k = \gamma_{k,i} \hat{d}_i^k$. For each layer $i$ inside the summation, we can bound the terms involving its update direction $\hat{d}_i^k$ exactly as in the single-layer case:
\begin{itemize}
    \item $\|\gamma_{k,i} \hat{d}_i^k\|^2_{(i)} \le (\gamma_{k,i})^2 (1+\err_{k,i})^2$.
    \item $\inp{\nabla_i f(X^k)}{\hat{d}_i^k}_{(i)} \le -\|\nabla_i f(X^k)\|_{(i)\star}(1-\err_{k,i})$.
\end{itemize}
Substituting these bounds into the main inequality gives the per-iteration descent guarantee for the layer-wise setting:
\begin{align*}
    f(X^{k+1}) \le f(X^k) + \sum_{i=1}^p \left[ -\gamma_{k,i} \|\nabla_i f(X^k)\|_{(i)\star}(1-\err_{k,i}) + \frac{L_i}{2} (\gamma_{k,i})^2 (1+\err_{k,i})^2 \right].
\end{align*}
Rearranging to isolate the measure of progress (the sum of gradient norms) yields
\begin{align*}
    \sum_{i=1}^p \gamma_{k,i} \|\nabla_i f(X^k)\|_{(i)\star}(1-\err_{k,i}) \le f(X^k) - f(X^{k+1}) + \sum_{i=1}^p \frac{L_i}{2} (\gamma_{k,i})^2 (1+\err_{k,i})^2.
\end{align*}
Summing this from $k=0$ to $K-1$ and using the telescoping property of the function values on the right-hand side, $\sum_{k=0}^{K-1}(f(X^k) - f(X^{k+1})) = f(X^0) - f(X^K) \le \Delta^0$, gives the final result stated in the theorem.
\end{proof}

We now analyze the convergence of the layer-wise method when using an adaptive step size for each layer, chosen at each iteration to maximize the guaranteed descent. This approach provides a theoretically-grounded, adaptive learning rate policy for each layer.

\begin{theorem}[Iteration Complexity with Layer-Wise Adaptive Step Sizes]
Let Assumption~\ref{beyond:assump:layerwise_smoothness} hold. Consider the deterministic layer-wise method where the inexact LMO for each layer $i$ satisfies a uniform error bound $\err_{k,i} \le \err_i < 1$. If at each iteration $k$, the step size for each layer $i$ is chosen adaptively as
\begin{align*}
    \gamma_{k,i}^* = \frac{\|\nabla_i f(X^k)\|_{(i)\star}(1-\err_{k,i})}{L_i(1+\err_{k,i})^2},
\end{align*}
then to guarantee finding an iterate $X^k$ such that the weighted squared gradient norm $\sum_{i=1}^p \frac{\|\nabla_i f(X^k)\|_{(i)\star}^2}{L_i} \le \varepsilon^2$ after at most $K$ iterations, it is sufficient to run the algorithm for a number of iterations $K$ satisfying
\begin{align*}
    K \ge \frac{2\Delta^0}{\varepsilon^2} \max_{1 \le j \le p} \left\{ \frac{(1+\err_j)^2}{(1-\err_j)^2} \right\}.
\end{align*}
\end{theorem}

\begin{proof}
The proof begins with the per-iteration descent inequality derived in the proof of the General Convergence Bound theorem:
\begin{align*}
    f(X^{k+1}) \le f(X^k) + \sum_{i=1}^p \left[ -\gamma_{k,i} \|\nabla_i f(X^k)\|_{(i)\star}(1-\err_{k,i}) + \frac{L_i}{2} (\gamma_{k,i})^2 (1+\err_{k,i})^2 \right].
\end{align*}
The adaptive step size $\gamma_{k,i}^*$ is the value that minimizes the term in the square brackets for each layer $i$. Substituting this optimal choice into the inequality yields the maximum guaranteed descent at each step
\begin{align*}
    f(X^{k+1}) \le f(X^k) - \sum_{i=1}^p \frac{(\|\nabla_i f(X^k)\|_{(i)\star}(1-\err_{k,i}))^2}{2 L_i (1+\err_{k,i})^2}.
\end{align*}
Summing this from $k=0$ to $K-1$ gives the global progress bound
\begin{align*}
    \sum_{k=0}^{K-1} \sum_{i=1}^p \frac{\|\nabla_i f(X^k)\|_{(i)\star}^2 (1-\err_{k,i})^2}{2 L_i (1+\err_{k,i})^2} \le \Delta^0.
\end{align*}
We assume a uniform (but potentially layer-specific) inexactness bound $\err_{k,i} \le \err_i < 1$. The inequality becomes
\begin{align*}
    \frac{1}{2} \sum_{k=0}^{K-1} \sum_{i=1}^p \frac{(1-\err_i)^2}{(1+\err_i)^2} \frac{\|\nabla_i f(X^k)\|_{(i)\star}^2}{L_i} \le \Delta^0.
\end{align*}
Let $w_i \eqdef \frac{(1-\err_i)^2}{(1+\err_i)^2}$ be the layer-specific error weight and let $G_k^2 \eqdef \sum_{i=1}^p \frac{\|\nabla_i f(X^k)\|_{(i)\star}^2}{L_i}$ be the weighted squared gradient norm. The bound can be written as
\[
\frac{1}{2} \sum_{k=0}^{K-1} \sum_{i=1}^p w_i \frac{\|\nabla_i f(X^k)\|_{(i)\star}^2}{L_i}.
\]
To derive a clear complexity, we can lower-bound the weights by their minimum value, $w_{\min} = \min_{1 \le j \le p} w_j$:
\begin{align*}
    \sum_{i=1}^p w_i \frac{\|\nabla_i f(X^k)\|_{(i)\star}^2}{L_i} \ge w_{\min} \sum_{i=1}^p \frac{\|\nabla_i f(X^k)\|_{(i)\star}^2}{L_i} = w_{\min} G_k^2.
\end{align*}
Substituting this into the main inequality gives
    $\frac{w_{\min}}{2} \sum_{k=0}^{K-1} G_k^2 \le \Delta^0.$
Let $G_{\min}^2 = \min_{0 \le j < K} G_j^2$. Since $\sum_{k=0}^{K-1} G_k^2 \ge K \cdot G_{\min}^2$, we have
\begin{align*}
    \frac{K w_{\min}}{2} G_{\min}^2 \le \Delta^0 \implies G_{\min}^2 \le \frac{2\Delta^0}{K w_{\min}}.
\end{align*}
To guarantee that the weighted squared gradient norm $G_{\min}^2 \le \varepsilon^2$, we require
\begin{align*}
    \frac{2\Delta^0}{K w_{\min}} \le
    \varepsilon^2 \implies K \ge \frac{2\Delta^0}{\varepsilon^2 w_{\min}}.
\end{align*}
Substituting back the definition of $w_{\min} = \min_{j} \frac{(1-\err_j)^2}{(1+\err_j)^2}$ gives the final complexity
\[
    K \ge \frac{2\Delta^0}{\varepsilon^2} \frac{1}{\min_{j} \frac{(1-\err_j)^2}{(1+\err_j)^2}} = \frac{2\Delta^0}{\varepsilon^2} \max_{j} \frac{(1+\err_j)^2}{(1-\err_j)^2}. \qedhere
\]
\end{proof}

\paragraph{Discussion}
This result for the layer-wise adaptive method is powerful as it provides a concrete, theoretically-grounded learning rate policy for each layer that leads to an optimal convergence rate.
\begin{itemize}
    \item \textbf{Optimal Rate:} The resulting complexity is $\mathcal{O}(1/\varepsilon^2)$, which is the optimal rate for deterministic non-convex optimization. This confirms that the layer-wise adaptive strategy is efficient.
    \item \textbf{Layer-Wise Adaptivity:} The step size $\gamma_{k,i}^*$ for each layer adapts to its own local properties: its current gradient norm $\|\nabla_i f(X^k)\|_{(i)\star}$, its smoothness $L_i$, and its LMO precision $\err_{k,i}$. This is a practical and intuitive update rule that naturally assigns smaller steps to sharper or less precisely updated layers.
    \item \textbf{Bottleneck Effect of Inexactness:} The final complexity is degraded by a ``worst-layer" factor, $\max_{j} \frac{(1+\err_j)^2}{(1-\err_j)^2}$. This shows that the global convergence speed is bottlenecked by the single layer with the worst combination of update quality and infeasibility cost. A single layer with very high inexactness ($\err_j \to 1$) will cause this factor to explode, dominating the iteration complexity regardless of how precisely the other layers are handled.
\end{itemize}

\section{Additional experiments and details}

This section provides supplementary details and results to support the empirical findings presented in Section~\ref{beyond:sec:experiments} of the main paper.

\subsection{Experimental setup: details}
\label{beyond:app:exp_details}

We provide additional details regarding the experimental setups for both the NanoGPT and CIFAR-10 benchmarks. Further details on model architectures and data processing can be found in the respective code repositories cited in the main text. All experiments were conducted on a system equipped with 4 $\times$ NVIDIA A100 GPUs.

\paragraph{NanoGPT on FineWeb.}
The hyperparameter grids used for the sweeps shown in Figure~\ref{beyond:fig:nanoGPT_heatmaps} are visible on the axes of the plots themselves. The LMO inexactness $\err$ is controlled by the number of iterations of the \approxalg{PolarExpress} algorithm; fewer iterations correspond to a higher error $\err$.

\paragraph{CNN on CIFAR-10.}
For the results presented in Figure~\ref{beyond:fig:cifar_variability}, we performed a comprehensive grid search over the following hyperparameters for each level of LMO precision (i.e., for each number of \approxalg{Newton-Schulz} iterations):
\begin{itemize}
    \item \textbf{Step Size ($\gamma$):} A logarithmically spaced grid from $2^{-12}$ to $2^{-5}$.
    \item \textbf{Momentum ($\alpha$):} A linearly spaced grid from $0.05$ to $0.90$ in increments of $0.05$.
\end{itemize}
The ``Best Accuracy" (solid lines) in Figure~\ref{beyond:fig:cifar_variability} corresponds to the maximum test accuracy achieved at a given step size by tuning over the momentum values. The ``Worst Accuracy" (dashed lines) corresponds to the minimum test accuracy from the same sweep.

\subsection{Additional results for NanoGPT}
\label{beyond:app:additional_nanogpt}

\begin{figure}[ht!]
    \centering
    \subfigure[Improvements saturate after $\sim$5 \approxalg{PolarExp} iterations.\label{beyond:fig:appendix_nanogpt_convergence_full}]{%
    \begin{minipage}[t]{0.49\linewidth}
        \centering
        \includegraphics[width=\linewidth]{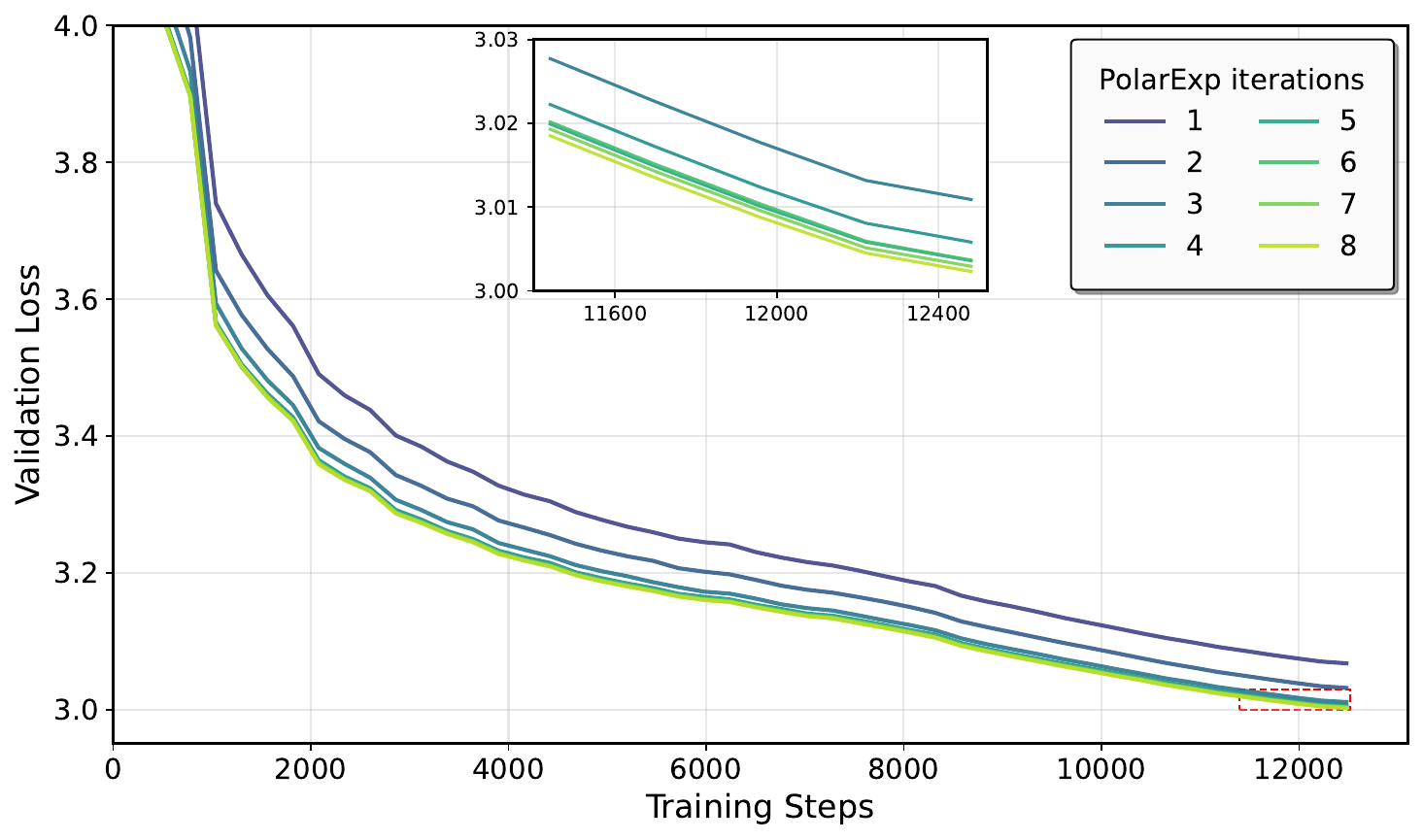}
    \end{minipage}}%
    \hfill
    \subfigure[Training time grows slightly with the iteration count.\label{beyond:fig:appendix_nanogpt_time}]{%
    \begin{minipage}[t]{0.49\linewidth}
        \centering
        \includegraphics[width=\linewidth]{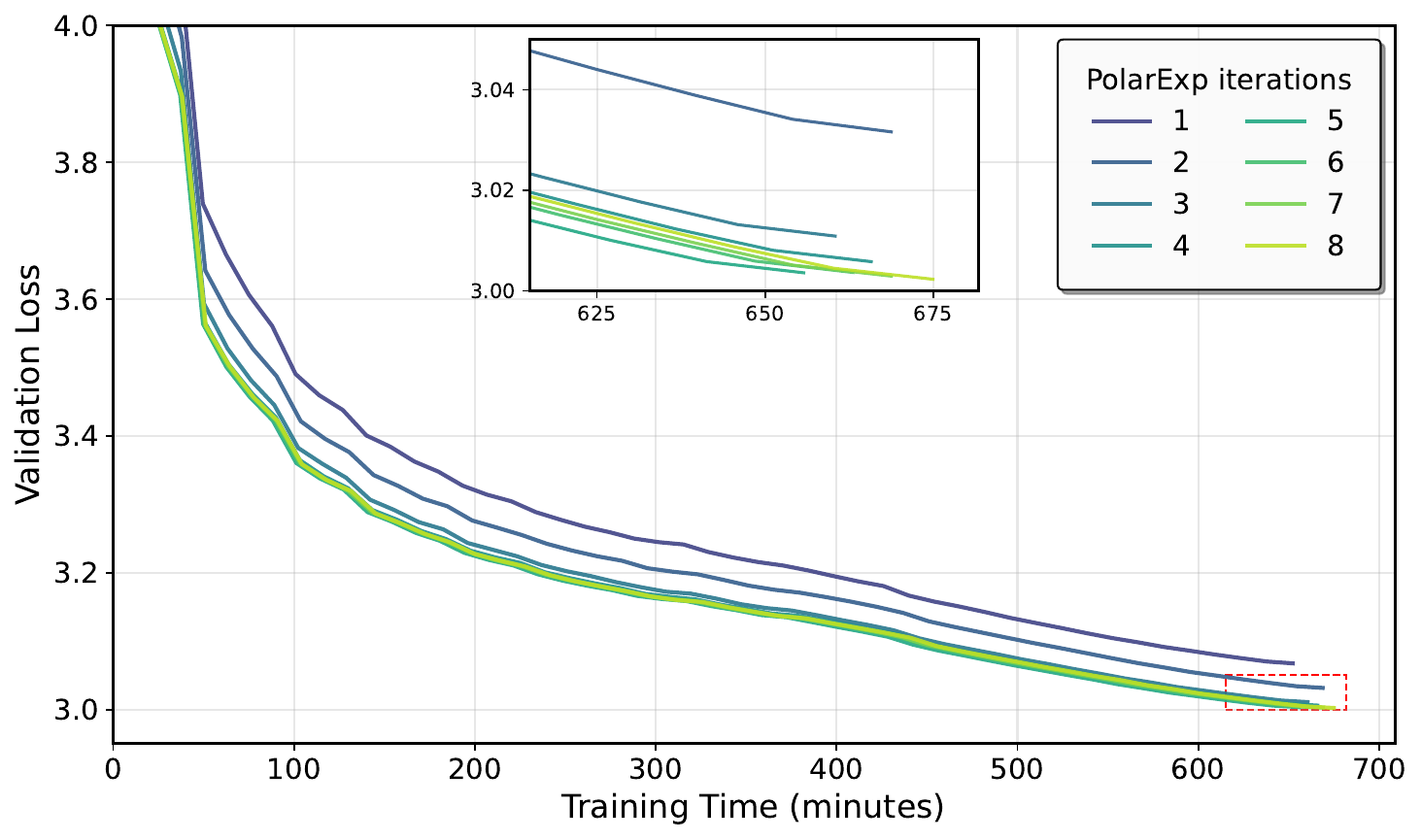}
    \end{minipage}}%
    \caption{Supplementary results for NanoGPT on FineWeb: convergence behavior (left) and runtime scaling (right) as the number of \approxalg{PolarExpress} (PolarExp) iterations varies.}
    \label{beyond:fig:appendix_nanogpt_combined}
\end{figure}

\paragraph{Performance Degradation with Increasing Precision.}
As mentioned in the main text, increasing the LMO precision yields diminishing returns. Figure~\ref{beyond:fig:appendix_nanogpt_convergence_full} provides the full validation loss curves for up to 8 iterations of the \approxalg{PolarExpress} algorithm. The plot clearly shows that while the most significant performance gains are achieved when moving from 1 to 3 iterations, further increases in precision (from 3 to 5, and 5 to 8) continue to improve the convergence speed and final validation loss, albeit with smaller margins. This observation confirms the trend of diminishing returns and empirically validates the degradation factor present in our theoretical bounds.

Table~\ref{beyond:tab:nanogpt_val_loss} complements these curves by reporting the final validation loss for each setting, highlighting the diminishing returns observed beyond five \approxalg{PolarExpress} iterations.

\begin{table}[ht]
\centering
    \caption{Final validation loss and total training time for NanoGPT on FineWeb across different numbers of \approxalg{PolarExpress} iterations used for the LMO approximation.}
    \label{beyond:tab:nanogpt_val_loss}
    \resizebox{\textwidth}{!}{%
    \begin{tabular}{@{}lcccccccc@{}}
        \toprule
        \approxalg{PolarExpress} iterations & 1 & 2 & 3 & 4 & 5 & 6 & 7 & 8 \\
        \midrule
        Validation loss & 3.0675 & 3.0316 & 3.0109 & 3.0058 & 3.0036 & 3.0037 & 3.0029 & 3.0023 \\
        Training time (minutes) & 652.2 & 668.8 & 660.4 & 665.8 & 655.7 & 663.1 & 668.8 & 675.0 \\
        \bottomrule
	\end{tabular}%
    }
\end{table}

\subsubsection{LMO error analysis}
\paragraph{Aggregate orthogonalization diagnostic.}
Figure~\ref{beyond:fig:lmo_error} reports an aggregate orthogonalization diagnostic over training, obtained by summing $\|\mathrm{PE}_p(M)-\mathrm{PE}_{100}(M)\|_2$ ($\mathrm{PE}_p$ stands for \approxalg{PolarExpress} run for $p$ iterations) over the tracked layers at each training step. The figure shows that higher inner-iteration budgets consistently reduce the aggregate discrepancy and that, for a fixed budget, this quantity remains relatively stable over training apart from occasional transient spikes. This aggregate view is useful for showing that the overall orthogonalization difficulty does not drift dramatically during training.
These results provide an explanation for the saturation of performance beyond certain LMO quality, as the theoretical benefits of higher precision diminish once the LMO error reaches sufficiently low levels.
However, it hides substantial heterogeneity across tensor families; Figure~\ref{beyond:fig:pe_layerwise_6000} provides the corresponding block-wise and family-wise breakdown at a representative checkpoint.

\begin{figure}[ht!]
    \centering
    \includegraphics[width=0.6\linewidth]{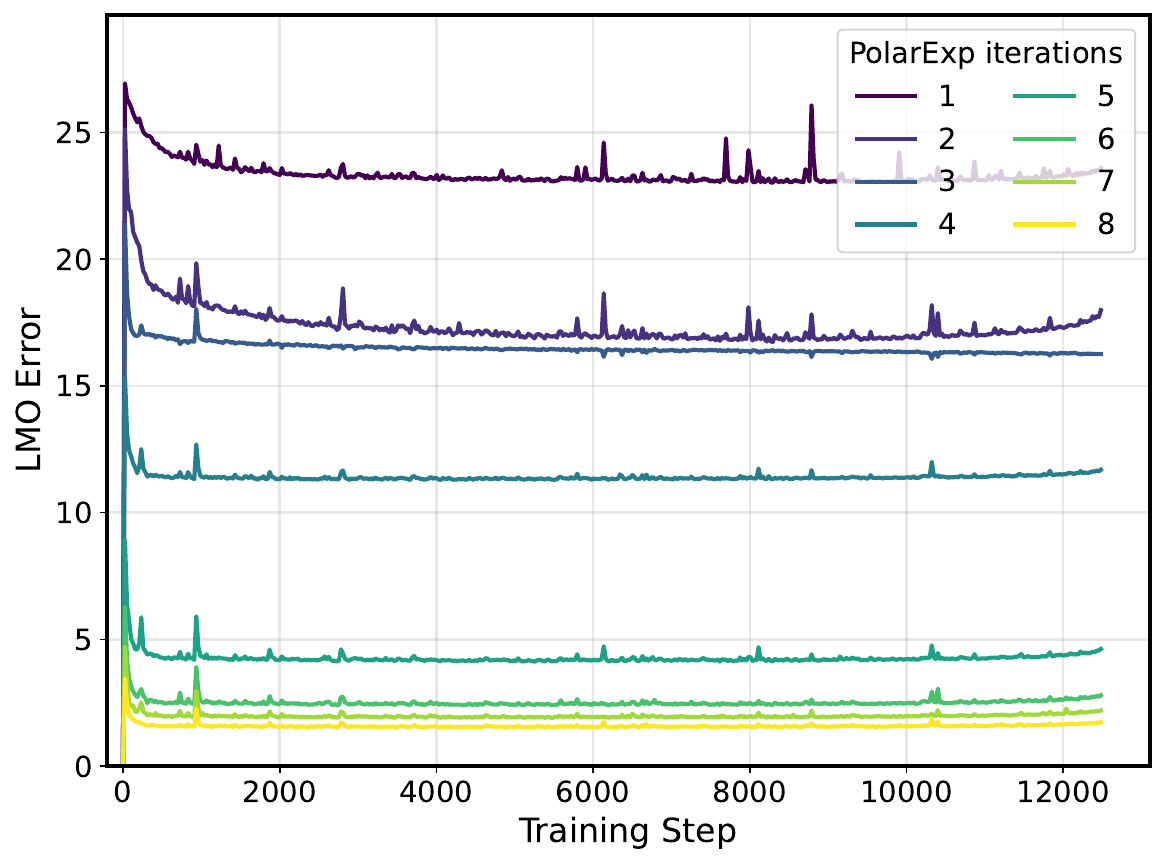}
    \caption{Aggregate \approxalg{PolarExpress} error over training, defined as the sum over tracked layers of
$\|\mathrm{PE}_p(M)-\mathrm{PE}_{100}(M)\|_2$
on the same normalized matrix $M$. Higher iteration counts maintain lower aggregate orthogonalization error throughout training.}%
    \label{beyond:fig:lmo_error}
\end{figure}

\begin{figure*}[t]
    \centering
    \includegraphics[width=\linewidth]{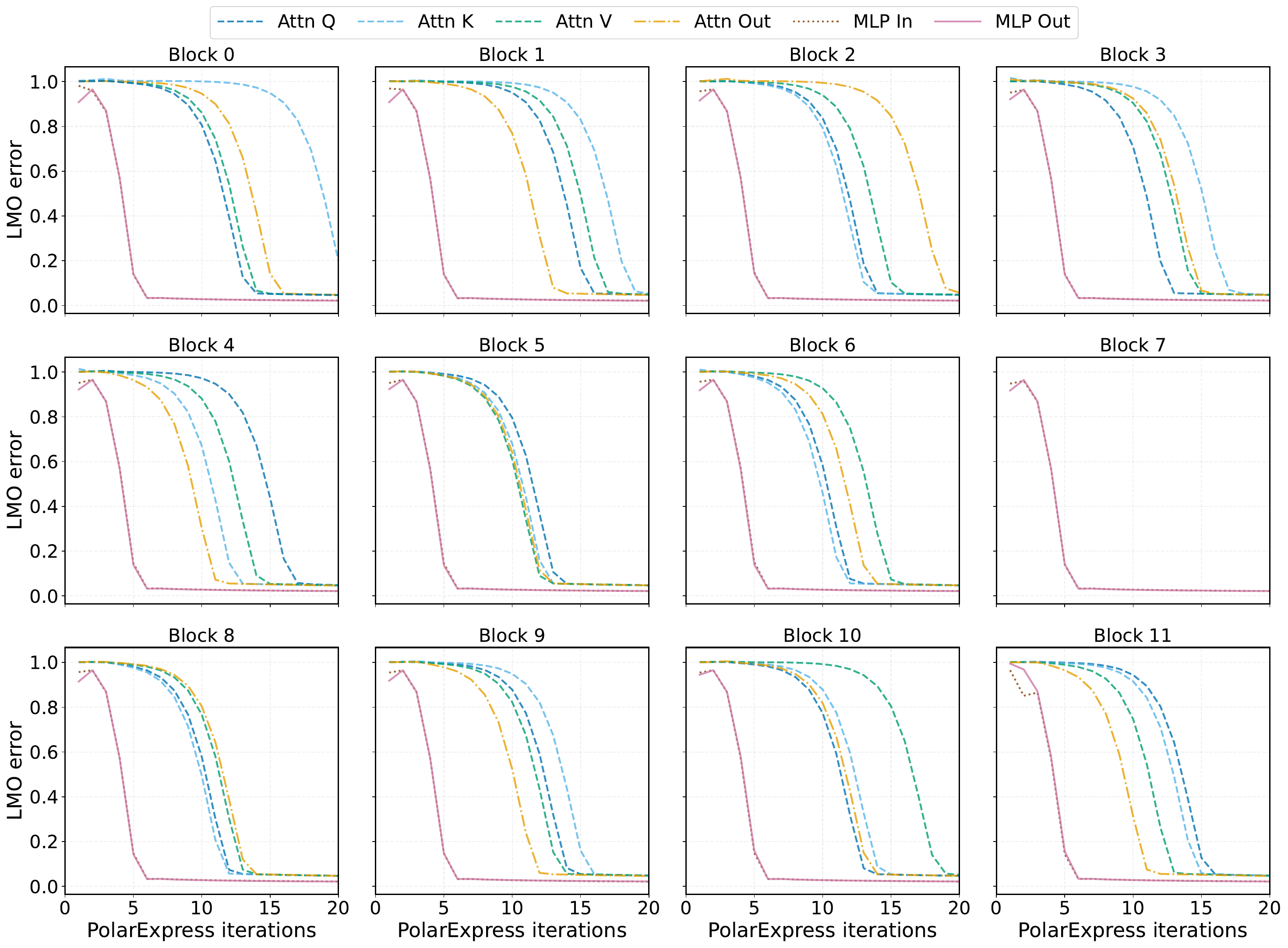}
    \caption{Layer-wise \approxalg{PolarExpress} convergence at validation step 6000 for NanoGPT. For each tracked matrix family in each transformer block, we plot the spectral-norm discrepancy
    $\|\mathrm{PE}_k(M)-\mathrm{PE}_{100}(M)\|_2$
    as a function of the inner-iteration count $k$, where $M$ is the normalized pre-orthogonalization momentum matrix and $\mathrm{PE}_{100}$ is used as a high-accuracy reference. MLP matrices converge in only a few iterations, while several attention matrices remain difficult until roughly $k\approx 10$--$20$, revealing strong heterogeneity across parameter families and blocks.}
    \label{beyond:fig:pe_layerwise_6000}
\end{figure*}

\paragraph{Layer-wise orthogonalization difficulty.}
Figure~\ref{beyond:fig:pe_layerwise_6000} complements the aggregate diagnostic in Figure~\ref{beyond:fig:lmo_error} by resolving the orthogonalization error at a representative intermediate training step (number 6000) into transformer block and parameter family. Two patterns are immediate. First, the dominant source of variation is the tensor family rather than the checkpoint: MLP In/Out matrices drop to a low-error band after roughly $5$--$6$ \approxalg{PolarExpress} iterations in every block, whereas attention matrices remain close to $1$ for the first few iterations and typically undergo their main transition only around $k\approx 10$--$20$. Second, this heterogeneity largely disappears once the inner budget is sufficiently large: by $k\approx 20$, most blocks and families have collapsed to a narrow low-error band. This explains why increasing the \approxalg{PolarExpress} budget beyond five iterations can still improve NanoGPT performance in Figure~\ref{beyond:fig:appendix_nanogpt_convergence_full}: five iterations are already sufficient for the MLP matrices, but several attention matrices are still far from their PE$_{100}$ reference at that budget. The figure also lends empirical support to the layer-wise discussion in Appendix~\ref{beyond:app:layerwise}, where the effect of LMO inexactness is predicted to be non-uniform across blocks.

Block~7 appears as an easy outlier because only MLP-type matrices are present in this diagnostic for that block. More generally, the figure shows that a single global summary statistic can hide a strongly bimodal low-budget regime: at small inner-iteration counts, the optimizer already orthogonalizes MLP matrices well, while many attention matrices remain substantially harder.

\paragraph{Why can the plotted error be slightly above 1?}
Small excursions slightly above $1$ at the earliest inner iterations are not contradictory. The plotted quantity is not the error to the exact polar factor, but the difference between two finite-step polynomial iterates,
$\|\mathrm{PE}_k(M)-\mathrm{PE}_{100}(M)\|_2$.
Moreover, our theory does not assume the approximate direction to remain feasible, so low-budget iterates can transiently have operator norm slightly above $1$. In practice, this effect is further moderated by the implementation: before applying \approxalg{PolarExpress}, we normalize the matrix by its Frobenius norm with a small $\varepsilon$ and a $1.01$ safety margin, and the degree-5 coefficient schedule is safety-scaled accordingly to improve numerical stability in bfloat16.

\paragraph{Measurement protocol for orthogonalization diagnostics.}
For the orthogonalization diagnostics in Figure~\ref{beyond:fig:lmo_error}, Figure~\ref{beyond:fig:pe_layerwise_6000}, and Table~\ref{beyond:tab:pe_iteration_discrepancy_summary}, we record the normalized pre-orthogonalization momentum matrix $M$ during validation and compute
\[
\|\mathrm{PE}_k(M)-\mathrm{PE}_{100}(M)\|_2
\]
on the same matrix, where $\mathrm{PE}_{100}$ serves as a high-accuracy reference. Thus, these measurements should be interpreted as a solver-to-solver calibration of the practical inner routine, rather than as the error to the exact polar factor. In \approxalg{PolarExpress} implementation the input is first scaled as
\[
M \leftarrow \frac{M}{1.01(\|M\|_F + 10^{-7})},
\]
which keeps the spectrum inside the convergence region of the polynomial iteration and improves stability in bfloat16.

\begin{table}[t]
\centering
\caption{Effect of the number of \approxalg{PolarExpress} iterations on the aggregate orthogonalization discrepancy
$e_k(M)=\|\mathrm{PE}_k(M)-\mathrm{PE}_{100}(M)\|_2$,
and the final NanoGPT validation loss on FineWeb. Discrepancy statistics are aggregated over all completed checkpoints and saved tensors. This is a solver-to-solver diagnostic relative to a high-accuracy PE$_{100}$ reference on the same normalized matrix, not the error to the exact polar factor.}
\label{beyond:tab:pe_iteration_discrepancy_summary}
\begin{tabular}{rccccc}
\toprule
PE iters & Mean $e_k$ & Std. $e_k$ & Median $e_k$ & Max $e_k$ & Val. loss \\
\midrule
1 & 0.9826 & 0.0311 & 1.0011 & 1.0200 & 3.0675 \\
2 & 0.9879 & 0.0220 & 1.0016 & 1.0079 & 3.0316 \\
3 & 0.9554 & 0.0644 & 1.0012 & 1.0143 & 3.0109 \\
4 & 0.8480 & 0.2260 & 0.9964 & 1.0055 & 3.0058 \\
5 & 0.6952 & 0.4047 & 0.9895 & 1.0032 & 3.0036 \\
6 & 0.6507 & 0.4321 & 0.9797 & 1.0030 & 3.0037 \\
7 & 0.6424 & 0.4387 & 0.9616 & 1.0031 & 3.0029 \\
8 & 0.6265 & 0.4469 & 0.9272 & 1.0028 & 3.0023 \\
\bottomrule
\end{tabular}
\end{table}

Table~\ref{beyond:tab:pe_iteration_discrepancy_summary} shows that increasing the inner PolarExpress budget consistently reduces the aggregate solver-to-solver discrepancy and improves the final validation loss, but also reveals a strongly heterogeneous low-budget regime: at small $k$, many attention matrices remain close to their PE$_{100}$ reference only after substantially more iterations than the MLP matrices, as seen in Figure~\ref{beyond:fig:pe_layerwise_6000}.

\subsection{Additional results for CIFAR-10}

\paragraph{Full Hyperparameter Grids.}
To complement the summary plots in the main text, Figure~\ref{beyond:fig:appendix_cifar_heatmaps} presents the full heatmap of final test accuracies across the entire hyperparameter grid for different levels of LMO precision. These plots provide a more detailed view of the optimizer's stability. For a highly precise LMO (e.g., 8 \approxalg{Newton-Schulz} iterations), a large contiguous region of high performance is visible, indicating robustness to hyperparameter choices. In contrast, for a highly inexact LMO (e.g., 1 \approxalg{Newton-Schulz} iteration), the high-performance region is smaller and more fragmented, highlighting the optimizer's sensitivity to tuning when the LMO is imprecise.

\begin{figure*}[ht!]
\centering
\includegraphics[width=\linewidth]{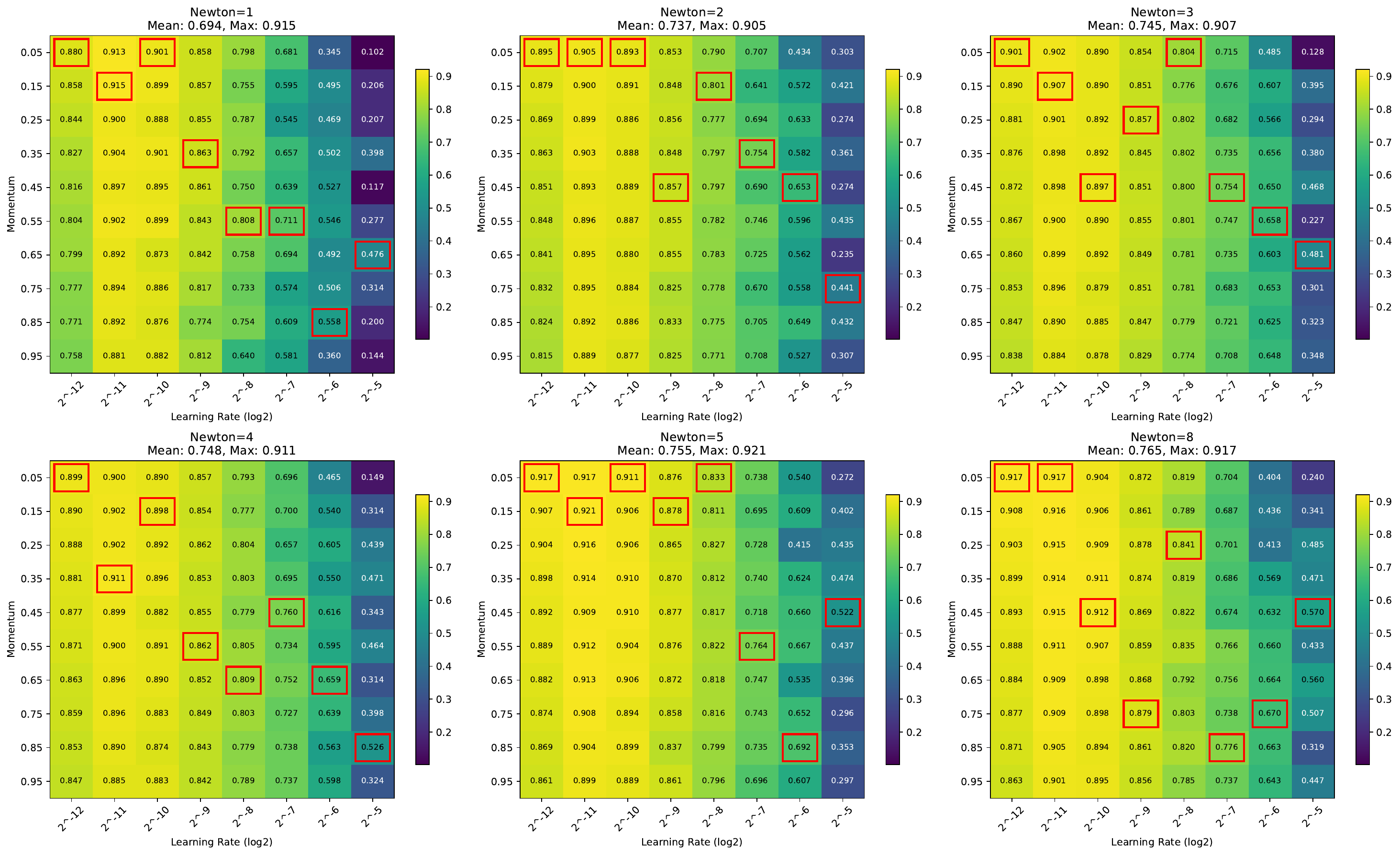}
    \caption{Full final test accuracy heatmaps on CIFAR-10 across a grid of step sizes and momentum values for different numbers of \approxalg{Newton-Schulz} (``Newton'') iterations. Brighter colors indicate higher final test accuracy. The region of high performance is visibly larger and more stable for higher numbers of \approxalg{Newton-Schulz} iterations.}
    \label{beyond:fig:appendix_cifar_heatmaps}
\end{figure*}
\endgroup

\end{document}